# Robots for Kiwifruit Harvesting and Pollination

Jamie Bell

A thesis submitted in fulfilment of the requirements for the degree of Doctor of Philosophy in Computer Systems Engineering

The University of Auckland

2019

# Abstract

This research was a part of a project that developed mobile robots that performed targeted pollen spraying and automated harvesting in pergola structured kiwifruit orchards. The project was called the "MBIE Multipurpose Orchard Robotics Project" and was a collaboration between the University of Auckland, the University of Waikato, Robotics Plus Ltd and Plant & Food Research.

Many of the contributions presented here specifically relate to automated harvesting, pollination and navigation in kiwifruit orchards. The contributions relating to harvesting include multiple fruit detection methods, testing multiple sensors for harvesting, a study of how to perform kiwifruit detachment, a calibration method for a kiwifruit harvesting robot and a path planning method for picking fruit that could not be reached by previous kiwifruit harvesting robots. In addition, multiple kiwifruit detachment mechanisms were designed and field testing of one of the concepts showed that the mechanism could reliably pick kiwifruit. Furthermore, this kiwifruit detachment mechanism was able to reach over 80 percent of fruit in the cluttered kiwifruit canopy, whereas the previous state of the art mechanism was only able to reach less than 70 percent of the fruit.

Artificial pollination was performed by detecting flowers and then spraying pollen in solution onto the detected flowers from a line of sprayers on a boom, while driving at up to 1.4 $ms^{-1}$. In addition, the height of the canopy was measured and the spray boom was moved up and down to keep the boom close enough to the flowers for the spray to reach the flowers, while minimising collisions with the canopy. The pollination system contributions described here include the design of flower detection systems, the calibration method used, a method for removing noise from stereo-matching algorithms and multiple generations of the boom height control algorithms. In addition, a dry pollination system was created, which produced fruit with similar statistics to commercial crops.

Mobile robot navigation was performed using a 2D lidar in apple orchards and vineyards. Lidar navigation in kiwifruit orchards was more challenging because the pergola structure only provides a small amount of data for the direction of rows, compared to the amount of data from the overhead canopy, the undulating ground and other objects in the orchards. Multiple methods are presented here for extracting structure defining features from 3D lidar data in kiwifruit orchards. In addition, a 3D lidar navigation system- which performed row following, row end detection and row end turns- was tested for over 30 km of autonomous driving in kiwifruit orchards. The row detection component of this navigation system worked well compared to an existing kiwifruit row detection method. Computer vision algorithms for row detection and row following were also tested. The computer vision algorithm worked as well as the 3D lidar row following method in testing.

The design of robust safety systems will be critical to the deployment of large mobile orchard robots. A risk assessment was performed for autonomous navigation with a large mobile orchard robot and found that the required risk reduction ($PL_r$) was a Performance Level of d or e, depending on the degree of interaction between the robot and untrained people. These are very demanding requirements and so an architecture was proposed that included multiple sensors with multiple layers of redundancy and diversity in order to reduce the risk of Common Cause Failures. Methods for pedestrian detection in camera and lidar data are presented here, including methods to detect pedestrians that are in hazardous situations. The most robust of the pedestrian detection methods tested was a high visibility vest detector using the intensity channel of a 3D lidar; this method was tested to have 100 percent true positives, 0 false negatives and 0 false positives within a range of 4 m for people wearing high visibility vests.

# Table of Contents

## Illustration Index

## Index of Tables

## Glossary

AMCL - Adaptive Monte Carlo Localisation.

AMMP - Autonomous Multipurpose Mobile Platform. The name given to the large mobile robot, which was developed by other team members for this research project.

Calyx (of fruit) - The blossom end of the fruit, opposite the stem.

CCF - Common Cause Failures, as discussed in ISO 13849-1:2015 [1].

CNN - Convolutional Neural Network.

CRC - Cyclic Redundancy Check.

FCN - Fully Convolutional Network.

FPS - Frames Per Second.

GPS - Global Positioning System.

GPU - Graphics Processing Unit.

HSV - Hue Saturation Value.

ICP - Iterative Closest Point.

ILG - Ionic Liquid Gel.

IMU - Inertial Measurement Unit.

INS - Inertial Navigation System.

IoU - Intersection over Union.

Leader (of kiwifruit vine) - The thick first branches of the kiwifruit vine, which in a pergola structured orchard, will typically run parallel to the directions of the rows, at the height of the canopy on both sides of a row.

Lidar - Light Detection And Ranging; scanning laser range finder.

LNA - Low Noise Amplifier.

LSTM - Long Short Term Memory.

NIR - Near Infrared.

PCA - Principal Components Analysis.

PCD - Point Cloud Data; a file format for point cloud data [2].

PID - Proportional, Integral, Derivative.

PL - Performance Level. As defined in ISO 13849-1 [1], a safety rating which is a measure of the probability of dangerous failure per hour.

$PL_r$ - Required Performance Level, referring to the required risk reduction, determined during a risk assessment [1].

RGB - Red, Green, Blue.

RNN - Recurrent Neural Network.

RPN - Region Proposal Network.

RTK - Real Time Kinematic.

SIL - Safety Integrity Level. As defined in IEC 61508 standards [3], a measure of the average probability of a dangerous failure on demand of a safety function.

SLAM - Simultaneous Localisation And Mapping.

SRP/CS - Safety Related Parts of the Control System [1].

SVD - Singular Value Decomposition.

SVM - Support Vector Machine.

ToF - Time of Flight.

YOLO - You Only Look Once [4]. A real time object detection CNN.

# 1 Introduction

This PhD research was a part of a wider project that aimed to develop mobile robots that would perform targeted pollen spraying and automated harvesting in kiwifruit and apple orchards. The project was largely funded by the New Zealand Ministry of Business, Innovation and Employment (MBIE) and was called the “MBIE Multipurpose Orchard Robotics Project”. The project was a collaboration between the University of Auckland, the University of Waikato, Robotics Plus Ltd and Plant & Food Research.

The University of Waikato and Robotics Plus Ltd developed various parts of the robotic systems, including spraying boom parts, some of the harvesting mechanisms and the Autonomous Multipurpose Mobile Platform (AMMP) (Figure 1). Plant & Food Research provided expertise in spraying systems, sensors and horticulture. The team members from the University of Auckland developed:

- Droplet spraying systems.
- Fruit and flower detection, tracking and targeting systems.
- Control systems for spraying booms and harvesting arms.
- Pollination and harvesting systems.
- The mobile robot navigation system.

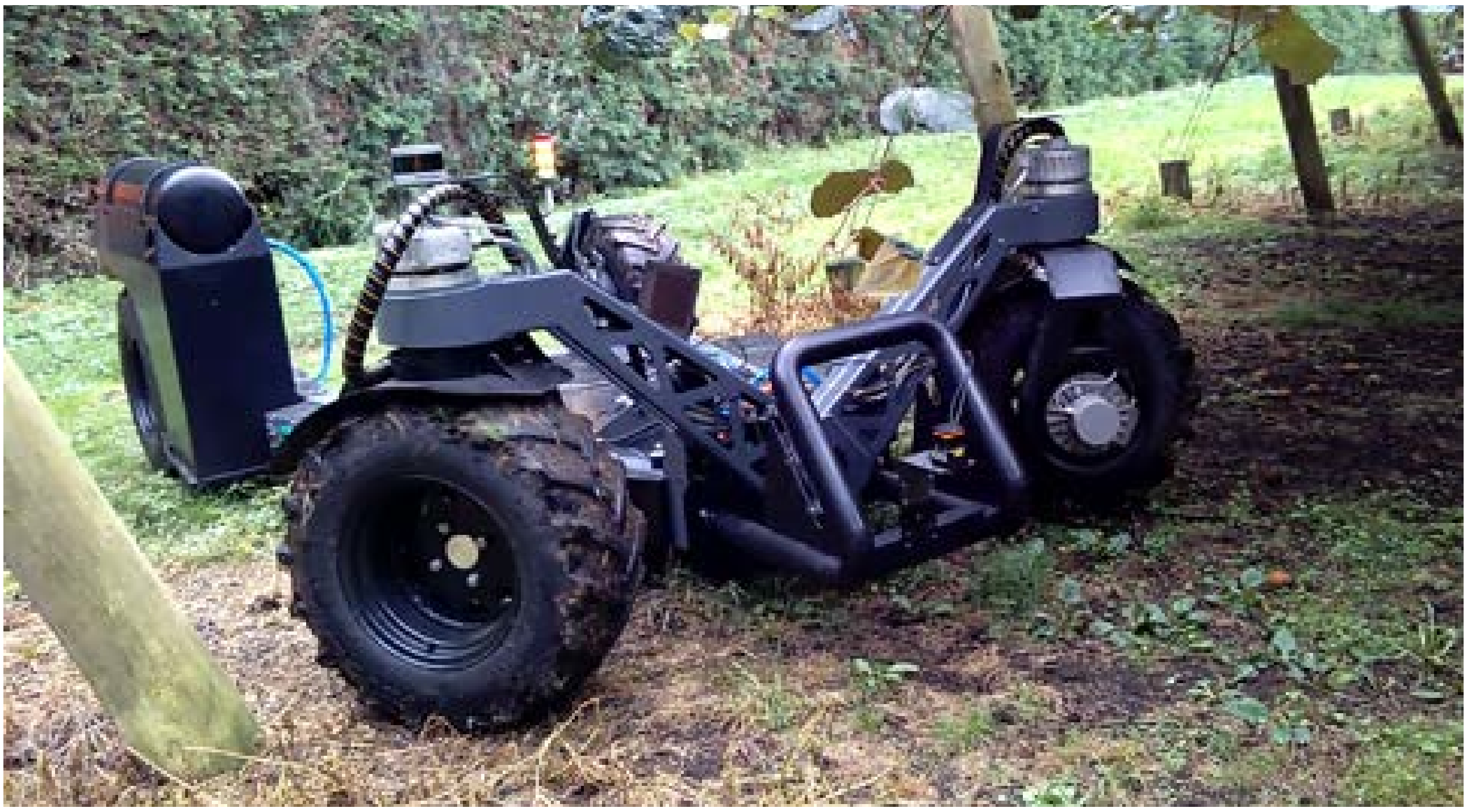

*Figure 1: The Autonomous Multipurpose Mobile Platform (AMMP), used in this project.*

## 1.1 Goals

Broadly, the goal of the research presented here was to develop improved methods for mobile robots to autonomously drive around orchards, while performing targeted pollination and automated harvesting. Kiwifruit orchards, in particular, were the focus. However, some work was also done for apple orchards.

The MBIE Multipurpose Orchard Robotics Project provided more specific goals for each part of the overall system. For the harvesting system, the key goal was:

> *"Greater than 80% of available fruit harvested with an average harvesting speed of 1 kiwifruit per second per robotic arm."*

This proved to be a challenging goal and, although over 50 percent of fruit were picked in three consecutive harvesting seasons, this goal of 80 percent of fruit harvested was not met by the team. Hence, a novel kiwifruit harvester was developed to try to approach this goal in a different way.

The pollination system detected individual kiwifruit flowers and sprayed those flowers with pollen in solution as the robot drove at up to 1.4 $ms^{-1}$. The pollen sprayers were mounted in a line on spraying booms and those booms were controlled to move up and down in order to stay close to the canopy so that the spray could reach the flowers (Figure 2). The key project goal for the pollination system was:

> "90% of targeted flowers hit."

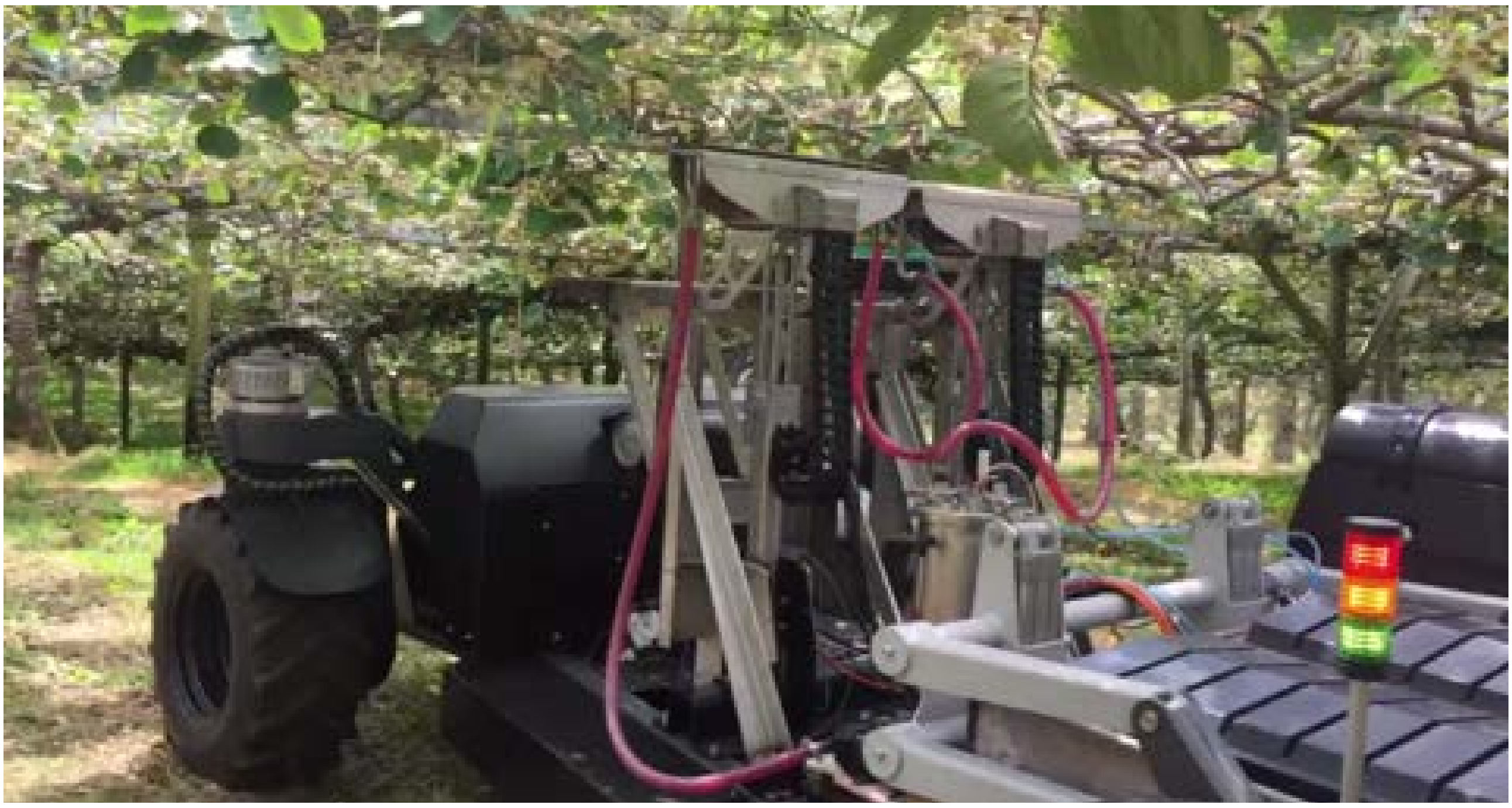

*Figure 2: The AMMP with two pollen spraying booms, which were moved up and down.*

This also proved to be a challenging goal and, although over 80 percent of flowers were hit during testing, this goal of 90 percent of flowers hit was not met by the team. Hence, a novel kiwifruit pollinator was developed to address the problem of robotic pollination in a different way.

The goals for autonomous driving were essentially to drive throughout kiwifruit and apple orchards, using multiple types of sensors, while performing mapping, localisation, object detection and obstacle avoidance. The required speed for the autonomous driving was 1.4 $ms^{-1}$.

## 1.2 Anticipated Impact

Kiwifruit and apples are very significant crops for New Zealand. In recent years kiwifruit and apples were New Zealand's second and third most valuable horticultural exports respectively, only being eclipsed by wine [5]. They are New Zealand’s most valuable fruit exports and their export value together has been approximately 1.4 billion New Zealand Dollars (NZD) per annum.

The New Zealand horticulture industry faces a number of challenges that mechanisation may be able to alleviate. For example, securing a reliable seasonal workforce has been a challenge for fruit growers in New Zealand [6]. Mechanisation has helped to reduce the workload, and hence the reliance on seasonal workforces, for the harvesting of many specialty crops. These harvesters typically shake the plant and the crop falls onto a collecting surface. Such harvesting methods are generally suitable for crops that are not sold fresh, such as frozen blueberries [7], wine grapes [8] or olives for oil [9].  Such harvesting methods can also be used for some crops that are not damaged by threshing or impact, such as walnuts [10] and almonds [11].

However, harvesting by shaking is not suitable for many fresh sold fruits that would be bruised and damaged by the process. A more gentle approach is required for such fruits. One approach that has been investigated by various groups is the use of cameras to localise the fruit and robotics arms to pick the fruit [12]–[14]. The work described here used a similar approach.

Another focus of this project is targeted spraying, which has many potential applications such as thinning or fertilising but, as an outcome of the work described here, will initially be focused at the problem of pollinating kiwifruit. Pollen from male kiwifruit flowers must be distributed to female flowers on separate vines [15]. The proportion of female flowers pollinated should be greater than 80% for some commercial crops [16]. The amount of pollen that each female flower receives is also important because the size of kiwifruit increases with the number of seeds [15].

Small amounts of wind pollination occurs naturally for kiwifruit and can be assisted by fans [15]. Honey bees can be used to achieve high rates of pollination; however, the quality of pollination of kiwifruit by bees is variable, depending on factors such as the health of the hives, the weather and other flowers in the vicinity that may attract the bees [17]. Kiwifruit can be artificially pollinated by hand spraying pollen suspended in water; however, this process is labour intensive and expensive. Kiwifruit can also be artificially pollinated by blowing clouds of pollen onto the kiwifruit canopy; however, this process is very wasteful since most of the pollen does not land on the flowers and pollen is expensive. Targeted automated artificial pollination, as presented in this thesis, may be able to improve the quality and reduce the cost of kiwifruit pollination.

It seems that mechanisation or automation of harvesting and pollination may be beneficial for New Zealand horticultural exports. Furthermore, the machines developed may themselves be exported or used to provide services internationally.

There are also potential global benefits of specialty crop mechanisation. Many tasks in orchards are repetitive and uncomfortable; mechanisation may help to free people from some of these tasks [18]. Improved task performance efficiency provided by specialty crop mechanisation should help to improve food supply for the growing world population and provide lower cost healthy diet alternatives. Food from animal products tend to have a much greater environment impact and contribute higher volumes of greenhouse gasses [19]; increasing the efficiency of specialty crop production should improve the supply of food for more sustainable diets.

There are many possible forms that automation in an orchard could take. This research is a part of a wider project to develop mobile robots for automation of pollination and harvesting in kiwifruit orchards. To a lesser extent, automation in structured apple orchards was also considered. It was expected that the development of these technologies would be a springboard for even more features in future projects.

## 1.3 Contributions

Many of the contributions presented in this thesis relate specifically to the problems of automated harvesting, pollination and navigation in kiwifruit orchards. This thesis is more closely related to biosystems engineering than general robotics or artificial intelligence, since many of the solutions presented here relate to engineering solutions that are specifically designed for and tested in kiwifruit orchards, in order to aid deployment of robots in kiwifruit orchards in the future. Nevertheless, some of the methods presented here take existing robotics and artificial intelligence algorithms and apply them to harvesting, pollination and navigation in kiwifruit orchards. The contributions of this thesis are:

1. Kiwifruit detection systems using various algorithms including a novel combination of an existing kiwifruit detection method with a classification Convolutional Neural Network (CNN) and multiple state of the art CNNs. These methods were tested both on ground truth datasets and in kiwifruit orchards, while harvesting fruit.
2. A novel study of different methods of kiwifruit detachment that lead to conceptual designs of multiple novel kiwifruit detachment mechanisms and field testing of kiwifruit detachment using different configurations of a novel kiwifruit detachment mechanism.
3. The design of all hardware and software for a novel kiwifruit harvester, which was tested and picked kiwifruit in a real world kiwifruit orchard. This included the successful testing of Time of Flight sensors in kiwifruit orchards for kiwifruit detection and a method for calibrating the sensor with a robot arm.
4. A novel method to detect and target unoccluded kiwifruit, which was combined with a novel mechanism design, sensor mounting and path planning methods for successfully harvesting kiwifruit that were obstructed from below.
5. The design of a novel kiwifruit flower detection system, including sensor selection, mounting design and algorithms for flower detection using hand engineered methods and CNNs. This system was tested and worked in real kiwifruit orchards.
6. A novel method for the calibration of the flower detection system with respect to the pollen sprayers. This system was tested with real flowers and also worked in real kiwifruit orchards.
7. Multiple versions of a novel boom navigation system, which kept the pollen sprayers in range of flowers, while avoiding collisions with the kiwifruit canopy and travelling at speeds that created tight margins. Multiple versions of this system were tested in real kiwifruit orchards.
8. A novel method for producing stereo data with less disparity errors by combining the output of different stereovision matching methods.
9. All hardware and software of a unique dry pollination system, which was tested in the real world and produced commercially viable fruit.
10. A novel navigation system for autonomous driving in kiwifruit orchards, using 3D lidar data. This system was tested in the real world on multiple platforms, including the AMMP. The navigation system included unique methods for row detection, row end detection and performing row end turns. The new row detection method performed well, when compared to an existing row detection method.

11. Multiple novel methods for feature extraction of structure defining features in kiwifruit orchards using 3D lidar. These methods were tested and compared using a ground truth dataset, which was generated using a unique labelling procedure, which included multiple innovations to save time during labelling and created multiple datasets.
12. Multiple methods for row detection using computer vision with camera images, including a method which was tested in the real world and had results comparable to the 3D lidar navigation system.
13. A high visibility clothing detector using 3D lidars. This pedestrian detection method is shown in this thesis to be very effective and may be more widely applicable to a range of industries where it is a standard operating procedure to wear high visibility clothing. This method reuses lidars that are also used for multiple other navigation functions.
14. Methods for detecting pedestrians in lidar data, including using state of the art capsule networks and CNNs.
15. Methods for detecting the presence of pedestrians in the path of a vehicle from a single camera image. These methods were developed using segmentation results from a stereovision pedestrian detection system and may be used for cross-checking the stereo detection results, particularly if the image from one of the stereo cameras is compromised.

Furthermore, multiple literature reviews, datasets, sensor selection analyses and robotic kiwifruit harvester studies are contributed. In addition, a risk assessment for the AMMP is contributed as a reference for others performing similar work.

# 2 Robotic Kiwifruit Harvesting

The New Zealand kiwifruit industry has faced difficulties in securing labour to perform harvesting. In the kiwifruit harvesting season of 2018, for example, the New Zealand Government's Ministry of Social Development declared an official labour shortage for the major kiwifruit growing areas in New Zealand [20]. A robotic kiwifruit harvester could be a valuable contribution to the kiwifruit industry in New Zealand in order to ease such labour shortages by supplementing human pickers. The MBIE Multipurpose Orchard Robotics Project goal for kiwifruit harvesting was:

> "*Greater than 80% of available fruit harvested with an average harvesting speed of 1 kiwifruit per second per robotic arm.*"

The key part of this goal is picking 80% of the fruit. The speed per arm goal is less important if the actual issue is speed per hardware cost. For example, it might also be acceptable for two arms to operate at half the speed, if each arm only incurs half the cost, without negative maintenance consequences and complexity effects. Furthermore, what is not stated in this goal is the acceptable percentage of fruit loss, through damage or being dropped. Ideally, there should be no fruit loss; however, even people cause fruit damage during kiwifruit picking so zero fruit loss may be unrealistic for a robot. In addition, there may be some trade off between harvesting speed and fruit loss, which the project goal did not take into account.

Because minimising fruit loss is important, bulk harvesting methods, such as shaking or threshing the whole canopy, are not considered in the scope of this thesis, since such methods can cause damage to fruit crops. The architectures of the robots considered in this section have sensors, which localise fruit, and robot arms with end effectors, which pick the kiwifruit.

I contributed to two of the robotic kiwifruit harvesting systems, which were developed as a part of the MBIE Multipurpose Orchard Robotics Project. The first was the main harvesting system, which was developed by multiple people from the University of Auckland, the University of Waikato and Robotics Plus. This harvesting system was largely based on the work of Scarfe [21] and is referred to in the rest of this document as the "original kiwifruit harvester". My contributions to the original kiwifruit harvester were in sensor selection, data collection, data labelling, detection algorithms and analysis of the failure modes.

As it became clear that there were limitations with the original kiwifruit harvester that prevented the project goals from being met, I decided to experiment with improvements to the original kiwifruit harvester. In doing so, I created my own kiwifruit harvester, for which all development related to kiwifruit harvesting was my own work. In the rest of this document, my kiwifruit harvester is referred to as the “second kiwifruit harvester”. My contributions to the second kiwifruit harvester were in sensor selection, arm selection, hardware layout design, data collection, data labelling, detection algorithms, fruit localisation algorithms, sensor to arm calibration algorithms, arm control algorithms, a kiwifruit detachment study, gripper and detachment mechanism design, building and tuning the entire harvester and the testing of the components.

This section describes my contributions to the original kiwifruit harvester and the second kiwifruit harvester. However, firstly, the following subsection surveys existing robotic harvesting systems.

## 2.1 Survey of Existing Robotic Harvesting Systems

A robotic kiwifruit harvester may consist of many components, including sensors, detection algorithms, algorithms for localisation in real world coordinates, arm trajectory planning algorithms, the end effector, amongst others. The literature review in this section is split into firstly determining where fruit are in the real world and secondly the design of the harvesting hardware, since these are key aspects of the contributions presented in this section. Determining where the fruit are in the real world includes sensor selection and detection algorithms as well as other related algorithms. The review of the harvesting hardware focuses mainly on mechanisms for fruit detachment.

### *2.1.1 Survey of Kiwifruit Localisation Methods*

This review begins by surveying works specifically related to kiwifruit detection. Earlier implementations of kiwifruit detection algorithms used hand engineered features. Fu et al. [22] performed kiwifruit detection at night using a colour camera and artificial lighting. Their kiwifruit detection algorithm firstly performed segmentation by pixel value thresholding, which was optimised beforehand by the Otsu method. Then dilation, area thresholding and Canny line detection were used to find the boundaries of the kiwifruit. Minimum bounding boxes were found for resulting connected lines and an elliptical Hough Transform was applied within the bounding boxes to find the elliptical shapes of the kiwifruit. The successful recognition rate of 88.3% achieved by this method may not be high enough for a commercial system, if it means that some of the remaining fruit are damaged as a result of the harvester responding to an inaccurate detection. Furthermore, harvesting kiwifruit at night is viewed as an unviable option because at night the fruit are damp with condensation, which has undesirable effects. It seemed that a method that can detect kiwifruit with daytime lighting variations would be a more suitable option for a kiwifruit harvester.

Cui et al. [23] presented similar work to Fu et al. [22] with similar issues. Wu et al. [24] performed segmentation in a similar way to Cui et al. [23] and Fu et al. [22] but also used the Watershed algorithm to segment adjacent fruit and performed classification using Haar features with a classifier. However, basing the segmentation on colour thresholding might make this method vulnerable to changes in lighting conditions.

Scarfe [21] also performed colour thresholding of the kiwifruit canopy. Following this thresholding, a Sobel filter was used to detect edges. A circular sliding window was moved across the colour thresholded regions of the image. In each window, the mass of edges was taken to be the sum of outputs from the Sobel filter and the inertia was calculated using these same values along with the radius of the pixels from the sliding window centre. The ratio of the calculated inertia to the mass was one feature used for classifying the window as a kiwifruit calyx. Another feature used was the ratio of pixel values above and below thresholds. The multiple uses of colour and intensity thresholds should make the method of Scarfe vulnerable to variable lighting conditions; this indeed was found to be the case in experiments with this method (Subsection 2.3).

In order to find methods for kiwifruit detection that were less vulnerable to variable lighting conditions, methods for the detection of other fruits were reviewed. De-An et al. [25] performed detection of apples and used multiple steps that were invariant to lighting including a median filter, shape feature extraction and classification using a SVM. However, they also performed segmentation using the hue in the HSV colour space by thresholding and region growing. It is unclear how robust this approach would be to different lighting conditions. In fact, similar methods were trialled in the MBIE Multipurpose Orchard Robotics Project and were found to require retuning with different ambient lighting.

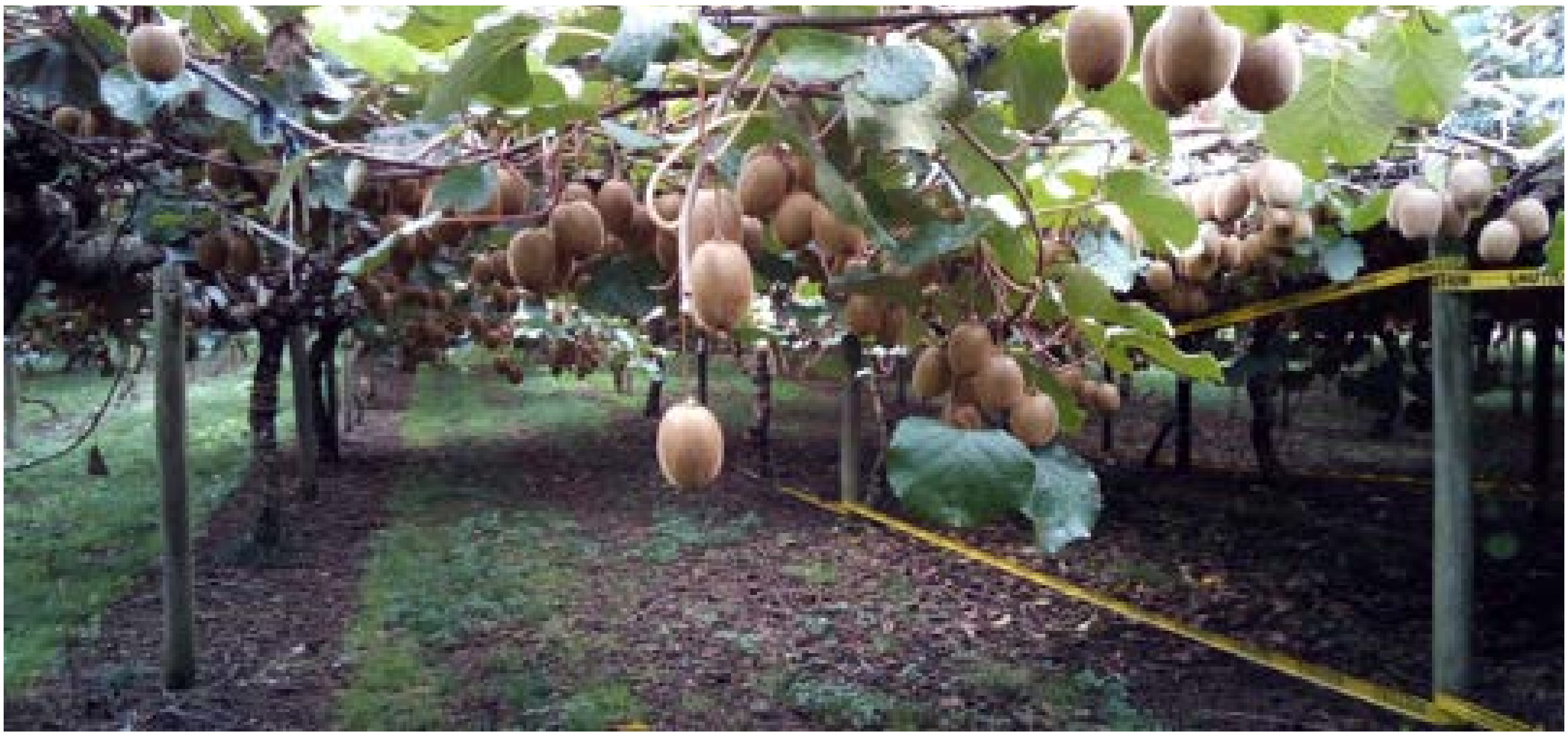

*Figure 3: Looking down a row of a kiwifruit orchard from within the pergola structure, where posts and kiwifruit trunks hold up a canopy of canes, beams, wires, fruit and leaves.*

Gongal et al. [26] also detected apples and successfully used multiple stages of colour filtering by blocking out ambient lighting using a shroud that encompassed a section of canopy. Nguyen et al. [27] also used a shroud in a similar way in order to allow apple detection with colour filtering. Because of the pergola structure of the kiwifruit canopy (Figure 3) and the considerable volume of uncontrolled canes that grow throughout a season before harvest, creating a similar shroud for kiwifruit seemed impractical and so this approach was rejected.

Wang et al. [28] used colour filtering successfully in their apple detection algorithm; however, their system was only intended to work at night without ambient lighting variations. Wachs et al. [29] used colour filters but also used a range of other methods, including circular Hough transforms, Haar wavelet feature extractors with fully connected neural network classifiers and combinations of different modalities. Fu et al. [30] used colour thresholding, morphological operations, edge detection and histogram equalisation for kiwifruit detection. Later work by Fu et al. [31] also used colour thresholding in the kiwifruit detection algorithm.

The prevalence of methods using colour as a feature for fruit detection led to a decision to review computer vision more widely. When this review was originally conducted in 2015, there were already many results that suggested that deep learning and Convolutional Neural Networks, in particular, may be the best performing methods for general image classification and object detection across a range of applications [32]–[35]. Now, at the time of finishing this thesis, there are many more results where deep learning performs better than other methods, including in the domain of fruit detection.

There has been a considerable amount of work in detecting fruit in RGB camera images using Convolutional Neural Networks to produce bounding boxes for the detected fruit. Fu et al. [36] used a smaller version of Faster R-CNN based on ZFNet to detect kiwifruit in images, which were taken looking up at a kiwifruit canopy from below. Bargoti and Underwood studied the use of Faster R-CNN to perform object detection of mangos, apples and almonds in orchards [37]. Stein et al. [38] used Faster R-CNN with a VGG-16 backbone to detect mangos in images for fruit localisation and yield estimation. Sa et al. [39] used Faster R-CNN with a VGG-16 architecture to detect rockmelons, strawberries, apples, avocados, mangos, oranges and sweetpeppers. Sa et al. [39] also experimented with fusing Near Infrared (NIR) and RGB data by adding the NIR data as an extra channel to the RGB data in the input of the CNN and also by averaging the outputs of two CNNs, one which took the RGB data as an input and the other which took the NIR data as an input. Based on this weight of evidence that Faster R-CNN can be used for detection of many types of fruit, including kiwifruit, it may now seem obvious that it could have been used for a kiwifruit harvester.

However, these works just discussed date to over 18 months after the start of the MBIE Multipurpose Orchard Robotics Project, which began in 2014.

Nevertheless, from surveying some object detection CNNs, Faster R-CNN [40] seemed like a good candidate for detecting kiwifruit because of its outstanding performance on publicly available datasets and its reasonable processing time. Faster R-CNN [40] improved on Fast R-CNN [41] by including a fast Region Proposal Network (RPN) after the main convolutional layers, instead of using slower region of interest proposal methods separately from the CNN. Fast R-CNN itself improved on the original Region CNN (R-CNN) [42] method, using shared computation between region proposals, by performing the calculations of the main convolutional layers only once, projecting region proposals to the output of those convolutional layers and using the resulting feature regions for ROI-pooling before performing bounding box regression and classification.

Faster R-CNN firstly calculates features from a succession of convolutional and max pooling layers [40]. These convolutional features are fed into a Region Proposal Network (RPN) and through ROI-pooling to the bounding box and classification fully connected layers. The RPN uses a sliding window approach to calculate region proposals from anchor boxes, which are set boxes of different sizes and scales that are centred at the centre of the sliding window. The resulting region proposals have bounding box coordinates and also a score for how likely that the region proposal is an object. These region proposals are used to determine the area of the convolutional features that are used for ROI-pooling, bounding box regression and classification [40].

Faster R-CNN achieved state of the art accuracy across multiple datasets in 2015 and was used in the first placed entries in the ILSVRC and MS COCO competitions [40]. Faster R-CNN runs at approximately 5 Frames per Second (FPS), depending on the variations and hyperparameters used.

Although it seemed that Faster R-CNN would work for detecting kiwifruit, it was unclear if it or any other object detection CNN would accurately detect long thin obstacles in the canopy, such as wires and branches, since the bounding boxes would enclose the objects, rather than detecting the boundaries of the objects. In addition, it was unclear if a highly accurate object detector like Faster R-CNN would be necessary for fruit detection. It seemed that the problem of kiwifruit detection in a canopy was a much more constrained problem than Faster R-CNN had been designed for, since there is only a small number of object classes to detect in the kiwifruit canopy, the artificial lighting conditions are designed to be favourable for detection during harvesting on the AMMP and there is only limited scale variability of the kiwifruit in the images. Hence, it seemed that a simpler CNN might be good enough to produce the accuracy required. A simpler CNN could produce faster results, which would have the advantage of enabling alternative control approaches such as visual servoing. Hence, it was decided to investigate the use of other object detection CNNs for fruit detection.

The You-Only-Look-Once (YOLO) CNN was one of the methods investigated for better real time performance [4], [43]. For YOLO a convolutional neural network is trained to predict a set number of bounding boxes for each of a set number of rectangular regions, in a grid across an image. For each bounding box, the probability that there is an object in the bounding box is calculated by the CNN. A single classifier is also trained for each rectangular region in the grid; the class probabilities produced in each region are conditional on there being an object in that region. The product of the probability of the class given that there is an object and the probability that there is an object in a corresponding bounding box is calculated to give a set of bounding boxes, weighted by the probability that the bounding boxes are of the predicted class. Then Non-Maximum-Suppression and thresholding are used to produce a set of object detection outputs. The YOLO method was shown to achieve higher mean average precision than other real time detection methods, when it was released. YOLO produced greater localisation errors than state of the art detection neural networks but produced less false positives on background and generalised between representation domains.

A key assumption made in the design of YOLO and derivative methods, such as DetectNet [44], is that there can only be a set number of objects in a set area. It seemed that with some care taken in the design of the computer vision hardware, that this could be a fair assumption for the kiwifruit harvester because of the relatively small variations in the size of kiwifruit in images, taken for the target field of view and from the target distance.

However, YOLO and DetectNet have the same issue as Faster R-CNN and other object detection methods: that they output bounding boxes that enclose the entire object. When detecting a thin object, such as a wire or branch, in the kiwifruit canopy and when that object is at a diagonal across the image, an enclosing bounding box would include a large area, which is not the thin object. For example, a wire which stretched from one corner of an image to the opposite corner of the image would have a bounding box that would be the entire image, even though the wire would only occupy a small proportion of the image; hence, this bounding box would not accurately describe the points in space to be avoided for obstacle avoidance, since there would be a large area that would unnecessarily be excluded from use. It would be better to use a method that detects the parts of the image that are associated with the thin object. Hence, it seemed that it would be beneficial to use a per pixel labelling technique, such as semantic segmentation, for determining the position and orientation of thin obstacles in the kiwifruit canopy.

The body of work considered here for implementing semantic segmentation was the fully convolutional networks of Shelhamer et al. [45]. Unlike earlier Convolutional Neural Networks, such as AlexNet [32] and LeNet [46], fully convolutional networks do not have fully connected layers near the output of the network. The fully convolutional networks of Shelhamer et al. [45] had multiple convolutional layers followed by multiple upsampling layers, which may be viewed as the opposite process of a convolutional layer. In addition, Shelhamer et al. [45] experimented with using different network architectures in the convolutional layers of the network and also implemented skip connections so that earlier layers could be aggregated with later layers without interaction with intermediate layers.

Further to semantic segmentation, it was postulated that instance segmentation may be useful. Instance segmentation produces individual segments for each instance of an object of a class; whereas, semantic segmentation only produces segments for each class, so that objects of the same class that are connected in an image would be segmented together as one mask. It seemed that instance segmentation could provide benefits for kiwifruit harvesting since the skin of each fruit could be segmented. It seemed that this could provide additional information about each fruit's pose, as opposed to just detecting the calyx and assuming that the orientation of the fruit is approximately vertical, which is not always true. State of the art results for instance segmentation have come from Facebook AI Research (FAIR) in the form of the Multipath Network [47] and Mask R-CNN [48]. Mask R-CNN is a modification of Faster R-CNN, with the following changes:

- A fully convolutional network is added in parallel with the classification and bounding box regression heads, in order to output the masks of the objects.
- RoiAlign is used to replace RoiPool, in order to maintain a consistent spatial relationship between the inputs and outputs [48].

Mask R-CNN outperformed 2015 and 2016 entrants of the COCO challenge for instance segmentation and object detection [48]. Mask R-CNN has also been used for fruit detection. Seabright et al. [49] used Mask R-CNN to detect calyxes on kiwifruit in colour camera images, which were taken from below the canopy with the cameras facing upwards. Liu et al. [50] used Mask R-CNN to detect objects in citrus canopies using a colour camera which faced horizontally. The classes Liu et al. used were unoccluded fruit, slightly occluded fruit, branch occluded fruit, leaf occluded fruit, overlapping fruit and the main branch of the tree. It seemed that some combination of the methods presented by Seabright et al. [49] and Liu et al. [50] may be a feasible approach for performing kiwifruit harvesting.

Mask R-CNN has also been extended to human pose estimation by training the fully convolutional network on one-hot encoding masks, representing keypoint positions, instead of segmentation masks. It seemed that a similar approach could be used to detect the poses of fruit for improved reasoning about the exact target pose of the hand, when picking. Other options considered for pose estimation using keypoint detection included Convolutional Pose Machines [51] and Part Affinity Fields [52]. However, these pose estimation methods are designed for human pose estimation, where there are multiple keypoints of different classes that are associated with each instance of a person. In contrast, kiwifruit pose estimation may be considered to be much simpler with only two keypoints (which could be the calyx and stem attachment point at opposite ends of the fruit) that are connected for each instance of a fruit. This would reduce kiwifruit pose estimation to estimating two points or four numbers per instance in an image, which is the same output type as object detection with bounding boxes and so it was postulated that an object detection CNN could be used for kiwifruit pose estimation.

### 2.1.2 Survey of Kiwifruit Harvesting Robots

Many of the existing kiwifruit harvesting robots have used an end effector which grips and rotates the fruit. Fu et al. [53] experimented with minimising the force required to detach kiwifruit from the stem and found that applying an upwards rotation to the fruit of 120 degrees (so that the fruit was 60 degrees from upside-down) required less force for detachment than with a rotation of 90 degrees or 0 degrees. Fu et al. [53] also developed a mechanism that approached the fruit from below, gripped the fruit between two contact surfaces and then rotated upwards in order to perform fruit detachment. A key issue of the mechanism of Fu et al. was its size, with the mechanism requiring approximately 0.1 of a metre to be clear behind a fruit in order to close the mechanism for gripping before rotating the fruit upwards for detachment. This amount of space is not available around many fruit in a kiwifruit canopy and attempting to occupy more space in a kiwifruit canopy can cause fruit to be damaged or knocked off, as is shown later in Subsection 2.4. In addition, the time required to pick a fruit for Fu et al. was 22 seconds, which was much slower than the goal rate for the MBIE Multipurpose Orchard Robotics Project.

Chen et al. [54] also created a mechanism that gripped and rotated kiwifruit, while performing detachment in a laboratory experiment; however, this mechanism was much larger than that of Fu et al. [55] and so would be impractical in a real world kiwifruit canopy. Furthermore, the mechanism of Chen et al. had a picking time of 9 seconds and a 90 percent detachment rate in laboratory conditions; preferably, both of these performance measures would be improved in a commercial system.

Mu et al. [56] presented an end effector that was less intrusive in the canopy and had a better cycle time of 4 seconds. As with some other kiwifruit harvesting robots, the picking mechanism of Mu et al. [56] gripped and rotated the fruit to perform detachment. An interesting aspect of this mechanism was the use of infrared proximity sensors in the gripper in order to improve the final pose of the gripper and the use of pressure sensors to adjust the gripping force. Mu et al. [57] presented updated versions of this work with a successful pick rate of 94 percent. However, it appears that their mechanism would significantly impact adjacent fruit in clusters.

Scarfe [21] developed an arm that approached kiwifruit from below and a detachment mechanism that gripped and rotated the kiwifruit in a single motion as opposed to having separate gripping and rotating movements. The arm and mechanisms of Scarfe were later imitated as a part of the MBIE Multipurpose Orchard Robotics Project [58], [59].

Barnett [60] developed an end-effector for kiwifruit detachment, which enclosed the fruit in a cup, closed the top of the cup with two shutters and rotated the cup before pulling down on the fruit and releasing the fruit. The key novel aspect of this design was instead of gripping the fruit, as with other fruit detachment mechanisms, the fruit was enclosed in the cup and the top of the cup was blocked, using two shutters. However, in order to get the fruit into the cup, the cup had to be manoeuvred from below the fruit and then upwards to enclose the fruit; this is an issue for any fruit which is obstructed from below by obstacles, such as branches and wires. As discussed later in Subsection 2.4 and Subsection 2.7.1, this is a significant proportion of the fruit. This issue of not being able to reach fruit, which are obstructed from below, applies to all of kiwifruit detachment mechanisms surveyed in this subsection up to this point because they all approach the kiwifruit exclusively from below.

Graham et al. [61] developed a kiwifruit detachment mechanism that uses a linear actuator to operate shears, which cut the stem. The key issue with this approach is that it leaves a part of the stem on the fruit and it is not acceptable practice to leave the stem on kiwifruit during picking because the stem can puncture other fruit in the storage bins, which leads to the punctured fruit being rejected. Furthermore, the mechanism of Graham et al. [61] was very large, requiring 0.33 of a metre of space to be available behind the fruit and 0.1 of a metre of space to be available beside the fruit; this amount of space is not commonly available adjacent to fruit in kiwifruit canopies.

Ting et al. [62] experimented with different detachment methods for kiwifruit. They rotated the fruit upwards about the stem attachment point at different angles before pulling downwards and measured the downwards pulling force; the benefit of this method of rotating the fruit upwards before pulling down was that it required less force than some other methods that they tried but it also could require the fruit to be moved into space that may be obstructed in the kiwifruit canopy. They also experimented with rotating the kiwifruit about the longitudinal axis of the fruit but found that an unacceptable length of stem was left on the fruit after detachment. They also conducted trials with a hook around the stem and pulling downwards, while measuring the force and found that this method required a moderate amount of force, compared to the other methods tested. They also experimented with applying a shear force at the stem attachment point and measured the force; however, this method was found to require the highest force out of the measured methods. Finally, they showed that fruit could be detached by striking the stem at a high speed; this method was demonstrated using a drill with soft tubing. It was noted by Ting et al. [62] that with stem striking a balance had to be struck between:

- Not detaching the stems due to there not being enough momentum at the contact point
- The ideal case where the stem was detached at the stem attachment point
- And cutting the stems with a surface that was too hard, sharp or fast.

The testing of the stem striker was performed later in the harvest season and it is unclear how this would affect the performance of this method, given that fruit can be easy to detach when they are more ripe. Also, it was noted that there was a risk of fruit damage due to the stem striker hitting the targeted or adjacent fruit, instead of the stems.

Other paradigms for picking fruit include gripping and pulling as described by Taqi et al. [63]; however, as studied by Ting et al. [62], using the approach of gripping and pulling kiwifruit requires more force for detachment than some other approaches and also causes the canopy to shake, which can cause damage from fruit hitting adjacent objects or fruit can fall uncontrollably. Fruit may also be removed by suction [64]; however, this approach has similar issues to gripping and pulling for kiwifruit. Targeted shake and catch harvesting systems show promising results for apple removal efficiency; however, even with the catching systems in place, between 6 and 14 percent of fruit can be downgraded due to damage [65].

## 2.2 Sensors for Kiwifruit Harvesting

The sensors trialled for kiwifruit harvesting were stereo camera pairs and a Time of Flight (ToF) sensor. Stereo cameras have previously been used by various researchers for kiwifruit harvesting [21], [66] and harvesting of other fruit [50], [67]–[69]. The use of the ToF sensor was similar to the use of 3D sensors by various researchers, including the use of Swissranger sensors for sweet-pepper detection [70] and the use of Kinect sensors for the segmentation of objects in apple tree canopies [71], [72] and on a tomato harvesting robot [73].

In the MBIE Multipurpose Orchard Robotics Project, various models of cameras were used for the stereo camera pairs, including Logitech C920 webcams [74], Basler ACA1920-40UC cameras [75] and a Stereolabs Zed [76], amongst others. The Time of Flight sensor used was a Basler tof640-20gm_850nm [77]. The use of Time of Flight as the sole sensor for kiwifruit harvesting, outdoors and during the day, was somewhat novel. There was some doubt as to whether this ToF sensor would give useful data, given that other experiments have shown that ToF sensors can give noisy data in orchards [78]. Nevertheless, it was postulated that under a well developed kiwifruit canopy, there might be enough shade to minimise the effects of ambient lighting- as may have been the case with apple orchard robotic harvesting in a V-trellis canopy with a ToF sensor [79].

For the stereo cameras, data collection was performed in orchards with the cameras facing up towards the canopy at heights of approximately 0.5 m to 1.0 m from the canopy. Different lighting was used in the stereo camera data collection, including with different LED set ups and also with no artificial light. For the Time of Flight data collection, intensity and depth data was collected with the sensor facing upwards and approximately 0.4 m to 0.8 m from the canopy. Data was collected at different times, including night time, and in several different orchards, including green and gold kiwifruit variety orchards. Over 20,000 images were collected, from which only a small proportion were selected for creating labelled datasets.

## 2.3 Kiwifruit Detection for the Original Kiwifruit Harvester

The original kiwifruit harvester (Figure 4) was developed by many people from the MBIE Multipurpose Orchard Robotics Project. Each arm of the original kiwifruit harvester had a single detachment mechanism, a chute for the fruit to roll down and a single pair of cameras for stereovision, with LED lighting to illuminate the canopy (Figure 5). The cameras used in the stereo camera pairs were Basler ACA1920-40UC [75] cameras with Kowa LM6HC lenses [80] (Figure 6). The approach taken to determine the position of fruit was to detect kiwifruit in individual images and then the position of fruit was determined using stereopsis.

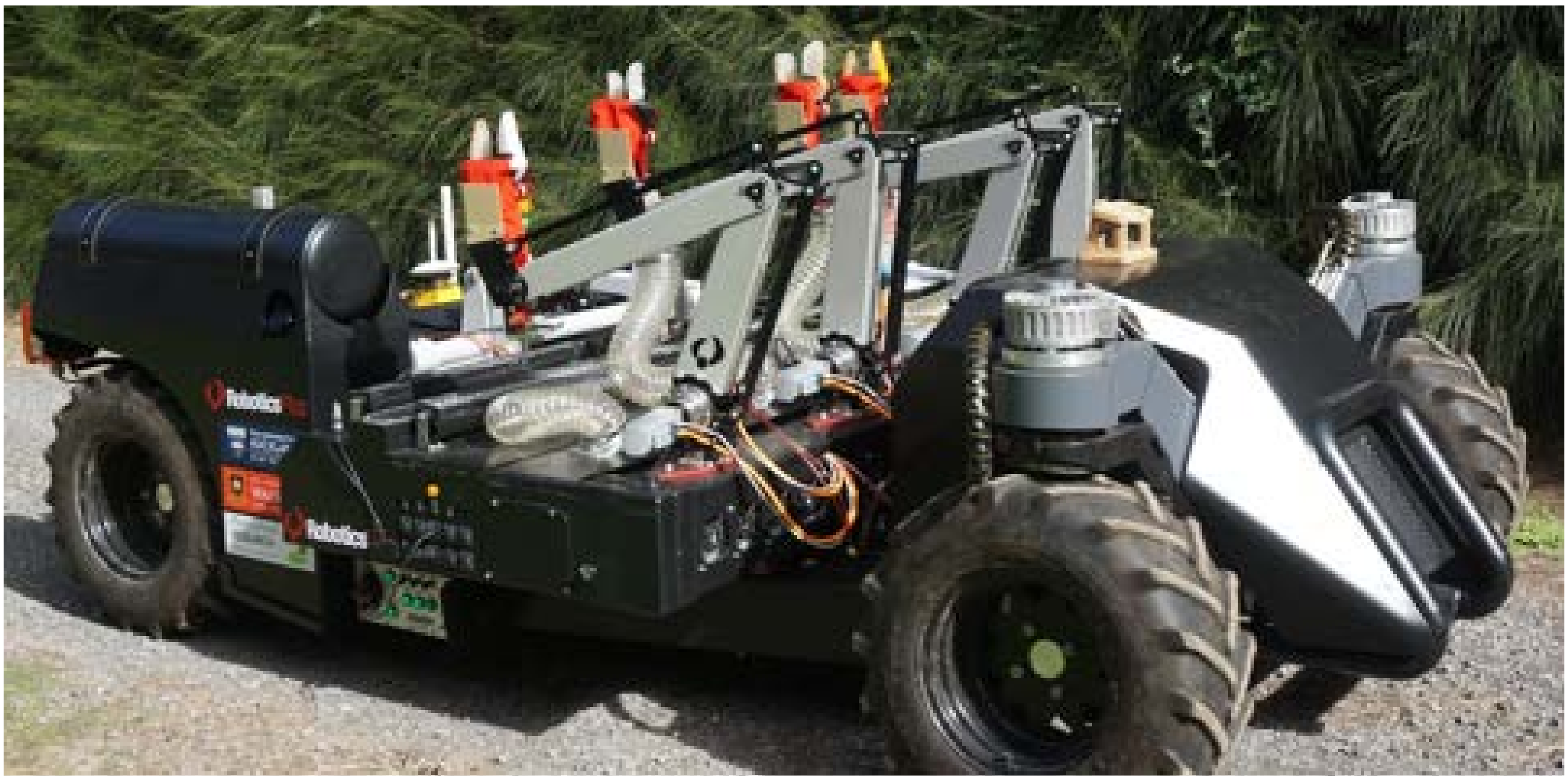

*Figure 4: The final version of the original kiwifruit harvester with four arms mounted on the AMMP. Photo courtesy of Matthew Seabright.*

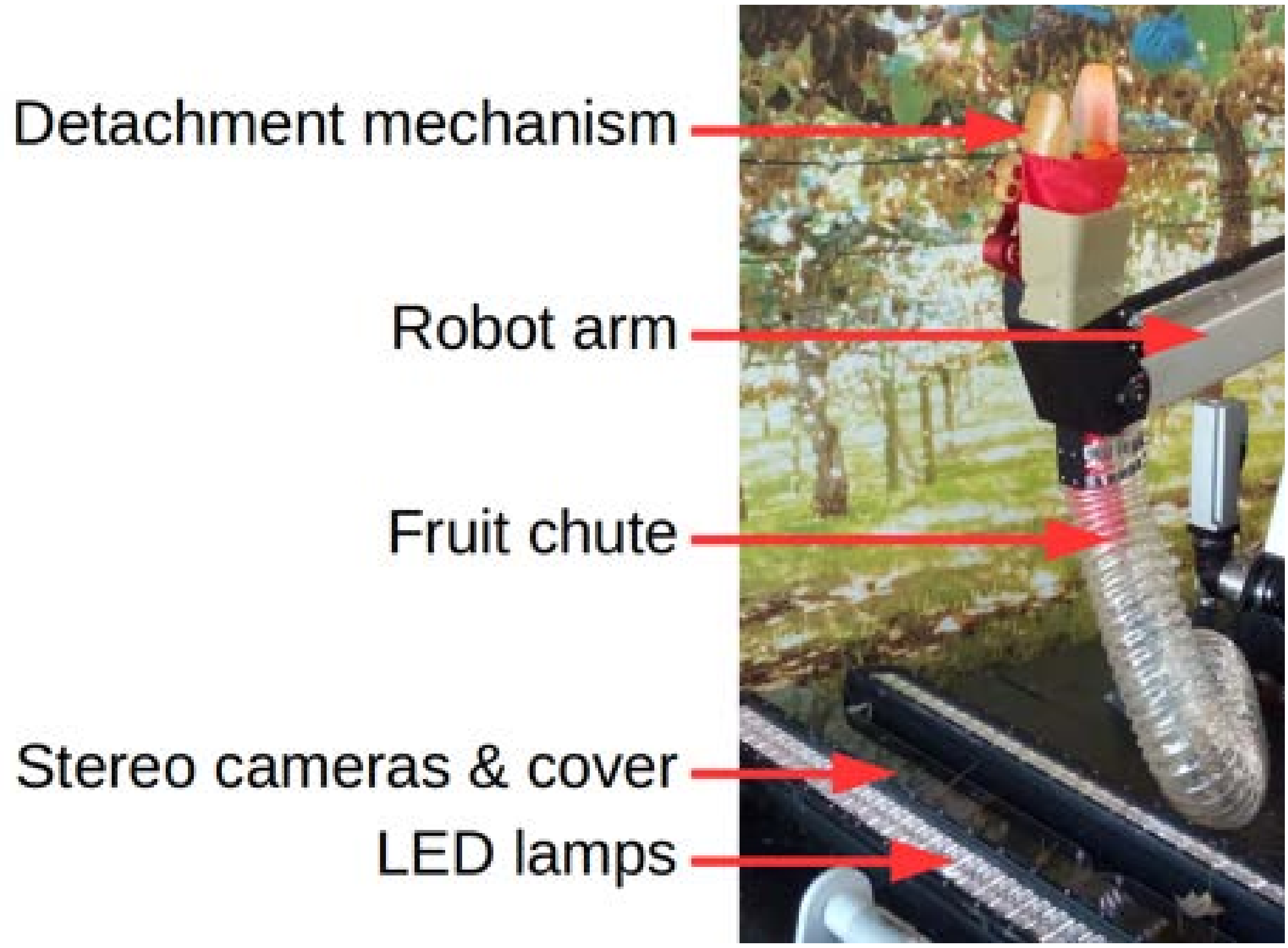

*Figure 5: Key hardware components of the original kiwifruit harvester.*

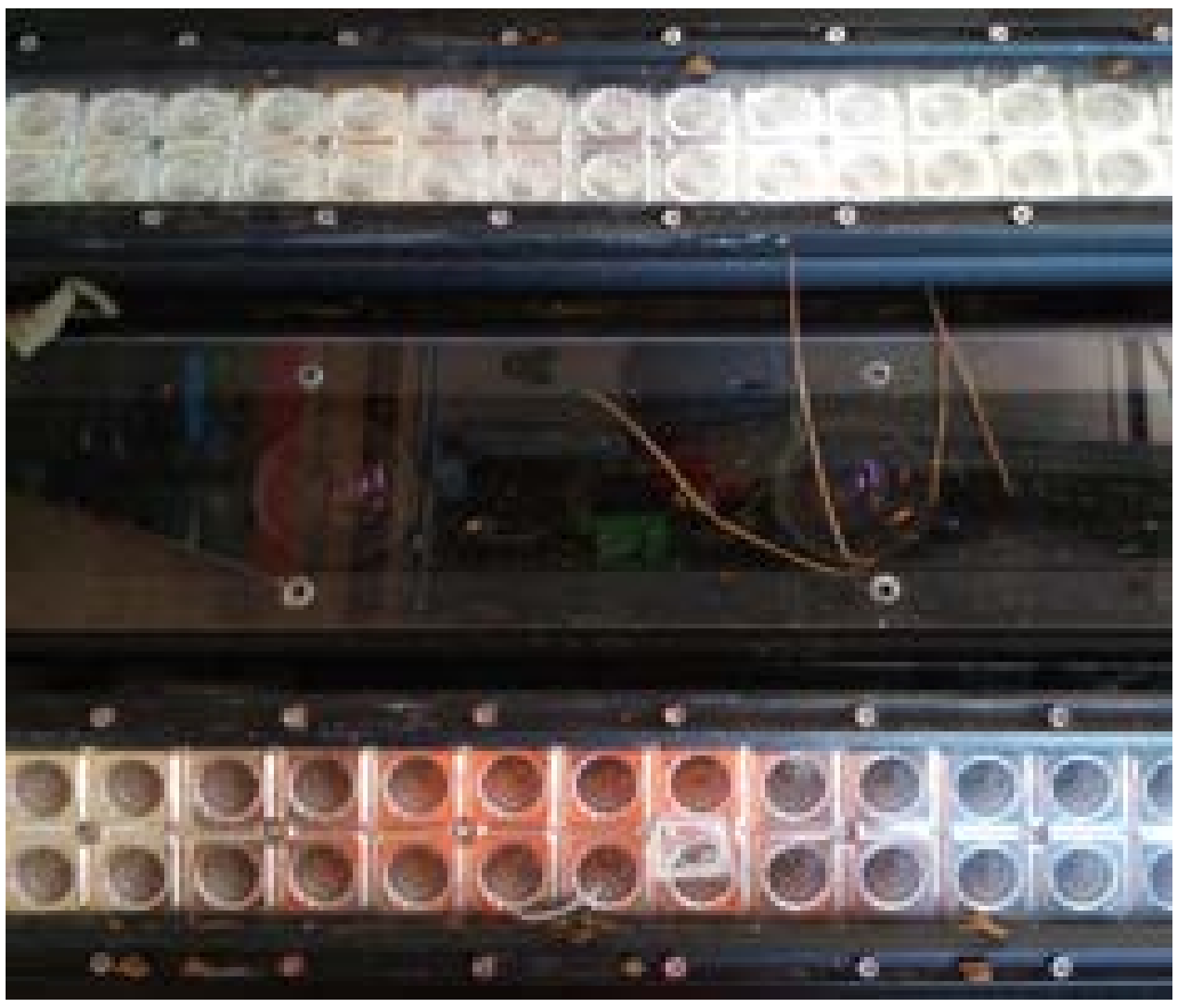

*Figure 6: A stereo camera pair used on a later version of the original kiwifruit harvester.*

The first season for testing the kiwifruit detection and harvesting algorithms for the Multipurpose Orchard Robotics Project started in April 2016. Initially, some of my colleagues reimplemented the kiwifruit detection algorithm of Scarfe [21] and collected results for this algorithm in kiwifruit orchards. I manually analysed some of these results by counting the fruit in each image, looking at the detections and classifying the detections as a true positive or a false positive. A result was classified as a true positive if a circular window classified as a kiwifruit had more than 50 percent overlap with a manually labelled ground truth bounding circle. The method of Scarfe used a circular sliding window and so ground truth circles were used for this analysis. This analysis is summarised in Table 1. The recall of 0.51 seemed too low to reach the goal of 80 percent harvested fruit and it seemed that the precision of 0.23 would cause the robot to try to pick many non-existent fruit. Hence, it seemed that something had to be done to improve on these results.

*Table 1: Summarised results from testing the kiwifruit detection algorithm of Scarfe [21].*

| **Number of Fruit** | **True Positives** | **False Positives** | **Precision** | **Recall** |
|---|---|---|---|---|
| 216 | 111 | 369 | 0.23 | 0.51 |

### 2.3.1 Kiwifruit Detection by Scarfe Region Proposals and AlexNet

It seemed that a high proportion of false positives were produced, when the parameters of Scarfe's method [21] were optimised to detect as many kiwifruit as possible across the full range of lighting conditions encountered. In order to improve on Scarfe's method, it was decided that extra processing would be required in order to reduce the false positive count. The approach adopted was to create a kiwifruit object detection algorithm by using Scarfe's method as an application specific region proposal method to produce input for a Convolutional Neural Network (CNN) classifier. The CNN used was an AlexNet [32] with only two classes: "kiwifruit" and "not kiwifruit". The training dataset was created from 5000 hand-labelled images, which were augmented using horizontal shifts, rotations and resizing to inflate the dataset to 200,000 images. 199,000 images from this dataset were used to train the AlexNet model, using Stochastic Gradient Descent with a base learning rate of 0.01 multiplied by 0.1 every 20 epochs for a total of 60 epochs. The deep learning framework used was Caffe [81], run within Nvidia Digits [82].

With the AlexNet trained, the validation accuracy on 1000 images, separated from the dataset of 200,000 images, was 100 percent. The steps used to perform kiwifruit detection were:

1. Scarfe's method was used to produce region proposals.
2. Each region proposal was resized to 227 pixels high and wide.
3. Each resized region proposal was classified by the trained AlexNet.

4. All region proposals classified as “not kiwifruit” were rejected for further processing.

These steps were performed on images from both cameras of a stereo pair. The resulting kiwifruit detection coordinates were fed into the stereo-matching algorithm. My contribution to this algorithm was the use of AlexNet as a classifier; the rest of the algorithm was implemented by other team members.

As can be seen in Table 2, the precision was much higher with the inclusion of the AlexNet classifier in the kiwifruit detection algorithm. The recall was still quite low; however, it was noted that this was lighting dependent as can be seen in Figure 7. This suggests that if the artificial lighting had been more uniform, the results might have been better.

Ultimately, the hybrid Scarfe/AlexNet method was the algorithm used on the original kiwifruit harvester for testing during the 2016 harvesting season from April to June. Testing of the original kiwifruit harvester was performed by other members of the team with 51.5 percent of fruit successfully picked and 11.4 percent of fruit damaged. The percentage of fruit picked was higher than the recall in Table 2 partly because the harvester was moved during testing so that there were multiple chances for each fruit to be detected and picked.

*Table 2: Results from testing Scarfe's method for region proposals and AlexNet for classification.*

| Number of Fruit | True Positives | False Positives | Precision | Recall |
|---|---|---|---|---|
| 212 | 108 | 4 | 0.96 | 0.51 |

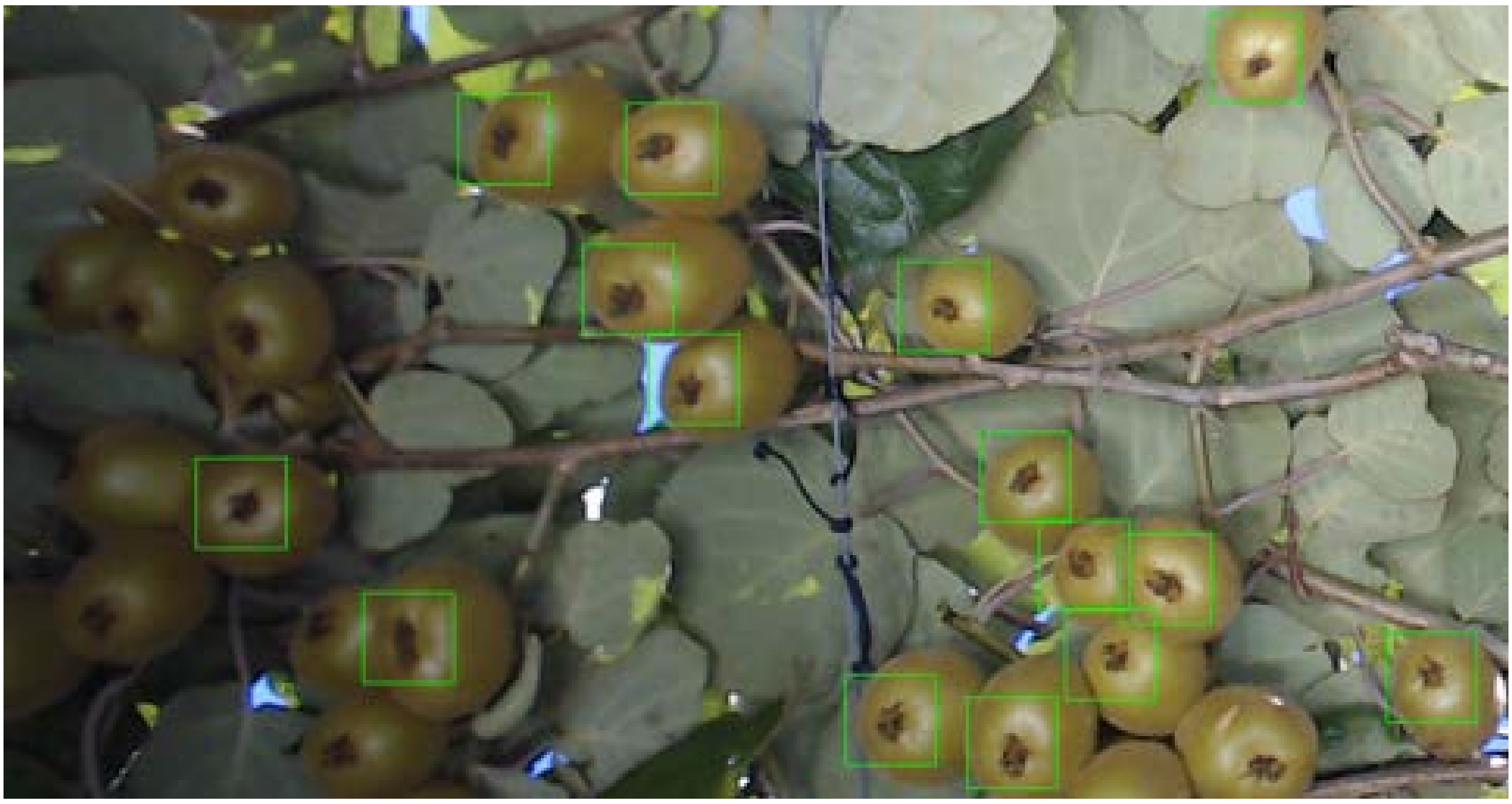

*Figure 7: Demonstrating the better performance of the Scarfe/AlexNet fruit detection method under stronger artificial illumination.*

### 2.3.2 Kiwifruit and Obstacle Detection by Semantic Segmentation

During the first season of testing with the original kiwifruit harvester, the robot arm and hand frequently hit wires and solid branches. The effects of these collisions included pushing the arm off course and blocking the arm from reaching its target. Hence, it was deemed essential that the detection algorithms should not just detect fruit but also potential obstacles.

Initial experiments were performed with using semantic segmentation for detecting obstacles for harvesting. The network architecture initially used was an FCN-AlexNet [45], using transfer learning from an AlexNet [32], which was pretrained and released as a part of the Caffe Model Zoo [81], [83]. A small exploratory dataset of 25 images was initially created. Only three classes were used: kiwifruit calyx, kiwifruit skin and wires. An inference result on a test image from this trial is shown in Figure 8. What was encouraging about such results was that a large selection of the wire was correctly, if somewhat roughly, detected in the images. Based on these results, it was hypothesized that segmentation with convolutional neural networks could be used for detecting the obstacles in the canopy with sufficient accuracy to establish zones where the robot arm should not go. Subsequently, semantic segmentation was used on the original kiwifruit harvester [58].

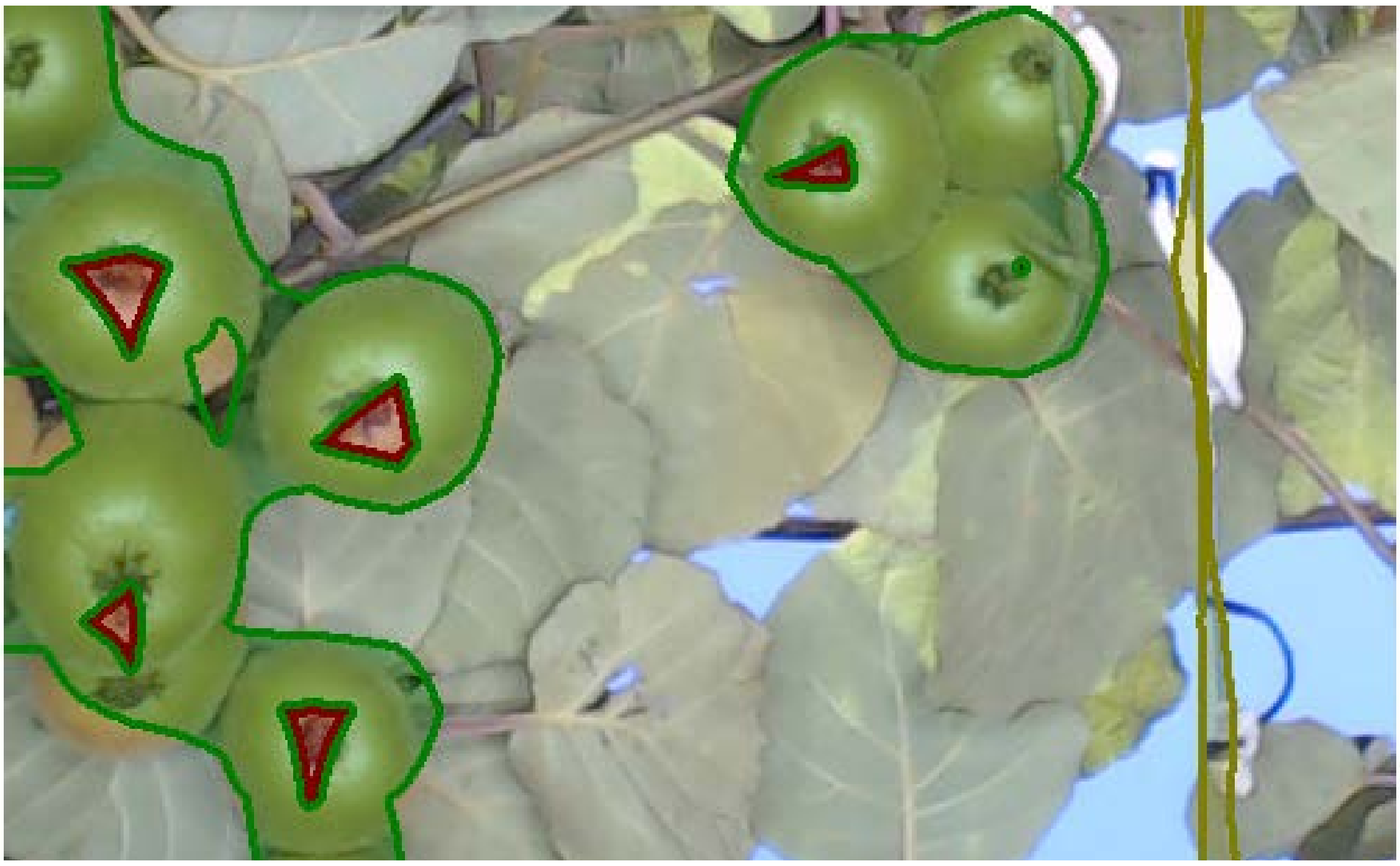

*Figure 8: An initial trial inference result from an FCN-AlexNet, trained on a small dataset of kiwifruit canopy images, with calyxes (red), kiwifruit skin (green) and wire (mustard).*

### 2.3.3 Kiwifruit and Obstacle Detection by Instance Segmentation

It was postulated that instance segmentation would provide advantages over semantic segmentation for harvesting by allowing the boundaries of individual fruit to be determined. It was thought that this could allow for the refinement of the calculation of the target pose of the harvester end effector as opposed to just using the calyx position, as was the case with semantic segmentation.

The instance segmentation CNN used was Mask R-CNN [48]. A Mask R-CNN model with a GoogLeNet [34] backbone, which was pretrained by others on the COCO dataset [84], [85], was used for transfer learning. This Mask R-CNN model was trained separately on RGB colour images and on the intensity channel of Time of Flight data from a Basler TOF640-20GM-850NM [77]. The properties of the Time of Flight and colour camera datasets are given in Table 3 and Table 4 respectively. The training hyperparameters used are given in Table 5 and were based on values that have previously worked well for other datasets [85].

*Table 3: The properties of the Time of Flight dataset for instance segmentation on the original kiwifruit harvester.*

| **Dataset Property** | **Values** |
|---|---|
| Image size | 640 x 480 |
| Number of training images | 6 |
| Number of validation images | 3 |
| Classes | Calyx, Kiwifruit Skin, Wires, Beams, Branches |

*Table 4: The properties of the colour camera dataset for instance segmentation on the original kiwifruit harvester.*

| **Dataset Property** | **Values** |
|---|---|
| Image size | 500 x 500 |
| Number of training images | 16 |
| Number of validation images | 8 |
| Classes | Calyx, Kiwifruit Skin, Wires |

*Table 5: Hyperparameters used for training Mask R-CNN.*

| **Hyperparameter Description** | **Hyperparameter Value** |
|---|---|
| Number of training steps | 10000 |
| Solver type | Momentum Optimiser |
| Learning rate | 0.0007 |
| Batch size | 1 |

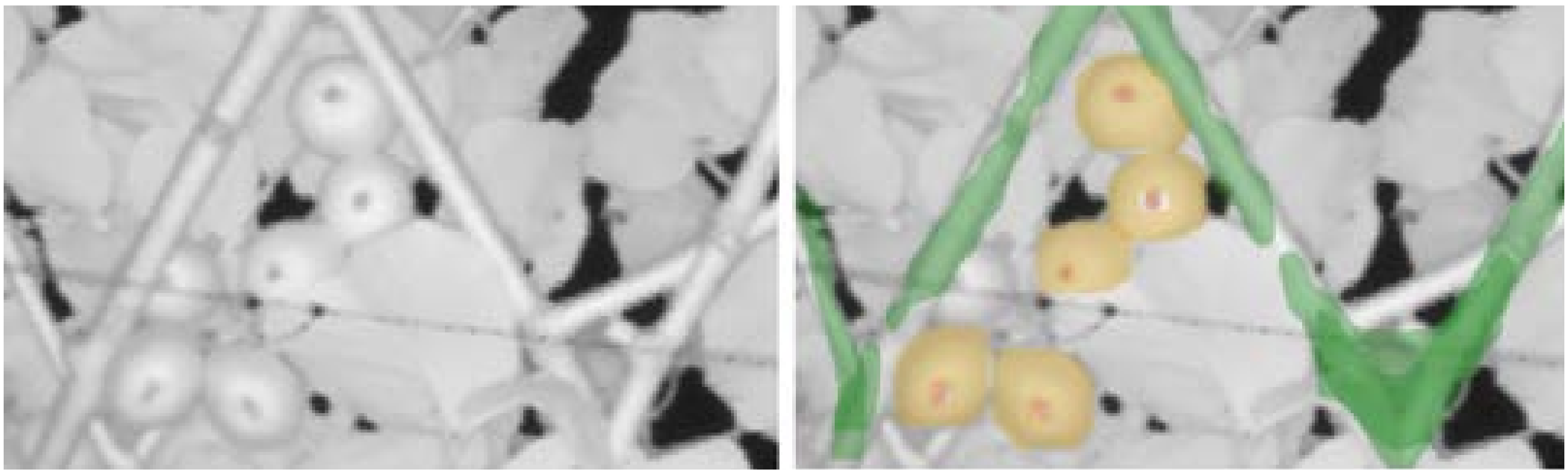

*Figure 9: Example inference result for an input Time of Flight intensity test image (left) with the output from Mask R-CNN (right).*

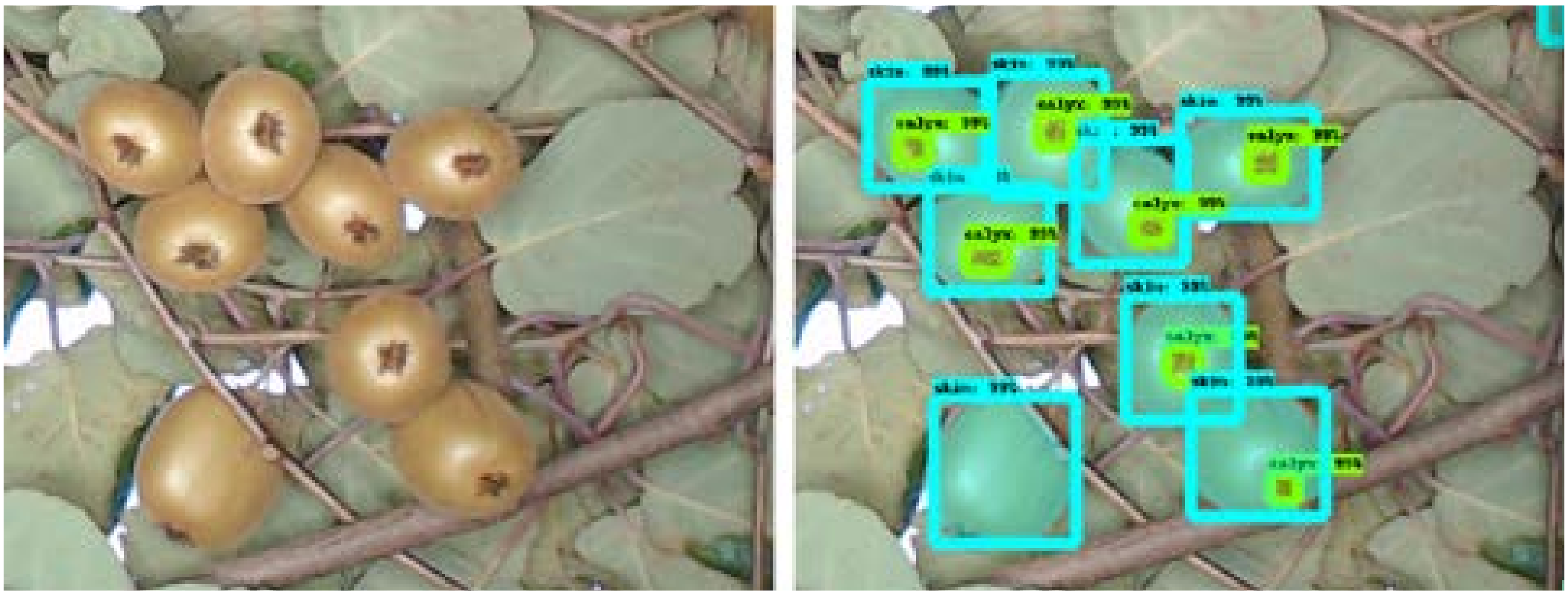

*Figure 10: Example inference result for a test image (left), with the output (right) from Mask R-CNN, which was trained with colour images to detect wires, kiwifruit skin and kiwifruit calyxes.*

Example inference results from unlabelled test datasets are shown in Figure 9 and Figure 10. The metrics used for assessing the performance of the instance segmentation were the Average Precision and Average Recall as defined by the COCO dataset detection evaluation criteria [86]. The validation results are given in Table 6 and Table 7. The Average Precision results are similar to results from other models and datasets [87]. The numbers of images in these datasets are small; however, the neural network is able to begin to learn to detect the labelled objects so well because of transfer learning and because instance segmentation labels provide many bits of signal per image due to the pixel level labelling of objects.

Although the instance segmentation approaches with colour cameras and ToF sensors were developed for the original kiwifruit harvester, they were never actually tested in the real world. However, the results presented here with small datasets were encouraging enough that instance segmentation with Mask R-CNN was used later with the second kiwifruit harvester (Subsection 2.6.2).

*Table 6: Validation results from training Mask R-CNN with Time of Flight intensity channel data for detecting kiwifruit canopy objects.*

| Metric Description | | | | Metric Value |
|---|---|---|---|---|
| **Type** | **IoU** | **Object Pixel Area** | **Given Detections per Image** | |
| Average Precision | 0.50 | Any | 100 | 0.44 |
| | 0.75 | Any | 100 | 0.34 |
| | 0.50:0.05:0.95 | Any | 100 | 0.25 |
| | | $< 32^2$ | 100 | 0.00 |
| | | $> 32^2, < 96^2$ | 100 | 0.21 |
| | | $> 96^2$ | 100 | 0.26 |
| Average Recall | | Any | 1 | 0.14 |
| | | Any | 10 | 0.24 |
| | | Any | 100 | 0.28 |
| | | $< 32^2$ | 100 | 0.01 |
| | | $> 32^2, < 96^2$ | 100 | 0.26 |
| | | $> 96^2$ | 100 | 0.28 |

*Table 7: Validation results from training Mask R-CNN with colour camera images for detecting kiwifruit canopy objects.*

| Metric Description | | | | Metric Value |
|---|---|---|---|---|
| **Type** | **IoU** | **Object Pixel Area** | **Given Detections per Image** | |
| Average Precision | 0.50 | Any | 100 | 0.53 |
| | 0.75 | Any | 100 | 0.15 |
| | 0.50:0.05:0.95 | Any | 100 | 0.20 |
| | | $< 32^2$ | 100 | 0.06 |
| | | $> 32^2, < 96^2$ | 100 | 0.25 |
| | | $> 96^2$ | 100 | 0.00 |
| Average Recall | | Any | 1 | 0.04 |
| | | Any | 10 | 0.23 |
| | | Any | 100 | 0.26 |
| | | $< 32^2$ | 100 | 0.10 |
| | | $> 32^2, < 96^2$ | 100 | 0.30 |
| | | $> 96^2$ | 100 | 0.00 |

## 2.4 Original Kiwifruit Harvester Analysis

Harvesting was tested on the original kiwifruit harvester in various orchards for three harvesting seasons from 2016 to 2018. Table 8 provides a summary of the 2018 tests of the original kiwifruit harvester; these and other details outside the scope of this thesis are described by Williams et al. [59].

*Table 8: Summary of harvesting results for the original kiwifruit harvester in 2018.*

| Successful picks | Dropped fruit | Fruit left in the canopy |
|---|---|---|
| 56% | 9% | 35% |

A fruit was classified as "dropped" if it did not fall into the chute for the fruit, which on the original kiwifruit harvester was below the hand mechanism. Dropping fruit would be a negative characteristic of a commercial robotic harvester because it would result in crop loss, which human pickers do not commonly incur by dropping fruit. On the original kiwifruit harvester, fruit dropping was observed to occur when:

1. The hand mechanism hit either the target or adjacent fruit during the up movement of the robot hand or during the rotation in the picking action of the hand (Figure 11).
2. The fruit was not fully enclosed in the hand as the hand closed and hence the fruit could be flung during the rotation in the hand's picking action or the fruit could miss the fruit chute when the hand released the fruit.
3. The fruit was pulled with such force that the canopy shook and some fruit dropped as a result.

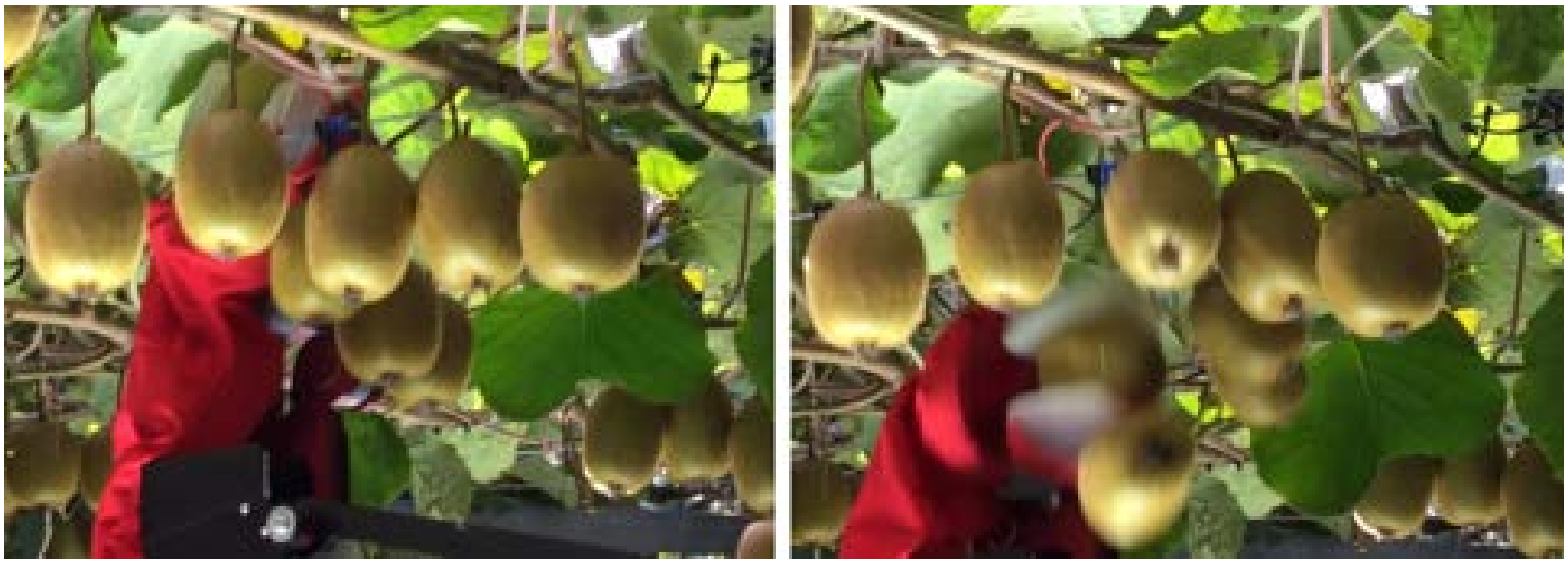

*Figure 11: A sequence showing an arm about to pick a fruit (left); however, it comes into contact with adjacent fruit (right), causing enough contact to dislodge the adjacent fruit.*

The issue of a fruit not being fully enclosed in the hand mechanism before initiating the picking action might be addressed by improvements to the fruit pose measurement system. In particular, it might be useful to have sensors closer to the fruit or sensors that can confirm that the hand is in a good position before initiating the picking action.

The issue of pulling fruit with such force that the whole canopy shakes and dislodges other fruit may be addressed by improving the picking action of the hand mechanism. It seems that a completely new detachment mechanism could be designed to improve the picking action. The issue of the hand mechanism hitting fruit on the original kiwifruit harvester could possibly be mitigated by:

- Improving the accuracy of the kiwifruit detection system. This could include software or hardware improvements. A lot of effort was put in throughout the project to improve the software systems with a progression of more accurate but computationally expensive Convolutional Neural Networks; however, despite this there was not an obvious corresponding improvement in the overall system. Less research was put into improving the computer vision hardware; in particular, using one pair of cameras below and in front of each robot arm was the configuration used throughout the harvesting testing. Other options, including more cameras per arm, cameras closer to the canopy and sensors on the arm, were not tested on the original kiwifruit harvester.
- Reducing the speed of the hand mechanism as it approaches the fruit. This measure was used throughout the project but could be improved so that the arm moves faster in low collision risk spaces and moves even slower when closer to the fruit.
- Using a smaller or softer hand mechanism. A smaller hand mechanism should reduce the chance of hitting adjacent fruit and a soft hand should mean that if fruit is hit, the impact should be lower. The use of soft hands was experimented with throughout the project; however, the hand mechanisms used were much larger than the fruit and without the ability to change the size of the opening, depending on the size of the fruit.
- Using a different approach trajectory for the hand mechanism. It seems that there might be optimal approach trajectories for entering a cluster of fruit in order to minimise collisions. It might be best for the hand to be oriented so that, when the hand picks a fruit, the hand moves directly away from the cluster, instead of towards other fruit, as shown in Figure 11. However, with the limited degrees of freedom of the arms that were used on the original kiwifruit harvester, optimising or selecting better trajectories in this way was not possible.

The successful pick rate of 56 percent was well below the project goal of 80 percent. However, this figure of 56 percent from the third and last planned harvesting season of the project was similar to the successful pick rates of the first and second harvesting seasons. For example, in the first season over 51 percent of fruit were successfully picked. There were a number of factors that were barely changed between the seasons, and, where changes were made, it seemed that they should have improved the system. The factors that had not changed much over the project harvesting seasons, included:

- The computer vision hardware. Throughout all three seasons one pair of cameras was used per picking arm, with the resolution and working area kept approximately the same.
- The computer vision software pipeline. The procedure for fruit localisation was:
    - Take an image of the canopy for each camera.
    - Detect the positions of the fruit in the images.
    - Perform stereo matching to determine the position of fruit.
    - Select fruit to pick and the order for picking- for more details see [58].
    - Command the arm to pick the selected fruit.
    - Wait for the arm to pick the fruit and then repeat.
- The design of the robot arms and detachment mechanisms. There were minor changes made to the dimensions and details of the arms; however, the number of degrees of freedom of the arms was kept the same.

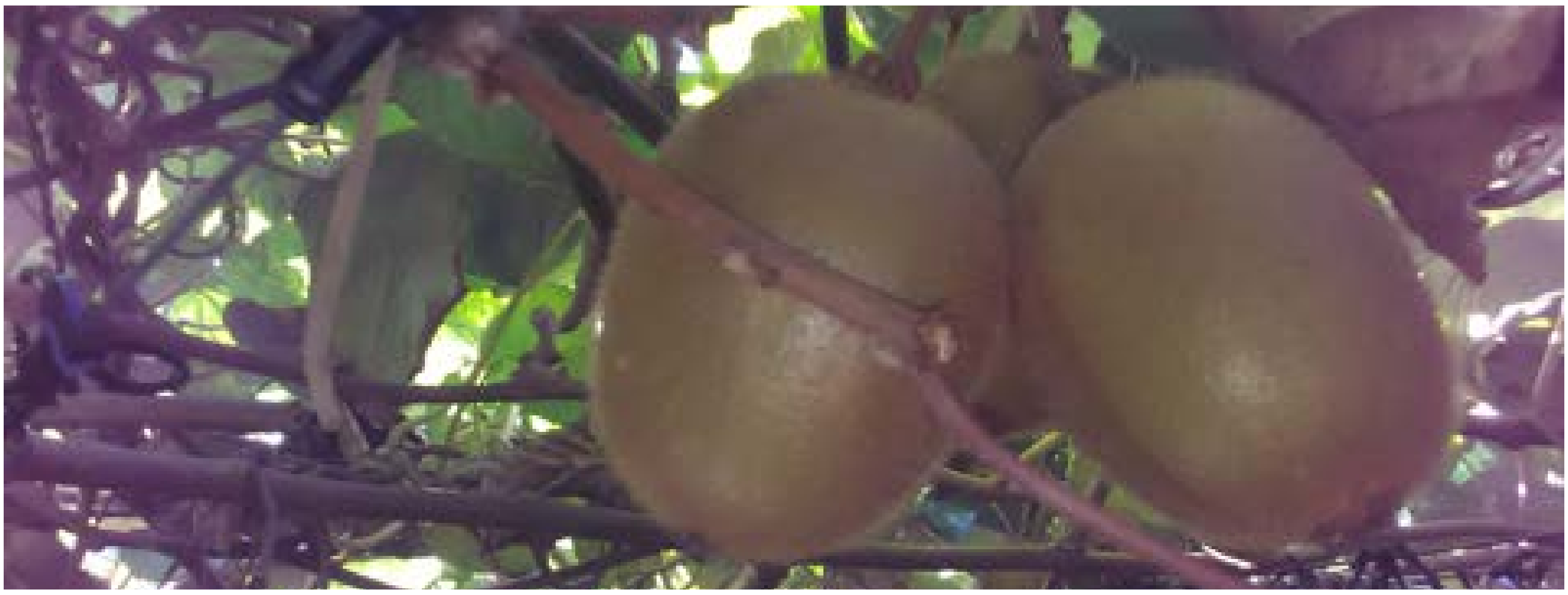
*Figure 12: Showing how in clusters of kiwifruit, the fruit may lean on each other.*

Since these factors were mostly unchanged and the end results were also unchanged, it seems that perhaps one or more of these factors might be important areas for improvement. Suggested improvements include:

- Changing the computer vision pipeline. The existing pipeline processed just one pair of stereo images for the picking of multiple fruit. This assumes that when a fruit is picked, other fruit do not move as a result; however, this is known to be wrong because the fruit grow in clusters and hence may lean on each other (Figure 12) in such a way that the fruit left behind may move when one is removed. Also, the act of picking the fruit can pull the connecting vine into a new position. In addition, the canopy moves and rises during picking because of the reduction of the weight of the canopy due to the removal of fruit. Therefore, various factors contribute to fruit movement during picking, which undermines the underlying assumption of the original kiwifruit harvester pipeline. Suggested improvements to this pipeline include processing one or more stereo-pairs per fruit picked, visual servoing to the fruit or using a hybrid of visual servoing and the existing pipeline; although, visual servoing would require faster image processing in order to not degrade the harvester speed.
- Changing the computer vision hardware. The existing stereo-pairs on the original kiwifruit harvester were designed to have a working distance of 0.9 of a metre. However, for a given pair of correctly matched features, the accuracy of the stereovision system was better at closer distances and hence having cameras closer to the canopy may have given better fruit pose measurement accuracy. In addition, only having one pair of cameras per arm meant that any errors in feature matching were not corrected, which would not have to be the case if more cameras were used per arm. Furthermore, there were five axes of rotation between the cameras and the end effectors on the original kiwifruit harvester and so there were some systematic calibration errors as well as unmeasured arm positioning errors. The effects of such errors might have been reduced using sensors on the arms and closer to the end effectors.
- Testing different designs for the robot arm and end effector. The robot arms used on the original kiwifruit harvester were based on the design by Scarfe [21]. However, this design was not able to pick fruit that was blocked by a branch, wire or beam from below because it was constrained to approaching from below and moving up towards the fruit. In addition, the end effector was much larger than the fruit and was not able to rotate independently about a vertical axis; these factors limited the ability of the end effector to pick fruit that was close to solid objects.

Changing the robot arm and end effector was estimated to be a significant undertaking compared to the other proposed changes, especially when considering the associated software changes required when changing hardware. Hence, it was decided to first understand how much of an issue was caused by obstructed fruit and fruit adjacent to solid objects. In order to do this, a survey was conducted of an area of canopy, which had been harvested using the original kiwifruit harvester. Each fruit left behind was subjectively categorised as given in Table 9. From Table 9 it seems that 67 percent of fruit left behind are obstructed from being picked by the original kiwifruit harvester arms. According to Table 8, 35 percent of fruit are left in the canopy after harvesting. If 67 percent of this 35 percent of fruit left behind are obstructed for the original kiwifruit harvester, that is approximately 23 percent of fruit that cannot be picked by the original kiwifruit harvester in the areas of canopy used for testing. It was deemed that this is a significant proportion and enough to justify trialling a different robot arm and end effector; especially since without this 23 percent, it would be impossible to get to the goal of 80 percent of fruit harvested, for the original kiwifruit harvester and the orchards that were used for testing.

*Table 9: Analysis of fruit left behind after picking with the original kiwifruit harvester, with categorisation performed according to why it seemed the fruit had not been picked.*

| **Apparent Issue Preventing Picking** | **Percentage of Cases** |
|---|---|
| Fruit obstructed by wire | 19 |
| Fruit obstructed by branch | 33 |
| Fruit obstructed by branch and wire | 8 |
| Fruit obstructed by beam | 7 |
| Fruit occluded by leaves | 3 |
| Unexplained; fruit appears pickable by the original kiwifruit harvester | 30 |

## 2.5 Second Kiwifruit Harvester Mechanical Design

The hardware for the second kiwifruit harvester was designed to reuse parts that were already available. A key aspect of the second kiwifruit harvester was the robot arm that was used. Due to the lack of availability of anything larger, an arm from a Kuka YouBot [88] was used. This arm has five degrees of freedom and also has a gripper mechanism (Figure 13). The additional degrees of freedom with this arm allowed for improved manoeuvrability of the detachment mechanism; although, with trade-offs of equipment cost and path planning complexity.

The kiwifruit detachment mechanism was a custom designed and built component. Typical kiwifruit dimensions were key to the detailed design of the kiwifruit grasping end effector. The dimensions used were taken from previous work [89] and are presented in Table 10.

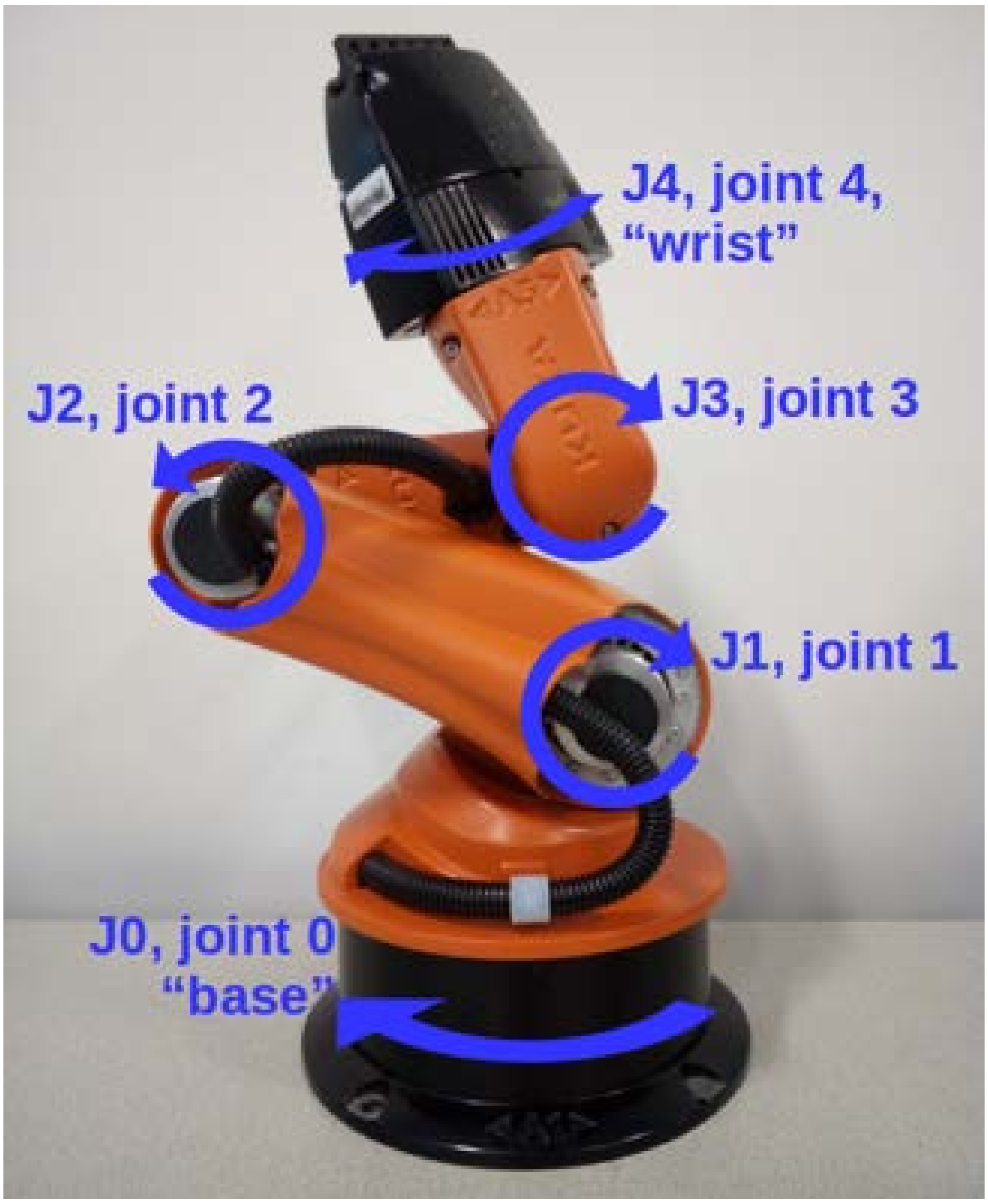

*Figure 13: The Kuka YouBot arm and joint terminology used here.*

*Table 10: Kiwifruit dimensions used for the kiwifruit detachment end effector design on the second kiwifruit harvester.*

| | **Length (mm)** | **Maximum Width (mm)** | **Minimum Width (mm)** |
|---|---|---|---|
| **Upper Bound** | 78.4 | 63.5 | 56.9 |
| **Lower Bound** | 54.8 | 40.8 | 38.0 |

### 2.5.1 Fruit Detachment Experiments

The original kiwifruit harvester arms were observed to sometimes pull the canopy with significant force in such a way so as to cause fruit to be shaken to the point of detaching and falling. In addition, those existing mechanisms caused contact with adjacent fruit with force during the fruit detachment action. Hence, in order to design a kiwifruit detachment end effector, it was decided that it was necessary to firstly propose new methods for fruit detachment and then select some of these. In order to assess different kiwifruit detachment methods, criteria were set for the assessment. These metrics are given in Table 11.

*Table 11: Criteria used for assessing kiwifruit detachment methods.*

| Description | Reason | Value | Importance |
|---|---|---|---|
| Damage to fruit must be minimised | Damaged fruit is wasted and the rotting of damaged fruit can affect adjacent fruit | Any damaged fruit is a failure | Very important |
| Stems must be removed from kiwifruit | Stems are known to puncture fruit in bins | Every stem not removed is a failure | Very important |
| The detachment method should work in tighter spaces | For many fruit, there is only a small space for accessing the fruit | This criteria may be assessed by the volume of space required to perform the detachment method | Important |
| The force applied may be minimised | The result of applying large forces may be undetected damage to fruit | This criteria may be assessed by the effort required to perform a detachment action | Somewhat important |
| The detachment method must be reasonably fast | The robotic detachment must be economically viable | There is a trade-off between speed and the cost of the equipment- so a slower method requiring cheaper hardware may be acceptable | Somewhat important |

The detachments experiments were conducted by first defining a trajectory of forces for each experiment (Table 12). Then these detachment actions were repeatedly performed by hand 30 times and observations were made from these tests (Table 12). Based on the observations from the experiments, it was hypothesized that the stem pushing, stem striking and fruit rotating methods would best satisfy the criteria of Table 11. Stem striking, where the stem is impacted with a relatively fast moving object, is similar to the method described by Ting et al. [62]. Fruit rotating, where the fruit is rotated upwards about the point where the stem and fruit meet, is similar to the method of Scarfe [21], although with more emphasis on rotating the fruit and less emphasis on pulling the fruit with high force. Stem pushing, where the fruit is held and the stem is pushed so that it rotates about the fruit, may be seen as a hybrid of stem striking and fruit rotating. Stem pushing is like stem striking because in both cases the fruit is stationary and the stem is contacted. Stem pushing is like fruit rotating because in both cases there is rotation about an axis through the connecting point of the stem and the fruit; although, with stem pushing the stem rotates, whereas with fruit rotating the fruit rotates.

*Table 12: The kiwifruit detachment methods tested and the results observed.*

| **Method name** | **Descriptions of the trajectory of forces applied to a kiwifruit for the method** | **Images showing the method** | **Observations of the experiments** |
|---|---|---|---|
| Stem pushing | The kiwifruit was gripped and held with enough force to keep the kiwifruit static, while the stem was pushed horizontally at a point 0.01 m above the top of the fruit with enough force so that the stem moved. | 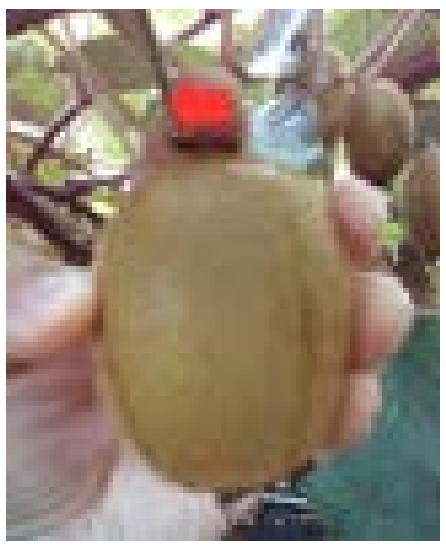 | The fruit detached cleanly with little force. |
| Fruit twisting | The kiwifruit was gripped and a torque was applied to the kiwifruit at the points of gripping contact with the axis of rotation being the axis of the stem. | 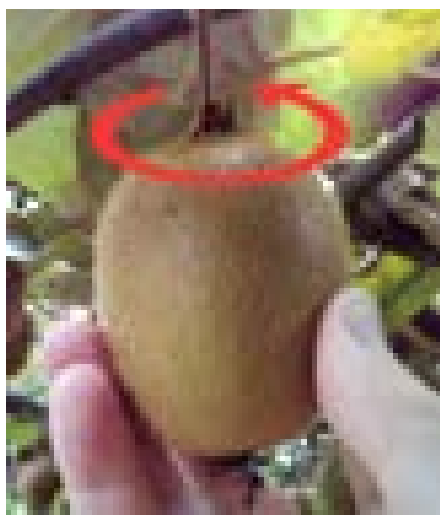 | Large parts of the stem remained after detachment: 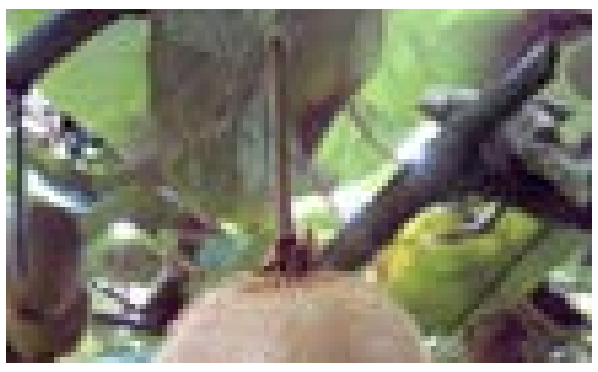 |
| Stem striking | The stem was impacted by a horizontally moving object with approximately 0.1 kg mass moving at a speed of approximately 3 $ms^{-1}$ at a point approximately 0.02 m above the fruit. This method is taken from Ting et al. [62]. | 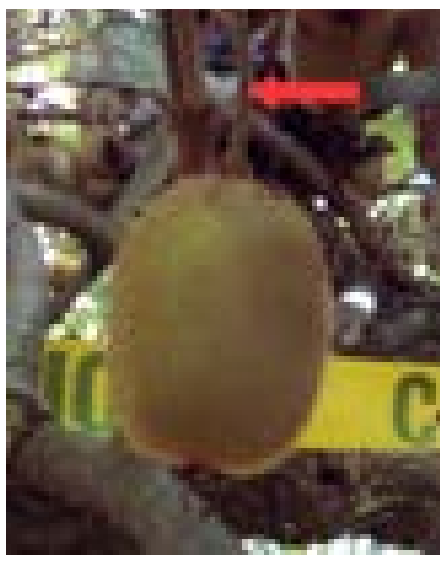 | The fruit detached with no remaining stem; however, this method was not successful in some attempts. |
| Fruit rotating | The kiwifruit was gripped and a torque was applied centred about a horizontal axis, located at the base of the stem/ top of the fruit, with the direction of torque causing the fruit to rotate upwards. After a rotation of more than 90 degrees, the fruit was pulled towards the ground. | 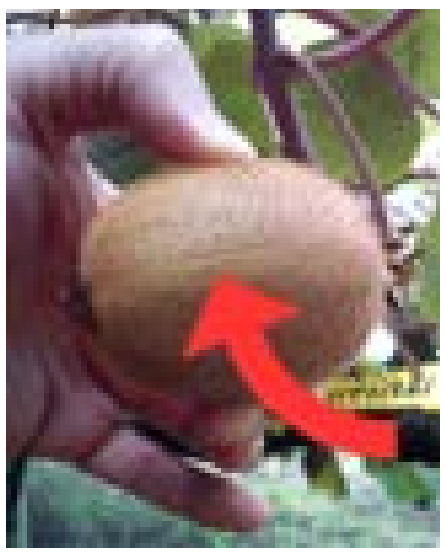 | The fruit detached cleanly with little force. |
| Fruit pulling | The kiwifruit was gripped and pulled towards the ground until the fruit detached. | 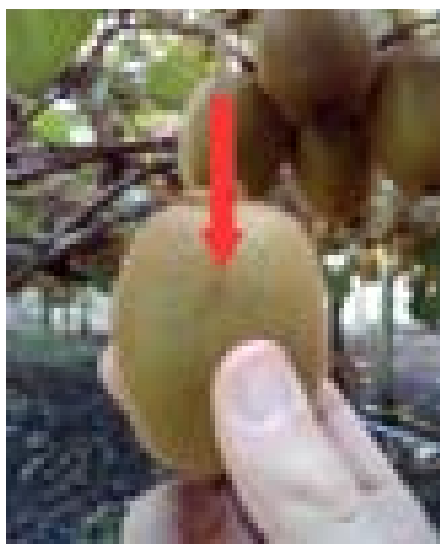 | This method required the most force out of all of the methods. The canopy shook at detachment to the point where other fruit fell to the ground. The stem remained attached for some fruit. |
| Stem cutting | The stem was cut by side-cutters with the side-cutters resting on the top of the fruit. | 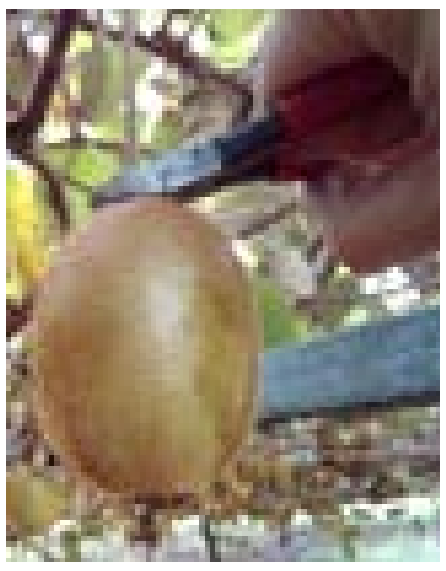 | Short parts of the stem remained after detachment: 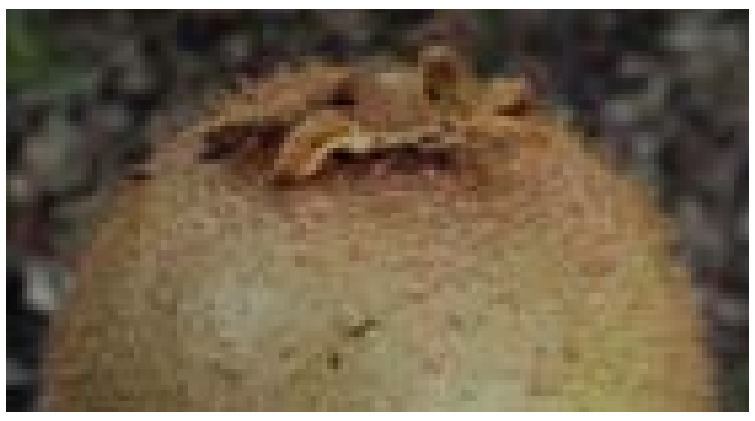 |

### 2.5.2 Stem Pushing Mechanism Design

One option considered for the stem pushing mechanism was placing an off the shelf linear actuator adjacent to the fruit. However, the linear actuators found occupied a relatively large volume of space adjacent to the fruit and hence alternatives were sought, since more space occupied suggests less capability to fit into tight spaces in the canopy.

Figure 14 shows a conceptual design of a stem pusher mechanism, where the motor has been positioned between the fingers of the Kuka YouBot gripper. A feature of this design is the use of a sliding pivot to keep the stem pushing part above the fruit being picked, as opposed to possibly contacting adjacent fruit if the stem pushing part was rigidly mounted to the shaft of the gearbox. The resulting motion of the stem pushing part with the sliding pivot is shown in Figure 15. An issue with this design is that the fruit may get in the way of the stem pushing part when the gripper is being moved into position; especially if the gripper is being moved vertically upwards into position with the gripping paddles around the fruit.

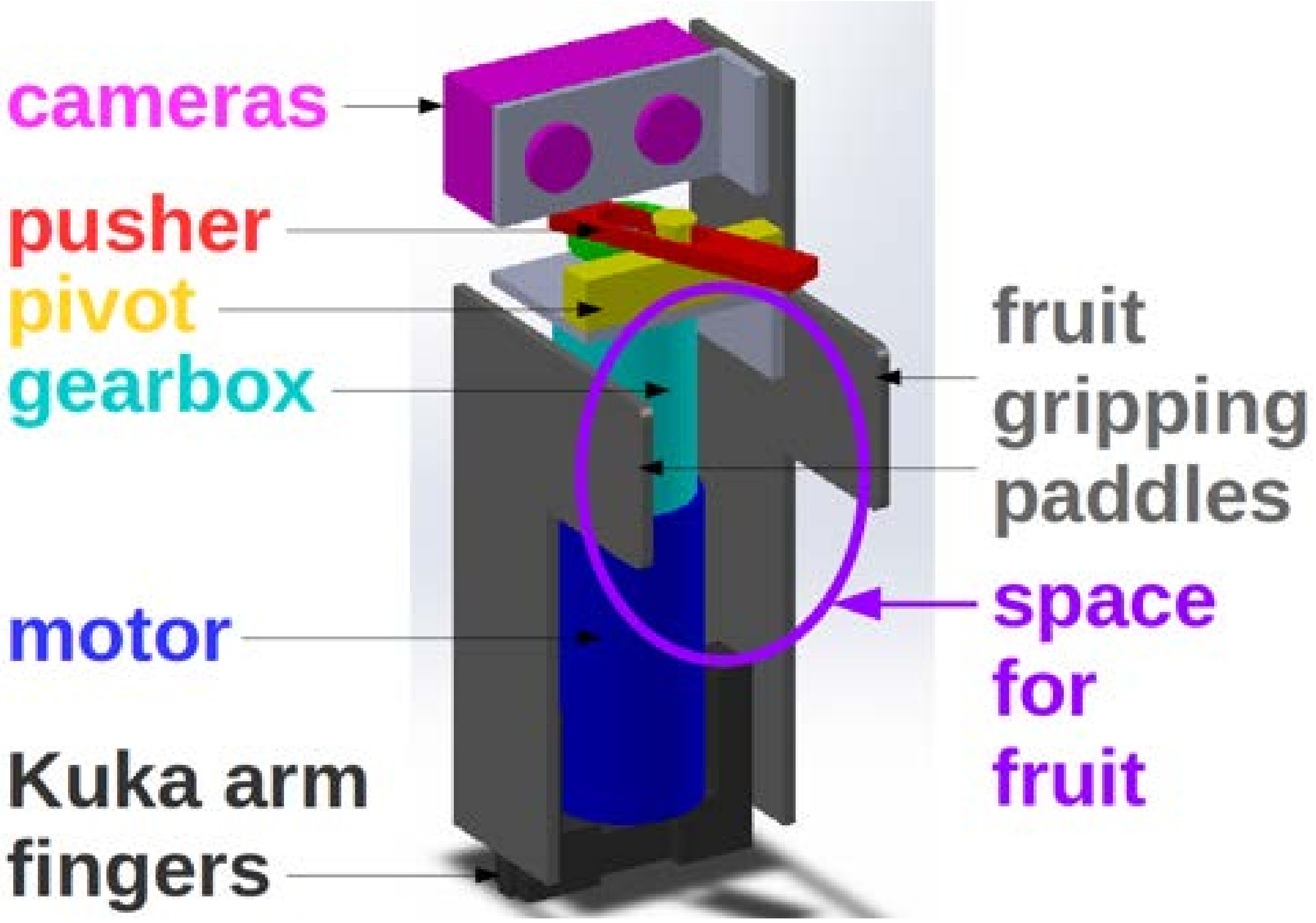

*Figure 14: Conceptual design for a stem pushing fruit detachment mechanism, where the motor fits between the Kuka fingers and the pusher moves about a pivot point on a horizontal plane.*

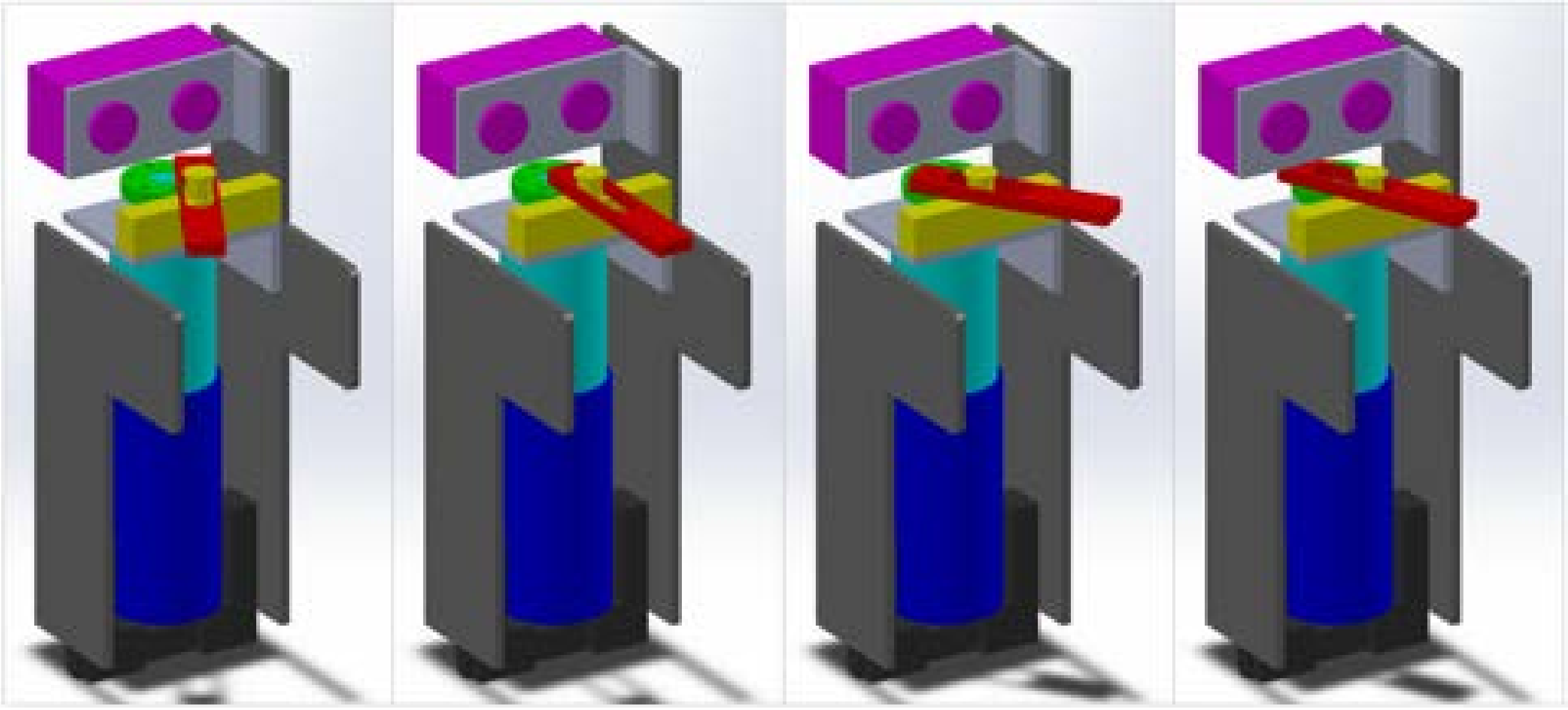

*Figure 15: Demonstrating the movement of the pivoting stem pusher on a horizontal plane.*

There was a trade off in the motor and gripper paddles placements for the pivoting stem pusher. If the motor and gripper paddles were placed lower, the motor would be moved behind the last link of the robot arm and it would increase the volume of the mechanism adjacent to the fruit- thereby decreasing the options for spaces that the fruit detachment mechanism could fit in. However, if the motor was placed between the fingers of the YouBot arm with the gripper paddles higher (Figure 14), the length of the moment arm created about the fingers would be increased, increasing the load on the fingers and reducing the stiffness in the gripper paddles. It was thought that a future fruit detachment mechanism design would not be so limited in terms of the arm hardware, as was the case in the design described here; therefore, factors such as stiffness and the load on the fingers could be compensated for by selecting a more robust gripper mechanism in future work. Hence, it was decided that it might be acceptable to firstly experiment with the motors between the fingers, in order to minimise the volume of the detachment mechanism directly adjacent to the fruit, which was the case for the concept shown in Figure 14.

Figure 16 and Figure 17 show alternative concepts for the stem pusher mechanism, where the stem pusher moves in the direction of the fruit gripping paddles, as shown in Figure 18. As shown in these images, there are different possibilities for the length of the pusher and the placement of the axis of rotation, resulting in different swept arcs for the pushing surface. For these concepts, the pusher moving in the direction of the fruit gripping paddles means that the pusher tends to push the fruit out of the paddles. This could result in the fruit being dislodged or might require the gripper to squeeze the fruit with more force, which could cause fruit damage.

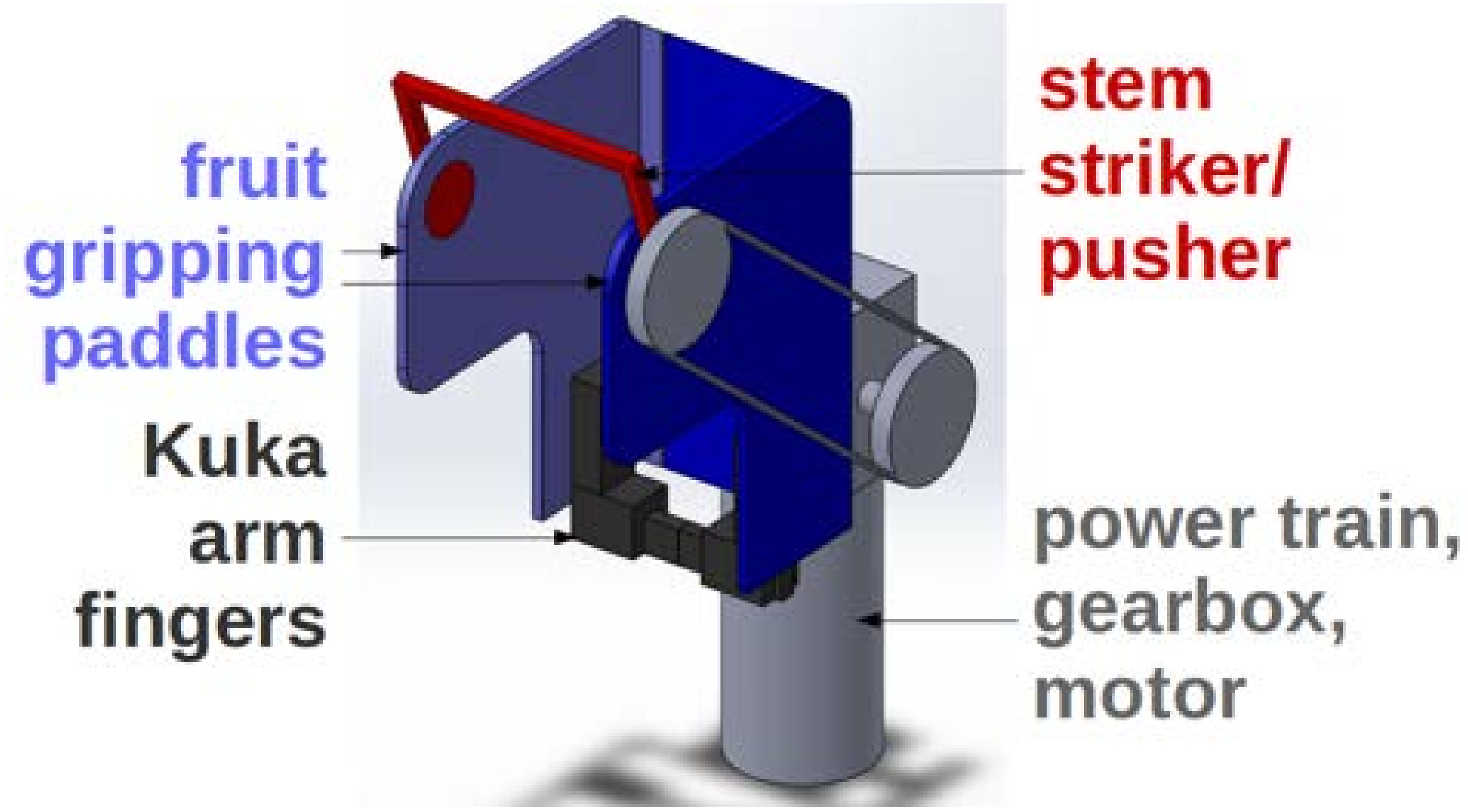

*Figure 16: A stem pushing concept where a shorter pusher rotates towards the fruit and the stem pusher telescopes when the Kuka fingers open.*

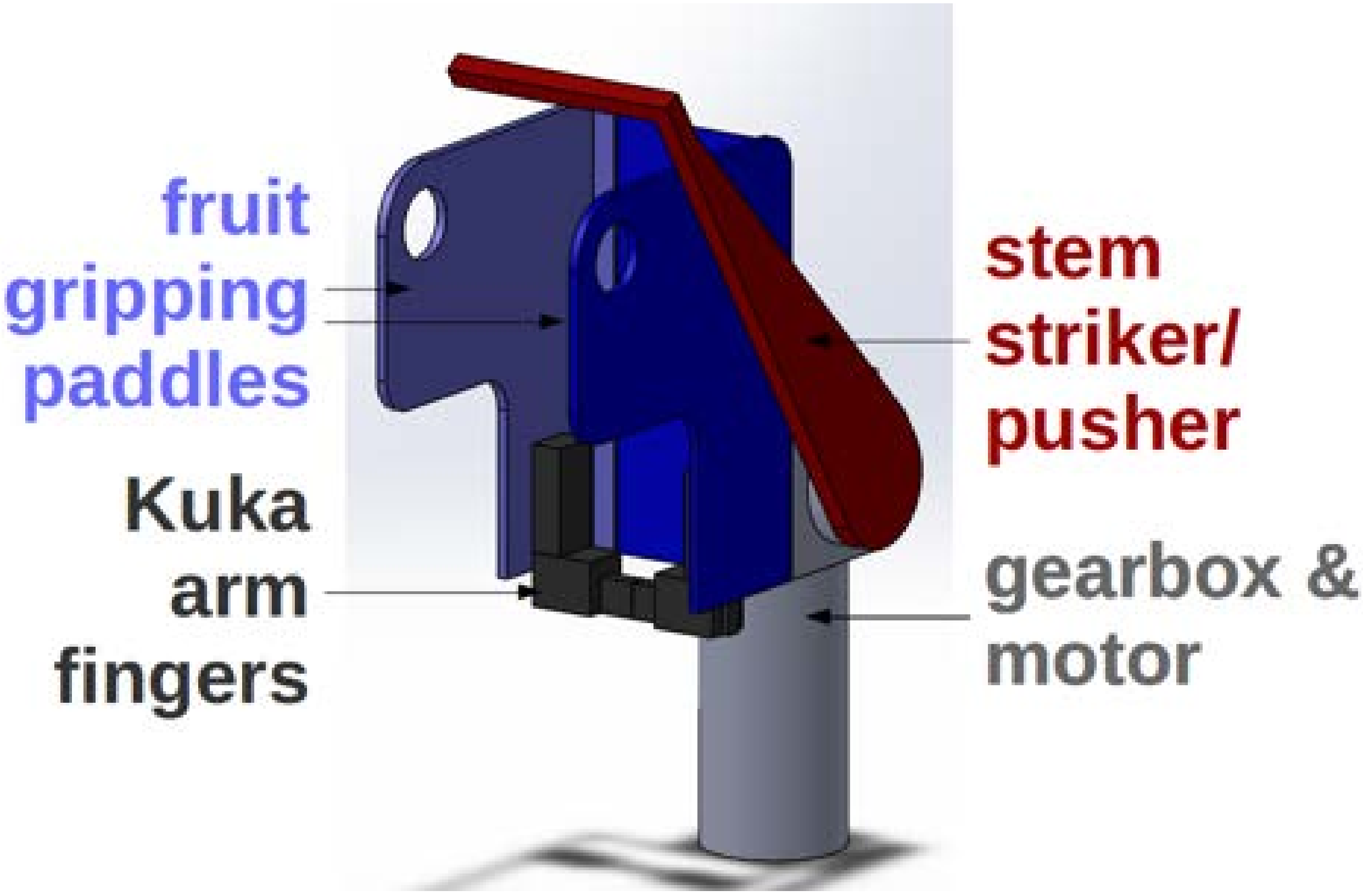

*Figure 17: A stem pushing concept where a longer pusher rotates towards the fruit.*

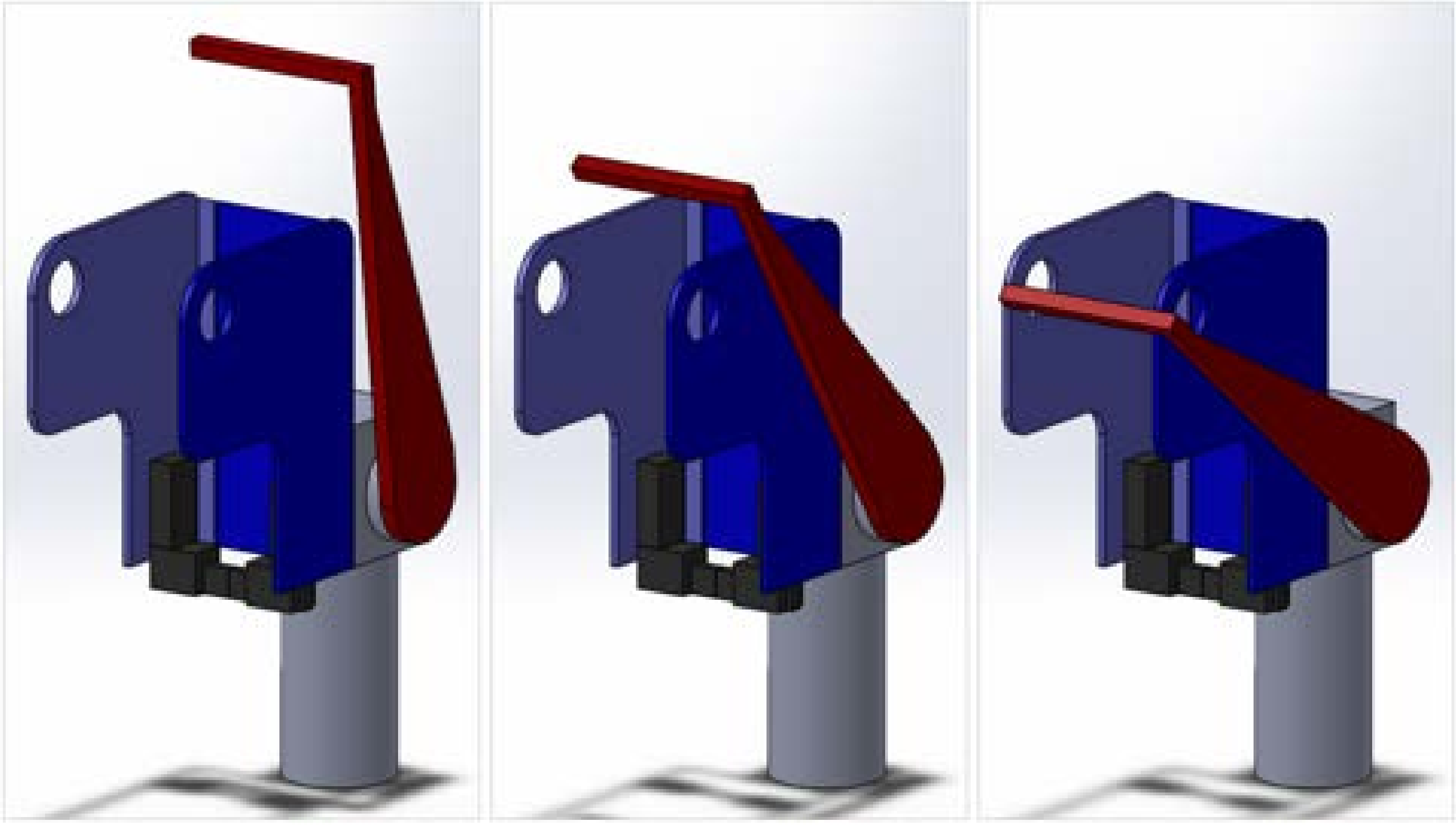

*Figure 18: Illustration of the movement of the stem pushing concept from Figure 17.*

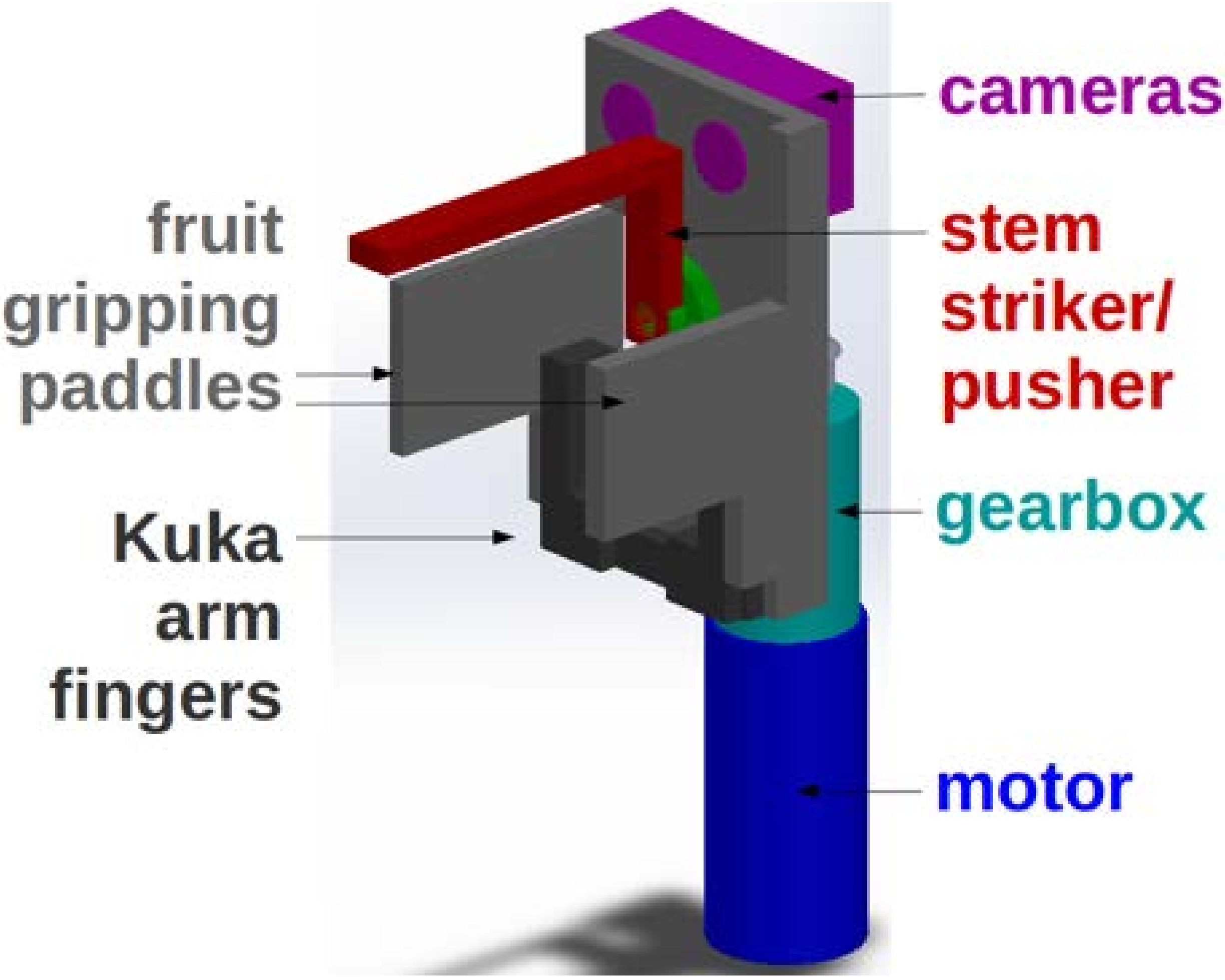

*Figure 19: A stem pushing concept where the pusher rotates from one gripping paddle to the other.*

Figure 19 shows a stem pushing kiwifruit detachment concept where the pusher rotates from one fruit gripping surface to the other. The movement of the stem pusher is shown in Figure 20. This concept avoids the tendency to push the kiwifruit out of the gripper, unlike the concepts with the pushers moving in the direction of the gripping paddles. In addition, with the pusher rotating from one gripping paddle to the other, this concept avoids the issues of the horizontal plane stem pushers, like that shown in Figure 14. As can be seen in Figure 20, the stem pusher can be moved into a position alongside a fruit gripping paddle, so that the pusher does not unduly hit the fruit during the arm movement into the picking pose. In addition, the stem pusher of this concept does not enter the space of adjacent fruit. Because of these advantages, it was decided to proceed with a stem pushing concept that rotates the pusher from one gripping paddle to the other.

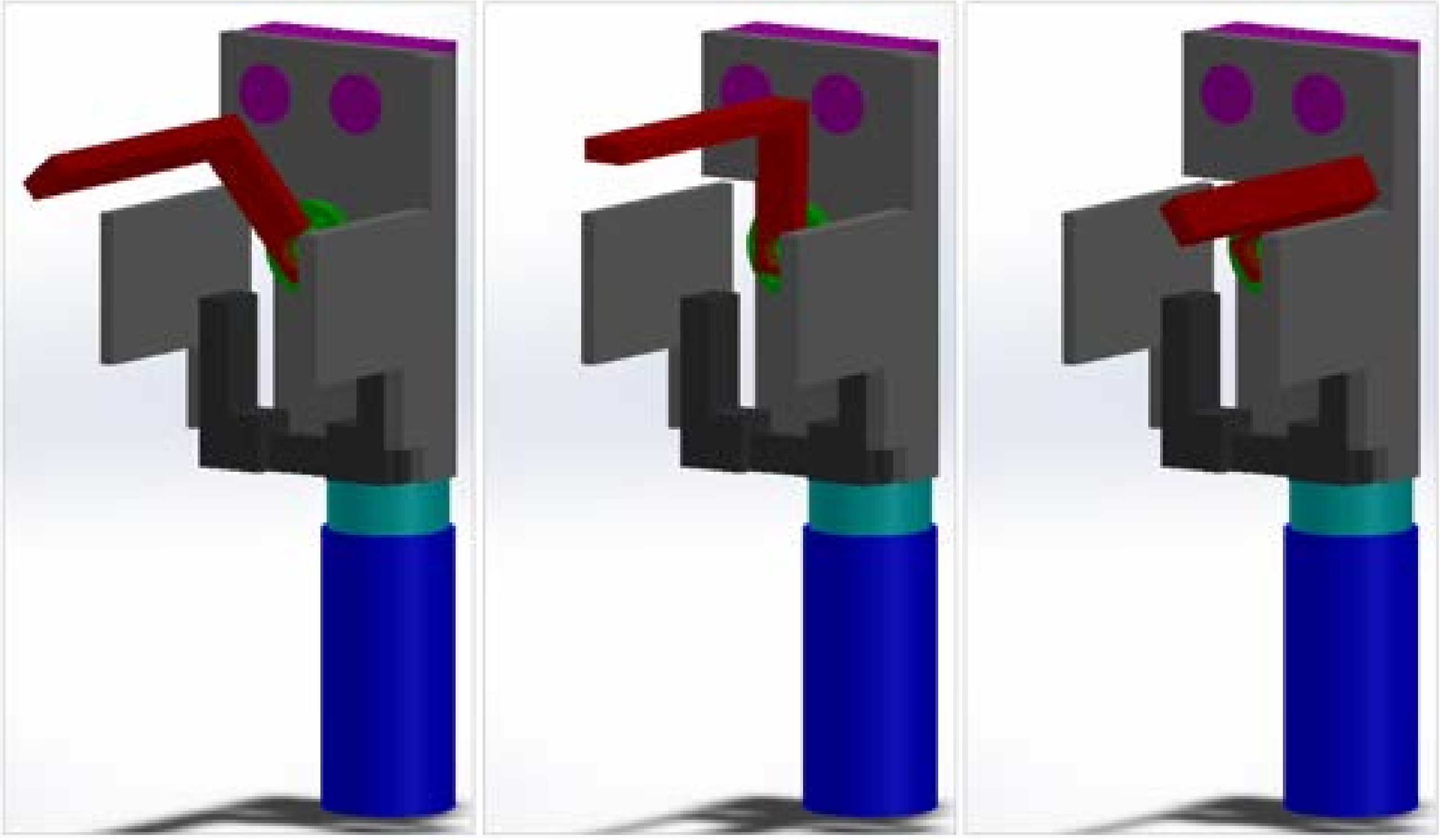

*Figure 20: Showing the movement of a stem pusher that rotates from one kiwifruit gripping paddle to the other.*

### 2.5.3 Stem Striking Mechanism Design

It was thought that the stem striking mechanism design could be very similar to the stem pushing design; although, it seemed stem striking might require a higher speed actuator, whereas, stem pushing implies a higher torque or higher force actuator. It seemed that stem striking may also allow for a ballistic mechanism and that gripping the fruit would be optional.

Ting [90] experimented with mechanisms that contacted and sheared the stems of kiwifruit. However, it was found that an important issue with this approach was the potential for fruit damage. The mechanism may strike fruit, causing direct damage to the struck fruit, or the mechanism may also strike adjacent canes or stems, causing indirect damage after the adjacent fruit falls.

It was thought that if stem striking was to be used, the risk of fruit damage may be reduced by detecting the fruit and stems with higher accuracy. However, it seemed that, even so, a mechanism that occupied and swept less space would be less likely to damage adjacent fruit. Mechanisms considered included linear actuators and motors connected to a piston; however, it was found that designs including such mechanisms occupied a relatively large amount of space adjacent to the fruit.

Other options considered included using a solenoid for impacting the stem. However, the stroke size of many solenoids is small compared to the overall size of the solenoid. For example, for a Pontiac Coil F0491A, the stroke length is 0.032 m and the length of the solenoid closed is 0.083 m [91].

Instead, it was decided to use the stem pushing mechanism of Figure 19 and adapt the motor or gearing to produce higher speeds at the stem contact point. In contrast, stem pushing would use a lower speed, higher torque version of the same mechanism.

### 2.5.4 Fruit Rotating Mechanism Design

Initially, it was thought that fruit could be gripped between two contact surfaces, which would be rotated upwards about an axis through the point where the stem meets the fruit. A conceptual design for such a mechanism is shown in Figure 21. In designing the mechanism to perform this action, the issues identified with this concept included:

- The volume swept by the mechanism and the fruit was relatively large (Figure 22); this volume was larger than the volume occupied by the stem striking and stem pushing design concepts. With fruit growing in clusters, this larger volume would increase the potential for contact with adjacent fruit.
- With the contact surfaces squeezing and rotating the fruit, there was technical difficulty in minimising the friction in the mechanism while also trying to keep the mechanism as small as possible.

Due to these issues, resource constraints and the potential benefits of stem striking and stem pushing designs, it was questioned whether there was value in developing the fruit rotating mechanism concepts.

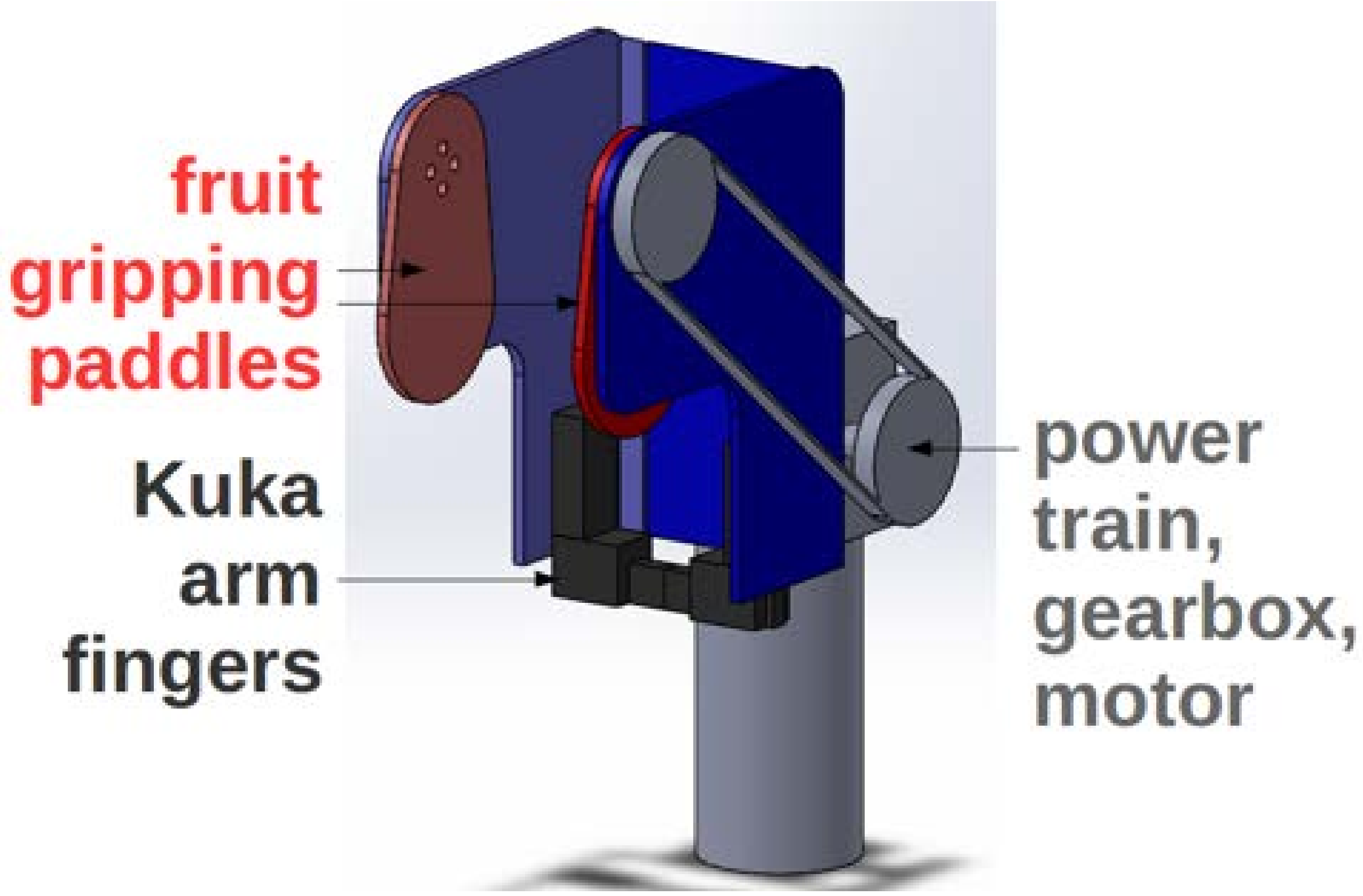

*Figure 21: Conceptual design for a kiwifruit upwards rotating mechanism design, where the kiwifruit would be squeezed between the two fruit gripping paddles.*

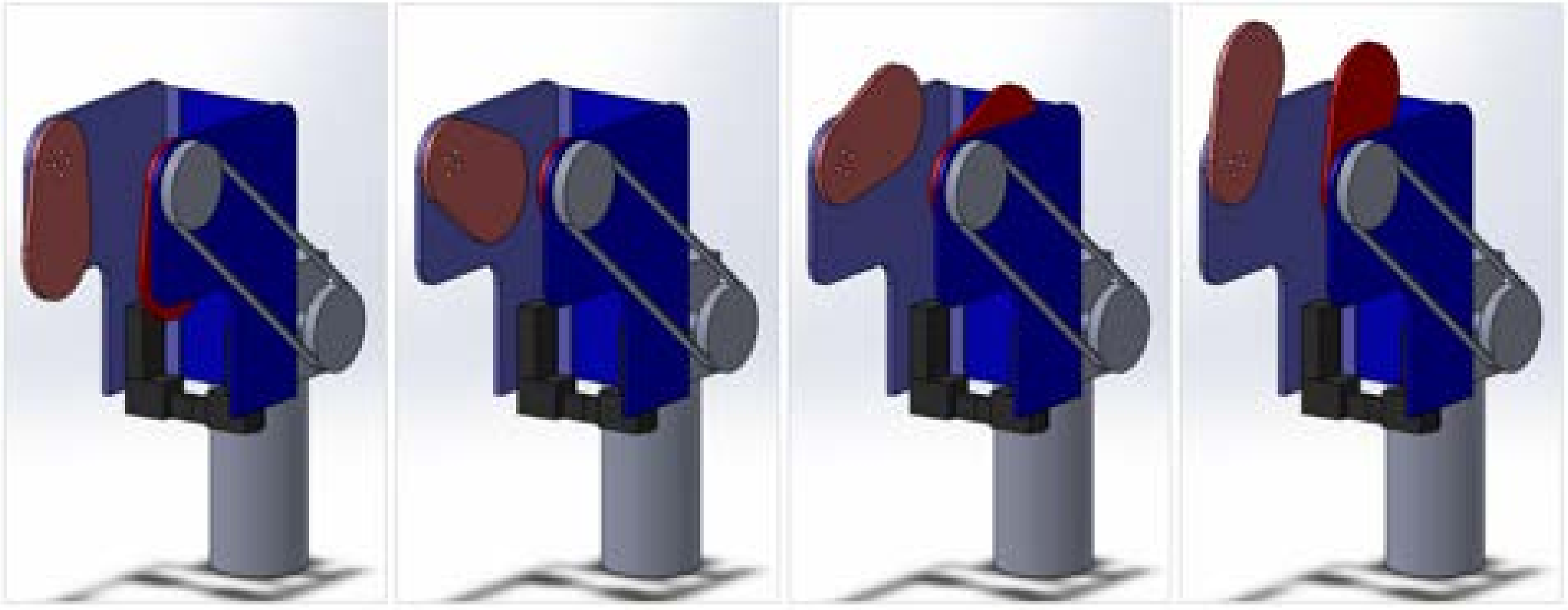

*Figure 22: An illustration of a kiwifruit upwards rotating mechanism movement, showing the space swept by the fruit gripping paddles.*

### 2.5.5 Consolidating Fruit Detachment Mechanism Designs

When considering why a kiwifruit detaches when it is rotated upwards about the stem attachment point, it was hypothesized that the important factor is the *relative rotation* between the stem and kiwifruit. It was thought that this *relative rotation* might create a shear force at the stem attachment point, which is sufficient to cause detachment. It was postulated that if it is indeed the *relative* *rotation* about the stem attachment point that causes the detachment, then it follows that it should be possible to rotate the stem (as opposed to the kiwifruit) about the stem attachment point, in order to achieve detachment. In order to rotate the stem about the stem attachment point, it was thought that the fruit could be held stationary and the stem could be pushed along an arc that followed the surface of the kiwifruit flesh. It seemed that the stem pushing concept shown in Figure 19 and Figure 20 would produce this arc for the stem rotation and hence it was hypothesized that this stem pushing mechanism would produce the same result as a fruit rotation mechanism by successfully detaching kiwifruit. Hence, it was decided to not manufacture a fruit rotation mechanism and to proceed with the stem pushing concept of Figure 19 instead.

Even with the decision to only manufacture the stem pushing concept of Figure 19, there were still remaining questions about the detailed design of this concept. These questions included:

- How much force was required at the stem pushing contact point?
- What stem pusher speed was required in order to achieve kiwifruit detachment?

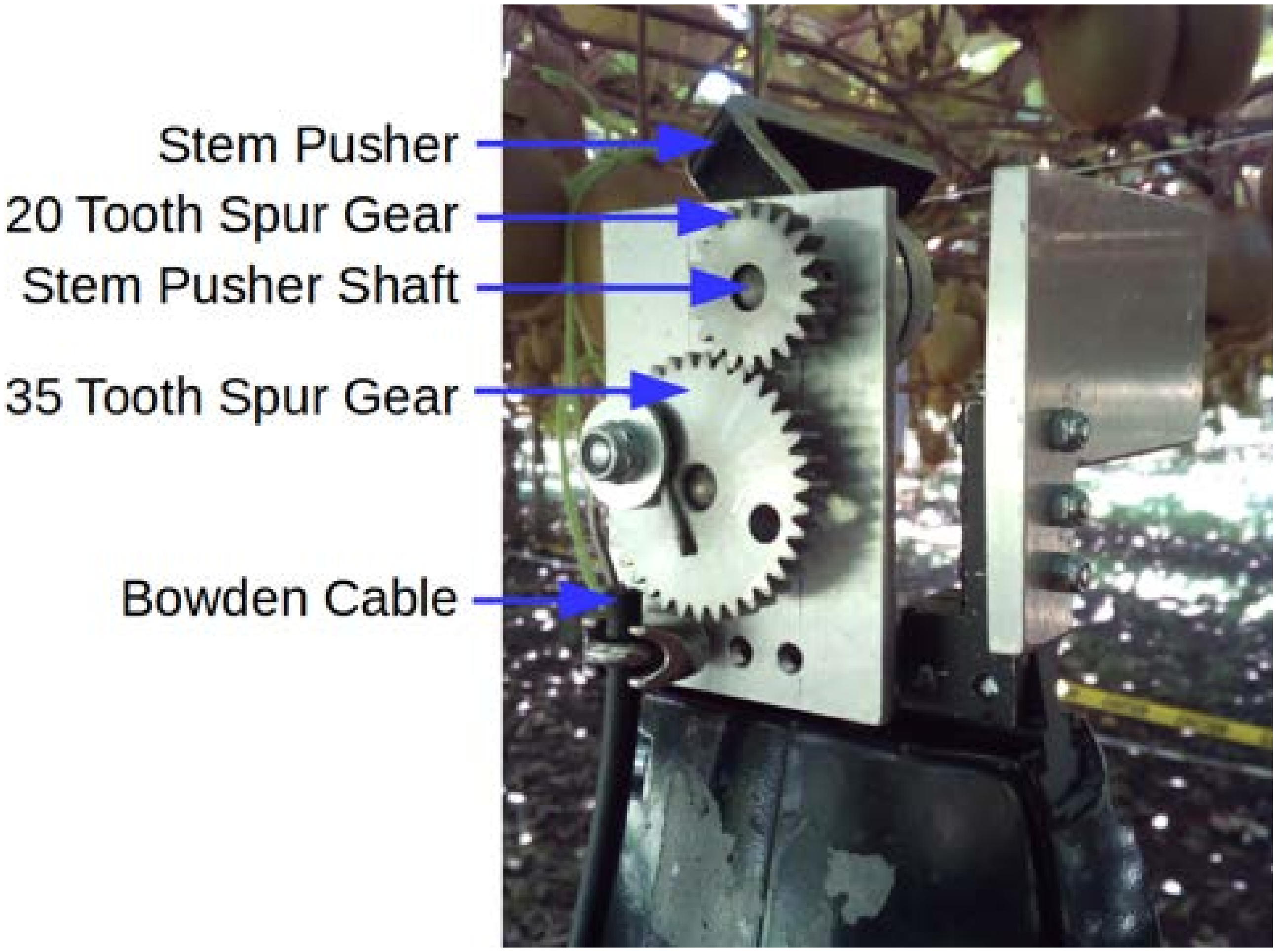

*Figure 23: The stem pusher mechanism with a 35:20 gear ratio.*

To answer these questions it was decided to design the stem pushing mechanism so that the speed and torque could be adjusted without time consuming changes to the assembly. To achieve this, it was decided to mount the stem pusher motor off the arm so that a larger motor could be used without overloading the arm; then the speed and torque could be adjusted through a wider range of values by changing the switching frequency, motor current or gearbox, depending on the motor used. The power transmission from the motor to the stem pusher was achieved with a Bowden cable and spur gears (Figure 23). For the first version of the stem pusher mechanism, the Bowden cable transferred the motion from the motor to a spur gear with 35 teeth, which turned another spur gear; this second spur gear was rigidly coupled to the stem pusher.

Two variants of the stem pusher detachment mechanism were made. A smaller variant was made with just the gripping paddles and the stem pusher parts. A larger variant was also made, which included a pair of cameras that would be able to view the stem from above the fruit. It was unclear if the cameras on the gripper mechanism would improve the fruit targeting enough to offset:

- The increase in volume of the gripper, which could affect how many kiwifruit could be reached by the gripper.
- Maintenance issues of having the cameras in the messy canopy, where lenses might get dirty or damaged.
- The added computational complexity of processing the data from the cameras.
- The cost of the extra hardware.

The effect on the ability of the detachment mechanism to reach the fruit was thought to be most important since it was estimated that even the smaller version of the stem pusher mechanism might be marginal in terms of being able to reach the goal of harvesting 80 percent of the fruit. Hence, in testing, the two stem pushing mechanism variants were compared in terms of their ability to reach and pick kiwifruit in real world kiwifruit orchard canopies (Subsection 2.7.1).

## 2.6 Second Kiwifruit Harvester Control System

A high level diagram of the pipeline of the second kiwifruit harvester control system is given in Figure 24. This entire pipeline was run for every fruit which was picked by the second kiwifruit harvester; whereas, on the original kiwifruit harvester, batches of fruit were detected and picked before restarting the pipeline.

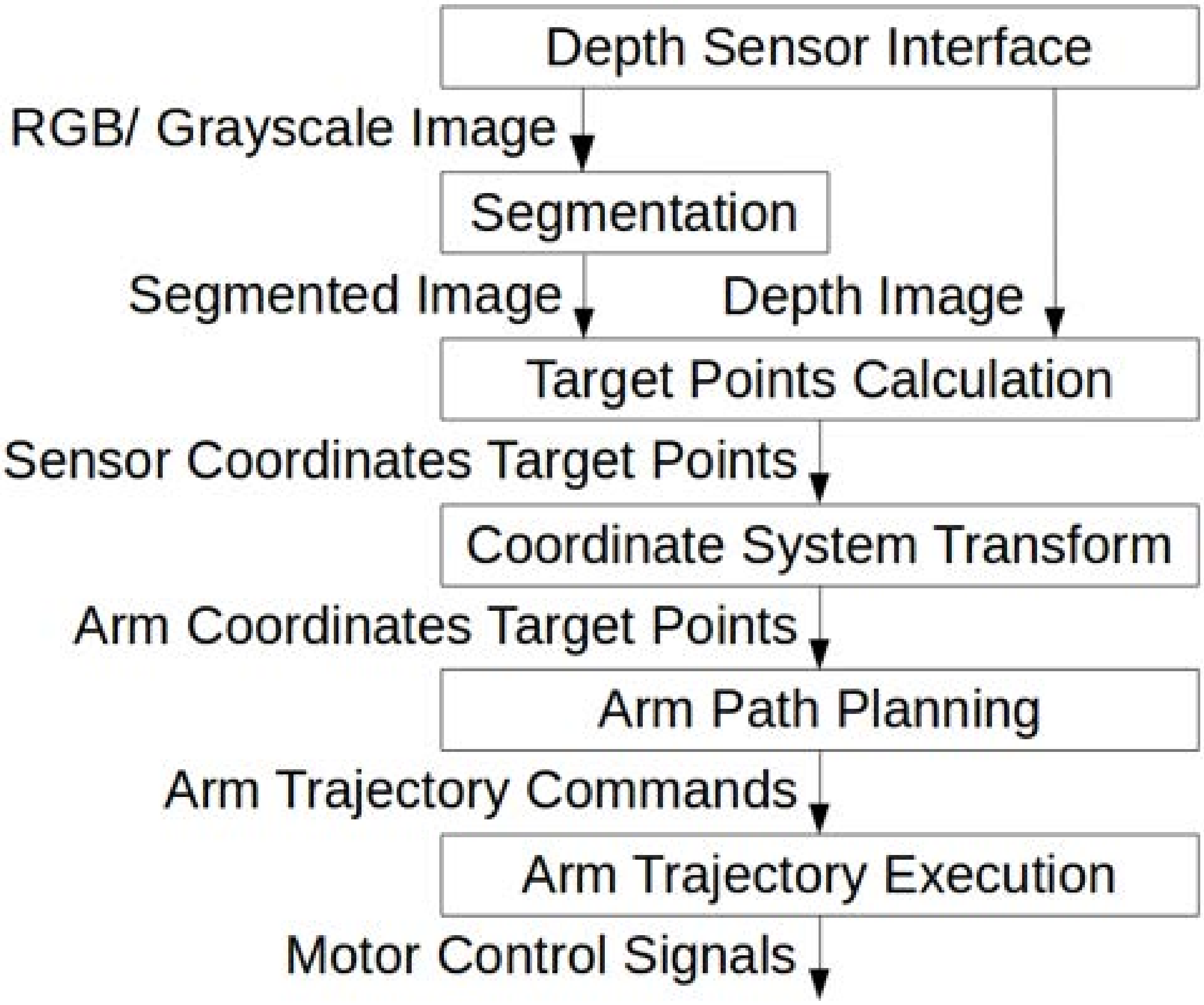

*Figure 24: High level pipeline for the second kiwifruit harvester.*

This subsection begins with a discussion of the sensor options considered for the second kiwifruit harvester. Then key parts of the pipeline from Figure 24 are described, including:

- Segmentation of the depth sensor data to detect kiwifruit.
- The target points calculation, which took the segmented depth sensor data and the raw depth data, to calculate a pose for the harvester end effector in the sensor's coordinate system.
- The transformation of the target points, calculated in the sensor's coordinate system, to the robot arm coordinate system, including how the transformation tensors were calculated.
- The arm path planning, including steps used to enable the second kiwifruit harvester to pick fruit that were occluded from below.

### 2.6.1 Sensor Selection and Mounting

This vision system development described here was performed independently of the rest of the MBIE Multipurpose Orchard Robotics Project and hence it was not constrained in terms of the hardware decisions made by other team members. This allowed for different sensors and different sensor placements to be considered. The sensors considered for the second kiwifruit harvester vision system were:

- A bespoke stereo camera system made from two Basler daA1600-60uc cameras [92]. This particular camera model was considered because of its small size and M12 mounting option, which would allow for small lenses. It was thought that a small size might allow for sensor placements that might not be possible with larger sensors. In addition, this camera model has a global shutter and hardware trigger, which are useful features for stereo cameras; in particular, these features would be important if the cameras were mounted on a moving component, such as the robot arm.
- A Basler Time of Flight TOF640-20GM-850NM [77]. This sensor outputs a single channel of intensity data and a single channel of depth data. When data was first collected from this sensor under the canopy of a kiwifruit orchard, it seemed to work well even on a sunny day and it did not require large LED lamps to illuminate the canopy for clear images of the kiwifruit, whereas the vision system of the original kiwifruit harvester used powerful LED lamps. It was also thought that using this ToF sensor would simplify the acquisition of depth data, when compared to the stereo cameras.
- An Intel Realsense D400 series depth sensor [93], which uses stereo cameras but also projects an infrared signal that enhances the stereo output. This sensor could be viewed as a hybrid sensor with the combined benefits of stereovision and an active 3D sensor. However, it was unclear if the infrared signal on this sensor would be strong enough to overcome ambient lighting and make a difference in the challenging lighting conditions of an orchard.

These sensors may all be viewed as *depth sensors* with a depth channel as well as a colour or grayscale channel in their output. In this respect, the options presented for sensor selection were quite similar and could be considered interchangeable. In contrast, the choice of sensor placement had potential system changing and performance altering effects. The sensor placement options considered for the second kiwifruit harvester included:

- Having the *depth sensor* facing vertically upwards, approximately 0.5 of a metre from the lowest kiwifruit, with the sensor underneath the working area of the arm. This configuration is essentially what was used on the original kiwifruit harvester. Since the fruit are approached from below on the original kiwifruit harvester, it may have seemed to make sense to view the fruit from below. However, issues with this sensor placement include:

  - Since the sensor is under the working area of the arm, the arm can block the view of the canopy, while the arm is in its working area. Hence, in order for the senor to view the canopy, the arm should be retracted. The retraction takes time that affects the speed of the system. Instead, the approach taken with the original kiwifruit harvester was to detect multiple fruit in the sensor data and pick multiple fruit before retracting the arm and performing detection again. The issue with this approach is that the fruit move during picking, as already discussed in Subsection 2.4. Since the fruit movement is undetected with the arm blocking the sensor field of view, the accuracy of the gripper pose relative to the fruit is affected. Hence, there is a speed versus accuracy trade off with this sensor placement.
  - Fruit may be occluded from below and so sensors that look up from below may not be able to view these fruit. Anecdotally, it was common for hanging branches to block the view of one or more cameras on the original kiwifruit harvester. In addition, fruit may be behind leaves, branches and wires that are higher in the canopy.

- Having a depth sensor in the palm of the end effector. Issues with this sensor placement include:
  - This sensor placement may increase the volume of the end effector, which may affect how many fruit the end effector can reach without collisions.
  - After a pick operation the sensor may be very close to fruit and hence may have to retract and turn in order to view more fruit to pick. This extra searching operation could cost time.
  - The sensors may be damaged in the cluttered canopy.
  - The sensor lenses may collect kiwifruit hair and other rubbish, requiring more frequent maintenance and cleaning.
- Having the sensor behind the robot arm, angled upwards towards the canopy with a view of the working area of the robot arm. With this sensor placement the robot arm may block the view of the sensor less so that there is not the same accuracy and speed trade off that there is with the original kiwifruit harvester. It also seemed that this sensor placement would allow fruit that were occluded from below to be viewed from the side. It was also thought that pose detection of the fruit might be improved since more area of the fruit would be visible from the side of the fruit.

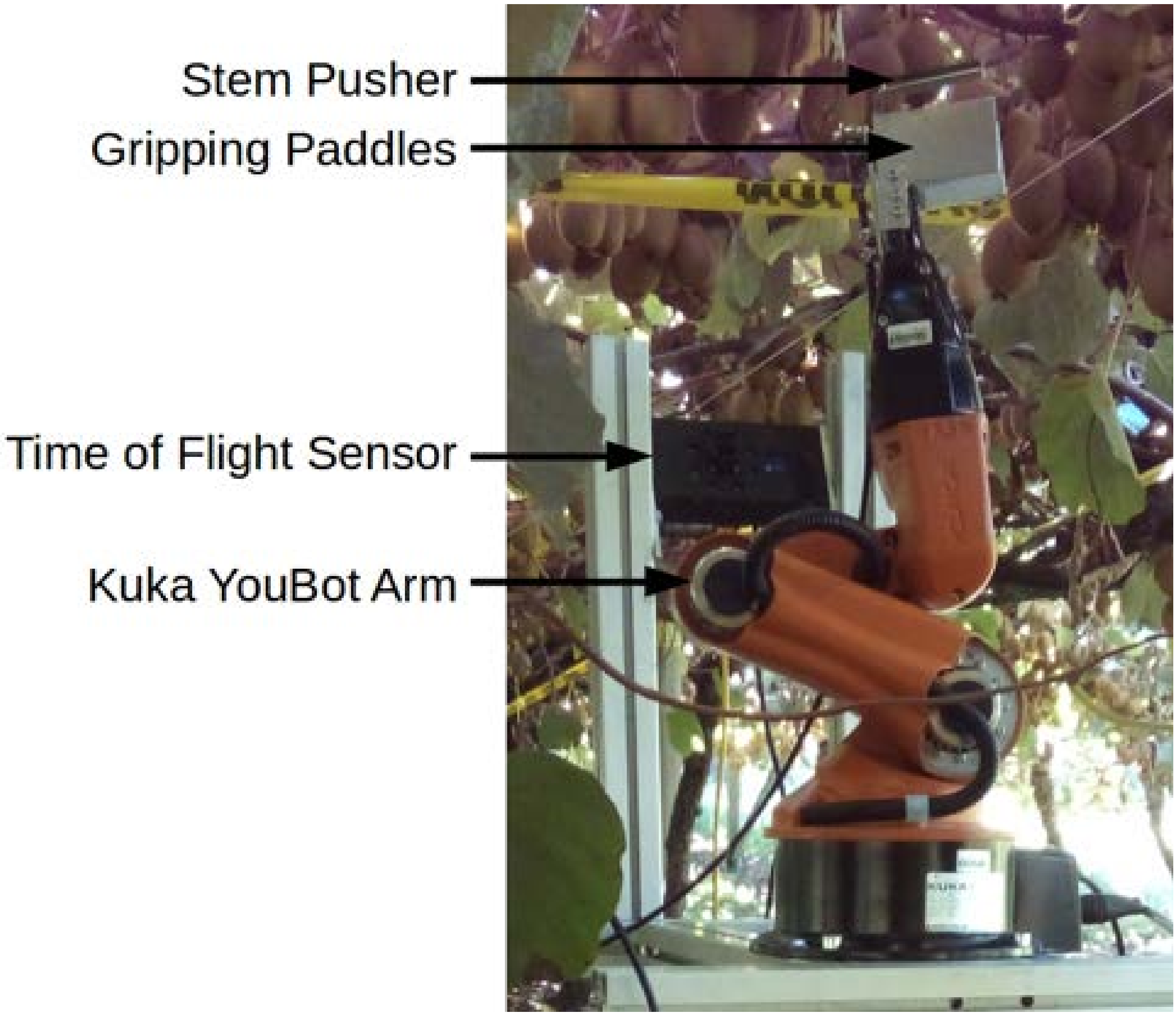

*Figure 25: The sensor and robot arm of the second kiwifruit harvester.*

The sensor finally selected for the second kiwifruit harvester was the Basler Time of Flight TOF640-20GM-850NM [77]. The sensor was behind the robot arm and angled up towards the canopy as shown in Figure 25. It was postulated that there might be some advantages to mounting the sensor to the first moving link of the robot arm, so that the sensor rotated with this first link. These advantages included:

- A narrower field of view of the sensor image could be processed or the sensor could be moved closer to the arm, since the sensor would only have to view the workspace of the remaining links in the robot arm. Another way of looking at this is that having the sensor moving means that the fields of view from different poses sum to a larger field of view.

- With the sensor on the first moving link, it would be possible to keep the sensor aligned with the plane of the middle 3 links of the robot arm. As a result, an obstruction in front of a fruit, as viewed from the sensor, would generally also be an obstruction to the robot arm, when picking this fruit. This would allow for simpler detection of unobstructed fruit, as further discussed in Subsection 2.6.2.

However, as shown in Figure 25 the sensor was not mounted to the first link of the robot arm and so these potential advantages did not apply. However, even for the pose of the sensor on the second kiwifruit harvester, a detected unobstructed fruit directly in front of the sensor would generally have also been an unobstructed fruit for the robot arm. For fruit less directly in front of the sensor, this would have applied to a lesser degree.

The final pose of the Time of Flight sensor was adjusted so that the working area of the arm was covered by the field of view. There was also a trade-off to consider between having the sensor higher, which meant having less occlusion from the robot arm, or having the sensor lower, which meant less chance of occlusion and collision with the canopy.

### 2.6.2 Kiwifruit Detection by Instance Segmentation

When segmentation was used in the computer vision systems for the original kiwifruit harvester, the kiwifruit calyx was a key feature in determining the target pose of the robot arm. The kiwifruit calyx was generally visible since it was typically at the bottom of the fruit and the sensors looked up at the fruit from below. However, having the Time of Flight sensor principal axis at an angle closer to horizontal meant that not all of the kiwifruit calyxes were visible. The feature that was visible on all of the fruit was the skin and so this was used as the key class for detection of the fruit. Semantic segmentation segments kiwifruit skin from adjacent fruit together in single connected masks as shown in Figure 8. Hence, it was decided to use instance segmentation to detect the kiwifruit skin of individual fruit on the second kiwifruit harvester.

When the instance segmentation systems were designed for the original kiwifruit harvester, multiple classes were used for the target objects, which were the kiwifruit calyx and kiwifruit skin, as well as the classes for the obstructions, such as the branches, steel beams and wires. This allowed obstructions adjacent to fruit to be detected and avoided. However, this approach does not directly address the issue of fruit occluding other fruit. For example, in the intensity data in Figure 26 there are many fruit that are partially occluding adjacent fruit. If the robot arm was commanded to pick an occluded fruit, it may contact and dislodge or damage the adjacent fruit. This failure mode was frequently observed on the original kiwifruit harvester.

For the second kiwifruit harvester, instead of explicitly detecting certain classes of obstructions, it was decided to instead detect unobstructed fruit. To illustrate this idea, an image and label pair from the training dataset is shown in Figure 26. The only fruit which are labelled are completely unobstructed. This approach assumes that the field of view of the sensor and the working area of the arm are aligned so that an unobstructed fruit in the sensor image is indeed an unobstructed fruit for the robot arm. This was approximately the case for the second kiwifruit harvester when the fruit was directly in front of the sensor, as discussed in Subsection 2.6.1.

The approach of only targeting unobstructed fruit relied on fruit becoming unobstructed after other fruit were removed or a fruit appearing unobstructed from another feasible sensor viewpoint. It was unclear how this reliance would affect the picking rate; however, it was assumed that this would be revealed with thorough testing of the second kiwifruit harvester.

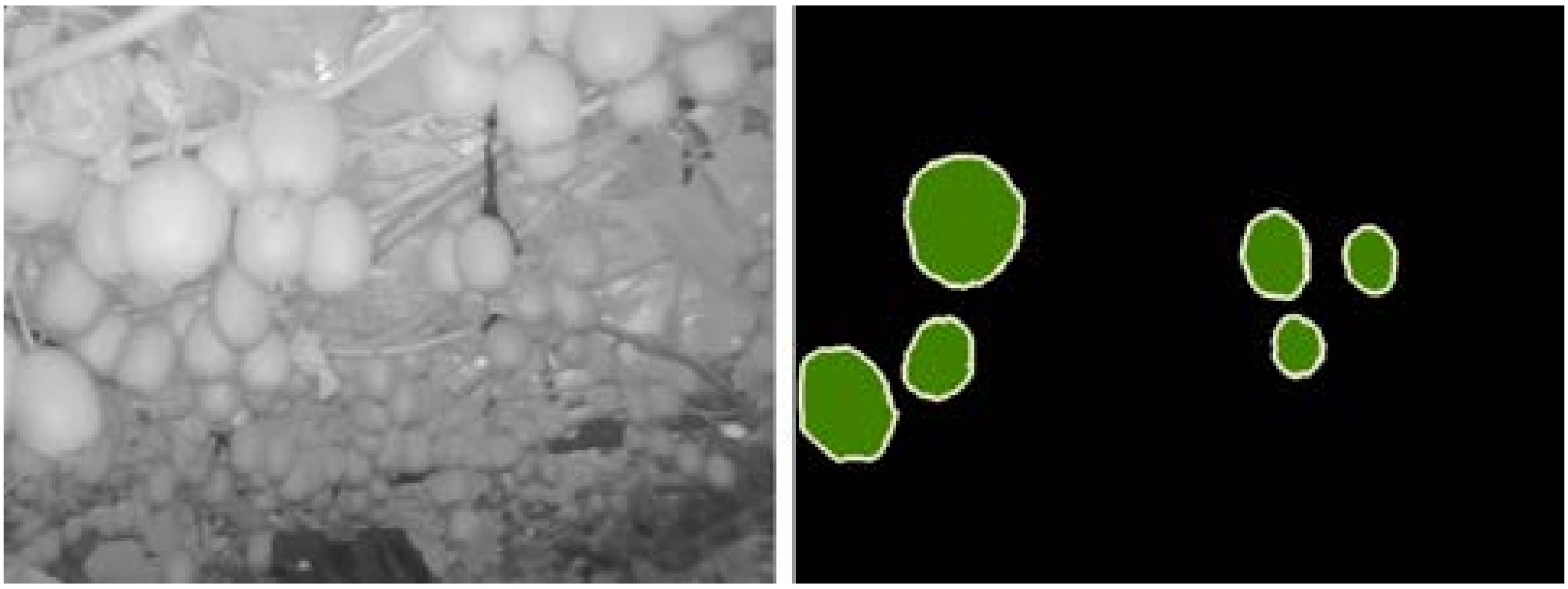

*Figure 26: An input intensity image (left) and label (right) pair from the training dataset for the second kiwifruit harvester, where only unobstructed and close fruit are labelled.*

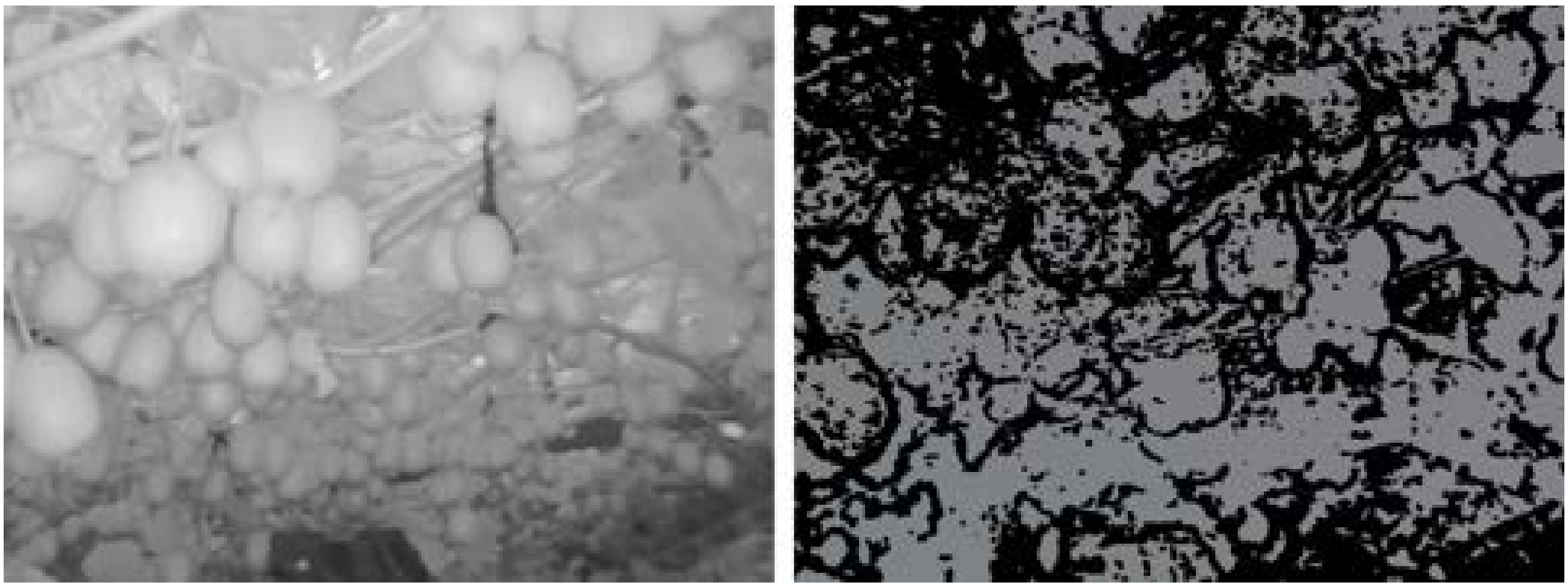

*Figure 27: Intensity data (left) and corresponding depth data (right).*

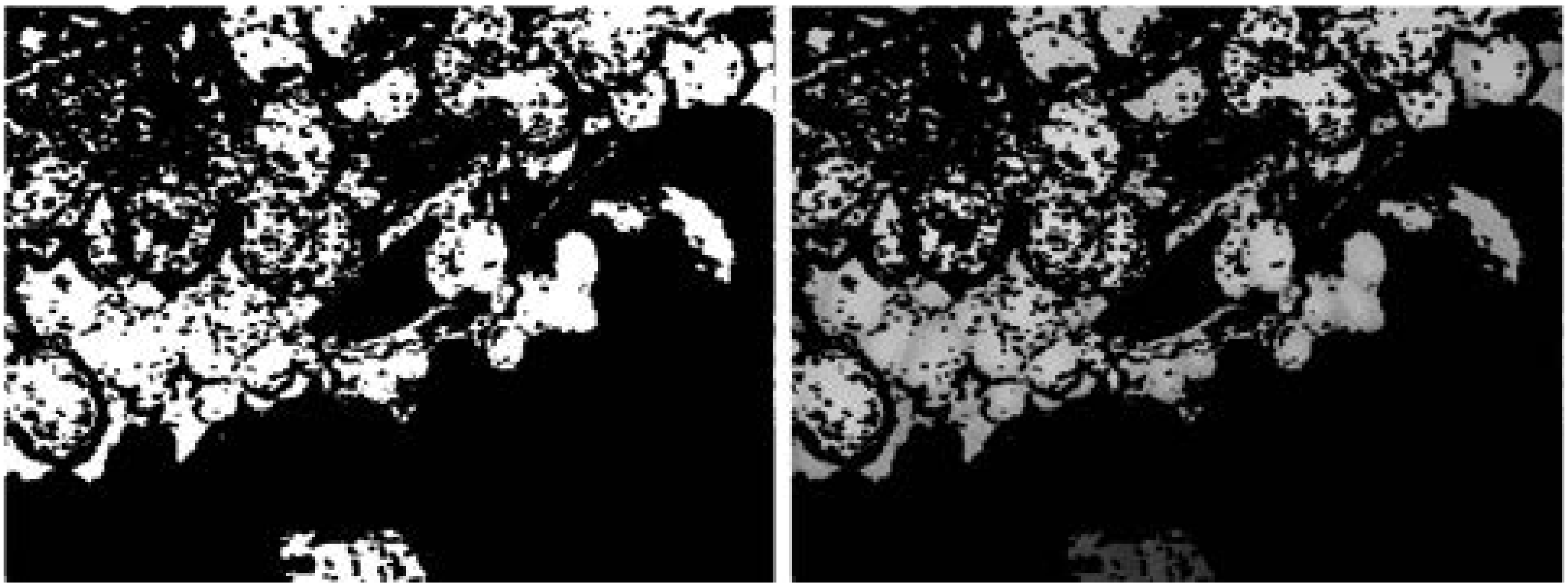

*Figure 28: A mask created by setting pixels high if the corresponding range value from Figure 27 was less than 1 m (left) and this mask applied to the intensity data from Figure 27 (right).*

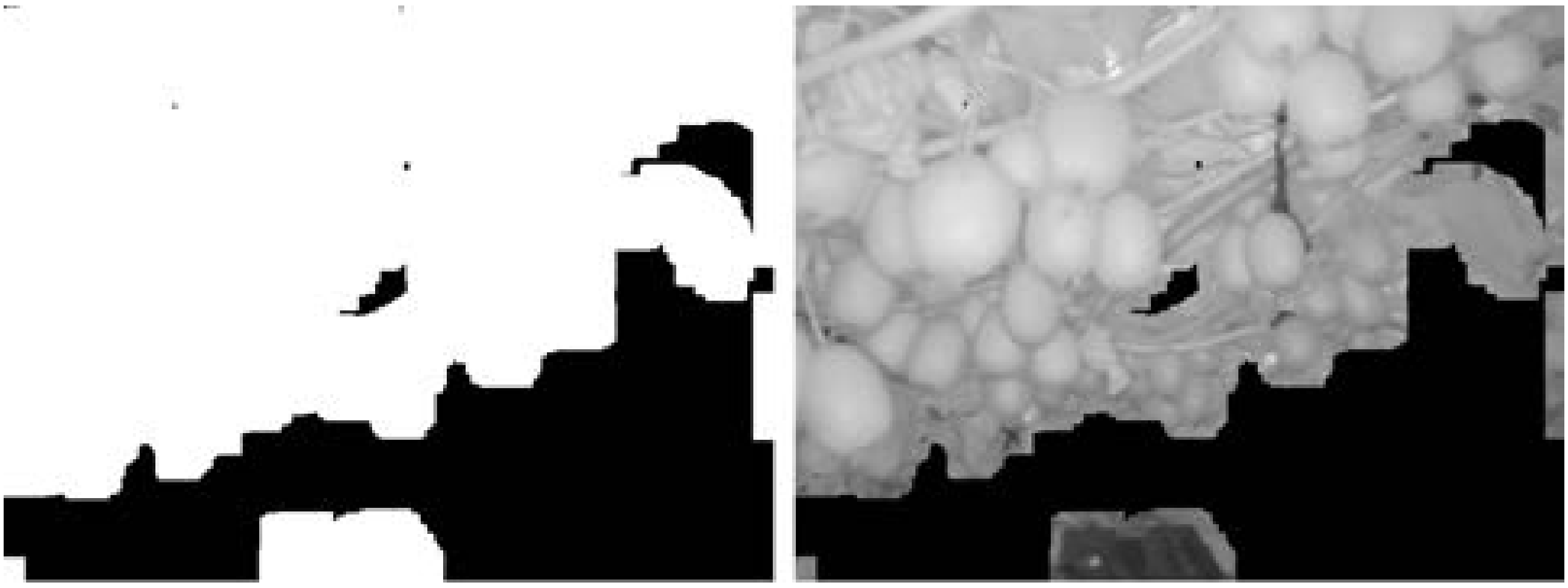

*Figure 29: A mask created by dilating the mask from Figure 28 (left) and this mask applied to the intensity data from Figure 27 (right).*

In addition to only detecting fruit that were unobstructed, it was proposed that during labelling it would be useful to disregard fruit that were clearly beyond the range of the arm. For example, considering the intensity data in Figure 27, many of the visible fruit are not in range of the arm. Using the depth image in Figure 27, for every depth pixel that was less than 1 m away from the Time of Flight sensor, the corresponding pixel was toggled from low to high in a mask and this mask was applied to the intensity data as shown in Figure 28. The resulting intensity data in Figure 28 was difficult to label since so much of the data was missing. As a result, the mask was dilated by 33 pixels per pixel and the resulting mask was applied to the intensity data from Figure 27, with the result shown in Figure 29. This data was labelled to give the input image and label pair shown in Figure 26. 60 images were labelled in this way, 50 of which were used for the training dataset and the remaining 10 were used for testing. A validation dataset was not required because no hyperparameter tuning was performed. The hyperparameters used were the same as those previously used in the development of the original kiwifruit harvester detection system (Table 5). The test set results are given in Table 13. An example inference result is shown in Figure 30.

Considering the results from Table 13, the average precision and average recall values are generally quite high, especially compared to the results for the original kiwifruit harvester and results for other datasets [87]. It is interesting to note in Table 13 that the average precision and average recall for the detection of small objects is zero; however, this was the expected result since very small kiwifruit were deliberately not labelled since these were the ones that were more than a metre from the Time of Flight sensor and beyond the working area of the robot arm.

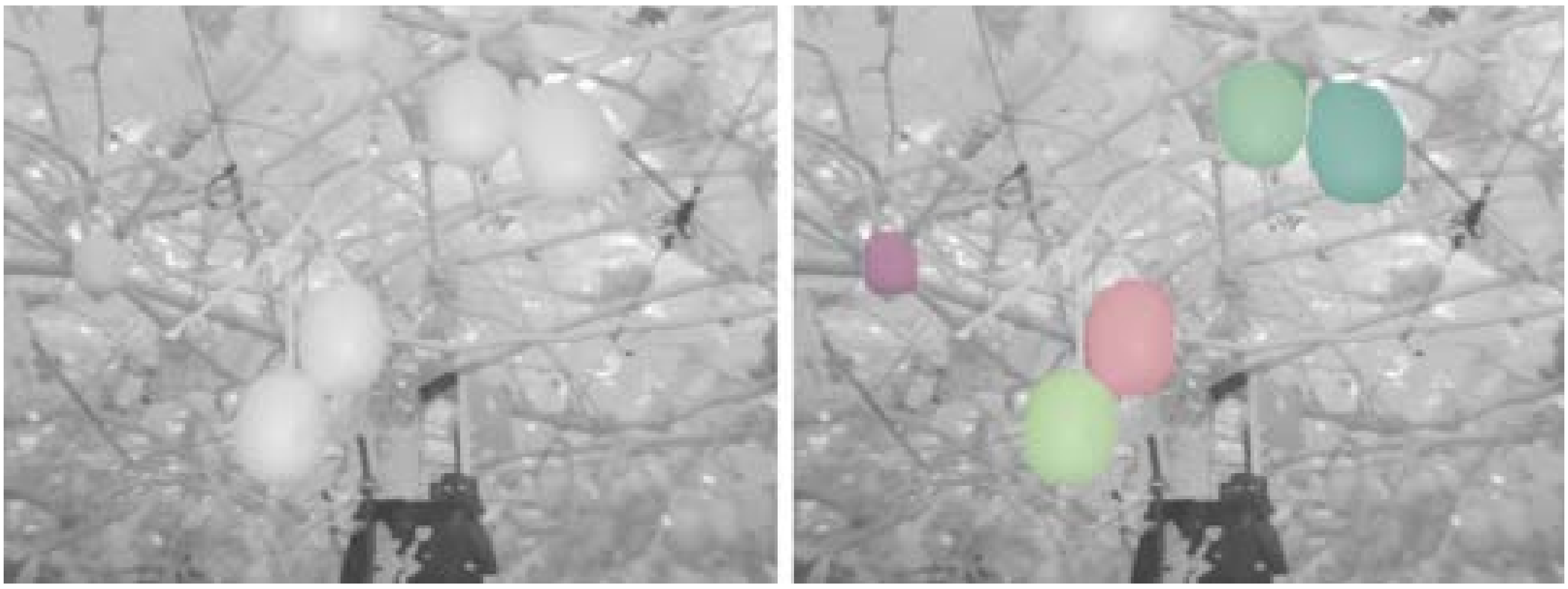

*Figure 30: An example inference result (right) from the second kiwifruit harvester on a test image (left), using Mask R-CNN on Time of Flight intensity data.*

*Table 13: Test results from training Mask R-CNN for the second kiwifruit harvester.*

| **Metric Description** | | | | **Metric Value** |
|---|---|---|---|---|
| **Type** | **IoU** | **Object Pixel Area** | **Given Detections per Image** | |
| Average Precision | 0.50 | Any | 100 | 0.88 |
| | 0.75 | Any | 100 | 0.86 |
| | 0.50:0.05:0.95 | Any | 100 | 0.67 |
| | | $< 32^2$ | 100 | 0.00 |
| | | $> 32^2$, $< 96^2$ | 100 | 0.67 |
| | | $> 96^2$ | 100 | 0.81 |
| Average Recall | | Any | 1 | 0.12 |
| | | Any | 10 | 0.67 |
| | | Any | 100 | 0.73 |
| | | $< 32^2$ | 100 | 0.00 |
| | | $> 32^2$, $< 96^2$ | 100 | 0.73 |
| | | $> 96^2$ | 100 | 0.83 |

### 2.6.3 Arm Target Pose Calculation

The goal of the arm target pose calculation was to use the segmentation output and depth data to produce a target pose for the robot arm to perform the picking of the detected kiwifruit. The premise of the first algorithms was to find the centre of the fruit and then to calculate a target pose by offsetting from the centre of the fruit. To find the centre of the fruit, it was postulated that it was important to take into account:

- Segmentation errors. In particular, it was thought that points on the boundaries of a segmentation mask could be points from adjacent objects in the data and hence the algorithms should not necessarily use these points.
- Noise in the depth data. In particular, it was proposed that more than one point of data should be used to reduce the effects of noise in the depth data and that this could be achieved by using a mean or percentile of points.

To find the centre of the fruit, a surface central point based on the visible skin of the fruit was firstly calculated. Then a fixed offset from this surface central point's polar coordinate radius was calculated as an approximate fruit centre. The algorithm to find the surface central point, while taking into account segmentation errors and depth data noise, was attempted by multiple methods, including:

- Extracting all of the depth data corresponding to the segmented mask for a detected fruit and finding a median point, which was composed of the median x, y and z coordinates in the Time of Flight sensor's coordinate system. Then all extracted depth coordinates were ranked in termed of their Euclidean distance from the median point and the closest 20 percent of points were extracted; an illustration of these extracted points is shown in Figure 31. The surface central point used was the average of the x, y and z coordinates of these extracted points. An issue with this method was that some of the most central points on the kiwifruit can be excluded from the final points extraction as shown in Figure 31. These central points are the points that are being targeted in the final extraction of points and so disregarding them in the calculation of the surface central point could be a weakness of this method.
- Another method was developed which began by finding the image coordinates of the centre of the segmented mask. Then the depth data of the associated point was extracted. Then the depth data around the extracted depth data was also extracted. This was repeated until a goal number of points had been extracted, where this goal number was a set proportion of the number of pixels in the segmented mask. An illustration of these extracted points is shown in Figure 32. The surface central point used was the median of the x, y and z coordinates of the extracted depth points. An issue with this method was that the points extracted formed a square; whereas intuition might dictate that the points extracted should form a shape that more closely fits the shape of the fruit- such as a circle, an ellipse or even an ellipse aligned with the principal components of the segmented mask. Nevertheless, this method was trialled because, unlike the results from Figure 31, the most central points were extracted. This method is referred to as the ‘central square method’ in the rest of this section.

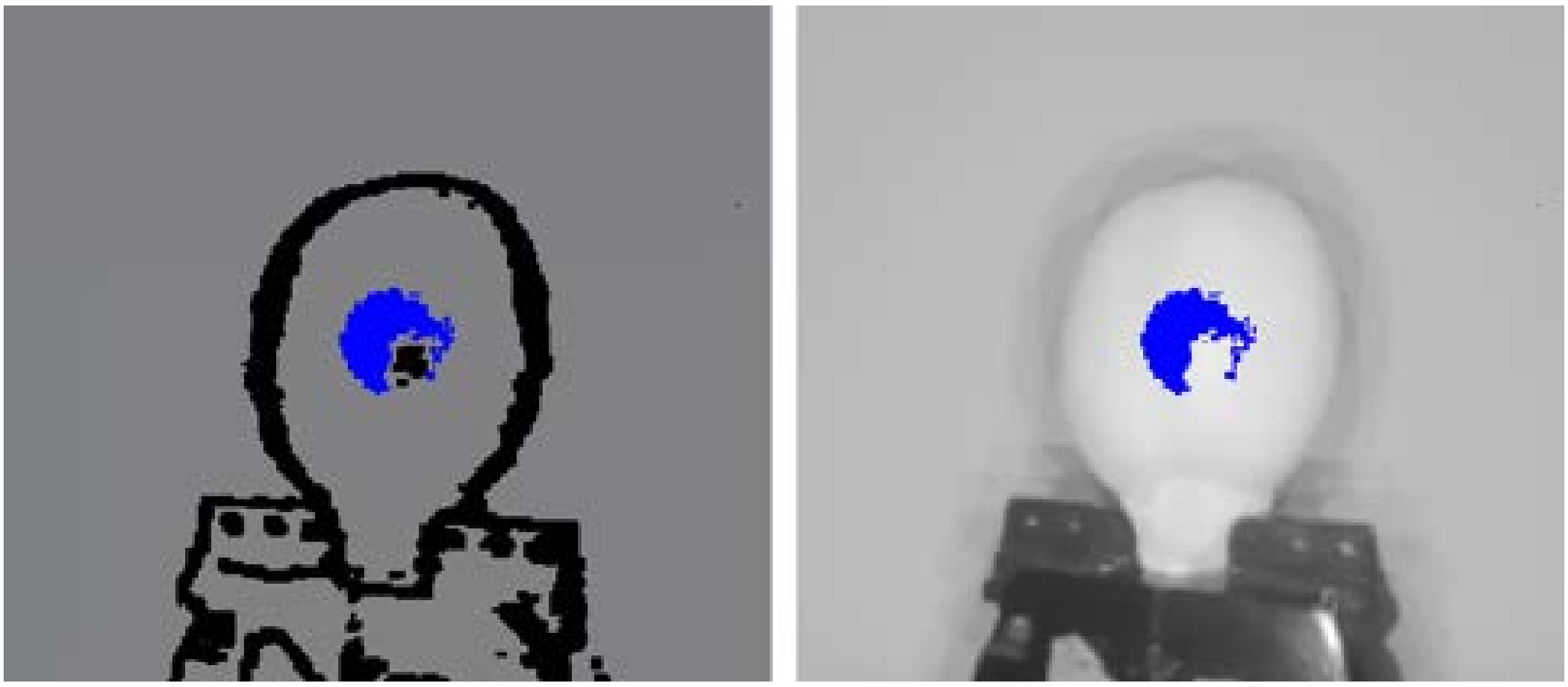

*Figure 31: The points extracted on the surface of a kiwifruit (blue), after ranking the closest points to the median of the x, y, z coordinates of points on the kiwifruit, shown with the depth data (left) and the intensity data (right).*

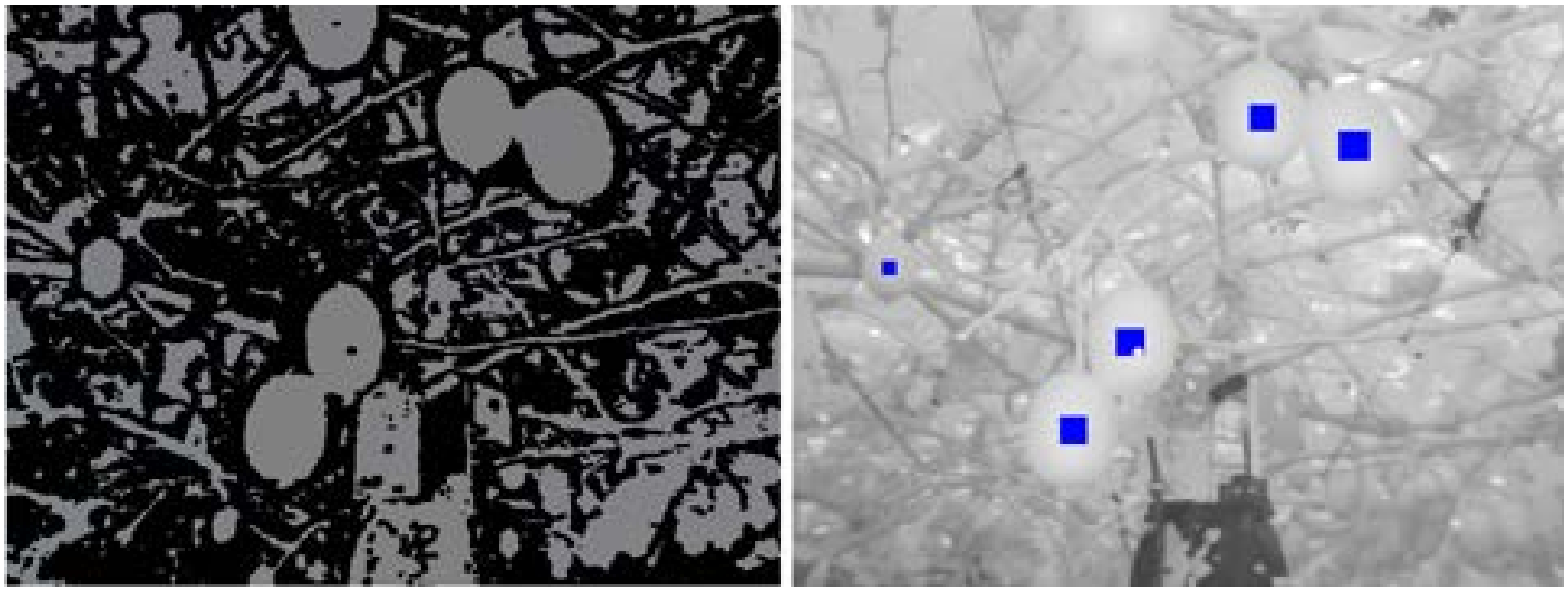

*Figure 32: Depth data (left) and intensity data (right) with the points extracted, using the central squares method, drawn in blue.*

As described in Subsection 2.7.3, the central square method was used during initial trials of the second kiwifruit harvester. However, it was found that the pose of the stem pusher was frequently too low or too high. After making these observations, it seemed that calculating the arm target pose just based on fixed offsets and the centre of the visible skin was a flawed approach, since the varying height dimension of the fruit was not taken into account. The stem pusher must be positioned above the top of the fruit and so it seemed that detecting a point closer to the top of the fruit would account for some of the variability in fruit heights. As a result, the central square method was modified so that instead of beginning by finding the centre of the segmented mask, a vertical percentile of the segmented mask was used, so that the extracted square would be closer to the stem. The percentile used was the 20th percentile of the image y coordinates, which was a point much higher on the fruit as can be seen in Figure 33. This modification of the central square method is referred to as the 'percentile square' method in the rest of this section.

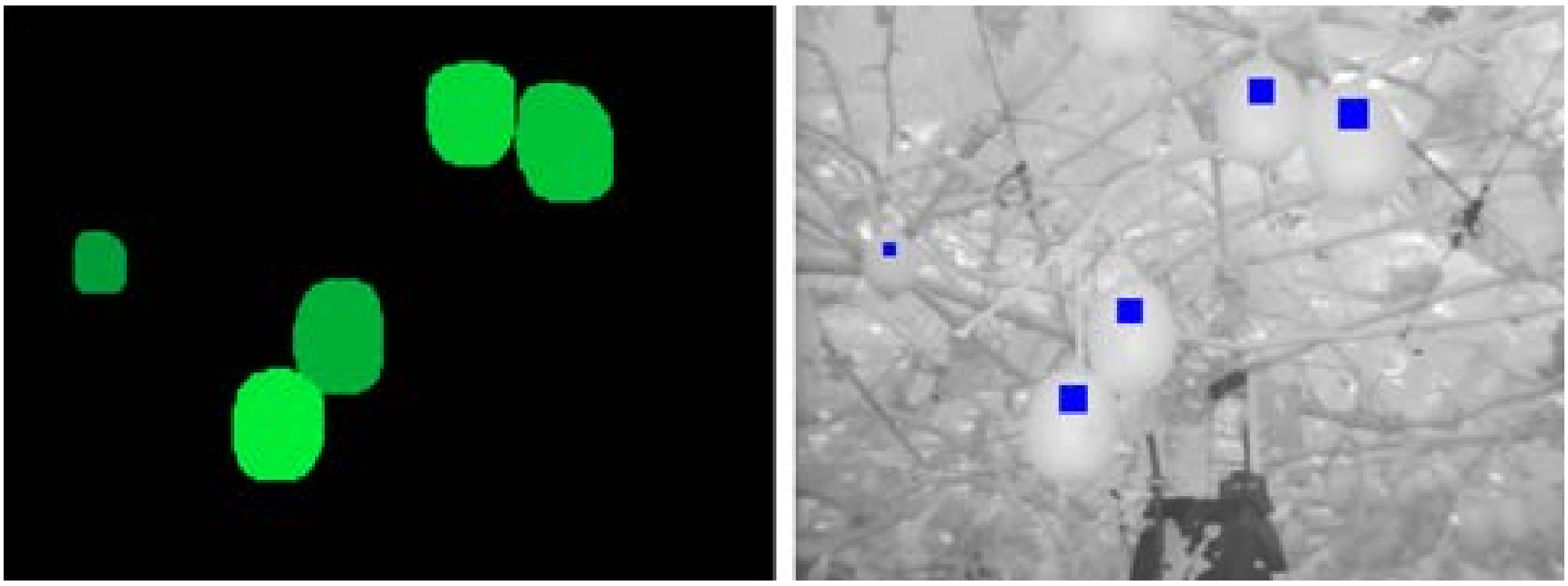

*Figure 33: The segmentation masks in shades of green (left) that were an input for the points extracted by the percentile square method in blue (right).*

### 2.6.4 Time of Flight Sensor to Robot Arm Calibration

The fruit segmentation and target pose calculation were performed using the point cloud data from the Time of Flight sensor. The transform between the Time of Flight sensor coordinate system and the robot arm coordinate system had to be calculated in order to command the robot end effector to targets calculated using the Time of Flight data. The steps used to find the coordinate system transformation were:

1. A spherical target was mounted at the centre of the end of the Kuka YouBot arm using an adhesive (Figure 34).
2. The position of the spherical target, relative to the end of the last link of the Kuka YouBot arm was measured using calipers.

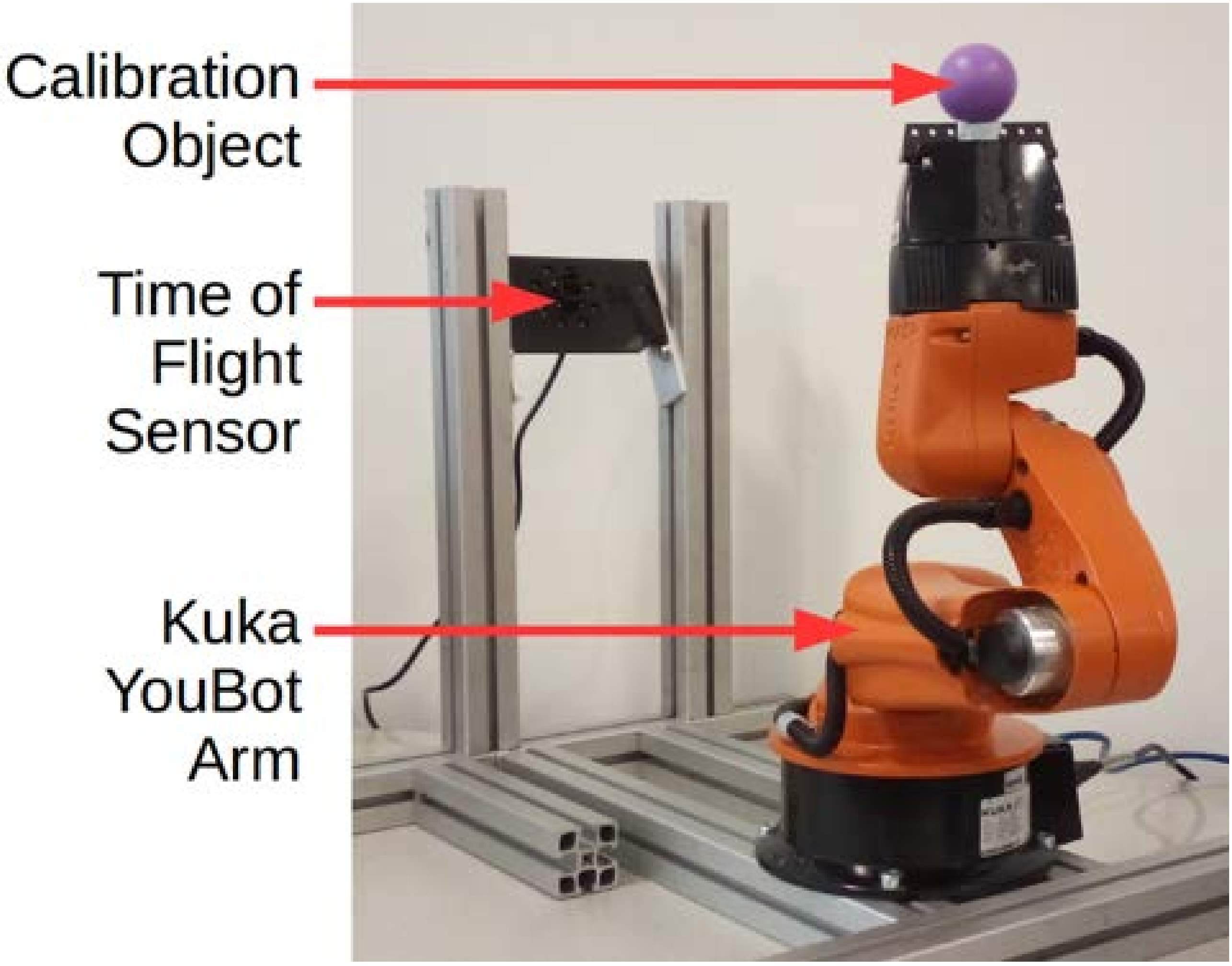

*Figure 34: The hardware setup used for the calibration between the Time of Flight sensor and Kuka YouBot arm coordinate systems for the second kiwifruit harvester.*

3. The robot arm was automatically controlled to move so that the spherical target was moved to and paused at 700 different positions, where the spherical target was fully visible in the Time of Flight sensor intensity channel and encompassing the working space of the robot arm for harvesting.

4. For each position of the spherical target from step 3, the joint angles of the robot arm and a frame of data from the Time of Flight sensor, including the intensity channel and the depth channel, were collected.

5. For each position from step 3, the position of the spherical target in the robot arm coordinate frame was found using the joint angles, forward kinematics, the dimensions of links taken from Kuka specifications [88] and the dimensions measured in step 2.

6. For each position from step 3, the position of the spherical target in the Time of Flight sensor coordinate frame was found using the central square method from Subsection 2.6.3.

7. The transform between the robot arm coordinate system and the Time of Flight sensor coordinate system was calculated using a Singular Value Decomposition (SVD) [94] of the spherical target positions calculated from steps 5 and 6.

The accuracy of the calibration method was tested by using the calibration system to collect more point measurements of the calibration object's position using both the forward kinematics calculation based on the Kuka YouBot arm joint angles to find points, *K*, and the Time of Flight data, processed by segmentation and the central square method. The points from the central square method, *C*, were transformed to points, *C'*, in the Kuka YouBot arm coordinate system, using the calculated rotation, *R*, and translation, *T*, tensors, according to:

$$C' = RC + T \tag{1}$$

The average Euclidean distance error for the transformed calibration object points was calculated according to:

$$\frac{1}{n}\sum_{i=0}^{n}\left(\sqrt{\left(K[i]_x - C'[i]_x\right)^2 + \left(K[i]_y - C'[i]_y\right)^2 + \left(K[i]_z - C'[i]_z\right)^2}\right) \tag{2}$$

The resulting error was 0.0024 m using $n$=100 points. This error includes the error in the segmentation and central square method. It was assumed that this error would be acceptable for the second kiwifruit harvesting system, since it was smaller than the error allowed for in the opening of the grippers, and hence the calculated rotation and translation tensors were used in transforming the kiwifruit targets to the Kuka YouBot arm coordinate system.

### 2.6.5 Second Kiwifruit Harvester Arm Control

If the canopy was static during harvesting, it may have been possible to use a multi-query planner in order to invest time into sampling the configuration space and building a roadmap of paths and points in the free space [95]. Although this approach could be time-consuming, it was proposed that it might be possible to perform this action ahead of time by having an additional set of sensors ahead of the arms so that harvesting could be performed in parallel with establishing roadmaps in the next volume of canopy to be harvested. However, as has already been discussed in Subsection 2.4, the canopy does move during harvesting and so the obstacle space and free space changes, making multi-query planning a computationally prohibitive option for adapting to the changing canopy.

Other arm control paradigms were considered including visual servoing and reinforcement learning methods (which are reviewed in Section 7). The method used for planning paths for the second kiwifruit harvester was similar to the method used for the original kiwifruit harvester; both path planners constructed a sequence of motion primitives, which were parameterised based on the starting pose, target pose and the work space of the arm for harvesting. In order to plan paths for the second kiwifruit harvester, a trajectory of arm configurations was calculated. Each arm trajectory consisted of a list of arm configurations, where each configuration consisted of the angles of the five rotational arm joints.

On the second kiwifruit harvester, the initial pose of the arm was set so that the end effector was low, in order to allow the sensor to view more of the canopy. Once a target was received, the first motion primitive in the trajectory consisted of rotating the base joint/ joint 0 of the arm to an angular displacement, where the middle three arm joints were on or approximately on the same plane as the target point.

In order to produce a trajectory of arm configurations for the middle three joints, the space of feasible configurations for these joints was calculated for a grid of end effector positions in 2D Cartesian space aligned with the middle three joints, given a fixed end effector angle. Example outputs of this calculation are shown in Figure 35, where the red-green-blue values at each pixel correspond to the angles of joint 1, joint 2 and joint 3 for an end effector position, which is given by the pixel coordinates. The inverse kinematics equations used to calculate the joint angles at each pixel are detailed in Section 8.

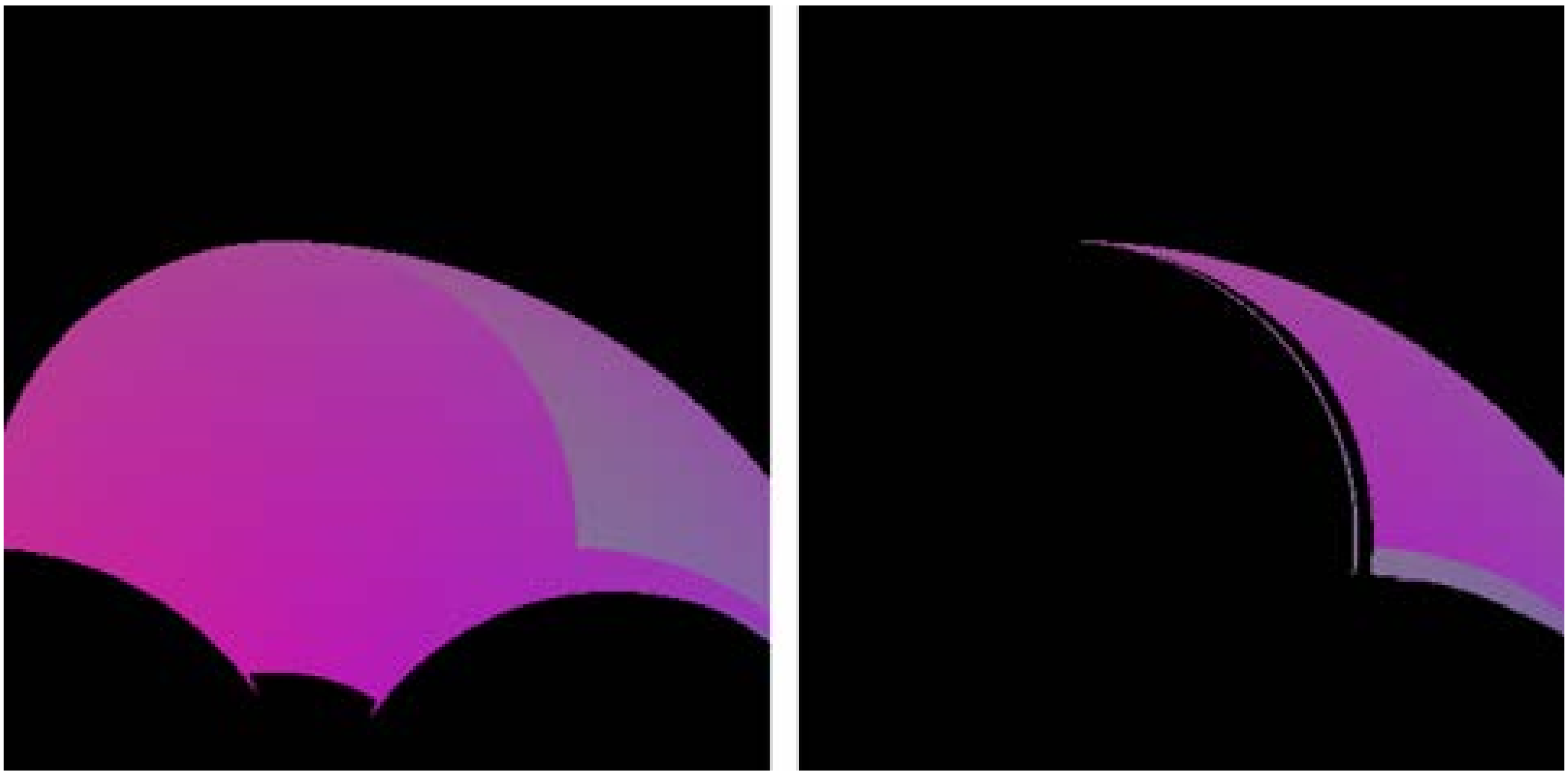

*Figure 35: Illustrations of the reachable points in space for the robot arm end effector, where the vertical axis of each image is the vertical axis in the real world, the horizontal axis of each image here is the horizontal axis aligned with the arm in the real world and the RGB values at each pixel are the middle three joint angles, calculated by inverse kinematics.*

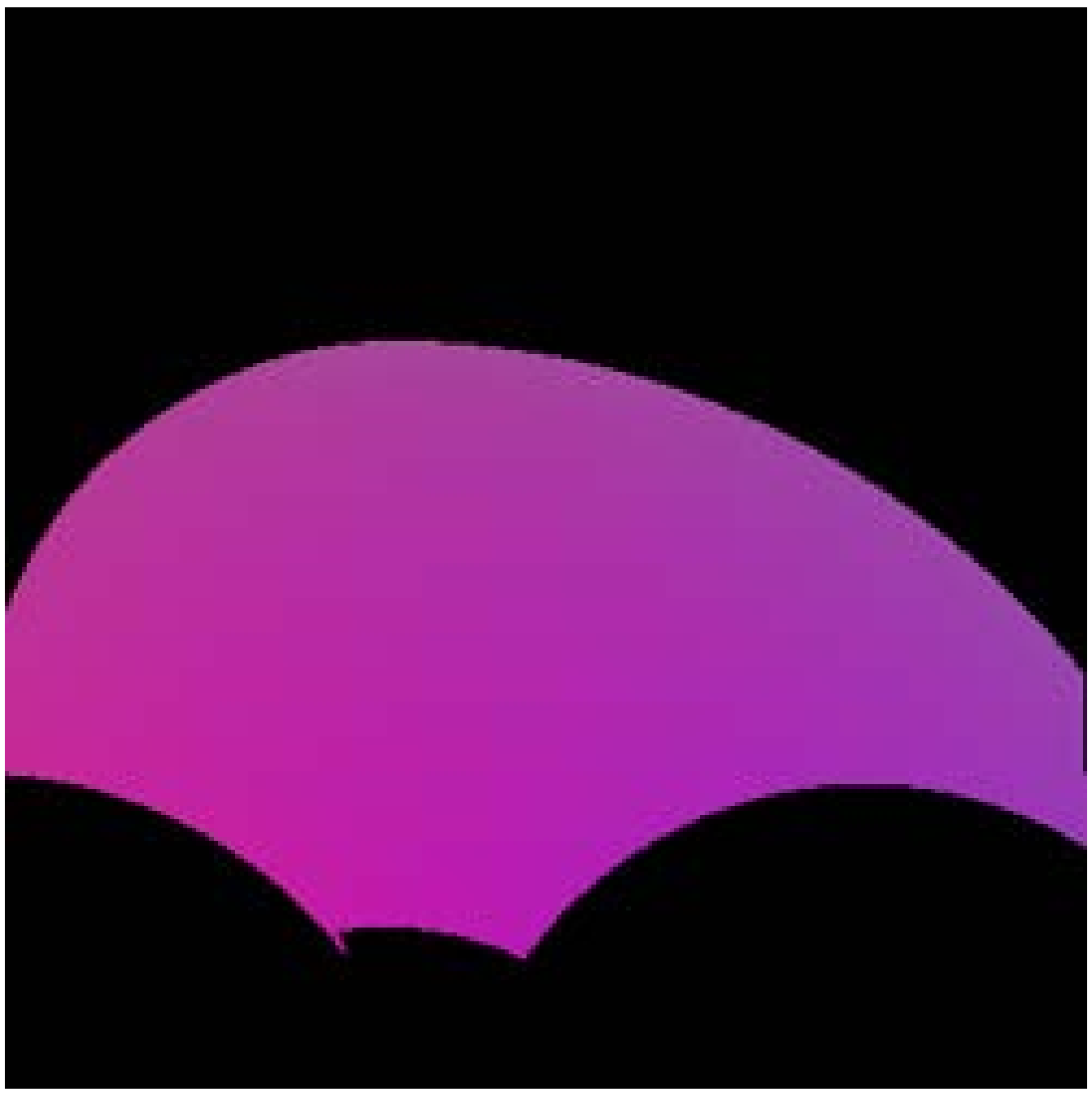

*Figure 36: A single work space of end effector positions on the plane aligned with joints 1, 2 and 3- with joint 1, 2 and 3 angles given by the pixel values- consolidating the outputs from Figure 35, with the angular displacement between adjacent points minimised and a fixed end effector angle.*

There are up to two outputs for each end effector position in Figure 35, because there are up to two sets of joint 1, joint 2 and joint 3 angles for an end effector pose. These two sets of outputs were consolidated to one set, where the joint angle displacements between adjacent points in the grid were minimised, as shown in Figure 36. The two dimensional space shown in Figure 36 represents a region of the plane aligned with joints 1, 2 and 3, with a height range from 0.4 m to 0.75 m above the origin of the YouBot arm and horizontal range of -0.09 m to 0.26 m from the origin of the YouBot arm, with the coordinate frames as defined in Section 8. This two dimensional work space was used to plan paths for joints 1, 2 and 3 by plotting waypoints in the work space according to motion primitives. If a waypoint was at a point where the pixel value was (0,0,0), then the waypoint was outside of the work space of the end effector and the path was rejected.

The waypoints for the end effector on the plane aligned with joints 1, 2 and 3 were found by starting at the initial pose (Figure 37). From the initial pose, waypoints were plotted on the two dimensional work space for the end effector (Figure 36), using the following steps:

- The first step was to subtract an offset from the target horizontal coordinate. The resulting difference was the horizontal coordinate at which the end effector would be raised. This approach of lifting the end effector in front of the target was used so that the end effector would be able to reach fruit that were obstructed from below by a solid obstacle. By raising the end effector in front of the fruit, an obstacle below the fruit could be avoided and once the end effector was at the target height, the end effector could be moved horizontally to engage with the fruit for picking.

- The next step was that waypoints were plotted in the direction of the horizontal coordinate at which the end effector would be raised. This was achieved by adding or subtracting small fixed horizontal displacements and keeping the vertical height of the waypoints low in the working space of the end effector (Figure 37). These waypoints were kept low in order to avoid collision with the canopy.
- When the waypoints were within range of the horizontal coordinate at which the end effector would be raised, the height of the waypoints was increased by adding small fixed vertical displacements until the target height was reached (Figure 37).
- The next step was to add small fixed horizontal displacements until the target horizontal and vertical coordinates were reached (Figure 37).

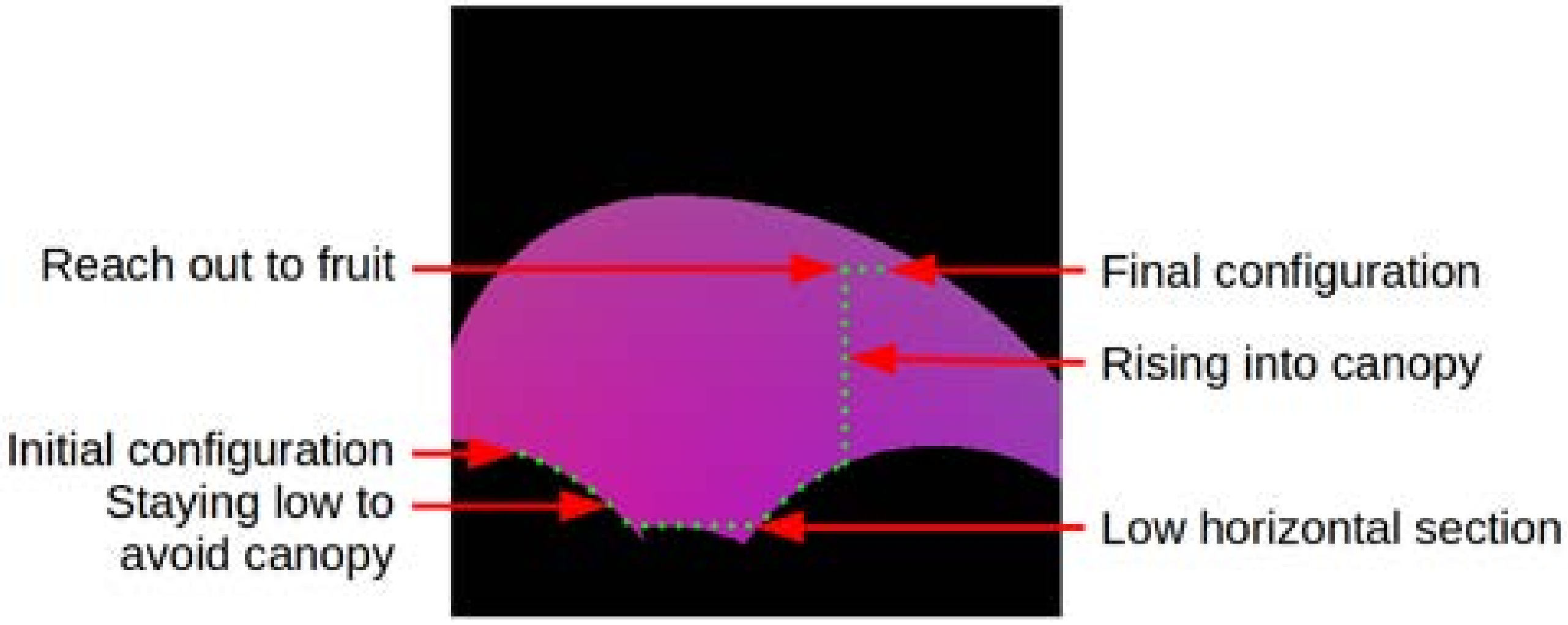

*Figure 37: A set of waypoints (green), where the end effector moves low during extension before moving up (towards the canopy) and finally out towards the fruit, where the planning is performed in this 2D space of reachable end effector positions on the plane of the middle three arm joints, given a fixed end effector angle.*

Figure 38 shows how the waypoints were plotted when moving away from a target position, after picking a fruit. The waypoints for moving away from the target were the reverse of the waypoints when moving towards the target. Plotting these waypoints stopped when the end effector had reached a vertical coordinate that was low for a given horizontal coordinate.

Once all of the waypoints were found, the pixel values were read in order to retrieve the joint 1, joint 2 and joint 3 angles for each waypoint. For arm control and getting the end effector to the fruit, the PID controller of the arm was sent consecutive joint angle sets as goals.

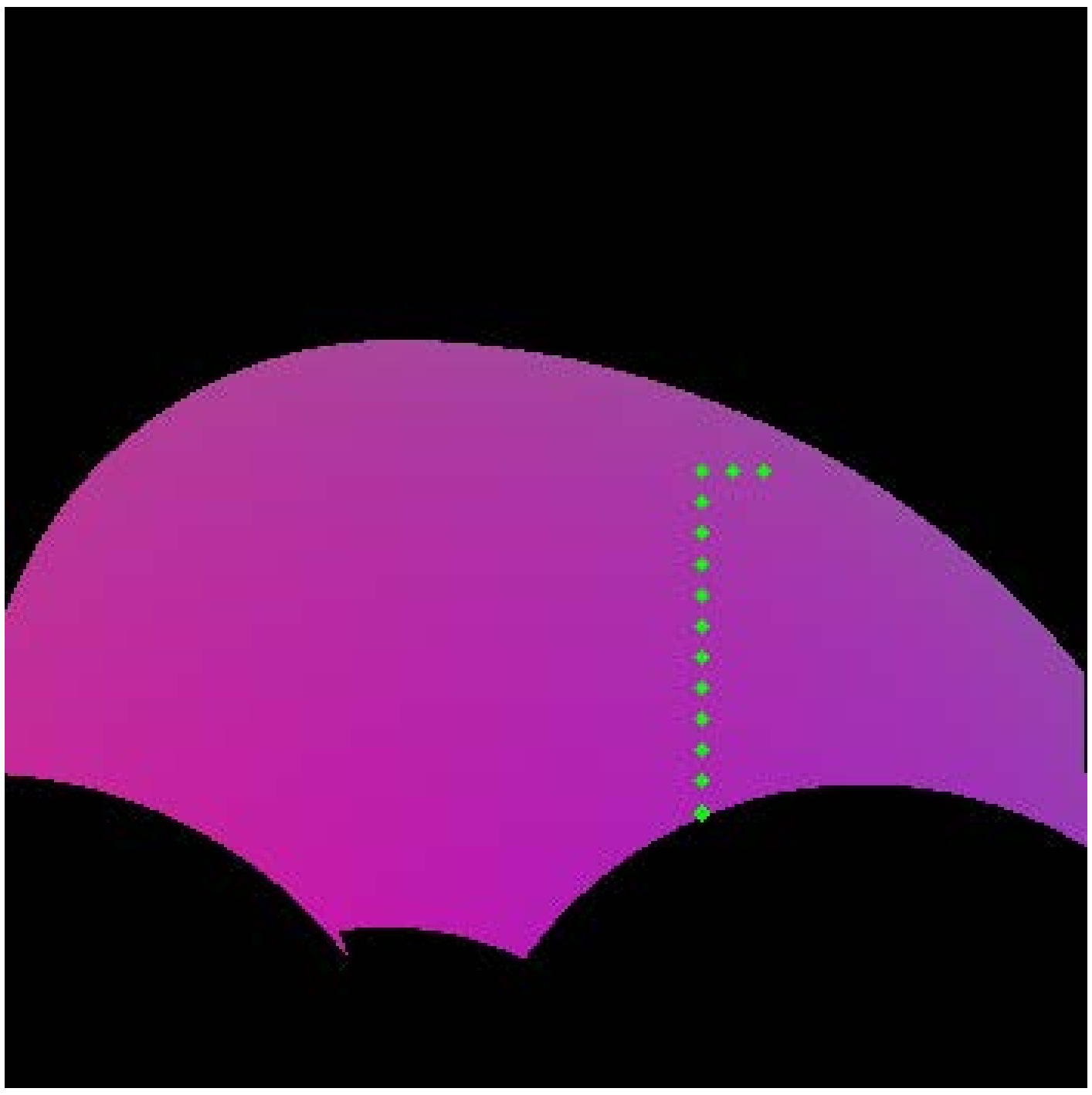

*Figure 38: The waypoints in the down movement were the reverse of the waypoints in the up movement, until the arm was at the minimum y coordinate for the current x coordinate.*

In addition to controlling the five rotational joints of the arm, the gripper and stem pusher were also controlled. The full sequence of actions is given in Figure 39. Note that this sequence of actions assumes that fruit chutes, like the ones on the original kiwifruit harvester, would be simple to develop for the second kiwifruit harvester in the future and that the kiwifruit would be dropped into such a chute.

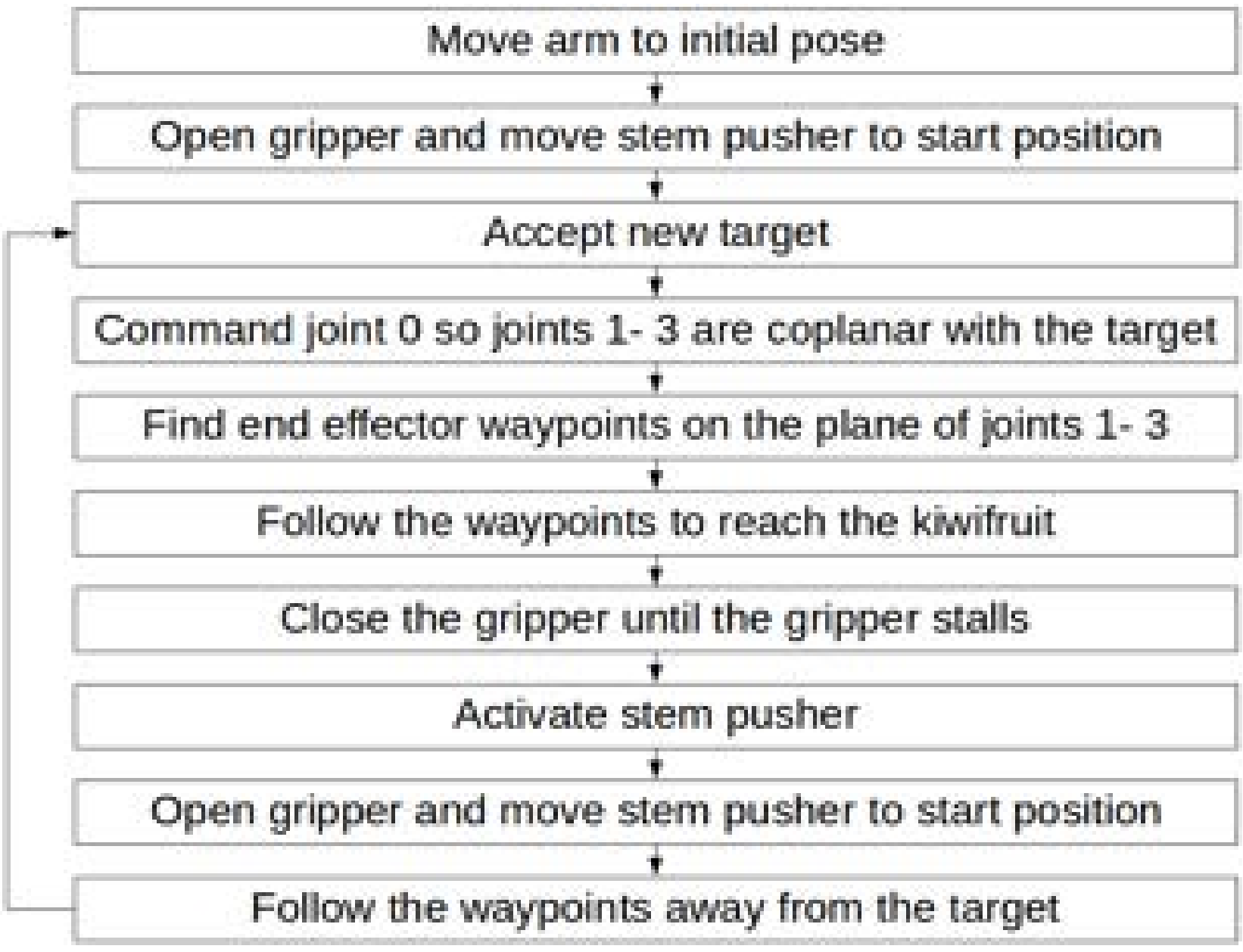

*Figure 39: Key steps in the arm, gripper and stem pusher control for picking fruit with the second kiwifruit harvester.*

## 2.7 Second Kiwifruit Harvester Orchard Experiments

Some of the parts of the second kiwifruit harvester were tested in the office, including the detection system and calibration method. However, other parts of the system, including the detachment mechanism and control system, were best tested in a real kiwifruit orchard.

The first question about the detachment mechanism was if it would work at all. There was also a question about whether higher speed or higher torque was more important for detachment. Another key point to be investigated was whether the second kiwifruit harvester detachment mechanism could potentially reach and pick more fruit than other kiwifruit detachment mechanisms; of particular interest was if the new detachment mechanism would be able to pick more than 80% of the fruit, in order to meet the MBIE Multipurpose Orchard Robotics Project goal.

Furthermore, the control system consisted of multiple parts, including the segmentation, target pose calculation, coordinate system transformation, path planning and arm control. There was a question about whether these parts worked together to form a control system that could pick kiwifruit in the real world. The control system and detachment mechanism were tested in three main experiments that are described in this subsection.

### *2.7.1 Kiwifruit Pickability Study*

An observed limitation of the original kiwifruit harvester arms was that the mechanical system could only approach the fruit from below. This meant that if a solid obstacle, such as a branch or wire, was below the fruit, the fruit could not be picked because the arm would be obstructed by the solid obstacle. In addition, even if the obstacle was adjacent to the fruit, the arm could be obstructed because the gripper could hit adjacent obstacles on the movement up into the canopy and the gripper could hit obstacles during the fruit detachment action.

Due to the hook shape of the stem pushing detachment mechanism, this mechanism could reach behind and beside some obstacles. A study was performed in the real world to determine what percentage of fruit could be reached by the stem pushing detachment mechanism with the Kuka YouBot arm and also to determine how this compared to the original kiwifruit harvesting system. This study was performed independent of the vision and control systems and so was only testing the mechanical designs of the systems. The goal of the study was to determine the maximum percentage of fruit that could be reached and picked by each mechanical harvesting system, regardless of the pose of the base of the harvesting system.

In addition to the original kiwifruit harvester, two variants of the stem pushing mechanism were tested; a larger variant included stereo cameras on the palm of the gripper and a smaller variant was made without the cameras. It was expected that the smaller mechanism would be able to reach into tighter spaces; however, it was unclear how the difference in mechanism size and shape would affect the percentage of fruit that could be reached.

The premise of the fruit pickability testing was to attempt to move each gripper mechanism to each fruit in an area and, for each fruit, to make a judgement as to whether the mechanism could be moved into a position to pick the fruit. The procedure used for this study was:

1. An area of kiwifruit canopy was marked with tape and solid structures in the canopy. This area was approximately 4 $m^2$ and contained 213 fruit. It is significant to note that the marked area extended from the centre of the row to the kiwifruit vine leaders at the edges of the row (Figure 40). This is significant because it was noticed that there were more obstructed fruit at the edges of the row.

2. For a kiwifruit in the marked area, the stem pushing mechanism with the cameras was moved by hand into a pose for picking and the action of picking was attempted without completing detachment.

3. If step 2 was successful, the kiwifruit was classified as “pickable” by the stem pushing mechanism (Figure 41). Otherwise, if it was not possible to find a feasible pose for picking, the kiwifruit was classified as “not pickable” by the stem pushing mechanism.

4. Steps 2 and 3 were repeated for the more compact stem pushing mechanism without the cameras and a mock-up of the original kiwifruit harvesting mechanism.

5. Step 4 was repeated for all 213 kiwifruit in the marked area.

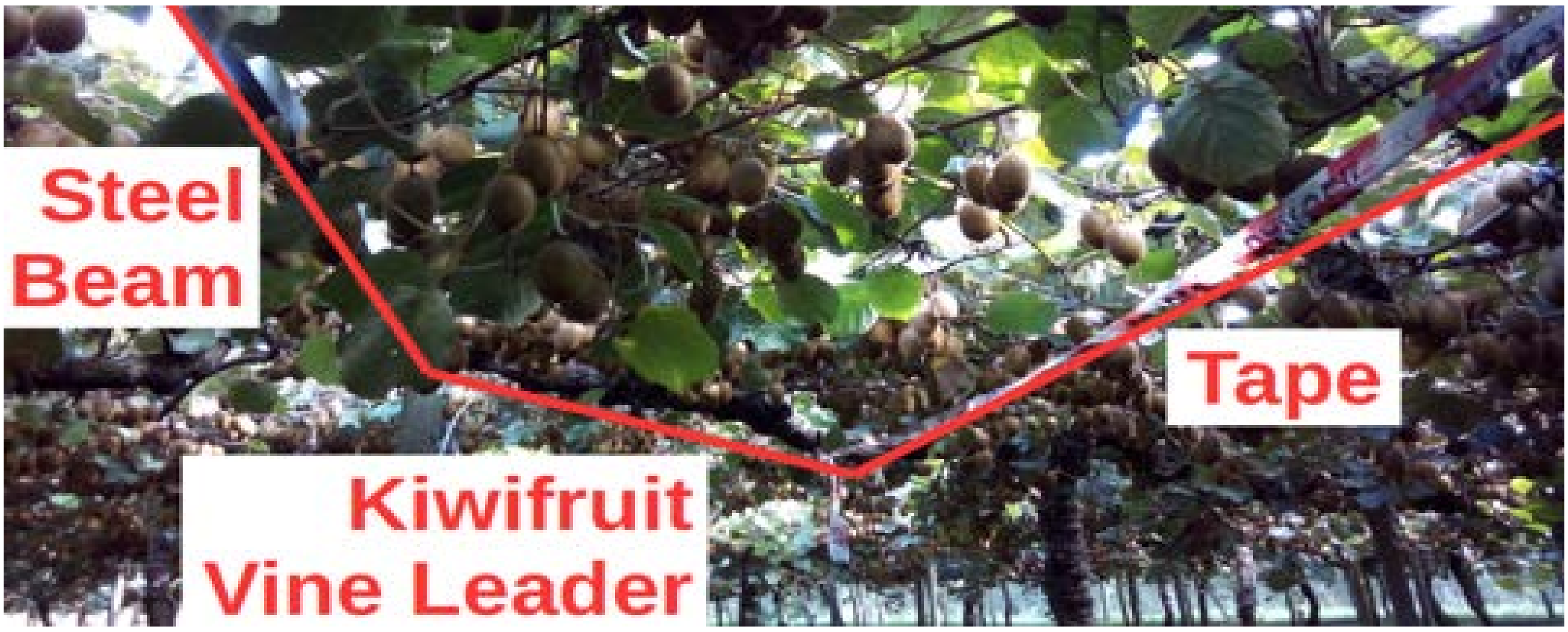

*Figure 40: Showing how the area used for the pickability study was marked and extended out to the edges of the row at the kiwifruit vine leaders.*

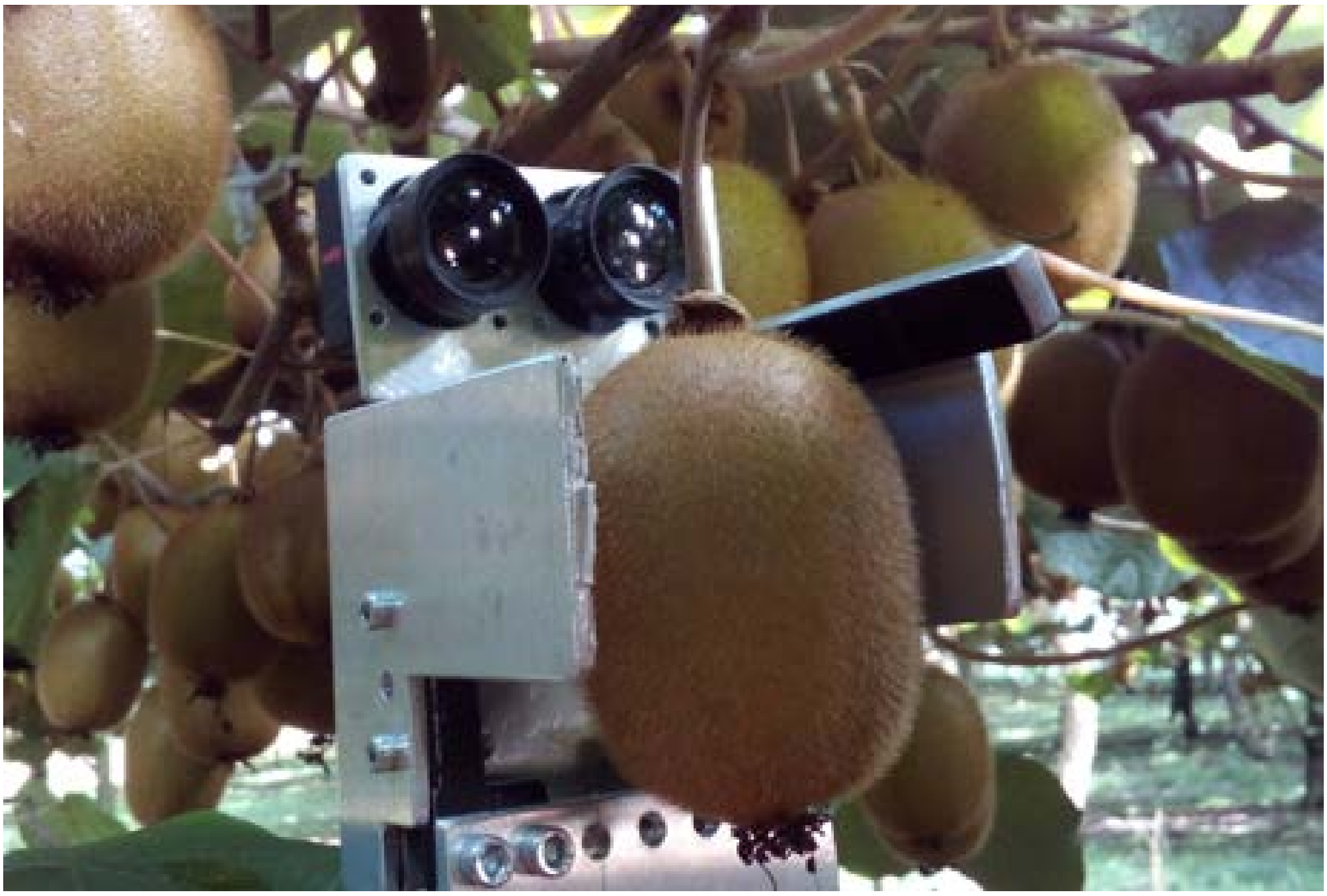

*Figure 41: The stem pushing mechanism with cameras, successfully moved into a pose for harvesting, which meant that this fruit was classified as pickable by this mechanism.*

The data gathered for the fruit pickability study is summarised in Table 14. The pickable percentage of 66 percent for the original kiwifruit harvester is similar to a result, which was earlier attained by a group of team members from the MBIE Multipurpose Orchard Robotics Project. Only the version of the stem pusher detachment mechanism without the cameras reached above 80 percent for pickability and so it was decided to proceed with the more compact stem pusher for subsequent tests of the detachment mechanism and the overall second kiwifruit harvester system. Note that the pickability study was completed without removing fruit. This aspect of the study may have affected results since fruit in clusters may move to more favourable positions after a fruit is removed.

*Table 14: Kiwifruit pickability for the second kiwifruit harvester stem pushing mechanism and the original kiwifruit harvester mechanical systems.*

| Kiwifruit Detachment Mechanism Variant | Pickable | | Not Pickable | |
|---|---|---|---|---|
| | Number | Percentage | Number | Percentage |
| Compact Stem Pushing Mechanism- No Cameras | 172 | 81 | 41 | 19 |
| Stem Pushing Mechanism with Cameras | 165 | 77 | 48 | 23 |
| Original Kiwifruit Harvester | 140 | 66 | 73 | 34 |

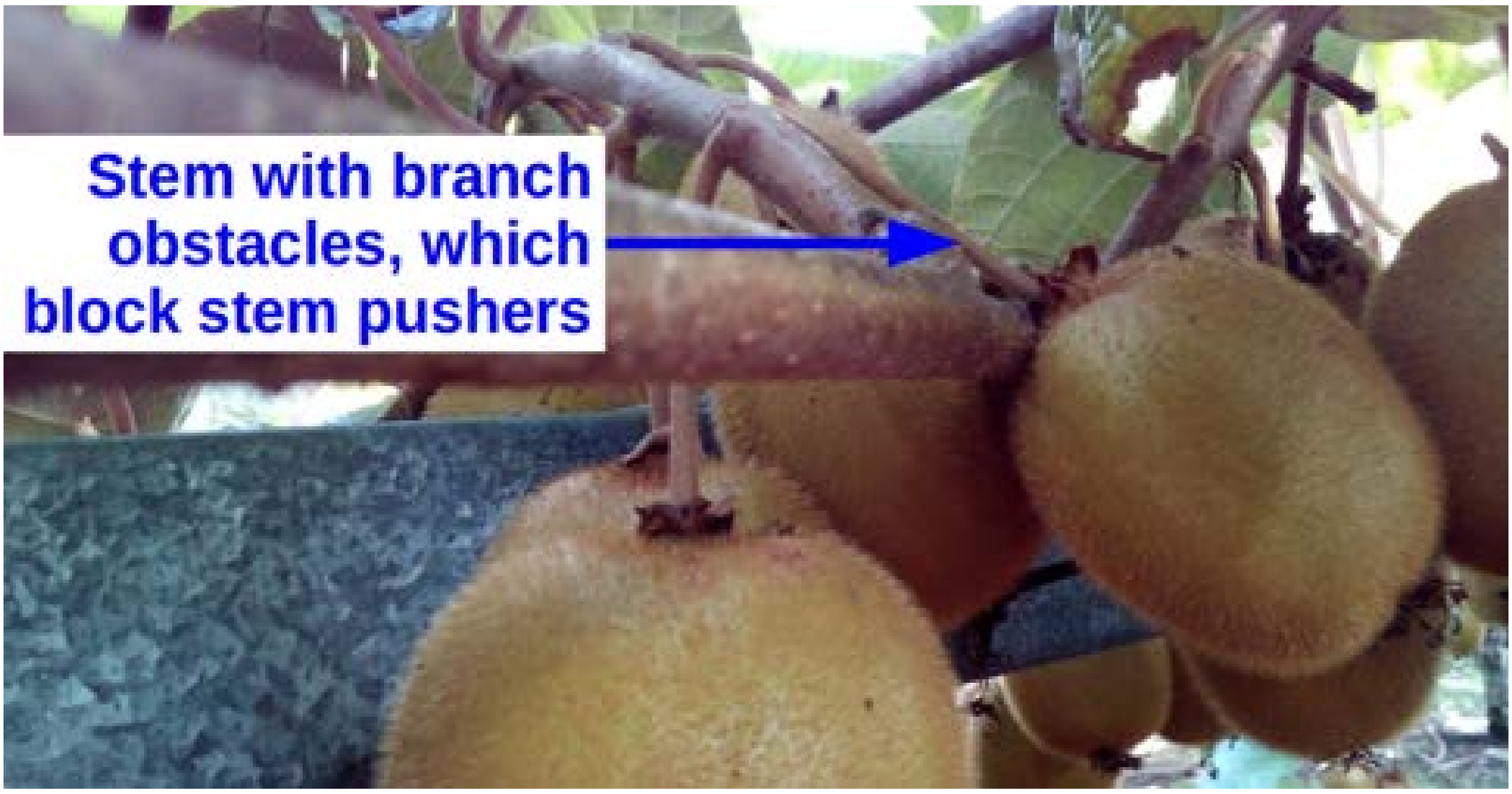

*Figure 42: An example of a stem which is blocked by a branch so that a stem pusher mechanism would fail to pick the fruit.*

Something unexpected that was discovered during the pickability study was that not every fruit that the original kiwifruit harvester could pick could also be picked by the stem pusher mechanisms. It was found that the most common cause of this was that some fruit had obstacles adjacent to the stem, which would have caused the stem pusher to fail (Figure 42). These fruit were classified as not pickable for the stem pusher mechanisms. However, it was thought that in future work some of the obstructed stems could become unblocked by moving the fruit from the resting position before actuating the stem pusher. Alternatively, for the fruit in Figure 42, it might be possible to grip and move the fruit straight down towards the ground, so that the branch adjacent to the stem acts as a stem pusher.

### 2.7.2 Testing Kiwifruit Detachment with Varying Speed and Torque

Before attempting to perform harvesting automatically, it was assumed that it would be important to determine if the stem pushing kiwifruit detachment mechanism worked at all. If the stem pusher did work, it seemed that the next step would be to ensure that the stem pusher worked reliably, so that this mechanism did not unduly affect the results of experiments with the whole second kiwifruit harvester system. However, it was unclear whether a higher speed or higher torque was required at the stem pusher. To resolve this issue, different gear ratios were tested. The gear ratios used in these experiments were 35:12 and 35:20 (Figure 23). In addition, an effective 1:1 ratio was tested by mounting the 35 tooth gear directly to the stem pusher shaft (Figure 43).

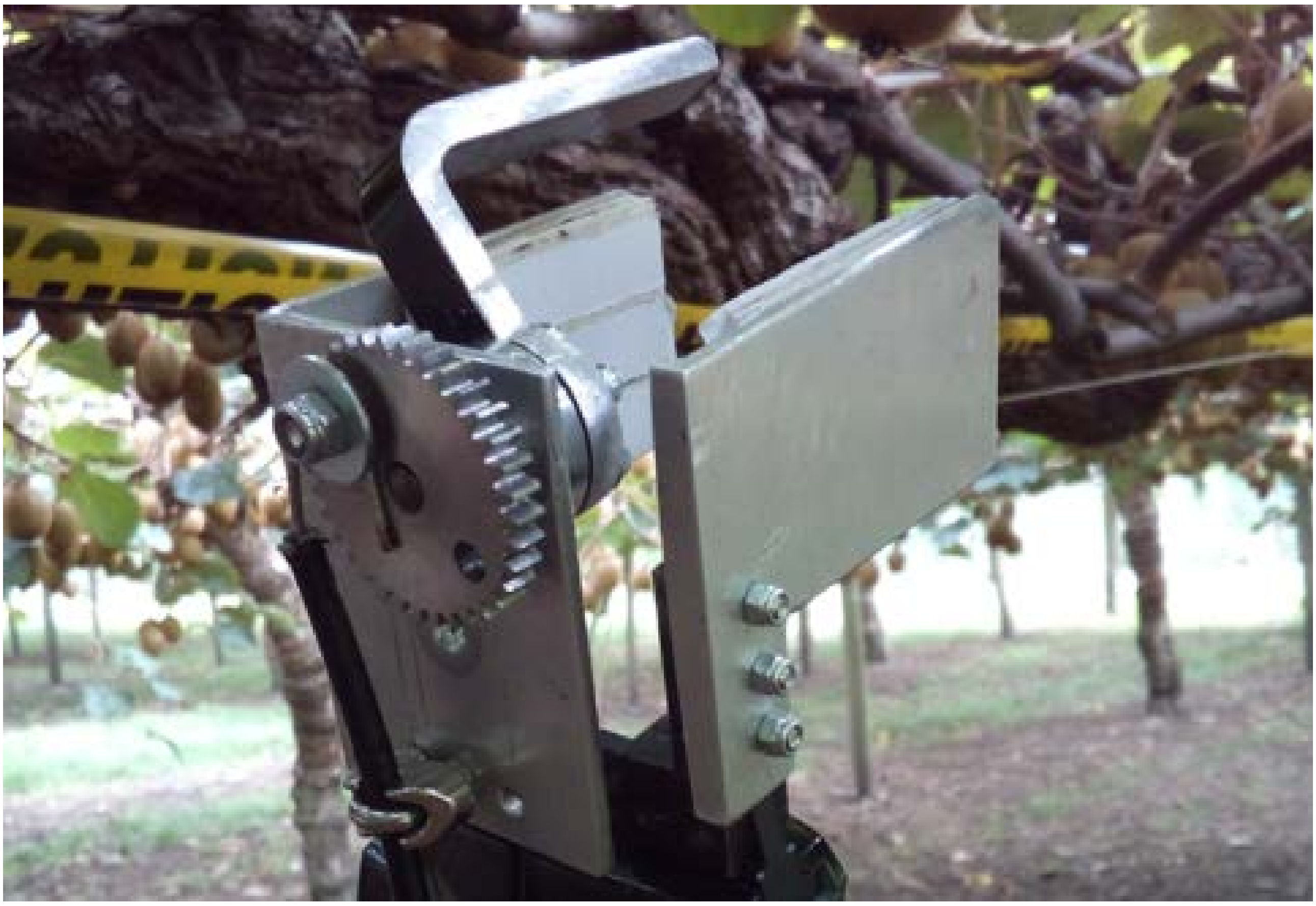

*Figure 43: Stem pushing mechanism with the 35 tooth spur gear connected directly to the stem pusher shaft.*

The steps in the test procedure for determining the best gear ratio for the stem pusher were:

1. The stem pusher was moved into position for kiwifruit picking. The distance between the pusher contact point with the stem and the stem attachment point to the fruit was set at approximately 0.01m.
2. Kiwifruit detachment was attempted by actuating the gripper and stem pusher.
3. The results were recorded in terms of successful or unsuccessful detachment.
4. Steps 1, 2 and 3 were repeated 20 times.
5. Step 4 was repeated for the different gear ratios 35:12 and 35:20 as well as with the 35 tooth gear directly coupled to the stem pusher shaft.

The results from the testing are summarised in Table 15. These results seem to be clear with the higher torque options producing more reliable detachment results. Hence, in subsequent tests of the second kiwifruit harvesting system no gearing was used and the Bowden cable was rigidly coupled to the stem pusher. Figure 44 and Figure 45 show the operation of this stem pusher configuration from the front and from behind, respectively.

Using the higher torque variant would have increased the cycle time for picking. Ideally, a mechanism with high speed and high torque would be used in order to have reliable and fast detachment. This could be achieved with a more powerful motor.

*Table 15: Results from testing the stem pushing mechanism with different gear ratios.*

| **Configuration** | **Successful Detachments** | **Unsuccessful Detachments** |
|---|---|---|
| 35:12 Gear Ratio | 2 | 18 |
| 35:20 Gear Ratio | 6 | 14 |
| 35 Tooth Gear on Pusher Shaft | 20 | 0 |

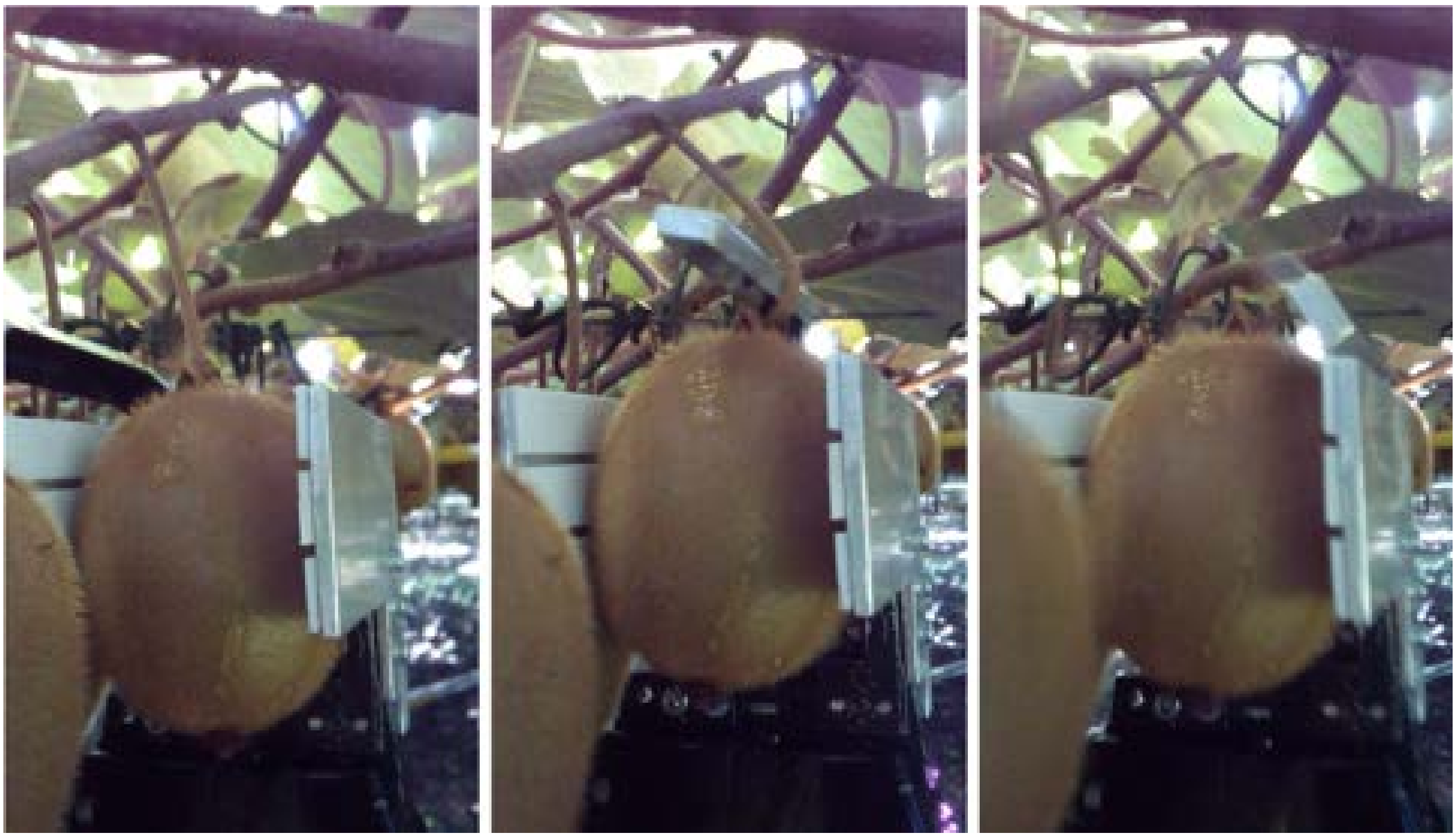

*Figure 44: Showing stem pushing kiwifruit detachment from the front.*

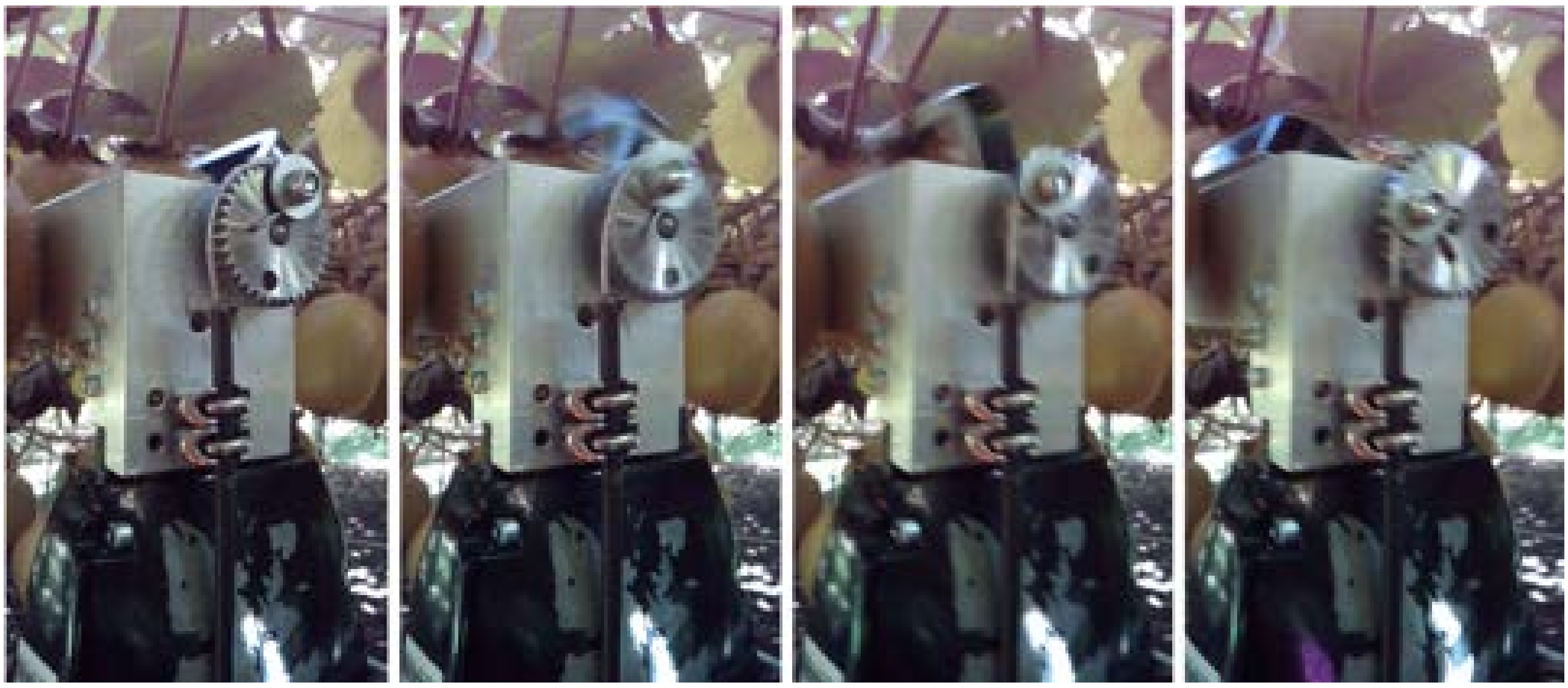

*Figure 45: Showing stem pushing kiwifruit detachment from behind.*

### 2.7.3 Kiwifruit Harvesting System Testing

The second kiwifruit harvester consisted of the stem pusher mechanism, Kuka YouBot [88] arm and Basler Time of Flight sensor, all mounted on a frame on top of a Clearpath Husky Robot [96] (Figure 46). A computer with a Nvidia GTX 1080 Ti [97] GPU was used to run the software. Most of the power came from a large 12 VDC battery and an inverter.

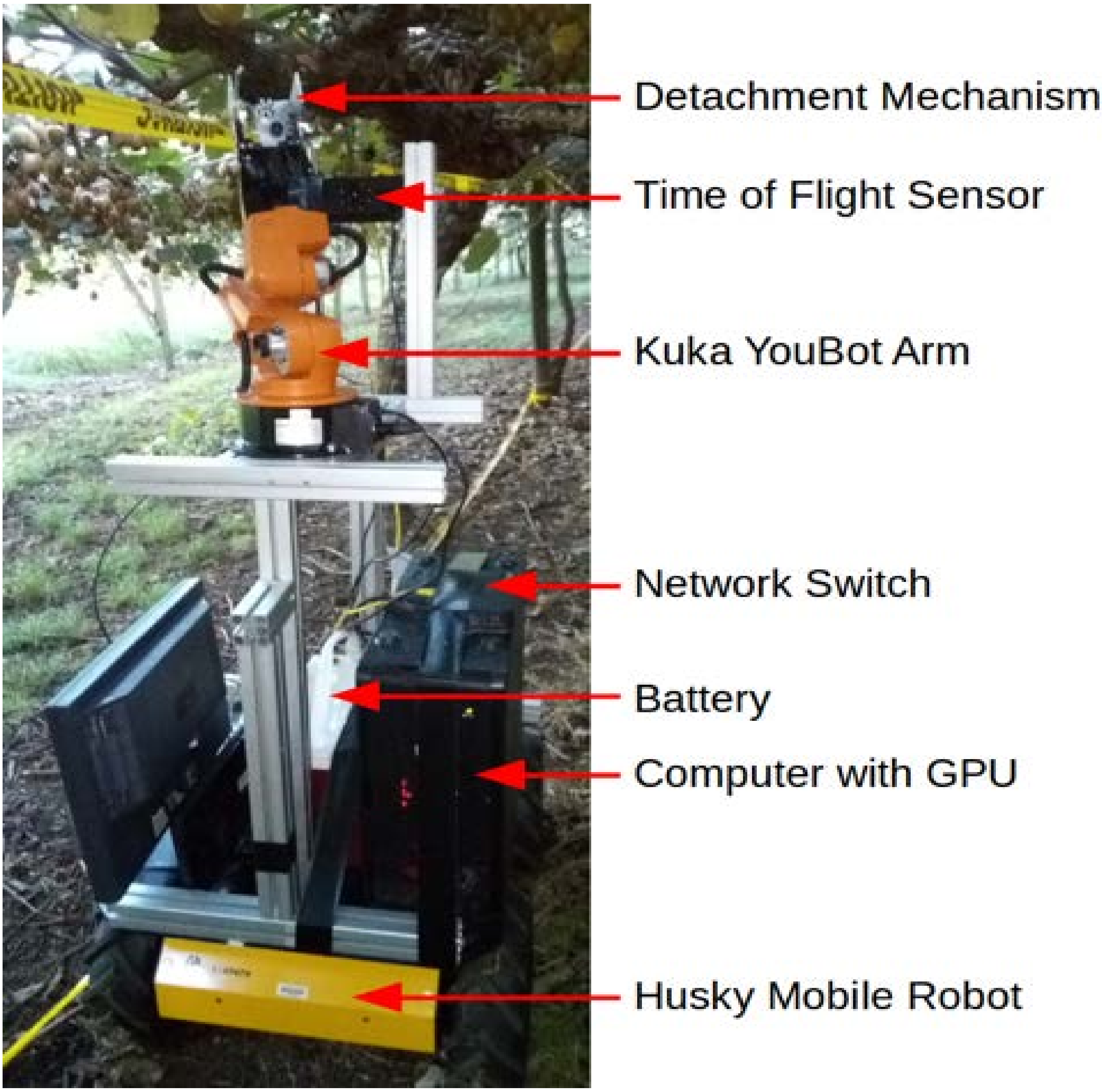

*Figure 46: Hardware components of the second kiwifruit harvester.*

The second kiwifruit harvester was tested by running the entire system in a real kiwifruit orchard. The aim of this testing was to discover the ways in which the system failed in order to direct future improvements. For every detected kiwifruit, it was noted if the arm successfully picked the fruit or not. If the targeted kiwifruit was not picked or an adjacent kiwifruit was damaged, the cause of the failure was noted. The observed failure cases were categorised as:

1. The arm collided with a solid obstacle and hence the arm did not reach the targeted fruit.
2. The detachment mechanism collided with an adjacent fruit, causing the adjacent fruit to be accidentally detached.
3. The arm target pose was too low so that the stem pusher hit the fruit instead of the stem.
4. The arm target pose was too high so that the stem pusher did not tend to shear the stem off the fruit.
5. The fruit was too small for the gripper opening so that the fruit was not held and the stem pusher rotated the fruit instead of shearing the stem (Figure 47).

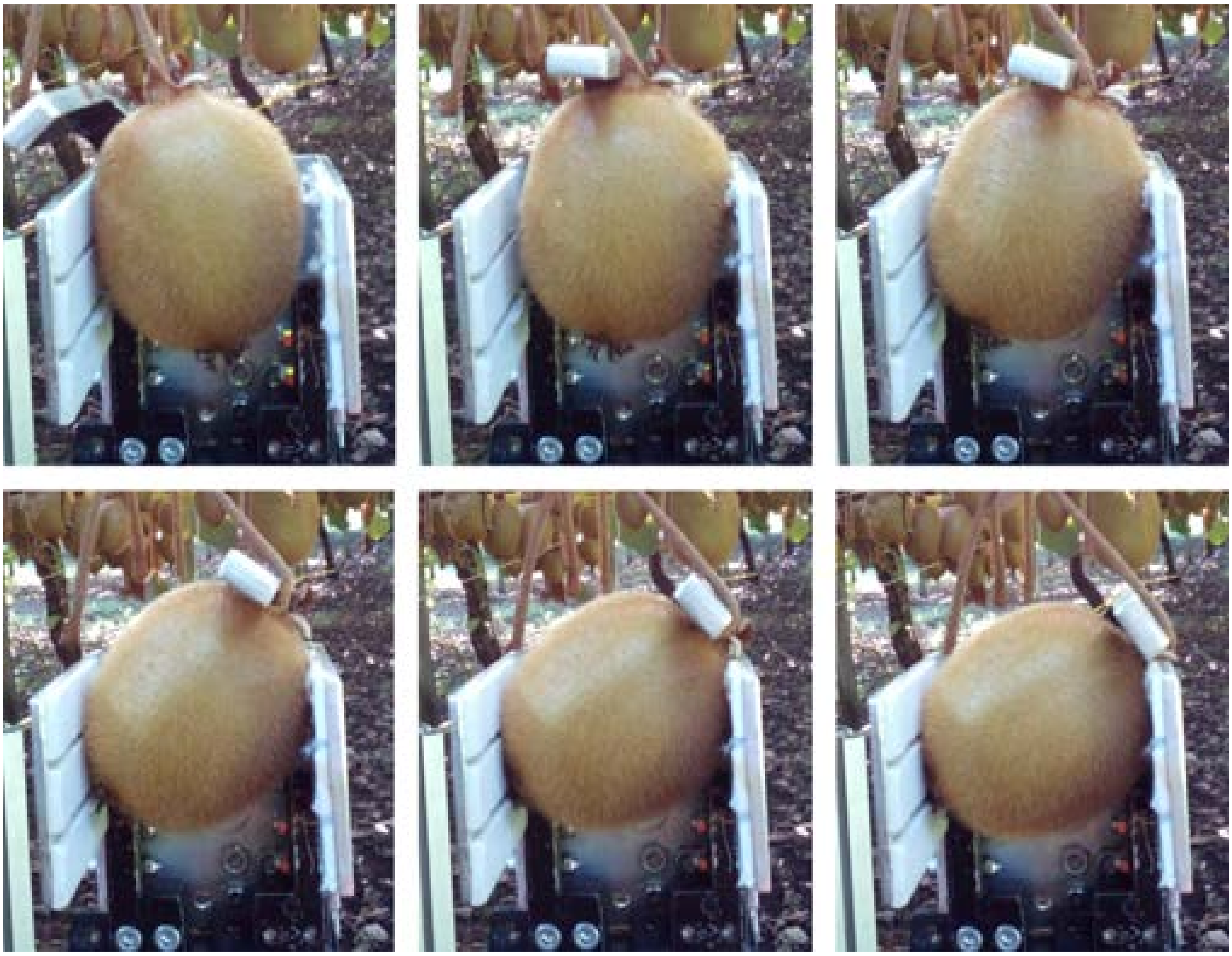

*Figure 47: Showing how a fruit, which is too small to be held tightly between the grippers, tends to be rotated by the stem pusher without detachment, instead of just having the stem move and break.*

In an initial test of the second kiwifruit harvester, 53 kiwifruit were successfully picked. Table 16 summarises the results of this testing. Out of all of the failure cases, it seemed that the target pose issues could be solved the easiest and as a result the percentile square method was developed, as described in Subsection 2.6.3. It seemed that the issue with the fruit not being held in the gripper could be overcome by using a gripper with a slightly longer stroke, that a path planner that avoided collisions could be developed and that the effects of collisions with adjacent fruit could be mitigated with a more refined gripper design; however, these measures were deferred as future work.

*Table 16: Results from initial tests with the second kiwifruit harvester.*

| **Observations** | **Number of fruit** |
|---|---|
| Fruit successfully picked | 53 |
| Arm collided with solid obstacle | 5 |
| Arm collided with adjacent fruit | 5 |
| Target pose too low | 8 |
| Target pose too high | 2 |
| Fruit not held by gripper | 5 |

The percentile square method was implemented to fix the issues of target poses being too high and too low. Then another round of testing was completed; this time with 60 kiwifruit picked. The results of this testing are given in Table 17. The percentile square method appeared to provide an improvement in terms of reducing the occurrence of bad targets. It may also have contributed to a reduction in collisions, since target poses that were too high or low could have led to collisions with adjacent objects; however, it was difficult to confirm this with observations in the field.

*Table 17: Results from final tests with the second kiwifruit harvester.*

| **Observations** | **Number of fruit** |
|---|---|
| Fruit successfully picked | 60 |
| Arm collided with solid obstacle | 2 |
| Arm collided with adjacent fruit | 4 |
| Target pose too low | 2 |
| Target pose too high | 1 |
| Fruit not held by gripper | 5 |

An observation that was made during the testing of the second kiwifruit harvester was that this system could pick fruit that the original kiwifruit harvester could not. The pickability study had shown that the second kiwifruit harvester mechanical components could reach and pick fruit that the original kiwifruit harvester could not reach and pick. However, the pickability study did not show that the second kiwifruit harvester control system, including the detection and arm control systems, could actually pick fruit that were obstructed for the original kiwifruit harvester.

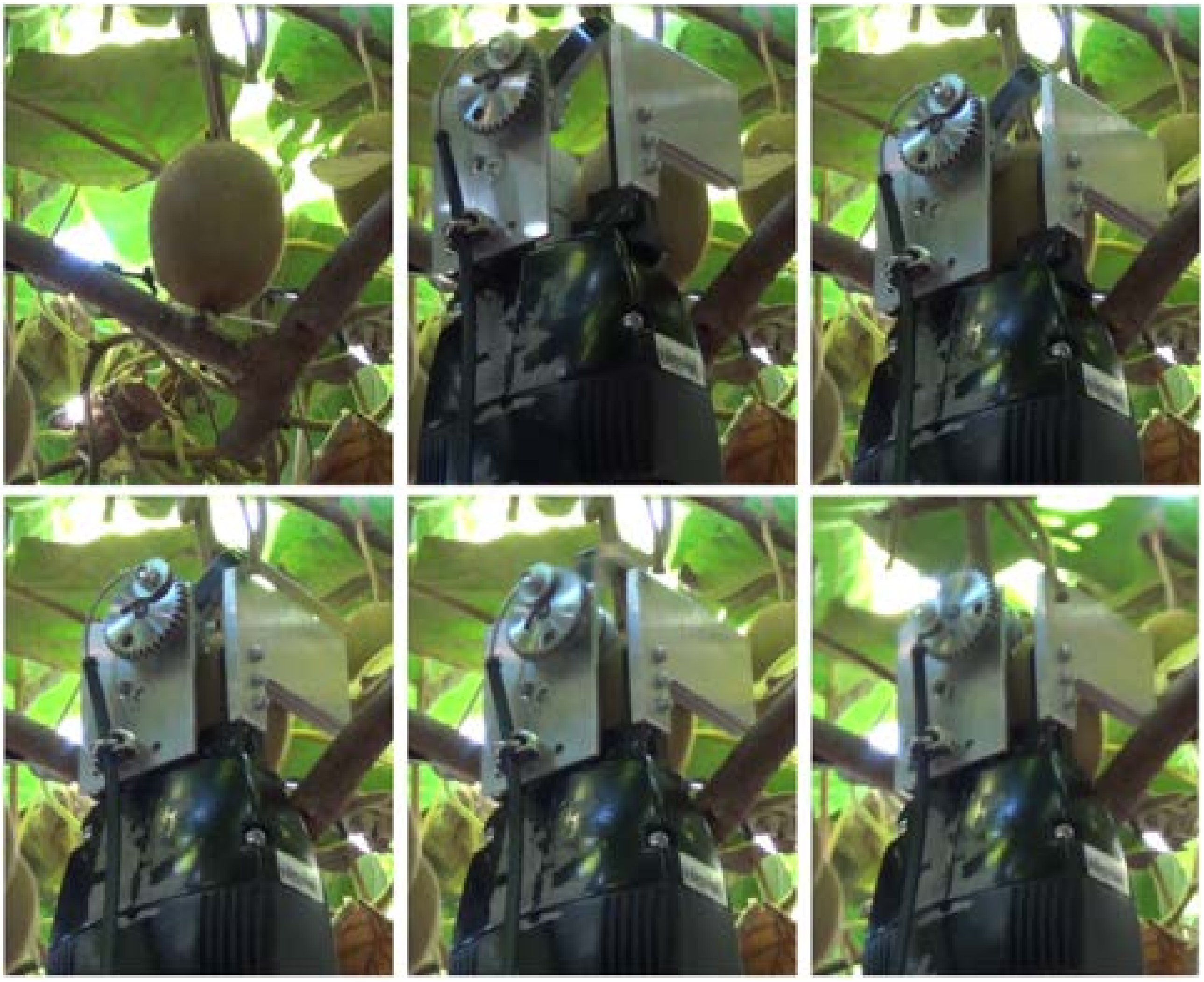

*Figure 48: These images show a kiwifruit that is obstructed from below, which the second kiwifruit harvester can detect, reach and pick.*

Figure 48 shows a kiwifruit that was obstructed from below. It is possible that the original kiwifruit harvester would not have been able to detect this fruit since the stereo cameras on the original kiwifruit harvester faced upwards; however, the second kiwifruit harvester can detect such fruit because of the angle of the mounting of the Time of Flight sensor. In addition, the second kiwifruit harvester arm control system allows for moving up to a pose that is adjacent to the target before reaching outwards, which allows the detachment mechanism to be moved above the branch; in contrast the original kiwifruit harvester would move upwards towards the fruit from below and would hit the branch in this situation. Furthermore, the shape of the gripper paddles and stem pusher on the second kiwifruit harvester allows this fruit detachment mechanism to reach above the obstructing branch, which allows the fruit to be picked.

## 2.8 Robotic Kiwifruit Harvesting Discussion

At the beginning of the MBIE Multipurpose Orchard Robotics Project, it seemed that hand engineered algorithms for fruit detection, which had colour thresholding steps in the algorithms, might work well despite lighting variations, since there were so many mitigating factors. These mitigating factors included limited variability in kiwifruit canopies, strong artificial lighting and shading from the kiwifruit canopy. However, the existing kiwifruit detection algorithm tested produced many false positives and required retuning of the colour thresholds. This led to an enhancement of this existing algorithm, using an AlexNet [32] CNN, which significantly reduced the false positives rate and the need for retuning. Following this, many CNNs were trialled for kiwifruit detection, including DetectNet [44] for object detection, fully convolutional networks for semantic segmentation [45] and Mask R-CNN [48] for instance segmentation. Mask R-CNN produced high Average Precision and Average Recall values in testing and was used to pick fruit on the second kiwifruit harvester. The use of Mask R-CNN for fruit detection may not be unique; however, the way it was used to detect just collision free and reachable fruit in Time of Flight data seemed novel and worked well in testing. In addition, the hybrid Scarfe [21]/ AlexNet [32] detection method was a novel algorithm, which produced comparable fruit picking results to the other later developed CNNs tested. Although these methods worked well, there would nevertheless be challenges with using these vision systems in the presence of extreme ambient lighting conditions from direct sunlight; however, it is expected that on a smaller mobile platform, the harvesting systems could be moved to different poses for different viewpoints of the canopy and hence the vision systems might be moved to avoid direct sunlight.

It was genuinely surprising that the Basler Time of Flight sensor worked as the sole sensor for kiwifruit harvesting in the real world. Previous experience with other active 3D sensors outdoors had led to the belief that this type of sensor would not work. In fact, as later described in Subsection 4.2.4, the Basler Time of Flight sensor was rejected for navigation. However, under the shade of a well developed kiwifruit canopy, it seemed to work well for detecting kiwifruit, to the extent that kiwifruit were successfully picked with the second kiwifruit harvester. The benefit of using the Time of Flight sensor was that multiple points from the surface of fruit could be accurately measured; whereas with stereo cameras, it is common to get mismatches, as is explored further in Subsection 3.5.5. Because the Time of Flight sensor worked well under the kiwifruit canopy, it is possible that it might also work well in other pergola structured orchards. However, this same sensor was tested in apple orchards and resulted in complete loss of depth measurements, even in overcast conditions, which suggests that Time of Flight sensors may not be useful in orchards with tree wall structures in daylight.

Multiple novel kiwifruit detachment mechanisms were proposed for the second kiwifruit harvester. Ultimately, only the stem pusher concept was manufactured; although, two versions of this concept were created: a version with two cameras onboard and a smaller version with no cameras. In the kiwifruit pickability study, only the smaller version of the stem pusher could have potentially picked more than 80 percent of the fruit in the area examined. It was found that the stem pusher benefited from higher torque but did not work so well when the fruit was not held firmly in place by the gripping paddles. In future versions of this mechanism, it is recommended that a gripping mechanism with a longer stroke should be used so that all kiwifruit can be firmly held; the stroke of the Kuka YouBot gripper was only 0.02 m.

The size and shape of the stem pusher kiwifruit detachment mechanism allowed it to reach fruit that were in tight spaces behind solid obstacles; previous robotic kiwifruit harvesters have not had this ability. In addition, the mounting pose of the sensor and the strategy for path planning, allowed for fruit to be picked behind solid obstacles. It was also found that detecting a point higher on the kiwifruit led to more accurate stem pusher placement for picking. Overall, the second kiwifruit harvester picked 74 percent of the fruit that it interacted with and this could be increased to 79 percent, just by using a gripper mechanism with a longer stroke. The second kiwifruit harvester shows potential for successful commercial operation and might be used alongside mechanisms like those from the original kiwifruit harvester, in order to exploit the strengths of both systems.

Other successes from the second kiwifruit harvester included the calibration method, which worked with an accuracy of 0.0024 m. The use of an arm with more degrees of freedom gave advantages to the second kiwifruit harvester in terms of being able to pick fruit that would have been obstructed for the original kiwifruit harvester. Finally, as a whole robot composed of novel software and hardware components, the second kiwifruit harvester was able to pick kiwifruit in a real kiwifruit orchard and hence shows potential for future development into a commercial system. This system need not be restricted to just kiwifruit harvesting. The same technology may also be applied to very similar tasks, such as kiwifruit thinning, but might also be applied to less similar tasks, such as apple thinning or tomato picking, where a stem pushing action is used by human pickers. Hence, the potential for this technology extends to multiple applications beyond those explored so far.

## 2.9 Robotic Kiwifruit Harvesting Conclusions

The contributions presented in this section have been the vision systems for the original kiwifruit harvester and all hardware and software for the second kiwifruit harvester. From experimenting with these systems, the following conclusions have been made:

- Convolutional neural networks can be used for detection of objects in kiwifruit canopies and, in particular, can be used to detect unobstructed fruit in a range of interest, with the use of bespoke labelling methods.
- A fruit gripping and stem pushing mechanism can achieve 100 percent detachment for well posed fruit when using higher torques and can be put into a pose for successful picking of 81 percent of fruit in a kiwifruit canopy.
- A hooked mechanism shape, a Time of Flight sensor viewing fruit from the side and a path planning method, incorporating a reaching motion at the end of a path, can all be combined to harvest fruit that are obstructed from below, which would be difficult to pick for previous robotic kiwifruit harvesters.

## 2.10 Future Work for Robotic Kiwifruit Harvesting

Collisions were a significant unresolved issue, observed during the system testing of the second kiwifruit harvester. Hence, it is recommended, that in future versions of the kiwifruit harvester, the path planner should incorporate collision detection and avoidance. The path planning strategies presented here may be used as one of many heuristics to speed up a future path planner. Further measures for avoiding collisions might include detecting and targeting the stem attachment point of the fruit. Also, mounting the sensor on the base joint of the robot arm, in order to keep the sensor in alignment with the other joints of the arm, would allow the detection of unoccluded fruit to work better for collision detection and avoidance.

Despite the results of the fruit pickability study, having sensors in close proximity to the stem pusher seems that it could offer many benefits including: the ability to target the stem to fruit attachment point, improvements in the accuracy of fruit targeting, detection of more fruit and detection of more obstacles. In general more sensors in different poses could have similar benefits. For sensors close to the stem pusher, in particular, smaller options might be investigated in order to not reduce fruit pickability so much.

A feature that was conceived but not implemented was detecting the size of fruit so that the gripper paddles would only be opened wide enough to accept the targeted fruit. In addition, some algorithms for collision detection and fruit pose measurement were started but not finished, including those in Section 9.

Other potential improvements include having a robot arm with more degrees of freedom and a larger working area. Also, shorter gripping paddles and a shorter stem pusher might improve the fruit pickability and reduce the probability of collisions.

Future testing of the second kiwifruit harvester should be more thorough by including the assessment of multiple metrics. Included in these metrics should be conventional factors, such as speed and the proportion of fruit successfully picked or lost. In addition, the height of the fruit picked might also be recorded so as to understand how well the robotic harvester picks low hanging fruit compared to fruit higher in the canopy. Furthermore, the offset of the fruit picked from the row centreline should be considered since it was observed that there were less obstructed fruit closer to the row centreline. In addition, each fruit in the targeted experimental area of the canopy should be assessed for fruit quality in order to understand how much damage the robotic harvester causes during picking. This step could be extended to assess the maturity of the fruit, since the maturity of the fruit affects how readily they detach and damage, which are factors that could affect the results when testing a robotic harvester.

# 3 Robotic Kiwifruit Flower Pollination

In addition to harvesting kiwifruit, performing automated targeted artificial pollination of kiwifruit flowers was a goal of the MBIE Multipurpose Orchard Robotics Project. It was intended that during pollination season, a pollination module would be deployed on the AMMP (Figure 2). Then during the harvesting season, the pollination module would be replaced on the AMMP by the harvesting module (Figure 4).

The need to develop an automated targeted artificial pollination system for kiwifruit flowers stems from the issues with existing means of pollination for kiwifruit. Current methods for pollinating kiwifruit flowers in orchards include using bees, pollinating individual flowers by hand and spraying clouds of pollen using machines. Bees can be effective pollinators for producing kiwifruit [98], [99]. However, flowers get visited by bees at different rates and some flowers may not be visited at all [98], [99]. In addition, kiwifruit flowers are a poor source of nutrition for bees [100] and bees cannot be controlled to pollinate kiwifruit flowers; so instead bees may favour other flowers in the local area [101]. Furthermore, on wet days the bees may stay in their hives, regardless of whether the kiwifruit flowers are open or not [101].

To overcome these issues with bees, pollen can be applied by hand in order to produce better quality fruit [102], [103]. Pollen can be mixed into solution, which is hand sprayed directly onto flowers [104], or pollen may be spread by hand by direct contact with flowers [102]–[104]. However, in both of these cases, hand pollination is a time-consuming and costly method of pollination.

Using products like the Quad Duster [105] or Pollensmart [106], dry pollen may be blown onto the canopy, while driving a quad-bike or tractor. Trials have also been performed for a similar approach with pollen in solution [107]. However, these methods of pollination are very wasteful since pollen is not applied directly to the flowers [108], [109].

## 3.1 Existing Robotic Pollination Systems

The issues with bees, hand pollination and existing machine pollinators have prompted the development of targeted robotic pollinators. Wood et al. [110] described their development of small flying robots that mimicked bees and were intended to perform tasks like pollination. However, these robots did not have a vision system for flower detection and such robots were only capable of a few minutes of untethered flight time [111]. In addition, wind is a significant issue for such a small flying robot and experiments have only demonstrated wind compensation for gentle gusts [112].

Chechetka et al. [113] used a small quadcopter to pollinate flowers. The contact surfaces tested to distribute the pollen were different types of fibres- including horse hair, nylon fibres and carbon fibres- which were either coated or not coated in Ionic Liquid Gel (ILG). Besides the carbon fibre, which did not hold any pollen, the coated fibres held more pollen than the uncoated fibres and the nylon fibres held more pollen than the horse hair. A patch of ILG coated horse hair was attached to the small quadcopter. The quadcopter was flown so that the horse hair contacted the stamen of one flower and then the pistil of another flower. However, this experiment was performed with a person operating a remote controller; the robot did not have its own navigation, detection or control systems [113], [114].

Potts et al. [115] argue that robotic drones are an inefficient pollinator. Williams [116] highlights that there are large technical challenges with flight time, navigation systems and detection systems on a small flying pollinator robot. For such reasons, it may seem that ground based robots may be a better near term solution for automated targeted artificial pollination.

A ground based robotic pollinator was developed by Ohi et al. [117], who described a mobile robot called BrambleBee, which used a robot arm with a custom flexible end effector to transfer pollen from a flower’s stamen to the same flower’s pistil. Wide angle detection of flowers was performed by a camera with a fisheye lense and then more refined localisation of flowers was performed using a SoftKinect Depthsense 525 Time of Flight camera. In a demonstration of BrambleBee [118] the robot only interacts with QR codes once in 30 seconds, whereas the robots described later in this section pollinate many more real flowers in this length of time by spraying pollen in solution rather than using robot arms, with time spent path planning and following. The task considered by Ohi et al. [117] is different to the problem considered here in that they address the pollination of berries that grow on bramble structured plants indoors, whereas the MBIE Multipurpose Orchard Robotics Project addressed the problem of pollinating kiwifruit on pergola structured vines outdoors. Kiwifruit present some different challenges since pollen must be transferred from male plants to female plants; as opposed to transferring pollen from one part of a flower to another part of the same flower. In addition, kiwifruit flowers require high pollen counts for large fruit and so enough pollen must be delivered for a commercially viable crop [17]; however, kiwifruit pollen is expensive and so it is important to not deliver too much pollen to female flowers.

Goodwin and Martinsen described RoboBee [108], [109], which was a predecessor of the robot pollinators developed in the MBIE Multipurpose Orchard Robotics Project. RoboBee detected flowers using optical sensors, with illumination provided by lasers. Detected flowers were sprayed with pollen as the RoboBee was manually driven through an orchard by an operator.

## 3.2 System Descriptions

The goal for the robotic pollination system was to deliver pollen to as many of the kiwifruit flowers as possible, while minimising the wastage of the expensive pollen. Four systems were developed in order to meet this goal. In chronological order, these systems were the following:

- An indoor test rig, which was pushed by hand, detected disks in a mock canopy and fired lasers to hit the disks (Figure 49). The disks were detected using colour thresholding and the position of the disks was determined by stereopsis. Overall, the system was designed for ten cameras, 144 lasers and two wheels with encoders for odometry. This system is referred to here as the “indoor pollinator”. The indoor pollinator was developed in order to allow testing of algorithms outside of the 4 week pollination season. My contribution to this system was in the calibration of the different coordinate systems.

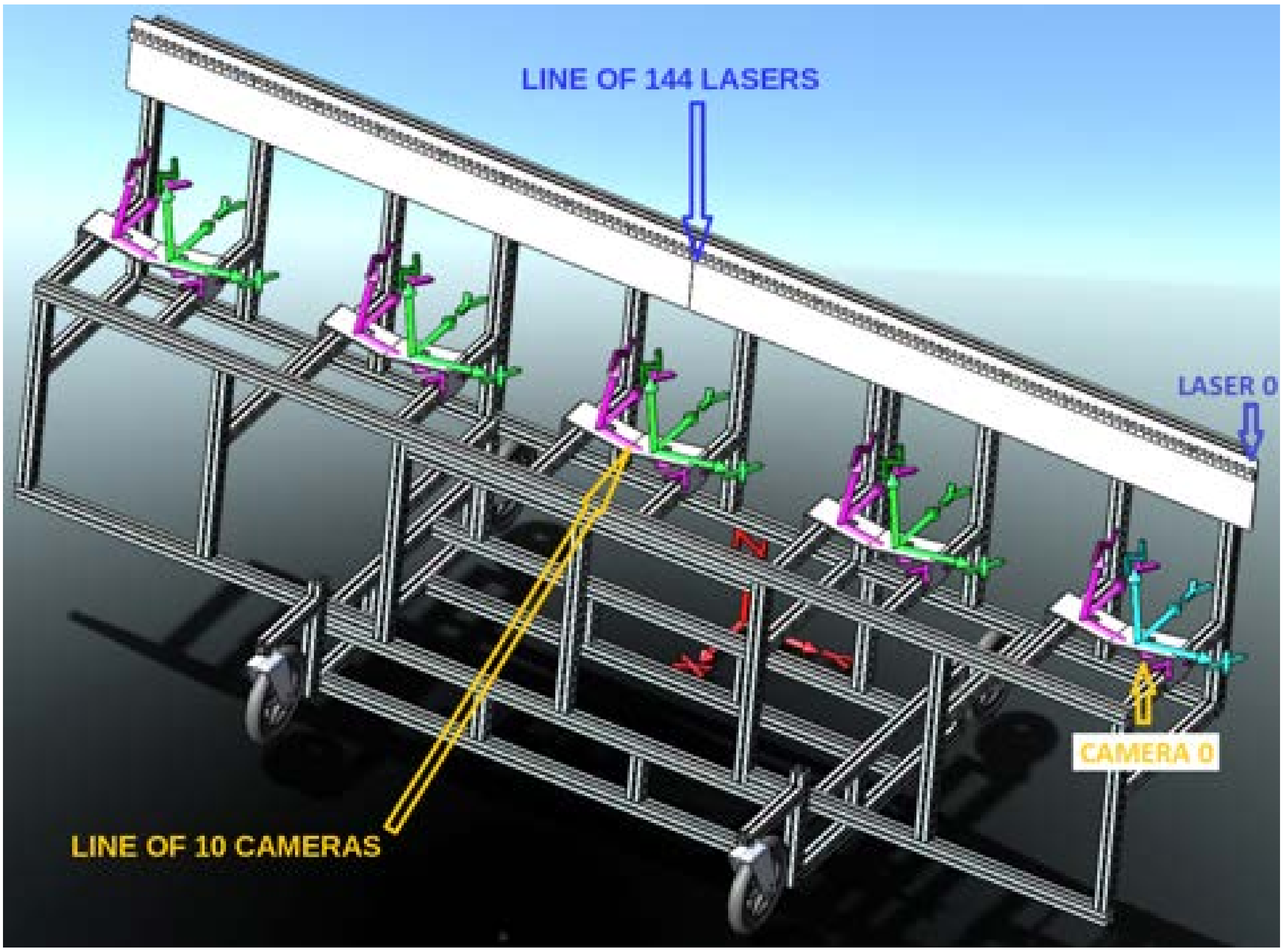

*Figure 49: A drawing of the indoor pollinator test rig with coordinate systems shown for the 10 cameras (magenta, green, cyan) and the overall system (red).*

- A single 2.1 m wide rigid spray boom was made, which detected and sprayed flowers (Figure 50). As with the indoor pollinator, this system used stereo cameras for detection and used wheels with encoders for odometry. The way the pollen was delivered was by mixing the pollen with water and pressure spraying the resulting solution through spray nozzles. The spray nozzles were arranged side by side in a line, much like the lasers on the indoor pollinator. It was intended that this spray boom would be actuated up and down in order to get close to the canopy and put the sprayers within the sprayers' range of the flowers. However, this movement functionality was never developed and so this boom was actually kept at a fixed height above the trailer that it was mounted on. This system is referred to here as the "big pollinator". My contribution for the big pollinator was in the calibration of coordinate frames, the design of the computer vision hardware, the refinement of the computer vision software and the integration to get the automated targeted pollination working.

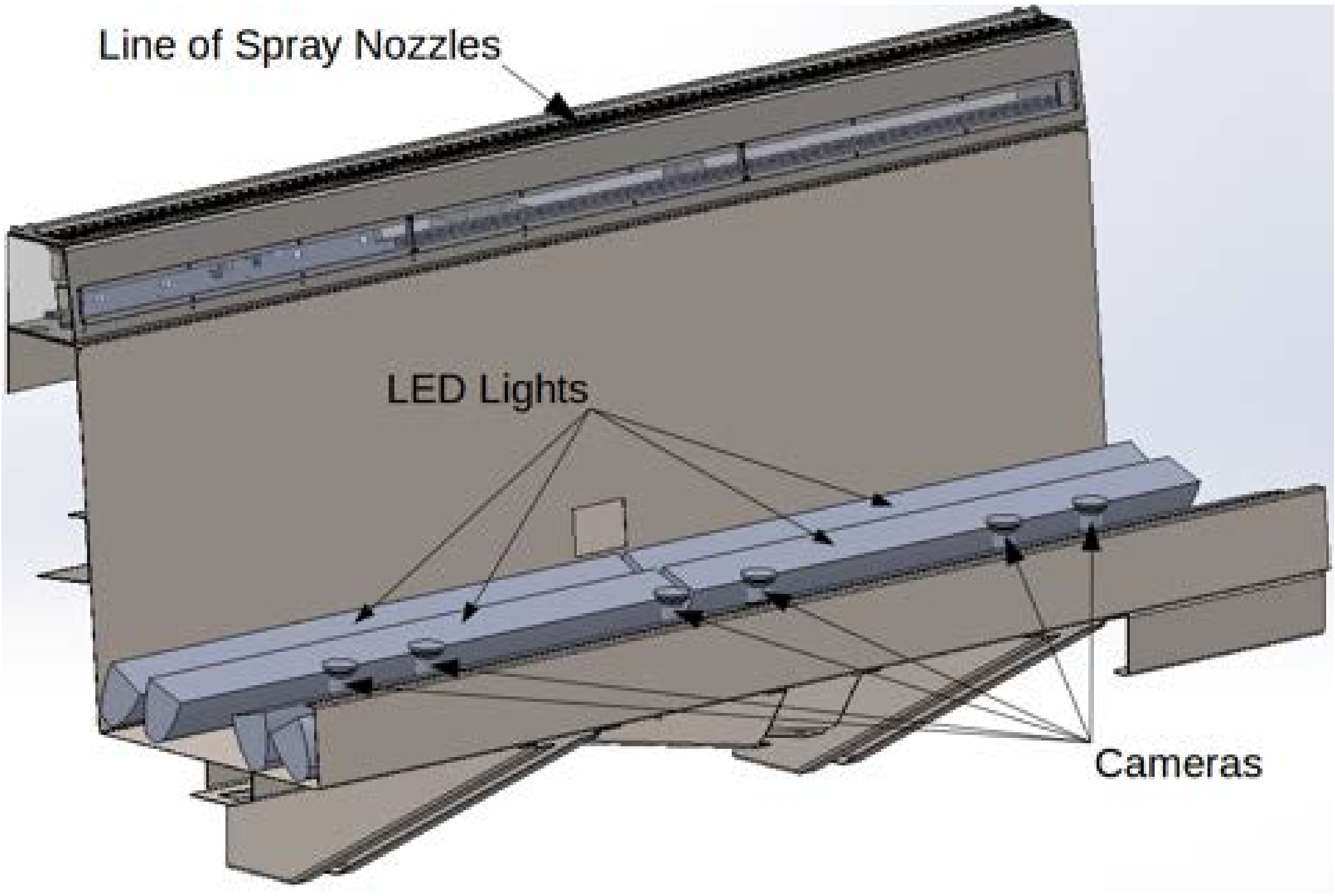

*Figure 50: Drawing of the key parts on the spray boom of the big pollinator test rig.*

- After the big pollinator, a different approach was taken of using more booms, which were smaller in size. Instead of having one long 2.1 m boom section, there were 2 independent shorter boom sections of only 0.5 m width each (Figure 51). Whereas the 2.1 m boom section had remained at a fixed height, the shorter boom sections were both raised and lowered, covering a full stroke of 0.4 m in 1 s. This system is referred to here as the "dual pollinator”. My contribution for this system was in the computer vision software and in the navigation of the boom sections.

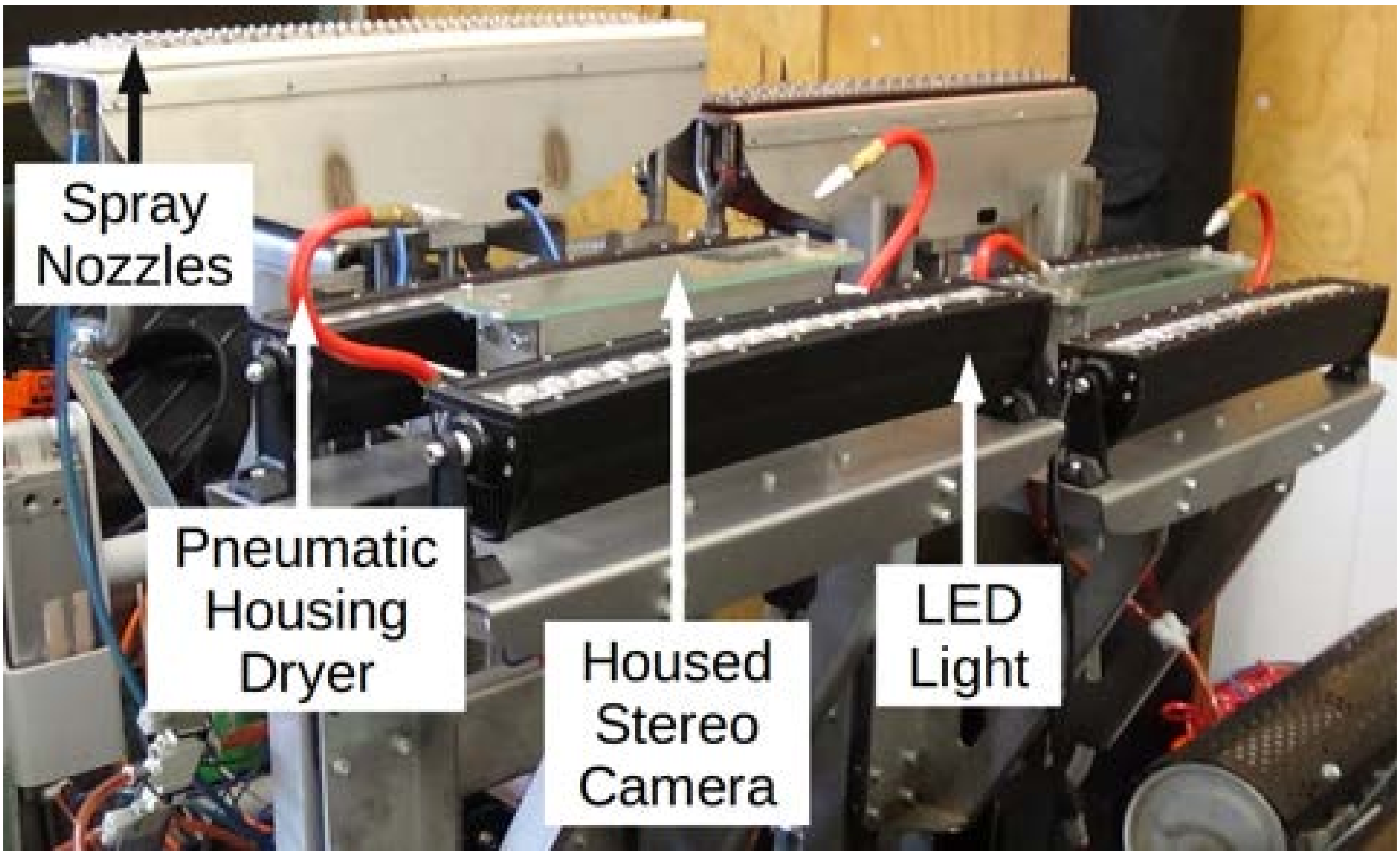

*Figure 51: Some key parts of the dual pollinator system.*

- I also created a robot arm dry pollination system (Subsection 3.6). This system is referred to here as the “dry pollinator”. My contributions for this system were all of the hardware, software and testing.

## 3.3 Flower Detection System

The goal of kiwifruit flower detection was to accurately determine the position of flowers in an image. For all of the pollinators except the dry pollinator, the 3D position of the flowers in the real world was then determined using stereopsis, in order to perform targeted spraying of pollen onto the flowers.

Flower detection had a more stringent real time processing requirement than kiwifruit detection. Kiwifruit detection processing could be performed with a period in the order of seconds on the original kiwifruit harvester without serious effects on the operational efficiency. In contrast, flower detection had to be performed within tens of milliseconds for reasonable performance of the pollination systems.

The goal was to perform pollination while travelling at at least 1.0 $ms^{-1}$ but up to 1.4 $ms^{-1}$. Flower detection had to output results at least before the sprayers had passed the flowers but ideally frequently enough for there to be multiple viewpoints of a given section of canopy. It was thought that this higher frequency might allow the cameras to capture a viewpoint of those flowers that were occluded from some other viewpoints.

### 3.3.1 Sensor Selection and Placement for Flower Detection

The cameras selected for pollination on the big pollinator were Basler Ace acA1920-40 colour cameras [75]. The key specifications that led to the selection of this camera were the resolution at 1920 pixels by 1200 pixels, the frame rate at 41 frames per second, a hardware trigger for synchronisation within a stereo pair and a global shutter.

The lense was selected based on the required field of view width. At a distance of 0.75 of a metre, each stereo pair had to view approximately a width of one third of the spray boom width, which was 2.1 m wide. The cameras in a stereo pair were separated by 0.18 of a metre. Based on these figures, an 8 mm focal length lense was selected, which was a Kowa LM8HC [119].

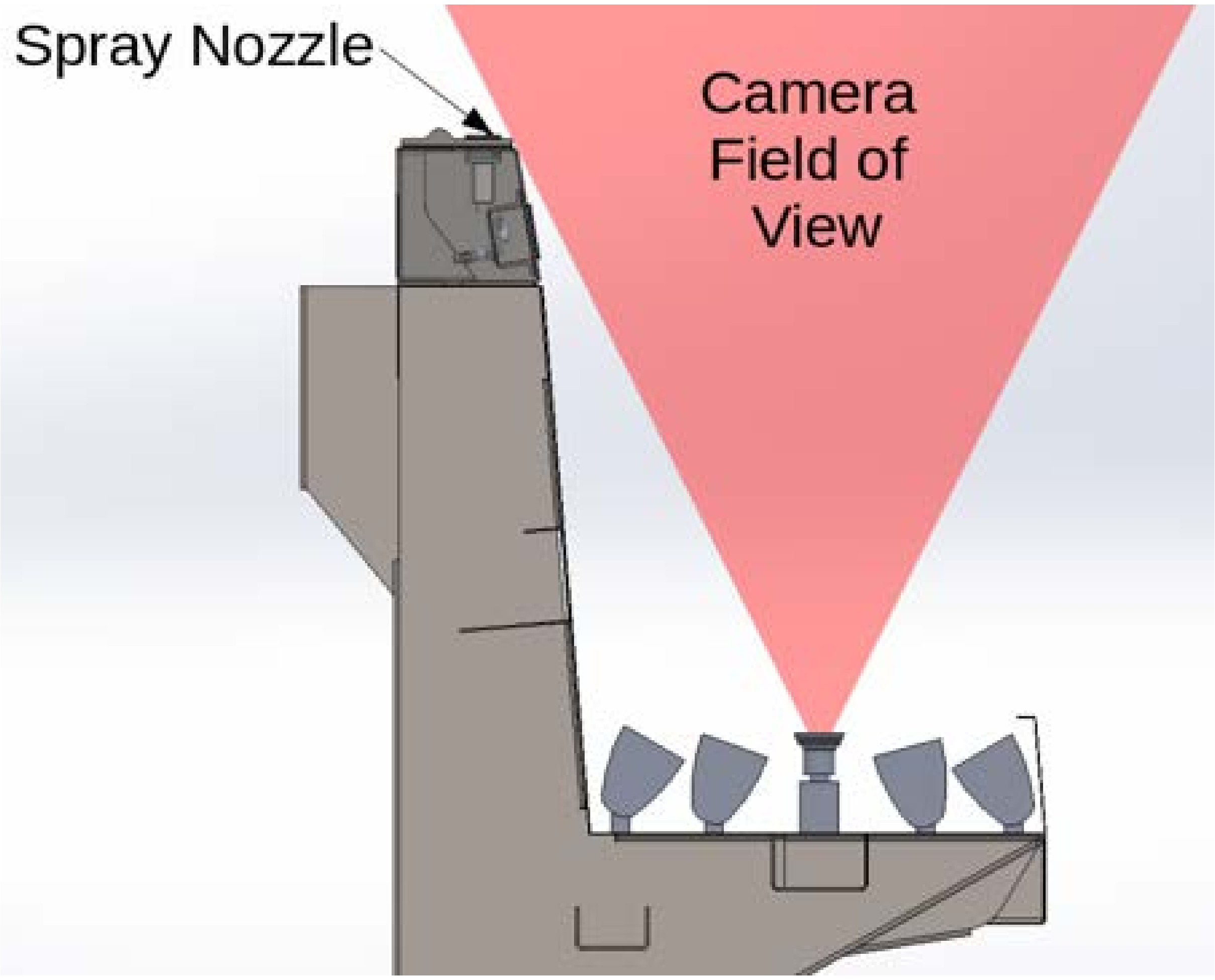

*Figure 52: Showing the camera field of view in relation to the spray nozzles on the big pollinator.*

The cameras were placed so that the field of view of the cameras would encompass the space above the spray nozzles, in the way shown from side on in Figure 52. This allowed a flower at a height of less than 0.1 of a metre directly above a nozzle to be detected when the pollinator was stationary. This was deemed to be a useful design guideline because it was assumed that the big pollinator would be no closer than 0.1 of a metre from the flowers, which it was spraying- which proved to be the case in field trials.

### 3.3.2 Colour Thresholding of Flowers

Initial attempts at detection of flowers used colour thresholding. Although, it was well known that colour changes with different lighting, it was hypothesized that the colour differences would be small enough to enable colour thresholding to work because of the following mitigating factors:

- The kiwifruit canopy is a constrained environment with only a limited amount of variability and a limited palette of colours.
- The canopy is viewed from the shaded side during pollination and the ground under the canopy is shaded, limiting the amount of ambient light and limiting the effect of ambient light on the canopy colour.
- Large amounts of artificial lighting was used. For example, on the big pollinator, approximately 400 Watts of LED lamps were used per metre width of canopy.

Before using colour thresholding for flower detection, the thresholds were manually tuned in the orchard. Colour thresholding was successfully used to detect and spray flowers (Figure 53). However, it was found that the system required retuning throughout a day.

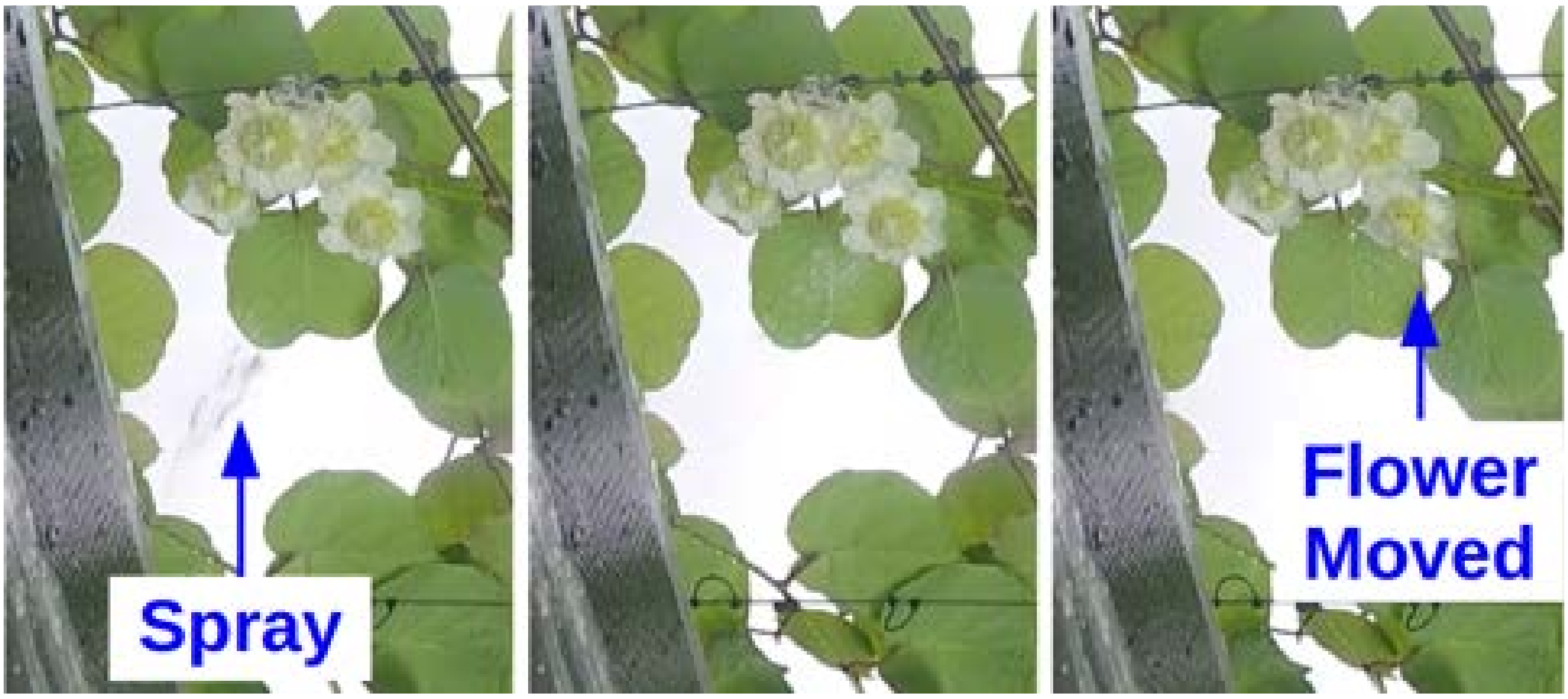

*Figure 53: A sequence showing the spray leaving the spray boom and hitting a flower.*

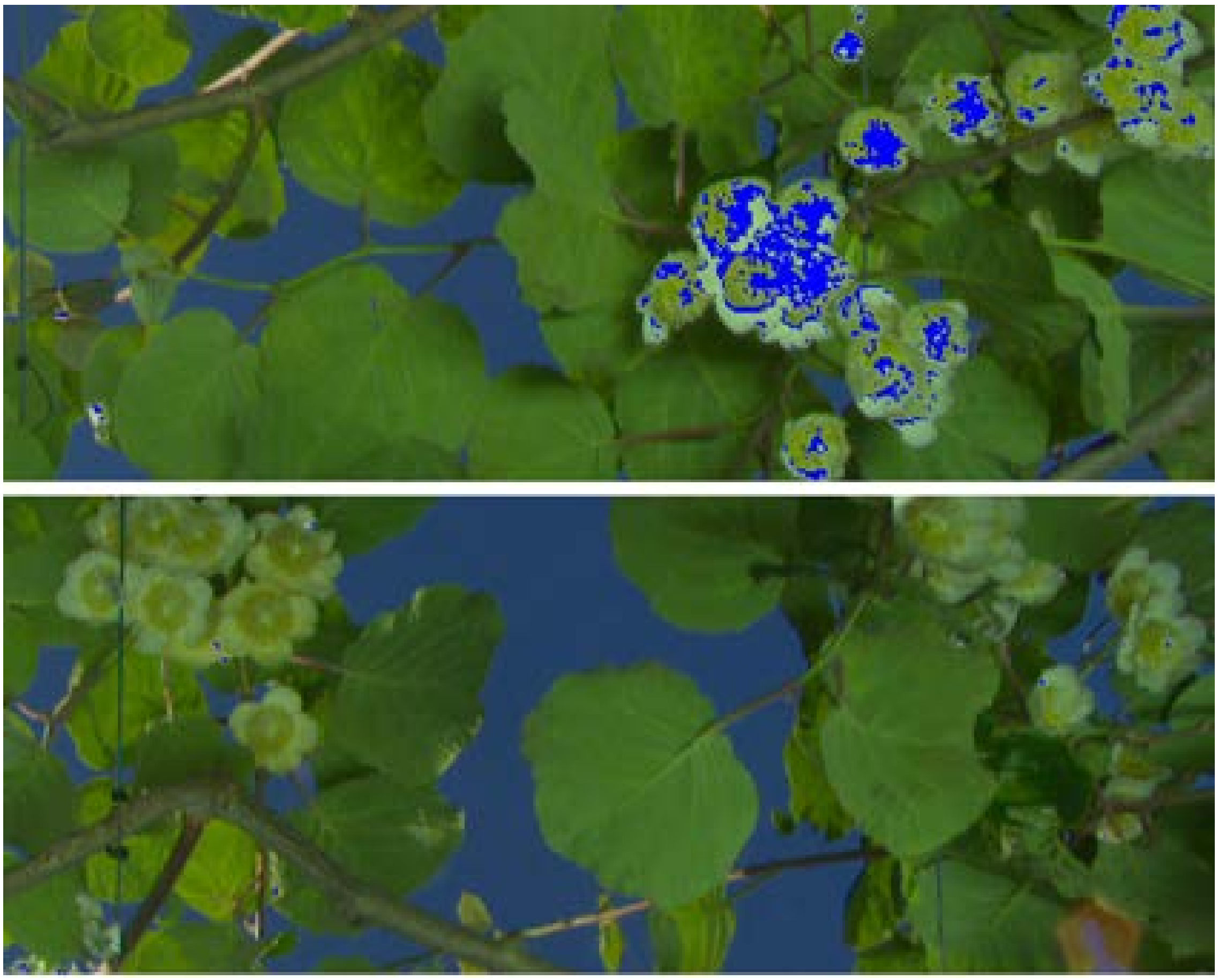

*Figure 54: Two colour thresholding results using the same thresholds but under different lighting conditions.*

Figure 54 shows how colour thresholds may work in one image but not another with seemingly small differences in the lighting conditions. Various techniques were tried to improve the robustness of the colour thresholding flower detection method including:

- Targeting different parts of the flower. Both the centre of the flower and the petals were thresholded for their colour; however, both were found to vary markedly in their colour with different ambient lighting.
- Performing median smoothing of the input image to try and reduce the range of colours of the flowers.
- Performing morphological operations such as erosion and dilation on the binary image after colour thresholding to try to filter false positives and connect parts of the flowers.

However, it was found that even with such measures, retuning the colour thresholds was required throughout a day. The requirement to retune the system created an additional burden of having to monitor the system performance, in order to determine when retuning was required.

### 3.3.3 Flower Detection Using Convolutional Neural Networks

Due to the frailty of the colour thresholding approach to flower detection, it was decided to investigate the use of more robust detection methods, at the cost of additional hardware and the robot power budget. In particular, the use of a Convolutional Neural Network (CNN) was investigated because of the previous success of deep learning on a range of computer vision tasks [35]. Faster R-CNN [120] was considered as a candidate CNN for flower detection because of its high accuracy and reasonable processing time. The VGG-16 [121] variant of Faster R-CNN was considered as possibly being too slow for flower detection at 5 Frames Per Second (FPS) for object detection on Pascal VOC Dataset images with resolutions in the order of 501 pixels wide by 375 pixels high [122]. The ZFNet [123] variant seemed more promising, at 17 FPS. It was decided to prioritise speed over accuracy, because of the high speed that wet pollination must be performed at, so object detection CNNs with even higher frame processing speeds were considered.

It was decided to try to use DetectNet for flower detection [44]. DetectNet is based on YOLO [4], which, at the time when the first pollination CNN was being developed, had state of the art real time performance. DetectNet is also derived from GoogLeNet [34], with readily accessible pretrained models, state of the art performance in 2014 and reasonable processing times. DetectNet was selected because its processing time seemed like it could be low enough for targeted pollination, with reported performance of 24 frames per second on images 1536 pixels by 1024 pixels on a Nvidia Titan X GPU [44]; this processing speed was much faster than the reported results for the other CNN architectures considered and this is why DetectNet was selected.

Initial experiments with using DetectNet [44] for flower detection were exploratory to investigate the feasibility of this approach. To create a trial dataset, 286 images of flowers were selected. The dataset included images collected:

- In the morning, noon, afternoon, evening and at night.
- With and without LED lighting and with different LED lighting.
- With and without polarising filters.
- Using different cameras from Basler, Point Grey and Logitech.
- At different distances from the canopy.

The images were cropped to a resolution of 1242 pixels wide by 375 pixels high. To define the labelling procedure, a rule was set that bounding boxes would be drawn around the visible parts of the flowers but only for those flowers that were estimated to be more than 50 percent visible. The coordinates of the bounding boxes drawn were extracted into KITTI [124], [125] format label text files. 267 of the images were used for training and the other 19 were used for validation. The hyperparameters used for training, which were selected by trial and error, are given in Table 18. A pretrained GoogLeNet was used for transfer learning [126]. The inference results for two of the validation images are shown in Figure 55, which shows how flowers in clusters can be occluded, resulting in some missed detections.

*Table 18: Key hyperparameters used for training DetectNet for flower detection.*

| **Hyperparameter Description** | **Value** |
| --- | --- |
| Number of Training Epochs | 600 |
| Batch Size | 2 |
| Solver Type | ADAM |
| Initial Learning Rate | 0.00005 |
| Learning Rate Multiplier | 0.9 |
| Epochs between Learning Rate Reduction | 25 |

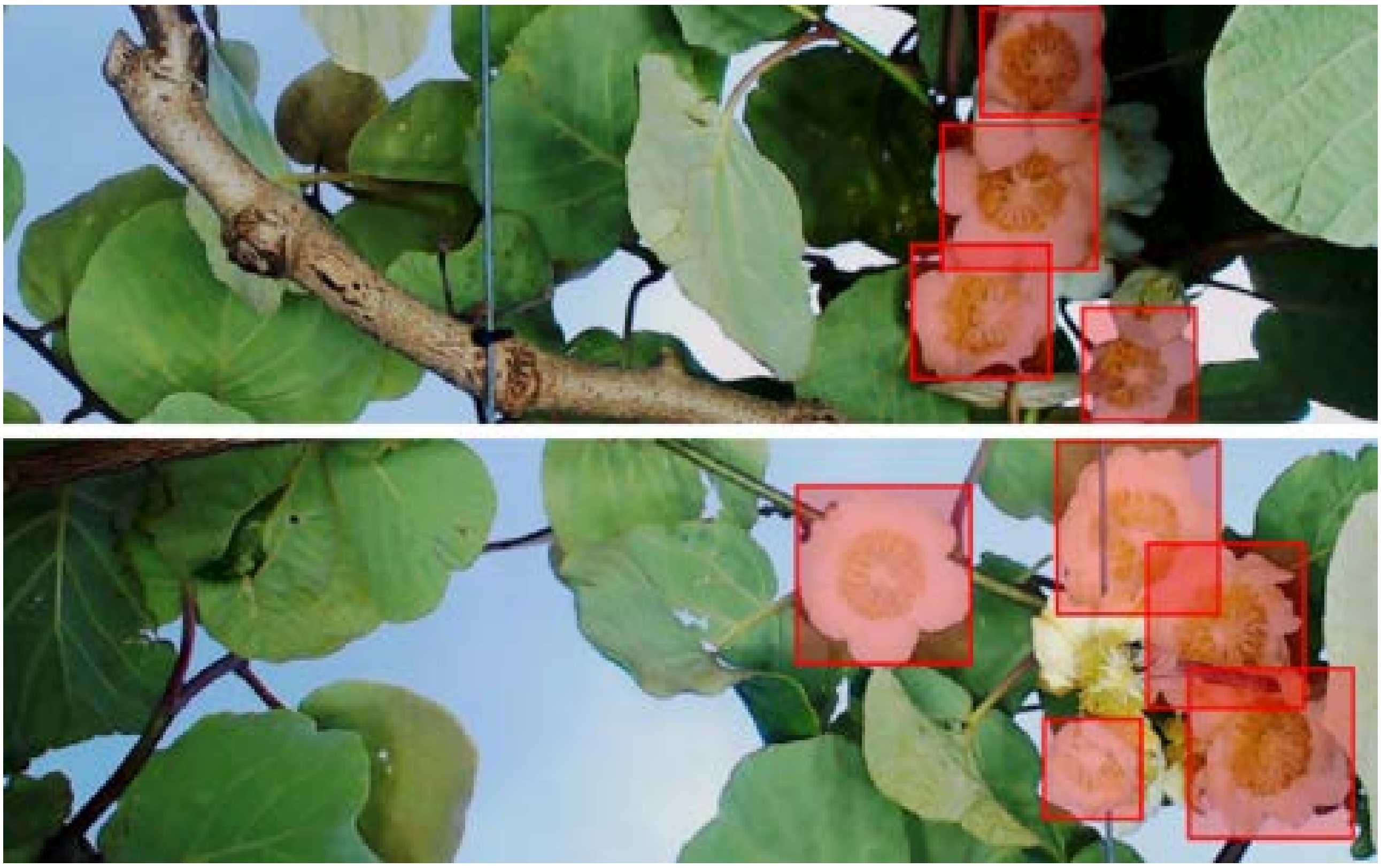

*Figure 55: Validation images from an initial trial with using DetectNet for kiwifruit flower detection.*

The Mean Average Precision, as defined by Barker et al. [44], achieved in the validation set was 83.4. The resulting model was tested in the field and detected 149 out of 162 of the flowers with open petals and more than 50 percent visibility; this corresponds to 92 percent of the flowers with no false positives in 326 images. If a high proportion of the flowers detected were subsequently sprayed, this detection rate of 92 percent might be high enough to meet the MBIE Multipurpose Orchard Robotics Project goal of hitting 90 percent of targeted flowers. A frame of the results, taken during field testing, is shown in Figure 56, which shows that even some partially occluded flowers may be detected accurately.

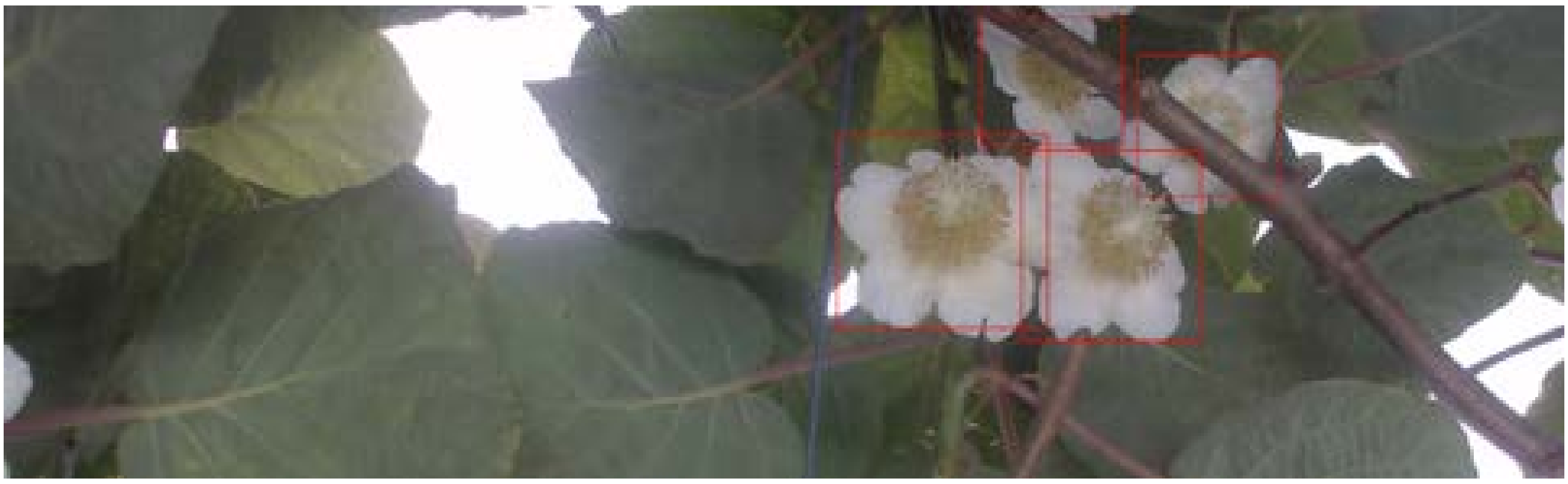

*Figure 56: A frame from field testing of flower detection using DetectNet.*

Subsequently, using DetectNet for flower detection was adopted and further tested by the wider team for pollination testing. This testing indicated that DetectNet far out-performed the colour thresholding methods and was adequate for the task of pollination.

Determining the feasibility of detecting apple flowers was also in the scope of the MBIE Multipurpose Orchard Robotics Project. Apple flowers would be detected for targeted spraying, similar to kiwifruit flower detection. However, there are some differences in the problems of detecting apple and kiwifruit flowers. Kiwifruit flowers in pergola structured orchards hang from the canopy and hence their orientation does not vary as much as flowers in an apple canopy, which commonly face any direction. In addition, an apple tree canopy is exposed to sunlight and high contrast shadows. Nevertheless, in a similar way to the kiwifruit flower detection system, DetectNet was trained to perform apple flower detection with promising results, as shown in Figure 57.

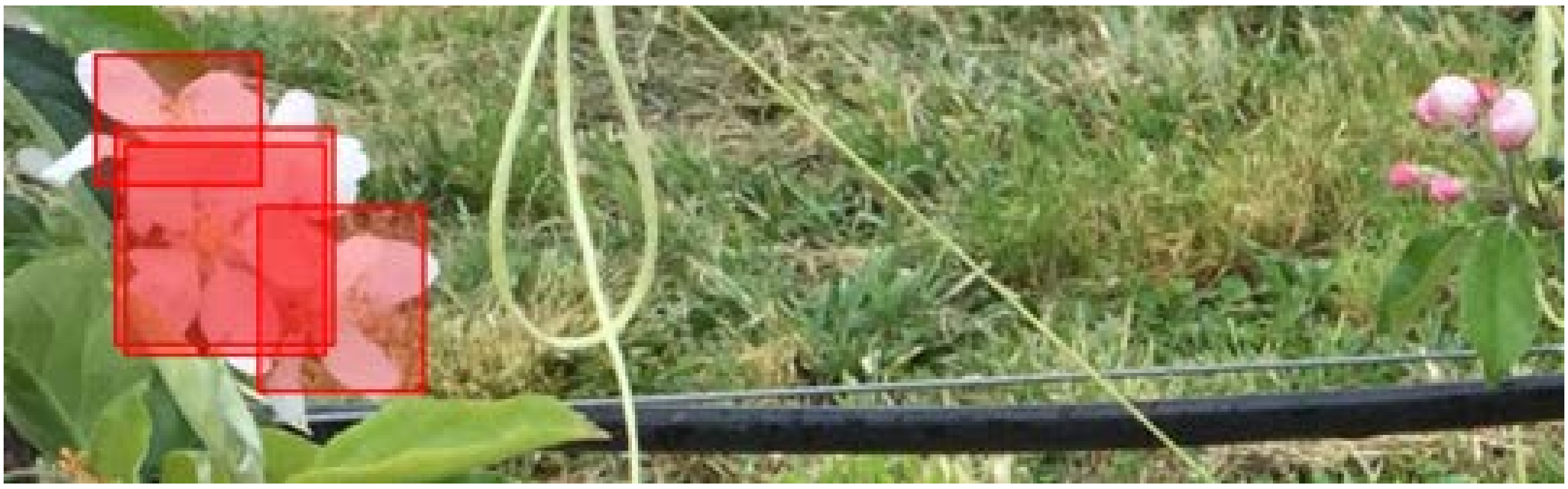

*Figure 57: Sample result given by DetectNet for detecting apple flowers.*

## 3.4 Spray Boom Calibration

After a flower was detected in an image, its coordinates relative to the spray nozzles had to be determined, in order to select which spray nozzle to trigger and to determine when to trigger the selected spray nozzle. In order to find the coordinates of a detected flower relative to the sprayers, two calibration steps were used:

1. Firstly, camera calibration was performed using methods from OpenCV [127] examples, in order to find the intrinsic and extrinsic parameters of each stereo pair.
2. Secondly, a new method was developed for determining the pose of the spray from the nozzles, relative to the cameras.

To determine the pose of the spray from each nozzle relative to the cameras, a calibration object was aligned with the spray. The calibration object was a metal sheet with notches laser cut into the bottom so that the metal sheet sat on top of the spray nozzles. Above each notch, corresponding to each spray nozzle, a series of holes were cut in a straight line at set heights. Every second series of holes was labelled with a number, corresponding to the designated number of the corresponding nozzle. The procedure for using this calibration object was:

1. Two people held the calibration object on top of the spray nozzles.
2. A selection of sprayers was turned on.
3. The two people, who were holding the calibration object, adjusted the pose of the calibration object so that the calibration object was aligned with the centrelines of the spray, when viewing the sprayers from side-on.
4. The sprayers were turned off.
5. Photos of the calibration object were taken from the stereo cameras.
6. For each series of holes, corresponding points were labelled in the left and right camera images (Figure 58). The point correspondences were used to find the 3D coordinates of the series of holes using OpenCV's triangulatePoints function and then a line was fitted to the 3D coordinates. The origin of the line was set at the top of the spray nozzle, using the 3D coordinates of and the known displacement to the lowest hole above the spray nozzle.

The camera to spray nozzle calibration was tested by running the kiwifruit flower pollination system and checking that the flowers were accurately targeted. The steps in this test procedure were:

1. 5 open kiwifruit flowers were picked from an orchard.
2. The pollination system was started.
3. A flower was held at a height of 0.1 m above a nozzle.
4. It was noted whether or not the correct nozzle was triggered and the flower was hit by the spray.
5. Steps 3 and 4 were repeated for the different flowers, for heights of 0.2 m and 0.3 m, and for all of the spray nozzles.

The result of this testing was that 100 percent of the correctly detected flowers were hit by the spray. The accuracy of the calibration was further validated by field testing as the kiwifruit flower pollination system was run in real kiwifruit orchards and was able to perform targeted spraying.

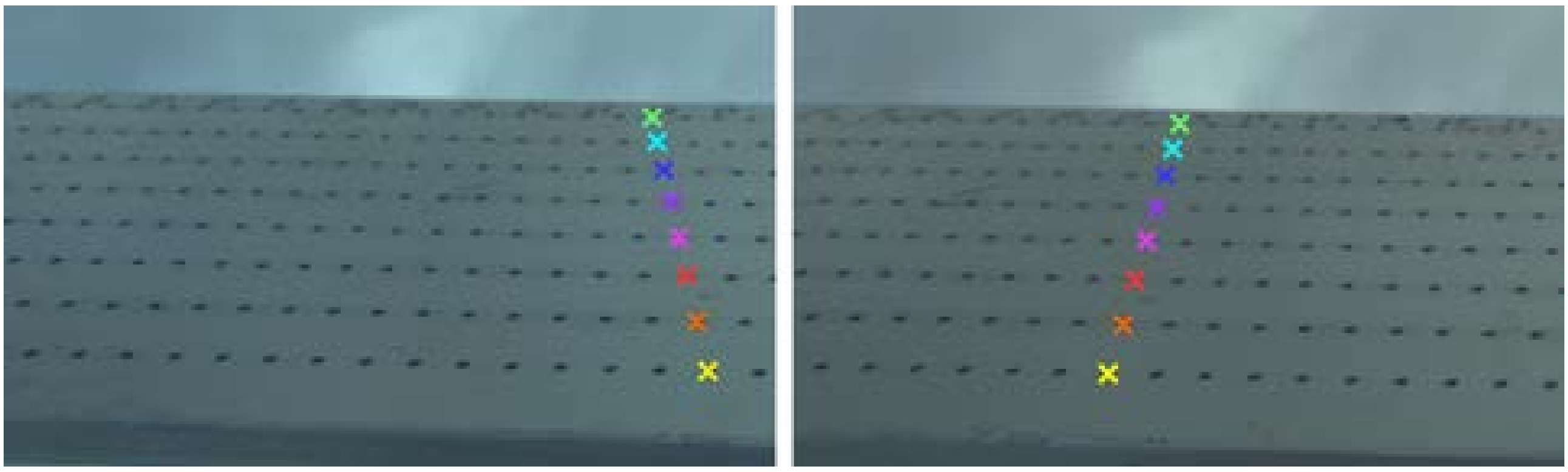

*Figure 58: Stereo images from the big pollinator, showing the top of the spray boom and the spray calibration target with hole correspondences marked.*

## 3.5 Boom Navigation

The spray nozzles only sprayed the pollen solution with a limited range and so boom navigation was developed in order to move the spray booms close to the canopy. However, at the same time boom navigation had to prevent collisions between the spray booms and the canopy. The boom navigation requirements evolved as the hardware developed and evolved further after making observations during testing in the orchard. However, initial requirements were formed based on the opinions of people in the MBIE Multipurpose Orchard Robotics Project. In order of priority, the key initial requirements of the pollen spraying boom navigation controller were to:

1. Prevent the boom from hitting anything other than a small proportion of hanging kiwifruit shoots and leaves.
2. Lift the boom to a fixed offset from a proportion of the canopy.

3. Lower the boom when no canopy was detected; this was the case when the pollinator was outside of the orchard rows.

Although it was assumed that a robotic pollinator commercial product would be driven autonomously throughout customer orchards, during testing of the robotic pollinators in the MBIE Multipurpose Orchard Robotics Project, all of the driving was done manually. As a result, the boom navigation system was developed assuming that the robotic pollinator would be manually driven. The differences during manual driving include:

- During manual driving, there is no autonomous control of the traction and steering so there is no opportunity to autonomously avoid boom obstacles by driving away from them or stopping.
- During manual driving, velocity planning information is not available so boom navigation has no knowledge of future velocities or trajectories.

These differences could appear to be quite fundamental and hence it may seem that boom navigation should have been tested in autonomous driving mode. However, developing for manual driving allowed for the possibility that the pollination module might be used on a trailer pulled by a tractor, instead of on the AMMP. This trailer option was used with the earlier testing and was also thought to be a potential product option. Because the manual driving was inherently unpredictable, the following assumptions were made for the development of the boom navigation algorithm during manual driving:

- At any moment, the driver of the AMMP could accelerate at maximum acceleration to the maximum travel speed.
- At any moment, the driver of the AMMP could turn left or right to some maximum steering angle.

At the start of the project, some team members had ideas about how the boom navigation might work. These views contributed to the initial sensor selection. The sensors that were initially used for boom navigation were:

- A laser scanner mounted ahead of the spray boom with a vertical scanning plane (Figure 59).
- Wheel encoders on the AMMP for odometry.

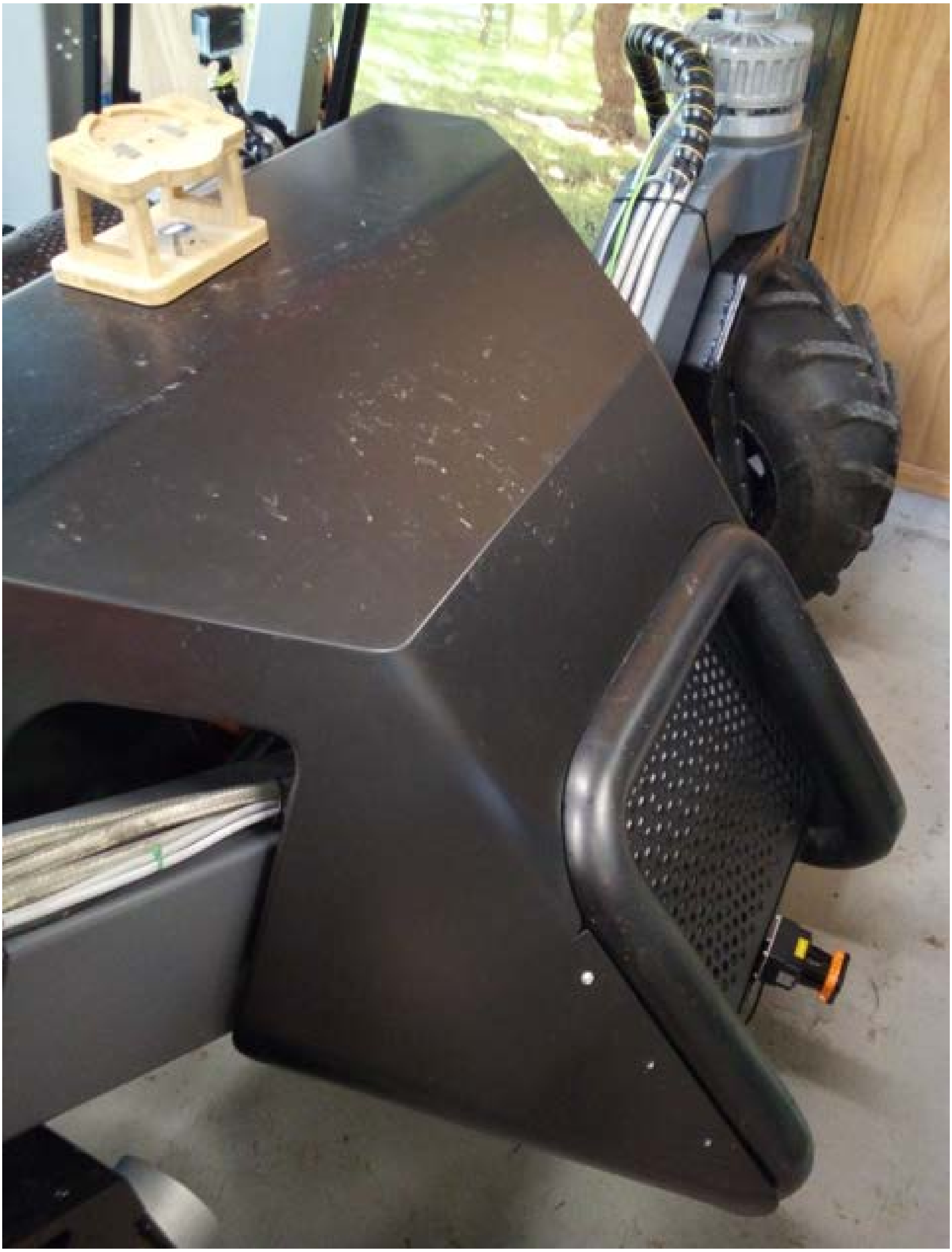
*Figure 59: The front of the AMMP with the lidar mounted so that the scanning plane was vertical.*

### 3.5.1 Vertical Lidar Boom Navigation

Initially just one boom navigation algorithm was developed for robotic pollination testing on the big pollinator and the dual pollinator. This algorithm used the lidar scanner at the front of the AMMP and the wheel encoders. For clarity, the algorithm is described for just one boom. For multiple booms the same algorithm was instantiated multiple times.

Firstly, the deviation of a spray boom from a straight path was considered. The possible future maximum deviation of the boom with respect to the boom's current pose was modelled as a curve in both directions. These curves accounted for the unknown future movement of the boom caused by the manual driver. These curves projected forward corresponded to a region of interest, where they intersected the current vertical lidar scan plane, as shown in Figure 60.

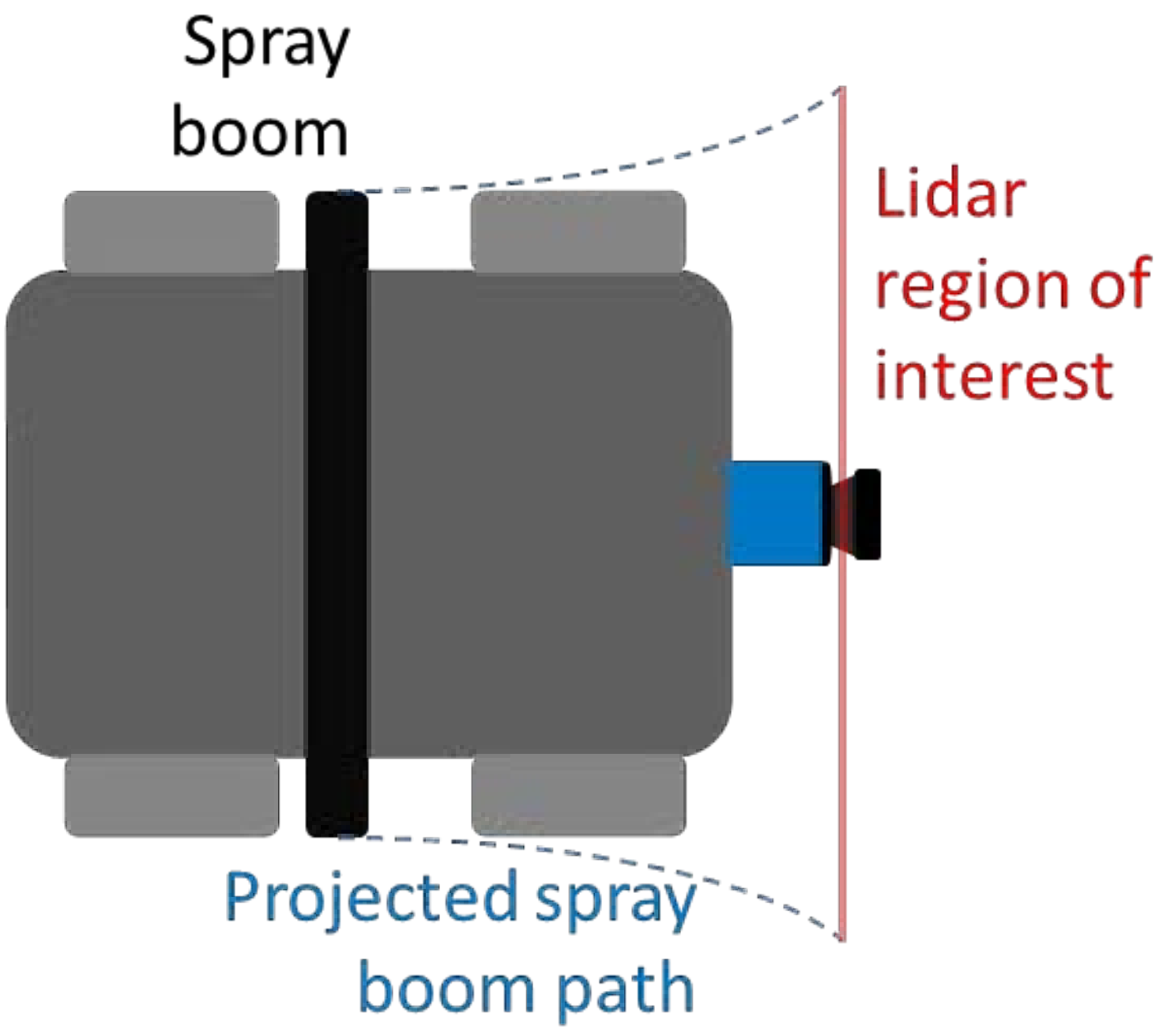

*Figure 60: Looking from above, the projected future path of the spray boom intersects the vertical lidar scan plane and forms the region of interest for the lidar data.*

An example of a vertical lidar scan and the region of interest are shown in Figure 61 with the lidar data shown as red points and with a 1 metre square grid overlaid to indicate the scale. Vertically, the region of interest started above the lidar in order to disregard points that were not a part of the bulk of the canopy.

The z-axis measurements of the data in the lidar region of interest were extracted and ordered as shown in Figure 62. A set percentile of the z-axis measurements was found. From this percentile, a fixed offset was subtracted, which took into account:

- The desired offset of the spray boom from the percentile of the canopy.
- The vertical, z-axis displacement of the lidar from the top of the spray boom, with the spray boom fully lowered.

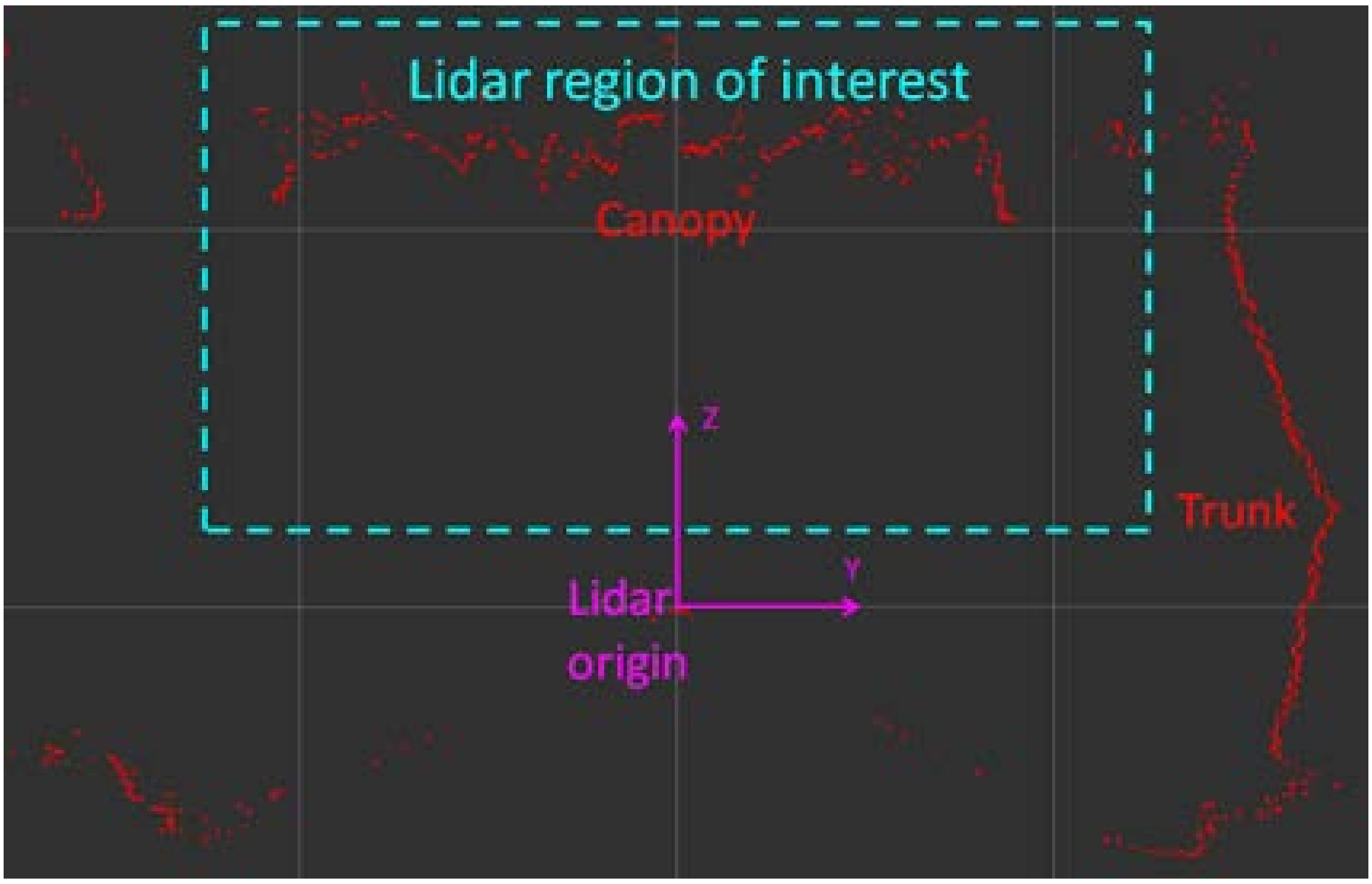

*Figure 61: A vertical lidar scan and the corresponding region of interest for boom navigation.*

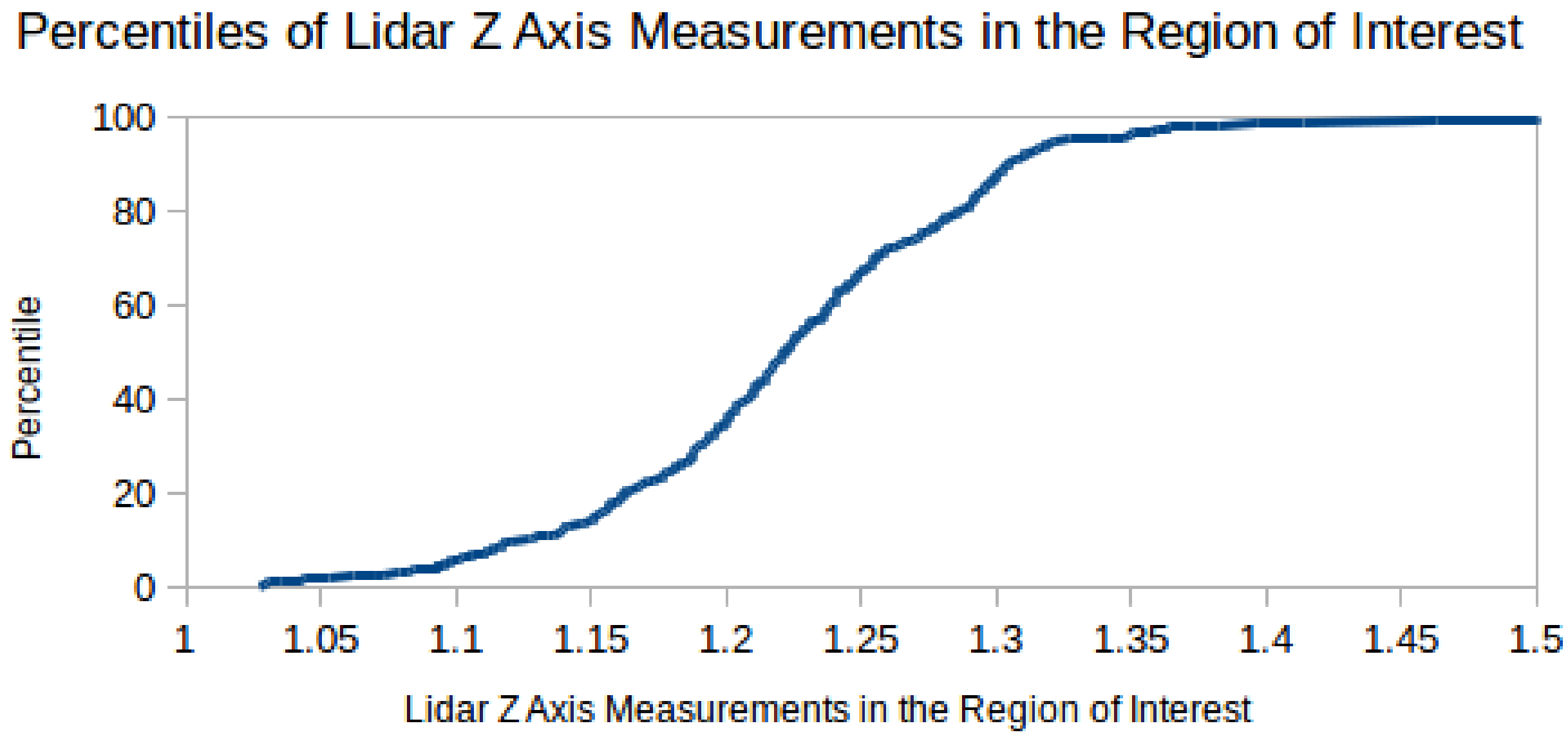

*Figure 62: Boom navigation lidar z axis measurements, ordered to find percentiles.*

The resulting height was taken to be the boom target for the current lidar scan. For each lidar scan, a new boom target was calculated. These boom targets were collated. When the spray boom moved past the plane, corresponding to a collated boom target, the boom target was deleted (Figure 63).

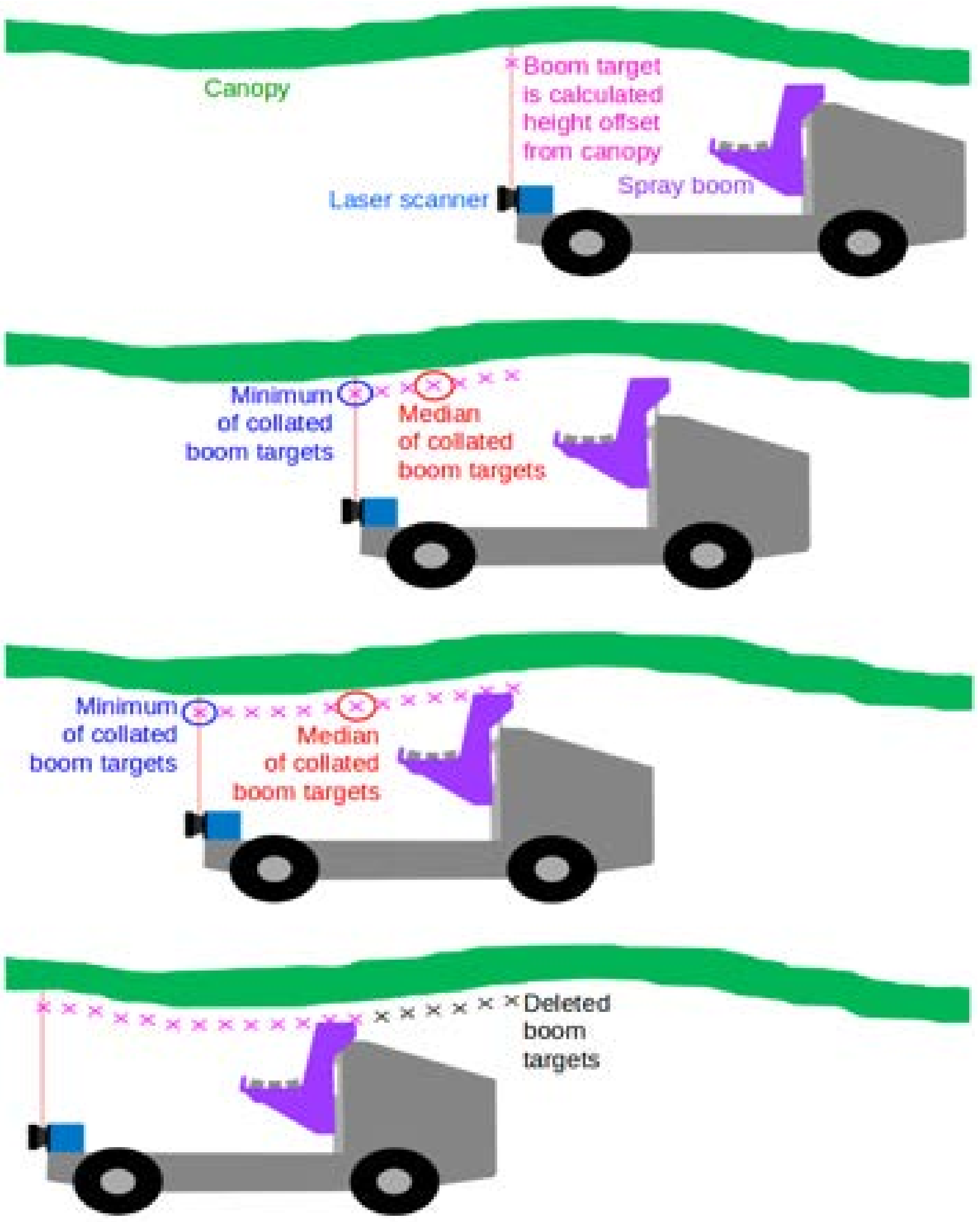

*Figure 63: Demonstrating lidar boom navigation from side on as the robot pollinator moves from right to left, while also showing the minimum boom targets and median boom targets sent to the boom height controller for different algorithms.*

In the first algorithm tested, the actual set-point sent to the boom height controller was the minimum of the collated boom targets (Figure 63). To distinguish it from other algorithm variants, this algorithm is referred to as the *minimum boom target* method. This method had some advantages over more complex approaches, given the constraints of the system, including:

- By design, at the maximum pollinator travel speed, when a low hanging area of canopy was detected, requiring the boom to be lowered from the maximum height of the boom to the minimum height of the boom, the boom controller would have to respond immediately with the maximum speed of the boom in order to avoid collision. The minimum boom target method allowed for this worst case scenario, which was appropriate because the manual driver could choose to accelerate to or even exceed the maximum travel speed at any point in time.

- By taking the minimum of the collated boom targets, the number of set-points that the boom height controller was sent was reduced, which was less demanding for the boom height control actuator and provided less height change dynamics for the pollen spray scheduler to take into account.
- The lidar scan was not guaranteed to accurately detect all features that it passed, so the minimum of the collated boom targets could be a conservative estimate of the height of the canopy ahead of the boom.

The minimum boom target method can be viewed as an algorithm that attempts to prevent all boom collisions. An alternative variant was also developed which attempted to track the general height of the canopy. This algorithm did not use the lowest of the collated boom targets. Instead a percentile of the collated boom targets was calculated. This percentile was used as the set point, sent to the boom height controller. The percentile used was the 50$^{th}$ percentile and so this algorithm is referred to here as the *median boom target* method (Figure 63). The anticipated behaviour of this algorithm was that it should not send the boom as low as the minimum boom target method because it does not use the lowest of the boom targets. It was also anticipated that using a percentile of the collated boom targets might tend to give the appearance of smooth canopy tracking.

### 3.5.2 Early Lidar Algorithm Testing with Post Processed Real World Data

Unfortunately, the big pollinator was never actuated up and down, because the hardware was not developed. In order to test the boom navigation algorithm at the time, real world data from a kiwifruit orchard was collected in order to perform mock runs of the algorithm.

The data collected was 2 sets of vertical lidar data and encoder data from wheels on a trailer, towed by a quad bike (Figure 64). The encoder data was used as the odometry input to update the set of collated boom targets. The forward most lidar data was also used as an input for the boom navigation algorithm; the output of the algorithm was the set point, which would have been sent to the boom height controller. The data from the second lidar was used to check that the set point, which would have been sent to the boom height controller, was accurately offset from the canopy at the plane of the second lidar. Effectively, the first lidar was used for boom navigation and the second lidar was used as a mock boom, to check boom navigation.

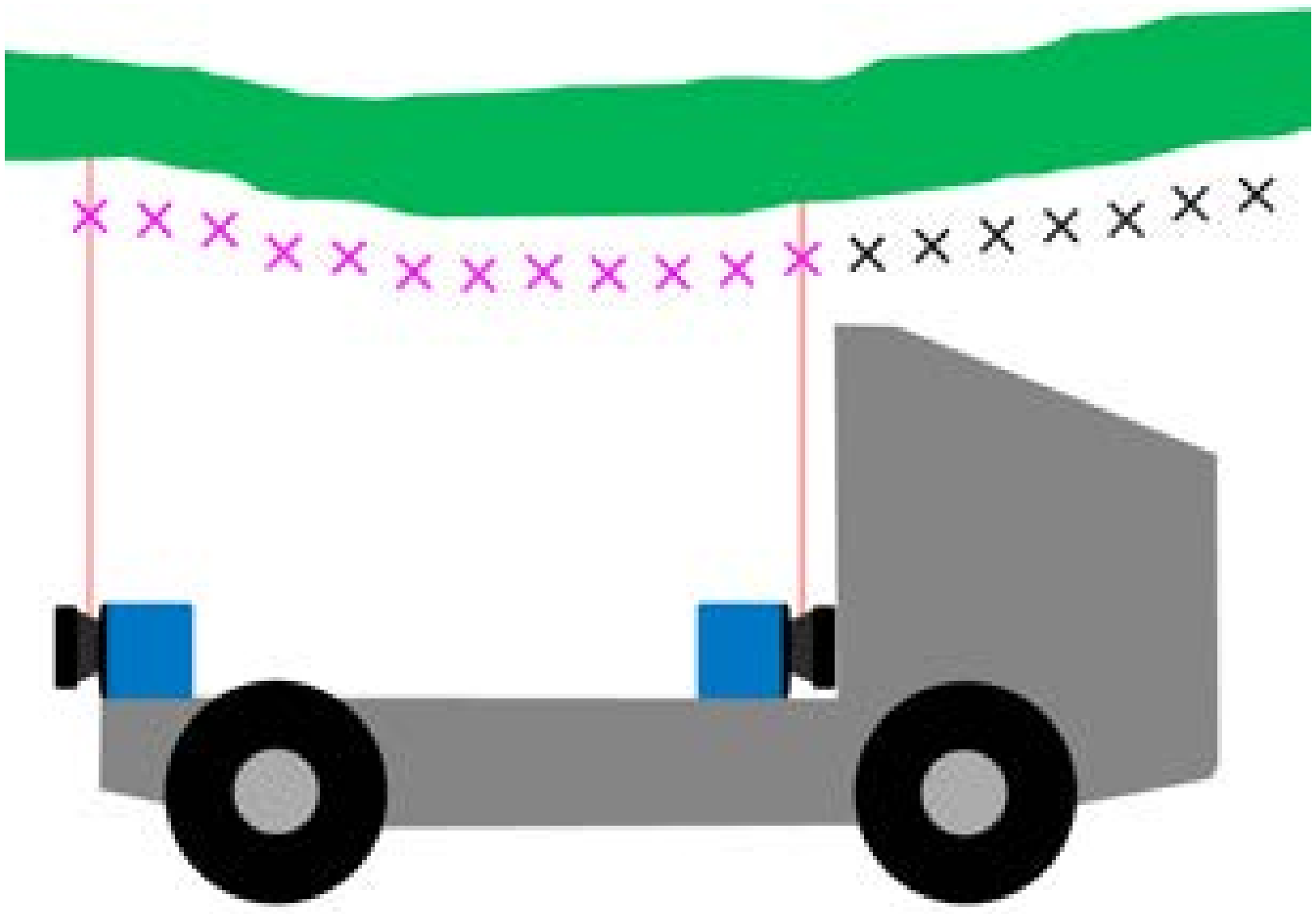

*Figure 64: An early lidar boom navigation test set-up, which travels from right to left with a second (right) lidar in place of the spray boom.*

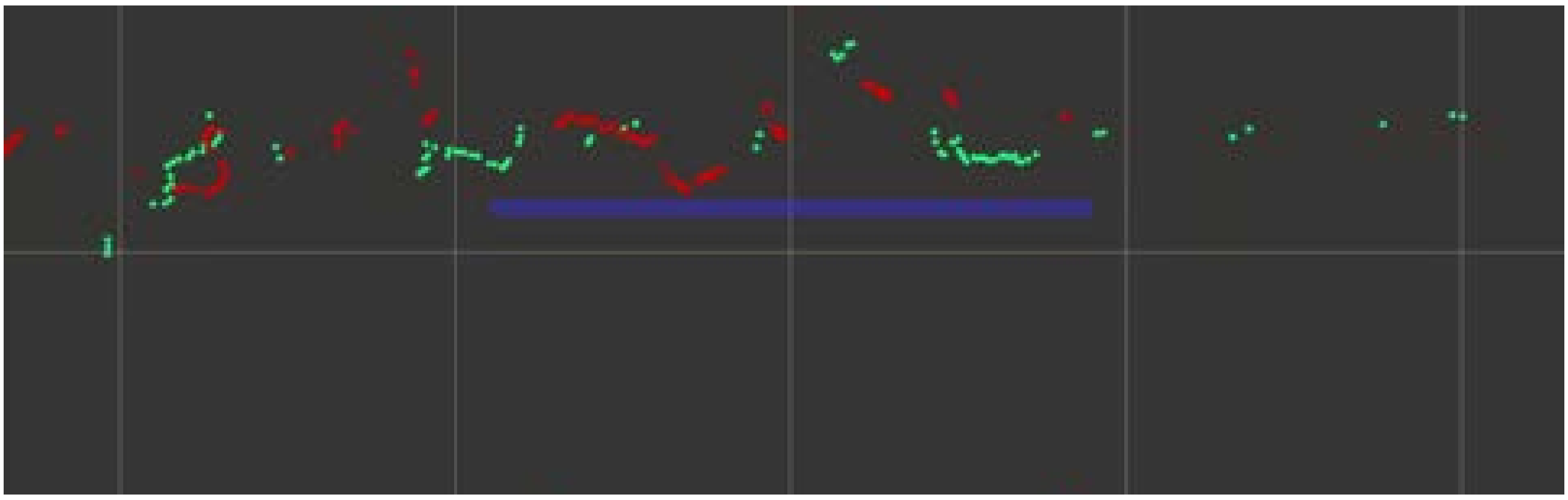

*Figure 65: Data from early boom navigation testing with the points of the boom navigation lidar (green), the points from the lidar at the plane of the boom (red) and the boom set-point sent to the boom height controller (blue line).*

Only the minimum boom target method was tested during this phase. To test the canopy tracking behaviour of the algorithm, the boom navigation offset from the canopy was set to zero. With this setting the expected behaviour in the field would be to have the boom just touching the canopy. With the canopy offset at zero, the number of times the canopy fell below the boom height set-point was counted because this was when the boom height set-point was too high. An example of a frame of the data is shown in Figure 65. In this image, the blue line represents the current height of the boom as commanded by boom navigation and the red points are the lidar measurements on the plane of the boom, representing the canopy where the boom would have been. The blue line is also drawn to the width of the lateral region of interest set for the boom as defined in Figure 60. A red point below the blue line indicates that there would have been a collision between the boom and the canopy. The worst such case in all of the data collected is shown in Figure 66.

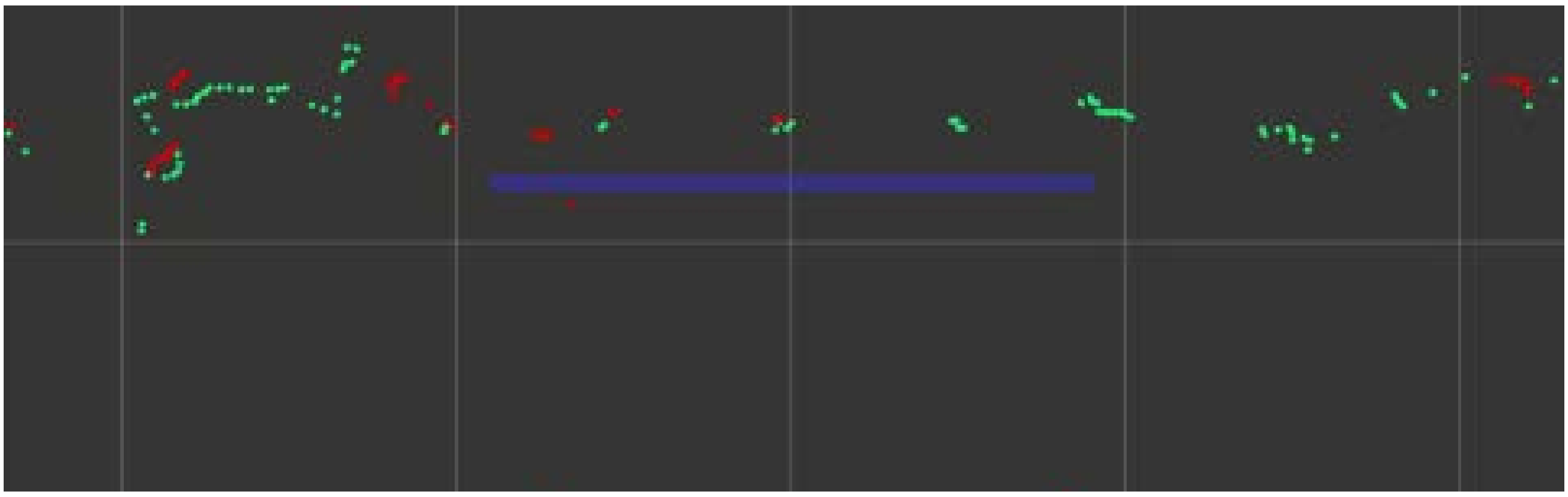

*Figure 66: The worst observed case where a lidar point on the boom plane (red) was below the boom target (blue line).*

The interpretation of the data in Figure 66 is that this is one frame of data with one point below the boom target. However, the point is not more than 0.1 m below the boom target, which means that if a canopy offset of 0.1 m was applied then a collision would not have resulted in this case. A summary of the analysis of all of the two lidar boom navigation test data is given in Table 19. From these results it seemed that with no canopy offset only a small number of minor collisions would occur and with a canopy offset of 0.1 m there might be no collisions. These results seemed acceptable because 0.1 m was the minimum canopy offset to be used in the field and having no collisions would be ideal.

*Table 19: Summary of results of lidar boom navigation testing with a second lidar scan on the plane of the boom.*

| **Total number of lidar frames observed** | **Frames with points below the boom target** | **Points observed below the boom target** | **Points observed 0.1 m below the boom target** |
|---|---|---|---|
| 1126 | 7 | 7 | 0 |

### 3.5.3 Lidar Algorithm Testing with the Dual Pollinator

Both lidar boom navigation variants were tested with the dual pollinator. The minimum boom target method was tested first by trialling the algorithm on the real world hardware and observing the behaviour in a real kiwifruit orchard during flowering. The median boom target method was subsequently tested in the same way.

The minimum boom target method avoided collisions with 100 percent of solid branches over approximately 400 metres of testing. However, the consensus of 5 team members observing its operation was that it was overly sensitive to low hanging soft branches that the boom could acceptably drive through. For example, for the hanging branch shown in Figure 67, the boom target was set as low as possible by the algorithm and hence the boom was outside of the spraying range of most of the canopy as the boom passed this hanging branch.

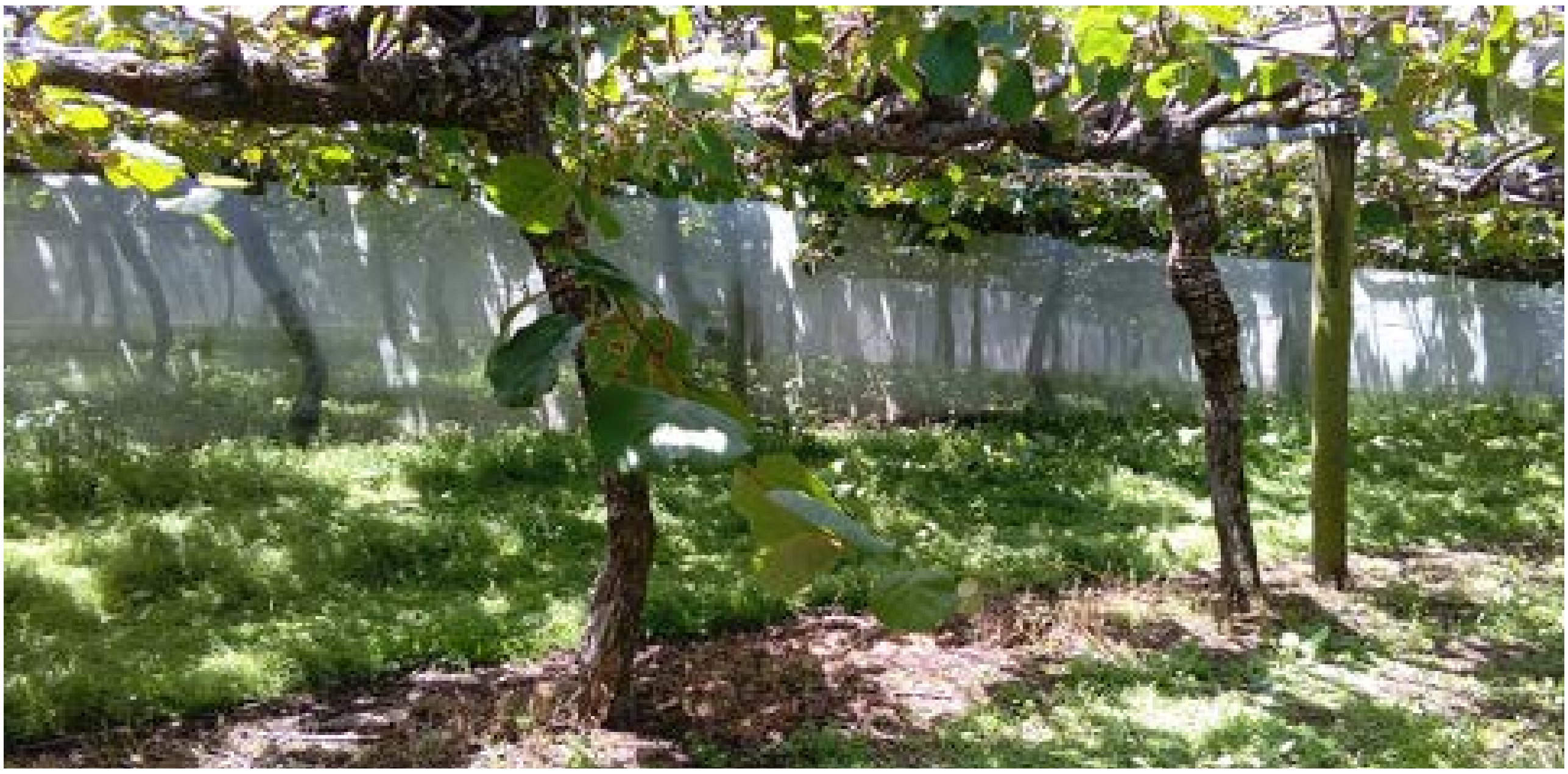

*Figure 67: Example of a hanging branch that may be driven through.*

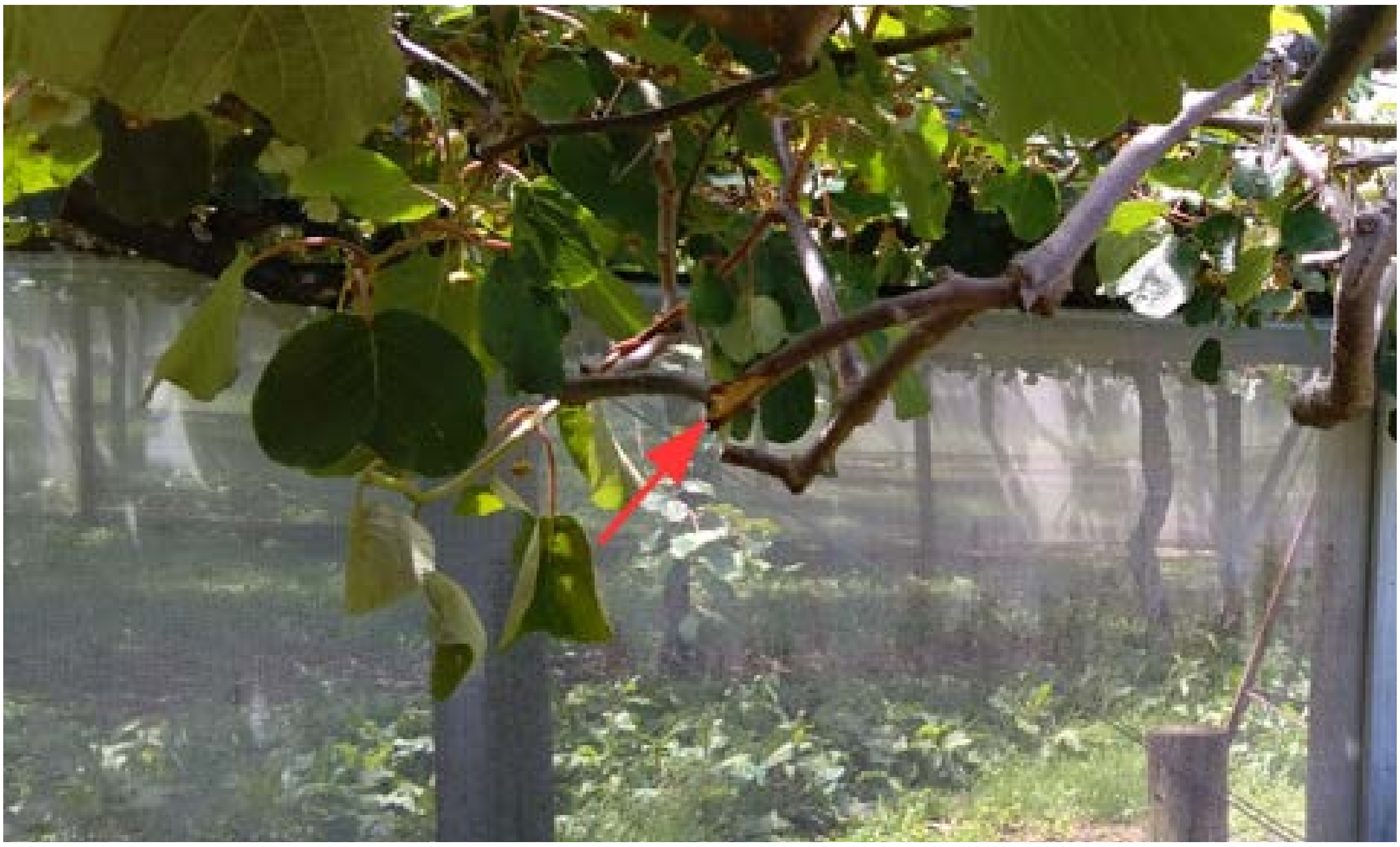

*Figure 68: A solid hanging branch that a spray boom collided with and became caught on, with the point of contact shown by the red arrow.*

The median boom target method followed the canopy more closely with an average boom height 0.2 m higher than for the minimum boom target method. Because the median boom target method kept the booms closer to the flowers, this algorithm was selected as the preferred algorithm for the first season of testing with the dual pollinator. However, in 400 metres of testing, there was a collision with the branch shown in Figure 68. This solid branch was low compared to the bulk of the surrounding canopy. This was a scenario that had not been noticed before as a potential risk for boom navigation. The difficulty in this scenario was that the lidar data from solid branches, which must be avoided, was not easily distinguishable from data from soft branches, which should be driven through. A sample of the vertical lidar data was collected with the lidar passing under the branch shown in Figure 68. This data was visualised on a screen and two people visually inspected the range and intensity data but both people were not able to identify the low solid branch from the data. From this inspection it seemed that it would be difficult to encode an algorithm to identify a solid branch from the lidar data. The visual inspection of the data also suggested that the data would be difficult to label by hand, which would be a barrier to supervised machine learning detection of a low solid branch in the lidar data.

After the collision with the branch shown in Figure 68, it seemed that a vertical 2D lidar might be a poor choice for a sensor to detect a low solid branch. The relatively low resolution of a 2D lidar and the fact that a 2D lidar only measures a 2D slice means that in any given frame of data there may only be a small number of points from a solid branch.

In order to reliably detect solid branches in the canopy, at the very least, the sensor performing the detection should receive some measurements from the solid branches. However, the vertical 2D lidar will be able to miss objects of width, $w_m$, between two consecutive scans of the lidar. This width is dependent on the period between lidar scans, $t_l$, the dimension of the effective detected area, $d_l$, and the speed of the AMMP, $v_a$. Assuming that any part of the lidar laser landing on an object is enough to register a measurement, the width of an object, which may be missed by a 2D vertical lidar, is given by:

$$w_m = v_a t_l - d_l \quad (3)$$

According to Hokuyo, the UTM-30LX [128] 2D lidar, which was initially used for boom navigation, has a detected area, which is rectangular with dimensions 0.03 m wide and 0.0016 m high for a point directly in front of the sensor at a distance of 1 metre. Interestingly, this rectangle has a different rotation for measurements at different angles around the sensor and is rotated by 90 degrees for points orthogonal to the forward direction of the sensor. For an AMMP with velocity 1.4 $ms^{-1}$, lidar with 0.025 s scan period and a detected area dimension of 0.0016 m, the width of an object not detected could be up to 0.033 m, which would be a substantial sized branch and should be avoided by the boom. If instead the detected area dimension of 0.03 m was used in this calculation, the output would be 0.005 m, which would be a thin branch that can be hit by the boom. However, in this case, a branch 0.03 m wide might be detected by a single lidar scan, the same as a smaller branch, and hence it would be difficult to measure the width of the branch in order to differentiate a thick branch from a thin branch. Hence, it seemed that the 2D lidar was not a good sensor for detecting and avoiding low thick branches. As a result, it was decided to investigate the use of other sensors for boom navigation.

It seemed that an upward facing camera or Time of Flight sensor could provide higher resolution details, which should allow for manual labelling and detection of solid branches. The upward facing camera could be part of a stereo pair in order to get the depth of the branches. There were some reservations about using the Time of Flight camera because it had not been tested when moving at over a 1 $ms^{-1}$. At such speeds, the fear was that the internal temporal filters of the sensor would cause the data to be blurred or with these filters turned off there would be a loss of data or accuracy.

### 3.5.4 Computer Vision for Boom Navigation

In order to meet the boom navigation requirements of avoiding low solid branches while driving through low soft branches, it was decided to investigate the use of computer vision techniques with a single camera or stereo camera pair. The single camera approaches considered included:

- Using the existing lidar boom navigation algorithms but switching between the minimum boom target and the median boom target methods, depending on whether or not a low solid branch was detected in a camera image. Generally, the median boom target method would be used but if a low solid branch was detected the minimum boom target method would be used to avoid an undesirable collision.
- Detecting solid branches in camera images and measuring the height of those branches with the lidar. This approach seemed feasible but would require a precise calibration and synchronisation between the sensors.
- It was proposed that it might also be possible to perform end-to-end learning of boom navigation targets from images taken when manually controlling the boom height. This approach would create a labelled dataset by controlling the boom height manually in the orchard, while collecting images, which might save time compared to hand labelling images as a manual post processing step.

It was decided to initially investigate the first of these options by detecting low solid branches in images. The methods considered for detecting the low solid branches from a single monocular camera included:

- Using a CNN to classify an image as “low solid branch present” or “low solid branch NOT present”.
- Using a CNN to perform segmentation of low solid branches.

In order to train a CNN by supervised learning for low solid branch detection, the first step taken was creating datasets. The input camera images were repurposed from the flower detection datasets, where images were collected with the cameras facing directly upwards, from below flowering canopies. However, it was discovered that, when faced with these images, the manual labeller struggled to differentiate a low solid branch from solid branches higher in the canopy. For example, consider the images in Figure 69. It was not clear to the labeller which image contained the low hanging solid branch; it was only known which solid branch was low hanging because the branches were observed in person in the orchard. Furthermore, because it was difficult for a human to label the data, it was proposed that a machine learning algorithm might also be inaccurate with this task. Hence, it was decided to abandon both of the single camera approaches for detecting and avoiding low hanging solid branches.

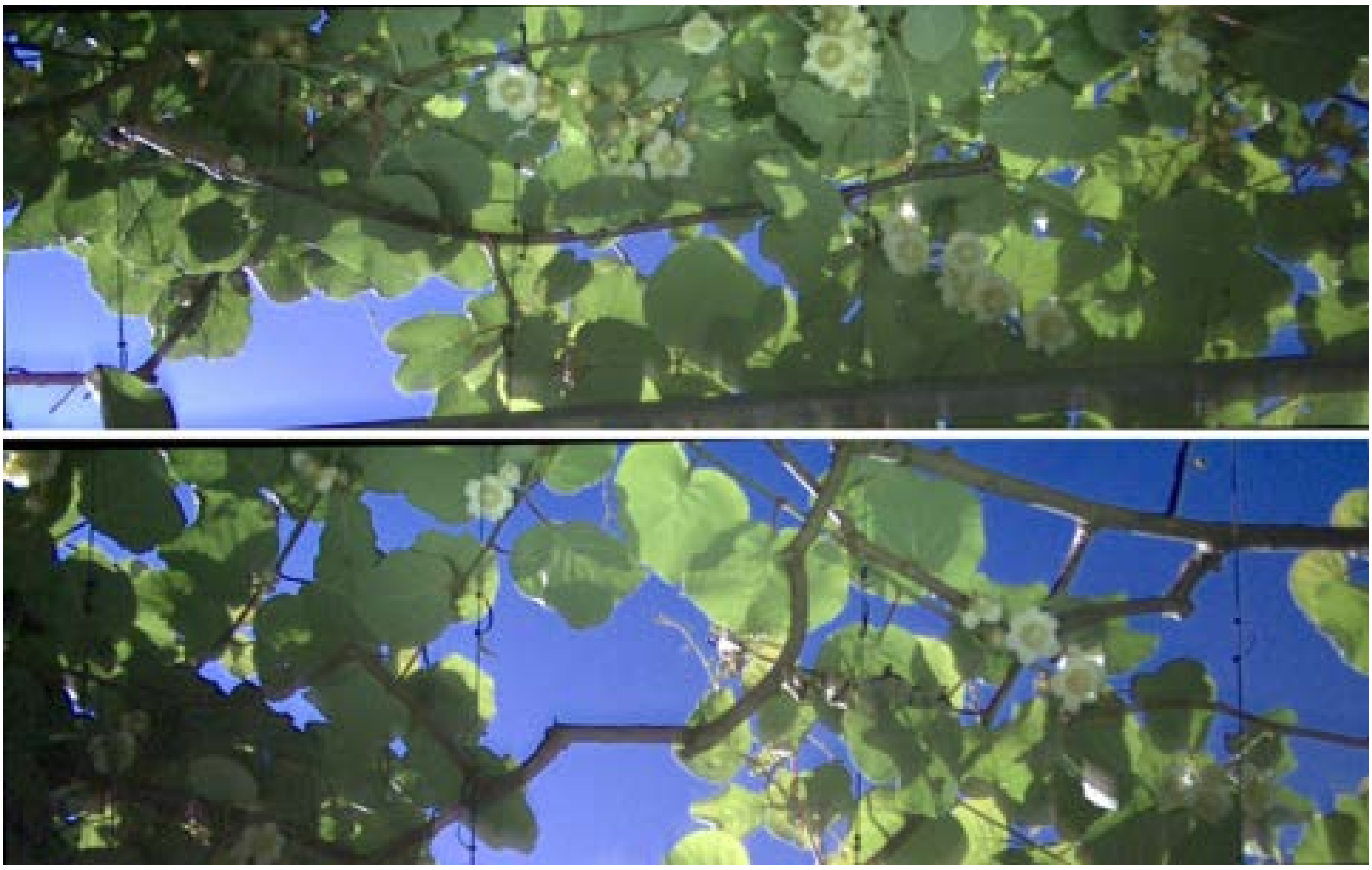

*Figure 69: Examples of images from the boom navigation dataset where one image contains a low hanging solid branch (top) and the other does not (bottom).*

Instead of detecting low solid branches in images with a single camera, it seemed it would be possible to use a stereo pair by detecting solid branches and then determining the height of those branches from the disparity. After the experience with harvesting, it seemed that solid branch detection would best be posed as a segmentation problem; then the points associated with the solid branch would be isolated for height measurement. The algorithms considered for segmentation were FCN-8s [45], for semantic segmentation, and Mask R-CNN [48], for instance segmentation. FCN-8s is a Fully Convolutional Network with skip connections from the pooling layers after convolutional layers three and four, which are fused with the seventh convolutional layer, in order to achieve higher accuracy in the upsampled output [45]. FCN-8s was used because it was postulated that instance segmentation might not be necessary for boom navigation, since the boundaries of individual branches might not be required. In addition, it was thought that it would be unnecessarily complicated to label and infer instances of branches, given that all that is required for avoiding the branches is knowing which pixels are branches; semantic segmentation would be sufficient for determining which pixels are branches. In addition, instance segmentation would require labelling according to a set of rules for where instances of branches started and ended; similarly, the neural network would be required to implicitly learn these rules. Furthermore, the inference results from instance segmentation would also have unnecessary complication with overlapping masks with different probabilities. To avoid these unnecessary complications with instance segmentation, semantic segmentation was thought to be the better solution.

However, even with semantic segmentation, as opposed to something more complex, there was a concern that the processing time would be too great, while driving at the top speed of 1.4 $ms^{-1}$. In the worst case of a very low obstacle, while travelling at 1.4 $ms^{-1}$, with the time taken to lower the boom at 1 s, the boom navigation sensor had to be at least 1.4 m in front of the boom in order to avoid a collision, assuming negligible processing time of the sensor data. A part of the concern with the processing time was that the processing time would add to this distance to the point beyond where all of the equipment would fit on the AMMP. The processing time of semantic segmentation using FCN-8s [45] on a Nvidia GTX-1080-Ti [97] for an image with 800,000 pixels was measured at 186 ms, including the image transfer time. This processing time would add 0.26 m to the equipment spacing, which was deemed to be marginally acceptable.

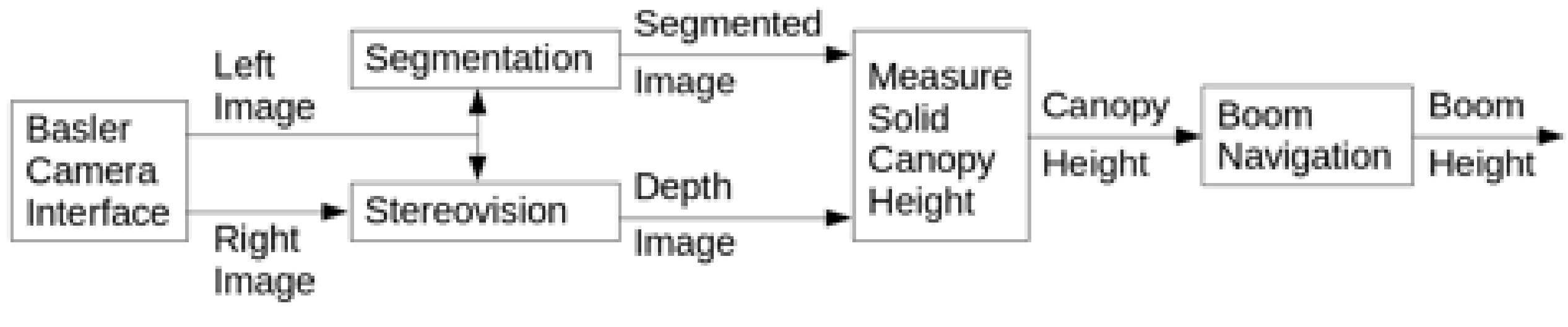

*Figure 70: Pipeline for a spray boom navigation system that uses stereovision to detect and avoid low hanging solid branches.*

A pipeline for a solid branch detection and avoidance system using just stereovision is illustrated in Figure 70. Many of the modules in this pipeline were reused from other systems. The *Basler Camera Interface, Segmentation* and *Stereovision* module were also used in the pedestrian detection systems (Subsection 5.4.8). The *Boom Navigation* module was similar to the method used with the 2D lidar. The module in this system that was unique to stereovision boom navigation was the *Measure Solid Canopy Height* module, which overlaid the outputs from the *Stereovision* and *Segmentation* modules and thus determined the lowest height of any pixel that had been classified as part of a solid branch in a volume above the stereo cameras. However, even this module had an analog in the stereovision pedestrian detection system.

An alternative pipeline for the stereovision boom navigation system is given in Figure 71. This pipeline uses more computational resource by performing segmentation on both images from the left and right cameras. In addition, the stereovision algorithm uses the segmentation output as an input and so the latency is increased with the processing of segmentation and stereo-matching in series as opposed to in parallel, which was the case for the pipeline of Figure 70. The benefit of the alternative pipeline of Figure 71 is that the input to the stereovision algorithm does not contain the noise of the raw images, which in turn may mean that the output of the stereovision algorithm contains less erroneous values. It was assumed that such errors in the stereovision output could have a significant effect if points on a low solid branch were incorrectly matched. Hence, it was proposed that the pipeline of Figure 71 might be employed if issues were detected using the pipeline of Figure 70.

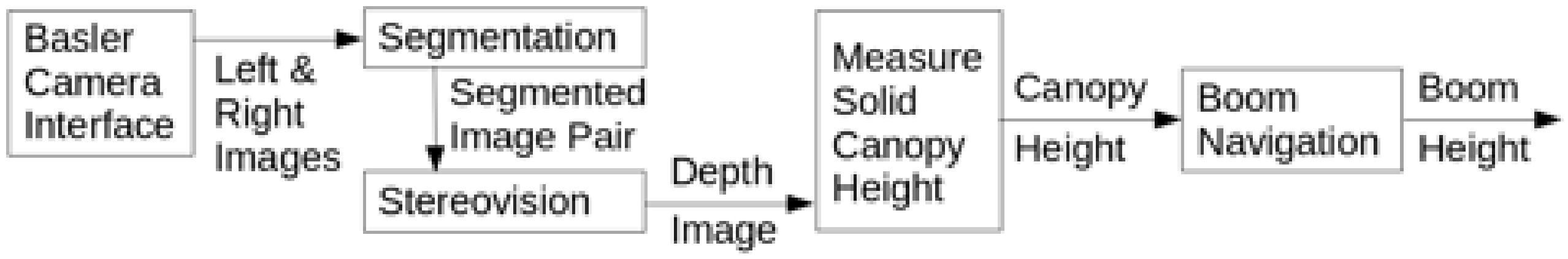

*Figure 71: An alternative pipeline for stereovision boom navigation, where images from both cameras are segmented and the results are processed by the stereo-matching algorithm.*

### 3.5.5 Stereovision for Boom Navigation

Different stereo matching algorithms from OpenCV [127] were tested in order to find a solution with a short processing time and high accuracy. Algorithms were tested running on a CPU, including Semi Global Block Matching and Semi Global Block Matching with just three directions. In addition, algorithms were tested running on a GPU, including Block Matching, Belief Propagation and Constant Space Belief Propagation.

The resolution of the images processed for stereovision boom navigation was 1600 pixels wide by 500 pixels high. With this resolution, the images were intended to have a working area of approximately 1.28 m wide and 0.4 m long in the canopy. The stereo system was custom built using 2 Basler Dart daA1600-60uc-s cameras [92] with a baseline of 0.1 m.

The CPU used for testing stereo matching algorithms was an Intel KabyLake Core i5-7600K [129]. The GPU used was a Nvidia GTX-1080-Ti [97]. Table 20 summarises the processing times measured for the algorithms tested. At 4.3 frames per second, Semi Global Block Matching with 3 directions would have processed 1.7 m of canopy per second (4.3 fps · 0.4 m of canopy in an image), which would have been sufficient for travelling at 1.4 $ms^{-1}$. However, this algorithm ran on a CPU, which would perform other tasks, including the camera interfacing, the preprocessing and postprocessing of the segmentation algorithm, the reprojection of the disparity to points in real world coordinates and the rest of the boom navigation algorithm. Hence, processing this algorithm in real time was marginal and testing with it indicated that sometimes the overall boom navigation system would run too slow for travelling at 1.4 $ms^{-1}$. As a result Block Matching and Constant Space Belief Propagation were the main algorithms considered for boom navigation because of their processing time on a GPU.

*Table 20: Processing times for stereo matching methods trialled for boom navigation.*

| **Stereo Matching Method** | **Processor** | **Time to Produce Disparity (s)** |
|---|---|---|
| Semi Global Block Matching | CPU | 0.665 |
| Semi Global Block Matching- 3 Directions Only | CPU | 0.230 |
| Block Matching | GPU | 0.013 |
| Belief Propagation | GPU | 0.808 |
| Constant Space Belief Propagation | GPU | 0.071 |

Figure 72 shows an image of the canopy taken from the left camera of the boom navigation stereo pair. Figure 73 shows the disparity output from Block Matching for the image in Figure 72. Figure 74 shows the disparity output from Constant Space Belief Propagation. Both of these outputs show obvious mismatches where the disparity is too high or too low. All sections of sky should be black in the disparity images; however, this is not the case in either disparity output. In addition, white patches in both images are likely to be mismatches as well since there are no objects hanging close to the camera in the input image; the Block Matching output in particular has multiple white patches, indicating mismatches. A particular concern with this type of mismatch, with the disparity output being too high, was that it could cause false low measurements, which could cause the boom to be lowered unnecessarily, which might put the flowers out of the range of the pollen sprayers. Hence, it was thought that for boom navigation it would be important to use the stereovision algorithm which produced the least amount of mismatches. However, tuning the parameters of both Block Matching and Constant Space Belief Propagation did not eliminate such high disparity mismatches completely.

It was postulated that since Block Matching and Constant Space Belief Propagation are such different algorithms, they should generally produce different mismatches. Hence, to reduce the mismatches beyond what any one algorithm was producing, it was decided to use both Block Matching and Constant Space Belief Propagation and then compare the outputs. The disparity values from the two algorithms were compared pixel by pixel. For corresponding pixels, where the difference was greater than a threshold value, an out of range value (zero) was assigned to the output pixel. For corresponding pixels, where the difference was less than the threshold value, the pixels were averaged and assigned to the output pixel. The output given the disparities in Figure 73 and Figure 74 is given in Figure 75, which has much less high disparity pixels than the outputs of Block Matching or Constant Space Belief Propagation alone.

The next concern was that the combined disparity output would contain too little data; however, the amount of data kept could be tuned, based on the threshold for the difference between disparities of the inputs. In addition, the important pixels for boom navigation were the solid branches. As Figure 76 shows, with the segmentation output overlaid on top of the combined disparity output, there are still disparity values covering most parts of the solid branches.

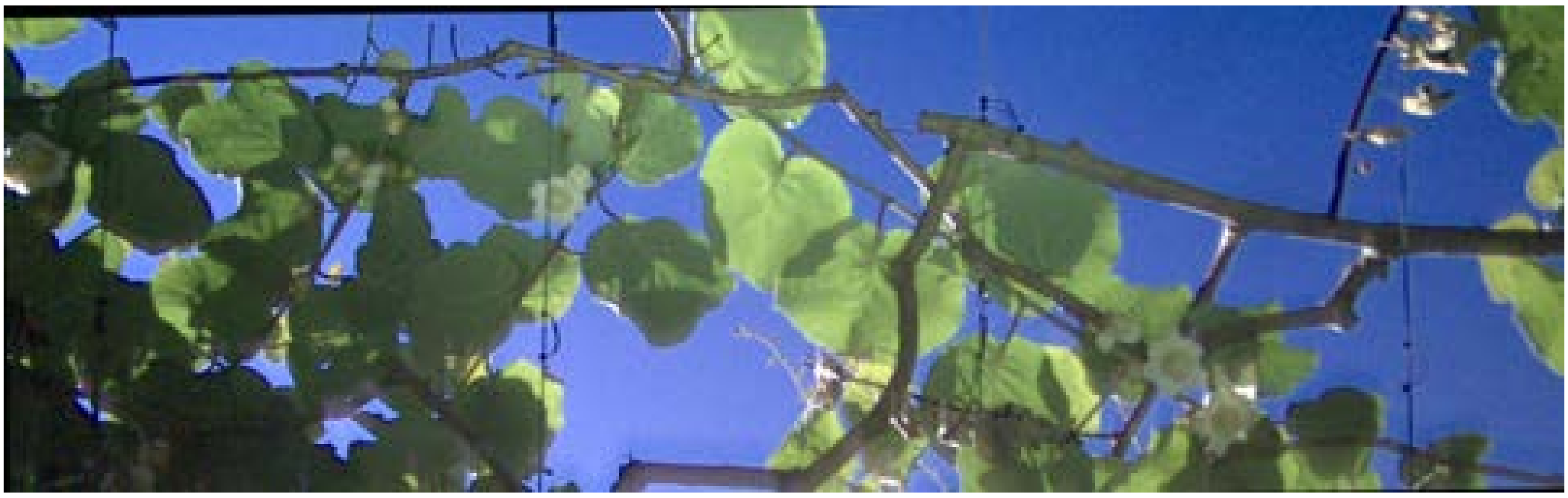

*Figure 72: A left image from the boom navigation stereo pair.*

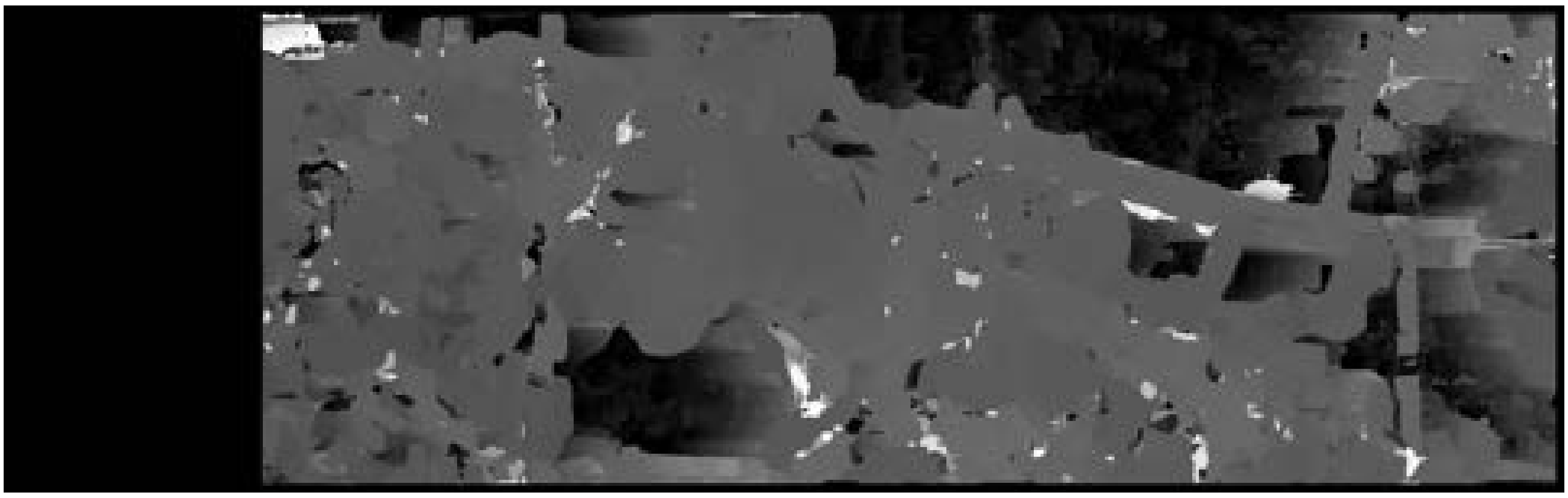

*Figure 73: The disparity output from Block Matching corresponding to Figure 72.*

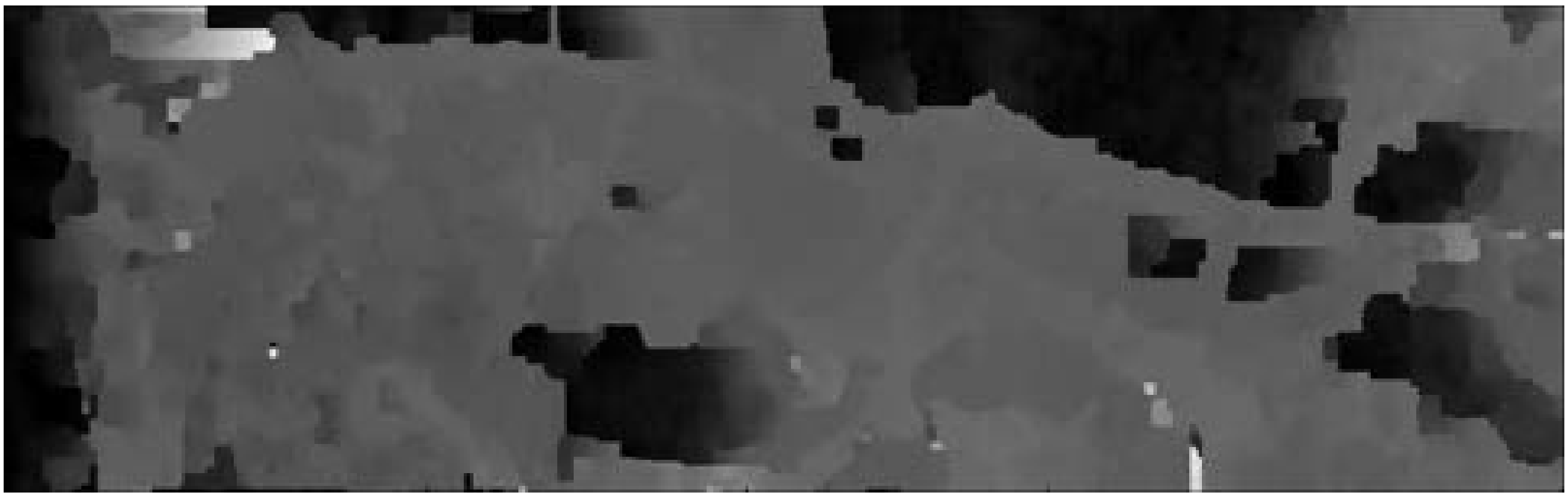

Figure 74: The Constant Space Belief Propagation disparity output corresponding to Figure 71

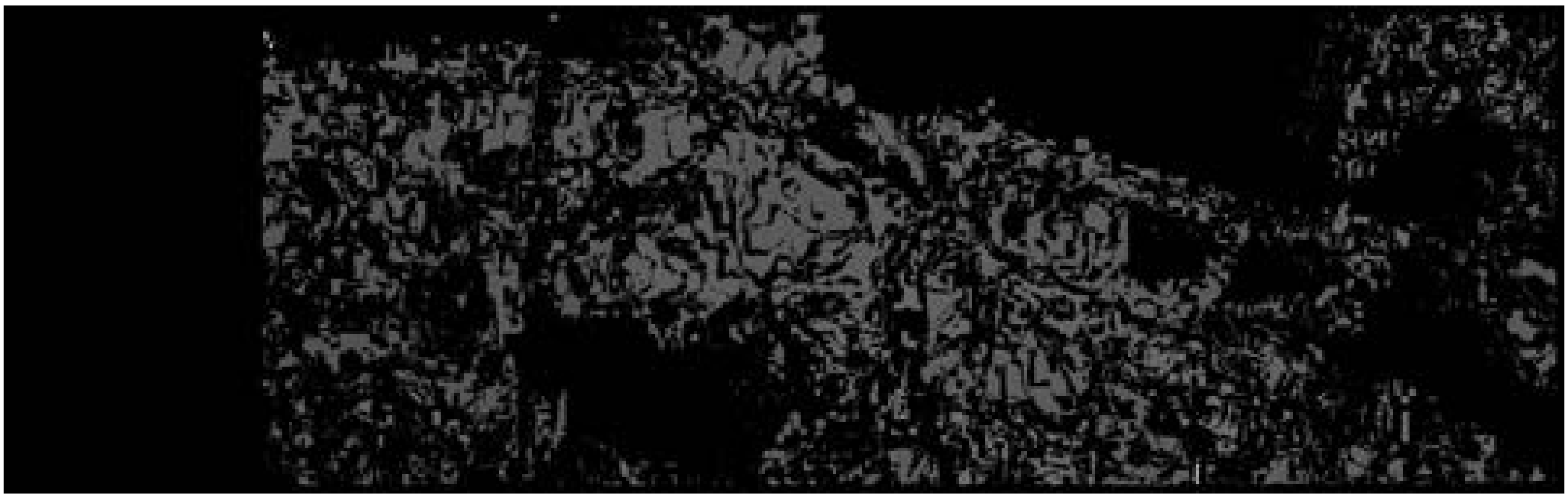

Figure 75: The result of only keeping disparity values that are within a set threshold of each other for the disparity outputs from Figure 73 and Figure 74.

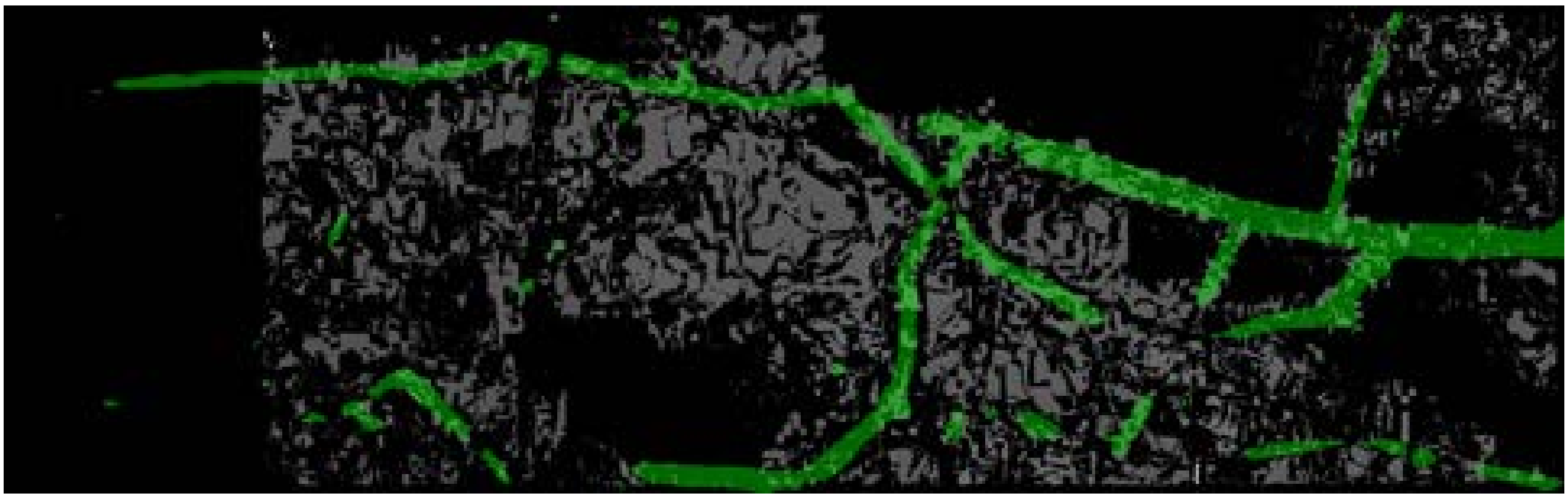

Figure 76: The segmentation result for solid branches (green) overlaid on the combined disparity result from Figure 75.

### 3.5.6 Final Lidar and Stereovision Boom Navigation System

The final boom navigation algorithm, which was used on the final version of the wet pollination system, was a hybrid of the lidar median boom target method and the stereovision boom navigation algorithms. Initially and when no canopy was detected, the boom was commanded to its minimum height. When canopy was detected, the boom targets were set by the lidar algorithm. However, when a solid branch was detected at a height lower than the lidar boom target, the boom navigation system used an offset from the height of the solid branch as a boom target for the minimum boom target method, which was in effect until the boom had cleared the detected low solid branch (Figure 77). Another way of thinking about this is the lidar algorithm was used for general canopy tracking; however, the stereovision system was used to detect and avoid the rare case of low hanging solid branches.

The sensors on the final boom navigation system are shown in Figure 78. The LED lamp was used to illuminate the canopy for stereovision. The lidar used was a Velodyne VLP-16 [130]; this was different to the Hokuyo lidar used earlier, simply because the Hokuyo lidar seemed to develop a hardware fault. The stereo pair consisted of two Basler Dart daA1600-60uc-s cameras [92] with Evetar N118B05518W 5.5 mm focal length lenses [131], with a spacing between cameras of 0.1 m. The images from the stereovision system were hardware synchronised and processed at 5 FPS.

The final boom navigation algorithm was tested with over 500 metres of driving. The paths driven included rows with low hanging solid branches. During this test driving, there were no observed collisions with solid branches.

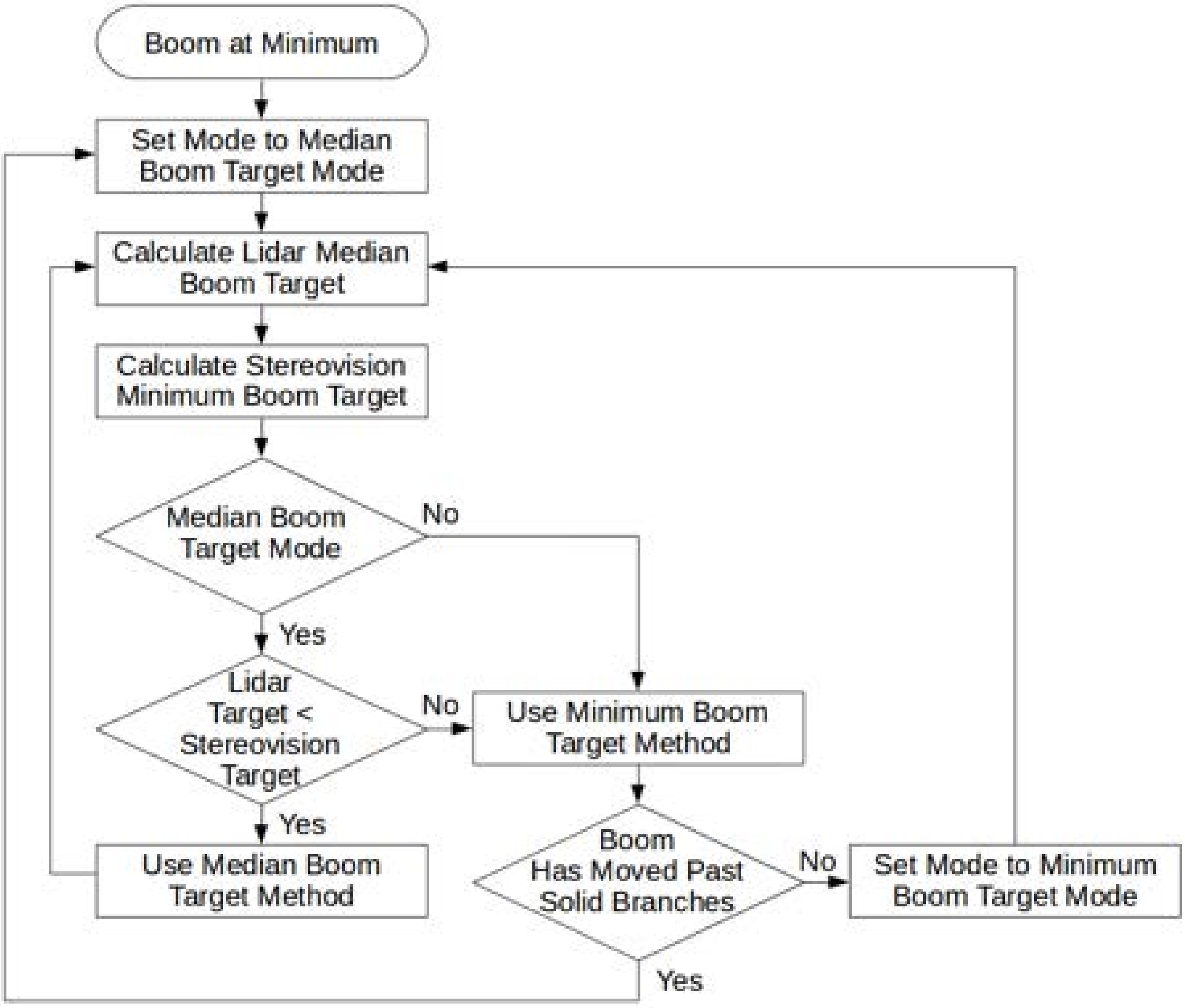

*Figure 77: Flowchart of the final boom navigation system, which was a hybrid of the lidar, stereovision, median boom target and minimum boom target methods.*

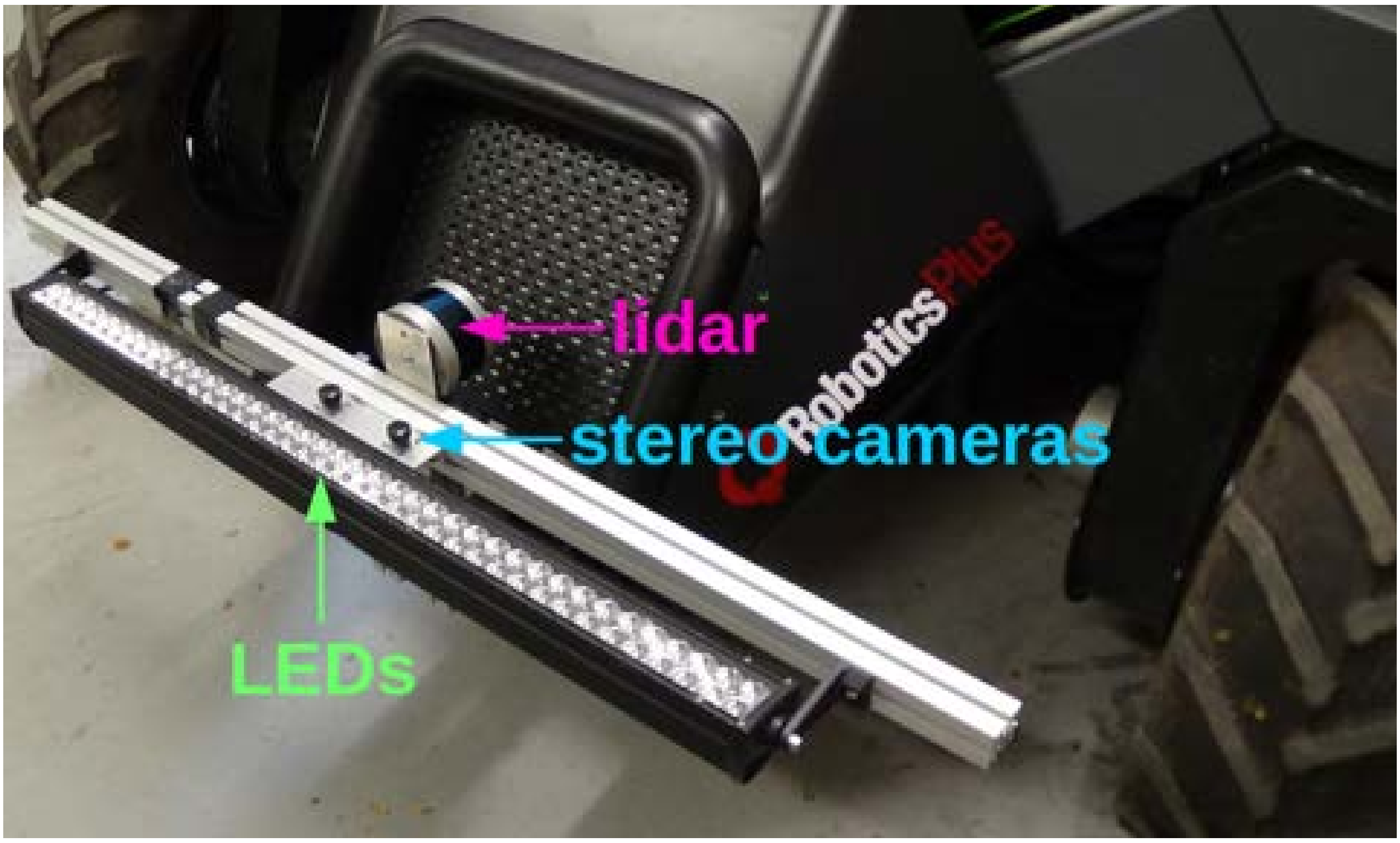

*Figure 78: Parts of the final boom navigation system with lidar and stereo cameras on the front of the AMMP.*

## 3.6 Dry Robotic Pollination System

For four seasons during the Multipurpose Orchard Robotics Project, pollination was conducted by detecting flowers and squirting the detected flowers with pollen in solution. Some of the issues with doing pollination in this way, included:

- Pollination had to be done relatively quickly because the pollen in solution would begin to germinate and hence lose viability. Because pollination had to be done quickly, the robot travel speed was increased, which made the task of successfully hitting the flowers more difficult.
- It was found that squirting pollen could be wasteful since water droplets could bounce off the target pistil area of the flowers or miss this area entirely. On average, less than 3 percent of pollen remained on the pistil area of the flowers.

A solution to the water droplets bouncing off the targets might be to reduce the pressure of the spray and to get closer to more flowers. This goal might be achieved by having more booms that can be controlled to move closer to flowers and driving slower so that the booms can be moved into position in range of the flowers before spraying. However, this approach might be slow and, according to the pollination experts on this project, the wet pollen could germinate and lose viability. To avoid the pollen germinating before delivery to the flower pistils, it seemed that the pollen should be kept dry for as long as possible before coming into contact with the pistils. Taking this idea to the extreme, the pollen might be applied to the flowers as dry pollen, instead of being immersed in water. However, there are trade offs with dry pollen application in terms of ease of delivery. These trade offs help to explain why both wet and dry pollination are used commercially.

The synergy of using dry pollen with natural pollinators is one potential benefit of automated dry pollination. With wet pollination, the pollen begins germinating relatively quickly and hence there is reduced potential for the wet pollen to be spread by natural pollinators, such as insects and wind. After automated dry pollen application, natural pollinators may continue to spread the pollen to other flowers that have been missed or the natural pollinator may even spread the pollen around the pistil of a flower that has been partially contacted with dry pollen. Hence, natural pollinators may improve the results from an automated dry pollen application system. The natural pollinators could even include hives of bees, which may be introduced to further improve the pollination in an orchard; this is already the practice with current dry pollination in kiwifruit orchards. Furthermore, it has been shown that when bees perform pollination, the density of well pollinated flowers is higher closer to the pollen source [98], [132]. Hence, with dry pollen being spread by a machine onto flowers throughout the orchard, the bees would be close to pollen sources at all times in the orchard and there should be a more even spread of pollen and well pollinated flowers.

For the experiments with dry pollination described here, the small Kuka YouBot robot harvesting system was adapted for dry pollen delivery. In a commercial system, using the harvesting arm for pollen application may provide benefits, in addition to the cost savings from the reuse of parts. One such benefit is the selection of flowers for automated fruit harvesting. If all of the flowers in an orchard were pollinated successfully, some of the fruit would be picked early by hand and discarded, in order to allow the remaining fruit to grow larger. Amongst the discarded fruit would be some of the low hanging fruit, which are simpler to harvest using the harvesting robot since they are less obstructed by branches and wires in the canopy. Possibly, a more efficient process would be to only pollinate less obstructed fruit to begin with. One way to achieve this might be to use the harvesting arms to perform the automated pollination, since the arms may tend to pollinate flowers, which will grow into fruit that the same robot arms can reach. For example, the robot arms may not pollinate flowers that are obstructed in the canopy and behind a cluster of branches since the robot arms may not be able to reach these flowers; however, this might be acceptable since the robot arms might not be able to reach or pick the fruit resulting from those flowers, had those flowers been pollinated. In fact, these fruit might also be missed by human pickers.

### 3.6.1 Dry Pollination System Design

The basic premise of the dry pollination system was that dry pollen on the robot arm would be applied to kiwifruit flower pistils by direct contact. The dry pollination system was similar to the second kiwifruit harvester system. The same Kuka YouBot robot arm [88] and Time of Flight sensor were used (Figure 79). The Time of Flight sensor was moved to a pose viewing directly upwards, because:

- Unlike with harvesting, the flowers did not move much after each pollination action and so performing pollination in batches was more acceptable. As a result, it was also more acceptable for the sensor to be more blocked by the arm while applying the dry pollen, since no detection needed to happen during that time.
- A greater area of the flowers was generally more visible from below.

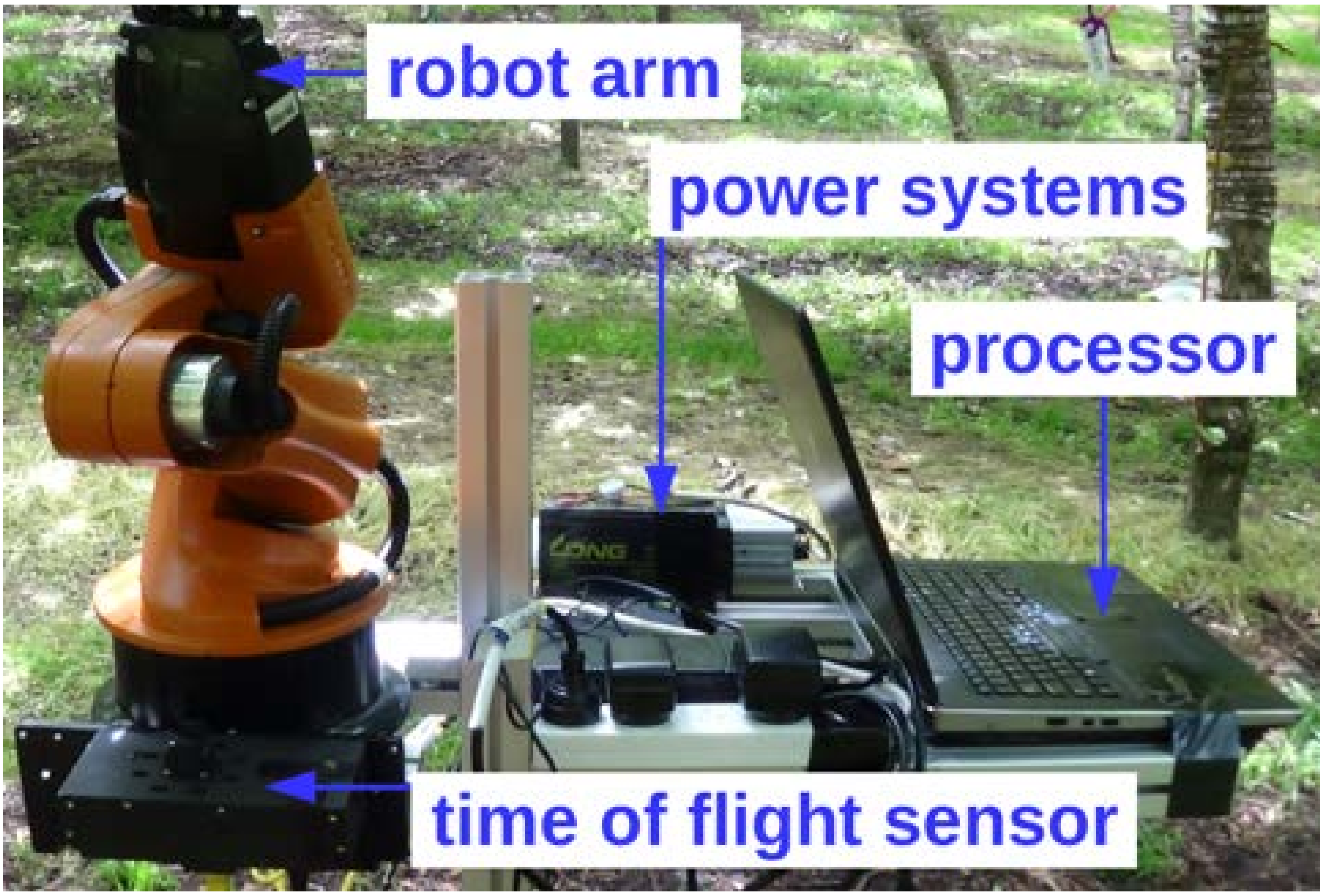

*Figure 79: Some key parts of the dry pollination system.*

The other key difference with the dry pollination system was that the harvesting end effector was replaced by a pollination end effector. The design considerations for the pollination end effector were that:

- The surface that was to contact the pistils should have a high surface area in order to carry the large numbers of pollen particles, required for complete successful kiwifruit pollination.

- The contacting surface should be soft so as to not damage the flowers on contact. Non contact dry pollen application was not considered because the goal was to minimise wastage and it was assumed that contact based methods would be better for this.
- The contacting surface could be similar to surfaces from natural pollinators in order to improve the probability that the end effector would be effective. Hence, it was thought that the contacting surface might be hairy like the thorax of a honey bee or the body of a bumblebee.
- The dimensions of the contacting surface should approximately match the area of the pistils, so that a single accurate movement of the end effector might be enough for the contacting surface to make full contact with the whole pistil area of a flower.

In order to determine suitable dimensions for the contacting surface, the diameters of 34 kiwifruit flower pistils were measured using calipers. The average diameter from the measurements was 0.0204 of a metre with a standard deviation of 0.0018 of a metre. Adding three standard deviations to the average diameter, the result was 0.0258 of a metre, which would be greater than 99.8% of pistil diameters, assuming: an ideal normal distribution, accurate measurements and a large enough sample taken. It was decided that 0.026 of a metre would be used as the outer dimensions for the contacting surface area of the dry pollination end effector during initial trials.

In order to meet the other design considerations listed above, a brush made out of goat hair fibres was constructed (Figure 80). It was thought that this might mimic the hairy body of a natural pollinator. The bristles of the brush were kept long in order to make the brush soft; the bristles were 0.034 of a metre long. It was found that at this length the fibres did not visibly damage the pistils of flowers on contact.

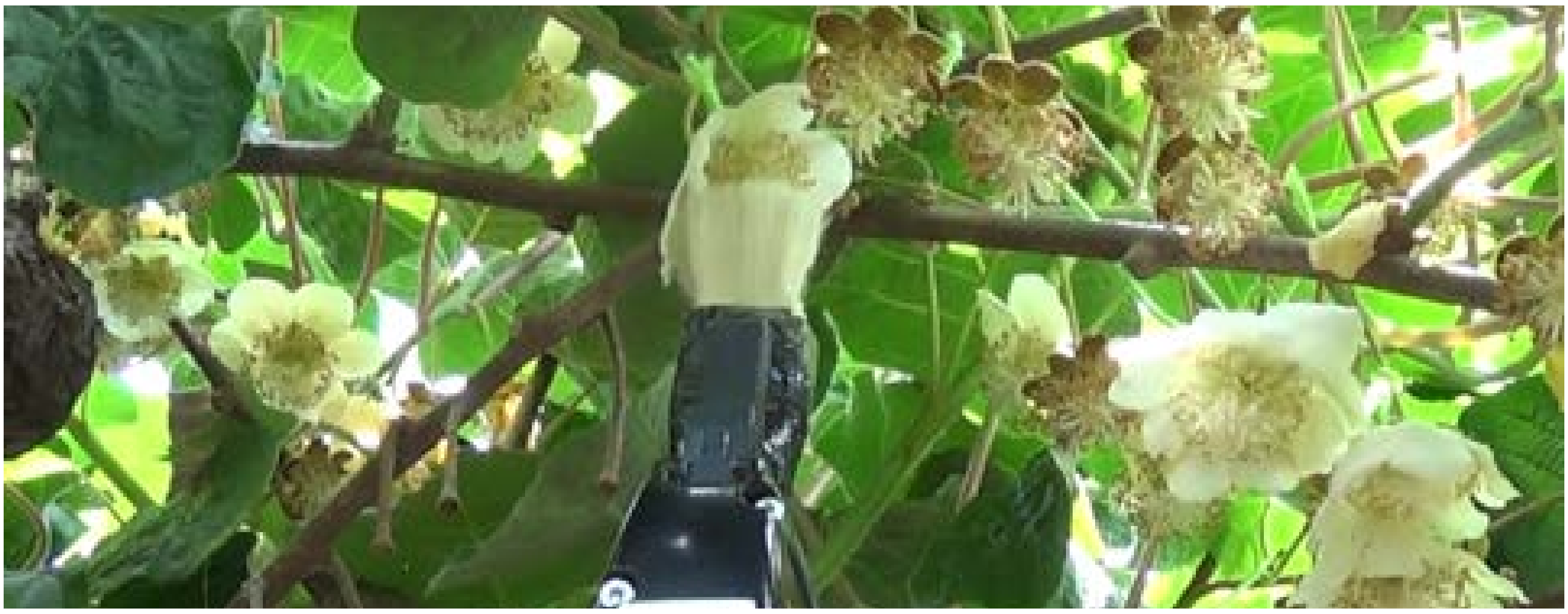

*Figure 80: The dry pollination end effector, made out of goat hair, in contact with a target flower.*

A survey of flowers was conducted to determine whether the pose variability of flowers could inhibit the dry pollination end effector from reaching the flower pistils. Occluded flowers were not included in this survey and were treated as a separate issue. The survey was aimed at determining if a significant number of flowers were angled upwards in such a way that the end effector reaching from below would not be able to access the flower pistils because of the angle. A total of 218 flowers were observed and it was judged that all 218 flowers hung in such a way that the flowers were accessible from below. As a result, it seemed that:

- The mechanical system should be able to push the end effector up against most unobstructed flowers.
- A strategy of approaching the flowers from below should work.
- Only position detection, as opposed to pose detection, would be required for calculating arm targets.

### 3.6.2 Dry Pollination System Software

The pipeline for the dry pollination system is given in Figure 81. A significant proportion of the software for the dry pollination system was shared with the second kiwifruit harvester. For example, the calibration procedure between the Time of Flight sensor and robot arm coordinate systems for the dry pollination system was the same as the equivalent procedure for the second kiwifruit harvesting system.

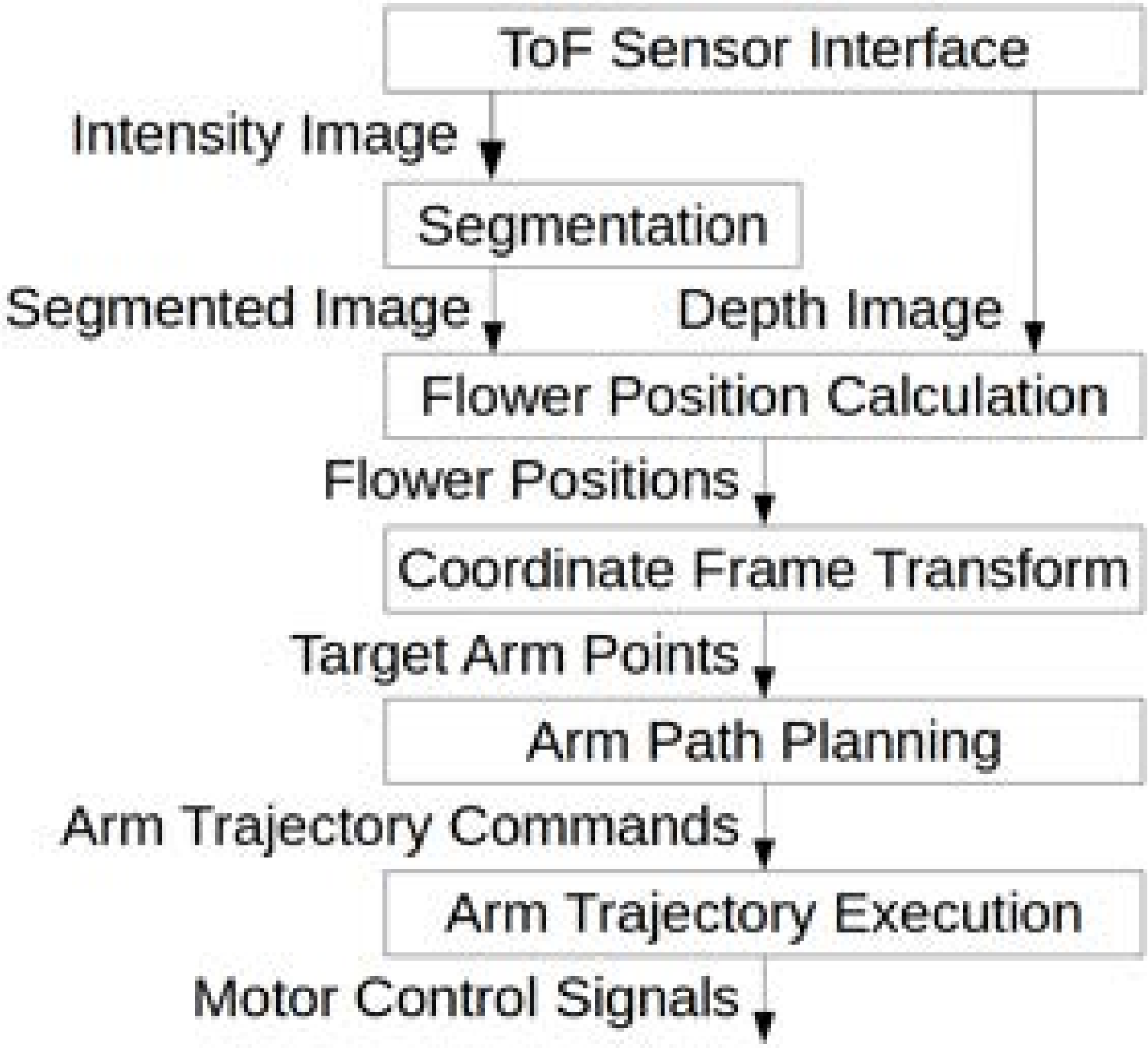

*Figure 81: The pipeline for the dry pollination system.*

There were some differences with the computer vision systems between the second kiwifruit harvester and the dry pollinator. It was noticed that flowers were quite difficult to distinguish from other objects in the canopy in the intensity channel of the Time of Flight sensor data. It was thought that this would be an issue for labelling the data and also might affect the ability of the CNN to perform segmentation. As a result, a contrast enhancement step was performed on the intensity channel of the Time of Flight data. For each pixel value, $p$, an offset, $k_o$, was subtracted and then the difference was scaled using the maximum value of a pixel for the image format, $p_{max}$, to transform the pixel value to a new value, $p'$, according to:

$$p' = \frac{p_{max}(p - k_o)}{p_{max} - k_o} \quad (4)$$

The offset, $k_o$, used for contrast enhancement was manually tuned using multiple images. An example of applying the contrast enhancement is shown in Figure 82. Flower detection was performed using semantic segmentation with the fully convolutional network FCN-8s [45], within the Caffe framework [81] and using Nvidia Digits [82], [133]. Just the intensity channel of the Time of Flight data was used for segmentation. For flower detection, the pistil area of each flower was labelled, as shown in Figure 83. A dataset of 125 image and label pairs was created, with 25 of those image pairs used for validation. The hyperparameters used for training are given in Table 21. The validation dataset accuracy was 99.4 percent. An example inference result is shown in Figure 84.

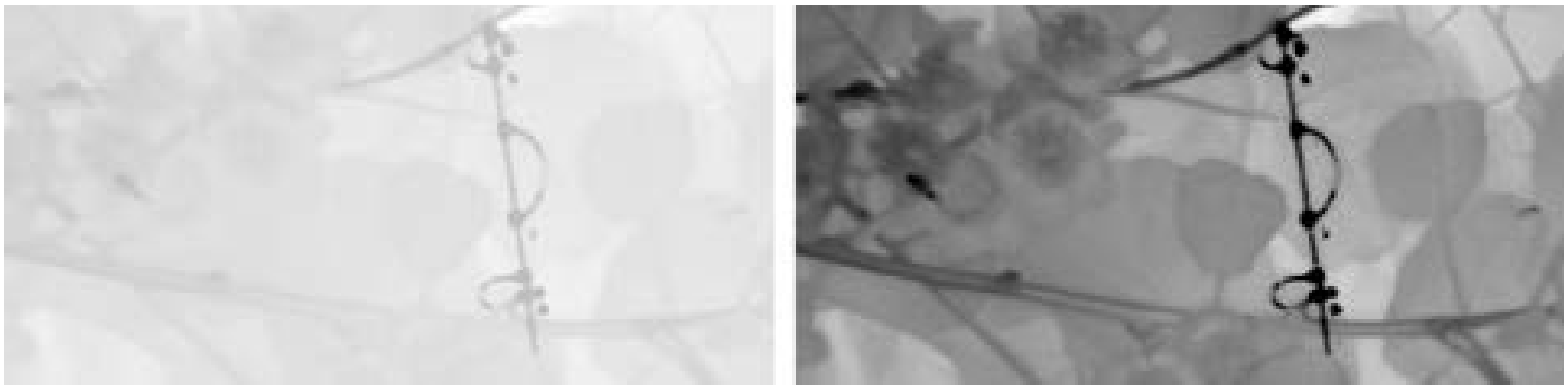

*Figure 82: A Time of Flight intensity frame before (left) and after contrast enhancement (right).*

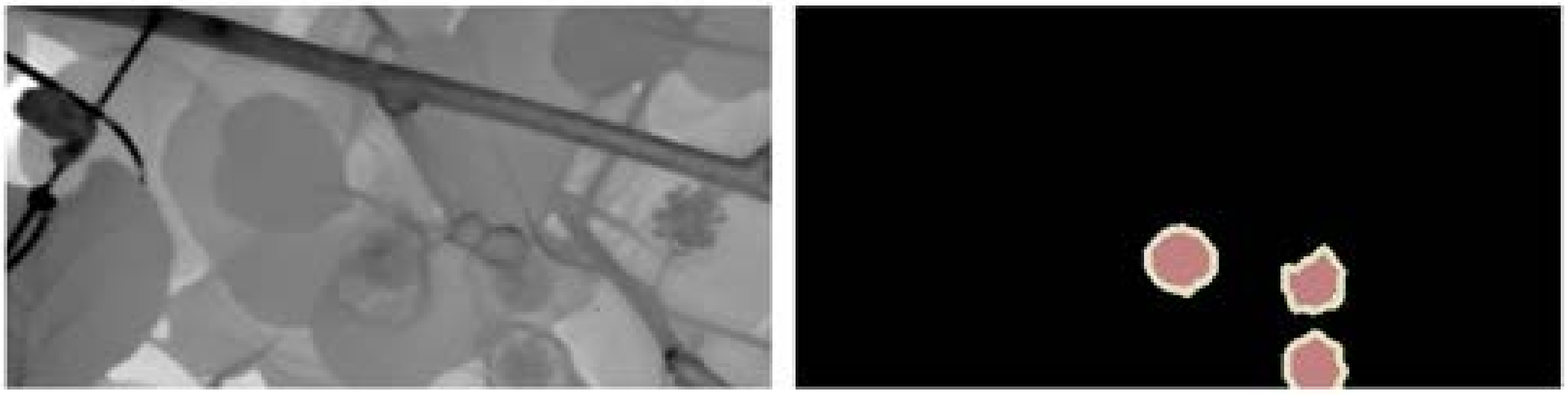

*Figure 83: An image and label pair, with the intensity channel from the Time of Flight sensor (left) and a paletted image with the pistil areas of the flowers labelled (right).*

*Table 21: Hyperparameters that were selected by trial and error and used for training FCN-8s for dry pollination flower detection.*

| **Hyperparameter Description** | **Value** |
|---|---|
| Number of Training Epochs | 100 |
| Solver Type | Adam |
| Initial Learning Rate | 0.001 |
| Learning Rate Multiplier | 0.85 |
| Epochs between Learning Rate Reduction | 5 |

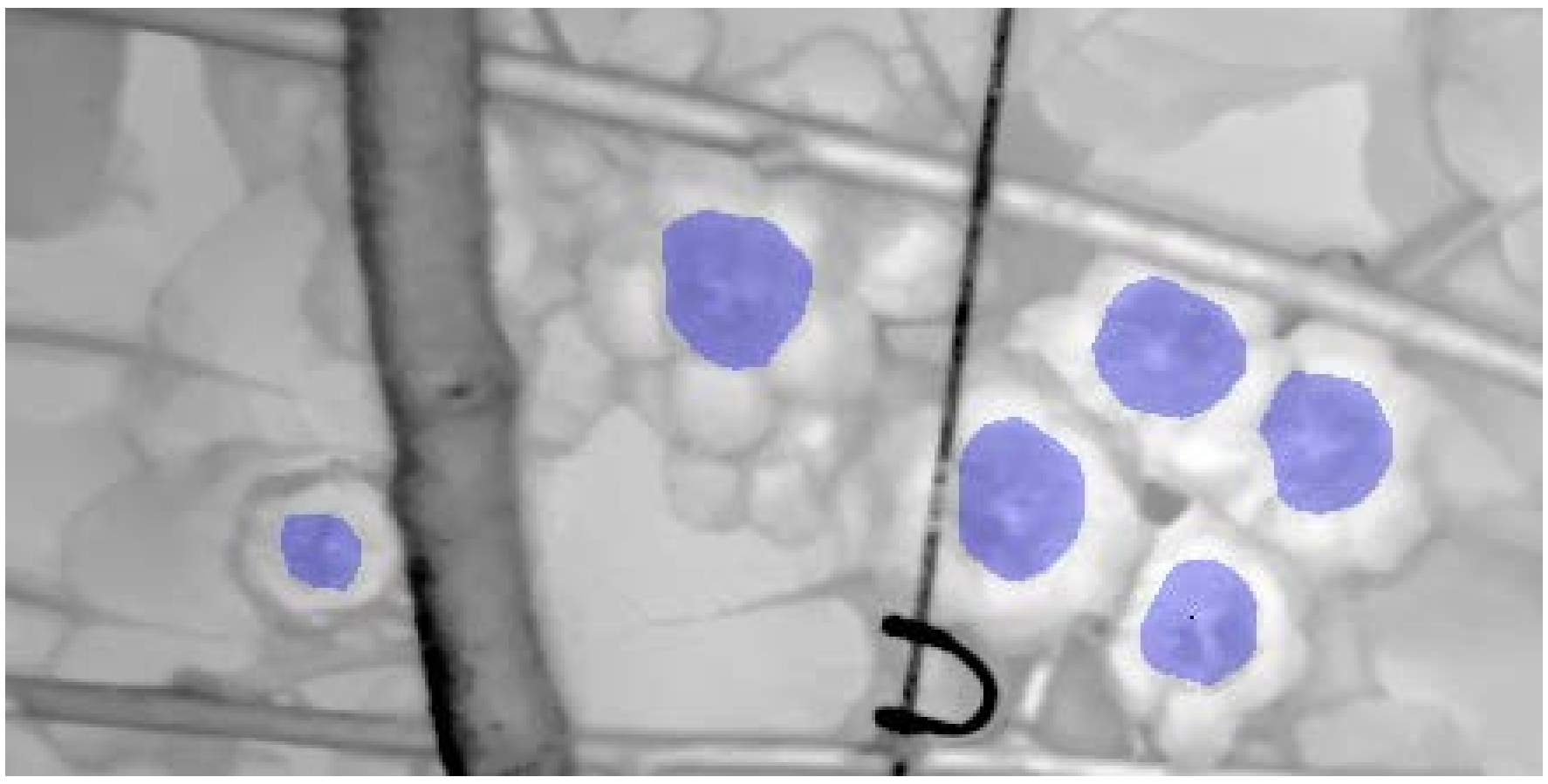

*Figure 84: Inference output of flower detection by semantic segmentation (blue) of the intensity channel of Time of Flight data.*

The flower position detection output was the median of the coordinates taken from the Time of Flight 3D points, which corresponded to segmented masks of more than a threshold area. Collision avoidance was not included in the dry pollination system. The arm control from harvesting was also used in the dry pollination system. The key difference was that an option was added to require all targets to be reached before accepting new targets. Other differences amounted to parameter changes, including:

- The length of the last horizontal end effector movement, which was used just before gripping and detaching fruit on the second kiwifruit harvester, was set to zero on the dry pollinator because the flowers were approached from below for pollination.
- There was no need to move the grippers or perform detachment on the dry pollinator.
- The end effector was not fully lowered after pollinating a flower, whereas the stem pusher on the second kiwifruit harvester was fully lowered after picking a fruit.
- The values of some offsets were changed to allow for the dimension changes with the different end effector on the dry pollinator.

### 3.6.3 Dry Pollination Fruit Quality Experiment

The primary question regarding the dry pollination system was if the hardware and software was capable of autonomously detecting flowers and delivering pollen to the flowers. This was tested in a real world orchard and the dry pollination system did successfully detect and pollinate flowers 25 times, separately from other experiments with the dry pollination system.

The secondary question was if the dry pollination system, including the robot arm and goat hair brush, was a proficient means of pollen delivery. An experiment was performed to determine if the dry pollination system was able to deliver pollen in such a way that the resulting fruit was of high quality. For this experiment, the metric of success was the weights of the resulting fruit, compared to fruit pollinated by bees. The steps in the experimental procedure were:

1. In order to prevent pollination by any means other than the robot arm dry pollination system, a selection of 50 flowers were covered before the flowers opened, using paper bags and twist ties (Figure 85). The flowers were covered when they were approximately four days away from opening, which was estimated based on the size of the flower buds at about 0.02 m wide. All of the covered flowers were marked with a piece of green wool tied around their stem.

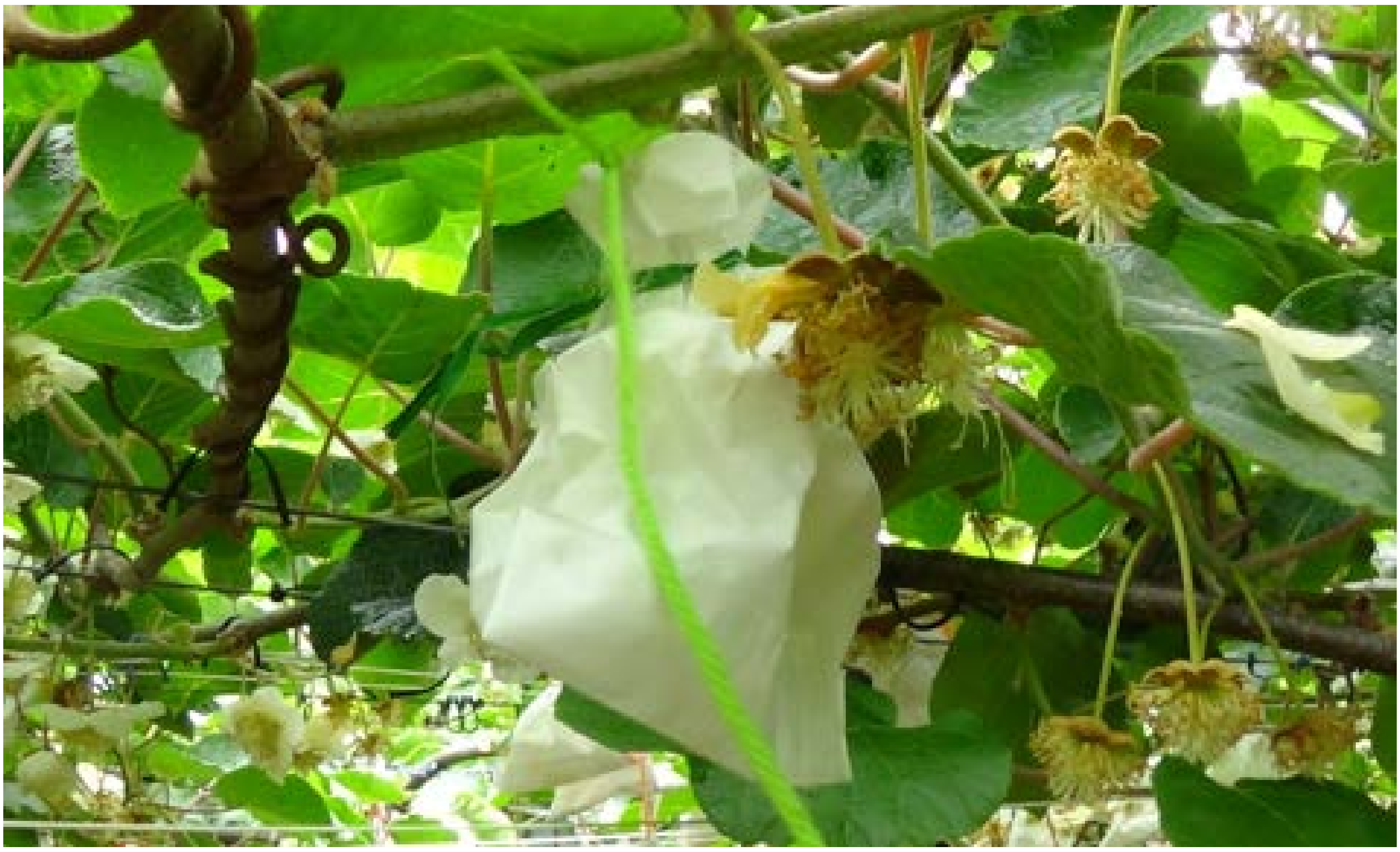

*Figure 85: A flower which has been bagged, using a paper bag and twist tie, in order to prevent pollination by wind or insects, including bees.*

2. The orchard was left for four days before returning.
3. A group of flowers in close proximity to each other were uncovered by removing the bags and twist ties.
4. The dry pollination system was positioned beneath and in range of the group of flowers.
5. The brush of the dry pollination system was dipped into a cup, which originally contained one gram of pollen. The brush was tapped against the side of the cup five times to remove loose pollen.
6. The dry pollination system was run so that the brush would contact the flowers.
7. In order to prevent any further pollination by any other means, the flowers were covered again using the paper bags and twist ties.
8. Steps 3 to 7 were repeated for all of the originally covered flowers.
9. After another week the bags were removed to allow the fruit to grow.
10. The fruit were allowed to grow for five months and were then harvested.
11. All of the fruit were weighed using scales.

The fruit produced using the dry pollination system varied markedly in size, as can be seen in Figure 86. The mean mass of the fruit was 100 g with a standard deviation of 32 g. This mean mass compares favourably with previously surveyed kiwifruit size distributions of commercial crops [134]. However, the standard deviation of 32 g is high and would be expected to be closer to 20 g [134]. This high standard deviation may be due to the fact that the areas of the orchard where these fruit were grown were not subject to thinning. Thinning may have resulted in some of the undersized and oversized fruit being removed, which would have resulted in a lower standard deviation. Thinning also may have resulted in the remaining fruit growing larger and hence may have resulted in a higher mean mass, as the kiwifruit vines would have put their limited resources into the remaining fruit.

6 out of the 50 flowers that were targeted by the dry pollination system did not form into fruit. This result was surprising because it seemed that all of the flowers targeted were contacted by the pollen carrying brush during pollination; however, it is possible that the pollen had reduced viability or that contact was not made between the pollen carrying parts of the brush and the flowers' pistils in some cases. It is also possible that the flowers may have been damaged either during the contact by the robot or during the handling of the flowers during the bagging and unbagging process.

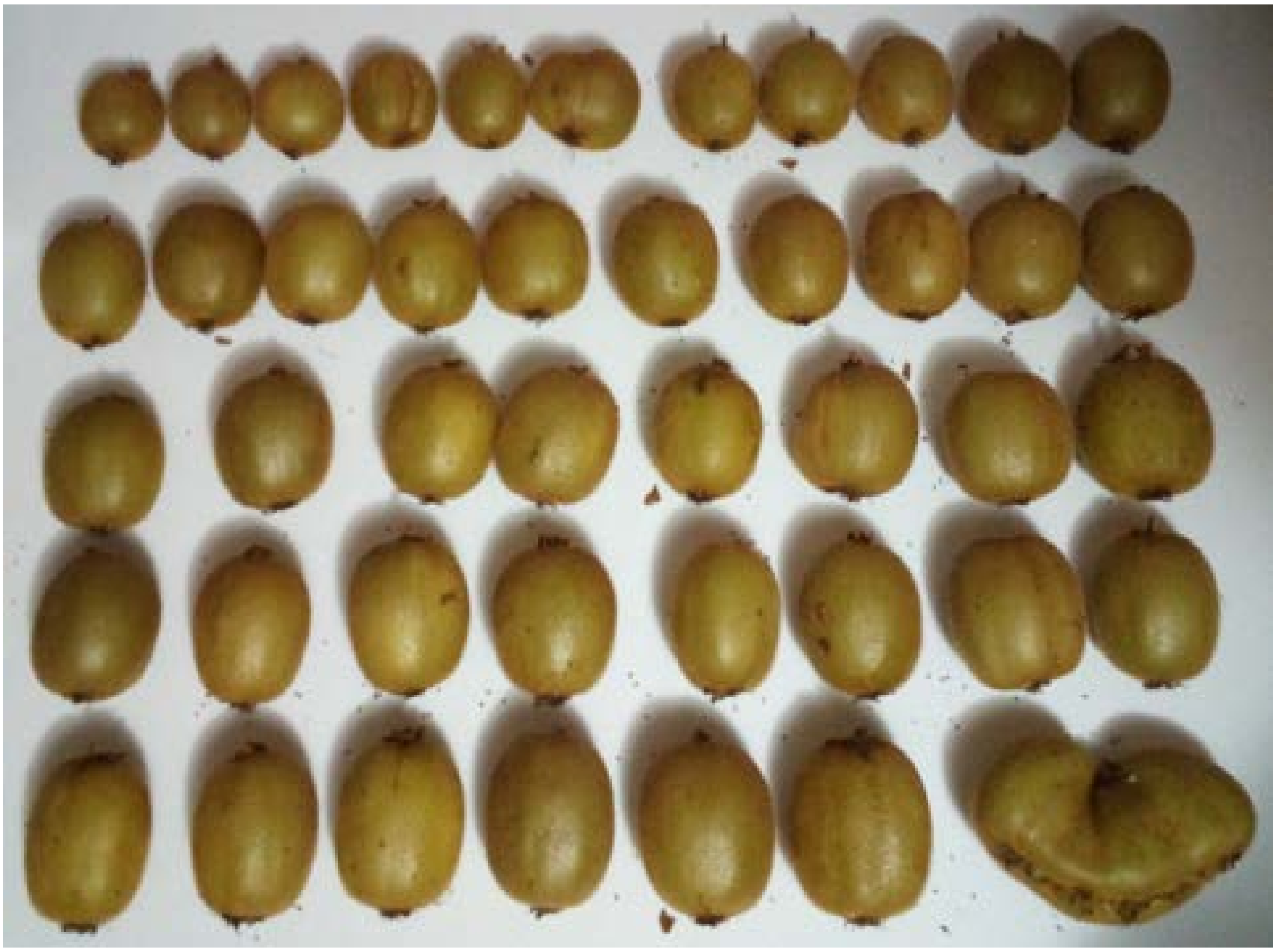
*Figure 86: The fruit produced after pollination by the dry pollination system.*

In future experiments determining the effect of bagging and unbagging the flowers might be determined with a control group, if there was a highly reliable form of pollination that could be used for this control group. Using the pollen carrying brush to gently perform manual pollination by hand is another experiment that might aid in determining the cause of the failed fruit set, since this test would remove the robot mechanical and control systems as a possible cause. In any case, the 88 percent fruit set achieved by the dry pollination system is not unreasonable with a general industry target of greater than 80 percent [16]. In addition, other pollinators, such as bees, may complement a dry pollination system.

## 3.7 Robotic Kiwifruit Pollination Discussion

In order to spray flowers with wet pollen, flowers were detected in colour camera images. Colour thresholding was used as one of the steps in the first algorithm trialled for flower detection for the pollination systems. Although these systems did successfully detect and spray flowers, it was found that, even within the constrained environment under the kiwifruit canopy and with powerful LED lighting, colour thresholding performance diminished when the ambient lighting changed.

Subsequently, flower detection was performed using CNNs, which worked much better and allowed the pollination system as a whole to achieve higher hit rates for flowers [135], [136]. However, only two cameras were used per set of spray nozzles for stereo matching and so, if either camera was compromised, it could be a potential cause of a failure of this system. A camera could produce useless data if it was affected by direct sunlight or was occluded by a hanging branch. Hence, a commercial system might require some sensor redundancy for a more robust system.

Boom navigation was performed to keep the spray boom of the pollination system close to the canopy, while colliding with only small amounts of the canopy. Using the minimum of canopy measurements from a lidar resulted in the boom navigation system not tracking the general height of the canopy well. Using the median of canopy measurements from a lidar generally tracked the canopy well; however, it was found that a rare case of low hanging solid branches could cause unwanted collisions with this system. In order to avoid such collisions, solid branches were detected by semantic segmentation and the height of these branches was measured using the combined output of two stereo matching algorithms. It is possible that the approach of comparing the output of two different stereo matching algorithms could be a useful approach for other applications, where mismatches should not be tolerated. The final version of the boom navigation system used lidar for canopy tracking and stereovision for avoiding collisions with solid branches. This system worked well in the orchards where it was tested. However, with different orchard maintenance practices, it is possible that a lidar only boom navigation system might be feasible. These practices could involve not training the canopy to have low hanging solid branches or pruning the low hanging soft branches in order to allow all branches to be avoided by boom navigation.

A separate dry pollination system was created, which was similar to the second kiwifruit harvester. It was found that contrast enhancement of the intensity channel of the Time of Flight data aided labelling of flowers. The final dry pollination system was able to detect and contact flowers with a goat hair brush for pollination. The fruit pollinated by the dry pollination system, without assistance from other pollinators, were comparable to existing commercial crops; it is assumed that this pollination performance could be even better with complementary pollinators, such as bees, other insects or wind. With such synergies it might be possible to also tolerate low rates of dry pollination without negatively affecting the overall quality of the crop and this could allow faster operation of the dry pollination system; however, such possible practical aspects are yet to be tested for their feasibility. The reuse of the hardware from the harvesting system is another potential benefit, which might greatly reduce the overall system cost and make kiwifruit orchard robots in general more financially viable.

## 3.8 Robotic Kiwifruit Pollination Conclusions

The contributions presented in this section were the vision, calibration and boom navigation systems for a wet pollination system and all of the components of a dry pollination system, which was based on the second kiwifruit harvester. The conclusions drawn based on experiments with these systems were as follows:

- Kiwifruit targeted pollination can be performed by calibrating using a solid target with fiducials aligned with the spray and by detecting flowers using convolutional neural networks.
- Kiwifruit pollination spray boom navigation can be performed using a 2D lidar for general tracking of the canopy and stereo cameras for detecting the height of solid branches.
- Robotic kiwifruit harvesting hardware, including Time of Flight sensors and robot arms, may be reused to perform pollination by replacing the harvesting end effectors with brushes.
- Robotic kiwifruit pollination using a brush and dry pollen can produce kiwifruit with statistics comparable to commercial crops.

## 3.9 Future Work in Robotic Kiwifruit Pollination

Existing research has suggested that nylon fibres may be a more effective surface for pollination by contact, particularly if coated with an Ionic Liquid Gel [113], [114]. It would be interesting to test if such a pollen carrying surface would produce better fruit with the dry pollination system. In addition, systems will have to be developed to automatically apply pollen to the contacting surfaces and to ensure that not too much pollen is wasted during the process of dry pollination so that all flowers receive the optimal dose of pollen.

Many performance metrics of the dry pollination system were not measured in this research. Assuming that autonomous harvesting is economically viable, autonomous pollination needs to be about three times quicker than harvesting, since the pollination season is about a third of the length of the harvesting season. Future testing of dry pollination should concentrate more on failure modes, failure rates and the speed of the system.

# 4 Autonomous Driving in Orchards

It was a requirement of the MBIE Multipurpose Orchards Robotics Project that the robots, performing harvesting or pollination, should drive throughout the orchards autonomously. It was also a requirement that the autonomous driving systems should use more than one type of sensor. As a result, this section is broken up into subsections, firstly covering navigation using lidar and then row following using computer vision. Another complicating factor was that autonomous driving in apple orchards was also a requirement; although, this was a much less important feature and so this section only touches on navigation in such environments.

## 4.1 Orchard Navigation Literature Survey

This literature review starts by discussing existing methods for autonomous navigation in orchards. It then becomes more general, exploring analogous areas and more widely applicable research.

In the MBIE Multipurpose Orchards Robotics Project, orchard navigation was considered in terms of two different orchard environments, which differed in their structure. These orchard structures were pergolas and tree walls. The distinction was made between pergolas and tree walls because some algorithms proposed for tree walls would not work well for pergolas and vice versa.

In pergola structures, there is a sparse array of posts and trunks which hold up a ceiling like canopy, which consists of branches, leaves, beams, wires and flowers or fruit (Figure 87). This type of structure has been used to grow kiwifruit, passionfruit, grapes and nashi pears [137]–[139]. However, navigation in pergola environments is not extensively researched.

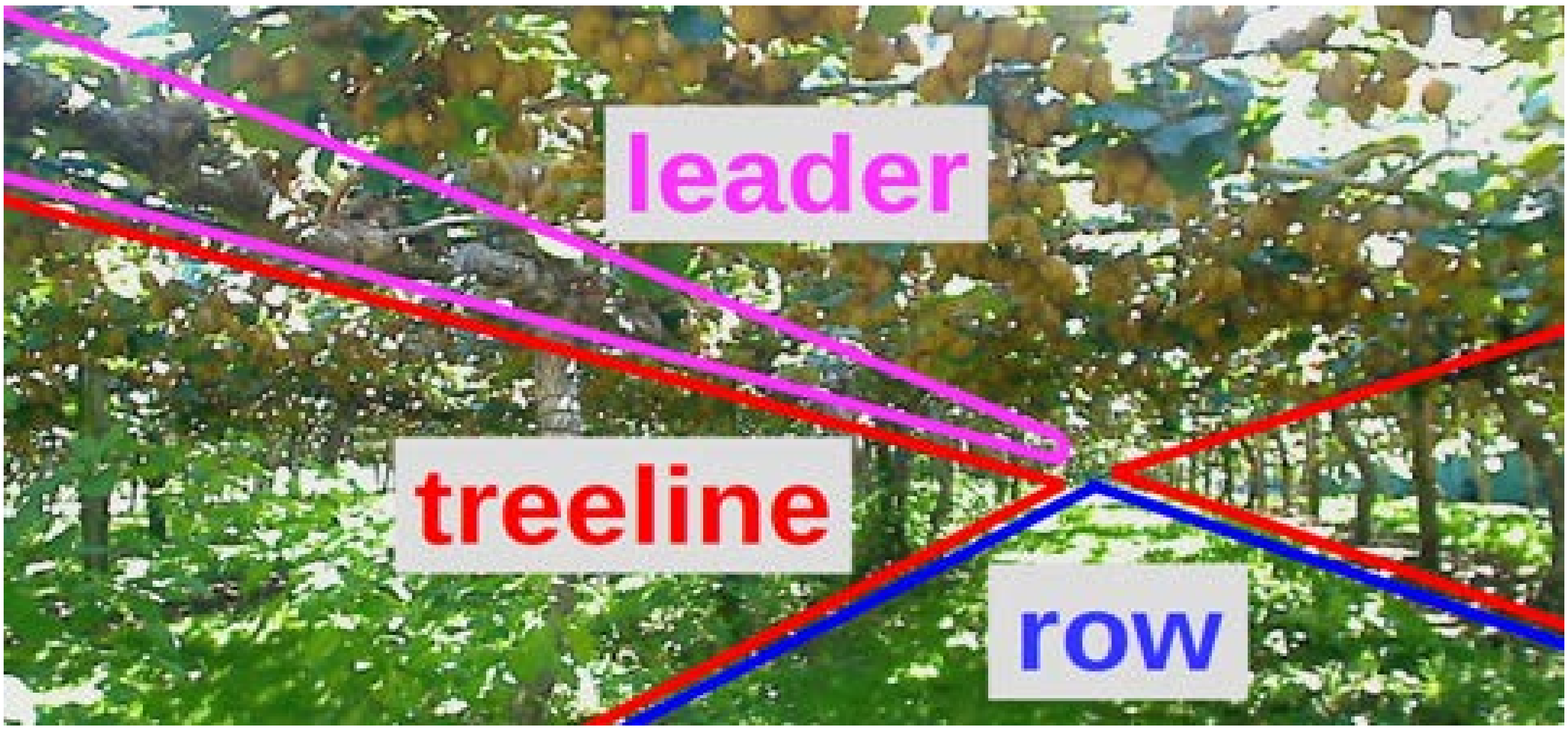

*Figure 87: Looking down a row in a pergola structured kiwifruit orchard.*

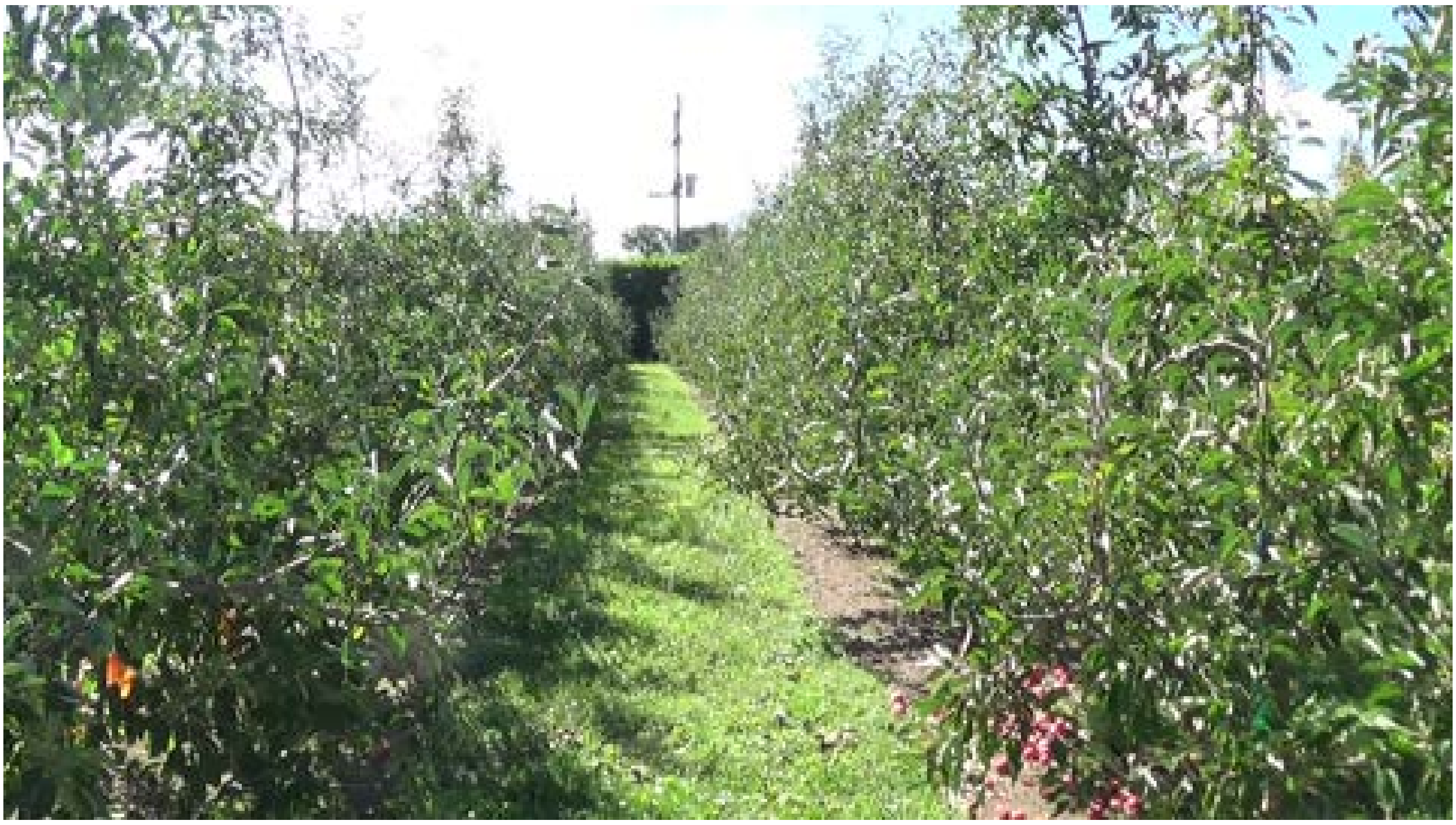

*Figure 88: An example of tree walls on both sides of a row.*

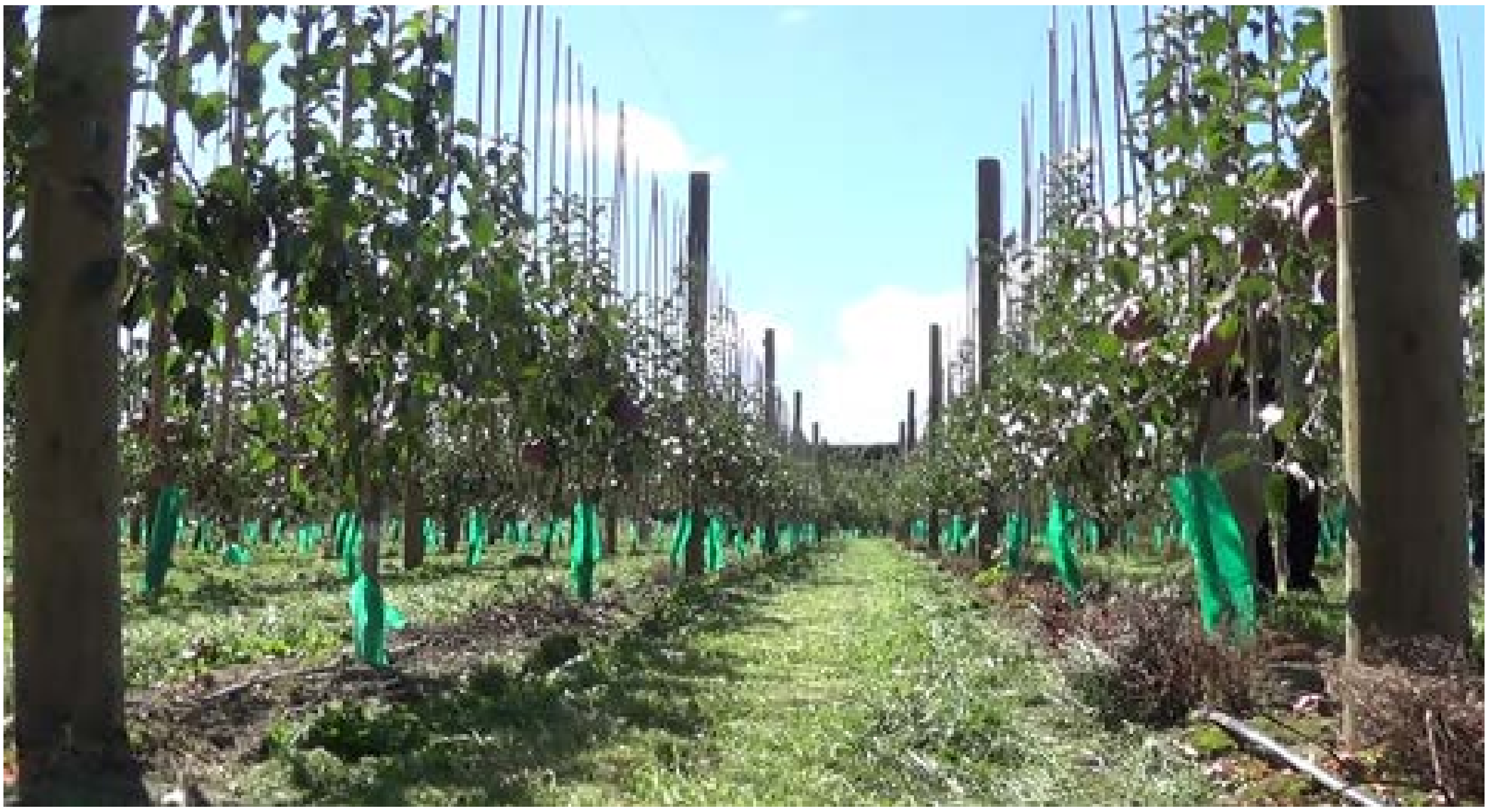

*Figure 89: Another example of tree walls on both sides of a row.*

Research on navigation in the more common type of orchard, where the trees form dense walls on both sides of a row, is much more common. Here, a general term of “tree wall” is used to encompass a variety of freestanding and trellis structures including those shown in Figure 88 and Figure 89. Many fruit trees are grown with tree wall rows, including the apple trees in Figure 88 and Figure 89.

### 4.1.1 Navigation in Pergola Structured Orchards

An early method for navigation in kiwifruit orchards used GPS and computer vision to detect trees and posts [66]. However, this method was later criticised as unreliable in high contrast lighting conditions and was superseded by a method using horizontal 2D lidars and a digital compass [21]. This method of lidar based navigation was developed for an autonomous mobile kiwifruit harvester. The navigation algorithms were designed to perform row following, turning between rows and navigation with respect to a fruit bin. The steps in the algorithm for determining the relative pose of the fruit bin were:

1. Data was collected from a horizontal 2D lidar.
2. The size of the bins was used to determine candidate clusters of points from the lidar data.
3. Hough transforms were performed on the candidate clusters and clusters that had one straight line or two perpendicular straight lines were kept.
4. The variance of points from the straight lines was measured and clusters with high variance were eliminated.
5. A final check of the bin dimensions was performed.

This algorithm did not account for the possibility that a bin might be close to another object, which could cause the algorithm to discard the data from the bin in the first size check and give a false negative detection. In addition, this algorithm was only tested with and without bins; other objects were not tested to determine if false positives could occur. As a result, it is unclear if the algorithm would give false positives in the presence of other objects or combinations of objects; for example, it seems possible that a group of people or a farm vehicle could be recognised as a bin.

In addition to navigating to a bin, a method was developed for row following, which used a digital compass and a horizontal 2D lidar to find the heading of a row in a kiwifruit orchard [21]. The steps in this algorithm were:

- Consecutive points from a lidar scan were clustered, based on the distance between them.
- Data points that were not clustered (singular points) were discarded from row heading detection.
- Clusters of more than 26 lidar points were discarded.
- Clusters spanning less than 0.1 m or more than 0.25 m were discarded.
- Clusters that were more than 4 m from the digital compass heading for the row were discarded.

- Clusters were grouped according to whether they were on the left or right side of the digital compass heading for the row.
- The centre of mass of each cluster was calculated.
- Every left cluster was paired to the closest right cluster.
- The distance between each left cluster and right cluster in a pair was checked against a range from 3.5 metres to 6 metres and pairs with distances outside of this range were discarded.
- The midpoint between the left and right clusters in a pair was calculated for each pair.
- A line of best fit for the midpoints was calculated as the row heading.
- Midpoints more than a threshold distance from the line of best fit were discarded and the line of best fit was recalculated.

This method above discards single points of data, which may not always be valid. It is possible that a single data point may represent a young tree with a thin trunk, a tree or post that is far away or a useful feature that is partially occluded. It may be useful to preserve such data to improve the accuracy and robustness of the row detection method.

It may also be possible to implement the row heading detection without the digital compass. The use of the digital compass not only incurs the equipment cost but also requires that the heading of the row is known beforehand. Furthermore, data from adjacent rows that is discarded may possibly be used to find the general heading of multiple rows, as a starting point for more precisely determining the heading of the current row.

### 4.1.2 Driving in Orchards with Tree Walls Using 2D Lidar

Subramanian et al. [140] described separate lidar and computer vision navigation systems for an autonomous tractor, which was developed for use in citrus groves. The camera and lidar used by the navigation systems were mounted on top of the cab of the tractor and were angled down. The lidar row detection method searched for the centre of two discontinuities in the lidar data. This method was tested with hay bales set up as walls on both sides of a path. Testing in an orchard was avoided because it was decided that the additional complexity and inconsistency of structure in the orchard environment could cause the system to not work; solutions to this problem were not provided.

In a kiwifruit orchard, the row detection system of Subramanian et al. [140] would have even less consistent data because the pergola structure does not form continuous walls like stacks of hay bales. In areas where poles or trunks in the pergola are not symmetrical about the row centreline, which are common in kiwifruit orchards, the lack of symmetry would affect the centreline calculation using the method of Subramanian et al. [140]. Furthermore, with the angled down lidar, there would be the possibility of no valid features being detected between the posts and trunks of a pergola structure. In a pergola, a 2D lidar plane would intersect more features on average if the lidar plane was horizontal and this is the configuration that has been used more commonly for autonomous orchard driving both in pergola and tree wall rows.

Bayar et al. [141] described the navigation systems of a family of autonomous orchard vehicles. Data for localisation was collected from steering encoders, wheel encoders and a horizontal lidar, which was mounted on the front of each vehicle. At the end of rows, turn paths were generated out of straight and curved sections to approximate the shape, given by a previously calculated clothoid. Localisation was performed by odometry down rows of known length; the odometry was reset at the start of a new row. Row following was performed using the lidar to measure the distance to trees on each side of the row, calculating a line fitting the trees and then calculating the linear and angular offset of the vehicle. For headland turns, the end of the row was detected as empty space.

Barawid et al. [142] used data from a horizontal 2D lidar on the front of a tractor. Hough transforms were used to find the treelines and hence the tree row centreline. Auto regression was used to reduce noise in the centreline measurement. Testing at $1.43ms^{-1}$, the RMS linear and angular errors for the measurement of the row centreline were 0.36 m and 4.9°, respectively. Greater errors were found where there were breaks in the treeline. These reported errors seem quite large and this work did not account for protruding branches or other objects in front of the tractor.

The row detection methods of Bayar et al. [141] and Barawid et al. [142] are similar in that they fit lines to tree walls. However, these methods could produce erroneous measurements in a pergola structured orchard. Potential causes of these erroneous measurements could include 2D planar lidar scans not intercepting the desired features on sloping ground. Consider the scenarios in Figure 90 and Figure 91. The lidar scan plane does not reach the desired features, which are the trunks and posts, because of the slope of the ground. Figure 92 shows a lidar scan of this type of scenario in a gently sloping kiwifruit orchard. The canopy appears as a cloud of points and clusters. If the lidar detects the canopy at a short range, this could affect the result given by line fitting or a row detection algorithm. Note that in an orchard with tree walls, there is no overhead canopy to be detected as a cloud of data; however, the ground may produce a similar effect. A potential solution to this would be to have the lidar higher in a tree wall orchard. However, this might increase the number of obstacles that can fit under the lidar scan plane, reducing the effectiveness of obstacle detection. Furthermore, raising the lidar plane might not work in a pergola because it would increase the amount of canopy affected data.

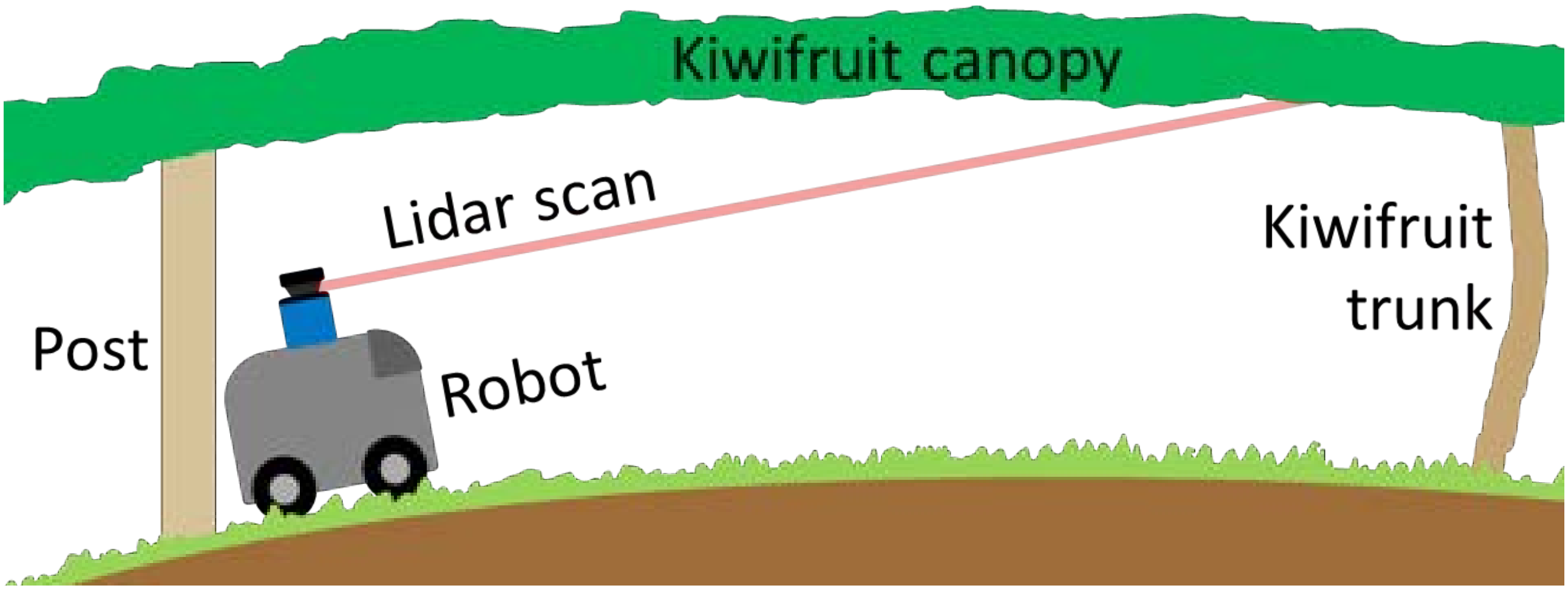

*Figure 90: In a kiwifruit orchard on a convex slope a 2D lidar scan plane may reflect off the canopy instead of posts and kiwifruit trunks.*

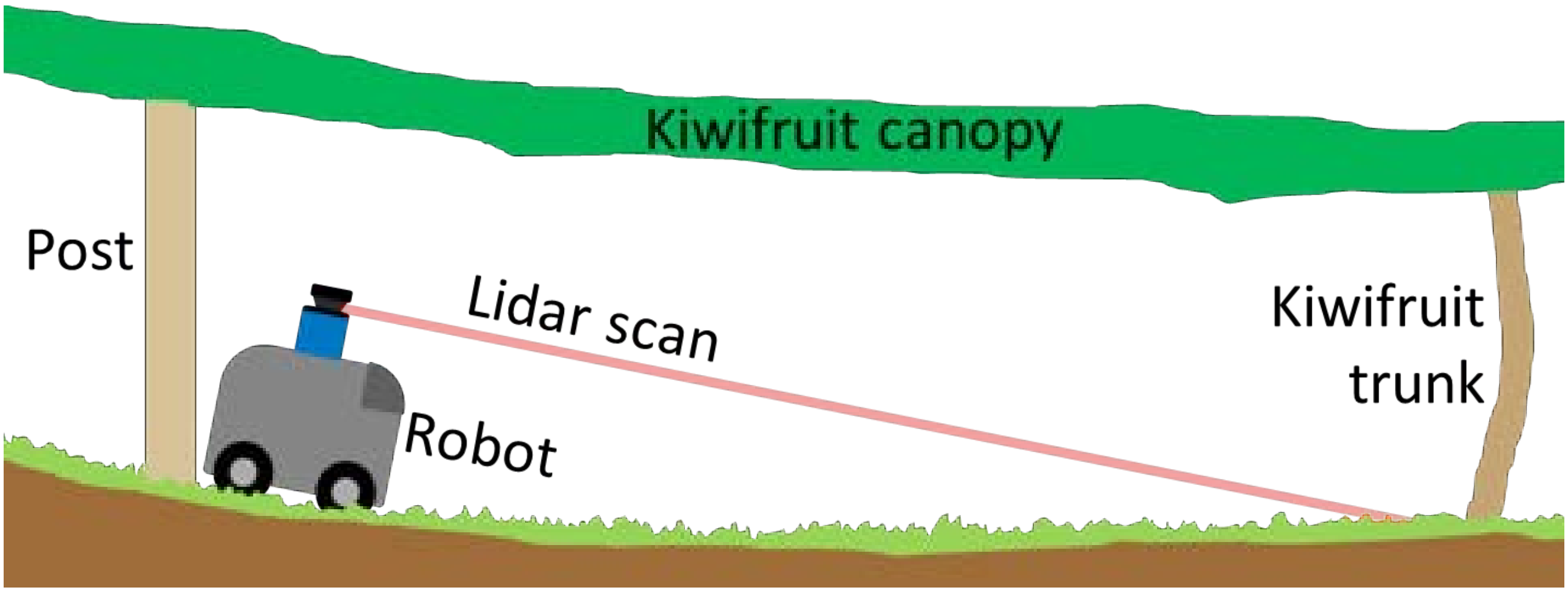

*Figure 91: On a concave slope a 2D lidar scan plane may reflect off the ground instead of posts and kiwifruit trunks.*

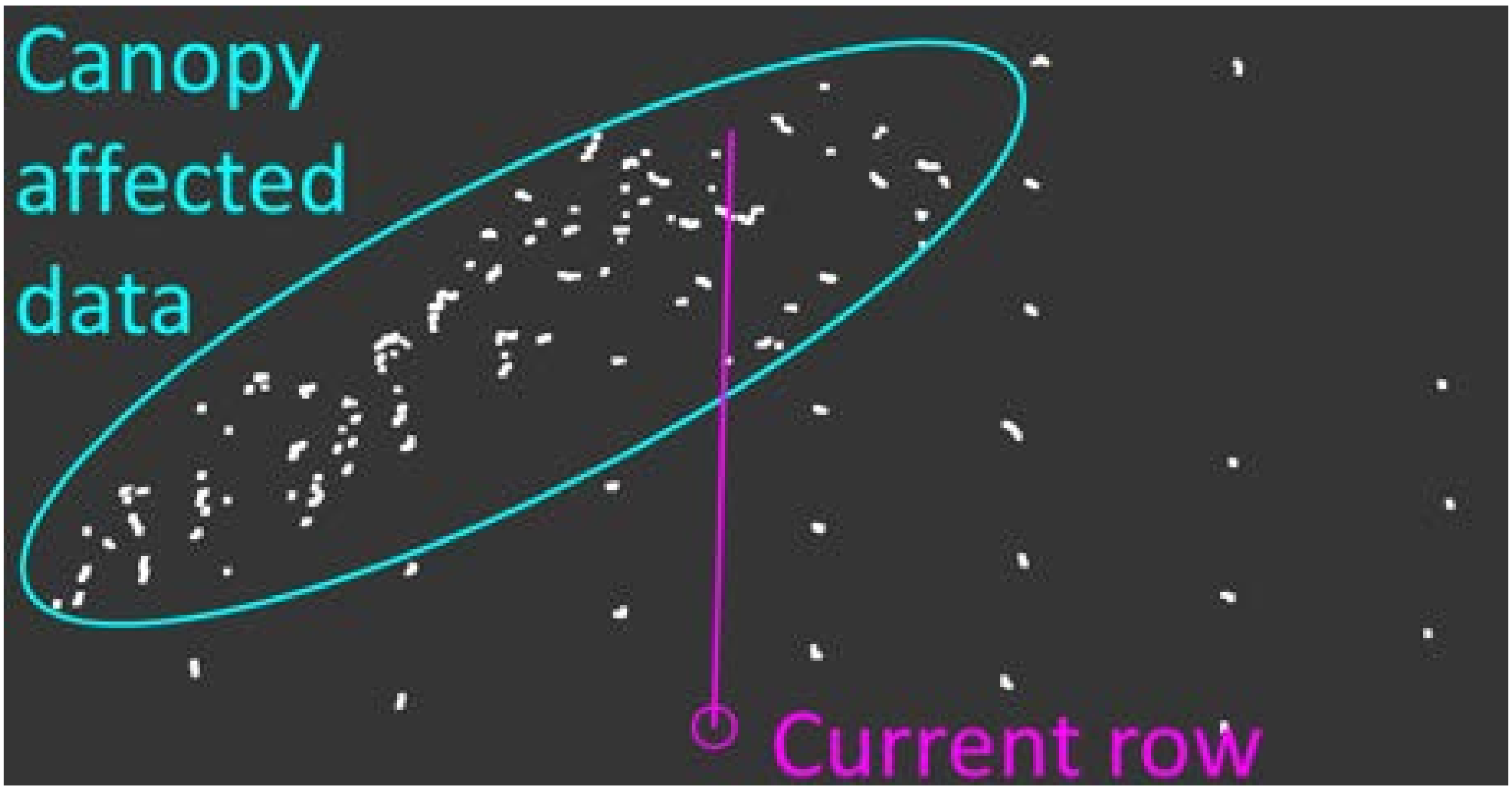

*Figure 92: A 2D lidar scan where a region of the data is formed by reflections off the canopy rather than from kiwifruit trunks and posts.*

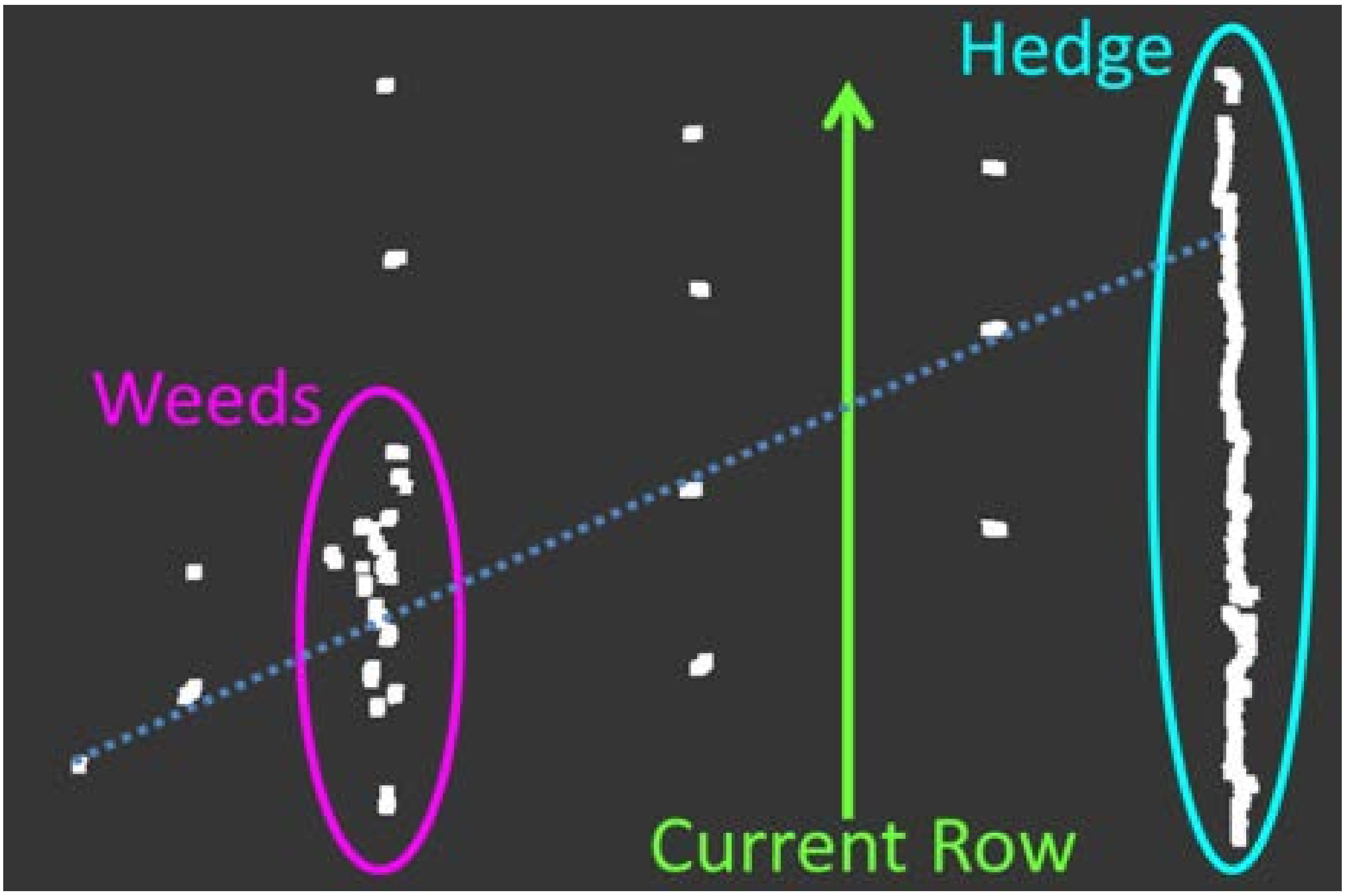

*Figure 93: A horizontal lidar scan from a kiwifruit orchard, where the strongest straight line candidates are not the current row.*

Furthermore, unstructured or unexpected objects in the current row or close by could cause the line fitting of a row detection method to be skewed or completely wrong. Examples of such objects could include weeds, hanging branches, farm vehicles, bins, hedgerows and people. These features may form straight lines. For example, in Figure 93, approximately straight lines are formed by the weeds in an adjacent row, the hedge on the orchard block perimeter and diagonals from a mix of features. As a result, the best fitted lines from a method such as the Hough transform may not be the current row. Note that Figure 93 demonstrates another issue with row detection in pergola structures; in general, approximately orthogonal to the direction of the rows, the regular planting structure of the orchard forms false rows, which are straight and apparently free space, according to the data. However, it is undesirable to drive through this space because it is commonly where irrigation pipes are run and leading branches of the kiwifruit form solid obstacles.

Another effect of weeds is shown with a camera image of the weeds and the corresponding 2D horizontal lidar data in Figure 94. In this situation the weeds are slightly to the right of the centre of the row. The weeds block the lidar from collecting data from the tree trunks and posts behind the weeds. As a result, there is less data to determine the treeline to the right of the row and the data from the weeds may be erroneously incorporated into the row detection estimate using the methods of Bayar et al. [141] and Barawid et al. [142]. Incidentally, the weeds also appear as obstacles and hence may cause the autonomous mobile robot to stop, even though it is safe to proceed. Just as weeds can affect row detection, so can hanging branches in a pergola. An example is shown in Figure 95, where hanging branches cause the row detection estimate to skew to the left, using an implementation of an existing row detection method [21], [143].

Riggio et al. [144] presented a navigation method for a vineyard with structural similarities to a pergola structured orchard. However, this work did not discuss how to overcome data from the ground below and canopy above on sloped ground with a 2D lidar.

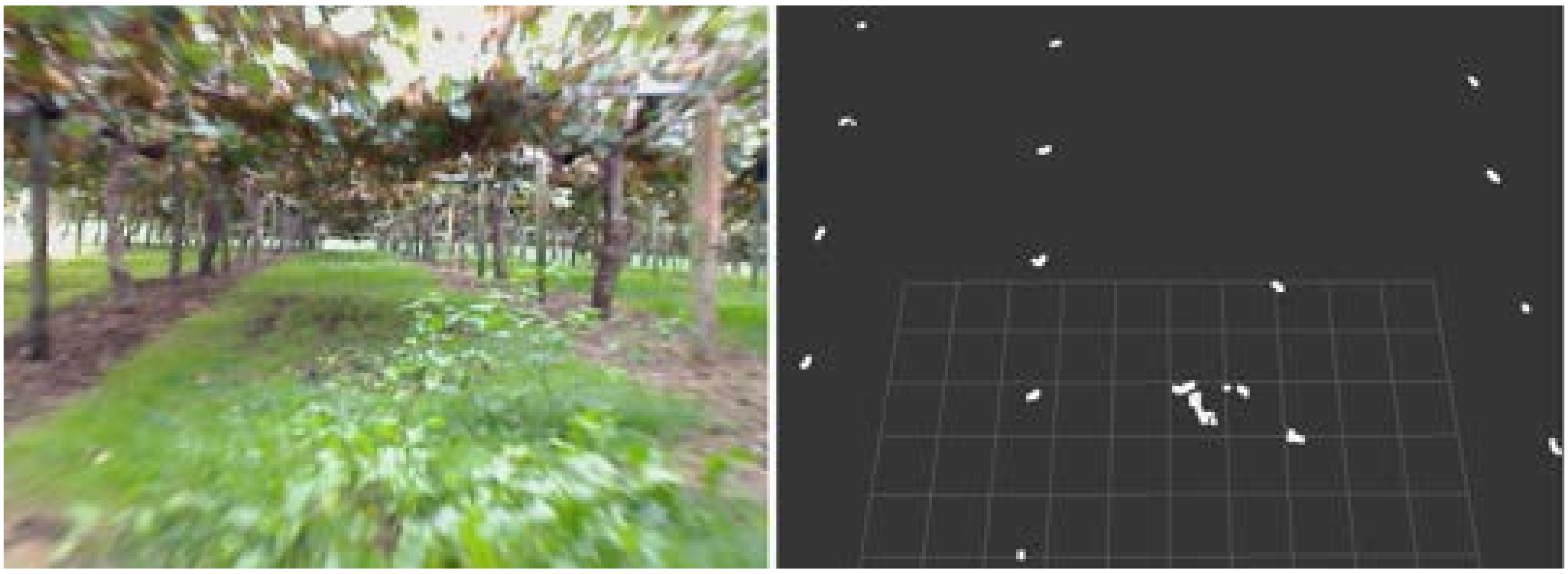

*Figure 94: A camera image of weeds blocking the path down a kiwifruit orchard row (left) and lidar data at the same point in time where weeds appear as obstacles and block the visibility of trunks and posts behind them in the current row (right).*

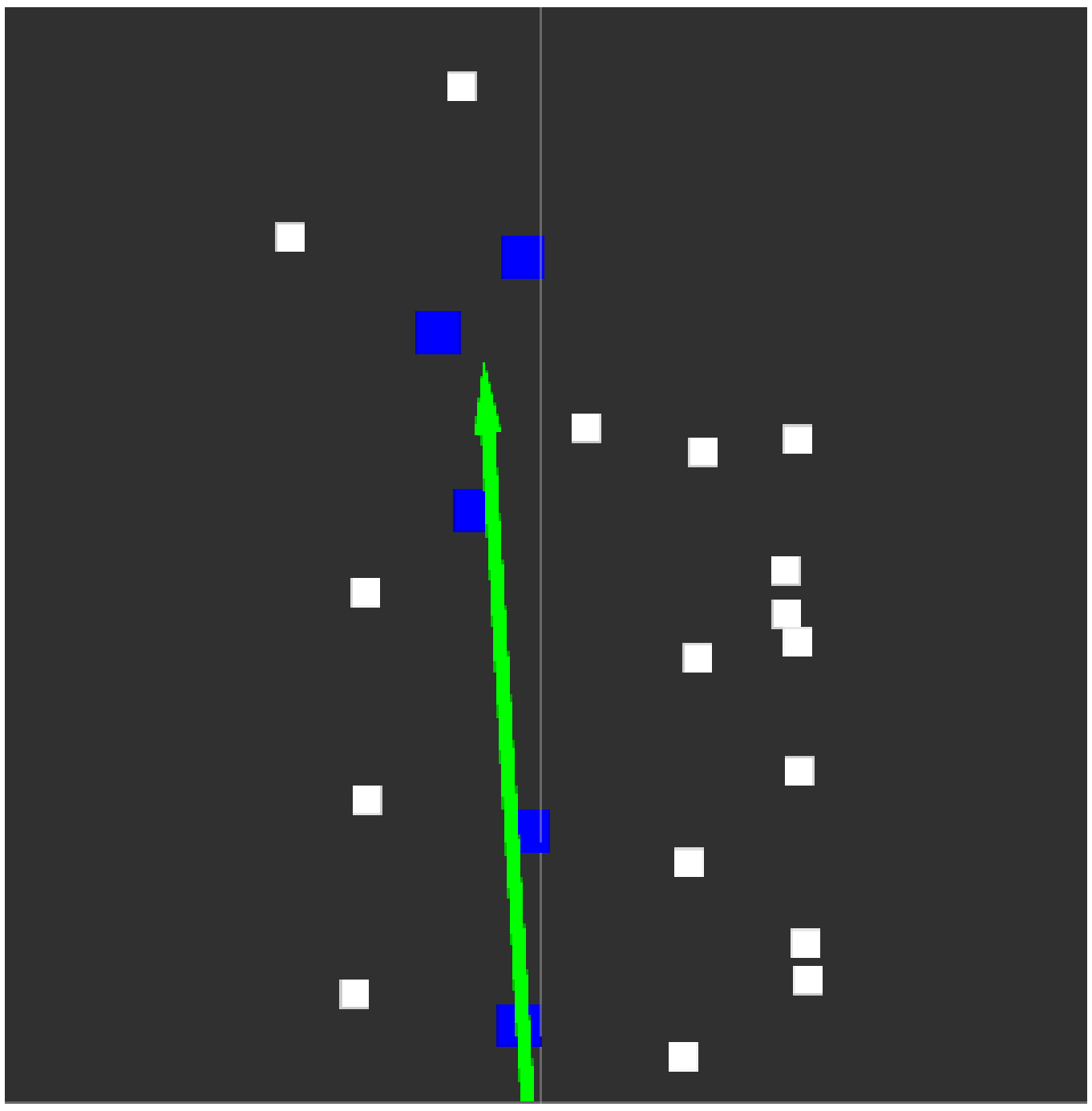

*Figure 95: An example of hanging branches affecting row detection, where white points are extracted data- of which only 7 appear to be posts or trunks and the other 10 are hanging branches- blue are calculated centre points and green is the resulting row centreline estimate.*

Hansen et al. [145] described a method that used a horizontal 2D lidar but did not use treeline detection and row following as described by Bayar et al. [141] and Barawid et al. [142]. This method was used in the navigation system of an autonomous tractor for use in apple orchards. The sensor feedback used for localisation was from a gyroscope, odometry and a lidar mounted horizontally on the front of the tractor. The pose state of the tractor was predicted using the gyroscope and odometry; this pose was corrected by matching line features to a map of the orchard. The method used to find line features from the lidar data involved firstly predicting a line using the map and the vehicle pose estimate and then taking points within a threshold distance from the predicted line, performing linear regression on the points and checking the result. An Extended Kalman Filter, second order Divided Difference Filter, Linear Regression Filter and Unscented Kalman Filter were all implemented and tested with similar results across all methods.

Hansen et al. [145] did not describe their method for line matching. However, given that the orchard changes markedly throughout spring and summer, due to the growth of plants, and also given that the lidar was targeted at the height of the branches, where there is the most growth, it seems it would be challenging to match features to indicate how far down a row a tractor was, based on maps formed much earlier in a season. In this case, the localisation along the axis of the rows might mostly be performed at the end of the rows, which is similar to the method of Bayar et al. [141]. However, by relying on driving according to the map, the method of Hansen et al. [142] does not allow for objects that are not in the map, such as branches growing towards the centre of the row, which can be seen in Figure 88, or other objects that may be in the row.

Marden and Whitty [146] presented similar work to Hansen et al. [145] for navigating in vineyards. However, this work did not describe methods for extracting useful data from a horizontal 2D lidar with the laser beams intercepting the ground below and the canopy above at short ranges in a pergola structured orchard.

### *4.1.3 Driving in Orchards with Tree Walls Using 3D Lidar*

There has been work done on using 3D point clouds from rotated 2D lidars to perform navigation for autonomous vehicles in orchards. In one example of such work, a Hokuyo 2D lidar facing the direction of travel was rotated about its forward heading axis from -55 to 55 degrees about horizontal by a motor and an encoder was used to measure the angle of rotation of the lidar [147]. Each lidar scan was registered with existing scans using wheel encoder odometry to calculate the displacement of the new scan from the existing data. The Iterative Closest Point (ICP) method was used to improve the concatenation of new lidar scans to the existing point cloud by matching the closest points from the lidar scans, while iteratively improving the linear and angular displacement calculated from odometry. RANSAC was performed on parallel line calculation to fit lines to the rows of trees. An Extended Kalman Filter, using odometry in the prediction step and the RANSAC line fit in the measurement step, was used to smooth out noise in the line fitting. Then for each side of the row, the point cloud data was separated into horizontal bands and for each band in a region around the fitted line the variance of point displacements with respect to the fitted line was calculated. The band with the lowest variance was deemed to be the band of interest. The middle of the band of interest was set as the height and the offset of the fitted line was refined to be the peak density of points in the band of interest. A particle filter was also used to find the closest trunks on each side of the row. Pure pursuit was used to follow the detected row. Experiments were conducted using an Applanix Pos-LV for ground truth measurement. This work demonstrated that the proposed use of 3D point clouds could work relatively well in structured and unstructured rows; whereas, for 2D scan data, the row detection only worked relatively well for structured rows.

This method attempted to remove the effect of branches that grow towards the row centre and treated such data as noise. However, such branches may be significant obstacles. In this case, it may be better to find the centre of the free space down the row, rather than the centre between the treelines.

Some other potential weaknesses of this work are the assumptions that the ground is relatively flat, which was not necessarily the case in many kiwifruit orchards observed. The flat ground assumption affects the accurate registration of the lidar scans. In addition, it is unclear if the step used to separate the left side and right side of the data would work with larger angular displacements, since for larger angular displacements, data from one side of the row may be beyond the halfway point of the lidar scan.

### 4.1.4 Driving in Orchards with Tree Walls Using Cameras

Subramanian et al. [140] used computer vision to find the centre of orchard rows. Their method used different colour thresholds for different lighting conditions, morphological operations, line fitting of the trees on each side and found the centreline between the treelines.

He at al. [148] performed row detection for Gala and Fushi apple orchards using monocular vision. An analysis of histograms of the colour channels of the images suggested it would be possible to distinguish between trees and the ground using the red and blue channels. Hence, the red and blue channels were extracted. Colour thresholding using the Otsu method was performed to give a binary image. Erosion was performed to reduce noise and dilation was performed to fill gaps. To further reduce noise, blobs were selected for removal based on a size threshold. To extract out the area with ground as the region of interest, a sum of values in each row of the image was taken and where the sum crossed a threshold was where the limits of the region of interest were set. Further noise removal was conducted with another stage of blob removal based on blob area. The features that were extracted for row detection were the transition between the bottom of the trunks and the ground. These were found by searching for the lowest corner feature in a vertical column region. The trunk base features were split into containers for the left and right of the image. Lines were fitted to the trunk base features using sum of least squares regression and the centreline was found as the average of the two treelines.

He et al. [148] noted that branches on the ground after pruning could cause noise and processing such noise was deferred as future work. They also noted that detections failed in some images with shadows and with the presence of objects of similar colour to the ground amongst the trees or in the background. These false detections expose some of the difficulties of using methods that rely on colour, which changes with lighting [149]. In images shown in the paper of He et al. [148], the ground was either all brown or had large brown patches, which made it distinctive; however, such colour differences might not be as strong in other orchards.

Other potential issues with this work included the use of the base of the trunk as a feature, which would commonly be occluded by weeds, branches or other objects in orchards. The method for splitting the image into left and right sides of the row would not work well for larger angular offsets.

Torres-Sospedra and Nebot [150] used the mean, energy and coordinates of regions of pre-processed images as inputs to an ensemble of fully connected neural networks with 3 layers. The neural network classified the regions in the image as ground, tree or sky. The resulting output was processed with morphological operations to reduce noise and edge detection was used to extract the treelines. At different horizontal slices, the Hough transform and row centreline calculation was performed.

Sharifi and Chen [151] also performed row detection on colour images of orchard rows. They used mean shift clustering and classified the resulting clusters as sky, ground or trees according to their colour. The ground pixels were extracted and the Hough transform was used to fit lines for both sides of the row. The row centreline was calculated from the two treelines. The results given in the paper show that the method does not necessarily work as intended, with the treelines being fitted to the mulch under the trees, instead of the trees themselves, in one case. As with the work of Subramanian et al. [140], He et al. [148] and Torres-Sospedra and Nebot [150], issues arising from shadows, colour changes, occlusions, weeds, stray branches and other objects, which might be present, were not addressed.

To solve issues with shadows in orchards, it seemed it may be possible to use one of many shadow removal techniques, some of which work in real time [152]–[154] and have been successfully applied to place recognition for autonomous driving [155]. Issues with lighting and colour changes may be solved by moving away from using colour to instead using Convolutional Neural Networks, which have successfully been applied to scene labelling [33]. Shadows, lighting and colour may be solved together by using stereovision. Issues with occlusion and stray branches, affecting the row centreline measurement, may be solved by considering the free space, as opposed to the planted structure of a row.

### 4.1.5 Driving in Orchards with Tree Walls Using Cameras and 3D Lidar

A practical implementation of an autonomous tractor system was described by Moorehead et al. [156]. Autonomous tractors were used for mowing and spraying in citrus groves. The tractors had accurate localisation data from a RTK-GPS unit, which enabled path following on a predefined map. However, RTK-GPS did not account for changes in the trees due to growth and hence data from a nodding 2D lidar was used to assist in the row following. Data from the 2D lidar was used to detect ground and objects by decomposing the data into a grid, where the height of each grid element was a probabilistic combination of sensor data, sensor noise and adjacent grid values. It is not clear how the algorithms of this system would work in a pergola structured orchard. Decomposing the data into a grid of ground and objects may be more challenging in a pergola since the canopy may be classified as objects, in which case there would be objects in all directions. Even, if most of the canopy was filtered out, branches hanging from the canopy might still be classified as objects, even when they can be safely driven through. In addition, the reliance on GPS might be an issue under the kiwifruit canopy since the GPS signal may be blocked by foliage.

### 4.1.6 Mapping and Localisation without Autonomous Driving in Orchards

Localisation has been performed in almond orchards using a SICK LMS-291, performing a two-dimensional lidar scan vertically at 75 Hz, by Jagbrant et al. [157]. The vertical scans were used to create 0.2 m wide slices of the orchard, from which height and volume was measured. Segmentation of trees was performed using a Hidden Semi-Markov Model with state transitions between trees, boundaries of trees, gaps between trees and small trees; this method was proposed as an improvement on previous methods that relied on the separation between objects, which was found to cause two objects to be detected as one in some cases. Individual trees were recognised using the sequence of heights from the 0.2 m wide slices. Localisation was performed by matching a sequence of recognised trees to a previously created map and controlling state transitions with a Hidden Markov Model. The map was created using GPS and INS sensors to register the lidar data; however, GPS dropout affected the accuracy of this method so it was suggested that odometry might be used in the future. During localisation, a row traversal constraint was enforced, which stated that a new row could not be entered until the current row was exited. Row transitions were handled by assuming that the next row could be any row in the orchard with heading being in either direction and hence each row entry point was given equal probability when exiting a row. Additional future work identified included adapting to seasonal changes in the orchard due to growth of the trees.

The general approach of Jagbrant et al. [157] for recognising individual trees seems promising and a similar approach has been applied in an apple orchard by Bargoti et al. [158]. However, apple and almond trees with a vertical growing structure have a different appearance to kiwifruit vines in orchards, which have a clear trunk and a ceiling-like canopy. The methods used to recognise individual kiwifruit vines might be quite different to those used for almond orchards, especially since there is no clear separation between vines in the kiwifruit canopy, which is dense with branches that overlap between adjacent plants.

Using the fixed vertical lidar configuration of Jagbrant et al. [157] and Bargoti et al. [158] for mapping and localisation removes the ability to use the same sensor for free space detection, obstacle detection and obstacle avoidance in the forward direction. A horizontal 3D lidar system may provide the benefits of both of these vertical lidar methods and horizontal 2D lidar approaches.

Nielsen et al. [159] used GPS, wheel encoders and an inclinometer with a Kalman filter for localisation in rows of a peach orchard. Stereo cameras and lidar sensors were directed towards the trees in order to collect information about the trees. Trees were segmented by iteratively fitting a mixture of Gaussians to the point clouds, given by the stereo cameras and lidar sensors. It is unclear how accurate and reliable the localisation system was using the GPS; however, other research has reported unreliable GPS signals in similar environments [157], [160]. The methods used to segment trees may be applicable to apple orchards because of the structural similarity to the peach orchards; however, these methods may be less applicable to kiwifruit orchards, because of the structural differences of kiwifruit orchards, compared to peach orchards. Again, by directing the sensors sideways, the opportunity to use the same sensors for obstacle detection and avoidance was lost.

### 4.1.7 Agricultural Mobile Robot Navigation

As already discussed, using a straight line Hough Transform [161] might extract unwanted features. Instead, Winterhalter et al. [162] used a Pattern Hough Transform to find multiple line matches at the same time with a variable offset, direction and spacing. By detecting the pattern of multiple rows, rather than individual rows, they were able to reduce errors from noise in the data. They used this method to find the direction of rows in broadacre crops. It seems that a Pattern Hough Transform might be used for detecting structure in a kiwifruit orchard; however, it seems that a Pattern Hough Transform could produce erroneous measurements with patterns in the directions transverse to the true row direction, which can be seen in pergola structures.

Zhang et al. [163] performed mapping using a bespoke 3D lidar and odometry data from steering and wheel encoders. Their algorithm used odometry and point cloud registration to map features down the rows. Loop closure was enabled with the use of installed landmarks, which were reflective beacons on posts at the end of rows or reflective material mounted on pipes, which were installed at the end of rows where posts were not available. The necessity to make, install and maintain the reflective beacons is an undesirable characteristic of this method because it requires additional site commissioning time, manual labour and maintenance; hence, ideally if this step could be avoided, it should be avoided. However, the necessity to introduce artificial landmarks suggests that perceptual aliasing and data association may be key problems to overcome for the front end of a SLAM algorithm in an orchard environment.

### 4.1.8 General Robotic Navigation

Kim and Eustice [164] used an active approach to SLAM, where the planning of the robot path took into account potential improvements to the performance of the mapping. The cost of revisiting feature rich areas for loop closures was balanced with the pose uncertainty in order to plan routes that produced improved maps efficiently. This approach may not be required in kiwifruit orchards because of the high visibility of many features- such as posts, trunks and hedgerows- from most poses in an orchard. However, it may be more useful in tree wall structures, which may have less widely visible and less distinctive permanent features to map.

Although there have been numerous successful demonstrations and implementations of SLAM algorithms, there are also open issues and common problems with current SLAM algorithms, including:

- Robots being able to share and update previously built maps over extended periods of time;
- Achieving robustness sufficient to achieve long term autonomy without human intervention;

- Automatic recovery from error states is a challenge for existing SLAM algorithms;
- Perceptual aliasing, due to different locations producing similar patterns in the sensor data, which can lead to errors in the data association;
- Excessive numbers of dynamic, complex and changing elements, which may result in maps containing a larger proportion of elements that do not represent the environment at a later point in time, which can affect the accuracy of data association [165], [166].

It was hypothesized that performing SLAM in orchards could be vulnerable to perceptual aliasing due to the repetition of structures throughout the environment. In addition, orchards contain a large number of elements that change over time. Orchards can contain people, vehicles, bins and other objects, which may move around as typically troublesome dynamic elements for SLAM. These elements can be mapped or may produce data association errors. In addition, various plants in the orchard undergo significant growth throughout a year, so that a map made in winter may not contain many elements that have grown by harvest time in the following autumn. The changes in the plants include:

- The canopy starts as sparse and tidy, with branches tied down and no leaves in winter. By harvest time, there is thick leaf cover and many more branches, which grow in various directions, including into the path of the robots. In addition, the canopy becomes heavy and sags in the order of 0.3 of a metre in places. As fruit is harvested, the canopy rises again.
- Weeds can grow very high in kiwifruit orchards (Figure 96). Some of these weeds may appear as vertical objects and could be mapped or may block the detection of mapped features. However, these weeds may be mown, sprayed or may die naturally and hence they are only temporary.
- Trees may be removed, replaced or grafted, affecting the appearance of features, which would otherwise only change slowly over time.
- Hedgerows, which form the boundaries and shelter belts of many orchard blocks, grow throughout the year and so reduce the space near the boundaries of each block. When the growth becomes excessive, the hedgerows are trimmed, causing a more immediate change and increase in the space in an orchard block.

These changes due to growth and maintenance suggest that if maps are going to be created by SLAM and these maps are later used to perform localisation, then the maps should use features that are more permanent. Features initially considered for use included the posts and kiwifruit trunks.

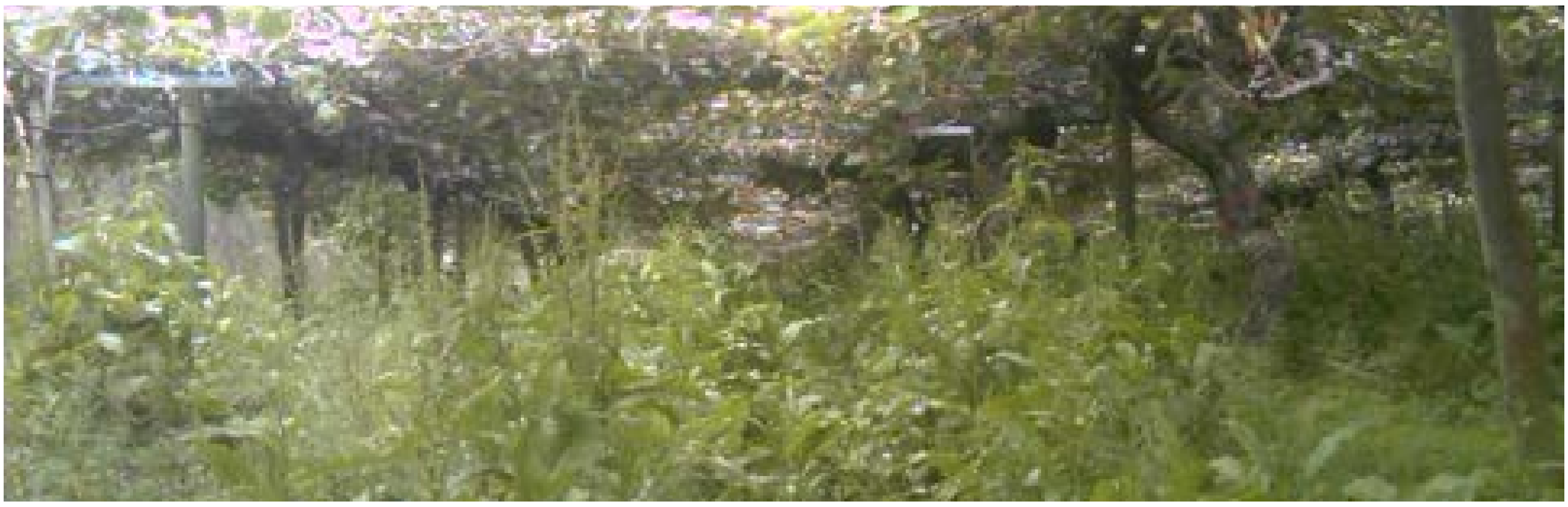

*Figure 96: Tall weeds in a kiwifruit orchard.*

#### 4.1.9 Driverless Vehicle Research

Driverless road vehicles are an interesting analogous application for autonomous orchard robots because:

- Many tasks that driverless cars perform must also be performed by autonomous orchard robots, such as mapping, localisation, lane following, object detection, object recognition, path planning, path following and obstacle avoidance.
- Many of the hazards are similar on the road and in an orchard, essentially amounting to avoiding collisions with stationary and moving objects.
- Like orchard robots, driverless cars must be designed for different weather conditions and highly variable environments.
- Various groups are working to produce driverless vehicles and many of these groups are testing in the real world and so have a practical focus, which is also important for orchard robot development.
- As driverless car technology matures, the hardware is becoming cheaper and could be leveraged by the agricultural robotics industry. This trend is already being seen with 3D lidar technology decreasing in price and developments in high performance computing hardware.

Hence, some driverless vehicle research is reviewed here as an analogous application. ALVINN (Autonomous Land Vehicle In a Neural Network) was early work that showed how a neural network can be trained to drive a vehicle [167]. For inputs ALVINN used a camera with 30 pixels high by 32 pixels wide resolution and a laser range finder with 8 pixels high by 32 pixels wide resolution. The neural network was fully connected as opposed to convolutional, with just one hidden layer. The neural network was trained using simulated data, due to practical limitations of using real world data at the time. However, the system was tested on a test vehicle in the real world; although, the top speed used was initially just 0.5 $ms^{-1}$.

This work with ALVINN later achieved road following at speeds of up to 55 miles per hour [168]. Tackling the problem of traversing intersections was also attempted. The approach taken was to use the road detection neural network to detect roads from virtual camera views ahead of the vehicle. Then traversal of the intersecting road was controlled by pure pursuit of a projected point on the intersecting road. Using virtual camera views would not always be possible where the views are occluded. In addition, Jochem et al. [168] pointed out that using pure pursuit and projecting a point down a road of unknown geometry was an unsatisfactory solution.

LeCun et al. [169] used a similar approach to ALVINN, in order to train a small robot to avoid obstacles in off-road environments, using images from two forward facing cameras as input into a 6 layer CNN, which produced steering angles as the output. Training data was collected by recording images, steering commands and throttle commands, while a person was remotely controlling the robot to drive straight, when the path was clear, and otherwise drive around obstacles. The CNN was trained to predict the steering commands; the actual outputs of the network were two values, corresponding to left and right turns. The resulting system was able to avoid obstacles under different conditions, while driving at 2 $ms^{-1}$.

CNNs have also been used to recognise traversable terrain from camera images [170]. This method used short range stereovision data to label terrain as traversable in a short range and this labelled data was used online to learn traversable terrain. This approach may be useful for orchard driving; although the online aspect may be unnecessary if a sufficient training data set was used for training beforehand.

A team from Nvidia Corporation trained a CNN to output the steering commands for a car, using input from camera images of the environment ahead [171]. During training data collection, the car steering angle, controlled by a person, and images from three forward facing cameras were recorded. During training, the error between the car steering angle from the person and the steering angle computed by the CNN was used for training the CNN. The training data was augmented by using the multiple camera images at each time step to apply random shifts and rotations to the input of the CNN and calculating corresponding steering angles for the error calculation. During runtime, input from a single forward facing camera was used to calculate the steering angle for driving a test car in a diverse range of conditions.

Describing this work from Nvidia, Muller [172] said that it would be possible to get data of the vehicle’s trajectory being corrected, by having a person swerving during data collection; however, he pointed out that this would require manual curating of the data to cut out the parts of the driving where the person was deliberately driving off path. Instead the multiple viewpoints from different cameras allowed the transformation of the viewpoint; this method allowed much more data than was collected to be created. Muller also talked about the benefit of end-to-end learning being that a new scenario does not require more hand coding but just more data collection and training; this may be beneficial in orchards because there are differences between orchards. Muller concluded by saying that he does not think that end-to-end learning is a complete solution and that there should be redundancy in algorithms and hardware.

Analysing the method from Nvidia, the data augmentation approach may be both a positive and negative aspect. It is positive because it creates a lot of data of path correction that would otherwise be difficult to create [171], [172]. However, in a controlled environment like an orchard, this data may be possible to collect by remote control swerving, in the way that Muller described [172]. The negative aspect of the data augmentation is that an algorithm is used to create artificial data of a path correction manoeuvre and therefore the learnt behaviour is redundant because the data augmentation already contains an algorithm that is capable of path correction.

From the existing literature it was unclear if it was feasible in a kiwifruit orchard to use the approach of having camera images as input and steering commands as output for a neural network. The kiwifruit orchard has a sparse array of posts and trunks that define its structure and it was not clear if this lack of defining features would prevent a neural network from detecting the structure well enough to robustly produce a useful steering command; especially since transverse directions appear similar to true rows in the pergola structure.

Banino et al. [173] performed navigation in simulation using agents trained by deep reinforcement learning. The first stage in the development of the agents was supervised training of a recurrent neural network with LSTM modules to perform odometry. During this training the agents learnt representation resembling cells from brains, including grid cells, so long as regularisation, such as dropout, was applied. In order to perform navigation in maze like environments, the recurrent neural network was incorporated into a larger architecture, including a CNN to process the input data stream and a policy recurrent neural network with LSTM modules to determine the actions. This work adds the ability to navigate in more complex environments. However, a path in an orchard may be relatively simple since the area can be covered by driving down one row and turning into the next row. Hence, the level of complexity employed by Banino et al. [173] may not be required for orchard navigation.

Levinson [174] extracted geometric curb features and reflectivity lane marking features from moving vehicles on roads using 3D lidar. The lane markings on roads were found to have relatively high reflectivity, which allowed the lane markings to be detected in the intensity data from lidar sensors mounted with their scan planes directed towards the road. The method for extracting curb features consisted of:

1. Maintaining a local 2D grid of accumulated lidar scan data centred around the lidar. For each new lidar point, the corresponding cell in the 2D grid was updated with the addition of a calculated score. This score was the dot product of the angle to the current lidar point from the lidar centre and the angle from the current lidar point to the next point in the same plane. This dot product would tend to give low values for the concentric circles that are commonly observed in lidar data from a flat ground plane.
2. The yaw of each lidar point to the nearest lane markers in a prior map was calculated.
3. A point spread function was applied to the lidar points in the direction of the calculated yaw values from step 2.
4. An edge filter was applied to the resulting data along the lateral direction and the high response values on both sides of the vehicle were found.

The methods of Levinson are not all directly applicable to navigation in orchards. There are no curbs in orchard rows; however, it seemed that the ground plane removal method of Levinson could be useful for the removal of the ground plane and the canopy in 3D lidar data in a kiwifruit orchard. In addition, there are no highly reflective painted lane markings in orchards; however, it seemed possible that some objects in orchards may produce higher intensity reflections. In particular, it seemed that the wires used to train the canopies could have higher reflectivity than surrounding objects; however, it also seemed possible that these wires would not produce higher intensity values since these wires only have a small cross sectional area and are commonly galvanised, which makes them appear to be relatively dull. Nevertheless, it was decided to survey lidar data in order to identify objects in orchards with distinctively higher reflectivity.

Cheng et al. [175] performed road centreline extraction using a bespoke Convolutional Neural Network, which they termed CasNet. Their approach used two CNNs, the first of which was a semantic segmentation CNN to segment road pixels. This first CNN was connected to a second centreline extraction CNN, whose function was to reduce the extracted road pixels to smooth single-pixel width lines, corresponding to the road pixels. Although a similar approach to that used by Cheng et al. [175] might be used for a problem such as kiwifruit row centreline extraction, there could be some issues with directly reapplying their method, including:

- Their proposed network is relatively deep with 34 layers, consisting of convolutional and deconvolutional layers. Their network was not designed to be run in real time on robots. The goal with kiwifruit orchard row centreline extraction would be to use a smaller network that can run at at least 5 Hz, if not faster, on consumer level GPU hardware.
- Their problem is quite different because they are detecting roads from aerial images, where roads appear as thin lines already; whereas, for kiwifruit row centreline detection, the features from sensors may not directly form lines. As a result of this difference in the problems being solved, CasNet may not work for kiwifruit row centreline detection.
- The output of the CasNet is an image; whereas, it would be preferable for kiwifruit row centreline detection to output the parameters of a line in order to avoid further processing.

Eustice [176] discussed research carried out at the University of Michigan for the development of autonomous cars. The driverless cars had GPS for coarse pose estimates and used four Velodyne 3D lidar sensors on the roof, for object detection and localisation. Multiple sensors were used for redundancy and to cover blind-spots and reduce shadowing. Cameras were used for traffic lights and object detection. Delphi radars provided speed and range measurements of objects and other cars. Odometry and Applanix INS sensors were also used. The range of sensor information was effectively extended by relaying information between cars. These sensors used for an autonomous car are also potentially interesting for an autonomous orchard vehicle. In addition, the idea of relaying information may be possible in an orchard, if there are multiple robots working together.

The Daimler AG Research and Development team used complementary radar and computer vision systems for autonomous navigation of a vehicle through city streets and highways [177], [178]. The radar system was used for detection of objects in a short range around the perimeter of the car and over a longer range in the forwards direction. The radar system was used at roundabouts, intersections and during lane changing manoeuvres. The stereovision system employed semi-global matching, implemented on FPGAs [179], [180]. The stereovision system was used for lane recognition, free-space detection and obstacle classification. The stereovision system had a relatively narrow field of view, so wide angle cameras were used for detection of features that required greater peripheral vision. The applications of the wide angle cameras included traffic light monitoring, localisation and pedestrian detection while the vehicle was turning. Similar to the rest of the navigation system, the localisation system also used complementary sensors, taking data from GPS, inertial sensors and cameras [179]. The Daimler team highlighted the importance of accurate maps for localisation as well as redundancy and robustness of the sensing systems. The navigation system on the autonomous vehicle ran in real time, with all modules operating at 25 Hz. The autonomous vehicle logged 6700 km in fully autonomous mode, during testing in a variety of lighting and weather conditions. The final test was over 100 km in busy traffic with no human intervention. The variety of sensors used and what they were used for in this project is potentially applicable to sensor selection for autonomous driving in orchards.

### 4.1.10 Autonomous Driving Literature Review Conclusions

Only one previous body of work has directly addressed navigation in pergola structured kiwifruit orchards [21]. However, this method was not well tested in the real world and used a 2D lidar, which can intercept the ground, canopy and weeds in a pergola structured orchard to give unstructured data, affecting the accuracy of the row measurements. Multiple navigation methods have been developed for tree-wall orchards and autonomous vehicles. However, none of these methods has addressed the particular challenges of pergola structured orchards with unstructured data coming from the ground, weeds and an overhead canopy. Hence, extracting useful data from the navigation sensors in a kiwifruit orchard is a key part of the work presented in this section.

## 4.2 Autonomous Driving Sensor Selection

Sensor selection was a critical early step in the development of the autonomous driving system for the AMMP because it dictated the direction of subsequent research. Three steps were used in the process of sensor selection:

1. A literature survey was conducted of sensors previously used in orchard navigation and similar applications.
2. The features, specifications and costs of sensors available to purchase were collated in order to perform a cost-benefit analysis.
3. Based on the cost-benefit analysis, some sensors were acquired and data was collected from these sensors in kiwifruit orchards. Then the data was analysed to ensure that the environment of the kiwifruit orchards did not noticeably inhibit the ability of the sensors to take measurements of the environment.

### 4.2.1 Sensor Literature Survey

From the existing literature, it seemed that the sensors that were commonly used for outdoor autonomous navigation outside of orchards included monocular cameras, stereo cameras, 2D lidar, 3D lidar, radar, GPS, IMU systems and odometry [174], [176]–[181]. These choices were consistent with the existing sensors that were commonly used for orchard navigation systems, which included cameras, 2D lidar, GPS, inertial sensors, wheel encoders, gyroscopes and magnetometers [182]–[184]; some examples are summarised in Table 22.

*Table 22: Examples of sensors used in previous orchard navigation research.*

| Reference | Application | Sensors |
|---|---|---|
| Subramanian et al., 2006 [140] | Citrus grove navigation | Separate angled down 2D lidar & camera (640x480); steering encoder |
| Barawid et al., 2007 [142] | Autonomous tractor | Horizontal 2D lidar |
| Hansen et al., 2011 [145] | Autonomous orchard vehicle | Horizontal 2D lidar, fibre optic gyroscope, odometry (encoders) |
| He et al., 2011 [148] | Orchard row detection | Camera (640x480) |
| Torres- Sospedra and Nebot, 2011 [150] | Orchard row detection | Camera (640x480) |
| Scarfe, 2012 [21] | Kiwifruit harvesting robot | Horizontal 2D lidar, fluxgate compass |
| Freitas et al., 2012 [186] | Orchard robot obstacle detection | Angled down 2D lidar, wheel & steering encoders, low cost IMU |
| Zhang et al., 2013 [147] | Orchard robot row following | Rotated 2D lidar; lidar rotation, wheel & steering encoders |
| Bergerman et al., 2015 [185] | Autonomous orchard vehicle | Horizontal 2D lidar, wheel & steering encoders |
| Bargoti et al., 2015 [158] | Apple orchard localisation | Vertical 2D lidar, Global Positioning Inertial Navigation System |
| Jagbrant et al., 2015 [157] | Almond orchard localisation | Vertical 2D lidar, Global Positioning Inertial Navigation System |
| Sharifi and Chen, 2015 [151] | Orchard row detection | Camera |

### 4.2.2 Cost Benefit Analysis of Navigation Sensors

The costs of different sensors for a single unit were collated by contacting various manufacturers and suppliers; these results were used to estimate cost ranges for each sensor type. The applications and issues for each sensor type were noted, based on the existing literature. This information is summarised in Table 23. This data seemed to indicate that the sensors that deliver the most functionality are lidar, cameras and Time of Flight sensors. In addition, these three sensor types are a reasonable price at the lower end of their cost range; especially for Time of Flight cameras, although it was assumed that these sensors might have a serious issue with degradation in performance in sunlight. Because localisation is such a key functionality, GPS also seemed like an important option to test and its cost can also be comparable or cheaper than other sensor modalities. From this data and since the AMMP has built-in encoders, it seemed to make sense to use encoders to assist with mapping, localisation and velocity feedback.

*Table 23: A sample of the sensors considered and parameters used for sensor selection for the AMMP navigation system.*

| Sensor Type | Manufacturers Considered | Cost Range (USD) | Navigation Applications | Issues |
|---|---|---|---|---|
| GPS | Ublox, Garmin, Omnistar | 50- 4000 per annum | Localisation, pose and velocity measurement | Signal loss under the kiwifruit canopy |
| IMU | InvenSense, Analog Devices | 50- 3000 | Acceleration and angular velocity measurement | Thermal drift, accumulated errors |
| Digital Compass | Honeywell, KVH | 50-500 | Heading measurement | |
| Encoders | CUI | 30-50 | Dead reckoning, velocity measurement | Accumulated errors |
| 2D Lidar | Hokuyo, SICK | 2000-10000 | Mapping, localisation, pose and velocity measurement, obstacle detection | Tend to not work well in fog and heavier rain |
| 3D Lidar | Velodyne, Quanergy, Neptec | 4000-90000 | Mapping, localisation, pose and velocity measurement, obstacle detection | Tend to not work well in fog and heavier rain |
| Time of Flight Cameras | Intel, Basler, Fotonic, Odos Imaging | 100-9000 | Mapping, localisation, pose and velocity measurement, obstacle detection | Tend to not work well in bright sunlight conditions, fog and heavier rain |
| Cameras | FLIR, Basler | 500-1500 | Mapping, localisation, pose and velocity measurement, obstacle detection | Tend to not work well when visibility is low |
| Thermal Cameras | FLIR, Optris | 3000-14000 | Pedestrian detection | In hot conditions, this sensor becomes less useful for pedestrian detection |
| Radar | Delphi | 3000-5000 | Object detection, possibly through foliage | Lower resolution data than some other sensors |

### 4.2.3 GPS Data Collection

Two GPS modules were tested. Both were connected via serial to a Beaglebone Black single board computer. The GPS set-ups tested were:

- Ublox Neo-M8N [187] module, selected for its superior navigation sensitivity of -167 dBm and internal Low Noise Amplifier (LNA), additional 20 dB gain/ 0.8 dB noise LNA and active circuitry for the 25mm square ceramic patch antenna.
- OmniSTAR 5120VBS [188] module, with AX0 series antenna, with 34 dB gain/ 1.4 dB noise LNA. High multipath rejection is claimed for the AX0 series antenna.

The procedure used for data collection was:

1. A route through a kiwifruit orchard was planned and plotted on a satellite image. The route was designed to include long stretches underneath the canopy and some areas with a good view of the sky.
2. The distance between post centres along the planned route was measured using a tape measure.
3. The GPS data collection setups were turned on at the starting point of the planned route and were left to initialise for 30 minutes.
4. The data collection unit was carried along the planned route, stopping at each post. At each post, the data collection unit was placed near to the post in an orientation that was repeated for every post along the route.
5. Step 4 was repeated until the entire route had been traversed and every post along the way had been (approximately) mapped by the data collection unit.

It was noticed during testing that the signal quality lights on both GPS setups were regularly indicating loss of GPS fix. Although the entire data collection procedure was performed, many of the results, such as the measurement of spacing between posts, are far less significant than the regular loss of the GPS fix.

The orchard used for the data collection was Bateman's at 48 Newnham Road, Whakamarama, New Zealand. The path followed and the corresponding GPS data collected from the different setups is shown in Figure 97.  Note that the path followed was traversed twice- once going out and once returning. The data was collected at a slow walking speed- the entire path of approximately 400 metres took in the order of 15 minutes to complete, including stops about every 5.5 metres within the orchard. Multiple sets of data were collected but Figure 97 is indicative of the results.

The Omnistar GPS setup appears to track the approximate path better but the data is sparse and there appears to be regular loss of signal in the orchard. The Ublox GPS setup collected more data than the Omnistar setup but was much less accurate. There were no quantitative results analysed for this work but, based on the spareness of the data, it was concluded that high quality GPS equipment may be useful to provide sanity checks of the approximate location in the orchard. However, it was decided that GPS could not be relied on to provide localisation and other feedback, under the kiwifruit canopy.

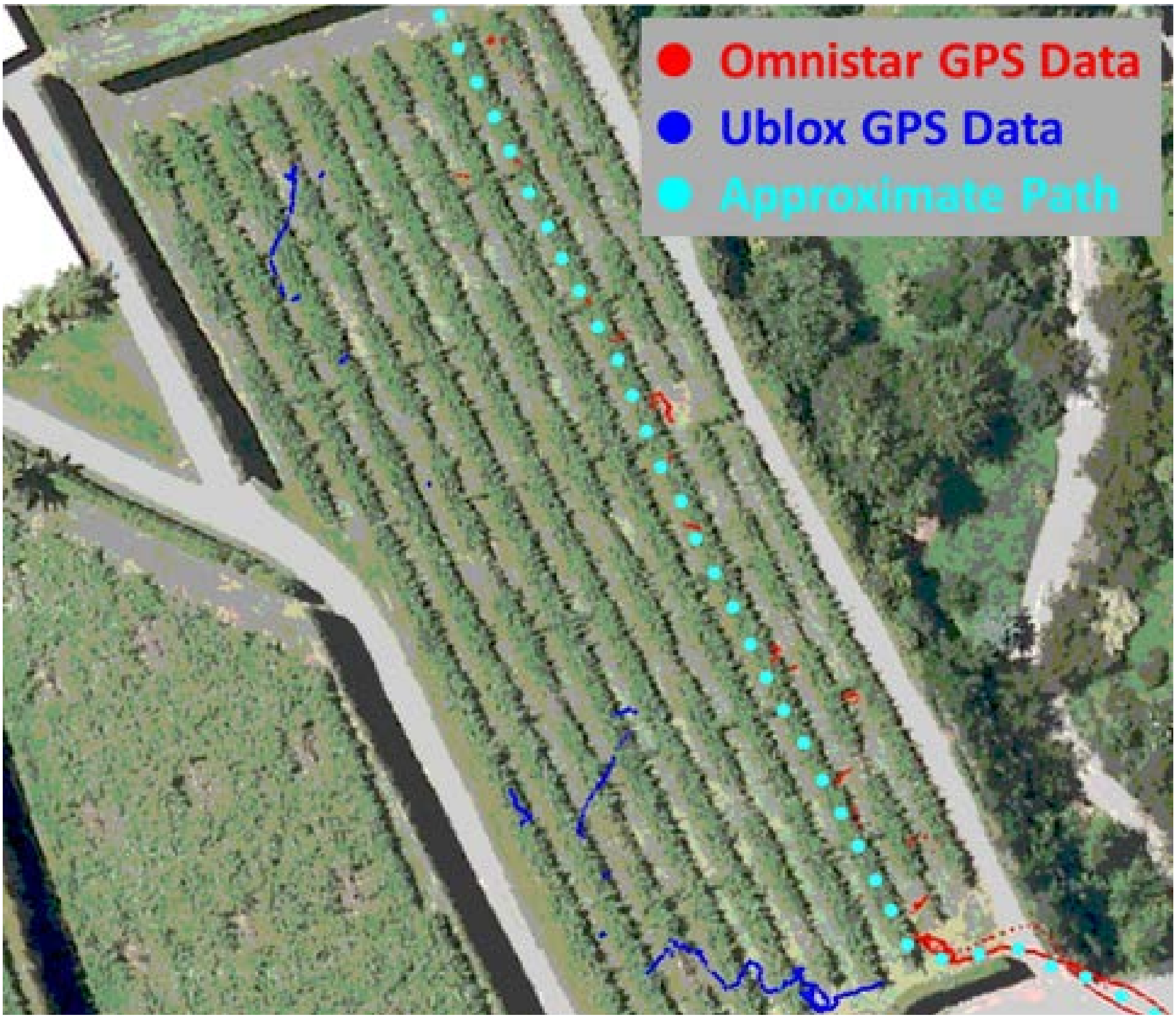

*Figure 97: Aerial view of a path through a kiwifruit orchard and the associated GPS data.*

### 4.2.4 Time of Flight Camera Data Collection and Inspection for Navigation

The Time of Flight camera tested was a Basler tof640-20gm-850nm [77]. This sensor provides depth and intensity data at a resolution of 640 pixels wide by 480 pixels high. It was assumed that this sensor might work well in sunlight because it had previously been successfully used to collect data from the underside of the kiwifruit canopy in different lighting conditions, with minimal occurrences of the data being washed out. However, it was found that when the sensor was viewing the post and trunks of the pergola structure, there was large amounts of data loss in the depth data, which was deemed unacceptable for important navigation functions such as object and obstacle detection. For example, in Figure 98, there are posts and trunks that are within the range of the sensor; however, there is no depth data for these objects.

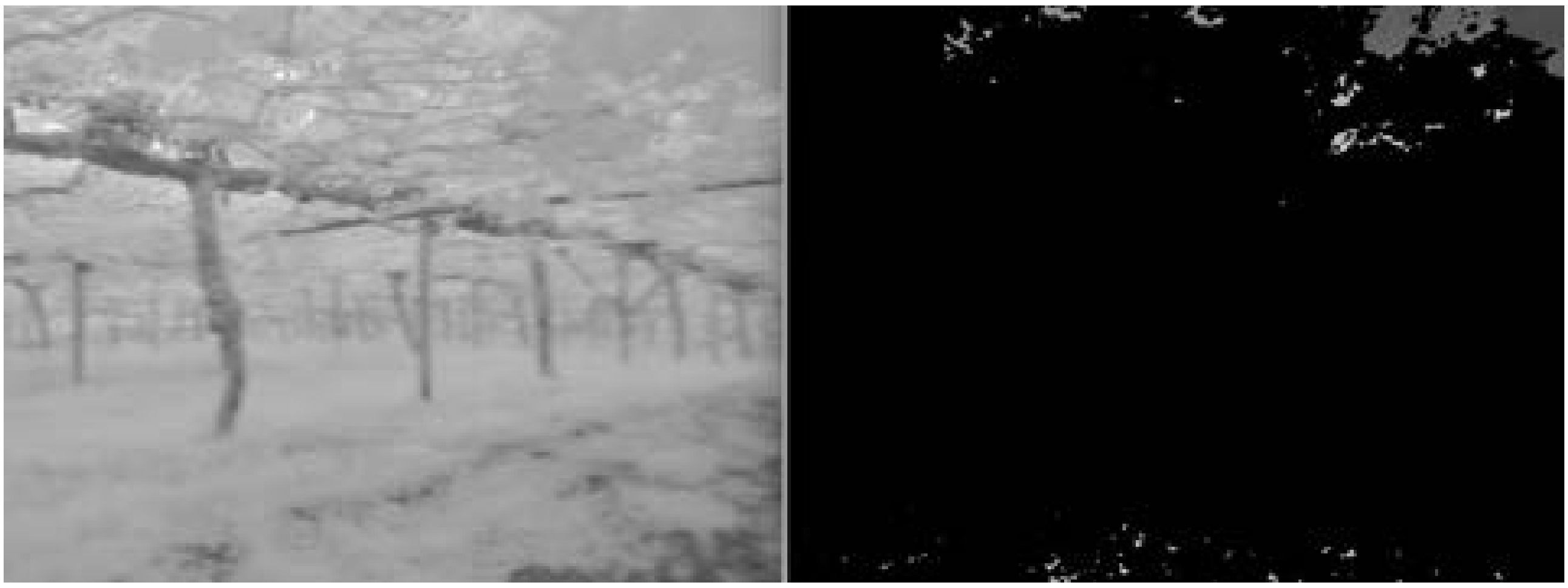

*Figure 98: Time of Flight sensor data, viewing the structure of a kiwifruit pergola, with intensity data (left) and the corresponding depth data (right).*

### 4.2.5 Lidar Data Collection and Inspection for Navigation

Data was collected from two 2D lidar sensors; these were a Hokuyo UTM-30LX [128] and a SICK LMS111 [189]. Data was also collected from a 3D lidar sensor, which was a Velodyne VLP-16 [130]. The data was collected by driving along rows in kiwifruit orchards with the sensors horizontal and at a height of 0.8 m, which was approximately midway between the ground and the canopy.

It was thought that the lidar sensors would measure the position of structure defining features in the orchard, such as posts, trunks and hedges. Detecting such features would then allow the boundaries of the row to be found for row following or these features could be mapped and used for localisation. However, both 2D lidar sensors produced clouds of unstructured data amongst the structured features, as shown in Figure 92. This was caused by the lidar plane intercepting the canopy on convex slopes (Figure 90) and intercepting the ground on concave slopes (Figure 91) at short ranges. In addition, the Hokuyo UTM-30LX exhibited some spurious unexplained measurements. The cause of these erroneous measurements was not determined.

Data was also collected from the 3D lidar sensor. The 3D lidar sensor had 16 layers of data vertically. As a result on concave slopes, some of the angled up planes had a longer viewing range facing forwards and on convex slopes some of the angled down planes had a longer viewing range facing forwards.

From these results it seemed that the 2D lidars could be a useful sensor at a relatively short range to use for an independent channel of processing for redundancy and increased reliability. However, it was decided to use 3D lidar as a primary navigation sensor because of its ability to be used at longer ranges on undulating ground in kiwifruit orchards.

### 4.2.6 Camera Data Collection and Inspection for Navigation

Data was collected from Logitech C920 [74], Basler Dart daA1600-60uc [92] and Flir CM3-U3-13S2C-CS [190] cameras. Data collection was performed at various times of the day and night, in different weather conditions and in different orchards. The data was collected on autonomous and manually driven platforms at a height of approximately 0.8 m from the ground and at speeds of up to 3 $ms^{-1}$.

It was noticed that the Logitech C920 cameras produced significant motion blur. In addition, these cameras did not provide a hardware trigger interface, which would be important if the cameras were used for stereovision. No image quality issues were noticed with the Basler and Flir sensors; however, the Basler camera was favoured for its later model image sensor.

### 4.2.7 Sensor Selection Conclusions

It was decided to proceed with using cameras, 3D lidar and encoders as the primary sensors for the navigation system. Algorithms developed and tested using these sensors are described in the rest of this section.

## 4.3 2D Lidar Row Following by Free Space Angle Search

Since the rows form such a high proportion of the space in orchards, it was thought that autonomously driving down rows was a key capability for navigating orchards. This section describes the method and testing of an algorithm to perform such autonomous driving down rows.

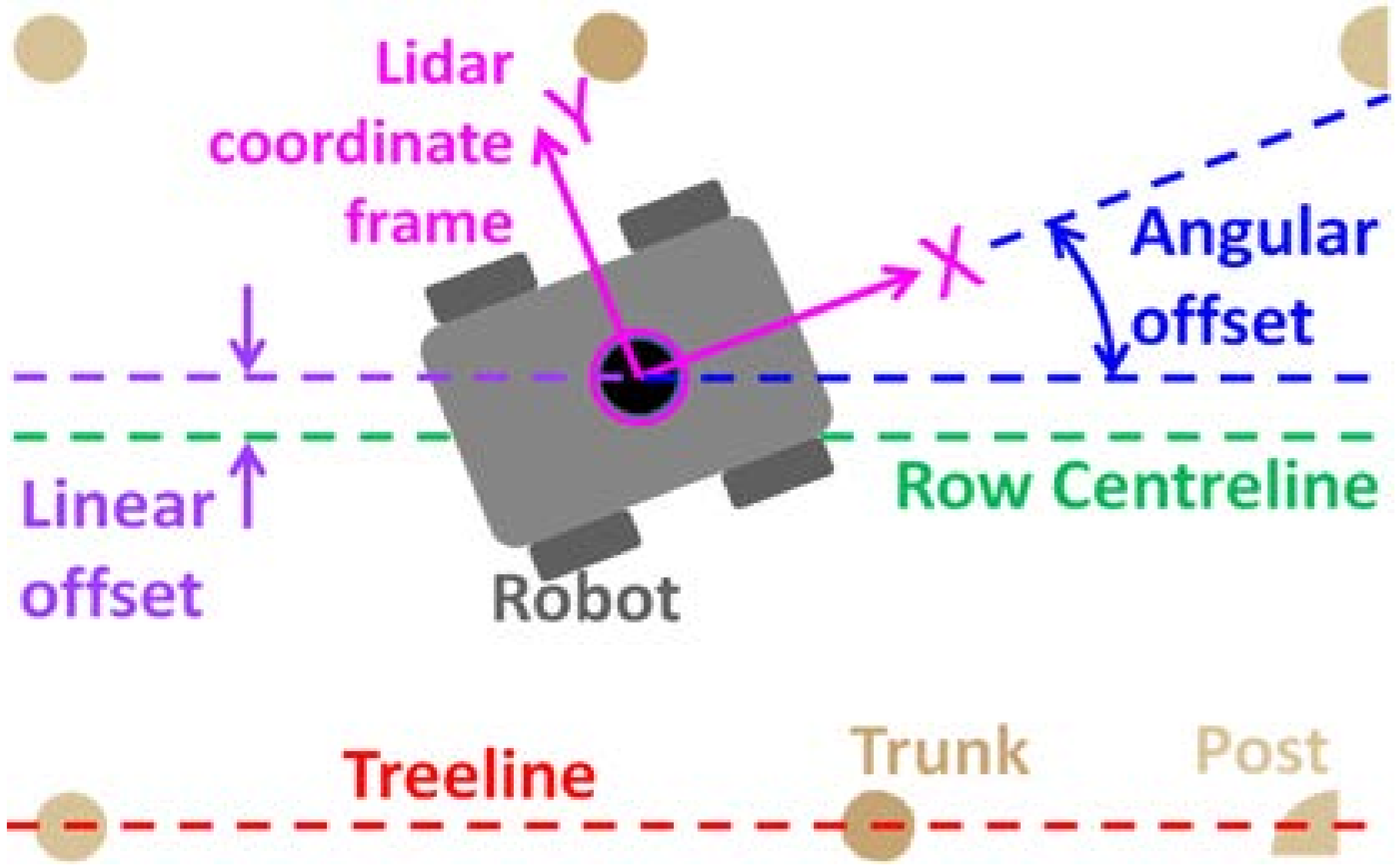

*Figure 99: A robot viewed from above, travelling left to right down a row- showing the definition of the linear and angular offset of the lidar coordinate frame with respect to the row centreline.*

Figure 99 shows a scene from above with a robot driving down a row and this figure defines the lidar coordinate frame and variable offsets of this coordinate frame from the row centreline. The problem of *row following* was formulated as travelling at a target speed down a row without collision, while minimising the linear offset, $o_l$, and angular offset, $o_a$, of the lidar coordinate frame with respect to the row centreline. Because these offsets must be minimised throughout an orchard, over multiple, $n$, time-steps and there are multiple objectives that require some weighting or scaling, $w_a$, the objective function to minimise for row following can be represented by:

$$\sum_{i=1}^{n} (o_l + w_a o_a) \tag{5}$$

Initially, a row following algorithm was developed for a mobile robot, using just a Hokuyo UTM-30LX 2D lidar [128] as the sensor. This lidar was mounted so that the scan plane was horizontal. The premise of the algorithm was to check different angles for the number of points in a set area in order to find the direction with the most free space. This algorithm is referred to in the rest of this section as the *free space angle search*. The free space angle search used the following steps:

1. Lidar points with a radius below an upper threshold were extracted for further processing.
2. The extracted lidar points were rotated about the lidar centre by a set angle.
3. The lidar y coordinates (Figure 99) of the rotated points were calculated.
4. The number of y coordinates within a set threshold of the lidar origin were counted and stored, along with the corresponding rotation angle.
5. Steps 2-4 were repeated for different angles.
6. If a single angle gave the lowest number of the y coordinates at step 4, this angle was taken as the target angle, $\gamma_i$.
7. If multiple angles gave the lowest number of the y coordinates at step 4, these angles were clustered into lists of consecutive angles and the median of the largest cluster was taken as the target angle, $\gamma_i$.
8. The new angular velocity command for the robot, $\omega_i$, was calculated using a gain, $k_a$, according to:

$$\omega_i = k_a \gamma_i \tag{6}$$

Row end turns were implemented by checking a set rectangular area of the lidar data for any points. When there were no points in the set area, this was taken to be a row end condition and a fixed radius turn was commanded at a set speed and for a set distance. After the turn was completed, row following was restarted.

### 4.3.1 Laboratory Testing

A test robot was created for the free space angle search algorithm using a Hokuyo UTM-30LX 2D lidar [128], mounted with a horizontal scan plane on a Pioneer P3-AT [191] mobile robot, with the processing performed on a laptop. Test rows were created using cardboard postal tubes to represent the posts and trunks of an orchard (Figure 100). These tubes were used to create 2 rows for the testing of row following. The algorithm was run continuously on this setup for 3 hours, while changing the width, length and curvature of rows. After 3 hours, a motor on the robot failed.

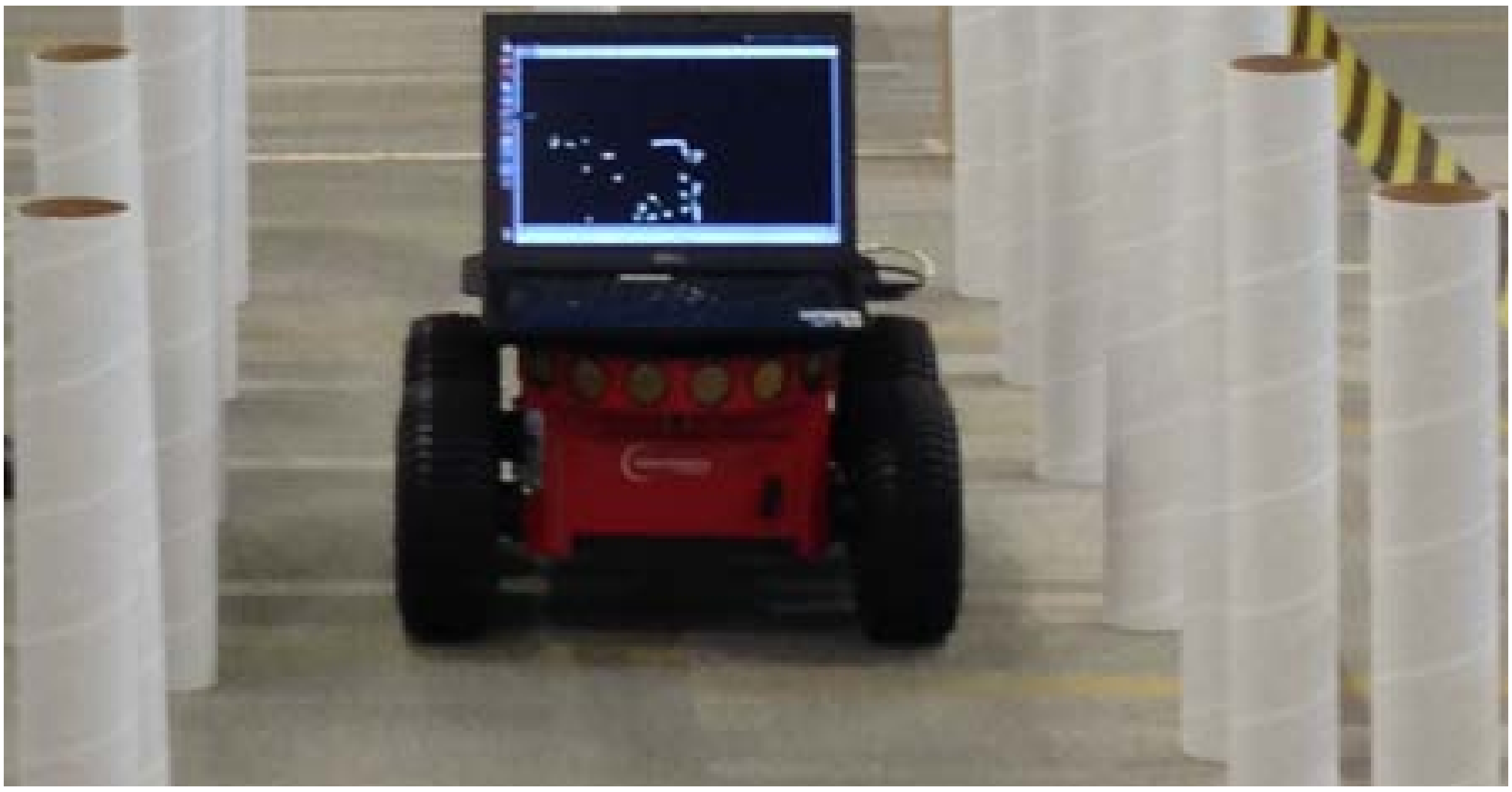

*Figure 100: Row created using postal tubes for testing row following in a laboratory.*

### 4.3.2 Testing in an Apple Orchard

The algorithm tested in the laboratory was further tested using a Clearpath Husky [96] robot platform and a Velodyne VLP-16 [130] lidar in an experimental apple orchard. To run the free space angle search algorithm with the Velodyne VLP-16 lidar, only a single plane of lidar data was used. An interesting feature of the orchard was it used multiple growing structures in the same row, including V-shaped and vertical trellises. The autonomous driving system was tested by successfully performing autonomous driving in 2 rows, of 1.5 m width and 105 m length each, for a total of 630 m of autonomous driving at 1 $ms^{-1}$. In addition, tests were performed to ensure that the algorithm could correct large linear and angular offsets. The steps in this test procedure were:

1. The robot was moved by hand into a pose with a significant linear offset and angular offset (Figure 101).
2. Row following was initiated and the robot was allowed to drive for 40 m, while observations were made.
3. Steps 1 and 2 were repeated 10 times.

For each of these tests, the robot consistently corrected the linear and angular offsets. After approximately 20 m, the driving had the appearance of the row following without the large initial linear and angular offsets.

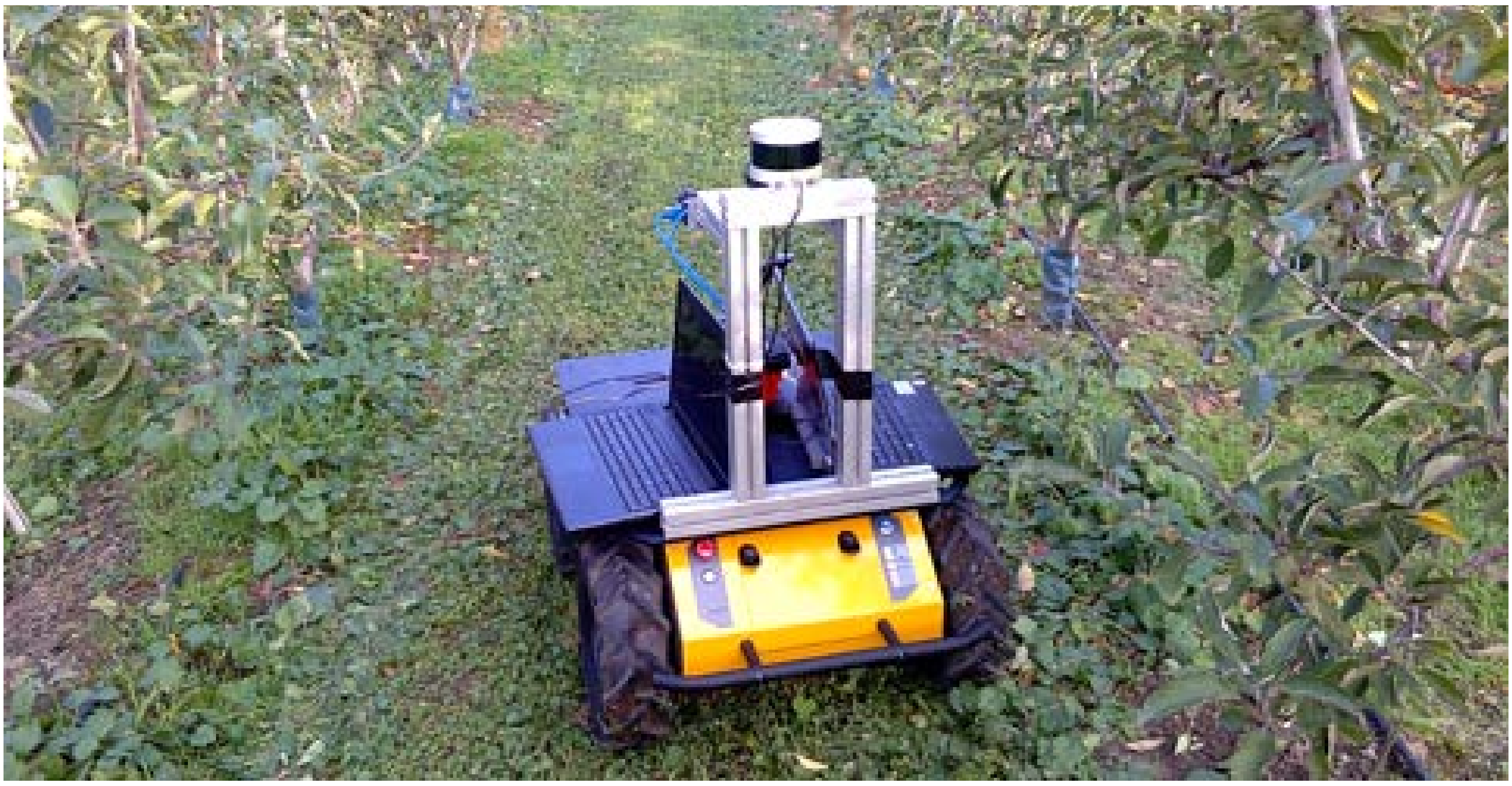

*Figure 101: Hardware used for testing row following in an apple orchard, with the robot shown in its initial pose.*

## 4.4 3D Lidar Kiwifruit Orchard Row Following and Navigation

The free space angle search method had known failure cases for some kiwifruit orchards because in these orchards there were some gaps in treelines that were greater than the row width. In these cases, free space angle search would be expected to allow the robot to drive out of the current row; however, this would not be acceptable because this could lead to damaged irrigation pipes or collisions with kiwifruit leader vines, which follow the direction of the treelines. Even when there are not unusually large gaps in treelines, directions orthogonal to the true row direction can appear as feasible paths for navigation in kiwifruit orchards (Figure 102).

Kiwifruit orchards present a number of challenges for a navigation system. The ground can be muddy, undulating and there may be hard fruit on the ground, which a robot may run over and slip on. Weeds and hanging branches form a cluttered environment at the level of the trunks and posts (Figure 103). A mix of shadow and sunlight under the kiwifruit canopy on a sunny day creates high contrast lighting conditions (Figure 104), which can be challenging for computer vision systems [21].

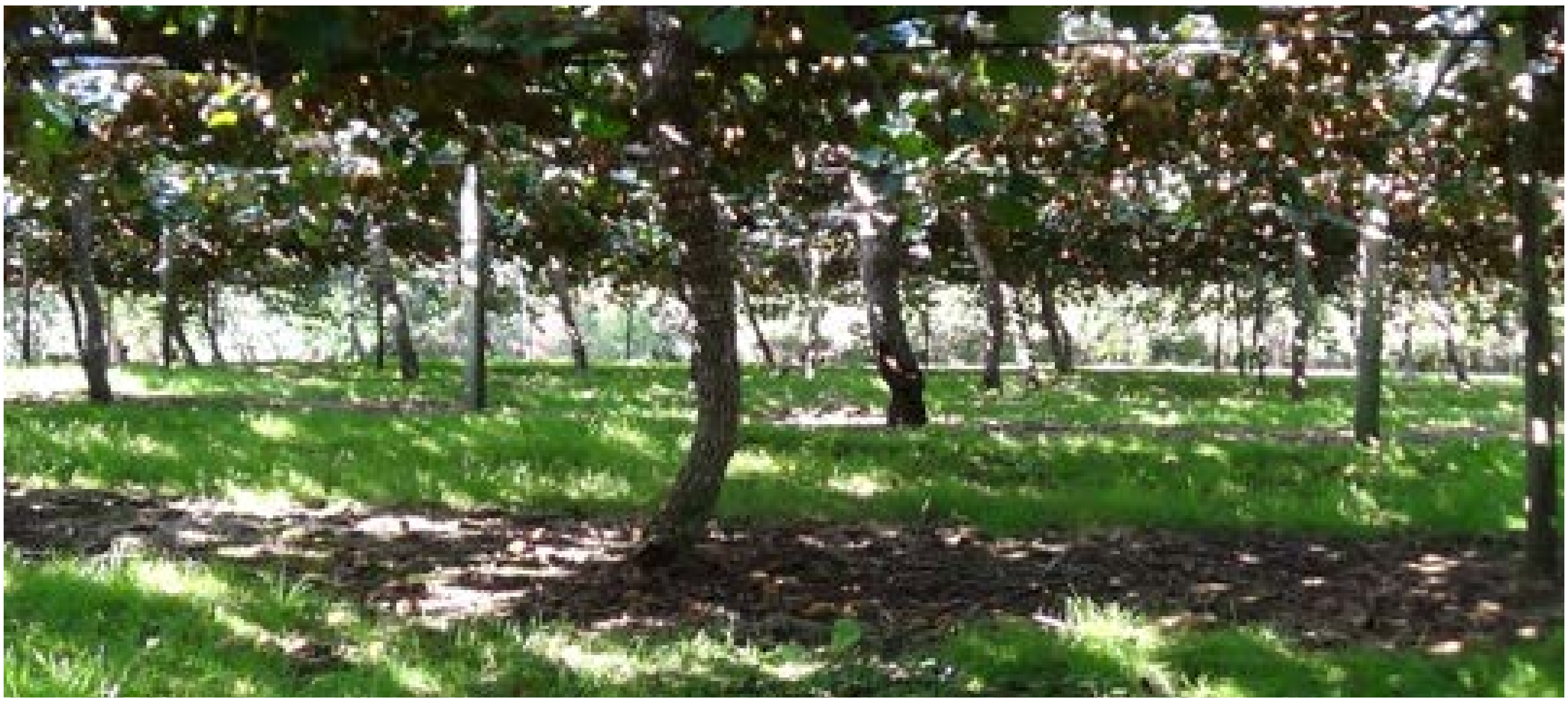

*Figure 102: Showing free paths in directions transverse to the true row direction.*

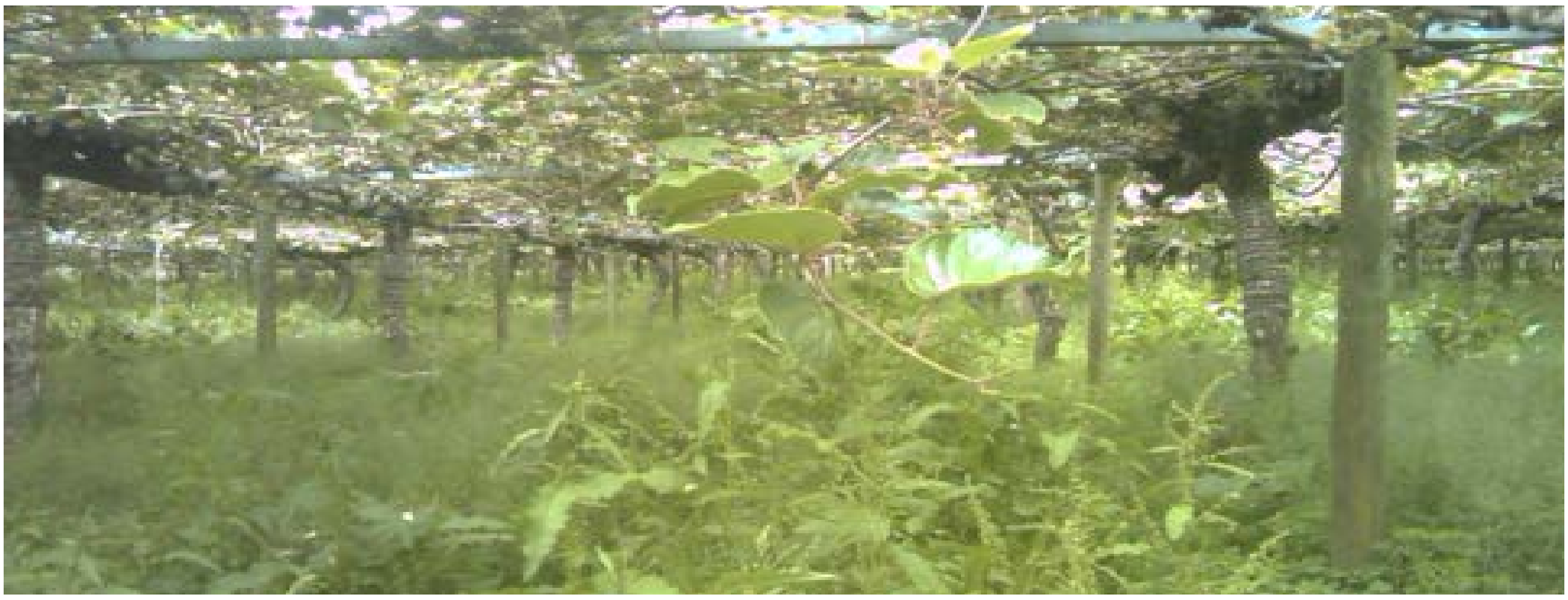

*Figure 103: Demonstrating occlusion from weeds and hanging branches in a kiwifruit orchard.*

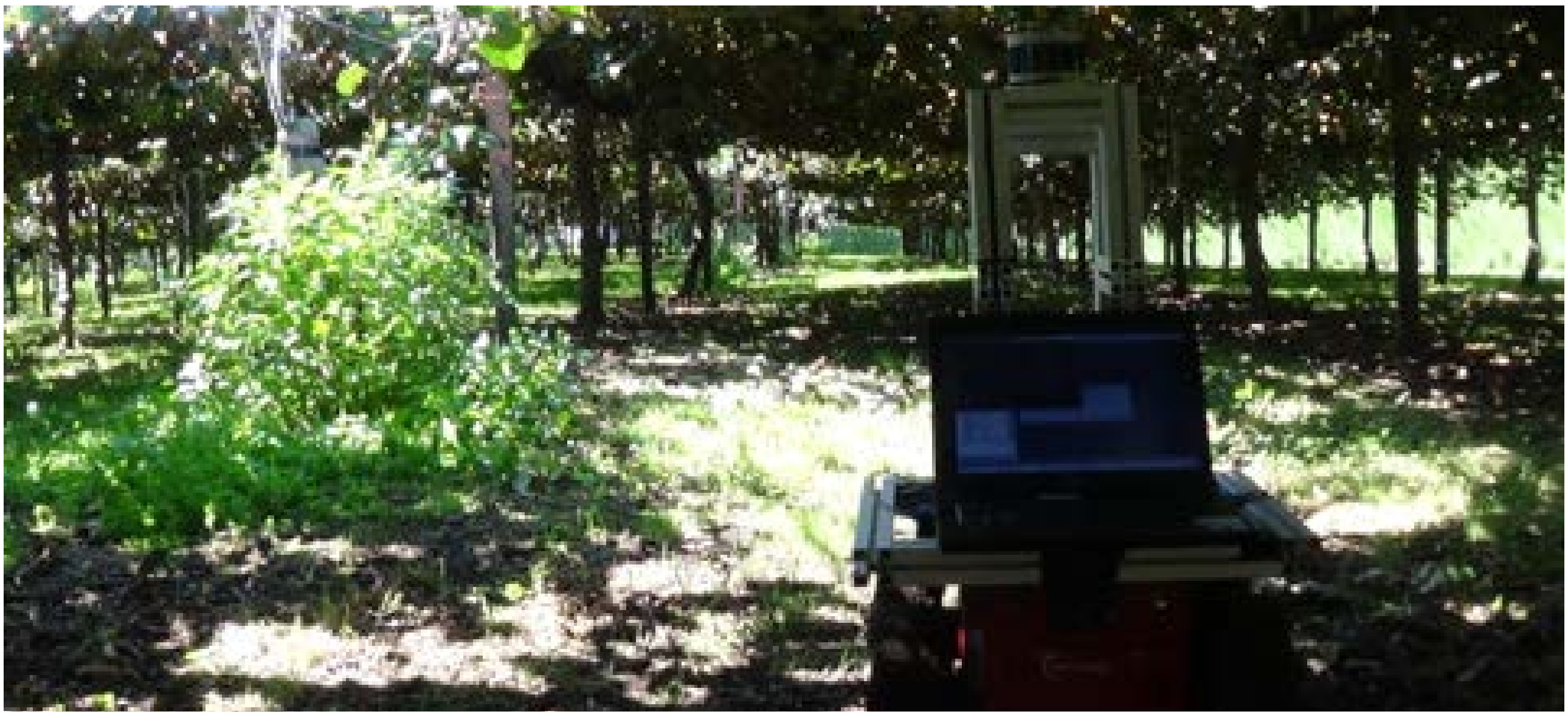

*Figure 104: High contrast lighting conditions under a kiwifruit pergola on a sunny day.*

As a part of this research, a 3D lidar navigation system for kiwifruit orchards was developed and tested [143]. This navigation system consisted of row detection, row following, row end detection, row end turn traversal and topological map execution components.

### 4.4.1 Algorithm for 3D Lidar Kiwifruit Orchard Row Detection and Following

The row detection algorithm estimated the linear and angular offsets with respect to the row centreline, as defined in Subsection 4.3. The row detection algorithm designed for kiwifruit orchards just used 3D lidar data as an input. The key part of this algorithm was the calculation of the angular offset of the lidar coordinate system from the row centreline. This algorithm attempted to extract the posts and trunks from the lidar data, while discarding points from other objects. Then the angles between pairs of posts and trunks in the same treeline were found. The key steps of the algorithm were:

1. For each plane of 3D lidar data, each consecutive pair of points was considered. If the distance between consecutive points was less than a threshold distance, the points were grouped into the same cluster; otherwise, if the distance between the consecutive points was more than the threshold, the points were grouped into different clusters. An example of using this clustering step is shown in Figure 105.

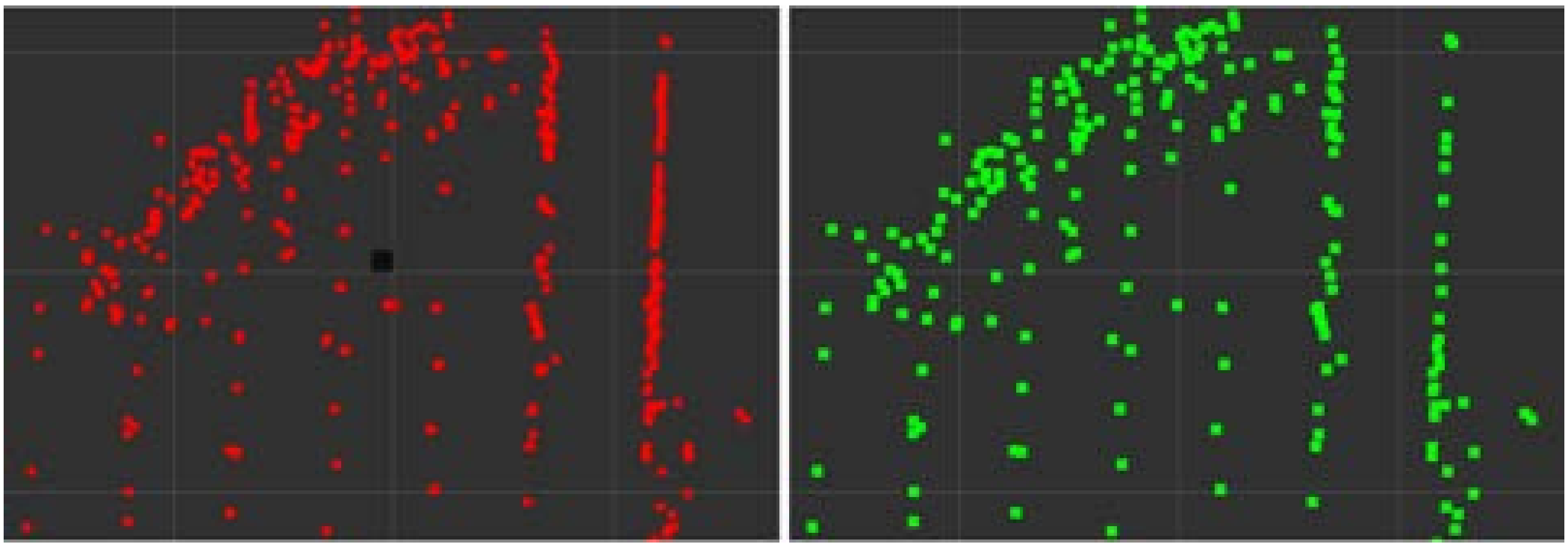

*Figure 105: Processing a single plane of 3D lidar data with raw data (left), clustered data with centroids shown (right) and a 10 m square grid overlaid.*

2. A threshold was set at approximately half the circumference of the thickest post and trunk. Clusters from step 1 with a contour length greater than this threshold were discarded since they were unlikely to be posts or trunks.
3. For each cluster remaining from step 2, the number of other clusters from the same lidar plane within a threshold distance was found, using the centroids of the clusters. If this number for a cluster was greater than a threshold, the cluster was discarded because such a cluster could be a part of a cloud of data, which can be returned from non-structure defining features such as a group of weeds, the kiwifruit canopy or the ground. An example result after this step for a single plane of the 3D lidar data is shown in Figure 106.

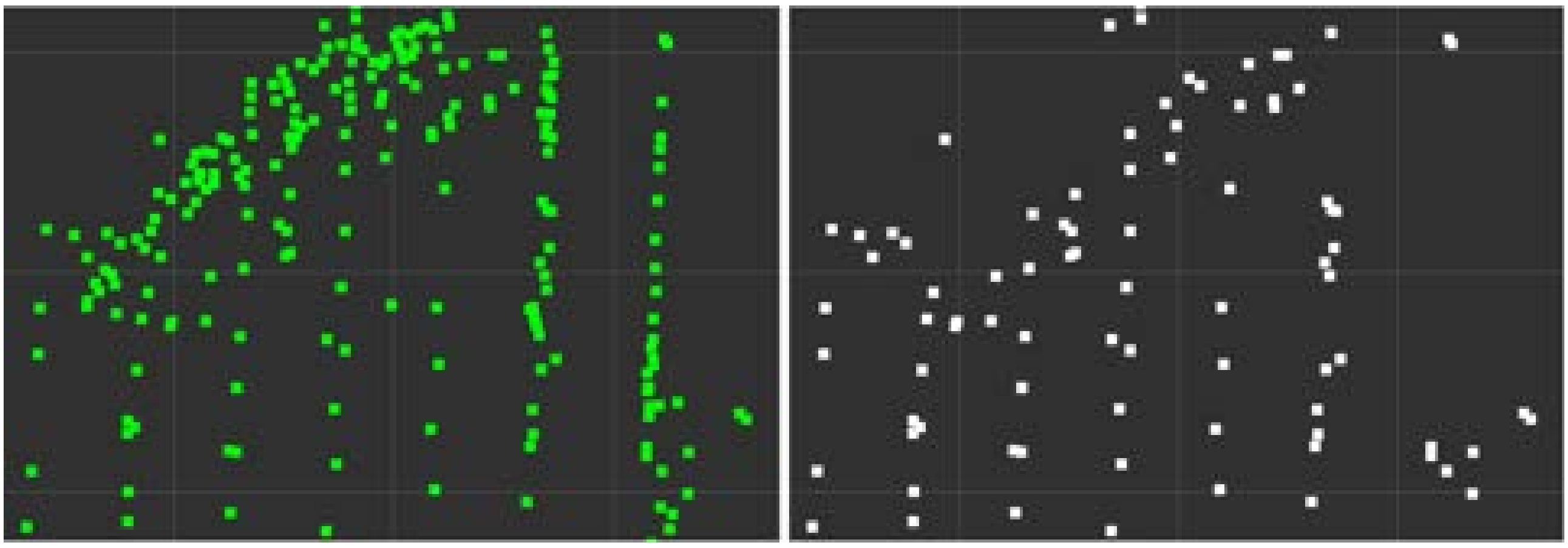

*Figure 106: Cluster centroids from Figure 105 (left) and the result of discarding oversized clusters and groups of clusters (right) from a single plane of 3D lidar data.*

4. For the first cluster remaining from step 3, a container of clusters was created and the cluster was added to the container. Then for each other cluster remaining from step 3, across all of the lidar planes, a new cluster was added to an existing container if that container had a cluster that was less than a threshold XY plane distance from the new cluster. If a new cluster could be added to more than one container, the containers were combined and then the new cluster was added to the combined container. This step combined the data from all of the lidar planes so that each container for a trunk or post should contain measurements at multiple heights from multiple planes on the same object.
5. The height of each container from step 4 was calculated using the difference between the maximum and minimum Z coordinates in each container. Containers which were below a threshold height or above a different threshold height were discarded (Figure 107).

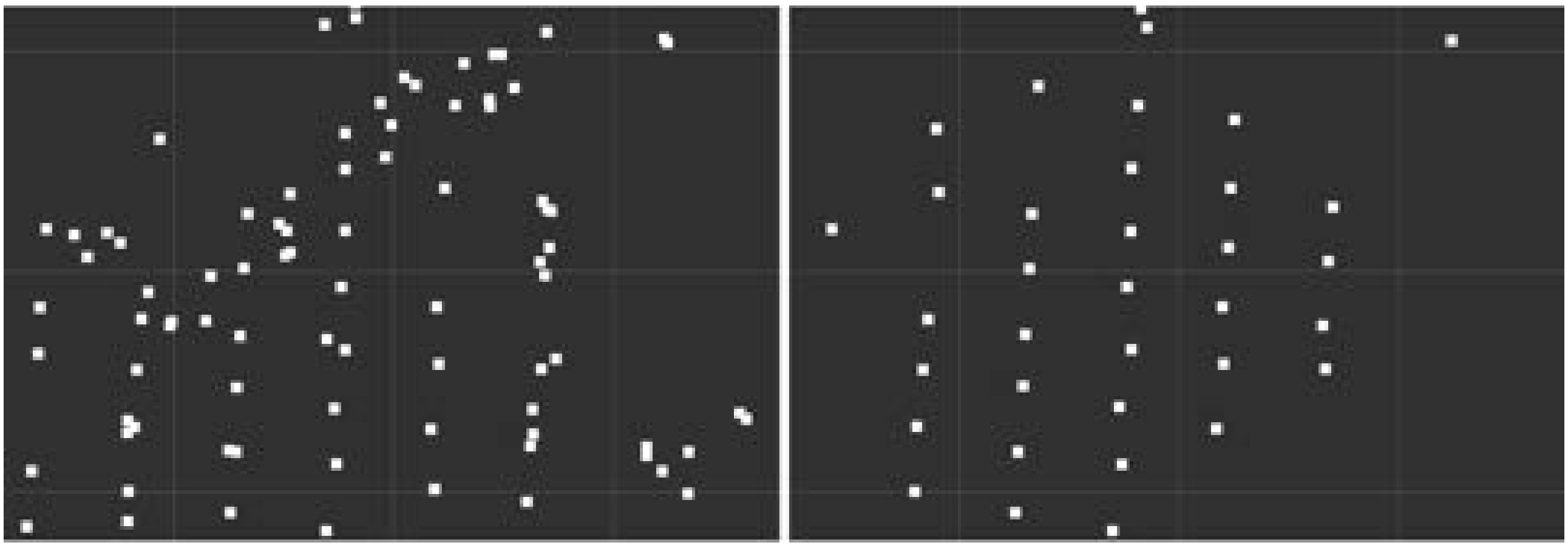

*Figure 107: Before (left) and after clustering data across multiple planes into containers and applying height thresholds to the data in the containers, with container centroids shown (right).*

6. Using the centroid of points in each container, each container's nearest neighbour was found. Nearest neighbour pairs were found because generally the nearest neighbour of a post or trunk is another post or trunk in the same treeline. When this is the case, the angle between nearest neighbours should also be the angle of the treeline that the nearest neighbours are in, which is also the angle of the rows of the orchard.
7. For each pair of nearest neighbours, the distance between the nearest neighbours was found. For a nearest neighbour pair, if this distance was greater than the row width, the nearest neighbour pair was discarded because the two neighbours in the pair might not have been in the same treeline. An example result after this step is shown in Figure 108.

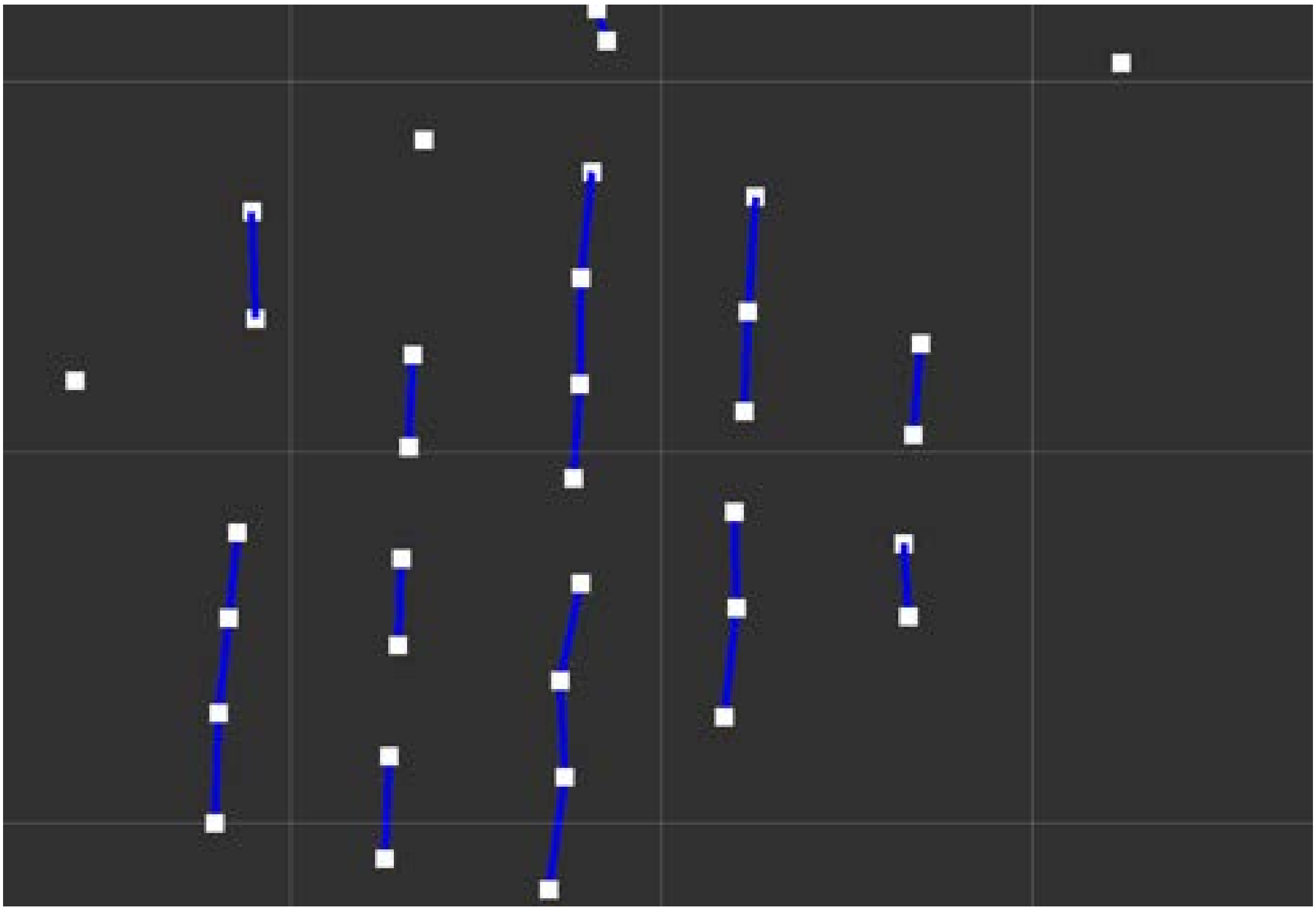

*Figure 108: Container centroids from Figure 107 (white) and nearest neighbour pairs with a separation distance less than a threshold (blue).*

8. For the nearest neighbour pairs remaining after step 7, the angle between the members of each pair was calculated.
9. The nearest neighbour angles from step 8 were discretized into angular ranges and the mode of the angular ranges was found. The mode was used in order to remove the effect of erroneous measurements. This mode was defined to be the current measurement of the angular offset of the lidar from the row centreline (Figure 109).

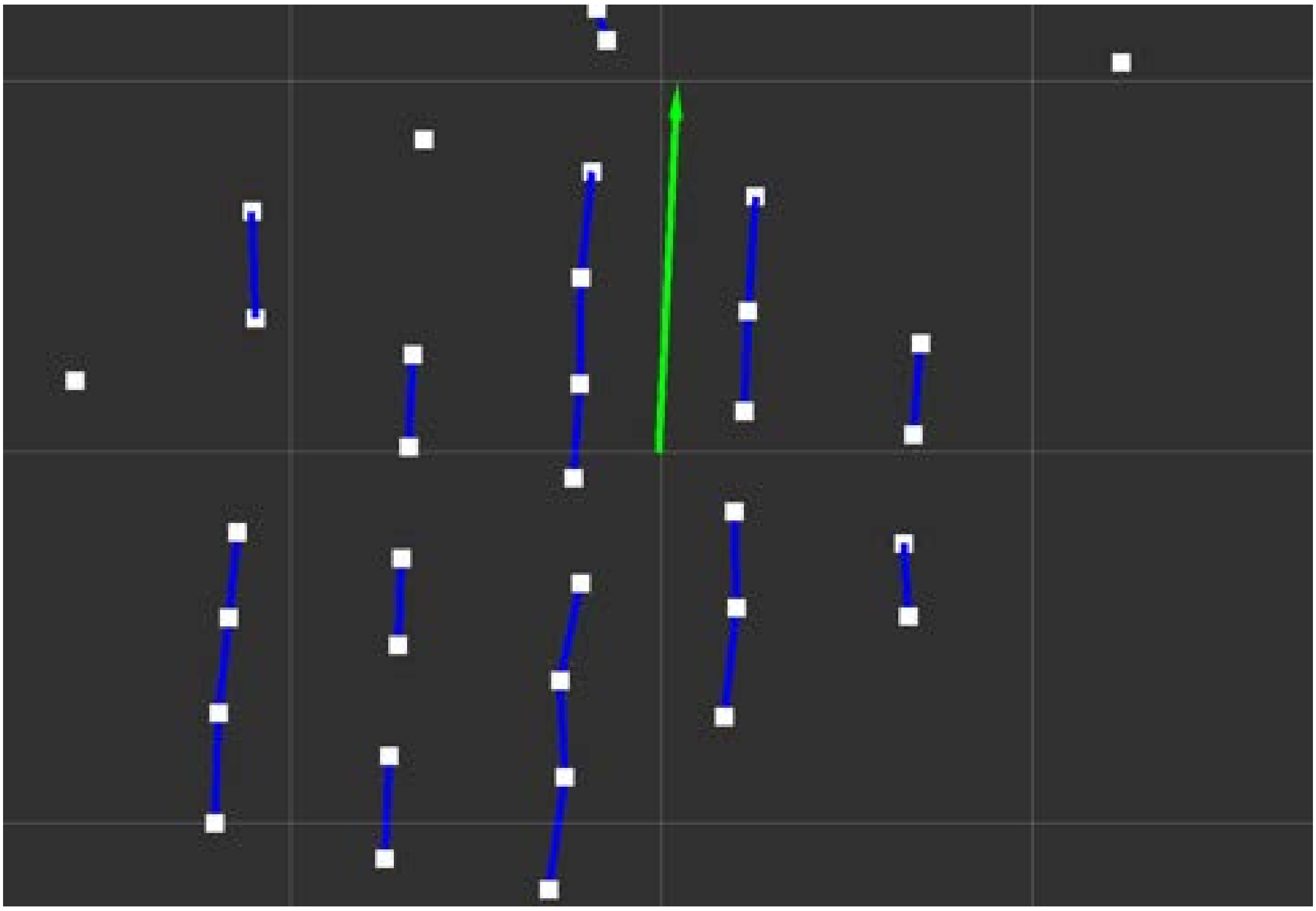

*Figure 109: Row centreline (green) calculated from the mode of nearest neighbour angles and the equation for the linear offset.*

Once the angular offset was found, the linear offset of the lidar coordinate system origin from the row centreline was calculated. The steps used were:

1. The containers of clusters, remaining at the end of the angular offset calculation, were rotated by the angular offset about the lidar origin (Figure 110). The transformed lidar axes were designated to be x' and y'.

2. The rotated data belonging to the current row was found by discarding data more than a threshold distance from the x' axis (Figure 110). The value of this threshold was the maximum row width for the orchard.

3. The data was split into left data points, $p_l$, and right data points, $p_r$, of the row with left side data having positive y' coordinates and right side data having negative y' coordinates (Figure 110). The number of left and right data points was $n_l$ and $n_r$, respectively.

4. The linear offset, $o_l$, was found by calculating averages of the left and right y' coordinates separately and then averaging the two results, according to:

$$o_l = \frac{n_r \sum_{i=1}^{n_l} (p_l[i])_{y'} + n_l \sum_{i=1}^{n_r} (p_r[i])_{y'}}{2 n_l n_r} \quad (7)$$

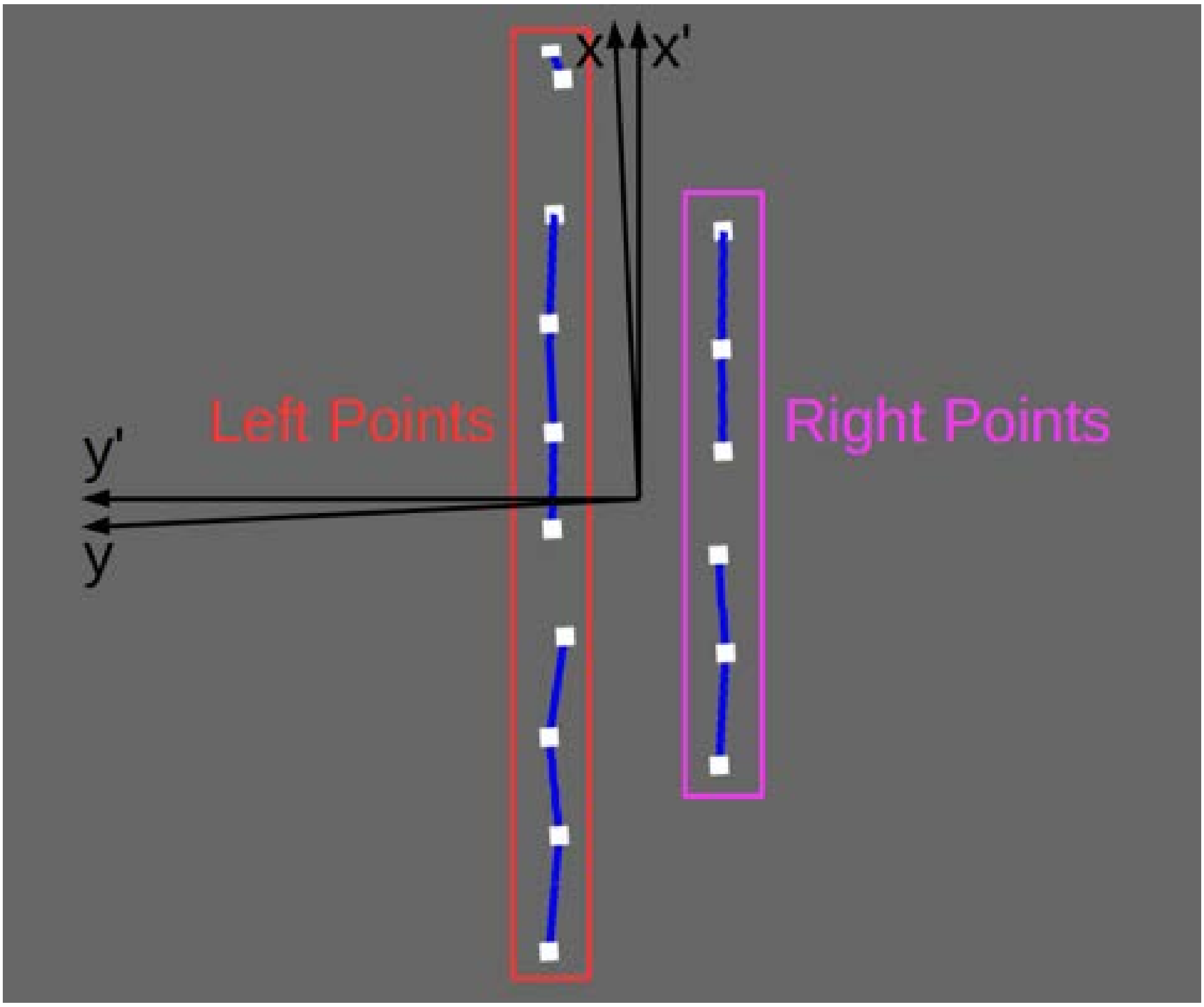

*Figure 110: The nearest neighbours from Figure 109, rotated by the angular offset, filtered based on transformed y (y') axis coordinates and grouped into left and right points.*

After the linear offset was found, the row centreline could be drawn as shown in Figure 109. Row following was performed by calculating the angular velocity from the sum of the proportional control signals for the linear and angular offsets. The linear proportional gain, $k_l$, and angular proportional gain, $k_a$, were manually tuned so that when there was a large linear displacement, the robot would turn sharply towards the centre of the row and, when the linear displacement was low, the robot would straighten with respect to the row centreline. The angular velocity command for the robot was calculated according to:

$$k_l o_l + k_a o_a \quad (8)$$

### 4.4.2 Tests of 3D Lidar Row Detection against Hand Labelled Ground Truth

A ground truth dataset of frames of lidar data was collected from a Velodyne VLP-16 [130] 3D lidar, which was mounted horizontally at a height of 0.8 m from the ground on a Pioneer P3-AT [191] mobile robot (Figure 111).

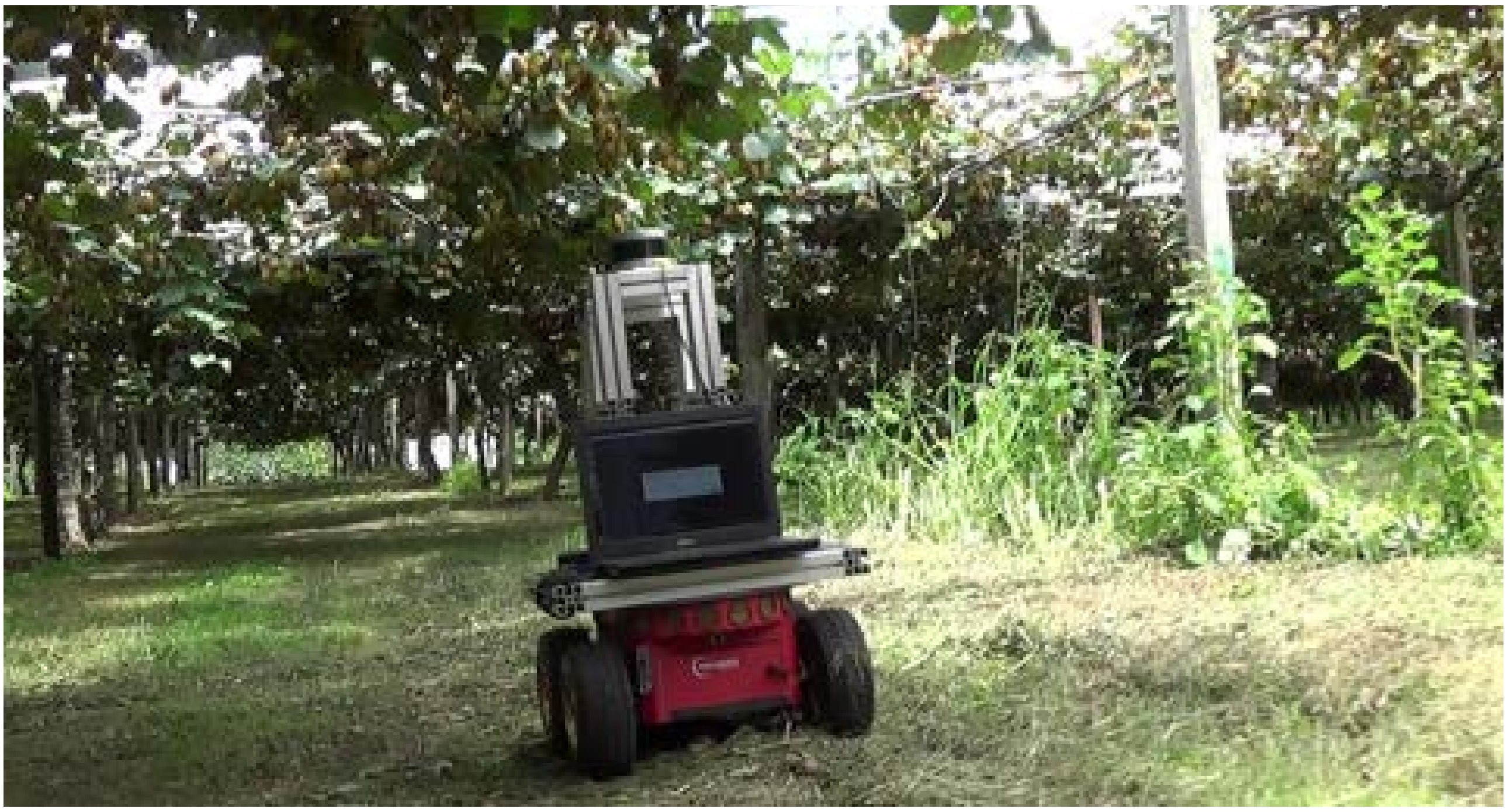

*Figure 111: Pioneer P3-AT robot with Velodyne VLP-16 3D lidar and a laptop for processing.*

The robot was manually controlled to swerve from side to side while driving through a kiwifruit orchard block. A small proportion of the captured lidar frames were labelled. Each ground truth lidar frame was manually labelled by:

1. Finding the coordinates of two points from the treeline on one side of the row.
2. Finding the coordinates of two points from the treeline on the opposite side of the row.
3. Calculating the straight line equations for each side of the row.
4. Calculating the row centreline gradient, $m_c$, and the row centreline axis intercept, $c_c$, in the lidar coordinate frame, by taking the average of the gradients and intercepts of the straight line equations for each treeline.
5. Calculating the ground truth linear and angular offsets according to:

$$o_a = atan(m_c) \tag{9}$$

$$o_l = c_c \cos(o_a) \tag{10}$$

The row detection algorithm, described in Subsection 4.4.1, was run on the same frames of lidar data that were hand labelled for ground truth. In addition, the kiwifruit row detection algorithm of Scarfe [21] was coded and run on the same lidar data. However, since this algorithm was developed for 2D lidar, only the 8$^{th}$ lidar plane was used because this plane is one of the two middle planes of the lidar. The errors in the calculation of the linear and angular offsets of each algorithm with respect to the hand labelled ground truth are shown in Figure 112. These results are also summarised in Table 24. It seems from these results that the kiwifruit row detection algorithm, described in Subsection 4.4.1 using 3D lidar, produces lower linear offset and angular offset errors than the algorithm of Scarfe [21] using 2D lidar data.

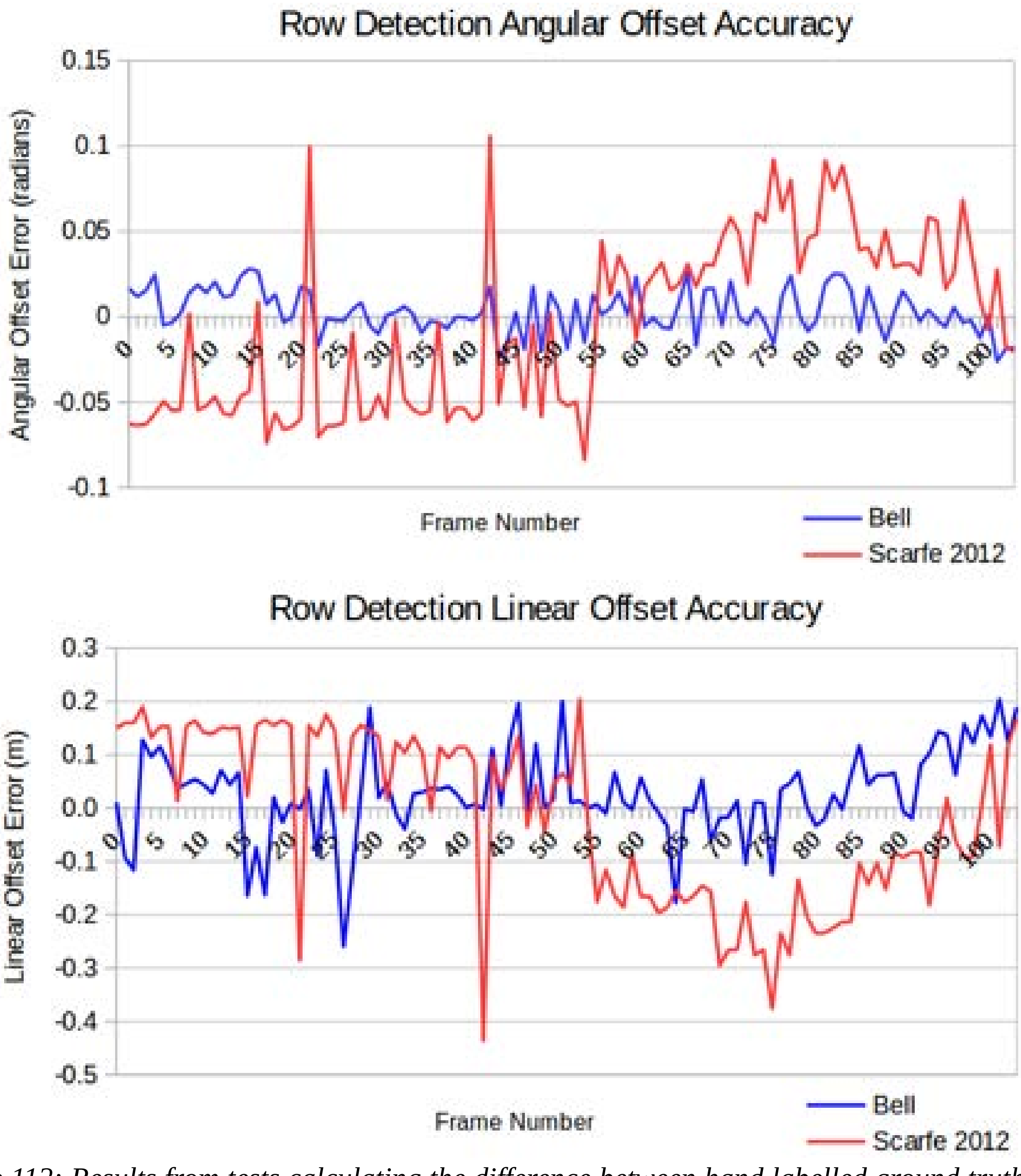

*Figure 112: Results from tests calculating the difference between hand labelled ground truth offsets and offsets calculated by different algorithms.*

*Table 24: Summarising results of comparing kiwifruit row detection algorithms to hand labelled ground truth.*

| | Mean Absolute Angular Offset Error (rad) | Mean Absolute Linear Offset Error (m) |
|---|---|---|
| **Bell** | 0.010 | 0.063 |
| **Scarfe 2012 [21]** | 0.045 | 0.142 |

Figure 113 shows data points from the last steps for the algorithm described in Subsection 4.4.1 and the algorithm of Scarfe [21] for the same lidar frame. The method of Scarfe incorrectly detects some hanging branches as posts or trunks. The method described in Subsection 4.4.1 has multiple steps that aim to discard such points by eliminating large clusters, discarding clusters that are close to multiple other clusters or using multiple planes of the lidar data, by measuring the heights of objects and thresholding objects based on those heights.

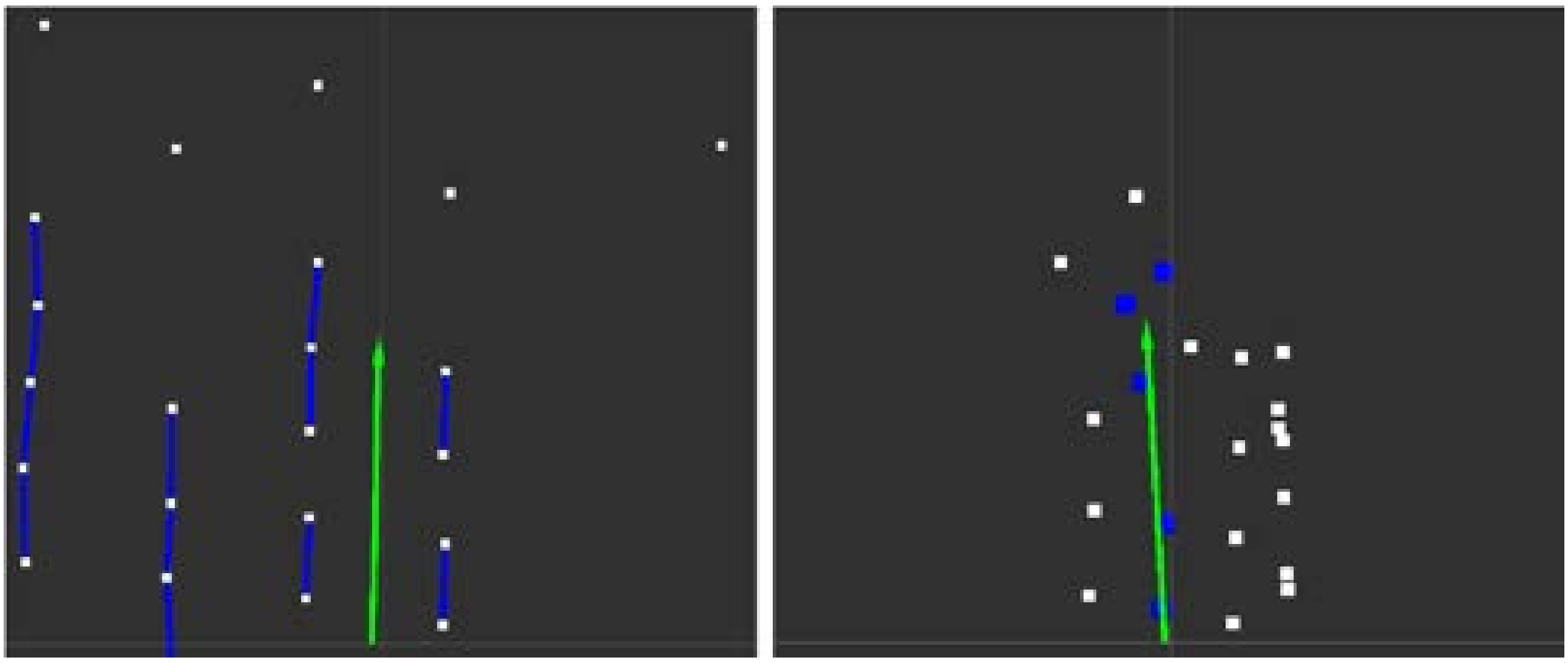

*Figure 113: Points from the last stages of kiwifruit row detection on the same frame of lidar data by the algorithm from Subsection 4.4.1 (left) and the method of Scarfe 2012 (right).*

### 4.4.3 Kiwifruit Orchard Row End Detection by Hedge Proximity Detection

Row end detection was implemented in order to allow for the localisation of the robot at the end of a row. Row end detection triggered a transition from row following to other behaviours, including row end turns. Row end detection was defined as detecting the passing of a set point on the robot beyond a set feature, which defined the end of the row. The point on the robot was initially defined to be the centre of the 3D lidar that was used for row end detection. The feature used to define the end of the row was developed with some experimentation.

It is common in kiwifruit orchards to have hedges or other barriers at the end of rows, bordering the headlands of the orchard blocks (Figure 114). When this is the case, the presence of the hedge and the distance of the hedge to the robot may be used as variables to determine that the row end has been reached. Because the robot's path is often orthogonal to the hedge before row end detection, the hedge may be modelled as an orthogonal plane; although the branches making up the hedge may create a surface that has some texture about this plane. As a result, in order to detect the hedge, a fixed volume in front of the robot was checked for a threshold number of points. The displacement of the volume from the robot was tuned to allow the robot to perform row end turns with maximum clearances.

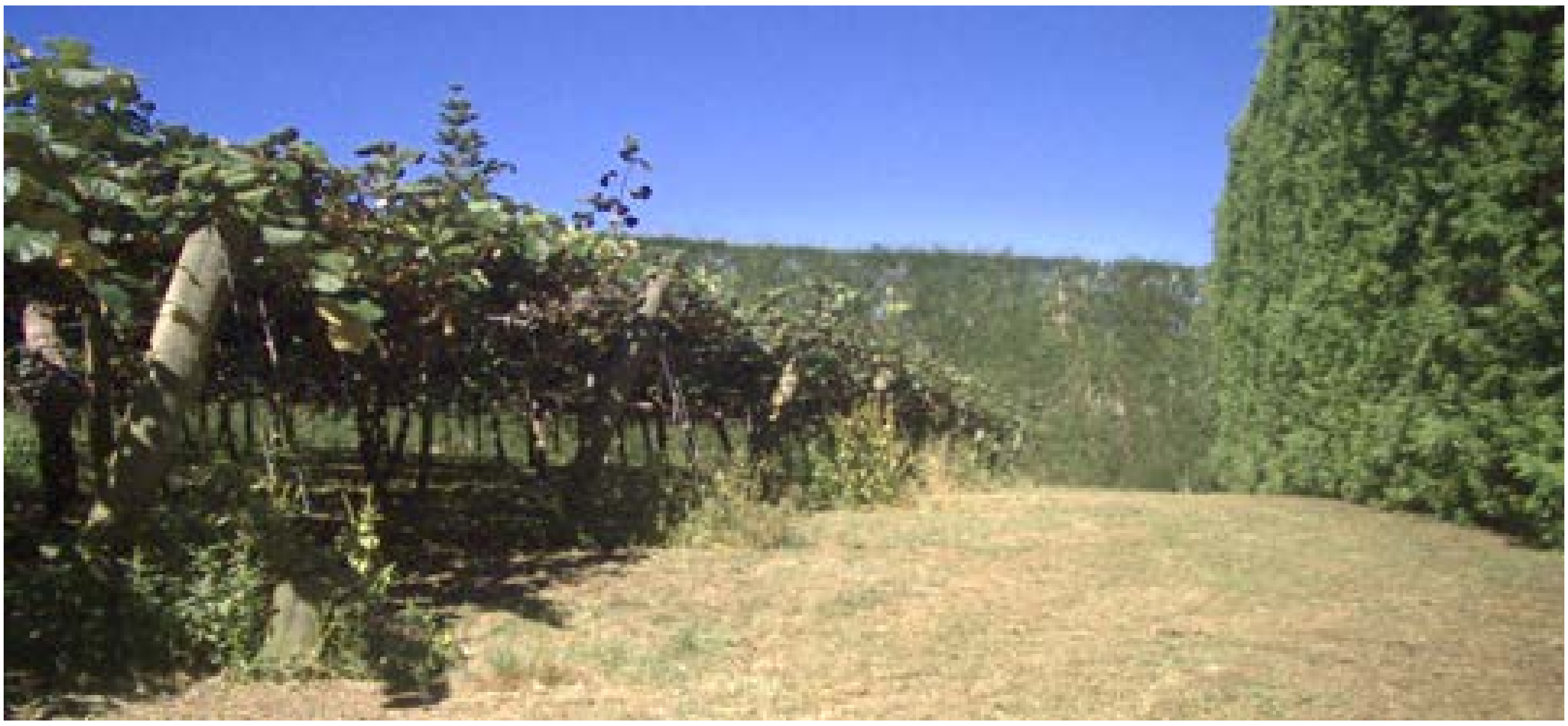

*Figure 114: A typical example of a hedgerow in the headland of a kiwifruit orchard.*

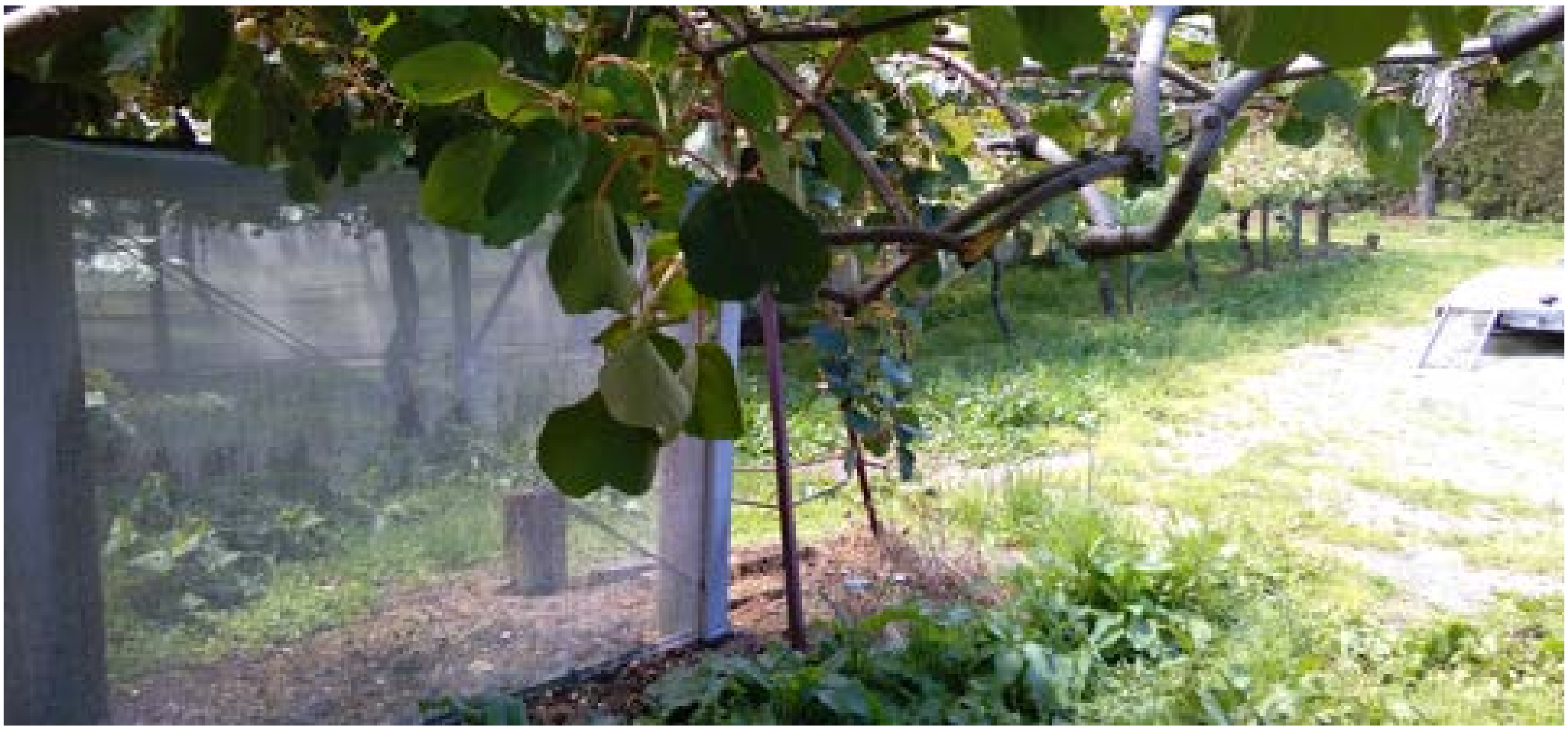

*Figure 115: An example of a row end that ends in a loading area, even though adjacent rows end with a hedgerow.*

However, there are other situations, where the row may not end in a hedge; instead, there may be unused space, a driveway, a gap in the hedge or an open space used as a loading area, as shown in Figure 115. This suggests that row end detection by hedge proximity may only be used at some row ends and that at other row ends there must be alternative methods used.

### 4.4.4 Kiwifruit Row End Detection by Sensing Canopy Absence Above

In a kiwifruit orchard with a pergola structure, where the canopy is overhead, a large gap in the canopy above and to at least one side is a sign that the end of the row has been reached. Canopy detection was performed by counting the number of points from the 3D lidar in two volumes of interest. The two volumes of interest were used for the left and right sides of the data. It was necessary to consider both sides separately because, in places where there was a hedgerow or other object adjacent to the row end on one side, there may have been more than the threshold number of points in the volume of interest. When there was less than the threshold number of points in one of the volumes of interest, canopy detection was false and this was used as a positive indication of row end detection. By setting the lowest point of the volumes of interest to a lower height, the system was also checking for the presence of posts and trunks, which was another indicator that the robot was in the row, when presence was detected. When the presence of posts and trunks was not detected, this was an indicator that the robot was in a row end.

This method of row end detection by counting points in volumes of interest had one detected failure after 10 km of testing. During the failure case, a false positive row end detection occurred. At that time the lidar was wet from rain and became covered by leaves from hanging branches; these factors may have meant that the lidar received so few measurements in the canopy detection volumes of interest, that the canopy detection threshold was not exceeded. The solutions that were implemented to fix this issue were:

- Defining that the end of row was not detected until canopy detection had been false for a threshold number of times consecutively, using a threshold of 5.
- Using multiple methods for row end detection at the same time, where possible.

In addition, it was proposed that multiple sensors could be used to overcome the issue of an occluded sensor. Options considered were:

- Using the data from multiple lidars so that a single 3D lidar becoming covered would not create a region of no data.
- Using the upwards facing sensors on the boom navigation, pollination and harvesting systems to detect the presence or absence of canopy.

### 4.4.5 Naive Row End Turns

The approach taken with row end turns for kiwifruit orchards was to drive into an adjacent row, after exiting the current row. Row end turns were initially simple when testing with the Pioneer P3-AT robot [191] (Figure 111) because this robot was small and hence the clearances in the headlands of the kiwifruit orchard could be made to be large. However, the row end turns became more complex, when developing and testing with the AMMP because the clearances were smaller.

The first strategy trialled for performing row end turns was to use a sequence of fixed open-loop motion primitives for each turn. It was hypothesized that if the row following was sufficiently robust- in the sense that the linear and angular offset of the robot from the row centreline was sufficiently and consistently small- then it may be possible to use open-loop control for a small section of driving, such as a row end turn.

The sequence of motion primitives initially trialled was a fixed length straight section of driving, to allow the robot to drive out of the row that it was exiting, followed by a fixed radius turn for a fixed length, as measured using wheel encoders. An issue with not having the straight section before the fixed angle turn for a larger robot is illustrated in Figure 116. The side of the robot on the outside of the turn tended to get very close to a post when entering the next row; in fact, the AMMP had to be stopped during testing to avoid these collisions in some instances. To reduce the probability of these collisions occurring, a straight section was added before the start of the turn so that the turn occurred more in the free space of the headlands of the orchard.

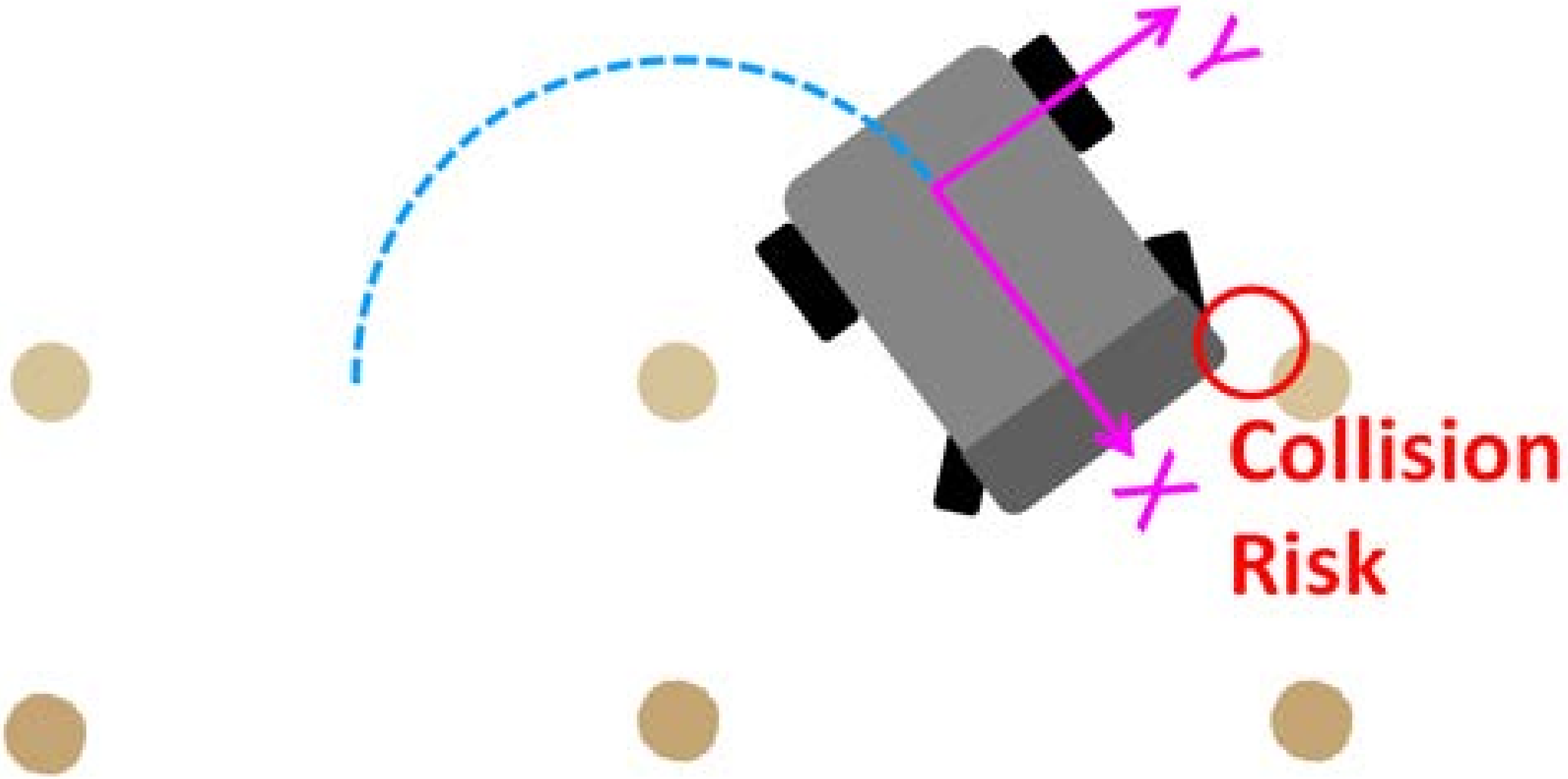

*Figure 116: A fixed angle turn, centred about the last post in a treeline, resulted in a collision risk.*

To further reduce the collision risk at the outer side of the robot during a row end turn, the diameter of the turn was reduced to less than the width of a row. However, in this case, if the robot was allowed to perform a full 180 degree turn, it would finish the turn with a linear offset and hence the row following behaviour would immediately have to correct this offset inside the row. Instead, the fixed radius turn length was tuned to be less than 180 degrees and just enough, so that after the turn the row following behaviour would smoothly correct the linear and angular offsets before entering the row. An example of the row end manoeuvre up to the point where row following recommenced is given in Figure 117.

Although simple, the strategy of using fixed radius turns allowed 100% completion of turns for the AMMP for 18 row end turns trialled. However, in 6 of those turns the AMMP came within 0.2 m of a post in the way illustrated in Figure 116. This inconsistency may have been caused in part by the variation in the row following behaviour at the end of the rows before the row end turns; small linear and angular offsets at the end of the row before the row end turn may have caused the AMMP to get closer to posts. Variability in the row end detection trigger point may have also had an effect because if the row end detection came later the robot would perform more of the row end turn in the orchard headland. Also, because the turn was performed without correction by a global reference, there would have been some drift in the turn itself which could have caused the variability in the proximity to posts.

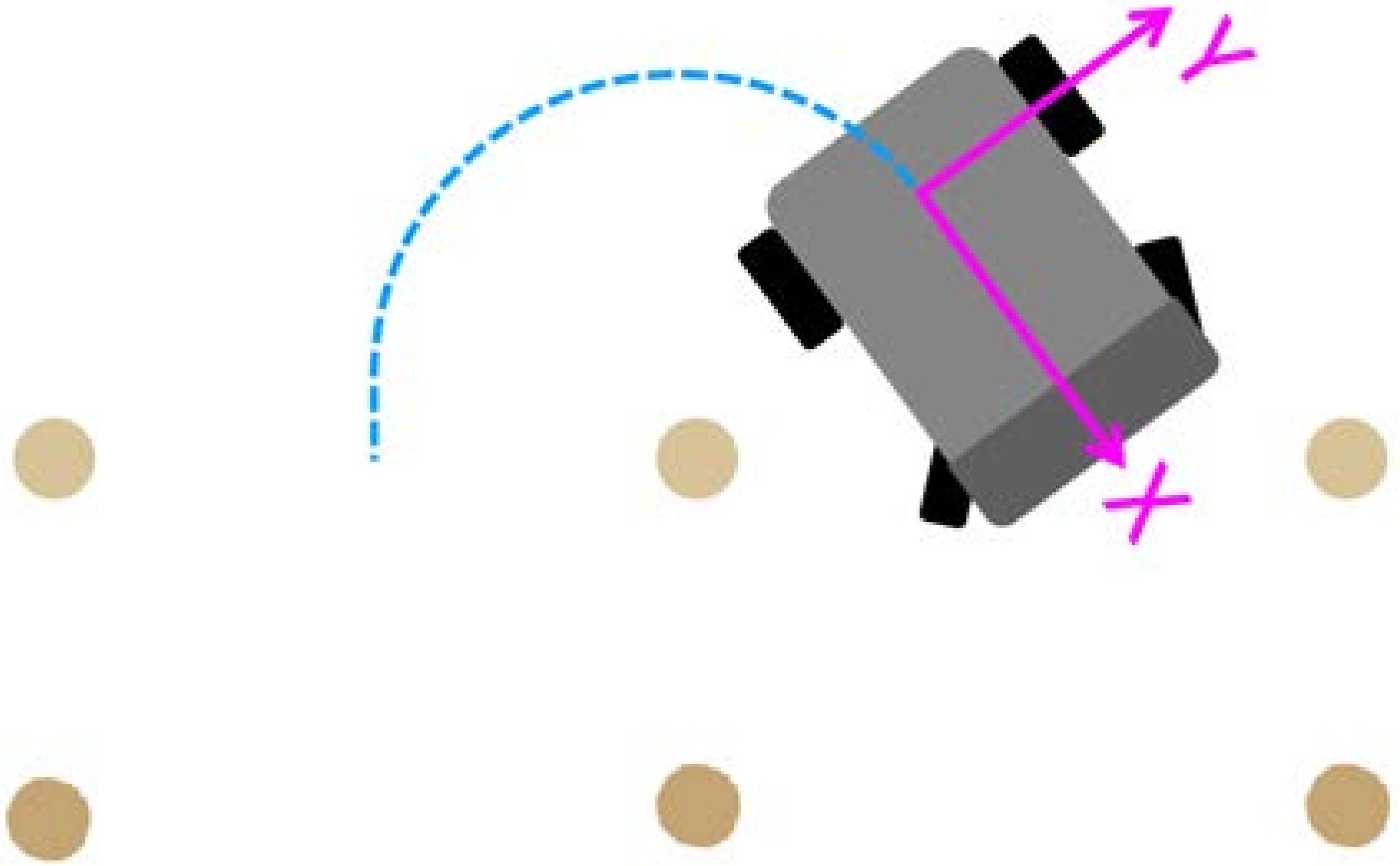

*Figure 117: The naive row end turn, where a short straight section is followed by a fixed curvature turn with diameter less than the row width and angular displacement less than 180 degrees.*

### 4.4.6 Naive Row End Turns with Bumpers

To further reduce the risk of collisions in the way illustrated in Figure 116, a method of adding some local adjustment to the fixed curvature row end turns was tested. To achieve this, a *bumper field*, which was a volume of interest in the lidar data, was checked in order to determine if the AMMP was getting too close to a post at the front of the AMMP (Figure 118). An object detection in the bumper field was defined to be more than a threshold number of points counted in the volume of interest. If an object was detected in the volume of interest, a fixed parameter was subtracted from the magnitude of the turn radius, in order to make the turn sharper. Using this method, the AMMP completed 36 consecutive row end turns without the AMMP coming within 0.2 m of a post.

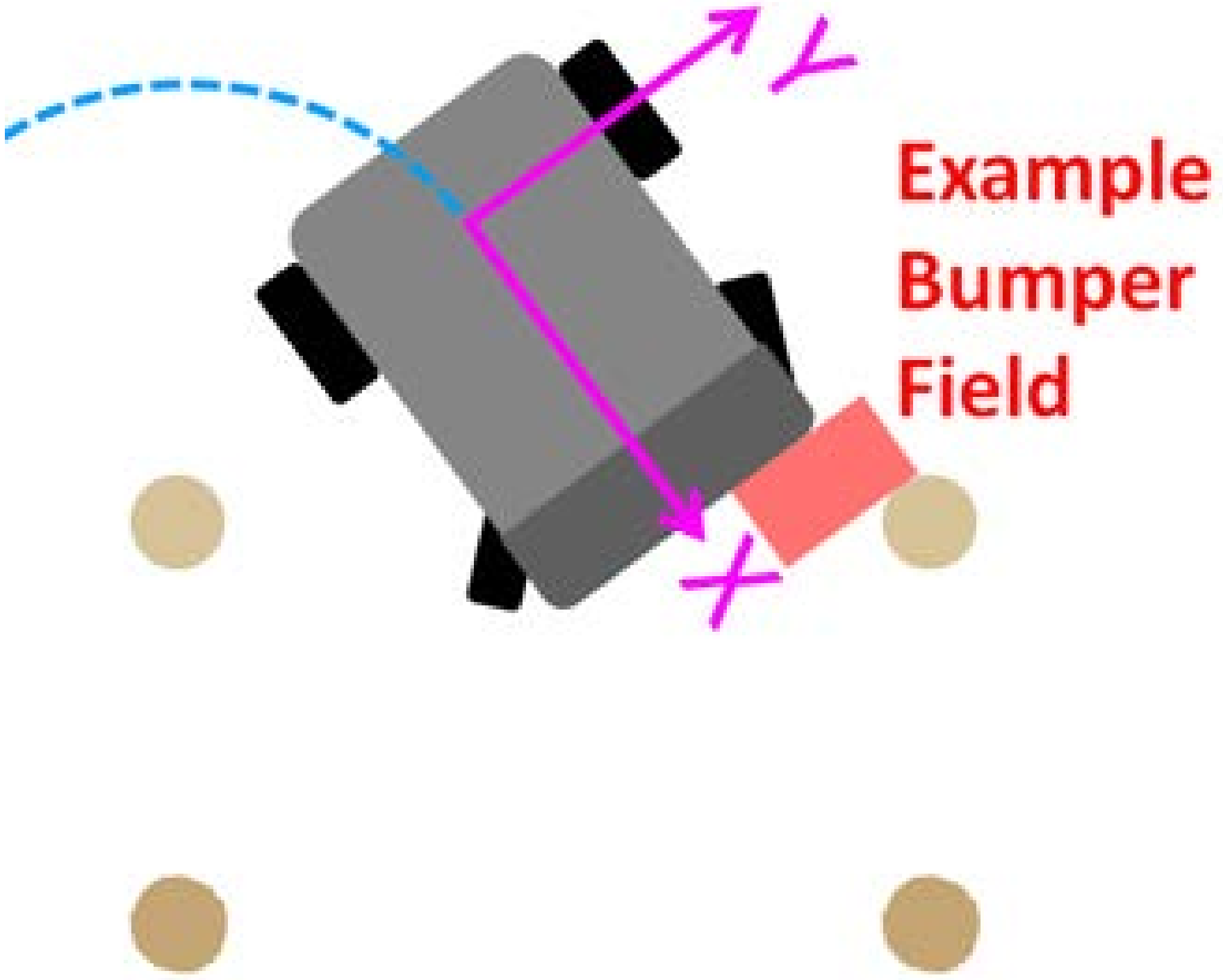

*Figure 118: An illustration of a bumper field, which is a volume of interest, where if there is an object detected, the robot turns more sharply to avoid the obstacle.*

### 4.4.7 Kiwifruit Orchard Topological Map and Paths

It seemed that the algorithms developed for row following, row end detection and row end traversal would work well in a topological map abstraction and did not require a geometric map. A topological map would not preclude the use of geometric maps as well; however, in the case where geometric maps were used, it was decided to use them in the overall framework of the topological map.

The topological map for an orchard block was represented with each end of each row forming a node (Figure 119). The vertices of the graph were row end manoeuvres and row traversal manoeuvres. To define a path in the topological map a sequence of row ends was selected according to the rules that:

- Each row traversal had to be followed by a row end manoeuvre.
- Each row end manoeuvre had to be followed by a row traversal.

Once the sequence of row ends was known, a corresponding sequence of states was generated with each row end manoeuvre and each row traversal consisting of multiple states. Each state had a set of associated parameters, including the driving speed, turn radius and boundary conditions. An excerpt from the start of a path is shown in Table 25 and an interpretation of this path is given in Figure 120. The path was read by the navigation program and the navigation system performed autonomous driving according to the current state and parameters until the boundary condition for the state was met, at which point the program moved onto the next state. This continued until the end of the path was reached.

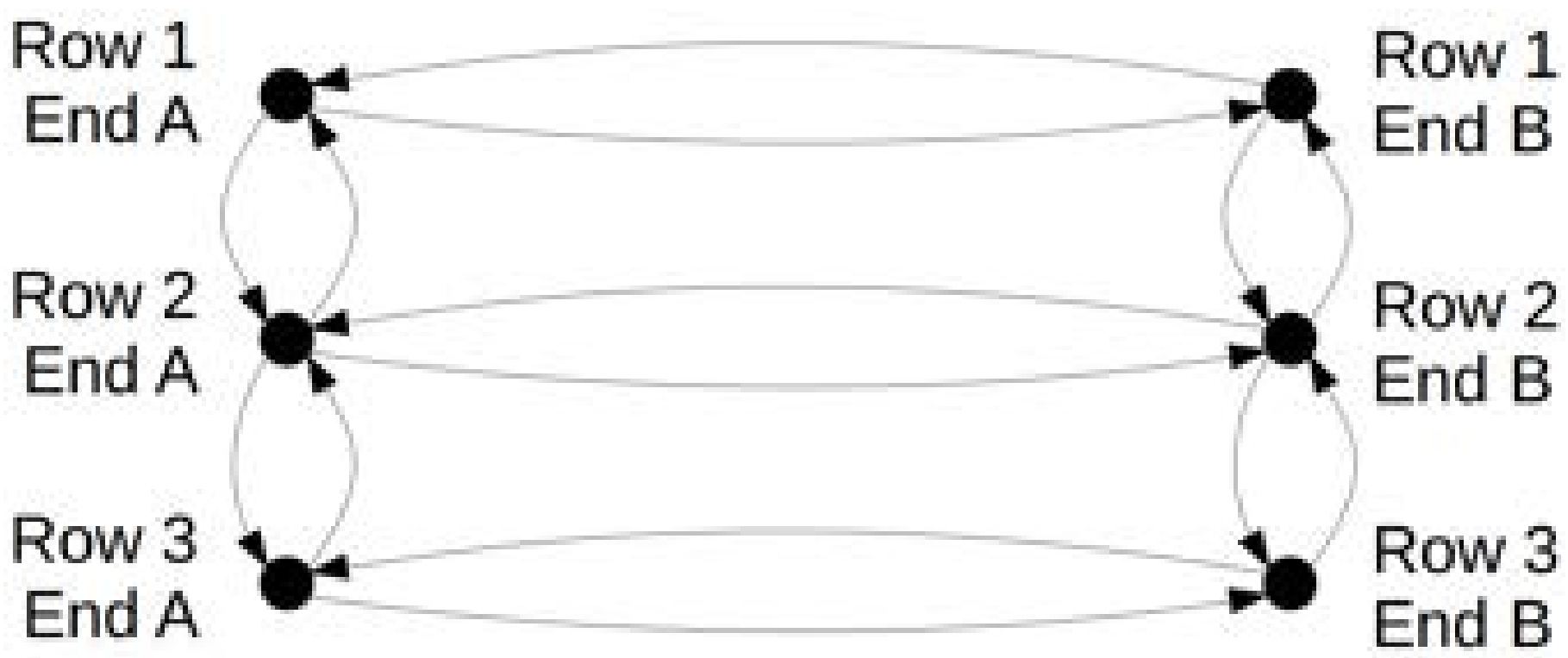

*Figure 119: The nodes in the topological map were the ends of rows and the vertices in the topological map were row traversal and row end traversal manoeuvres.*

*Table 25: An excerpt from a fixed path taken from the start of the first row to an entry point of the second row of an orchard block, where each column represents a state in the path.*

| **State ID** | 10 | 11 | 12 | 13 | 14 | 15 | 20 |
|---|---|---|---|---|---|---|---|
| **Vertex Type** | IN_ROW | IN_ROW | ROW_END | ROW_END | ROW_END | ROW_END | IN_ROW |
| **Offset in Location (m)** | 0.0 | 0.0 | 0.0 | 0.0 | 0.0 | 0.0 | 0.0 |
| **Target Speed ($ms^{-1}$)** | 0.3 | 1.0 | 0.3 | 0.3 | 0.3 | 0.3 | 0.3 |
| **Goal Turn Radius (m)** | 1000000000 | 1000000000 | 1000000000 | 2.0 | 1.7 | 1.7 | 1000000000 |
| **Boundary Condition Criteria** | FIXED_ LENGTH | NO_CANOP Y | HEDGE_ OFFSET | FIXED_ LENGTH | FIXED_ LENGTH | DEFINITE_ ROW | FIXED_ LENGTH |
| **Boundary Condition Parameter** | 5.0 | 5 | 2.3 | 2.5 | 2.7 | 5 | 5.0 |
| **Additional Parameter** | NONE | NONE | NONE | NONE | USE_BUMPE R | NONE | NONE |
| **Additional Parameter Value** | 0 | 0 | 0 | 0 | 0.05 | 0 | 0 |

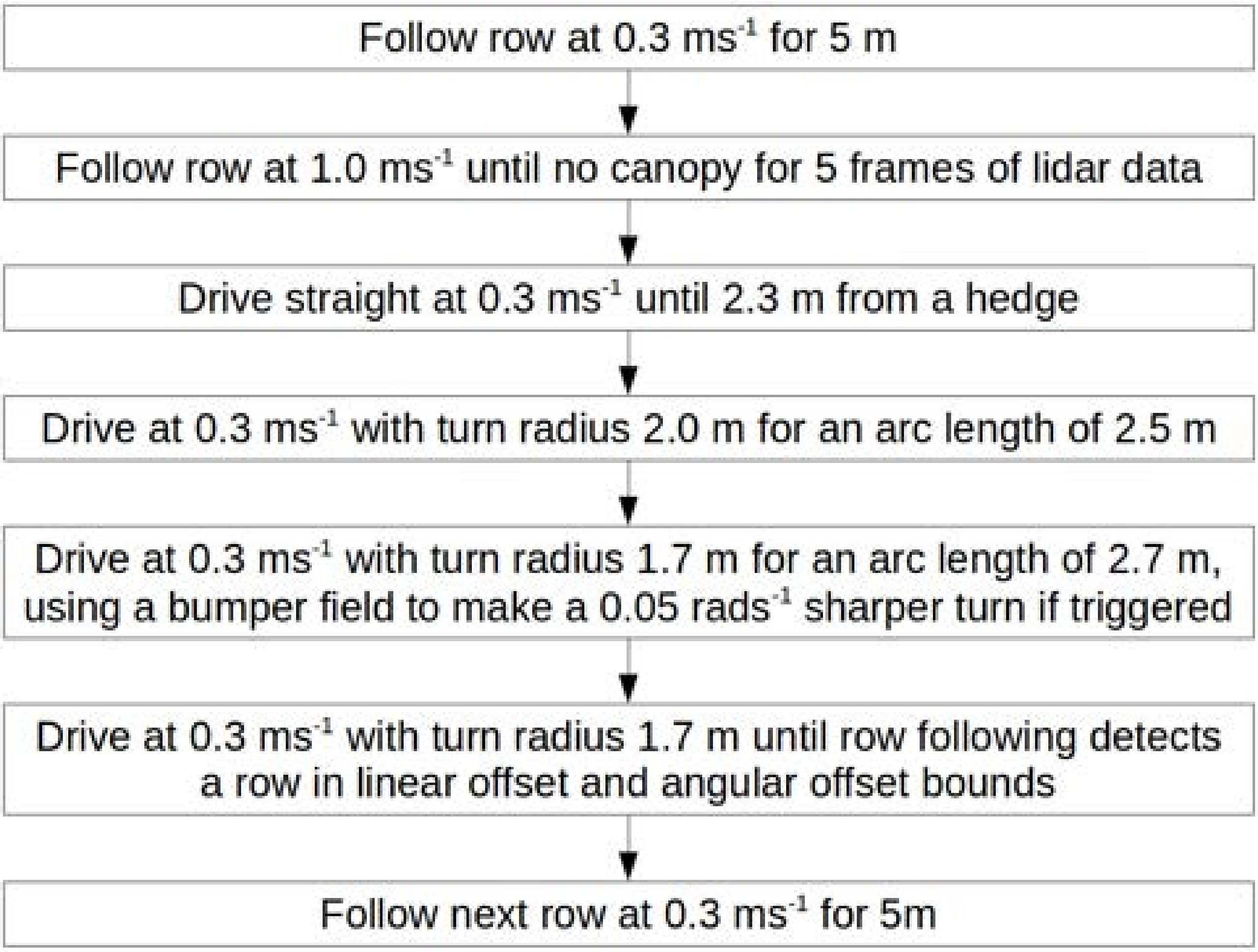

*Figure 120: Interpretation of the path excerpt from Table 25, where each block here represents a column/ state from Table 25.*

### 4.4.8 3D Lidar Kiwifruit Orchard Autonomous Driving

The 3D lidar navigation system, consisting of the row detection, row following, row end detection, row end traversal and topological map traversal algorithms, was tested in two different kiwifruit orchards. Three different robotic platforms were used, starting with the Pioneer P3-AT robot [191] (Figure 111), followed by testing on a Clearpath Husky robot [96] (Figure 121) and with most of the testing performed on the AMMP. The navigation system was run on different laptops. The autonomous driving was performed using just 3D lidar and wheel encoder data. The lidar was mounted at a height of approximately 0.8 m from the ground and was set to output data at 10 Hz. The wheel encoders were only used to measure the distance travelled during manoeuvres that had a fixed length.

Overall, 30 km of autonomous driving testing was performed. For this testing, the top speed set varied between 0.5 $ms^{-1}$ and 1.4 $ms^{-1}$. The key issues observed and fixed during this testing were regarding row end detection and row end manoeuvres, as described in Subsections 4.4.3, 4.4.4, 4.4.5 and 4.4.6. After the final version of the navigation algorithm was frozen, over 10 km of autonomous driving was performed with the AMMP in kiwifruit orchards without any detected errors.

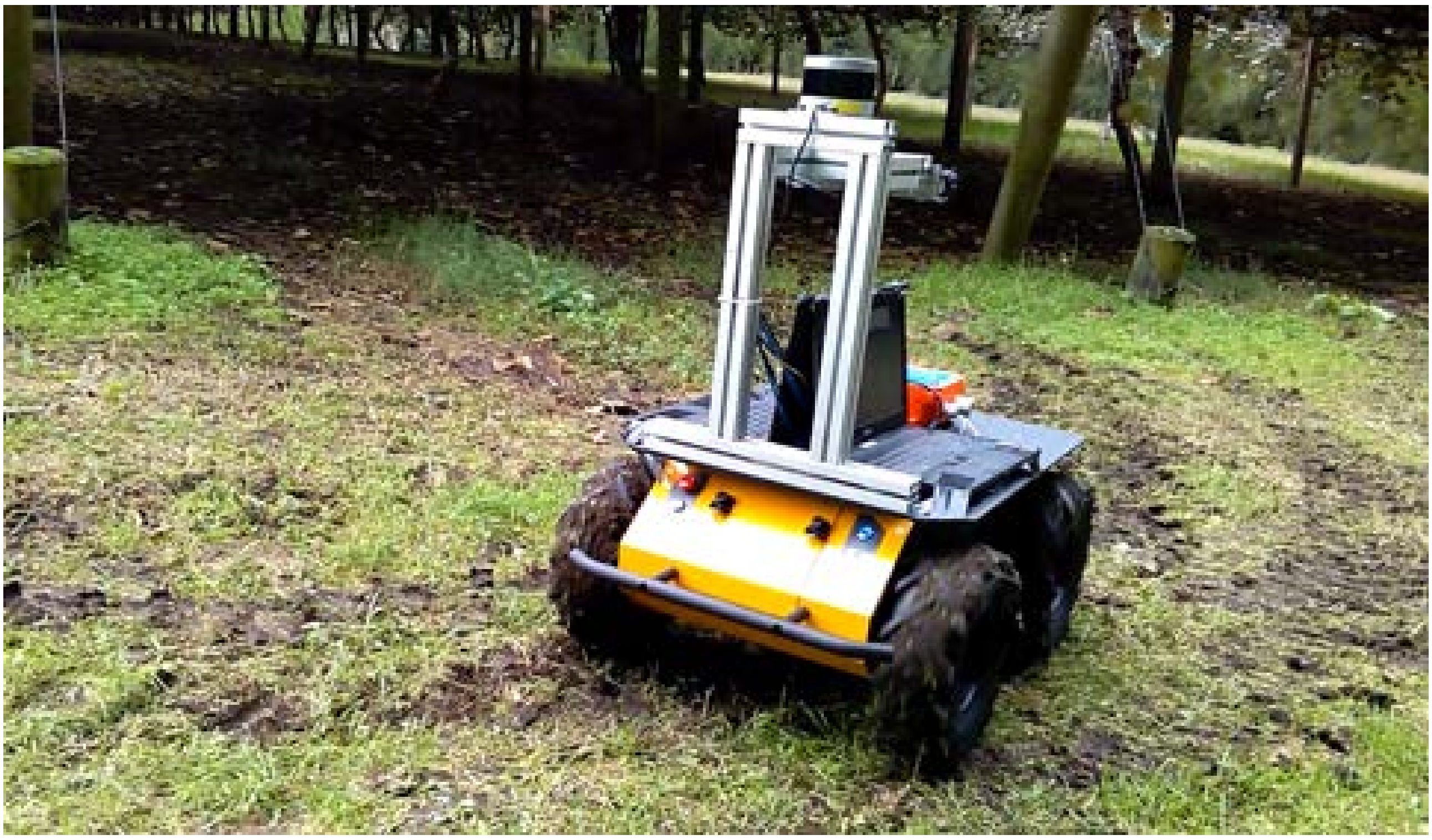

*Figure 121: The Clearpath Husky robot used for early kiwifruit orchard autonomous driving testing with a Velodyne VLP-16 3D lidar.*

### 4.4.9 3D Lidar Kiwifruit Orchard Autonomous Navigation Conclusions

The 3D lidar row following method presented here used multiple steps to extract the data from the posts and trunks, while filtering out data from the canopy, hedges and weeds. The approach of using the nearest neighbours from multiple rows allowed for the calculation of the heading of the orchard rows, which was key to calculating the linear offset of the robot from the row centreline. This method compared favourably with an existing 2D lidar and digital compass kiwifruit orchard row detection method. This result may be due in part to the extra information that can be gleaned from 3D lidar. However, besides the performance increase, the navigation system presented here had the advantage, over the existing kiwifruit row detection system, of not requiring the digital compass and the prior measurements of row angles in order to calculate the angular offset.

The method of kiwifruit row end detection using the absence of canopy proved to be robust, especially when multiple consecutive frames of lidar data were used. In addition, row ends were detected using the proximity to hedgerows for many rows.

Row end turns using a series of motion primitives worked well for smaller robots because there was ample space for the robots to complete the turns without collision. With the AMMP, objects were detected close to the front corners of the AMMP in order to perform obstacle avoidance and allow the AMMP to complete row end turns consistently. The final navigation system, based on row following and row end turns, was able to repeatedly navigate kiwifruit orchard blocks without intervention.

## 4.5 3D Lidar Feature Extraction for SLAM and AMCL

The key weaknesses of the 3D lidar kiwifruit navigation system were that the parameters in the paths, like those in Table 25, required manual tuning and there was no strategy for driving between orchard blocks or across the headlands of an orchard. It was proposed that a more general navigation algorithm could be used to overcome these issues. It was decided to experiment with using Gmapping [192], [193] to create a map and using Adaptive Monte Carlo Localisation (AMCL) [194], [195] to enable localisation in the map, while traversing waypoints. The implementation of Gmapping and AMCL used both took odometry and a 2D laser scan as input. However, it was thought that using a 2D laser scan would have the issues of data from the non-structure defining features, such as the canopy, ground, weeds, hanging branches and other objects in a kiwifruit orchard. Hence, it was decided to use a 3D lidar scan but extract 2D data from structure defining features, including posts, trunks and the hedges. This subsection presents methods for performing feature extraction of such features from a single 3D lidar sensor for SLAM and AMCL.

### *4.5.1 3D Lidar Data Observations*

An example of 16 planes of lidar data from a Velodyne VLP-16 sensor [130] in a kiwifruit orchard is shown in Figure 122. From examining thousands of similar frames of lidar data, the following observations were made:

- The features that are commonly recognisable in the lidar data are the posts, trunks and hedges. The posts and trunks appear as points in a skewed grid pattern; for example, posts and trunks are visible in planes 4 to 15 in Figure 122. The hedges appear as the strongest and longest straight lines; for example, there is a hedge visible towards the right in planes 6, 9, 10 and 11 in Figure 122.
- The highest and lowest planes do not provide as much information about the rows when compared to the middle planes. This can be seen in Figure 122, where in planes 5 to 12 there are more posts and trunks visible than in planes 1 to 4 or planes 13 to 16.

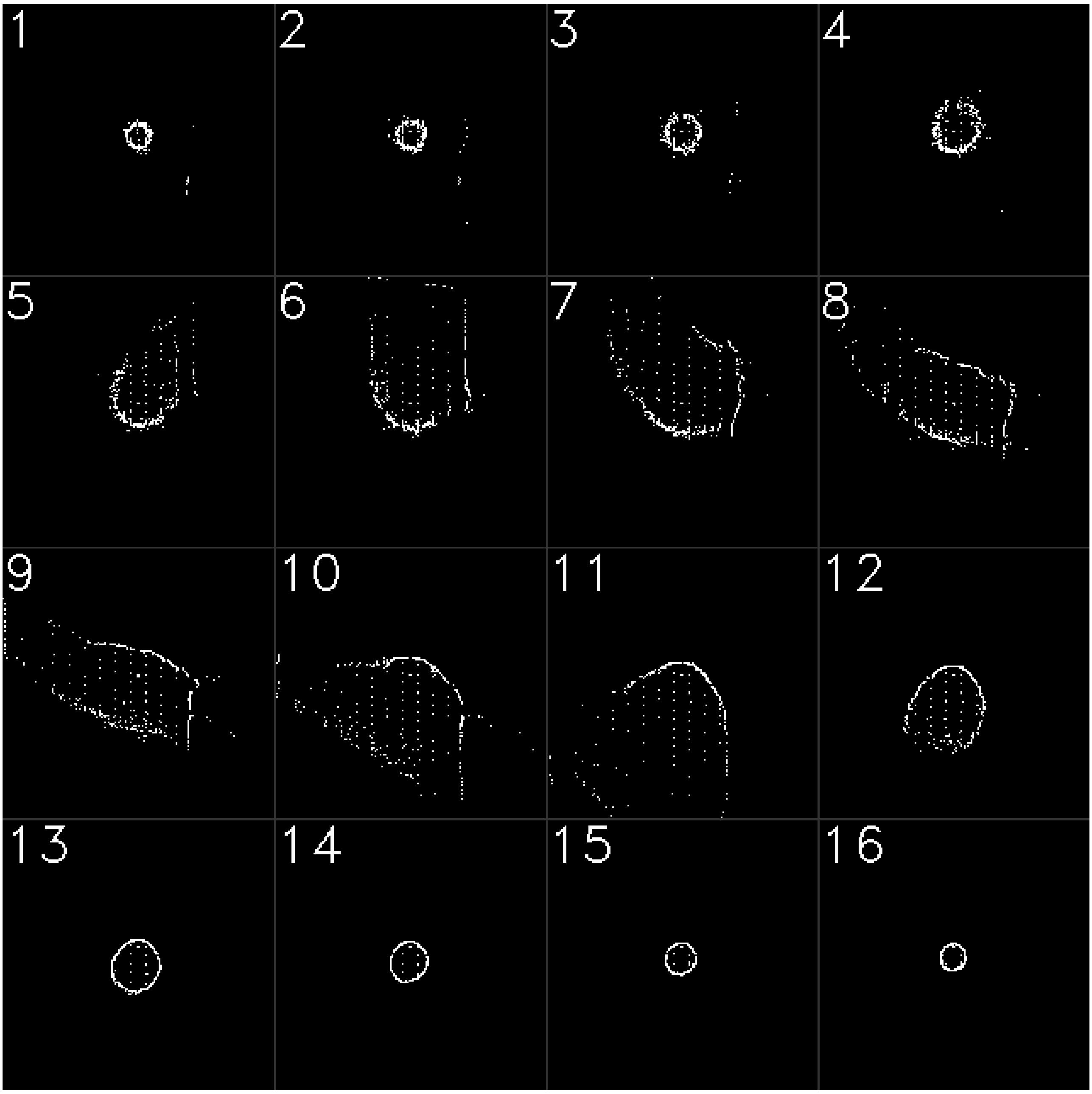

*Figure 122: Lidar data from a bird's eye view from an approximately horizontal Velodyne VLP-16, with the planes ordered from the highest plane (1) to the lowest (16), with the sensor at a height of approximately 0.8 of a metre from the ground, showing an area 80 metres by 80 metres with the lidar at the centre, near a hedge (to the right) in a kiwifruit orchard.*

- The lidar planes intersecting with the ground and canopy, instead of reflecting off features that define the structure of the orchard, such as posts and trunks, limit the range of useful data. This can be seen in Figure 122, where the data from the ground and canopy forms curves that limit the grid of points, which represents the posts and trunks.
- In different directions, different planes provide more information about the position of posts, trunks and hedges. For example, in Figure 122 planes 5-8 provide more details about the posts and trunks in the upper half of the images of the lidar data; whereas, planes 9-12 provide more information from the posts and trunks in the lower half of the images of the lidar data.

- The planes that provide more data from the posts and the trunks tend to have longer range values. For example, in Figure 122 plane 1 has shorter range values and provides less data from posts and trunks than plane 7, which has longer range values.

### 4.5.2 Hypotheses for 3D Lidar Feature Extraction

From the observations of the 3D lidar data the following hypotheses were formed:

1. Posts, trunks and hedges are the features that should be extracted from the 3D lidar data for detecting the structure of kiwifruit orchard blocks.
2. The highest and lowest planes may be disregarded during feature extraction for posts and trunks without adversely affecting results.
3. For different segments of the data it may be possible to select different lidar planes to extract the data from in order to output data with a better ratio of structured features to unstructured features.
4. The plane of the 3D lidar that has the highest range values is less obstructed by short range obstacles, such as the ground or canopy, and hence, if a single plane of the 3D lidar data was to be used, the plane with the highest range values would yield more data from posts and trunks.

These hypotheses were used to develop algorithms for extracting features from 3D lidar data, as described in the following subsections.

### 4.5.3 Plane Selection by Maximum Ranges

This subsection considers the previously proposed hypothesis: that the plane of the 3D lidar that has the highest range values is less obstructed by short range obstacles, such as the ground or canopy. Hence, if a single plane of the 3D lidar data was to be used, the plane with the highest range values would yield more useful data, which may include more measurements from posts and trunks. Essentially, what this hypothesis amounts to is selecting just one plane of the 3D lidar data to use for further processing and performing the selection of that plane by selecting the plane with the maximum range values. There are multiple metrics that could be used to define the plane with maximum range values. The metrics for determining the plane with the maximum ranges, which are considered here, are:

- The median of the range measurements in a plane. To use this metric to select a plane, the median range measurement for each plane would be found and then the plane with the highest median range measurement would be selected as the plane to use.
- The average of all ranges in a plane. To use this metric to select a plane, for each plane the average of the range measurements for that plane would be found. Then, the plane with the highest average range measurement would be selected as the plane to use.
- The maximum of all ranges in a plane. To use this metric, the maximum valid range measurement for each plane would be found and then the plane with the highest range measurement would be selected as the plane to use.

A comparison of these metrics applied to the data from Figure 122 is given in Figure 123. The average range metric method selects plane ten. The median range metric method selects plane nine. The maximum range metric method selects plane eight. For the frame of data from Figure 122, these metrics select different planes; however, for other frames of lidar data, it has been observed that two or all three of the metrics select the same plane. It is unclear just from observation of Figure 123, which metric has selected the most useful plane for further processing and so these metrics are compared along with other feature extraction methods in Subsection 4.5.7.

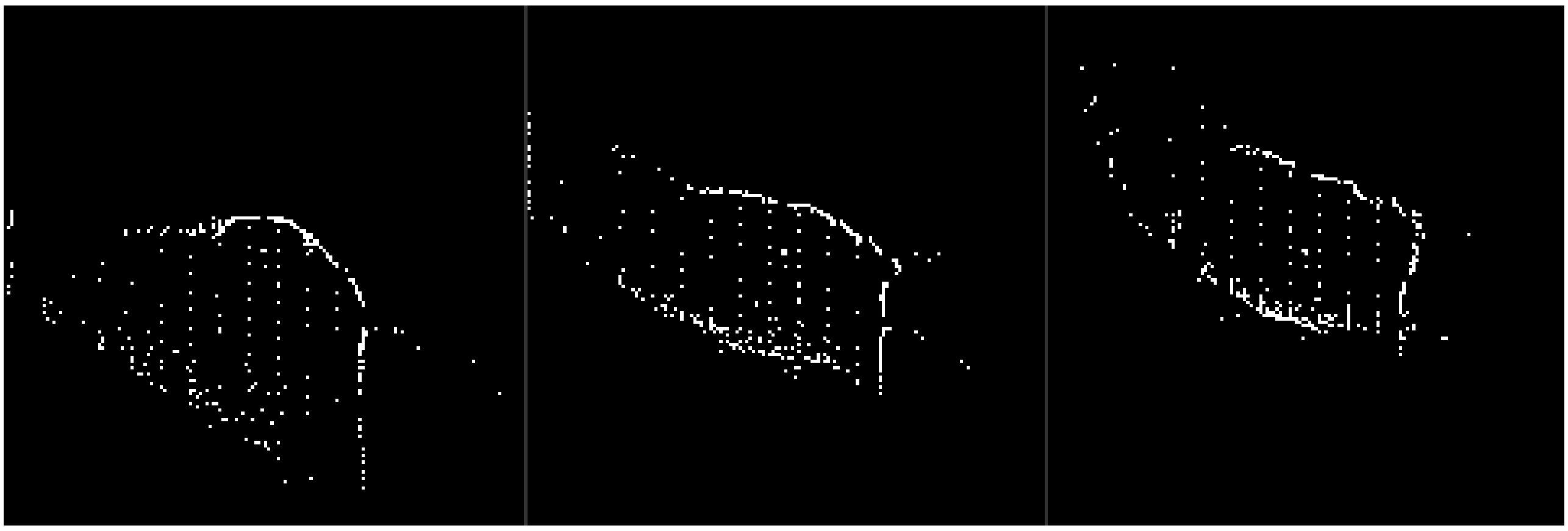

*Figure 123: Planes selected by selecting the plane with highest average range (left), the highest median range (middle) and the highest maximum range (right).*

### 4.5.4 Plane Selection by Maximum Ranges in Segments

This subsection considers the hypothesis that for different angular ranges of the data it may be possible to select different planes to extract the data from in order to produce data with a better ratio of structured features to unstructured features. Observing a variety of planes as in Figure 122, it is clear that a single plane does not necessarily provide the maximum ranges in all directions. To overcome this issue, it may be better to split the horizontal field of view into multiple segments.

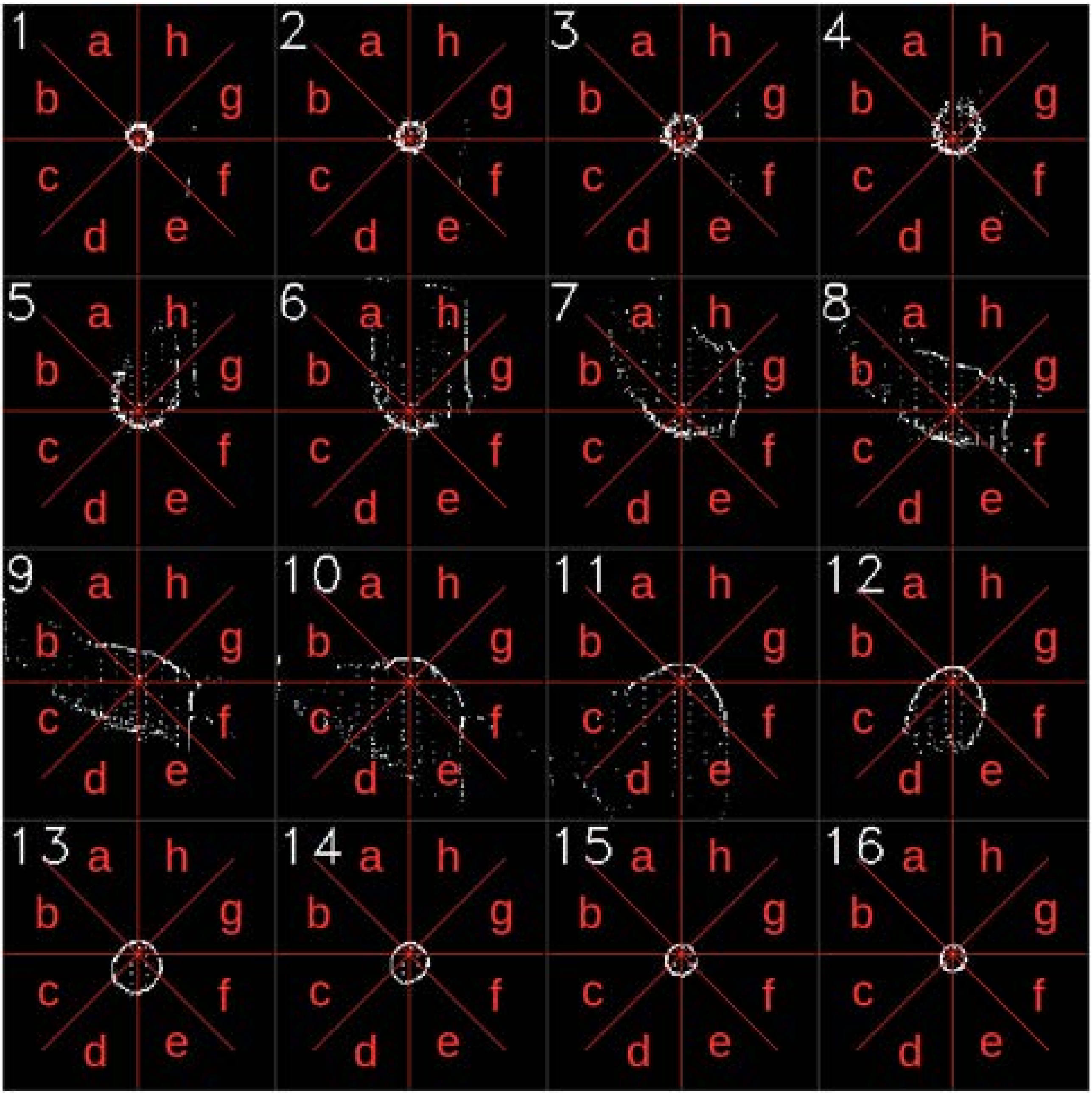

*Figure 124: Splitting a 16 plane lidar scan (1 to 16) into 8 segments (a to h) for the purpose of selecting the best plane in each segment and producing an output that maximises post and trunk features, while minimising ground and canopy data extracted.*

Figure 124 shows the lidar data from Figure 122 split into 8 segments (labelled "a" to "h"). In order to produce cleaner data with more features from posts and trunks and less features from the canopy and ground, the best plane for each segment is selected. For example, for segment "a", the maximum range values for each lidar plane are found, using one of the metrics from Subsection 4.5.3 and then the plane with the maximum range values for segment "a" is selected as the plane to take data from. This procedure is repeated for segments "b" to "h". Figure 124 demonstrates splitting the lidar data into 8 segments but it is possible to use any number of segments from 1 (which is the case considered in Subsection 4.5.3) up to the full horizontal resolution of the data.

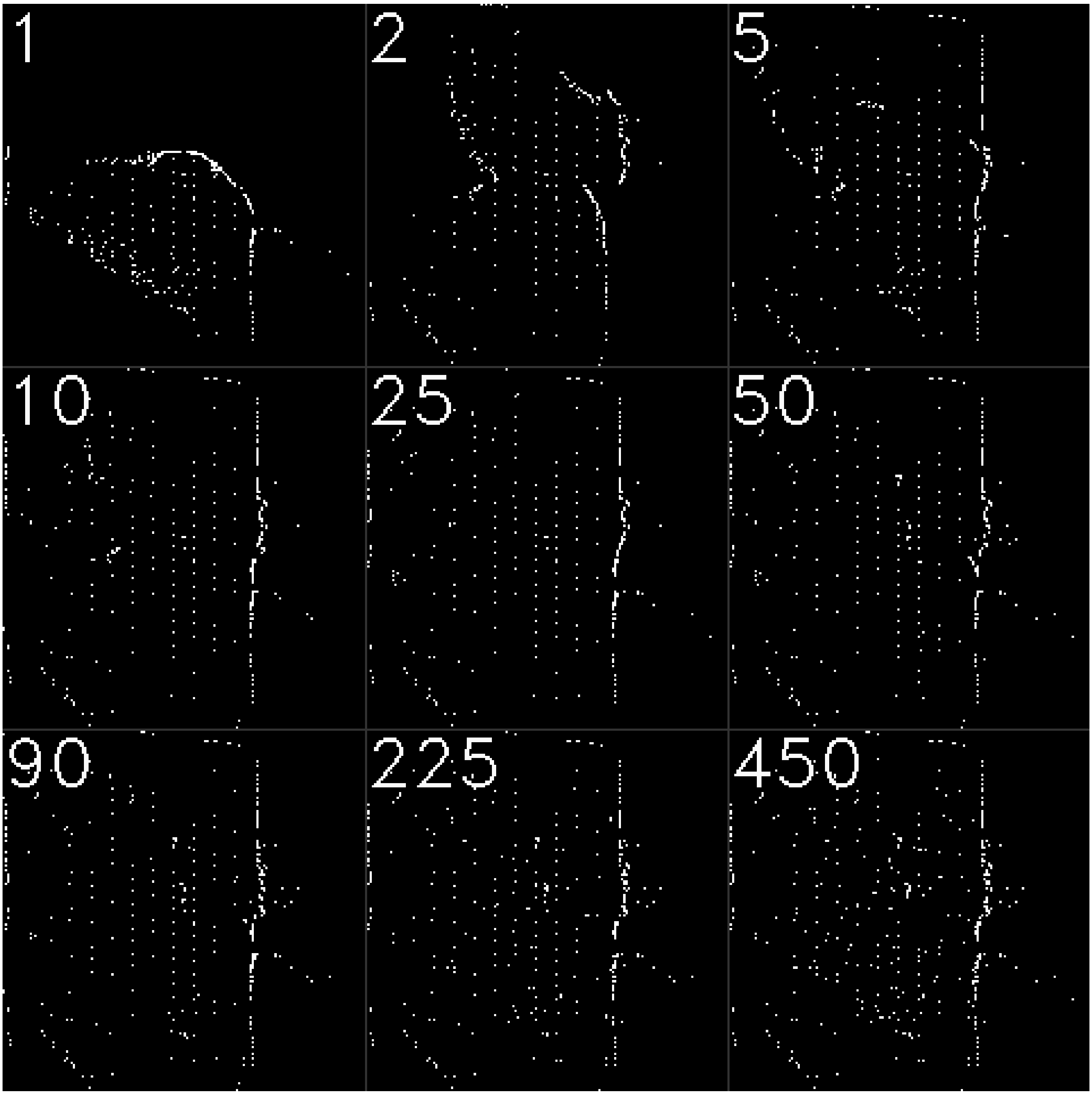

*Figure 125: Example results of selecting the plane that gives the maximum average range in each segment of lidar data, where the number of segments used is given with the associated result.*

Example results from performing plane selection by maximum ranges using the mean range metric for different numbers of segments are given in Figure 125. The use of plane selection from a moderate number of segments appears to give an output with more posts and trunks, while reducing the amount of canopy and ground data, when compared to selecting just a single plane. For example, in Figure 125 using 25 segments seems to give many more posts and trunks, while only leaving a small amount of data from the ground and canopy, when compared to the one segment case in the top left hand corner of the same figure. However, using too many segments such as 225 or 450 seems to give very noisy data. This may be due to the posts and trunks being relatively thin objects and hence for a given azimuth or angle of data it is possible for a measurement to be taken on a post or trunk but also behind it, in which case the measurement behind will be the selected point of data (Figure 126).

Figure 125 appears to show that selecting the plane with maximum ranges for a moderate number of lidar data segments, numbering in the order of 10 to 50 segments, produces an output with an improved ratio of structured to unstructured data- especially, compared to any single plane selected by maximum ranges or the aggregate of all 16 planes without any such plane selection. However, Figure 125 does not show whether such results also hold for other frames of lidar data or how well this method removes unstructured data, compared to other feature extraction methods. Such testing and results are described in Subsection 4.5.7.

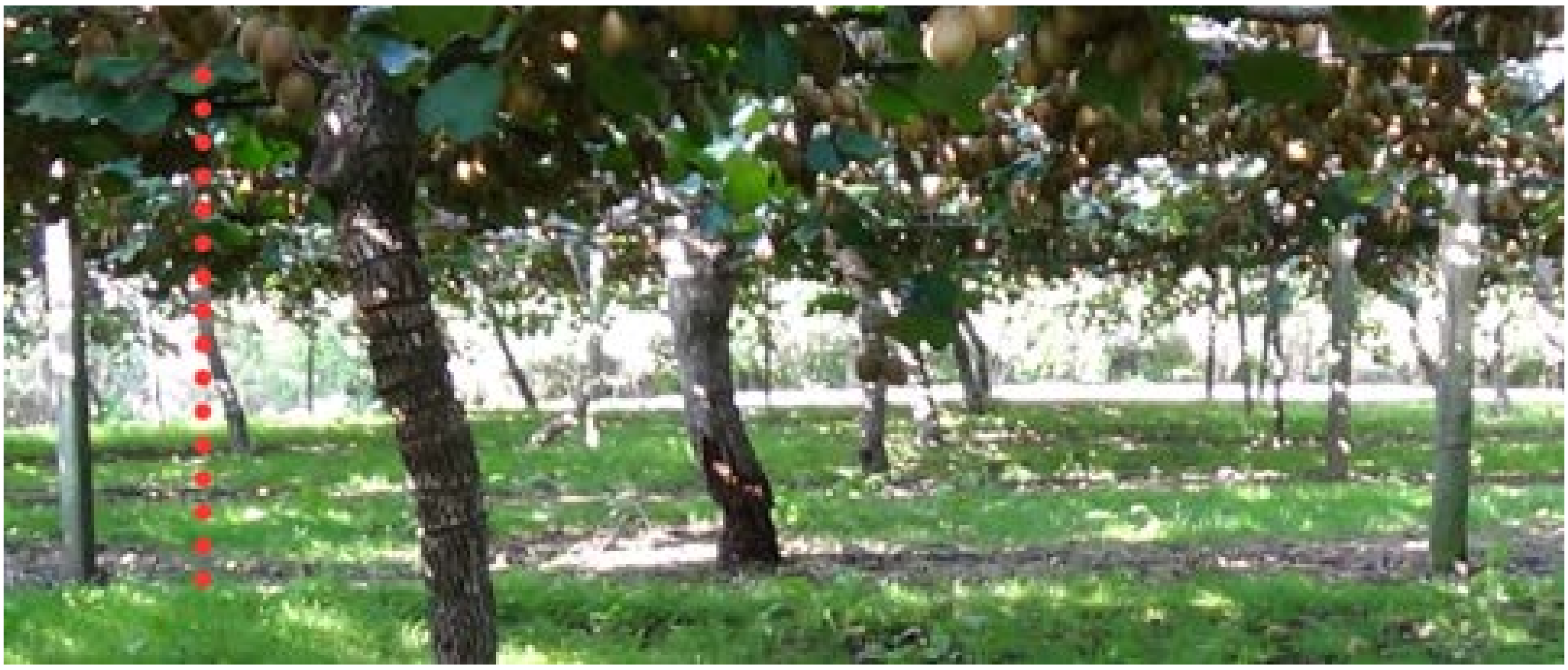

*Figure 126: Showing how a single azimuth of lidar measurements (red) may fall partly on a feature of interest like a trunk but also on non-structure defining features in behind and hence plane selection by maximum range for this azimuth would select data from a non-structure defining feature.*

### 4.5.5 Horizontal Lidar Density Feature Representation

It was hypothesized that in 3D lidar data, when viewing the data from above, the posts, trunks and other vertical objects would appear as areas of high densities of points. It was postulated that this property could be used to separate out the vertical features. As a result, the following algorithm was developed:

1. An image was initialised with pixel values of zero. The dimensions used were 1000 pixels by 1000 pixels. This image was defined to represent an area of 100 m by 100 m in a horizontal plane, so each pixel was to represent an area of 0.1 m by 0.1 m. The centre of the image was defined to be the axis of rotation of the 3D lidar.

2. For each lidar point, the horizontal coordinates of the point were calculated and the value of the corresponding pixel was incremented by a set value. An example of a result of this step is shown in Figure 127. In this image it seems that posts or trunks close by and the person standing close to the lidar are brighter, corresponding to higher density at such points.
3. A threshold was applied to the density image from step 2. An example of this step is shown in Figure 128, which is the result of processing Figure 127. The result shows that mostly posts, trunks and the person remain. However, only points very close to the lidar were retained.

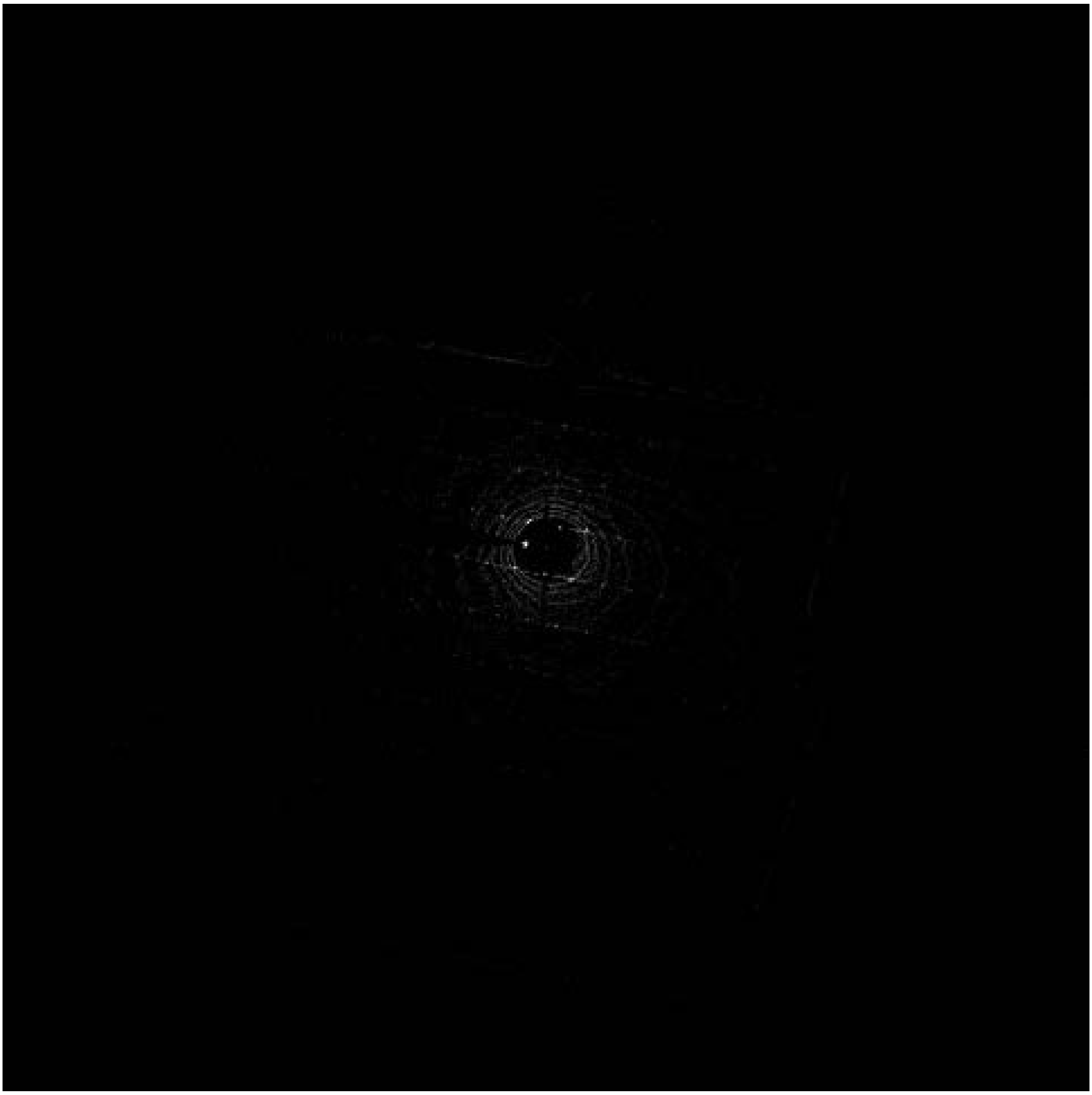

*Figure 127: Bird's eye view of 3D lidar data in a kiwifruit orchard, where each pixel value represents the number of lidar points at that location.*

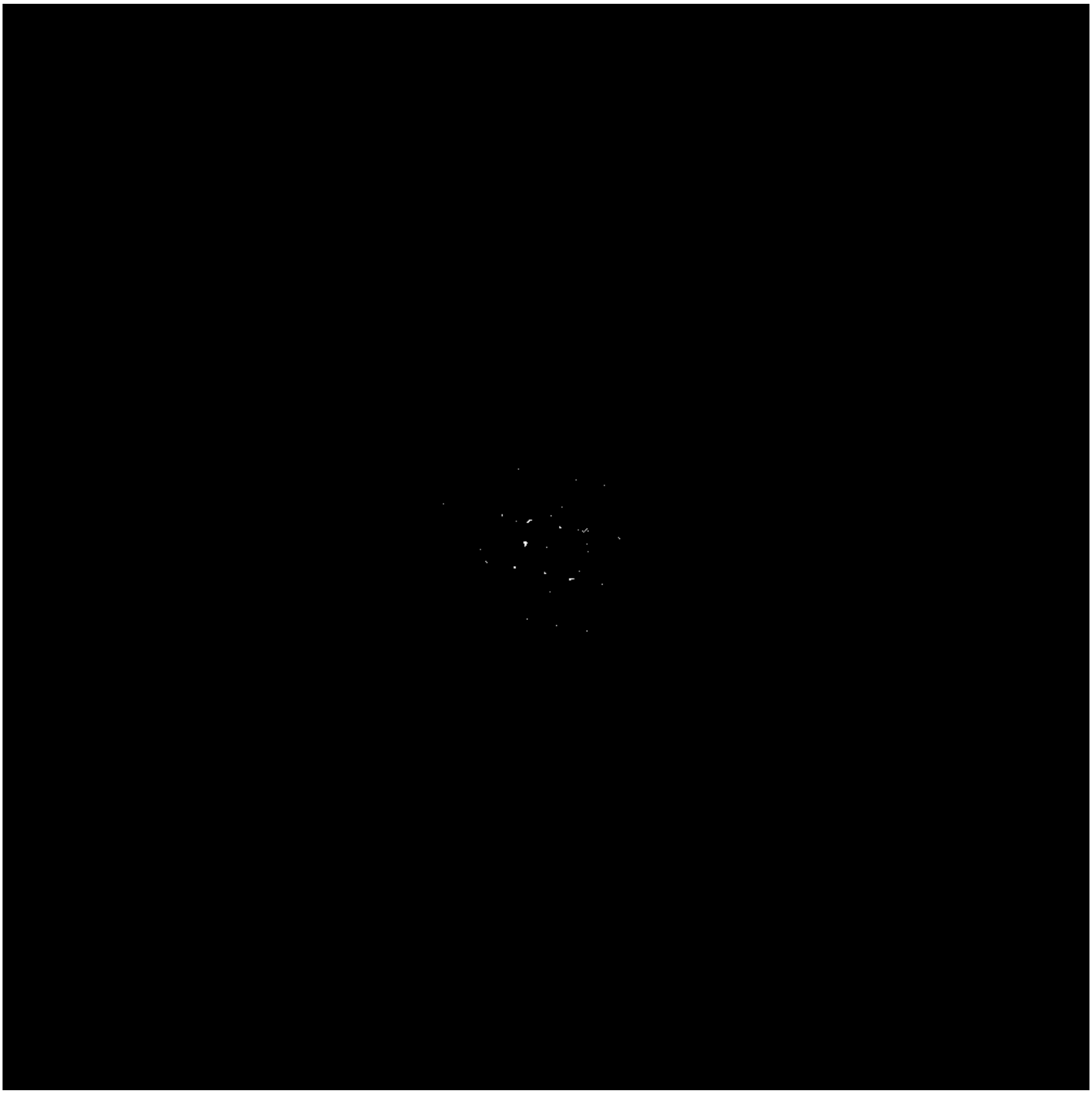

*Figure 128: The result of applying a threshold to the density image in Figure 127.*

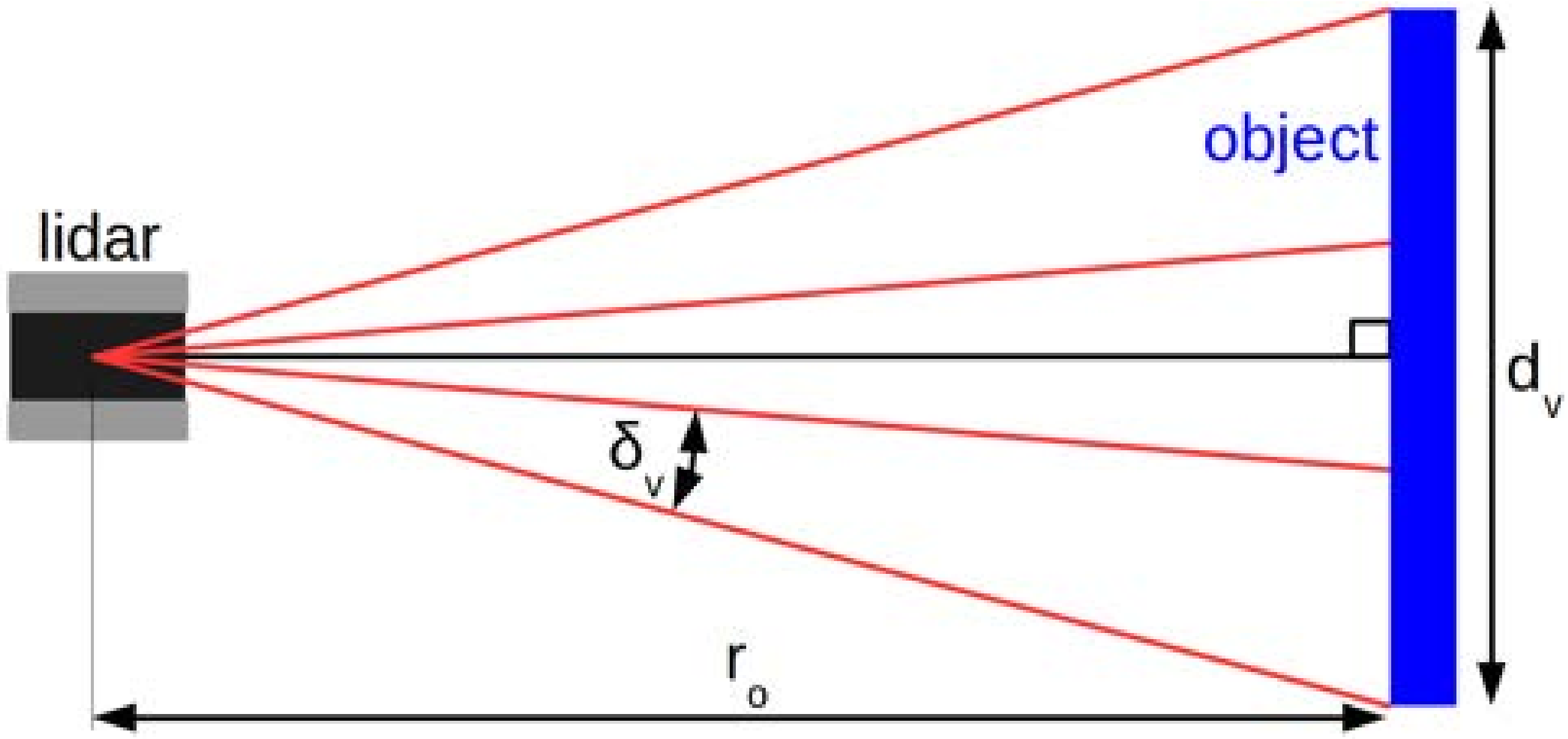

*Figure 129: A diagram of a lidar shown from side on, with some of the lidar rays intersecting an object.*

What the outputs in Figure 127 and Figure 128 do not take into account is the change in point density with range. Consider the diagram in Figure 129, which shows a 3D lidar from side on with a vertical angular resolution of $\delta_v$. Also in this diagram, there is an object of height $d_v$, which a number of lidar planes, $n_v$, intersect at a range of $r_o$. By considering a right angle triangle on half of the object, one may find:

$$\tan\left(\frac{(n_v-1)\delta_v}{2}\right)=\frac{d_v}{2r_o} \tag{11}$$

The number of lidar planes on the object, $n_v$, is related to the density of points on the object. Making $n_v$ the subject gives:

$$n_v=1+\frac{2}{\delta_v}atan\left(\frac{d_v}{2r_o}\right) \tag{12}$$

Taking into account the fact that the number of vertical planes of the lidar used was limited, gives:

$$n_v=min\left(argmax(n_v),1+\frac{2}{\delta_v}atan\left(\frac{d_v}{2r_o}\right)\right) \tag{13}$$

In a similar way to considering the scene from side on, the scene may also be viewed from above. With an object of width $d_h$ and a lidar of horizontal resolution $\delta_h$, the number of lidar points on the object, $n_h$, may be calculated according to:

$$n_h=1+\frac{2}{\delta_h}atan\left(\frac{d_h}{2r_o}\right) \tag{14}$$

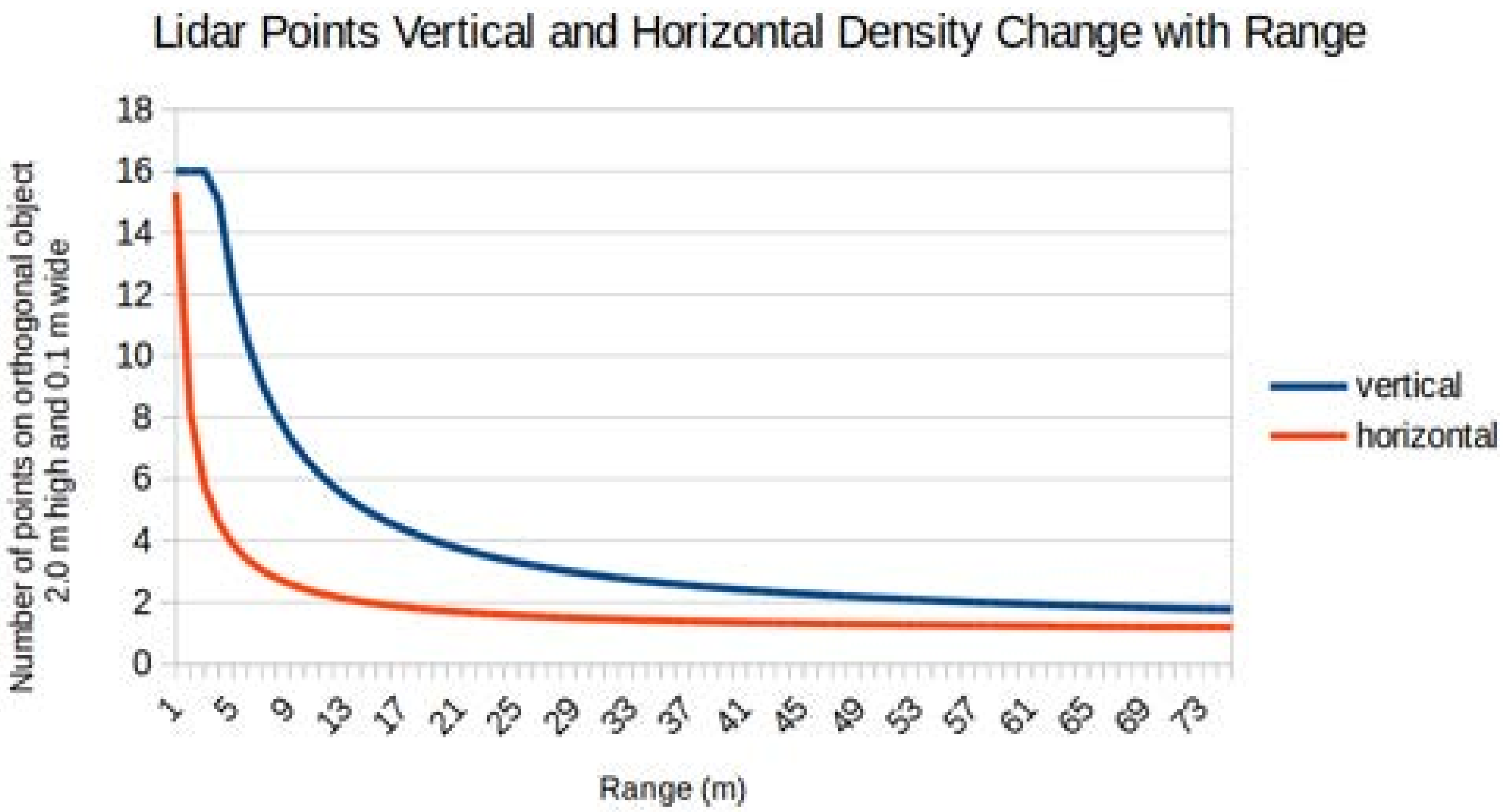

*Figure 130: The number of points on an object orthogonal to the lidar with size 2.0 m high and 0.1 m wide, when viewed from side on and above, for different ranges.*

The number of points on an object of a set size, when viewed from side on and from above, for different distances is shown in Figure 130. This graph shows that the number of points on an object reduces steeply at the shorter ranges. If the object was a rectangle, directly facing the lidar, the total number of lidar points on the object, $n_{3D}$, could be estimated by:

$$n_{3D} = n_v n_h \qquad (15)$$

A scaling factor, $s_d$, was calculated using the maximum value of a pixel, $P_{max}$, according to:

$$s_d = \frac{p_{max}}{n_{3D}} \qquad (16)$$

This scaling factor may be drawn in an image as shown in Figure 131. This image is the same dimensions as the density image in Figure 127. Multiplying the scaling factor image by the density image gives the result in Figure 132, which has clearer data at points further from the centre, when compared to the original density image in Figure 127. Applying a threshold to the scaled density image of Figure 132, gives the result in Figure 133, which retains more points than the unscaled equivalent in Figure 128.

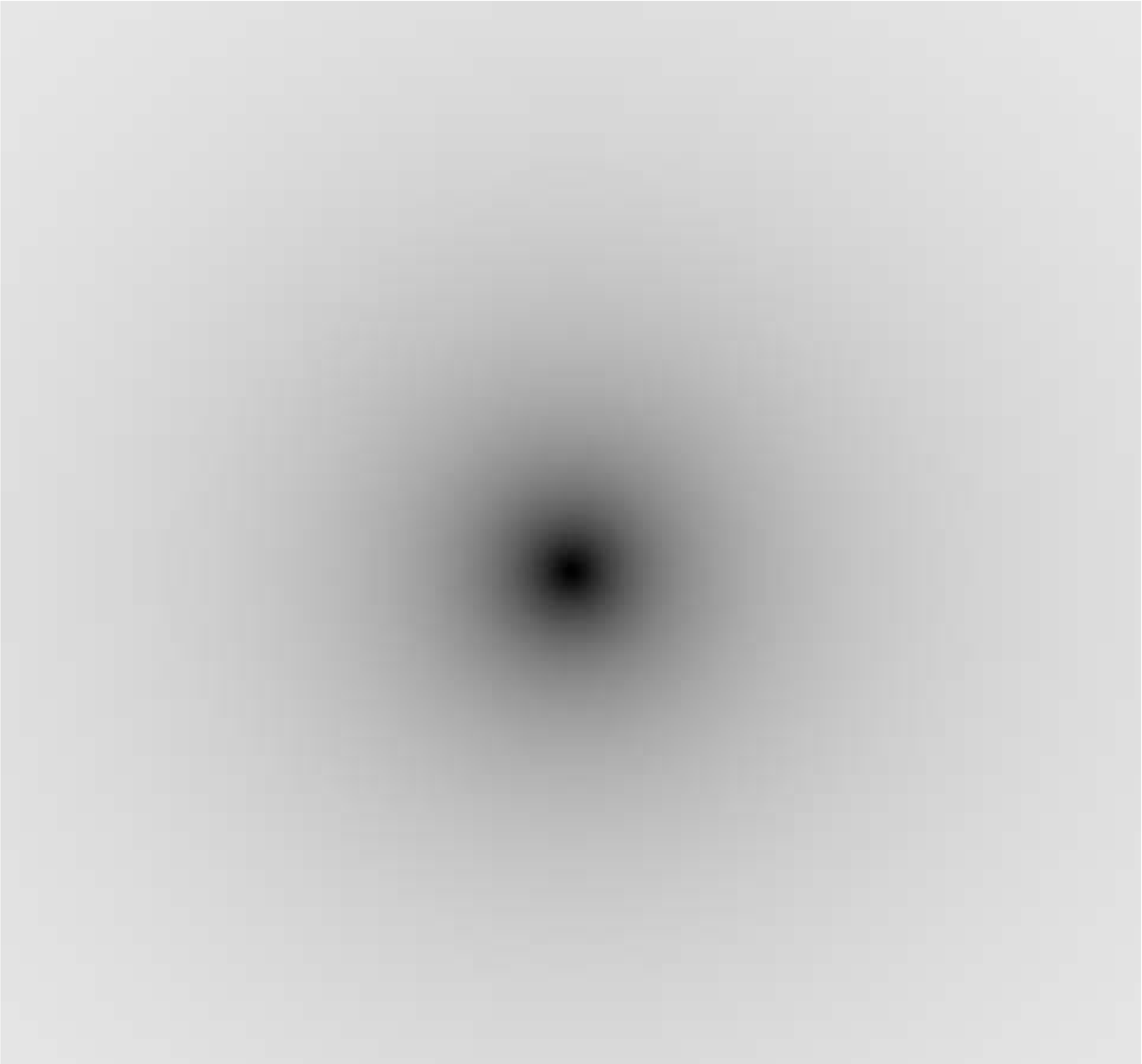

*Figure 131: A scaling factor for 3D lidar data to take into account the changes in point density with range, for a lidar with the centre of rotation at the centre of the image.*

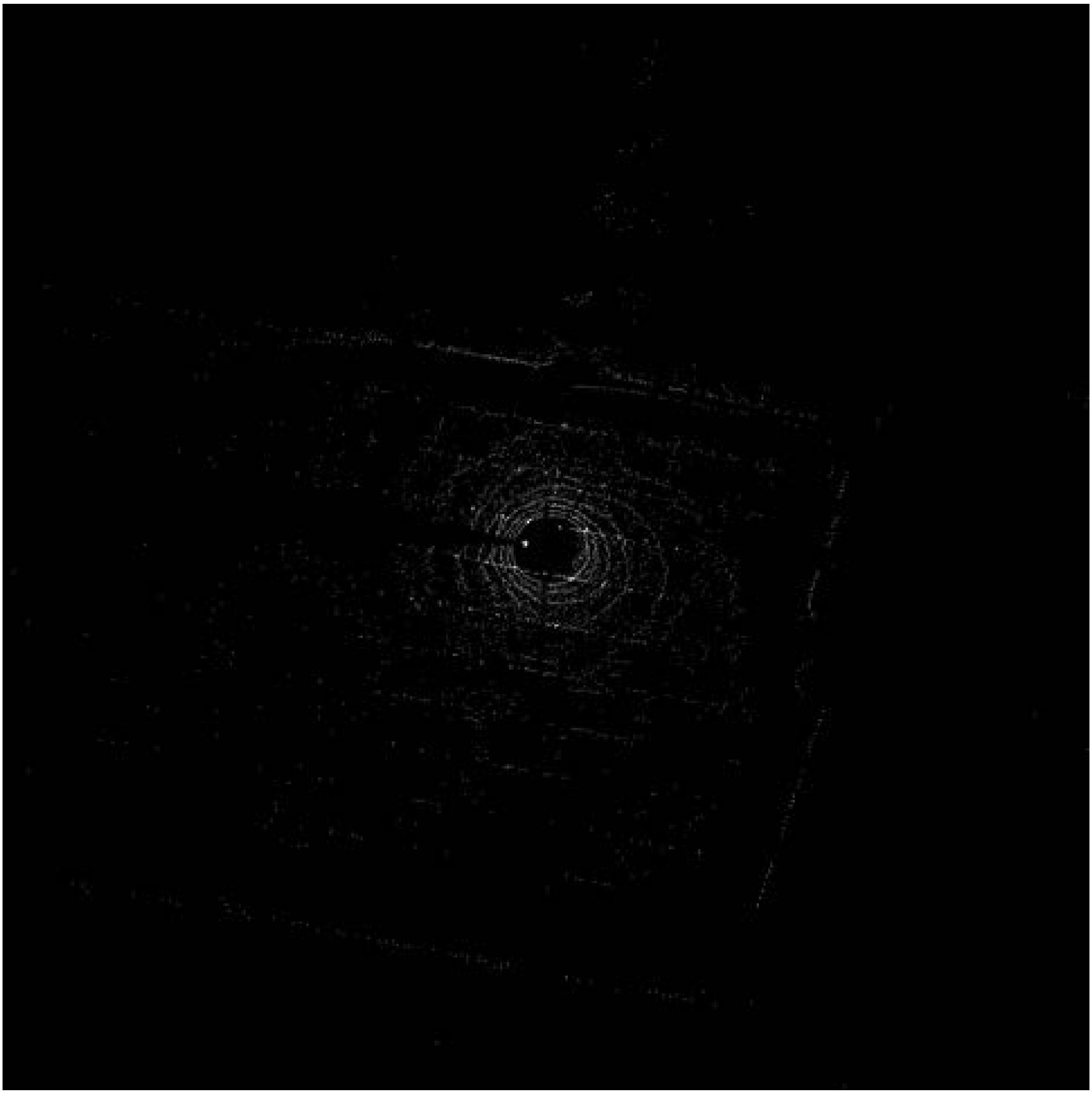

*Figure 132: An image of lidar point density, multiplied by a scaling factor to take into account the change in point density with distance from the lidar.*

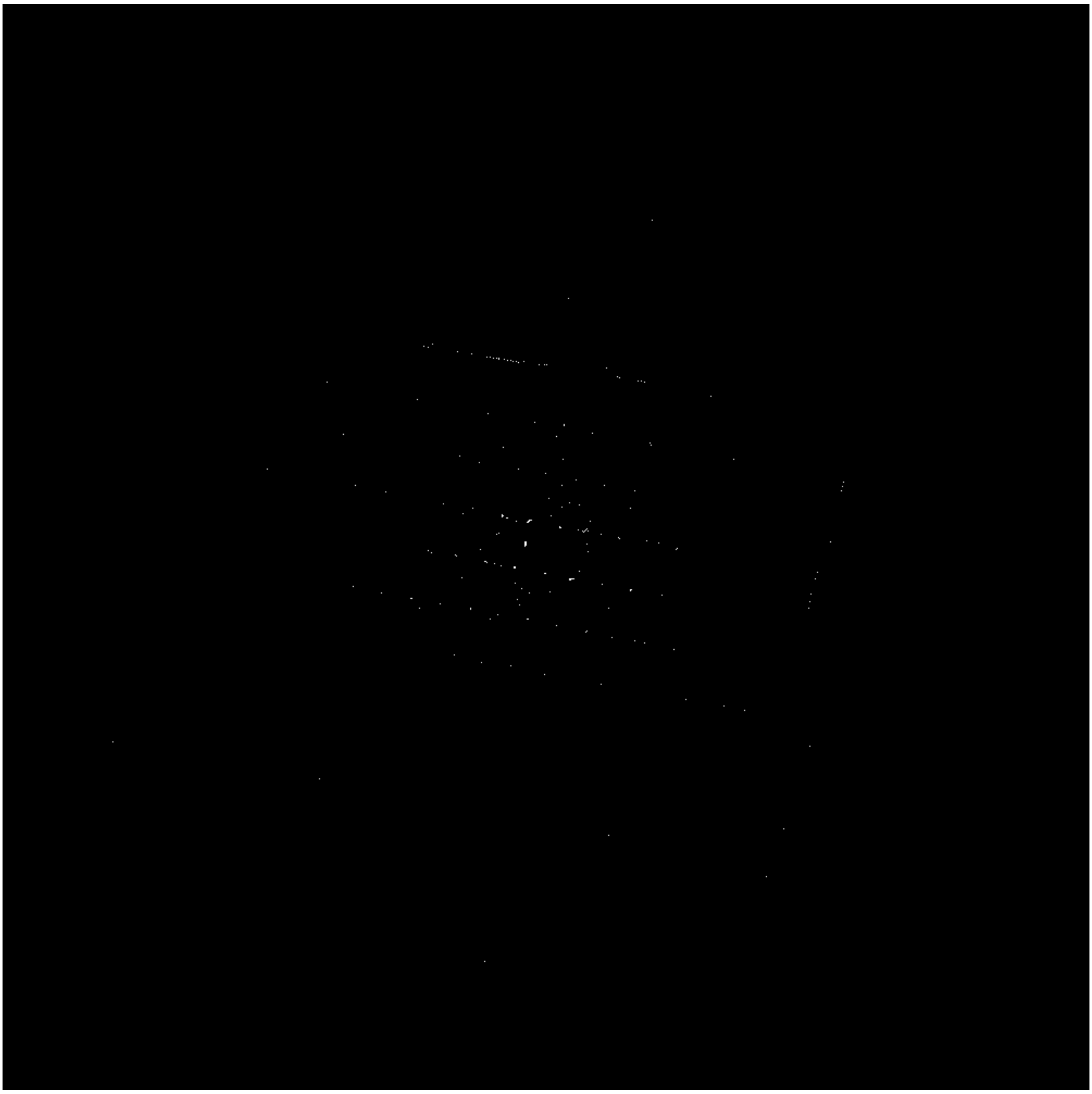

*Figure 133: The range scaled lidar density image after thresholding.*

### 4.5.6 Vertical Connected Objects in Compact Image Representation

It was hypothesized that in 3D lidar data, vertical objects of interest- such as posts, trunks and hedges- would have angles close to vertical between vertically connected points. A method for extracting vertical objects was developed based on this hypothesis. This method began by organising the 3D lidar data into rows and columns, which can be visualised as an image, as demonstrated in Figure 134. In this image, each pixel represents the range of the point, with closer objects shown as being brighter. The actual resolution of the data was 900 columns horizontally and 16 rows vertically, so the image in Figure 134 has been stretched vertically, just for visualisation.

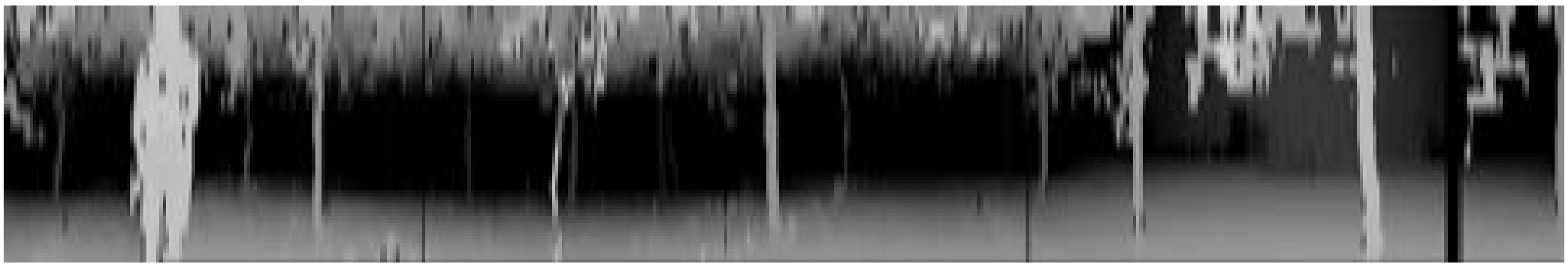

*Figure 134: 3D lidar data from a kiwifruit orchard, which has been organised into rows and columns, with each pixel representing the range of the point.*

It seemed looking at Figure 134 that vertical objects define the structure in kiwifruit orchards for this type of data. To detect vertical structures, the angle between vertically adjacent points was calculated. Figure 135 shows the variables used to calculate the angle between two vertically connected points on an object, $\alpha_{12}$; these variables include the ranges to the points, $r_1$ and $r_2$, as well as the vertical angles to the points, $\alpha_1$ and $\alpha_2$. Using these variables, the angle between adjacent vertical points was calculated according to:

$$\alpha_{12} = atan\left(\frac{r_2 \sin\alpha_2 - r_1 \sin\alpha_1}{r_1 \cos\alpha_1 - r_2 \cos\alpha_2}\right) \qquad (17)$$

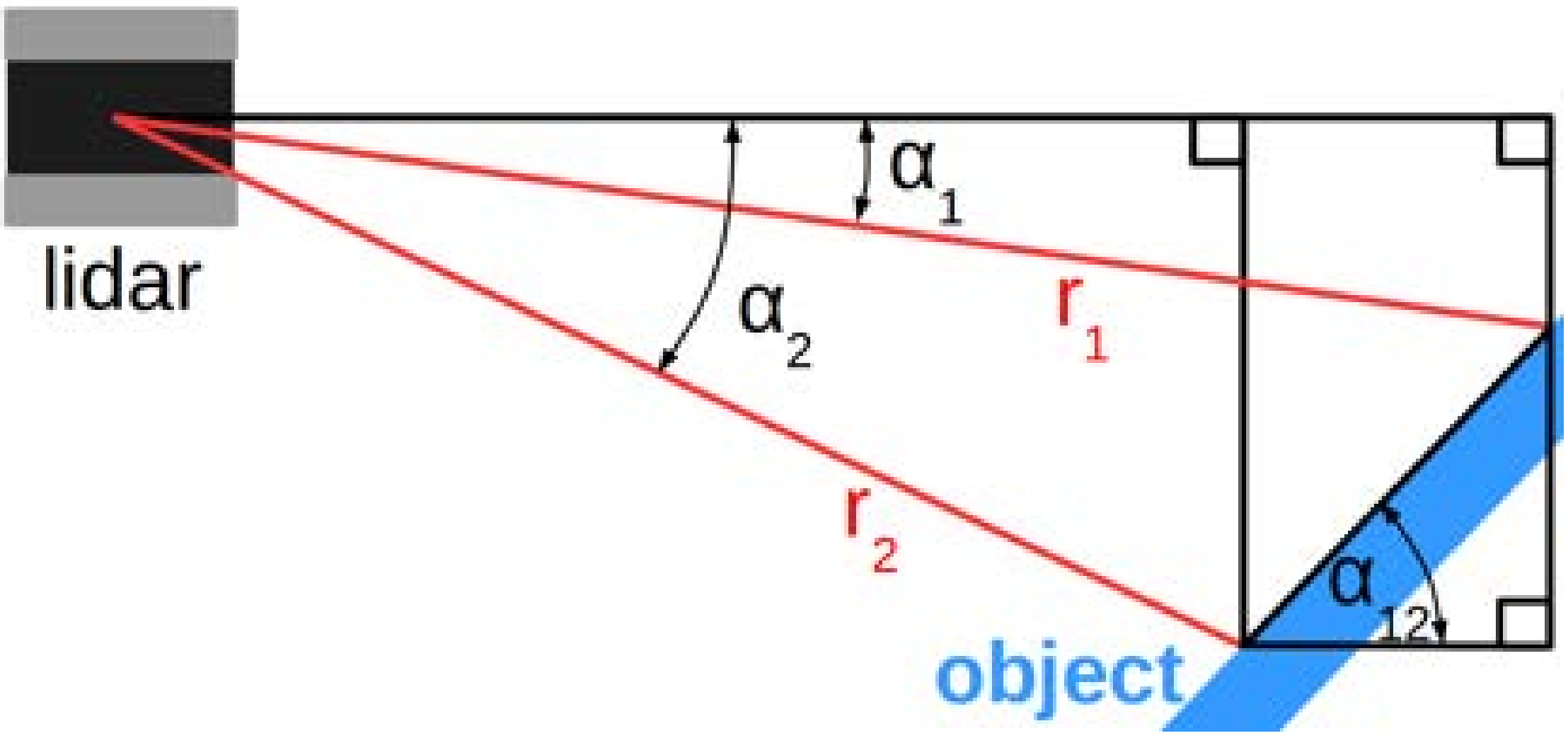

*Figure 135: Showing a 3D lidar from side on with two of its rays intersecting an object, with the angle between the two points from the rays on the object, $\alpha_{12}$.*

To show how the angle between vertically adjacent points was used to segment vertical objects, consider the small patch of points shown in Figure 136. The angle between the top left point (0,0) and the points below it are calculated. For this angle calculation, 0° is horizontal and 90° is vertical. If the angle threshold used for vertical objects was 45°, then points (0,0) and (1,1) would be considered part of a vertical object. The algorithm would then check if any point in the vertical object already had a class; if no point already had a class, a new class would be created and this classification is represented by the blue fill for these points.

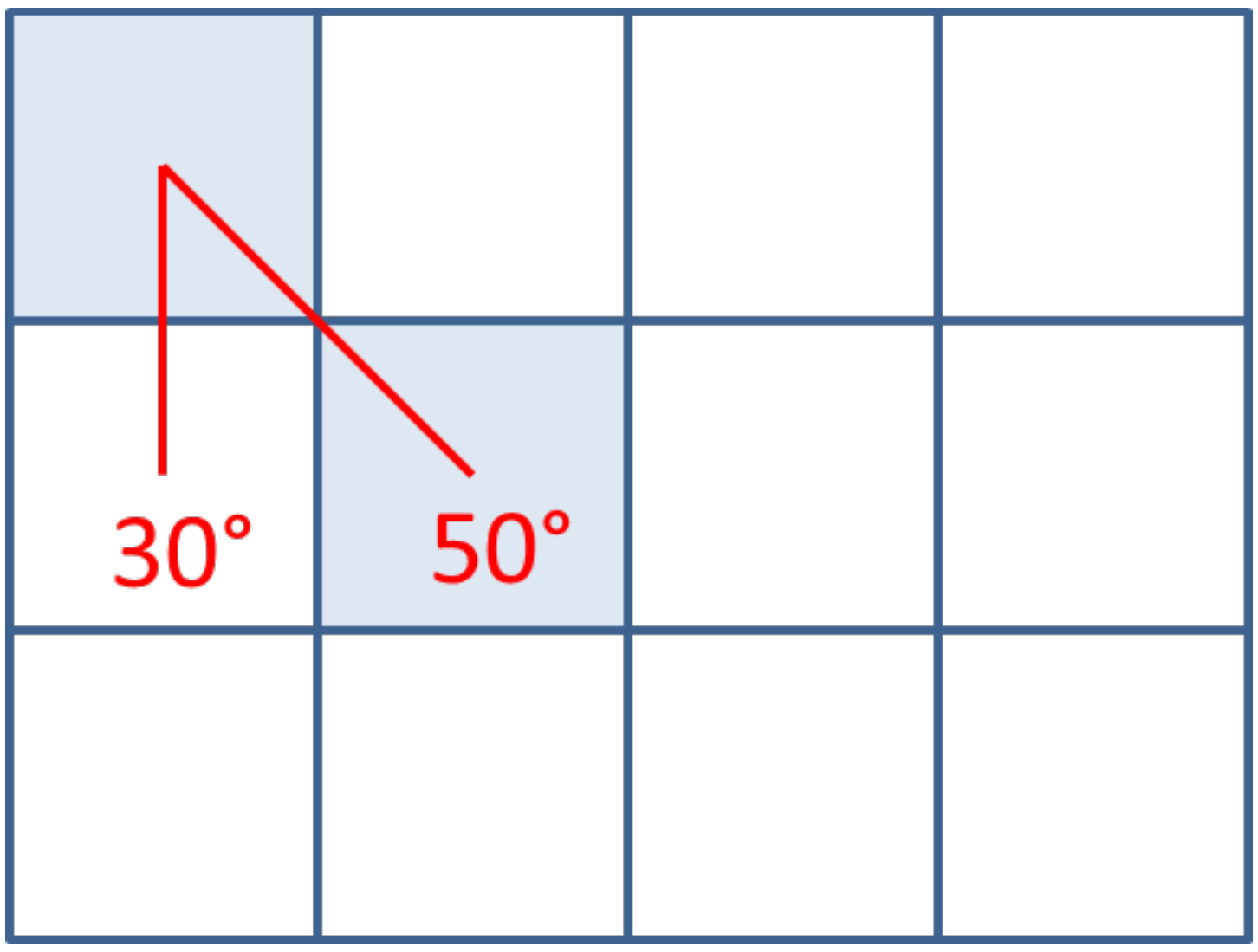

*Figure 136: Angles between a top left point and points below.*

Continuing on from Figure 136 at the next algorithm step, point (1,0) is considered as shown in Figure 137. Points (1,0), (1,1) and (2,1) are found to be on the same vertical object. Because (1,1) already has a class, (1,0) and (2,1) would be given the same class.

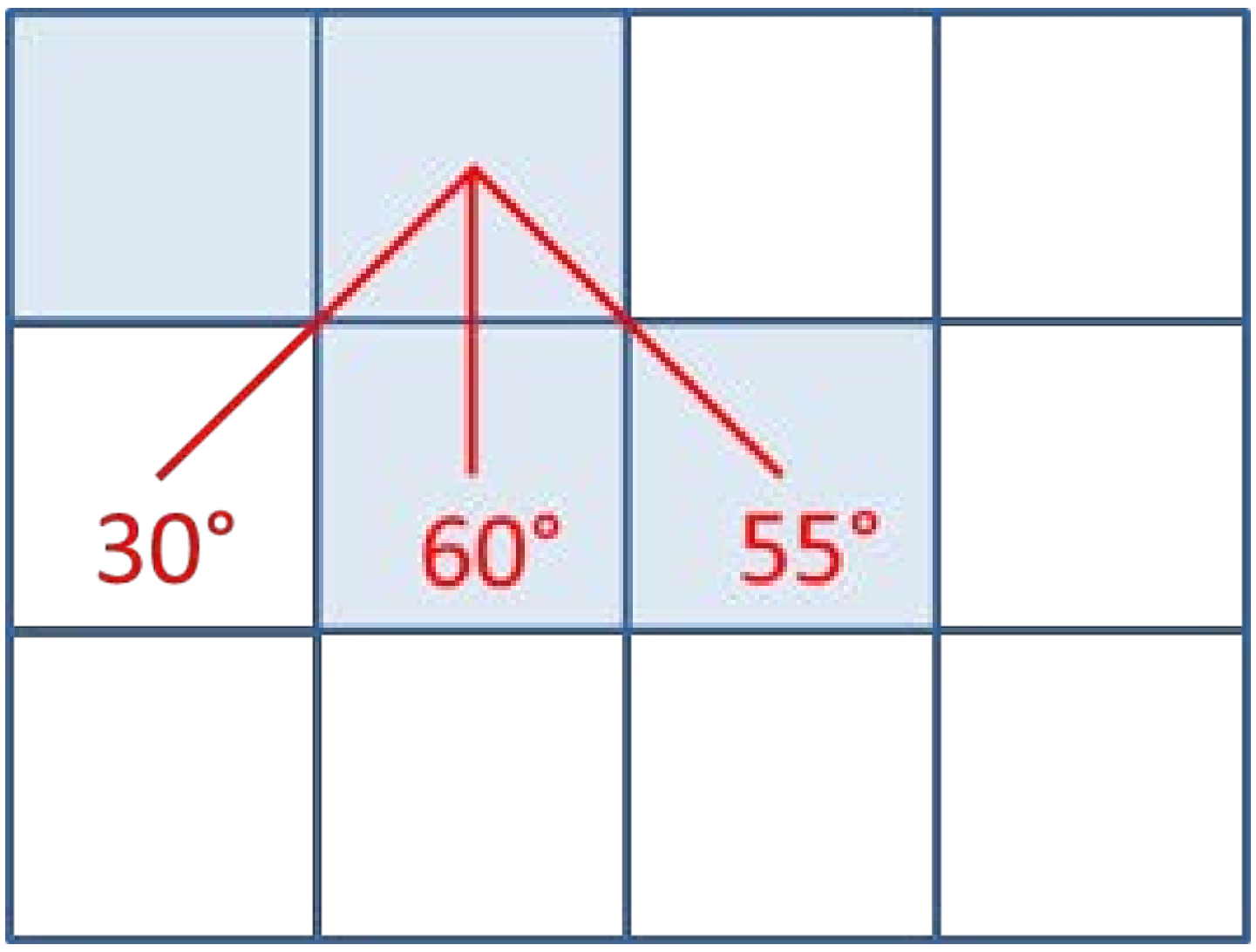

*Figure 137: Angles calculated for a second algorithm step of the vertical connected objects method.*

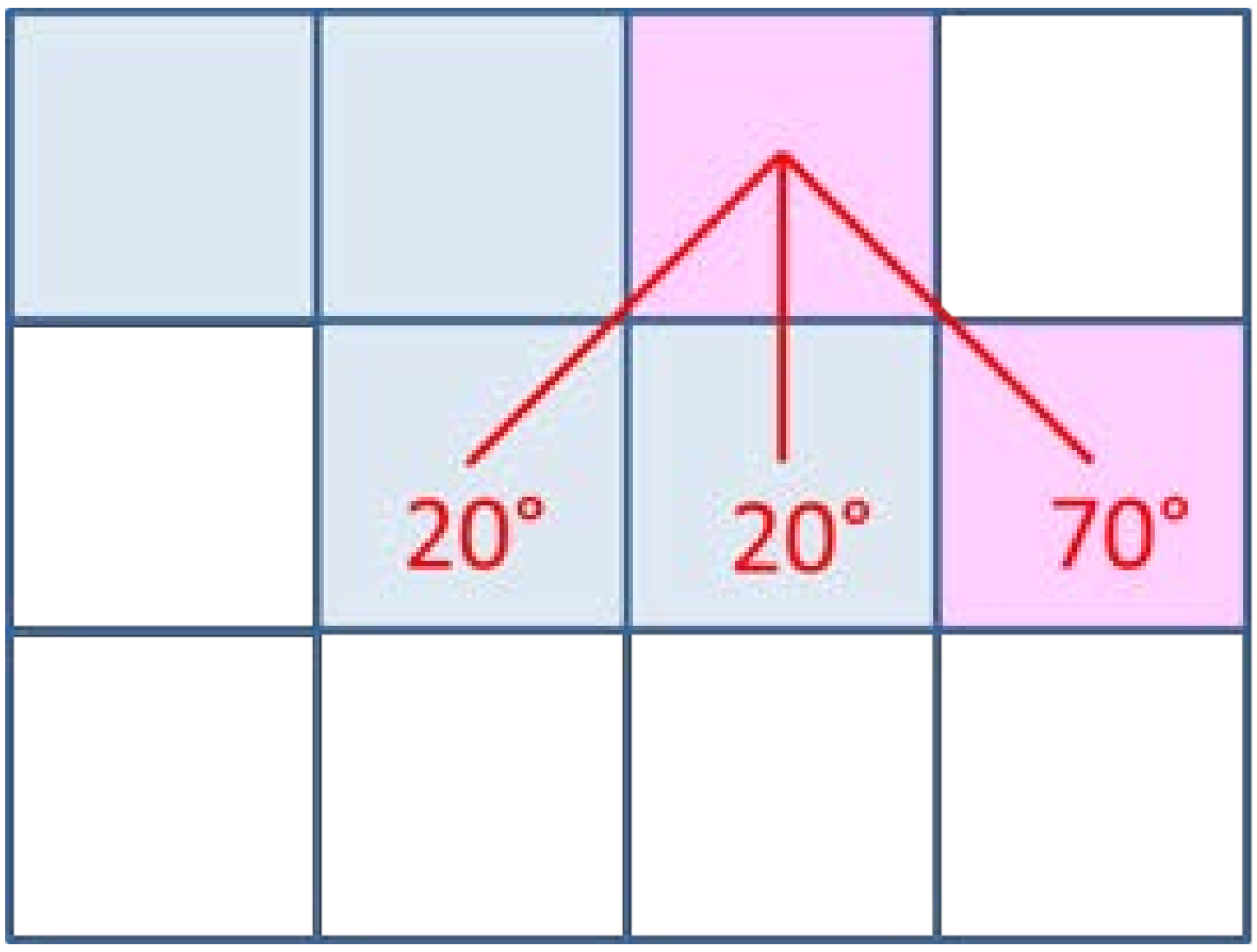

*Figure 138: Angles calculated for a third algorithm step of the vertical connected objects method.*

Continuing on from Figure 137, at the next algorithm step, point (2,0) is considered as shown in Figure 138. Here, points (2,0) and (3,1) are the only points found to be on the same vertical object but neither has a class already so a new class is assigned to those points. Continuing this procedure throughout 3D lidar data gives results like those shown in Figure 139, where objects that are more horizontal, like the ground, are removed. Note that much of the kiwifruit canopy above is retained because it contains many hanging branches, which are approximately vertical.

For each vertical object in Figure 139, the highest and lowest point is found to calculate the height of the object. Then, a height threshold is applied to each object to give the result shown in Figure 140. Inspection of Figure 140 shows that what are predominantly left behind are trees, posts and hedges, which represent the structure of the orchard. Most non-permanent features, like weeds and hanging branches, have been eliminated. An example of such an output plotted in a bird's eye view is shown in Figure 141. This bird's eye view output seems comparable to the density image and plane selection methods; however, which method produces the best output was unclear so a labelled dataset was created and tests were performed, comparing the different feature extraction methods.

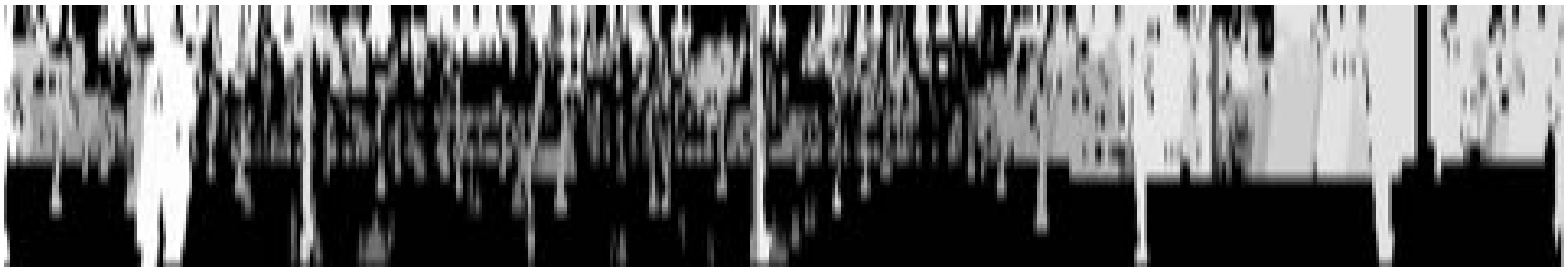

*Figure 139: 3D lidar data with a threshold applied to the angle between vertically connected points.*

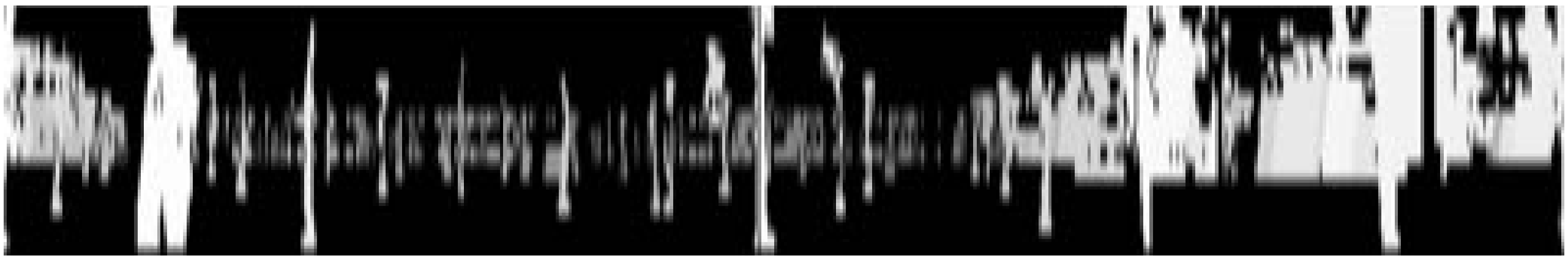
Figure 140: Objects remaining after a height threshold is applied to vertical connected points.

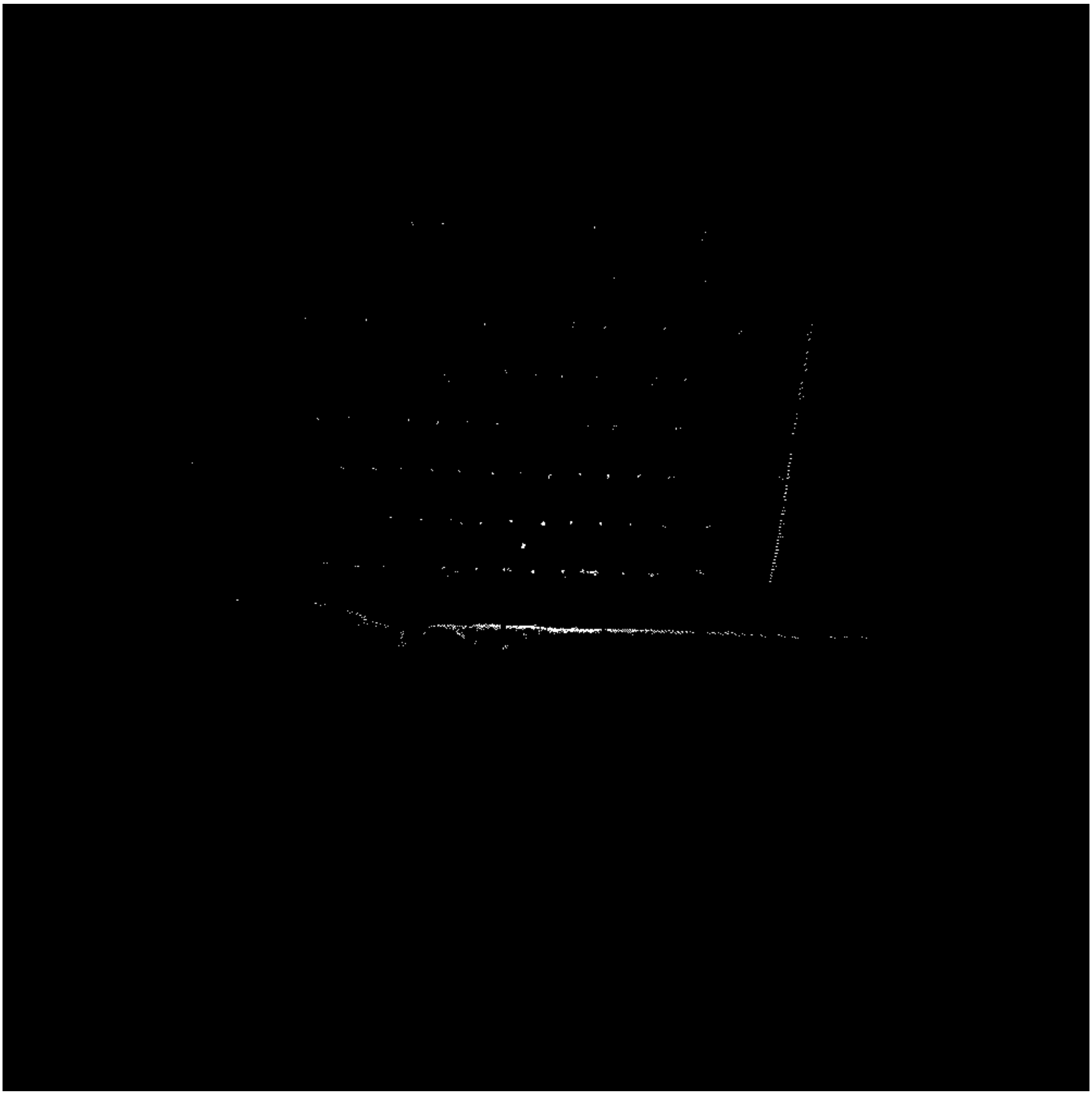
Figure 141: Bird's eye view of kiwifruit orchard lidar data, where angles between vertically connected points were used to segment objects and height thresholds were applied.

### 4.5.7 3D Lidar Feature Extraction Ground Truth

Three methods for extracting post, trunk and hedge features from 3D lidar data have been presented. In order to decide which method would be used for tasks, such as mapping and localisation, an appropriate metric had to be formulated. The metric had to reflect the need to ensure that as many structure defining features of the orchard were preserved, while non-structure defining features were filtered out. Structure defining features included posts and trunks but also, to a lesser extent, hedgerows.

Hedgerows are substantial objects that surround the orchard block, which could make them useful features for navigation. However, even though hedgerows are fixed in place, they change in shape due to shoot growth and maintenance to control the growth. Hence, hedgerows change more rapidly with time compared to trunks and posts, so this should be considered when using hedgerows in navigation algorithms.

Non-structure defining features are considered to be those that change markedly from season to season. The non-structure defining features that are of particular concern are those that are likely to remain fixed for long enough to be mapped. These include weeds and hanging branches; eliminating these objects from an orchard completely represents a significant amount of labour. Other non-structure defining features can include people, animals, and vehicles; however, it was anticipated that the presence of such objects may be easily controlled without too much labour.

As introduced before, the metric used to compare the different algorithms for feature extraction had to retain structure defining features of the orchard and filter out non-structure defining features. The metric used was the counts of structure defining and non-structure defining features in the output. However, there were difficulties in assessing this metric for the different algorithms, including:

- The different algorithms used different representations for processing the 3D lidar data, including point-clouds, image bird’s eye views and image horizontal views. To overcome these differences, all outputs were converted to a single representation, which was the bird’s eye view images.
- There was no ground truth input stream. In order to produce a ground truth, hand labelling was used.
- A single frame of the lidar data contained thousands of points so producing counts by hand would have been time-consuming.

In order to reduce the time labelling the data, a tool was created for selecting points in the lidar data, corresponding to the structure defining features. The selected points were automatically counted and used to create a mask so that the non-selected points could also be counted automatically.

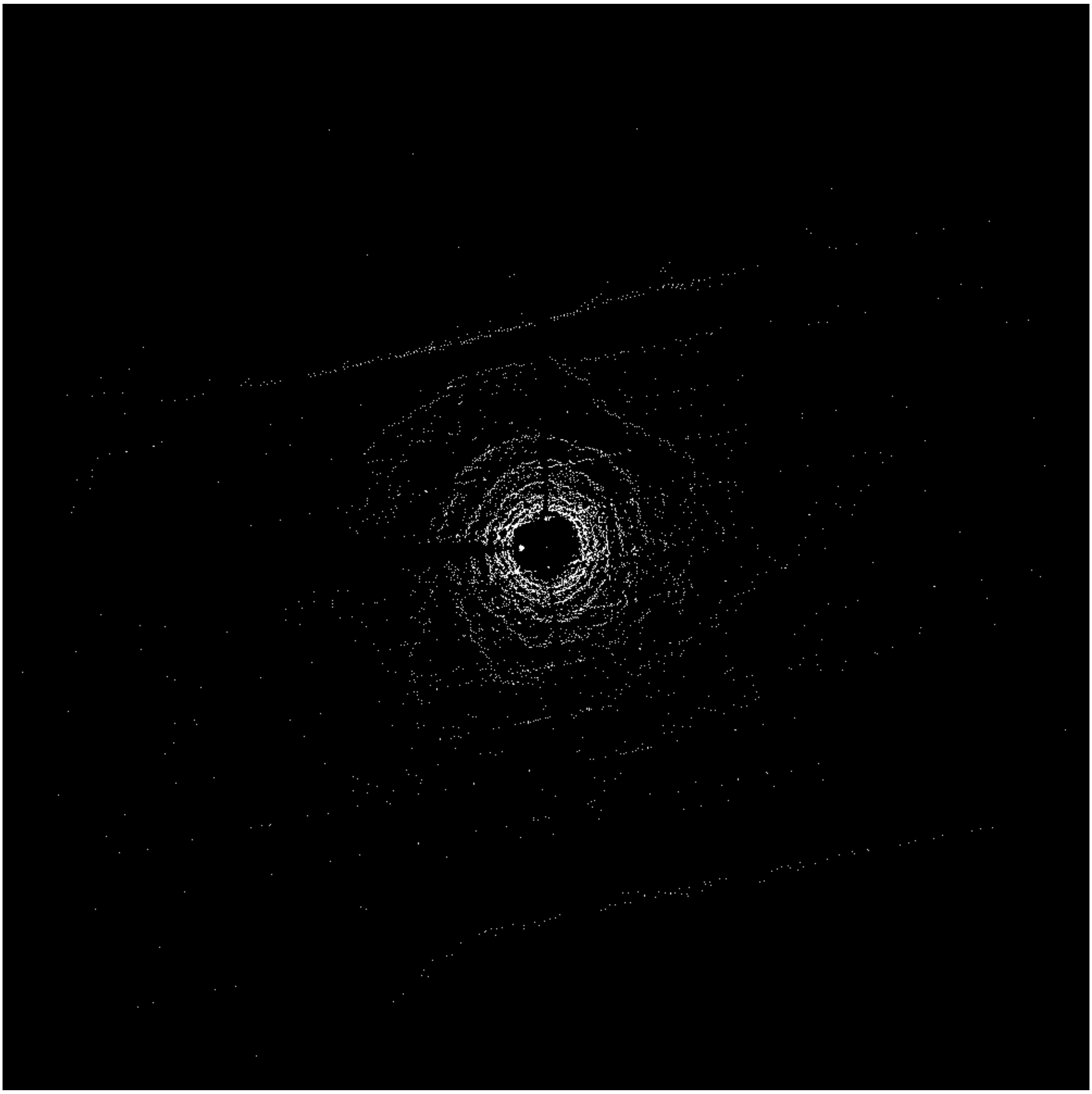

*Figure 142: An example of a frame of data from the testing dataset of lidar processing for navigation in kiwifruit orchards, where the lidar data has been plotted in a bird's eye view.*

The tool for labelling the lidar data saved considerable time and so the approach is contributed here. For each frame of lidar data, the labeller had access to all planes of the data. The way this was achieved was only one of the middle planes was initially displayed but the labeller was able to continuously cycle through the planes to label points of interest in each plane, with the results being aggregated into a single set of labels. This approach of viewing one plane at a time was adopted after first trialling a generic point cloud labelling tool that displayed the data in 3D [196], which was found to be approximately 3 times slower to label with. Also, viewing all of the data at once in a 2D bird’s eye view was tried; however, this caused the labeller to miss many features. As can be seen in Figure 142, when all of the planes are visible at once, it is difficult to make out the posts and trunks in the region around the lidar centre in a 2D bird’s eye view. Hence, the approach of viewing one plane at a time was adopted during labelling.

Time savings were made by exploiting the parallel lines of the pergola structure so that multiple posts and trunks could be extracted as the intersection between the treelines and a transverse direction in the pergola. In addition, three datasets were created from the labelling procedure, including:

- The structured feature extraction dataset- for testing the methods that have been presented in Subsections 4.5.3 to 4.5.6.
- A dataset where just the posts and trunks were the positive labels.
- A dataset that was used for row detection ground truth, where the linear and angular offset of the current row from the lidar coordinate frame was the output.

The test dataset was 100 frames of lidar data, which were collected from a Velodyne VLP-16 3D lidar [130], which was mounted on a robot with the axis of rotation vertical and at a height of 0.8 of a metre from the ground. The robot was driven through the rows of an orchard block by manual remote control and the driving was deliberately erratic, with extreme swerving, in order to allow for worst case driving. After driving through the orchard block, there were over 20,000 frames of lidar data. This was reduced to 100 frames by sampling every 200$^{th}$ frame of data. These 100 frames was stored as Point Cloud Data (PCD) files; however, for labelling, the data was converted to bird’s eye view image files, representing an area of 100 m by 100 m in a resolution of 1000 pixels by 1000 pixels, with one image per lidar plane of data.

Multiple stages of labelling were used in order to create the multiple datasets. The six stages of labelling used were as follows:

1. The first stage of labelling marked the two treelines on either side of the current row by mouse clicking on two points in one treeline and then doing the same for the other treeline. An example of labelled data for this first stage is given in Figure 143. This first stage of labelling served multiple purposes. From the two treelines labelled, the angles of all of the treelines were calculated, under the assumption that the treelines were parallel. The angle of the treelines was then used in the second stage of labelling. In addition, labelling the current row treelines also provided ground truth for the row detection methods. Note that the full treeline is not drawn in Figure 143, because the treeline would have been labelled based on a limited number of planes of the data and this was sufficient to get the pose of the treeline.

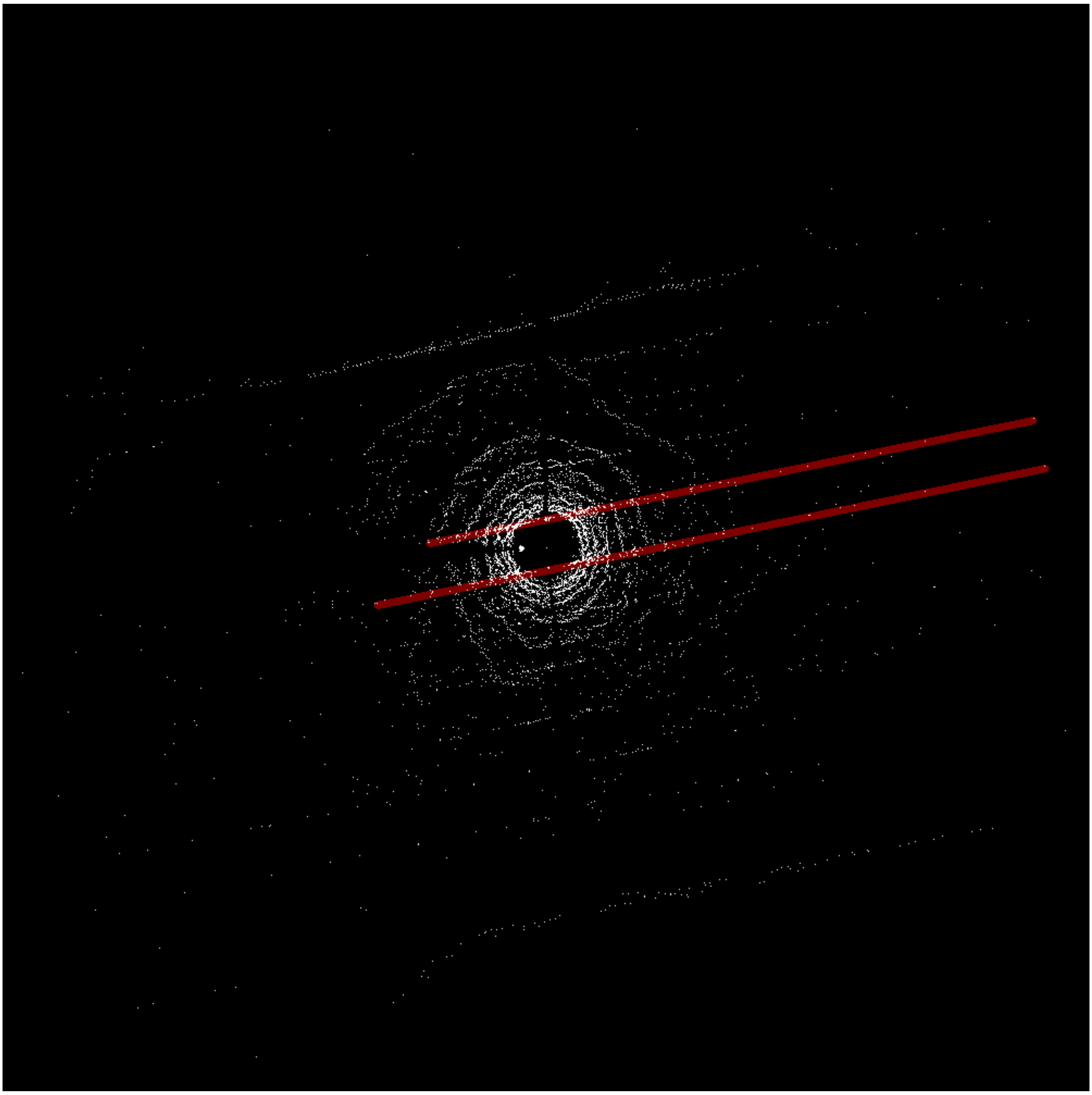

*Figure 143: The first stage of hand labelling for lidar feature extraction ground truth datasets, where the treelines of the current row are drawn with four point clicks.*

2. In the second stage of labelling, all of the visible treelines were marked by a single mouse click per treeline on a point in each treeline. The treeline angle from the first stage of labelling was used to draw the treeline based on the single point clicked per treeline (Figure 144); hence there was time saved in labelling the treelines.

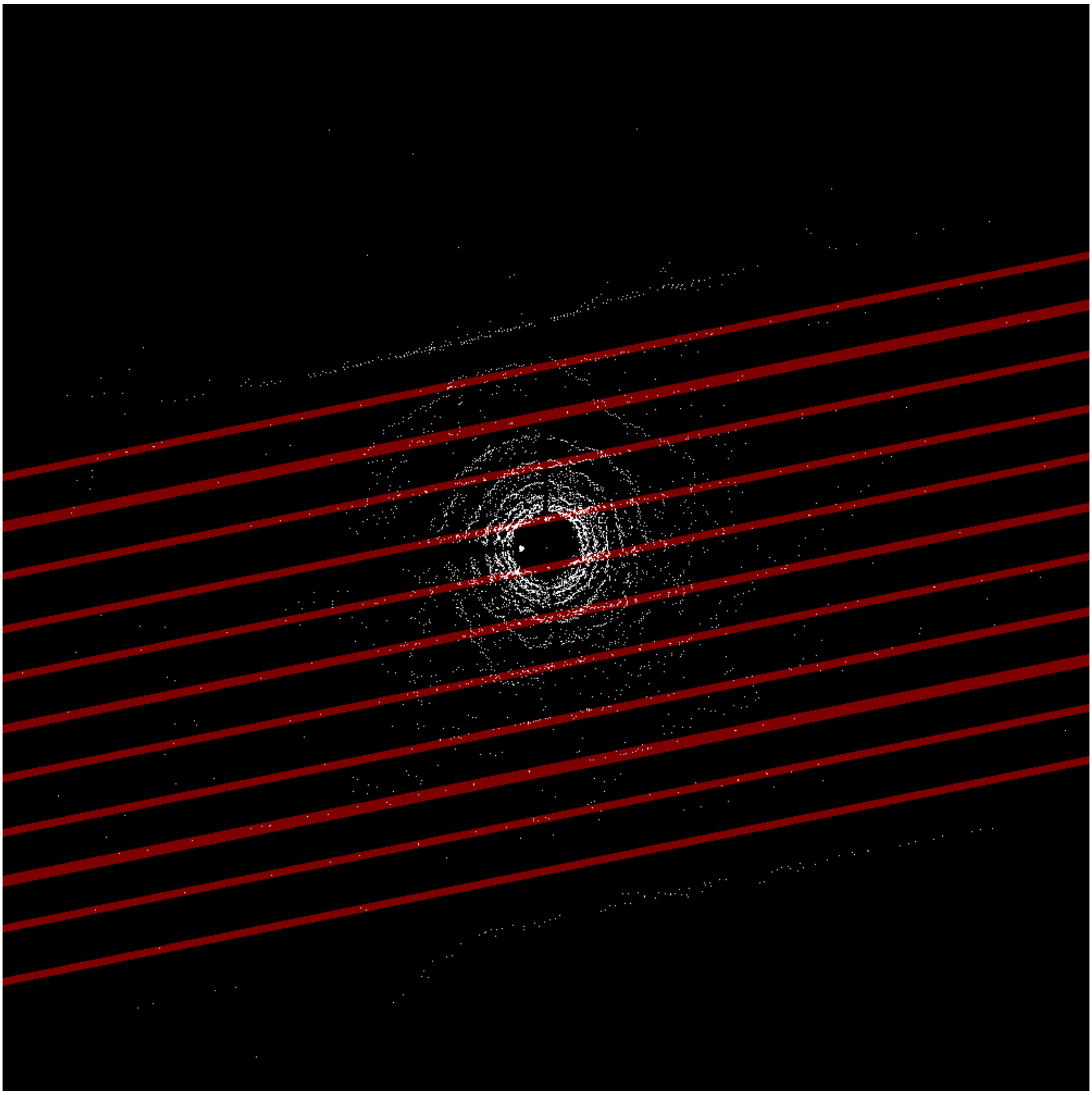

*Figure 144: Labelling multiple treelines in lidar data using the angle of the already labelled current row and a single mouse click per treeline.*

3. In the third stage of labelling, a transverse direction formed by the posts and trunks was selected by mouse clicking on two points. In a similar way to the labelling of the current treeline, the transverse direction angle was calculated and then each subsequent single mouse click on posts and trunks was used to draw lines for the transverse direction (Figure 145). The intersections of the treelines and the transverse direction lines were supposed to be the positions of the posts and trunks in the kiwifruit pergola.

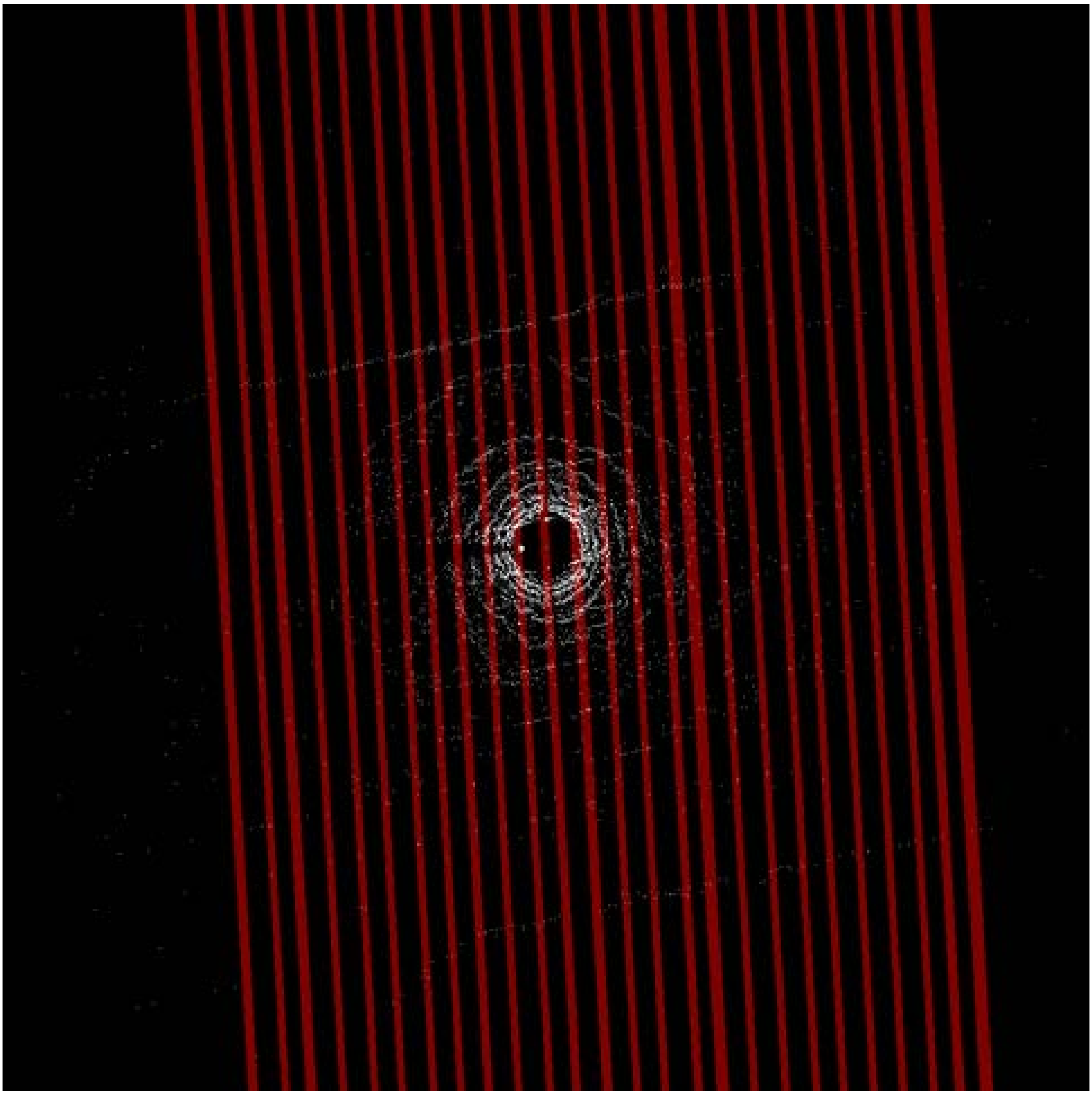

*Figure 145: Labelling parallel lines of trees and posts, where the parallel lines are in a transverse direction to the rows of the orchard.*

4. In the fourth stage of labelling, the points corresponding to people were marked with mouse point clicks.
5. In the fifth stage of labelling, the data corresponding to hedges and other vertical objects on the border of an orchard block (including water tanks) were marked with two mouse clicks per line (Figure 146). The people, hedges and other objects were labelled separately so that a dataset could be created for specific post and trunk segmentation testing.

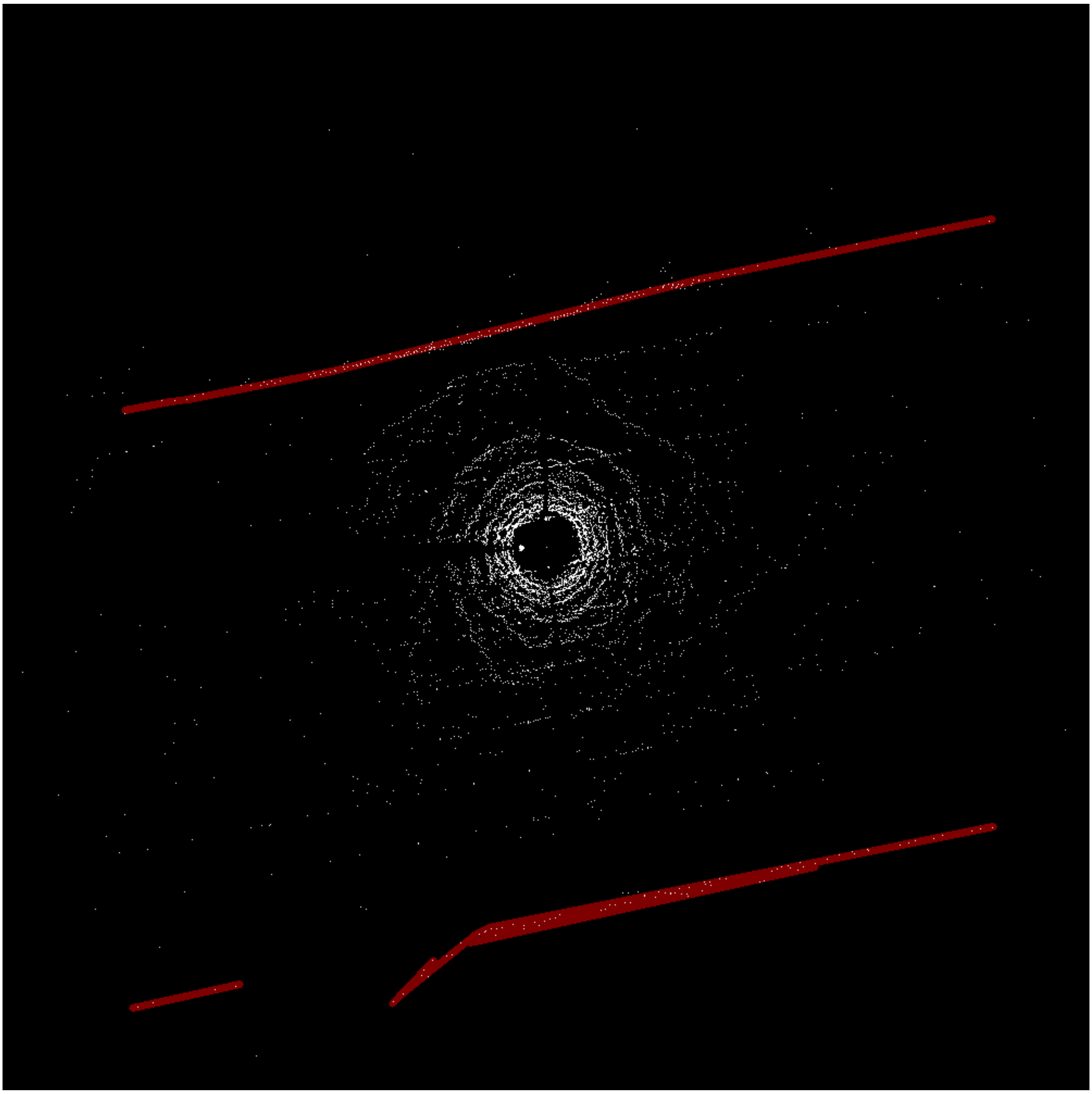

*Figure 146: Orchard block boundary objects labelled with straight lines.*

6. In the final stage of labelling, the labeller cycled through the different planes and decided which planes were contributing unique data in the bird's eye view and which planes were redundant. This action was performed so that canopy and ground data could be better removed from the final datasets. After this stage of labelling, outputs including all of the structure defining features and separate outputs with just the posts and trunks alone were extracted by masking the labels with the selected layers of the lidar data (Figure 147).

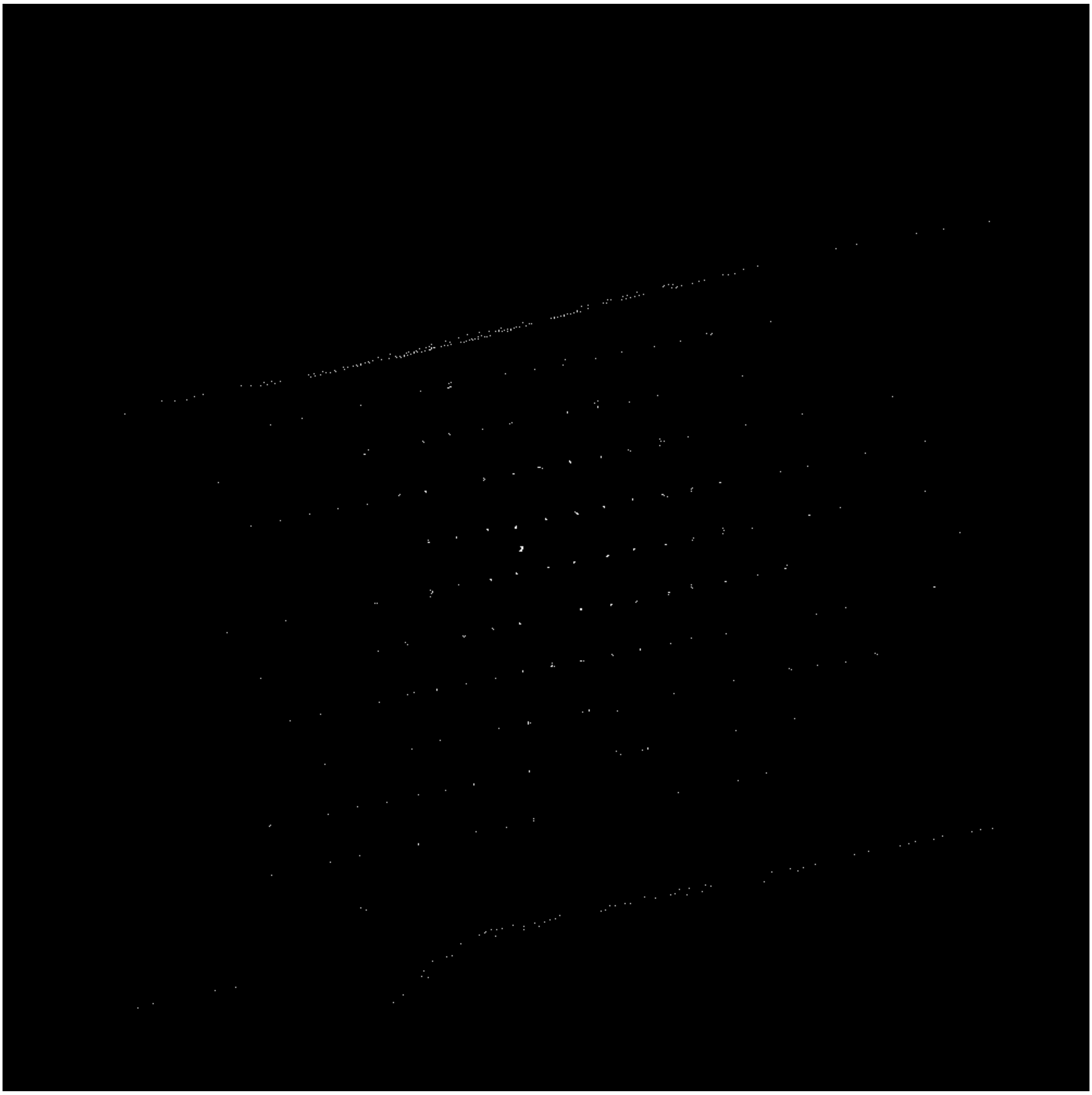

*Figure 147: An output of labelling a frame of lidar data (Figure 142) where many of the labelled points correspond to posts, trunks, people or hedges.*

### 4.5.8 3D Lidar Feature Extraction Testing

The ground truth datasets, as described in Subsection 4.5.7, were used for testing the different 3D lidar feature extraction methods from Subsections 4.5.3 to 4.5.6. The procedure for testing was to:

1. Run the algorithm to be tested on the 100 test lidar frames and output the results in the same bird's eye view format as the labelled images.
2. A bespoke testing application was used to inspect each pixel of the outputs to determine if the results were positive or negative, while each pixel of the ground truth was inspected to determine if the results were false or true. Thus the true positive, false positive, true negative and false negative pixels were counted for each test image.
3. Step 2 was repeated for ground truth data where only the posts and trunks were labelled as well as for ground truth data where posts, trunks, people and hedges were labelled.

The plane selection methods had multiple results because there were three metrics proposed and tested for plane selection. In addition, for each of these plane selection metrics, it was unclear how many angular segments should be used and so multiple values have been tested. For simplicity, the numbers used for the number of angular segments for plane selection were the whole number factors of the horizontal resolution of the lidar.

Because the lidar data was sparse in terms of the ratio of points to blank space in the bird's eye view, there were relatively few false positives (FP) compared to a relatively large number of true negatives (TN). This resulted in miniscule differences in the specificity for different methods, since specificity is given by:

$$\frac{TN}{TN+FP} \tag{18}$$

Hence, using the specificity to interpret results was difficult and instead the precision and recall were used to summarise the results. The results from testing the plane selection methods using the test procedure are illustrated in Figure 148 for ground truth including all structured features and in Figure 149 for the ground truth labels that only included the posts and trunks. These results suggest that the mean of the values in an angular segment gave better performance for plane selection for the tested dataset. Across the different metrics the optimum number of angular segments of the lidar data was between 2 and 30, as shown by the highlighted values in Table 26. For the mean plane selection metric, the optimum number of angular segments was between 2 and 10, with an average maximum of 5, across precision, recall and the different datasets. Hence, a value of 5 angular segments was used for subsequent testing with plane selection, using the mean of the values in a segment.

Figure 150 shows the precision recall curves for feature extraction by plane selection using the maximum mean metric across a range of segment numbers for the posts and trunks dataset as well as the dataset including the orchard boundaries. These curves show a limited peak region as opposed to a trade off between precision and recall, indicating there is a limited optimal region for the number of segments used for plane selection.

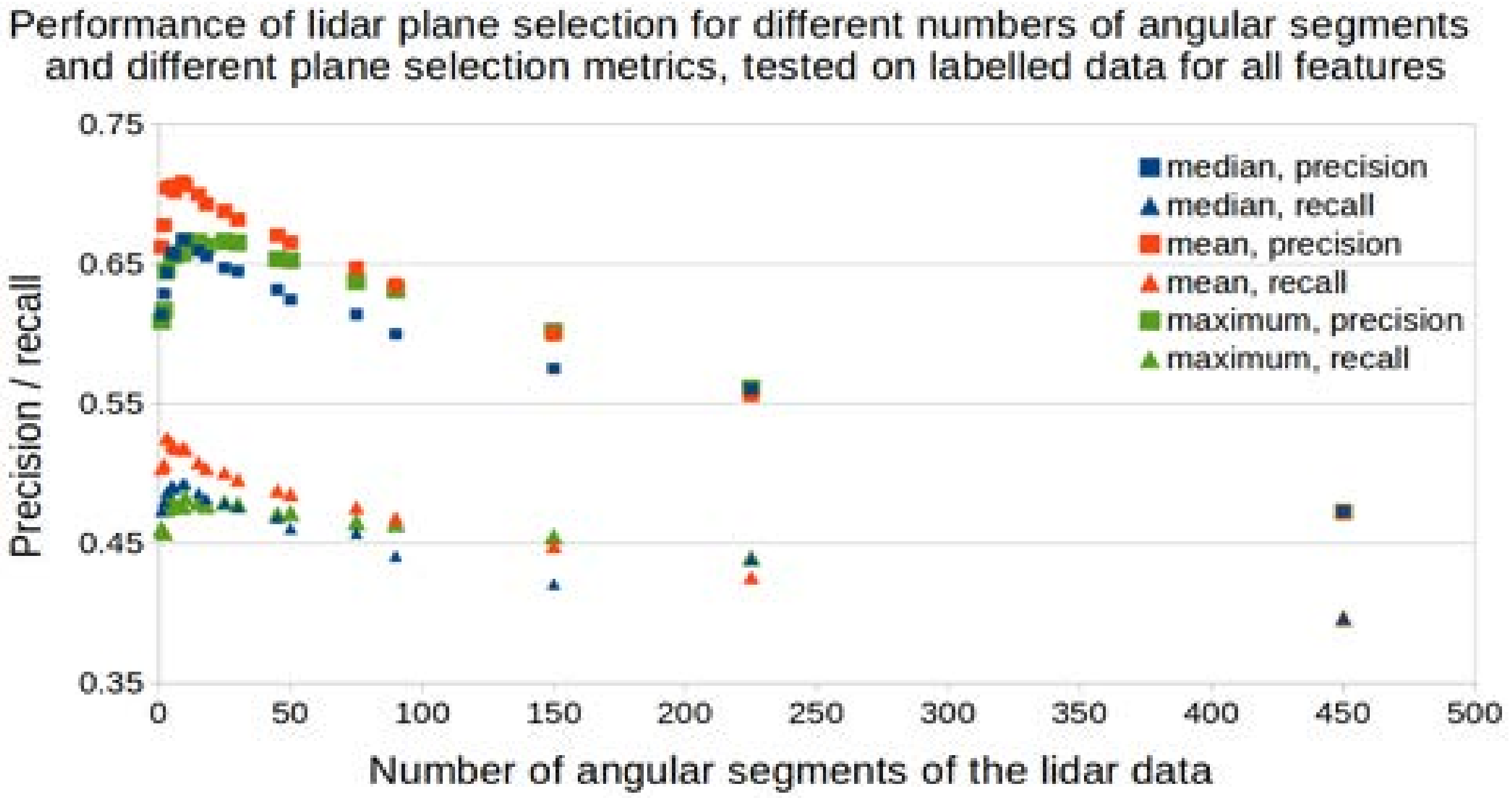

*Figure 148: Results from plane selection methods for feature extraction from lidar data, tested against ground truth points labelled as posts, trunks, people and orchard block boundaries.*

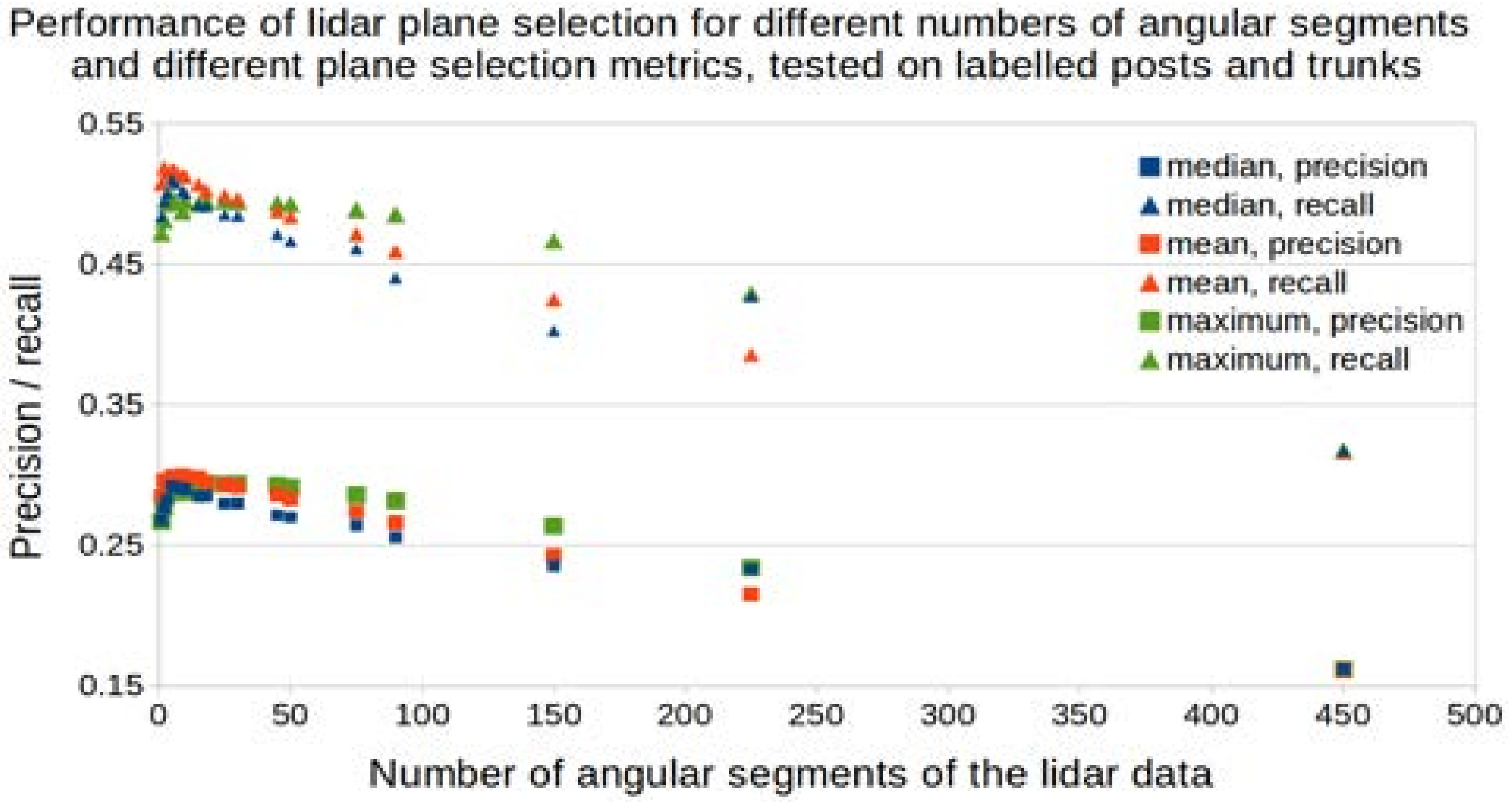

*Figure 149: Results from repeating the lidar plane selection testing with data labelled just for the posts and trunks.*

*Table 26: Results from testing the plane selection method for lidar feature extraction with different datasets, numbers of angular segments and plane selection metrics.*

| *Number of angular segments* | **All labelled objects dataset** | | | | | | **Posts and trunks only dataset** | | | | | |
|---|---|---|---|---|---|---|---|---|---|---|---|---|
| | **median metric** | | **mean metric** | | **maximum metric** | | **median metric** | | **mean metric** | | **maximum metric** | |
| | *precision* | *recall* | *precision* | *recall* | *precision* | *recall* | *precision* | *recall* | *precision* | *recall* | *precision* | *recall* |
| 1 | 0.614 | 0.473 | 0.662 | 0.504 | 0.609 | 0.460 | 0.268 | 0.484 | 0.285 | 0.507 | 0.267 | 0.472 |
| 2 | 0.629 | 0.480 | 0.678 | 0.506 | 0.617 | 0.458 | 0.277 | 0.494 | 0.296 | 0.518 | 0.277 | 0.481 |
| 3 | 0.644 | 0.486 | 0.704 | 0.525 | 0.645 | 0.475 | 0.283 | 0.500 | 0.295 | 0.514 | 0.286 | 0.493 |
| 5 | 0.659 | 0.491 | 0.706 | 0.521 | 0.656 | 0.481 | 0.292 | 0.510 | 0.299 | 0.516 | 0.290 | 0.497 |
| 6 | 0.657 | 0.489 | 0.702 | 0.518 | 0.656 | 0.478 | 0.291 | 0.507 | 0.299 | 0.516 | 0.289 | 0.494 |
| 9 | 0.667 | 0.492 | 0.708 | 0.518 | 0.658 | 0.476 | 0.290 | 0.502 | 0.299 | 0.513 | 0.287 | 0.487 |
| 10 | 0.667 | 0.492 | 0.706 | 0.517 | 0.667 | 0.483 | 0.290 | 0.500 | 0.299 | 0.512 | 0.292 | 0.495 |
| 15 | 0.660 | 0.486 | 0.700 | 0.508 | 0.666 | 0.479 | 0.285 | 0.492 | 0.298 | 0.507 | 0.292 | 0.492 |
| 18 | 0.655 | 0.482 | 0.693 | 0.504 | 0.663 | 0.477 | 0.285 | 0.491 | 0.295 | 0.502 | 0.294 | 0.495 |
| 25 | 0.647 | 0.479 | 0.688 | 0.501 | 0.666 | 0.479 | 0.280 | 0.485 | 0.293 | 0.499 | 0.293 | 0.495 |
| 30 | 0.645 | 0.476 | 0.682 | 0.495 | 0.665 | 0.478 | 0.280 | 0.484 | 0.291 | 0.496 | 0.294 | 0.494 |
| 45 | 0.632 | 0.468 | 0.670 | 0.488 | 0.653 | 0.471 | 0.271 | 0.471 | 0.286 | 0.487 | 0.292 | 0.493 |
| 50 | 0.625 | 0.461 | 0.665 | 0.485 | 0.653 | 0.472 | 0.270 | 0.466 | 0.283 | 0.483 | 0.291 | 0.493 |
| 75 | 0.614 | 0.457 | 0.647 | 0.476 | 0.637 | 0.465 | 0.264 | 0.461 | 0.274 | 0.471 | 0.286 | 0.488 |
| 90 | 0.600 | 0.441 | 0.635 | 0.468 | 0.631 | 0.464 | 0.256 | 0.440 | 0.266 | 0.459 | 0.282 | 0.485 |
| 150 | 0.575 | 0.421 | 0.600 | 0.448 | 0.602 | 0.455 | 0.235 | 0.403 | 0.243 | 0.425 | 0.264 | 0.467 |
| 225 | 0.561 | 0.439 | 0.556 | 0.426 | 0.561 | 0.440 | 0.233 | 0.427 | 0.215 | 0.385 | 0.234 | 0.429 |
| 450 | 0.473 | 0.396 | 0.473 | 0.396 | 0.473 | 0.396 | 0.162 | 0.317 | 0.161 | 0.316 | 0.162 | 0.317 |

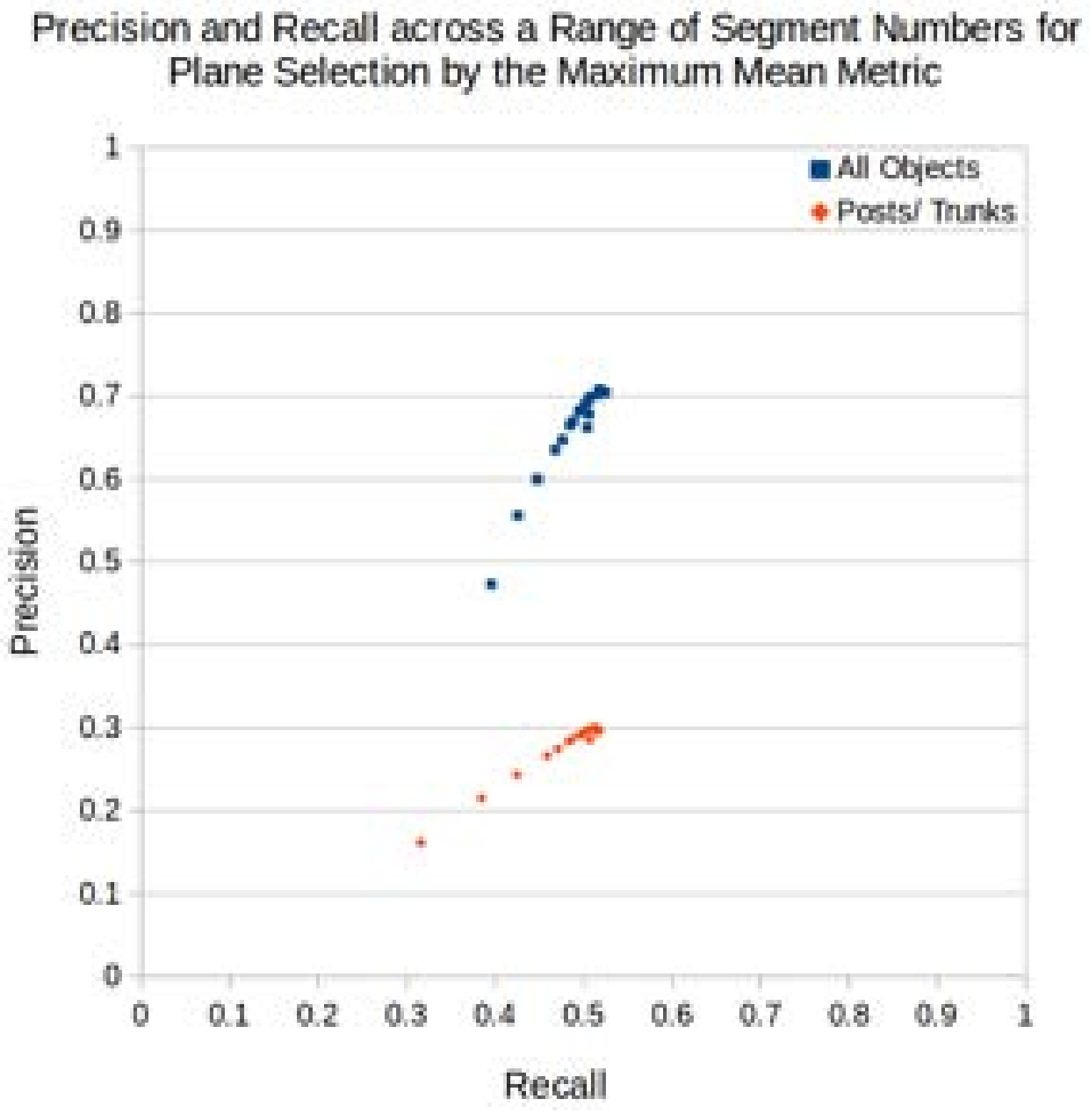

*Figure 150: Precision versus recall for feature extraction by plane selection using the maximum mean metric across a range of segment numbers.*

Table 27 and Figure 151 give the precision and recall results for feature extraction using the scaled density image method with different thresholds, for the dataset with just posts and trunks as well as for the dataset with orchard boundary objects included. For these results, the different thresholds do produce a trade off between precision and recall. The precision recall curves do not reach high into the top right hand side of the graph, indicating low performance for the thresholded scaled density image method.

*Table 27: Precision and recall results for feature extraction by scaled density image with different thresholds and for the two datasets with and without orchard boundary objects.*

| | **All Objects Dataset** | | **Posts and Trunks Dataset** | |
|---|---|---|---|---|
| **Threshold** | **Precision** | **Recall** | **Precision** | **Recall** |
| 0 | 0.104 | 1 | 0.044 | 1 |
| 10 | 0.104 | 0.998 | 0.044 | 0.998 |
| 20 | 0.145 | 0.991 | 0.062 | 0.988 |
| 30 | 0.264 | 0.901 | 0.107 | 0.857 |
| 40 | 0.457 | 0.518 | 0.168 | 0.445 |
| 50 | 0.598 | 0.315 | 0.251 | 0.310 |
| 60 | 0.684 | 0.279 | 0.274 | 0.262 |
| 70 | 0.737 | 0.216 | 0.282 | 0.194 |
| 80 | 0.763 | 0.156 | 0.292 | 0.139 |
| 90 | 0.788 | 0.129 | 0.295 | 0.113 |
| 100 | 0.821 | 0.104 | 0.296 | 0.088 |
| 110 | 0.840 | 0.086 | 0.296 | 0.071 |
| 120 | 0.864 | 0.071 | 0.300 | 0.058 |
| 130 | 0.883 | 0.059 | 0.308 | 0.048 |
| 140 | 0.900 | 0.049 | 0.307 | 0.039 |
| 150 | 0.914 | 0.040 | 0.318 | 0.033 |
| 160 | 0.928 | 0.033 | 0.329 | 0.028 |
| 170 | 0.942 | 0.028 | 0.351 | 0.024 |
| 180 | 0.951 | 0.024 | 0.362 | 0.021 |
| 190 | 0.959 | 0.020 | 0.372 | 0.018 |
| 200 | 0.968 | 0.017 | 0.382 | 0.016 |
| 210 | 0.973 | 0.015 | 0.390 | 0.014 |
| 220 | 0.981 | 0.012 | 0.403 | 0.012 |
| 230 | 0.983 | 0.010 | 0.424 | 0.010 |
| 240 | 0.987 | 0.009 | 0.436 | 0.009 |
| 250 | 0.987 | 0.007 | 0.438 | 0.008 |

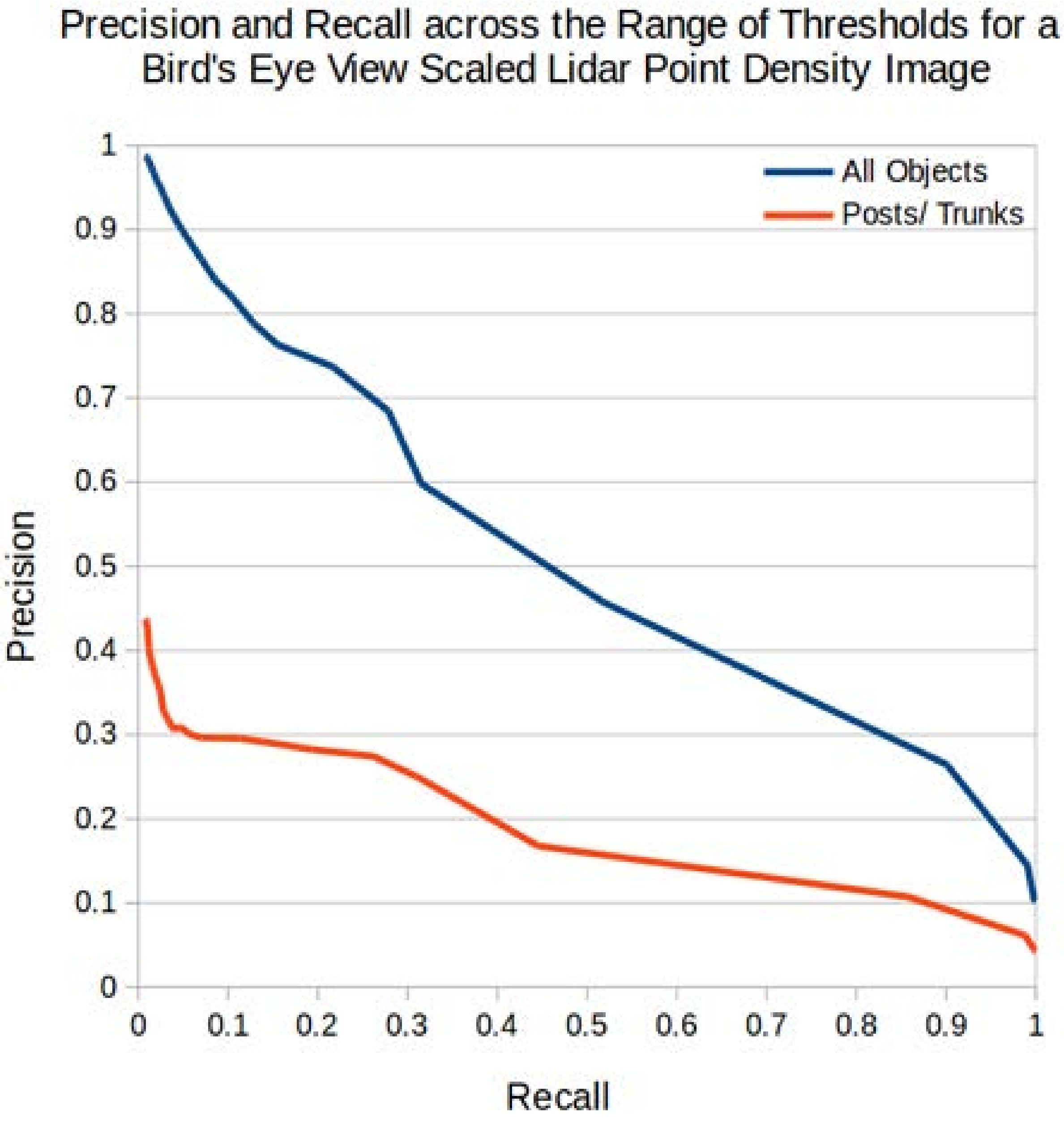

*Figure 151: Precision versus recall for feature extraction by scaled density images, using different thresholds for the datasets with and without the orchard boundary objects.*

Figure 152 shows the precision recall curve for the vertical connected objects method for feature extraction, using different object height thresholds with the two feature extraction datasets. As with the plane selection method, there is a limited peak region for the vertical connected objects method. This peak region corresponds to optimal values for the height threshold used for the objects. Examining the raw data showed that the optimal height threshold was in the range of 0.4 m to 0.5 m. The value of this optimal height threshold might correspond to the height of many weeds or branches hanging from the canopy. Hence, for a less well maintained kiwifruit orchard, it is possible that this threshold might be higher.

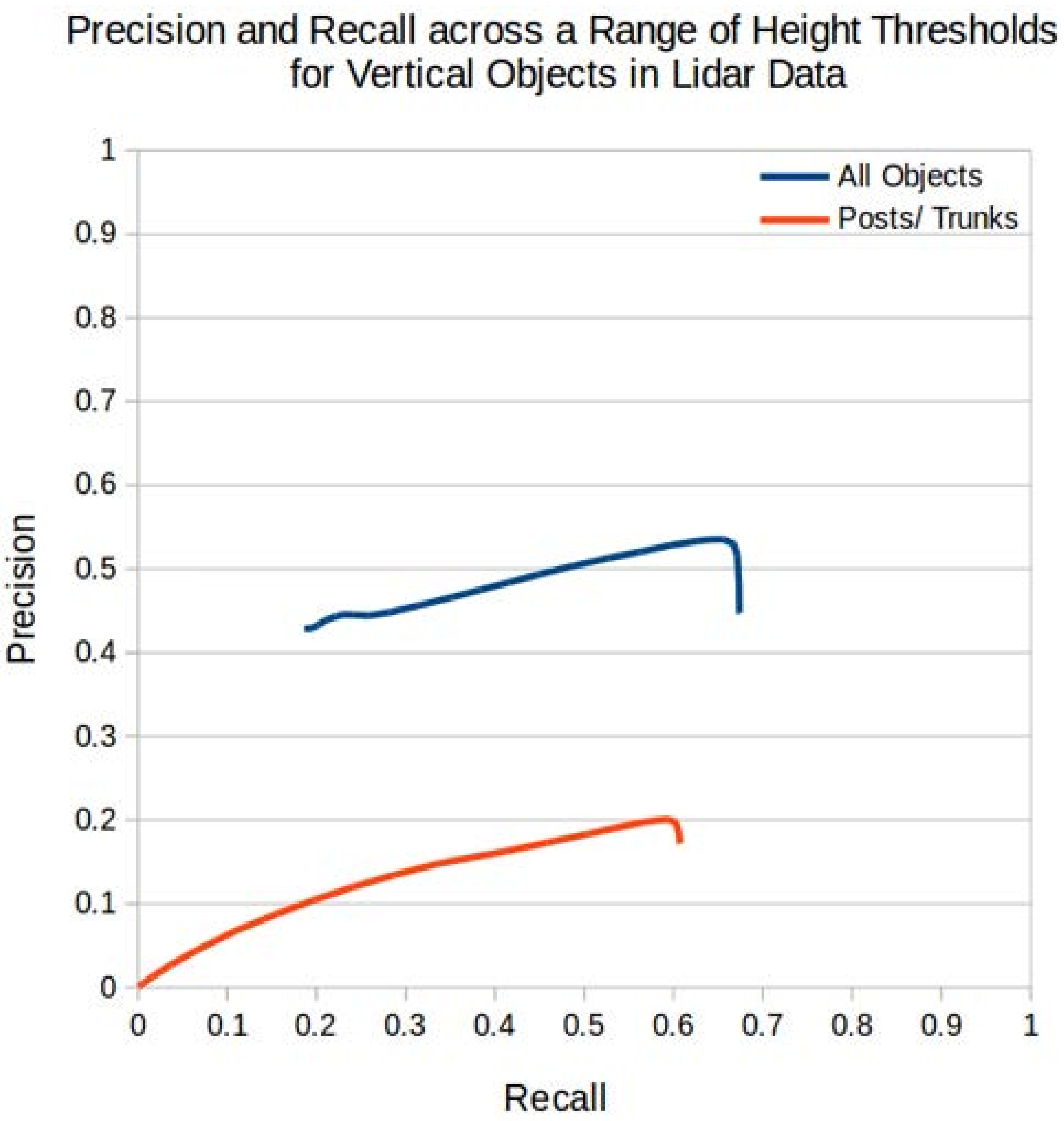

*Figure 152: Precision versus recall for the vertical connected objects feature extraction method, using different object height thresholds with the posts and trunks dataset as well as the dataset with the orchard boundary objects.*

Figure 153 summarises the results of testing the different feature extraction methods with the two feature extraction datasets. From these results, it seemed that there was no clear best method out of the methods tested. Depending on what precision or recall was required, a different method might be selected. Other factors to take into account for selecting a feature extraction method include the processing time. Table 28 lists the processing times for the different feature extraction methods. The plane selection method was found to be the fastest method, which could be important when using feature extraction in real time for mapping and localisation on the robot.

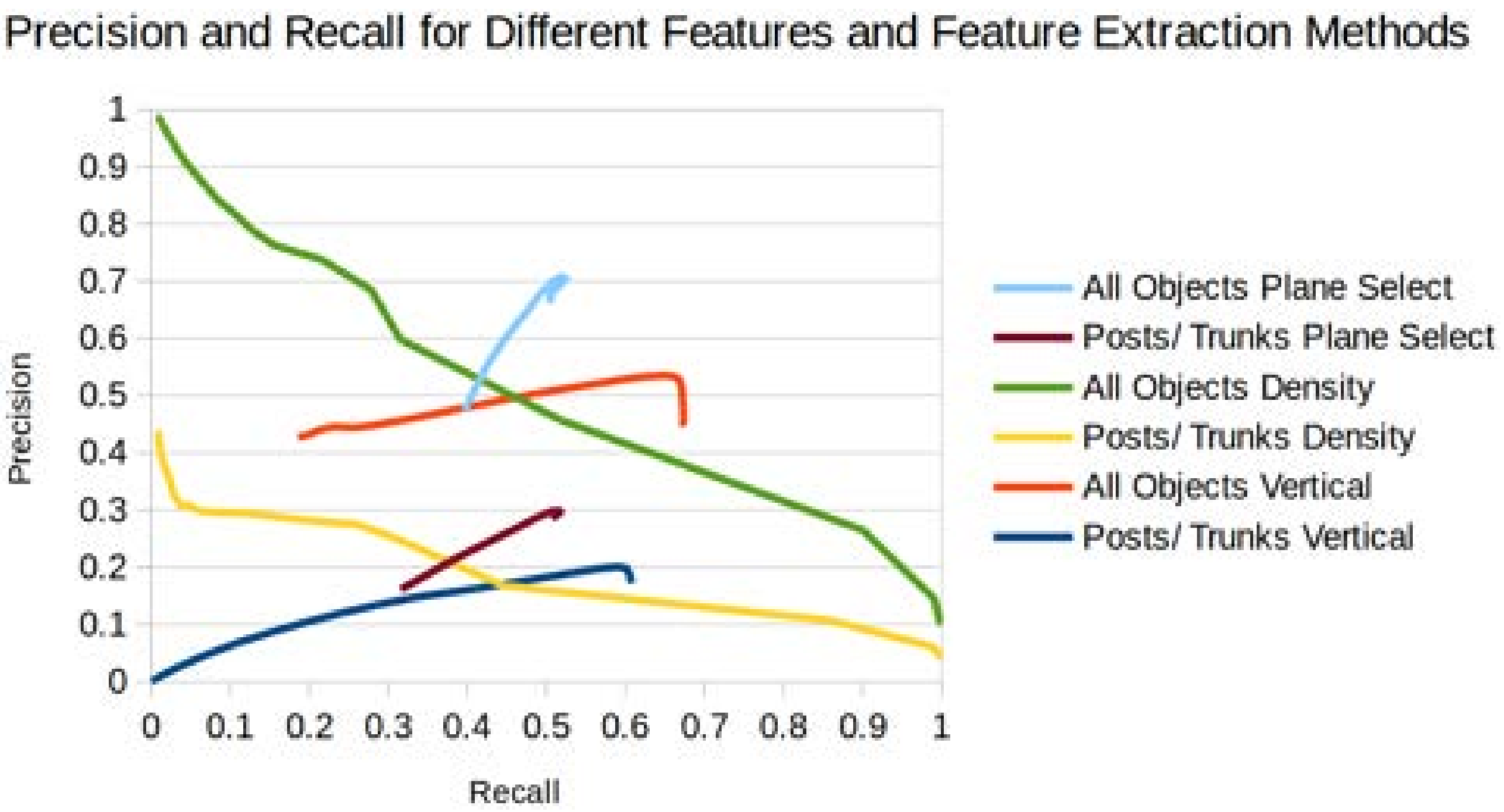

*Figure 153: Comparing all of the precision and recall results for the different 3D lidar mapping and localisation feature extraction methods and the two feature extraction datasets.*

*Table 28: Processing times for the 3D lidar mapping and localisation feature extraction methods.*

| Feature Extraction Method | Processing Time (s) |
|---|---|
| Plane Selection by Maximum Mean | 0.018 |
| Thresholded Scaled Density Image | 0.119 |
| Connected Vertical Objects | 0.063 |

### *4.5.9 3D Lidar Feature Extraction Methods Discussion*

The goal of 3D lidar feature extraction was to produce 2D data that contained as many structure defining features as possible, while removing non-structure defining features. It was intended that the extracted data would be used for SLAM and then localisation from the map, in order to supersede the row following based navigation system.

A method for selecting the plane of the 3D lidar data that had the highest ranges was developed. This method was extended by splitting the data into multiple segments. Multiple variants of the algorithm were tested with the plane selected using the mean of the ranges, the maximum of the ranges or the median of the ranges in each segment. When testing against a ground truth dataset, the variant using the mean of the ranges produced the highest precision and recall. The optimal number of segments of the lidar data was between 2 and 30.

Another method for extracting the structure defining features considered the lidar data from a bird's eye view and created a count of the points in grid cells. The counts were scaled by a non-linear function related to the change in density of points with distance away from the lidar. The scaled density of points was thresholded to give the output.

In addition, another method considered the angle between vertically connected points and clustered points based on an angle threshold. The height of the resulting clusters was thresholded to produce the output.

The different feature extraction methods were compared using a hand labelled ground truth dataset; however, the method selected would depend on the precision and recall acceptable for the application. The higher processing speed and balance of relatively good precision and recall make plane selection, using the mean metric with 5 segments, a potential candidate for future work.

## 4.6 Kiwifruit Orchard SLAM Gmapping and AMCL

It was thought that a feature extraction method from Subsection 4.5 would be used to create a 2D lidar scan input to SLAM Gmapping [192], [193] and AMCL [194], [195]. Indeed, Gmapping was used to create maps, as shown in Figure 154, and AMCL was used to localise off such maps. The development of an AMCL based navigation system for kiwifruit orchards was deferred as future work for a commercial development project. Although, mapping and localisation using Gmapping and AMCL were used for kiwifruit orchards, it is unclear if these approaches could also be used for other types of orchards with different structures; such an investigation may also be future work.

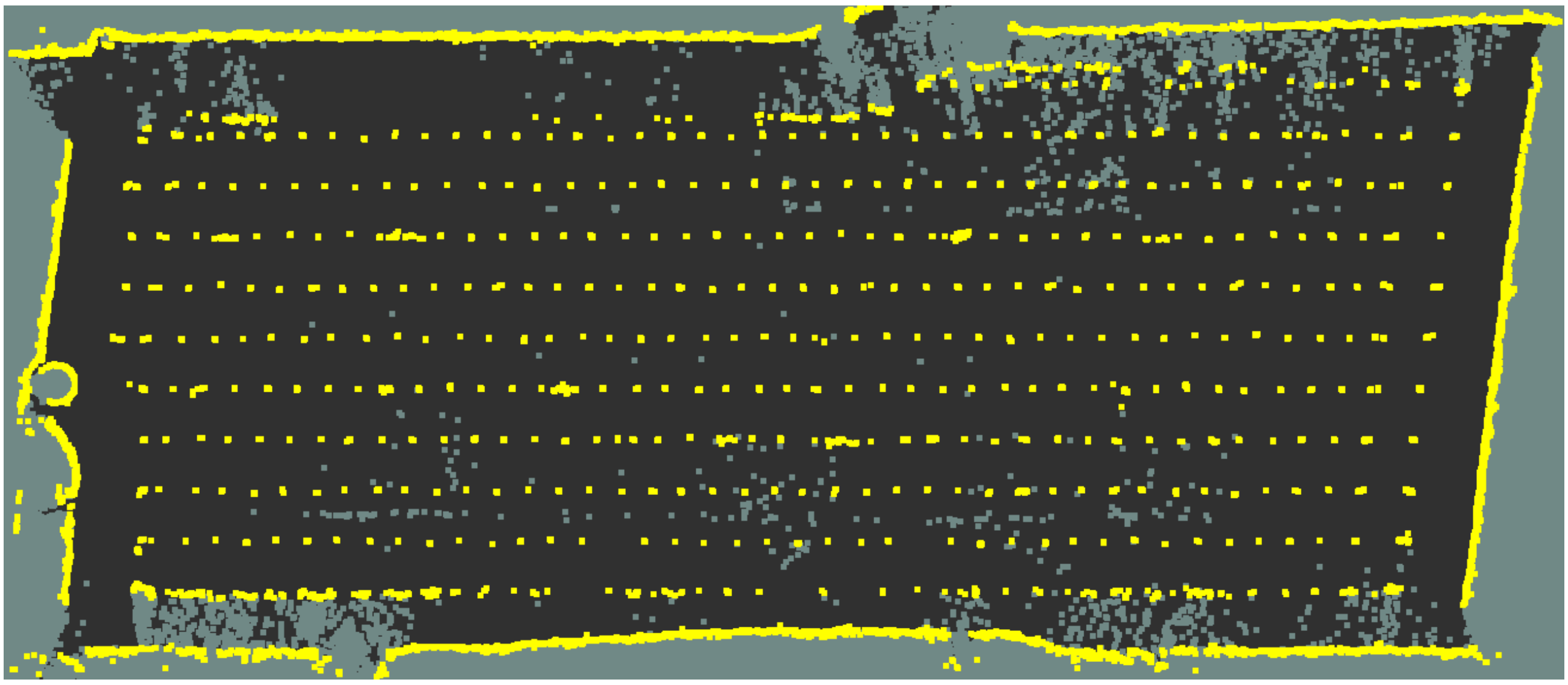

*Figure 154: A map created using GMapping and 3D lidar data.*

## 4.7 Kiwifruit Orchard Row Following Using Computer Vision

The requirements for the navigation system, which were stipulated by the MBIE Multipurpose Orchard Robotics Project goals, included the use of multiple types of sensors. The motivation for this requirement was to achieve greater robustness through the use of diverse and independent sensing channels. This is a common approach for creating more robust systems, as outlined in international standards, such as ISO13849-1:2015 [1].

It was decided to experiment with colour cameras since they are substantially different to the 3D lidars that had been used to achieve the previous orchard navigation results. Colour cameras are passive sensors, offer higher spatial resolution and provide colour information. These differences in the sensors that form the navigation system are important because they provide diversity, which helps to reduce the risk of Common Cause Failures [197].

### *4.7.1 Row Centreline Regression Problem Formulation*

It was hypothesized that a CNN could be used to detect the kiwifruit row centrelines in colour images and it was thought that this would allow for row following. The options considered for how to formulate the output of the CNN for row centreline regression included:

- A centreline could be defined by an image intercept and inverse gradient. The intercept could be the intercept of the bottom of the image, measured in pixels. The inverse of the gradient might be used in order to avoid a divide by zero issue for vertical centrelines in the image.
- A centreline could be defined by two points for the start and end points of the line. The end point could be defined as the centre of the end of the visible part of the row. The start point could be defined as the closest row centre point, where both sides of the row were visible. It was thought that it would be necessary to define the start point constrained by where the row sides were visible because it was difficult to label the row centre, where either side of the row was not visible.

The second option seemed advantageous because it seemed that having the endpoint could give information about the distance to the end of the row. Hence, the CNNs described here for row centreline regression had two points as their output.

In order to create datasets for supervised learning, images were collected from cameras mounted at a height of approximately 0.8 m from the ground, with the principal axis of the cameras parallel to the direction of travel. Various measures were employed to introduce variability in the data, including:

- Using different camera makes and models.
- Collecting data at different times of the day and in all seasons.
- Collecting data in different orchards.
- Collecting data in different weather conditions.
- Swerving in order to collect camera images from various poses in the row.

It was decided that a wide format image was advantageous because it could reduce the chance of the row edge not being visible in the foreground of the image. The image size initially selected was 800 pixels wide by 300 pixels high. This selection was made based on inspecting 1600 pixels wide by 1200 pixels high images from a Basler Dart daA1600-60uc [92] camera, with a 5.5 mm focal length lense. Regions of interest were selected in the images, corresponding to the features that defined the current row, including posts, trunks, beams and kiwifruit vine leaders. The size of the regions of interest were found to be approximately 1600 pixels wide by 600 pixels high. It was decided that this was quite a large image to perform training on using GPUs with 8 GB of memory and also quite large to perform inference with in real time, so it was decided to scale the image by a half, giving the selected image size of 800 pixels wide by 300 pixels high.

From the tens of thousands of collected images, 1132 were scaled and cropped to 800 pixels wide by 300 pixels high. These images were hand labelled and 100 of them were kept separate for validation.

It was found that it was not easy for a person to mark the row centreline by simply clicking on two points in an image because it was difficult to gauge the exact centre of the row in an image. Instead a tool was created which displayed the image and with four clicks the person performing the labelling could mark the row centreline. The first point clicked was the furthest visible point where one of the treelines met the ground; the second point clicked was the closest point where the same treeline met the ground. This process was repeated for the opposite treeline. The centre of the two far treeline points was used as the far centre point and similarly the centre of the two close treeline points was used as the close centre point of the row; the image coordinates of the two resulting centre points were used to define the row centreline for the image.

### 4.7.2 Row Centreline Regression with Bespoke CNN Architecture

Unlike more generic image processing tasks, like object detection and semantic segmentation, there were no clear classical or state of the art models for the specific task of row centreline regression and so it was decided to experiment with a bespoke network architecture. As a starting point, it was decided to use a series of convolutional, pooling, ReLU and normalisation layers, followed by fully connected layers.

An initial trial of row centreline regression was conducted using the bespoke CNN architecture given in Table 29. The hyperparameters used for training the CNN are given in Table 30. At the end of the training, the Euclidean loss for the training data was 0.7; the Euclidean loss for the validation data was 14000.

*Table 29: Bespoke CNN architecture trialled for row centreline regression.*

| **Layer Type** | **Kernel Size** | **Stride** | **Output Depth** |
|---|---|---|---|
| Convolutional | 5 | 1 | 20 |
| ReLU | | | |
| Max Pooling | 2 | 2 | |
| Local Response Normalisation | | | |
| Convolutional | 5 | 1 | 50 |
| ReLU | | | |
| Max Pooling | 2 | 2 | |
| Local Response Normalisation | | | |
| Convolutional | 3 | 1 | 100 |
| ReLU | | | |
| Max Pooling | 2 | 2 | |
| Local Response Normalisation | | | |
| Fully Connected | | | 200 |
| Fully Connected | | | 4 |

*Table 30: Hyperparameters used for training the bespoke row centreline regression CNN.*

| **Hyperparameter Description** | **Value** |
|---|---|
| Number of Training Iterations | 1500 |
| Batch Size | 21 |
| Solver Type | Stochastic Gradient Descent |
| Initial Learning Rate | 0.0000001 |
| Learning Rate Multiplier | 0.8 |
| Iterations between Learning Rate Reduction | 500 |

The network architecture and hyperparameters used for training were tweaked but regardless a large gap persisted between the validation and training loss. This large gap and the very low training loss suggested over-fitting on the training data.

It was proposed that more data could be a solution for the over-fitting. However, because labelling more row centrelines would be time consuming for the number of images required to make a difference, it was decided to first experiment with using a pretrained network and thereby leverage the vast amounts of data from large established datasets. In this case, a CNN might be trained on a dataset such as ImageNet, with millions of images and thousands of categories [198]. The resulting CNN would be expected to have generic weights and filters in its first layers, which may be applicable to many images, including those of orchards. Because many pretrained networks that are openly available are for well established and tested neural networks, the decision to use pretrained networks led to moving away from using a bespoke architecture in favour of an established CNN.

### 4.7.3 Row Centreline Regression with a Pretrained CaffeNet

A pretrained CaffeNet, sourced from Caffe [81] and the Caffe Model Zoo [199], was the CNN chosen to replace the bespoke CNN. The changes made to the standard CaffeNet network were:

- The reduction in outputs from 1000 to 4 numbers, which defined the start and end points of the row centreline.
- The use of the Euclidean loss instead of the softmax loss, since finding the points on the row centreline is a regression problem as opposed to a classification problem.
- Removal of dropout layers from the neural network since dropout will randomly affect the calculation of the points on the row centreline. Experiments were performed with dropout layers; however, it was found that removing dropout improved results.

The modified CaffeNet was trained with pretrained weights, except for in the fully connected layers because of the dimensionality differences of the row centreline regression problem, with larger images and only 4 outputs. The hyperparameters used for training are given in Table 31.

*Table 31: Hyperparameters used for training a CaffeNet CNN for row centreline regression.*

| **Hyperparameter Description** | **Value** |
|---|---|
| Number of Training Iterations | 50000 |
| Batch Size | 50 |
| Solver Type | Stochastic Gradient Descent |
| Initial Learning Rate | 0.000000001 |
| Learning Rate Multiplier | 0.8 |
| Iterations between Learning Rate Reduction | 10000 |

After training, the Euclidean loss for the training dataset was 300 and for the validation dataset it was 710. This result was encouraging because of the reduction in the validation dataset loss and the reduction in the gap between the training and validation dataset losses. However, these losses were considered to be quite high. The Euclidean loss was the sum of squared differences between the output coordinates, Y, and the corresponding ground truth labels, L, calculated according to:

$$\sum_{i=0}^{4} \left(Y[i]-L[i]\right)^2 \tag{19}$$

Using this equation, a validation loss of 710 corresponds to an average error for each line point coordinate of over 13 pixels. This produced a mix of inference results ranging from encouraging to erroneous, as demonstrated in Figure 155. Hence, more solutions for better accuracy were sought. The next approach used was to increase the amount of training data using data augmentation.

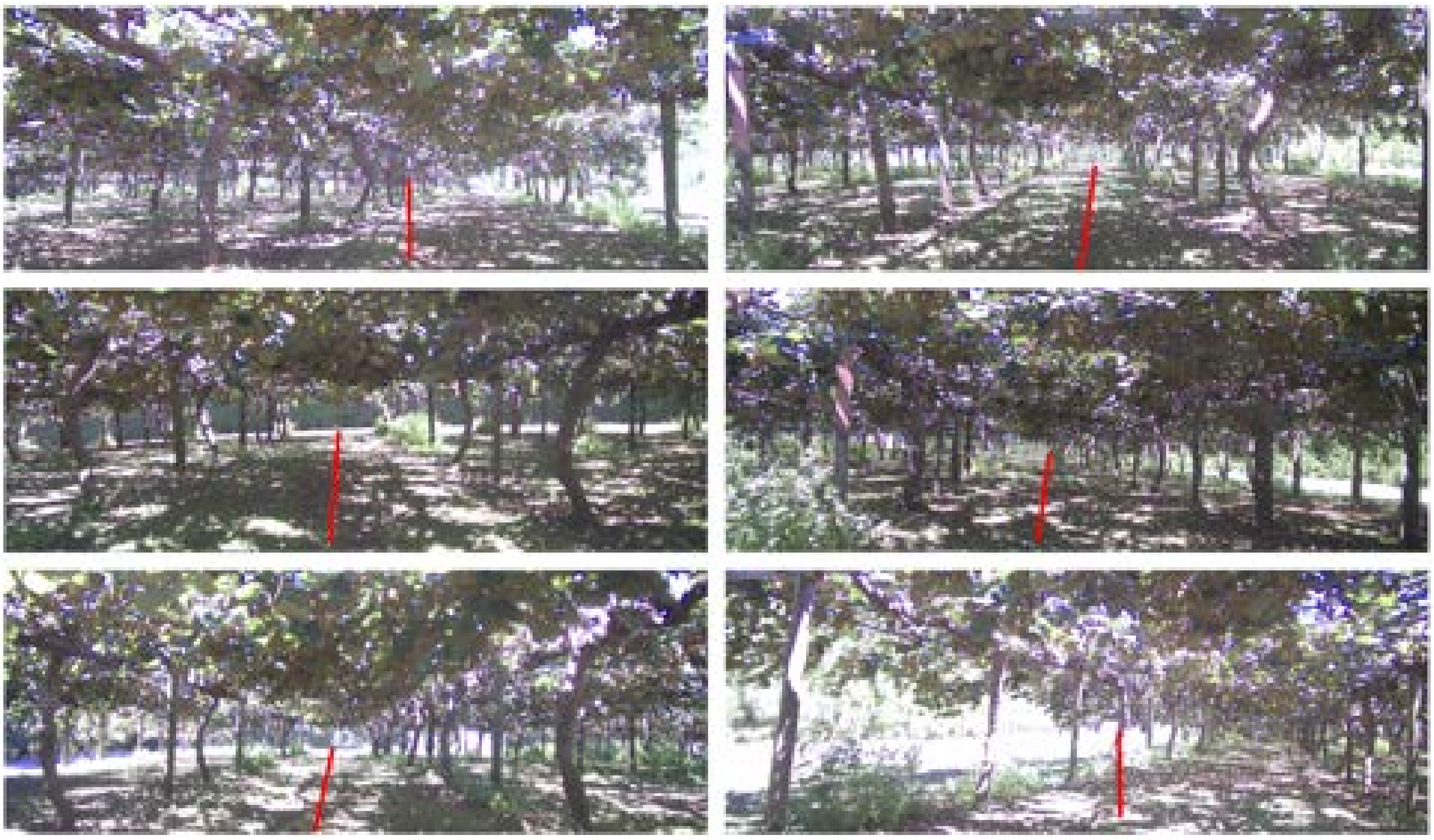

*Figure 155: A sample of CaffeNet row centreline regression validation dataset inference results.*

### 4.7.4 Data Augmentation for Row Centreline Regression

To inflate the dataset for training, it was decided to use data augmentation of the already hand labelled data. It was postulated that some common data augmentation techniques cannot be used for row centreline regression because they alter the position of the ground plane from the bottom of the image, which is something that is not seen in the real world. For example, the canopy should never be at the bottom of the image with fruit hanging upwards and the ground of the foreground should never be at the top of the image for row centreline regression. Data augmentation techniques which could cause such violations include large rotations and flips or mirroring about the x-axis of the image.

Colour alteration techniques can be applied to images without changing the corresponding label of the coordinates of the endpoints of the row centreline. However, other data augmentation techniques that change the pose of the row in the image do require the label to be recalculated, including:

- Scaling with cropping.
- Mirroring about the y-axis.
- Small rotations.
- Translations.

For rotations and translations there may be blank regions left in the image. It was not clear what to put in these blank spaces. It was postulated that leaving them blank may provide the CNN with artificial cues about the transformation performed on the image.

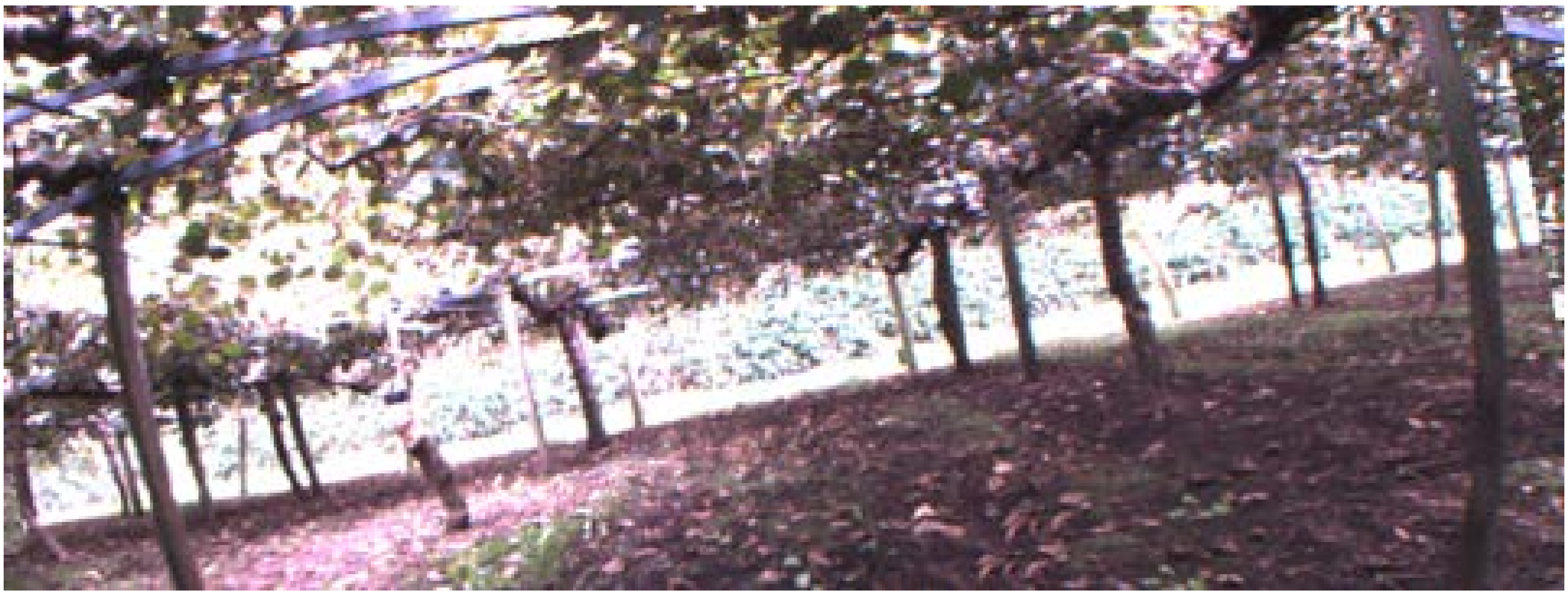

*Figure 156: An example of an image rotated about the bottom centre of the image with the original image as the background.*

It was postulated that arbitrarily filling the bottom of the image might have an adverse effect on the row centreline regression because this space may include features used for the regression. In contrast, the top of the image typically consisted of the canopy, which might be less influential in row centreline regression. Hence, it was postulated that arbitrarily filling the top of the image may have less of an effect on the detection than filling the bottom of the image. For example, consider the rotated image in Figure 156. The top left, top right and bottom left corners all show artefacts of the original image, which forms the background or fill for pixels that might otherwise be black. Because these artefacts are to the side or top of the image, they may not add features to what might be perceived to be the cues to the pose of the current row. However, the bottom right of the image shows a rotated post that joins to the end of the post from the original image. This combined post may be perceived as a part of the structure of the current row that is more towards the centre of the row and hence this created artefact at the bottom of the image could affect the calculation of the row centreline. To avoid such issues, the centres of rotation used were the two bottom corners of an image and the direction of rotation was clockwise for the left centre of rotation and anti-clockwise for the right centre of rotation.

For similar reasons to those just described for small rotations, it was postulated that arbitrarily filling to the sides and top should have a minor effect for small translations, because the row and most of the features that define it are towards the centre and bottom of the image. Translations used were up to 40 pixels long and were only applied horizontally. The fill for the image after translation was the original image.

For scaling with cropping, the images were scaled up by a set percentage. Then a randomly positioned crop, the same size as the original image and within the scaled image bounds, was taken and used for the training dataset.

Trials were conducted with data augmentation by horizontal mirroring, scaling with cropping, horizontal translations and small rotations. Each data augmentation method was applied to the 1032 hand-labelled images and trained with CaffeNet, using the hyperparameters from Table 31.

*Table 32: The Euclidean losses resulting from training a pretrained CaffeNet with datasets, which were inflated using different data augmentation techniques.*

| **Augmentation Method** | **Training Set Size** | **Training Set Loss** | **Validation Set Loss** |
|---|---|---|---|
| **No Augmentation** | 1032 | 300 | 710 |
| **Horizontal Mirroring** | 2064 | 360 | 610 |
| **Scale 20% & Crop** | 2064 | 340 | 740 |
| **Scale 10% & Crop** | 2064 | 330 | 620 |
| **Horizontal Translation** | 3096 | 370 | 640 |
| **Rotations, ±3º and ±6º** | 5160 | 410 | 710 |
| **Rotations, ±3º** | 3096 | 400 | 710 |

The final loss values were calculated using the average of losses from 10 epochs at the end of each training run. The results are summarised in Table 32. These results show that the improvements in the validation dataset losses with data augmentation were only modest. In addition, some tests were performed with combining multiple data augmentation techniques; however, the best results only produced validation dataset losses of 580 or higher, which corresponds to errors of over 12 pixels per line coordinate on average and only 1 pixel of improvement compared to not using any data augmentation. Reflecting on these results, it seemed that there was still a large number of options to consider related to network architectures and hyperparameters. Using another network architecture such as GoogLeNet [34] seemed likely to provide some improvement, since GoogLeNet has provided better performance than CaffeNet in benchmark testing on other tasks [200], [201].

### 4.7.5 Row Centreline Regression with a Pretrained GoogLeNet

GoogLeNet has multiple algorithmic advantages over CaffeNet, including greater depth, the use of training loss layers at multiple stages of the network and a range of convolutions, which are aggregated together in modules [34]. Hence, it was thought that GoogLeNet may provide some improvement in performance for the task of row centreline regression.

As with CaffeNet, the number of outputs from GoogLeNet was reduced from 1000 to 4 and the Euclidean loss was used. Also as with CaffeNet, a pretrained GoogLeNet network was used from Caffe [126] but not for the fully connected layers and the layers interconnected with the fully connected layers. The learning rate was multiplied by a factor of 10 for layers that were not pretrained. In addition, dropout layers were removed to assist with regression. The hyperparameters used for training the GoogLeNet were slightly different to those for CaffeNet and are given in Table 33. The learning rate was set higher because it was found that GoogLeNet could take a higher learning rate without diverging. The batch size was set lower because it was found that when using an 8 GB GPU, the memory limit was immediately exceeded while trying to train GoogLeNet with higher batch sizes. The data augmentation steps used were horizontal mirroring, horizontal translation and cropping an image that was scaled by 10 percent; this gave a dataset that had 12,384 examples.

*Table 33: Hyperparameters used for training GoogLeNet for row centreline regression.*

| **Hyperparameter Description** | **Value** |
|---|---|
| Number of Training Iterations | 100000 |
| Batch Size | 12 |
| Solver Type | Stochastic Gradient Descent |
| Initial Learning Rate | 0.00000001 |
| Learning Rate Multiplier | 0.8 |
| Iterations between Learning Rate Reduction | 10000 |

The validation dataset loss after training was 185, which corresponds to an average line coordinate error of less than 7 pixels. This was a marked improvement over the CaffeNet results and this improvement seems to be visible in the inference results, as can be seen in a representative sample shown in Figure 157. However, Figure 157 shows that even with the improvement from GoogLeNet, there can still be some inaccurate results. It was thought that a more robust method for row centreline detection should be found in order to avoid possible erratic driving from errors in row centreline regression.

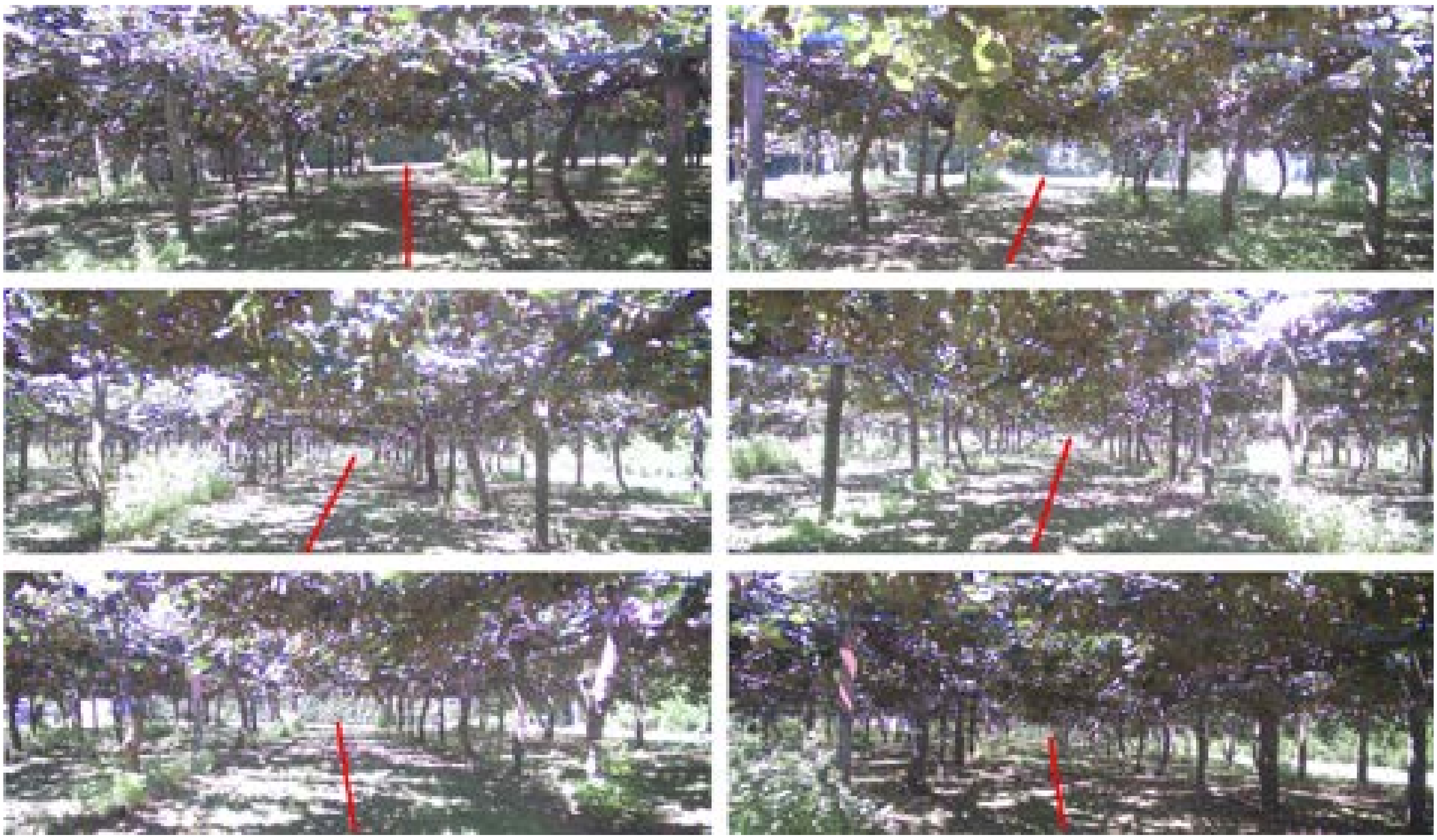

*Figure 157: Example validation dataset inference results from row centreline regression using GoogLeNet.*

It was speculated that some of the errors in the GoogLeNet results could be caused by some known issues with CNNs, including the loss of spatial resolution through the convolutional and down sampling layers as well as the lack of an intrinsic coordinate system in a CNN. Rather than solving these issues, it was proposed that these issues could be overcome by outputting many more values for the row centreline, as opposed to just 4 numbers for 2 points. By representing the row centreline with more values and having the CNN output these, it was thought that a post processing step could average the outputs of the CNN to find a more accurate row centreline.

One representation considered for the row centreline was to have 300 outputs, corresponding to the full height of the input image so that there would be one output for each y coordinate of the image. Each output would be the x coordinate of the row centreline for the given y coordinate. It was thought that this approach might work; however, instead it was decided to experiment with a more established approach by using semantic segmentation for objects in the kiwifruit orchard rows.

### 4.7.6 Semantic Segmentation of Kiwifruit Orchard Rows

As for harvesting and pollination, it was decided to experiment with Fully Convolutional Networks (FCNs) [45] to perform semantic segmentation and detect objects. Since FCNs can perform semantic segmentation of generic classes of objects in other datasets, it seemed likely that they would also be able to perform semantic segmentation of objects in kiwifruit orchard rows. It was thought that if the semantic segmentation did work, the labelled objects might then be used to identify the direction of rows for row following. The initial questions for this phase of research were:

1. Can FCNs be used to perform semantic segmentation of objects in images taken in kiwifruit orchard rows?
2. What objects could be segmented to allow row detection?
3. How could the segmented objects be processed to perform row detection and row following?

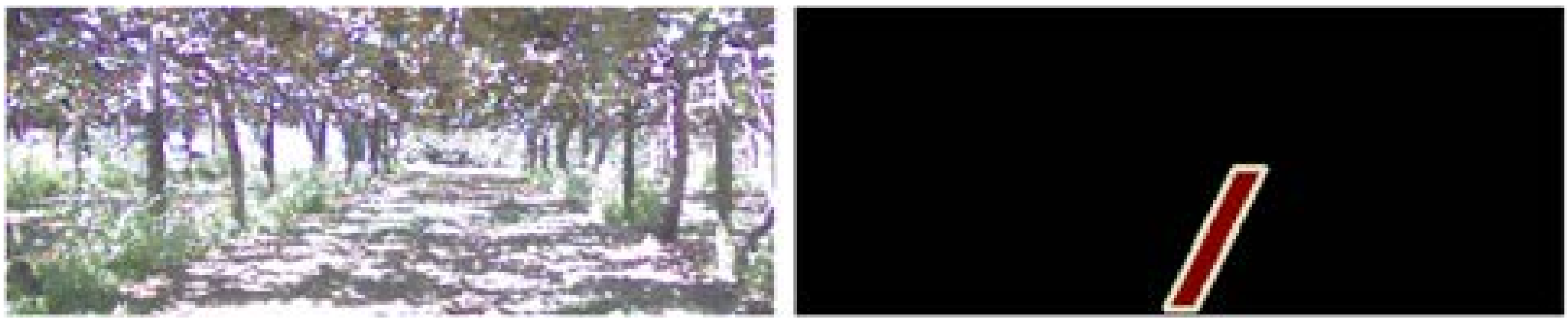

*Figure 158: An input (left) and label (right) pair for semantic segmentation of the row centreline.*

It was decided to follow on from the row centreline regression work by trying to use semantic segmentation to directly draw a row centreline. The approach used was to draw the row centreline label as a thick line, since it was postulated that the row centreline would need to have significant area in order to be produced by the up-sampling layers of the FCN. The thick line for the row centreline was drawn as a parallelogram. An example of an input and label pair is shown in Figure 158. The label was created by:

1. Clicking points on the input image and storing the coordinates of these points. The first two points clicked corresponded to either side of the end of the visible part of the current row. The next two points clicked were again on either side of the row in the image but as close as possible to the bottom of the image, while keeping the points vertically aligned in terms of their image y coordinates.
2. The centre of the pair of row end points and the centre of the close/ bottom points were both found by averaging. For both of these centres, points 10 pixels to the left and right of the centres were calculated. These points formed the corners of the parallelogram, representing the row centreline in the label as in Figure 158.

3. The parallelogram was flood-filled with the index-palette colour for the row centreline.

An FCN-8s network [45] was trained with the hyperparameters given in Table 34. The validation accuracy reached was 99%. An example of a typical inference result is shown in Figure 159. In this example and the other validation results observed, the output did not form a parallelogram like the training labels and the output did not seem to capture the direction of the row well. It seemed that semantic segmentation did produce outputs with some overlap of the label but the shape of the inference output did not reflect the row direction accurately and so it was thought that it might be difficult to use this output for row detection and row following.

*Table 34: Hyperparameters used for training FCNs for row features segmentation.*

| Hyperparameter Description | Value |
|---|---|
| Number of Training Images | 25 |
| Number of Validation Images | 5 |
| Input Image Height in Pixels | 200 |
| Input Image Width in Pixels | 500 |
| Number of Training Epochs | 500 |
| Solver Type | Adam |
| Initial Learning Rate | 0.0001 |
| Learning Rate Multiplier | 0.8 |
| Epochs between Learning Rate Reduction | 25 |

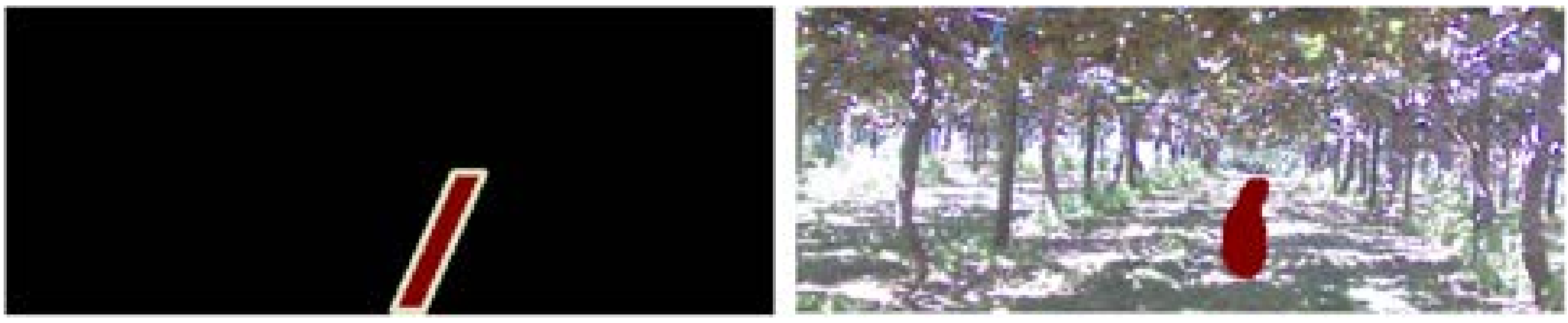

*Figure 159: A validation dataset label (left) and inference result (right) from row centreline segmentation using FCN-8s.*

### 4.7.7 Multiple Row Object Semantic Segmentation

Rather than trying to directly segment the row centreline, it was decided to experiment with segmenting classes of objects and then using the results to calculate the row centreline. A trial was performed using FCN-AlexNet [45], trained with images that were labelled with 5 classes: trees, posts, treeline, the ground of the current row and the end of the current row. A separate treeline class was used because it was found that it was difficult to label individual trees and posts that were far away in the image. An example of an input image and its label are given in Figure 160.

Some of the key training hyperparameters used are given in Table 34. A pretrained AlexNet, provided by the Caffe framework, was used for transfer learning [83]. After training, the validation accuracy, averaged across 3 attempts at training, was 90%. A typical example of inference on a validation image is shown in Figure 161. The AlexNet FCN appears to perform well for the current row end and row ground classes but not so well with the treeline and post classes. Performance on the trunks class was worst out of all of the classes, with no trunks segmented.

To try to improve the results from FCN AlexNet, FCN-8s was trained using the same images and hyperparameters. Existing results suggest that FCN-8s should deliver greater accuracy [45]. After training, the validation accuracy, averaged across 3 attempts at training, was 95%. A typical example of inference on a validation image is shown in Figure 162.

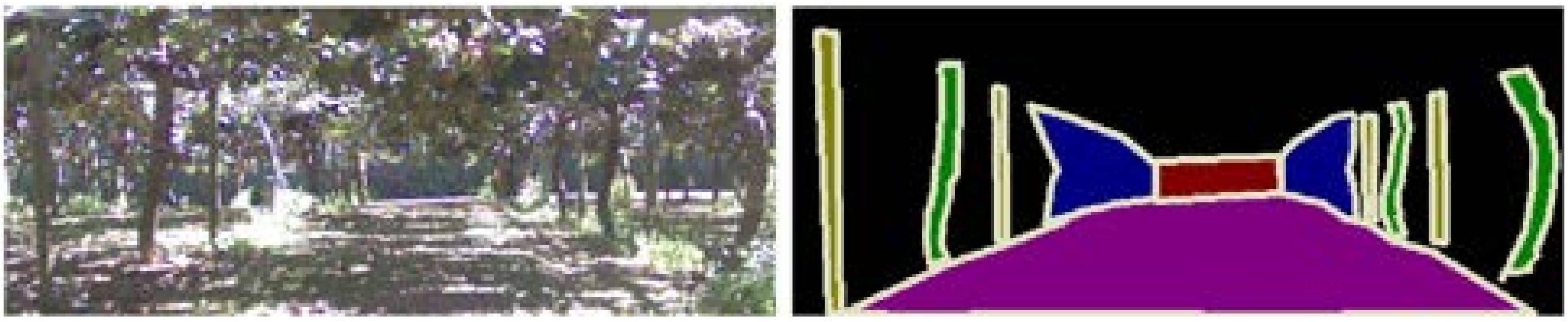

*Figure 160: An input image (left) and label (right) pair with five classes, which were trunks (green), posts (khaki), treeline (blue), current row end (red) and current row ground (magenta).*

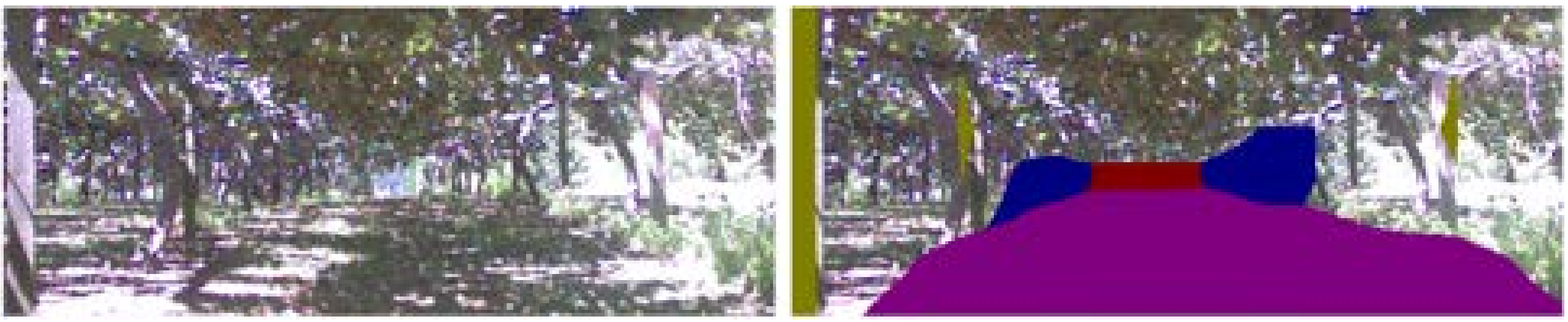

*Figure 161: Input image (left) and inference result (right) for kiwifruit row object segmentation using FCN-AlexNet.*

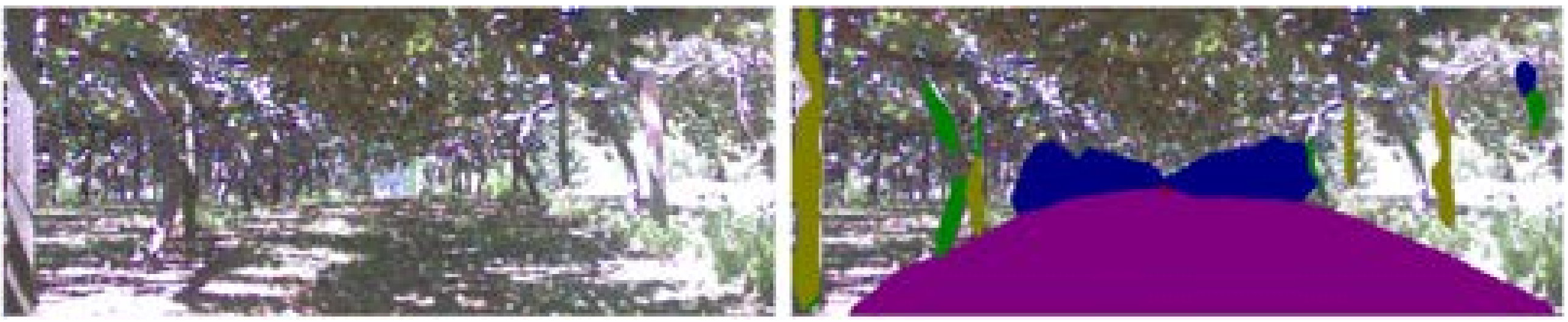

*Figure 162: Example input image (left) and inference output (right) from a FCN-8s network trained with 5 classes of objects in kiwifruit orchard rows.*

FCN-8s appeared to segment more posts and trunks than FCN-AlexNet. However, a key issue that FCN-8s had was distinguishing between the trunks, posts and treelines. A separate treeline class was used because trees and posts further away are harder to distinguish. However, from the inference results, it seemed possible that the difference between the treeline class, posts and trees was ambiguous. Hence, it was postulated that it might be better to have a single class that represents the entirety of the treelines in the current row without additional classes for posts and trunks. A dataset was created without the posts and trunks classes; an example input and label pair is shown in Figure 163. As before, a FCN-8s model was trained using the hyperparameters given in Table 34. An example inference result is shown in Figure 164. It seemed from such results that the approach of labelling the whole treeline as a single class was a more useful representation for extracting the direction of the row.

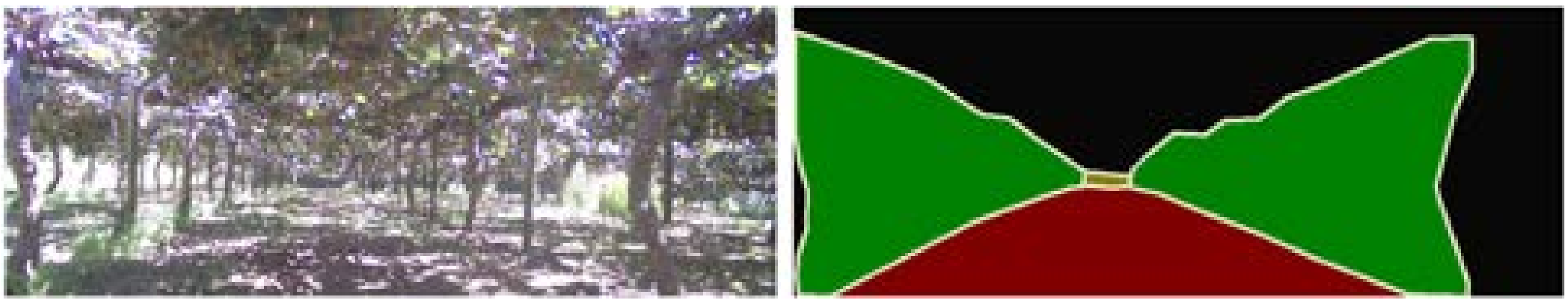

*Figure 163: An input image (left) and segmentation label (right) pair for a kiwifruit orchard row, using 3 classes for the row end, the ground in the current row and the treelines in the current row.*

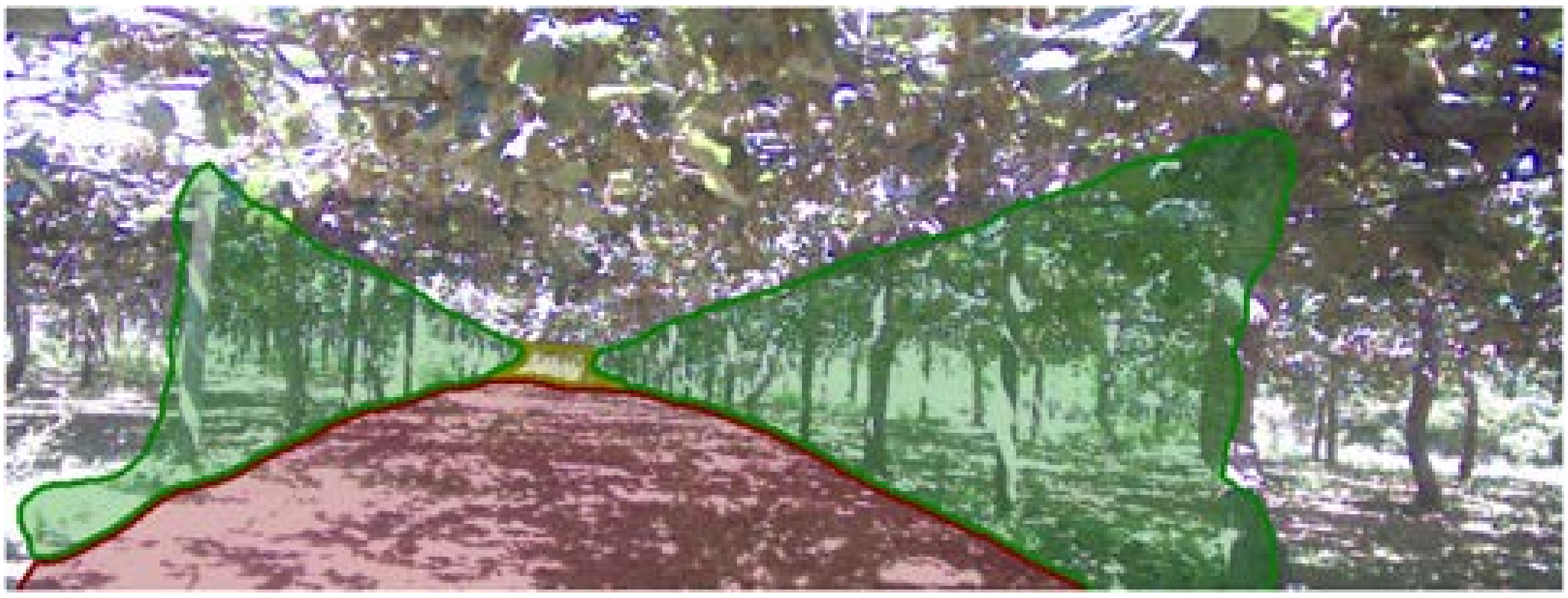

*Figure 164: A typical inference result from using FCN-8s to segment the current row ground, current row treelines and current row end.*

The treelines were the features used in the lidar row following algorithm in the navigation system for the AMMP. Hence, it was postulated that the treelines would be useful features for navigation using cameras. In this case, it was thought that it would be possible to use the current row treelines as the only class that was segmented. A treeline in this case included all visible posts and trunks on one side of a row. Another dataset was created with the treelines as the only class to segment; an example input and label pair is shown in Figure 165. FCN-8s was again trained using the hyperparameters given in Table 34.

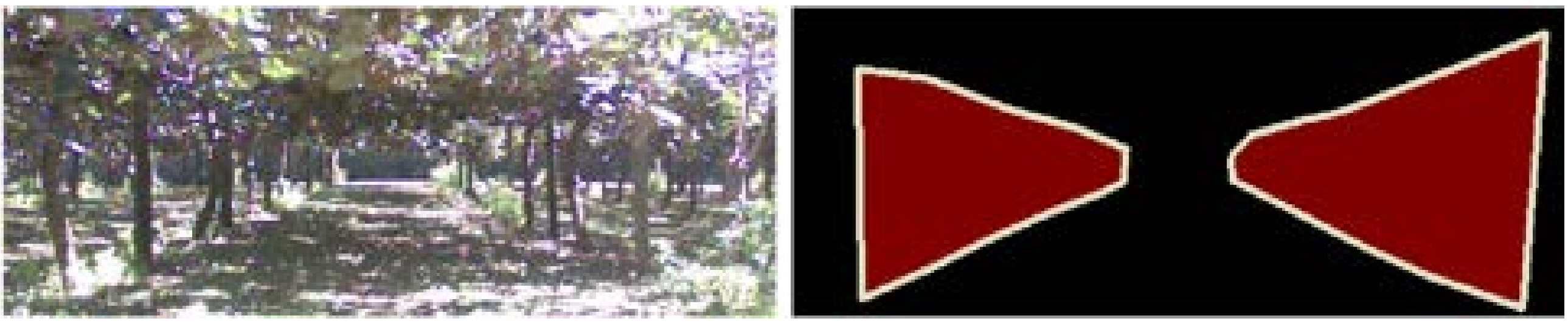

*Figure 165: An example of an input image (left) and segmentation label (right) pair, where the only class is the treeline of the current row.*

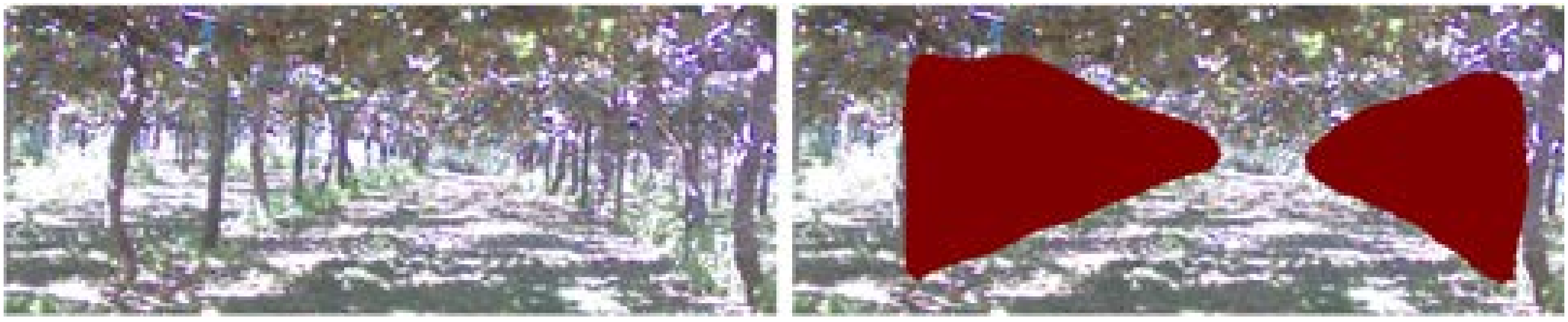

*Figure 166: Input image (left) and inference output (right) from a FCN-8s model, trained to detect the treelines in the current row of a kiwifruit orchard.*

After training, the validation accuracy, averaged across 3 attempts at training, was 95%. An example of the output of the trained model is shown in Figure 166. Generally, the model seems to detect the treeline well.

Segmenting the treeline seemed to make sense because it seemed that the treelines were distinctive features that defined the geometry of the row. The results from 3D lidar row following indicated that the treeline as a structure could be used for row following in kiwifruit orchards. However, it was also thought that the treeline in a colour camera image was not an output that translated directly to the row centreline for row detection and row following. For example, consider the output in Figure 166. To extract the row centreline from the segmented output, you might first extract the coordinates corresponding to the bottom of the treelines and fit lines to those points. However, the bottom of the treeline is just the outline of the ground of the current row, so it was thought that it might be simpler to segment the ground of the current row.

The results from FCN-AlexNet and FCN-8s for multiple classes, as shown in Figure 161, Figure 162 and Figure 164, seemed to indicate that the traversable space in the current row should be a feasible class to be segmented accurately. With the traversable space segmented, a line might be fitted to the centre of the output to find the row centreline.

### 4.7.8 Monocular Camera Semantic Segmentation Row Following

Another FCN-8s network [45] was trained but this time the data was labelled with the "immediately traversable space", which was taken to mean the space that a large mobile robot might be able to reach if translated in a straight line from the pose of the camera [202]. This more abstract class was used, instead of the current row ground class, because the traversable space class was used outside of the row and was also extended to monocular camera pedestrian avoidance (Subsection 5.4.9). An example of the labelled data is shown in Figure 167. The hyperparameters used for training were, again, those given in Table 34.

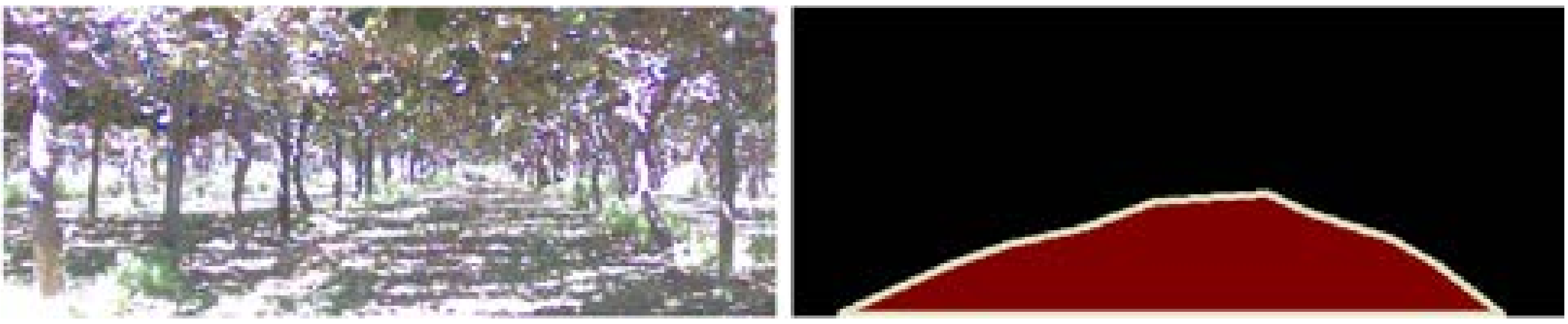

*Figure 167: An example of an input image (left) and segmentation label (right) pair, where the only class is the immediately traversable space.*

After training, the validation accuracy, averaged across 3 attempts at training, was 98%. An example of the output of the trained model is shown in Figure 168. An interesting aspect of segmenting the traversable space is the lack of false positives from adjacent rows, where the ground is also visible. Those rows are actually traversable but only when the robot is in those rows. In this sense, the traversability is conditional on the current state of the viewpoint. Despite this conditionality, the segmentation accuracy was high.

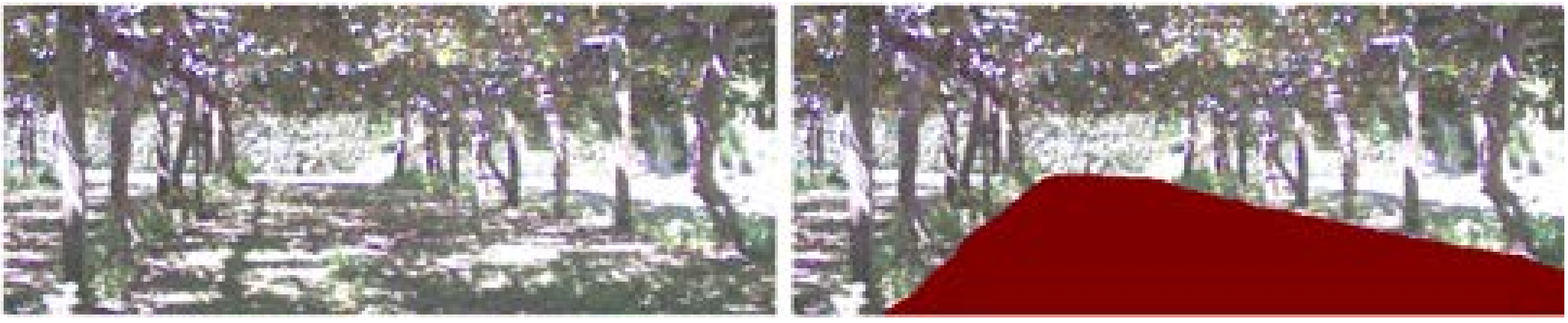

*Figure 168: An example input image (left) and output inference result (right) of a FCN-8s network trained on labels of traversable space.*

The results were encouraging enough that it was decided to proceed with this method for labelling and hence a larger dataset with 276 labelled images and 15 validation images was created. An FCN-8s model was trained as before but with this larger dataset. It is interesting to note that the final traversable space segmentation model tended to not produce false positive detections, even when only a small area of the current row was visible in the camera image (Figure 169). This was a useful outcome because it meant that the segmentation could work even with high angular displacements. It was also found that when no traversable space of the current row was visible, the false positives rate was low.

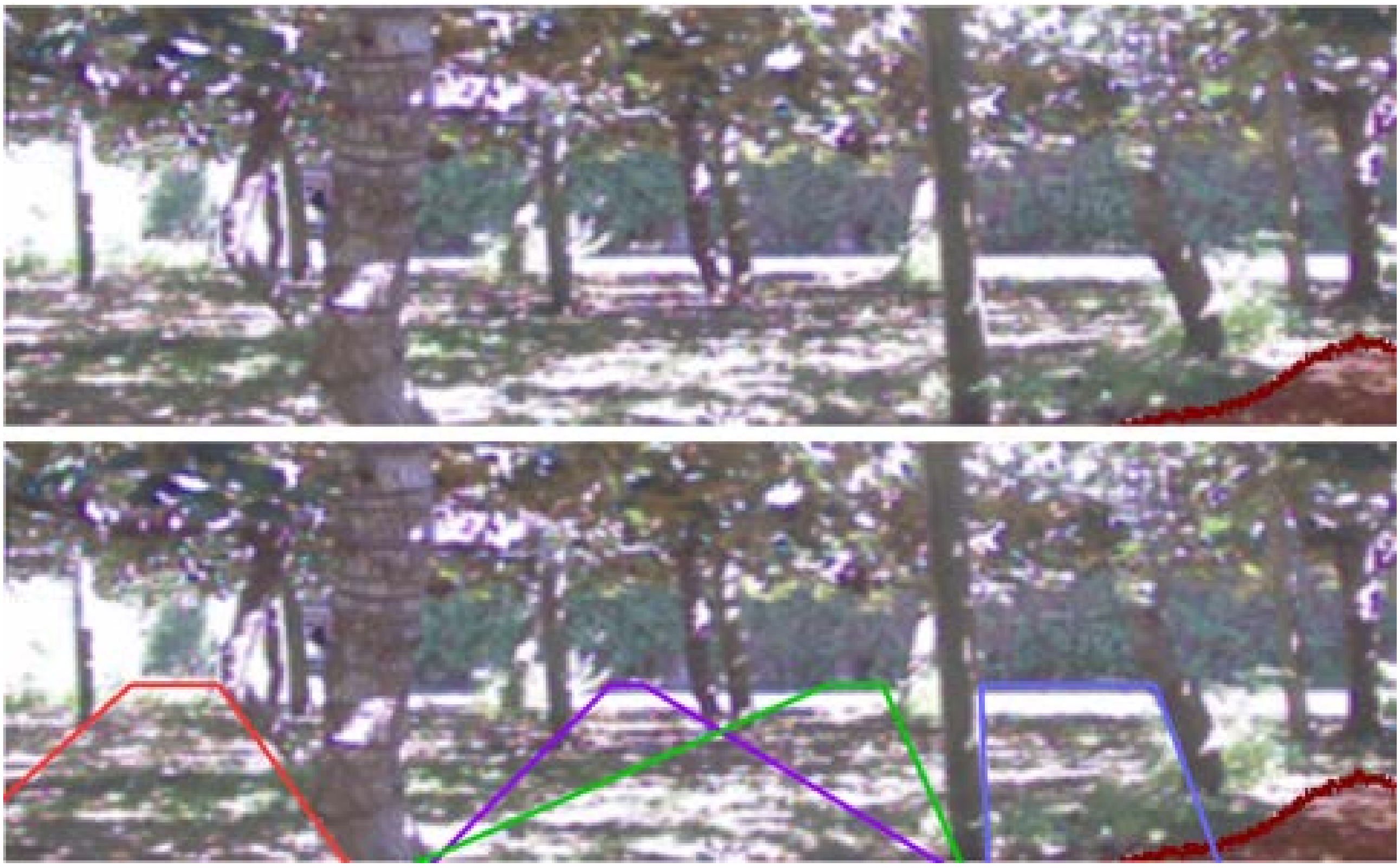

*Figure 169: An inference result where only a small area of the current row is visible and is mostly segmented (top), even though multiple feasible false rows are visible as outlined (bottom).*

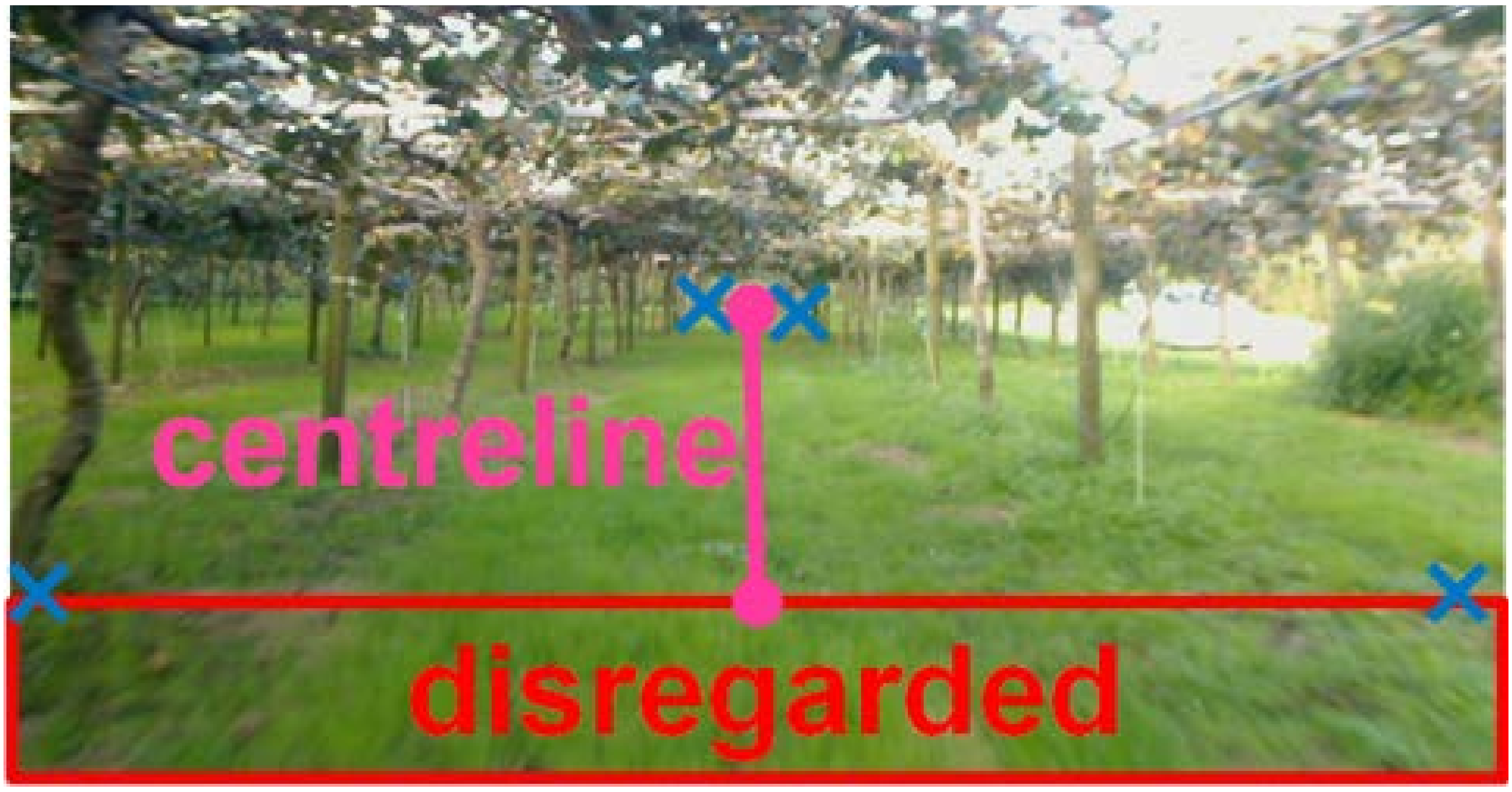

*Figure 170: Illustrating how the defined row centreline starts where both sides of the row are visible in the image and ends at the furthest visible point in the row.*

In order to convert the segmented traversable space into a command for a mobile robot, the first step used was to extract row centre points. As shown in Figure 170, the two row centre points extracted were:

- The row centre at the furthest visible point in the row.
- The row centre at the closest point, where both sides of the row were visible in the image.

To find these row centre points the segmented image was post-processed with the following steps:

1. For every y coordinate of the segmented image, the x coordinates of the start and end of the segmented region were extracted together along with the y coordinate. If more than one segmented region was found for a given y coordinate, the start and end x coordinates with the greatest difference were extracted for the next step, along with the y coordinate.
2. For every start and end x coordinate from step 1, the x coordinates and the associated y coordinate were extracted for further processing, if the following condition was met for both x coordinates, where $w_i$ is the image width:

$$0<x<(w_i-1) \tag{20}$$

3. For every start and end x coordinate from step 2, the centre of the segmented region for each remaining y coordinate was calculated as the average of the start and end x coordinates. An example of the output of this step is shown in Figure 171.

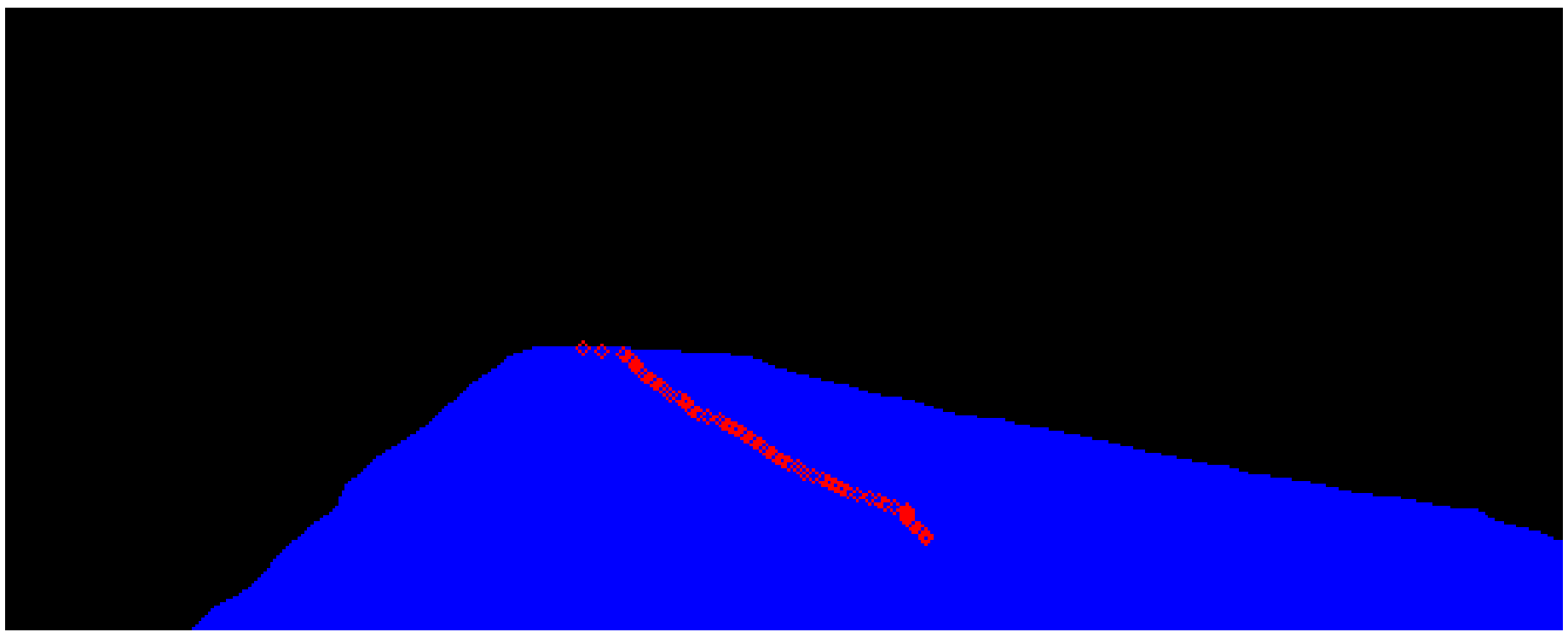

*Figure 171: The segmented output (blue) with the centre points used for centreline calculation (red).*

4. The five centre points from step 3 with the lowest y coordinates were extracted. The median of the x coordinates and the median of the y coordinates of these centre points was taken to be the row centre point at the furthest point in the row.
5. The five centre points from step 3 with the highest y coordinates were extracted. The median of the x coordinates and the median of the y coordinates of these centre points was taken to be the row centre point at the closest point in the row. An example of a row centreline formed by drawing a line between the closest and furthest row centre points is shown in Figure 172.

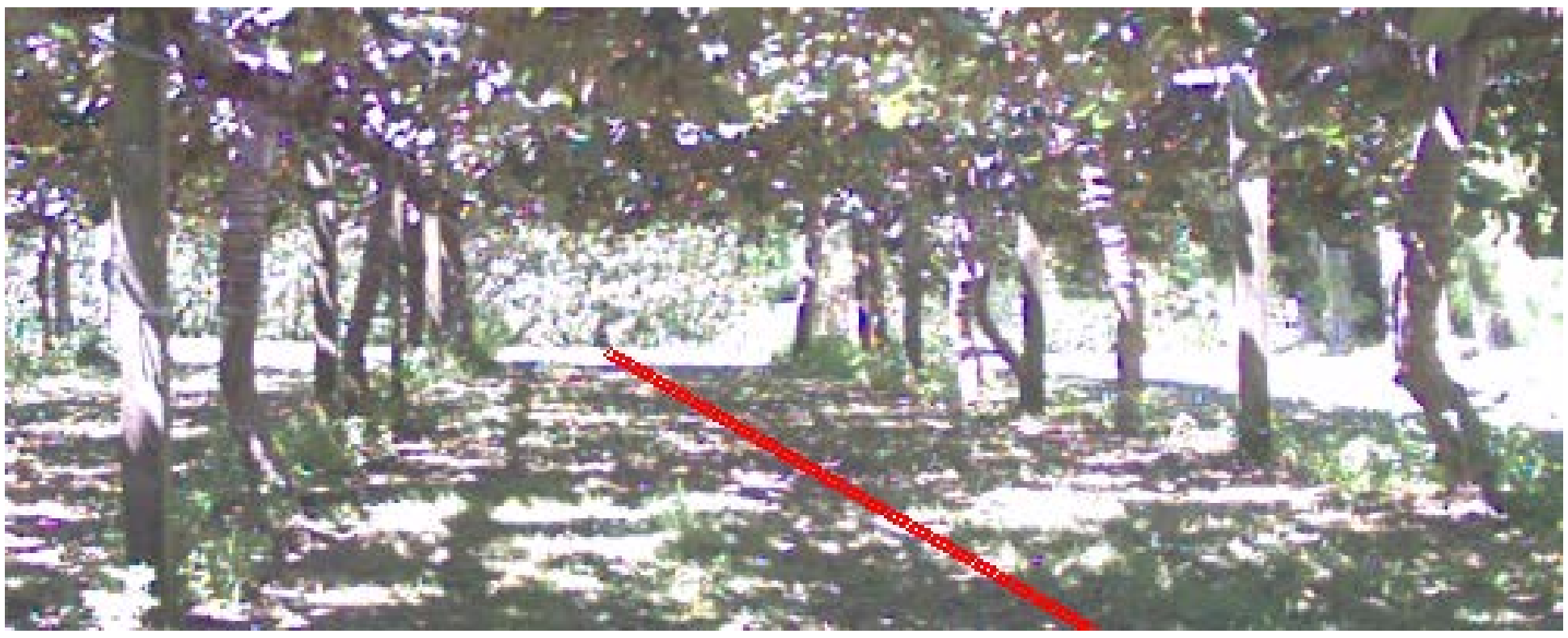

*Figure 172: The estimated row centreline (red) drawn onto the input image.*

To use the calculated centreline to control the robot for row following, it was assumed that the camera was in the centre of the robot and facing directly forwards. Hence, the goal for driving was to have the centreline intercept the bottom of the image near the horizontal image centre and this centreline had to be close to vertical. To develop the algorithm for row following, a sequence of images from a camera on a robot was collected, while a person was driving the robot by remote control and performing a row following manoeuvre from an extreme initial linear and angular offset. The images were studied and it was observed that the behaviour of the human driver was to:

1. Prioritise correcting the linear offset by driving towards the middle of the row.
2. Correct the angular offset after the linear offset was mostly corrected.

In a similar way, the control algorithm for driving the robot during autonomous row following used the following steps:

1. The difference between the image horizontal centre x coordinate, $x_c$, and the closest centre point x coordinate, $x_f$, was calculated.
2. If the magnitude of the difference from step 1 was greater than a threshold, the steering command, $\omega_s$, was set based on this difference, using a positive constant, $k_f$, according to:

$$\omega_s = k_f (x_c - x_f) \quad (21)$$

3. If the magnitude of the difference from step 1 was less than the threshold, the steering command was set based on the difference between the closest centre point x coordinate and the furthest centre point x coordinate, $x_e$, as well as a positive constant, $k_e$, according to:

$$\omega_s = k_e (x_f - x_e) \quad (22)$$

This algorithm can be thought of as a two stage tracking procedure. At first the robot aims for the centre of the row in the foreground and then the robot aims for the centre of the end of the row; this method is referred to as "*two stage tracking*" in the rest of this section. In addition, a variant of this algorithm was created that did not perform step 3. This variant effectively only aimed for the centre of the row in the foreground and hence is referred to as "*foreground tracking*" in the rest of this section.

The computer vision row following system was tested in the real world using a Clearpath Husky [96] robot with a single Basler Dart daA1600-60uc [92] camera mounted front and centre at a height of 0.8 m off the ground (Figure 173). The camera had 1600 pixels wide by 1200 pixels high resolution and the lense used had a 5.5 mm focal length. A laptop with a Nvidia M5000M [203] GPU was used for the processing. With this hardware the algorithm ran at approximately 5 frames per second with the image resized by a scaling factor of a third and cropped to 500 pixels wide by 200 pixels high. To measure the pose of the camera with respect to the orchard row, a Velodyne VLP-16 [130] 3D lidar was mounted above the camera.

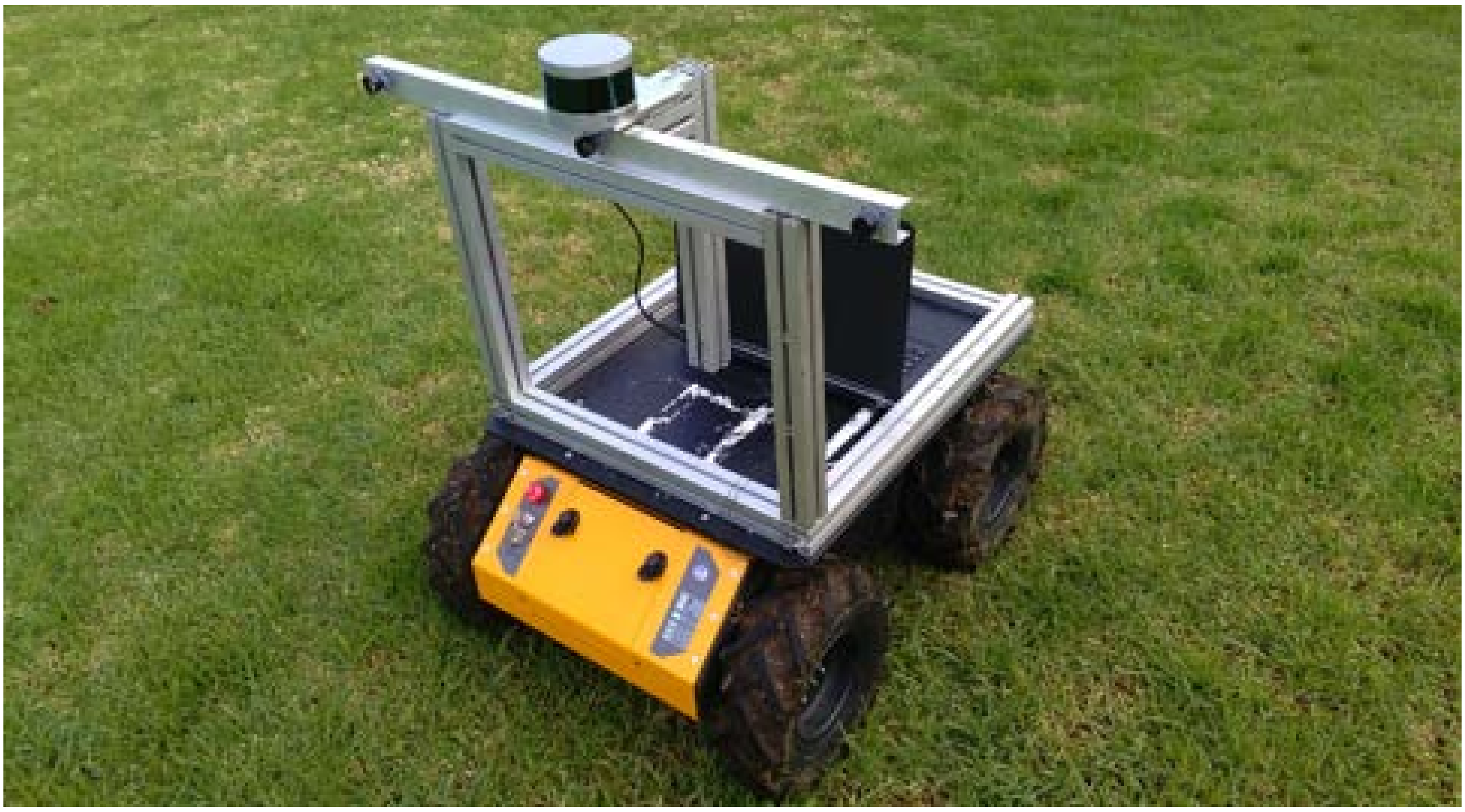

*Figure 173: The robot used for testing computer vision row following with a centrally mounted forward facing camera and a Velodyne VLP-16 3D lidar above the camera- additional cameras shown were not used for the row following testing.*

The values of the thresholds and steering constants, $k_e$ and $k_f$, were manually tuned by running the computer vision row following system on the robot and changing the values so that the robot corrected large linear and angular offsets, which were created by manually setting the initial pose.

After the tuning was completed over 600m of row following using the computer vision system was successfully completed.

During testing, the robot was started with an initial pose, which was marked by pegs next to the wheels. The initial pose was set so that the robot had a large linear and angular offset with respect to the row centreline. The computer vision based row following system was run until the robot passed a marker 50 m from the initial start pose. The speed of the robot was set at 1 $ms^{-1}$. While the row following system was running, the 3D lidar was collecting data so that the pose of the robot during the test run could be determined in post processing.

The test procedure was repeated for both variants of the computer vision row following algorithm. After both computer vision row following systems had been run for a given start pose, another row following algorithm was run from the same start pose. This third row following algorithm was used to compare with the computer vision based row following systems. The row following algorithm used for this comparison was the lidar row following algorithm, described in Subsection 4.4.1. As with the computer vision row following systems, when using the lidar row following system, the 3D lidar was collecting data so that the pose of the robot could be measured with post processing.

The pose of the robot during testing was measured by defining the robot coordinate system as being the same as the coordinate system of the 3D lidar and measuring the pose of the lidar with respect to the row. The linear and angular offsets of the lidar from the row centreline were determined by hand labelling the lidar data, using the following steps:

1. Each frame of lidar data was read in one at a time and displayed in a purpose built application with a bird’s eye view of all of the lidar points (Figure 174).
2. A person clicked one point on the treeline on one side of the current row and then another point on the treeline on the opposite side of the current row. The coordinates of these two points were averaged to give a point on the centreline of the current row.
3. Step 2 was repeated so that there were two points from the centreline of the current row, $(x_{c1},y_{c1})$ and $(x_{c2},y_{c2})$.
4. The angle between the two centreline points was calculated for the angular offset.
5. The linear offset was calculated according to:

$$\left(y_{c1}-x_{c1}\left(\frac{y_{c2}-y_{c1}}{x_{c2}-x_{c1}}\right)\right)\cos\left(atan\left(\frac{y_{c2}-y_{c1}}{x_{c2}-x_{c1}}\right)\right) \tag{23}$$

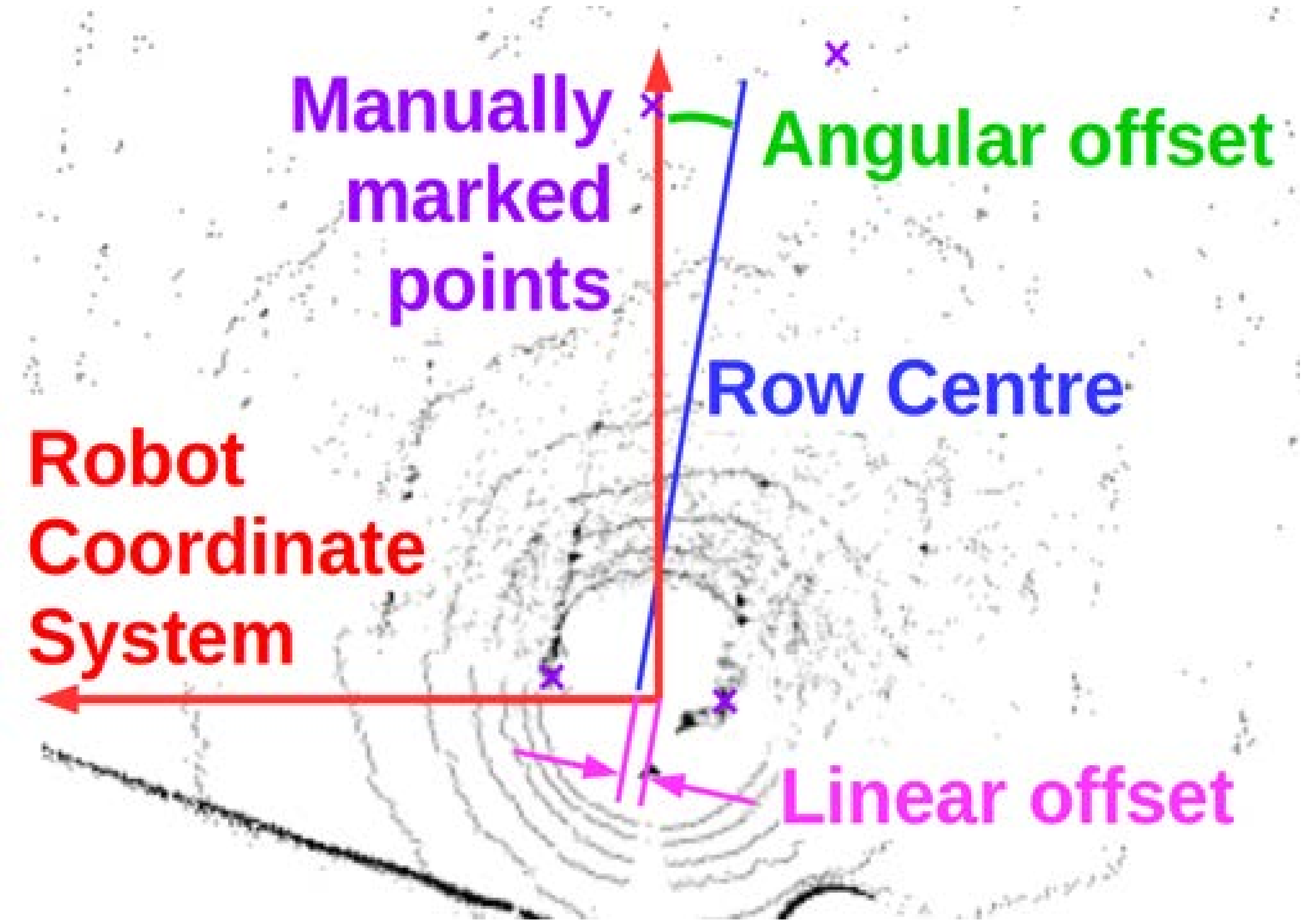

*Figure 174: Demonstrating the labelling procedure to find the linear and angular offsets of the robot from the row centreline, using points clicked on both sides of the current row.*

The linear and angular offsets measured during row following test runs with the different algorithms are given in Figure 175. These results are summarised in Table 35. From these results it seems that the two stage tracking method performs better than the foreground tracking method and the 3D lidar row following method; although the results for all methods are similar. The desired result for the computer vision row following system was that it should be at least as accurate as the 3D lidar row following method so that cross checking can be performed. In this respect, it seems that the computer vision row following system is satisfactory.

*Table 35: Results from experiments comparing computer vision and lidar row following algorithms.*

| | **Mean Absolute Offset** | |
|---|---|---|
| **Algorithm Description** | **Linear (m)** | **Angular (rad)** |
| Camera Foreground Centreline Tracking | 0.3 | 0.07 |
| Camera Two Stage Centreline Tracking | 0.2 | 0.06 |
| 3D Lidar Row Following | 0.3 | 0.07 |

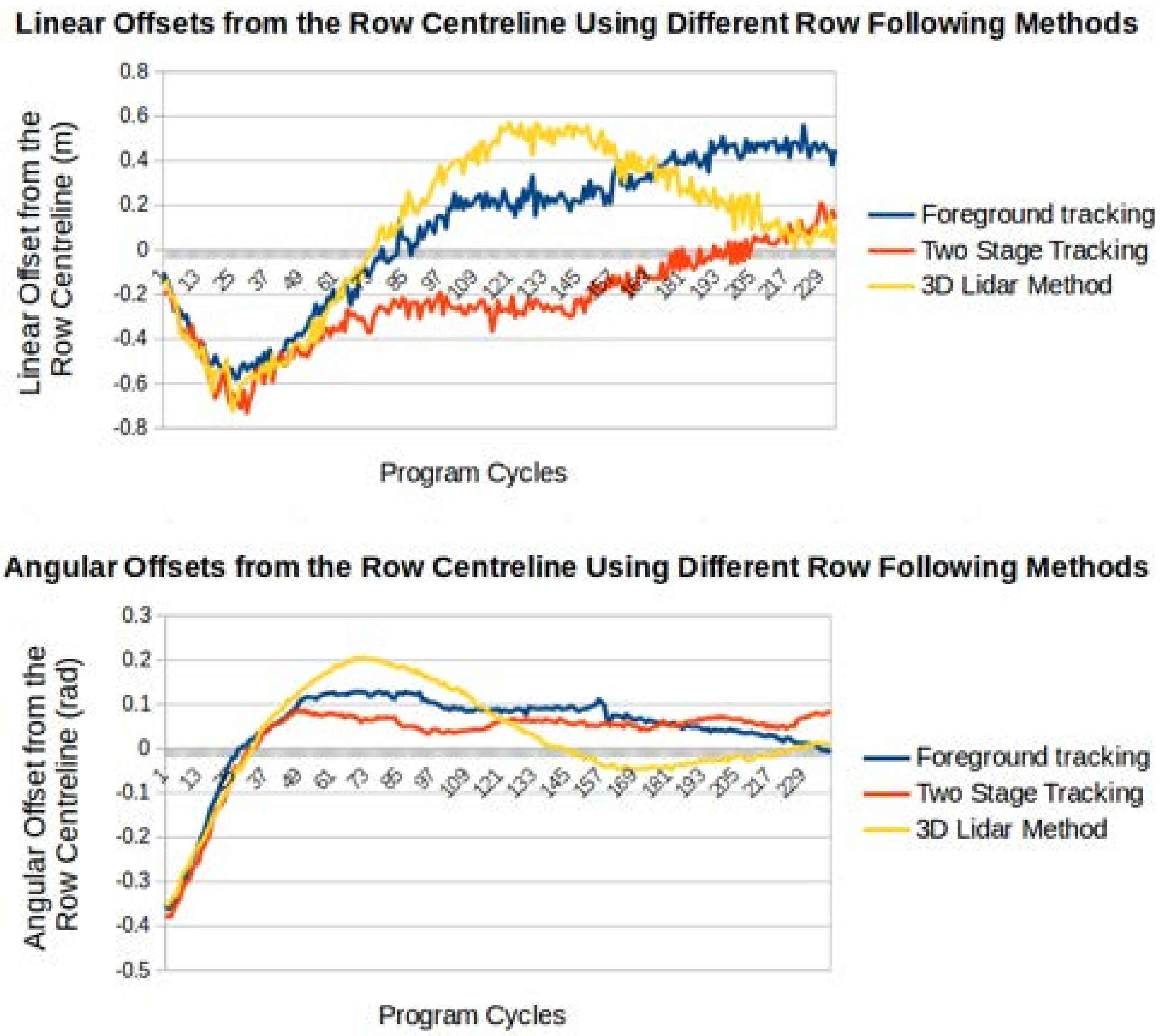

*Figure 175: The linear and angular offsets measured from test runs of different computer vision and lidar row following algorithms.*

### 4.7.9 Computer Vision Row Following Discussion

The computer vision row following system using semantic segmentation was demonstrated to follow rows for over 600 m of driving. In addition, in experiments the computer vision row following system seemed to perform as well as the 3D lidar row following system. The approach of row detection by post-processing the output of semantic segmentation may not seem as elegant as row centreline regression or other possible methods, such as that described in Section 10. However, the important metric for the MBIE Multipurpose Orchard Robotics project was that the computer vision system should produce linear and angular offsets on par with the 3D lidar row following method and this was the case. Only a monocular camera was used, under the assumption that the camera was centrally mounted and facing forwards. However, stereo vision systems without this assumption but using similar methods would be feasible also; although, a monocular system does have advantages over a stereo system in terms of hardware costs, computational costs and because if any of the cameras in a stereo system were to fail then the whole system could fail. In such a failure case, the navigation system might revert to using a method such as that presented here.

The fairness of the row following experiment with the computer vision and 3D lidar methods was limited somewhat by the extent of the tuning of each tested method. It may have been that one method was not tuned as well as another and this could have affected the results. The 3D lidar row following method was tuned over many months of development and testing, whereas the computer vision based systems were tuned for less than two hours. It is possible that if the same amount of time was spent for tuning both row following systems, the computer vision system might significantly outperform the lidar row following system. If the computer vision system was superior, it might be explained by the convolutional neural network performing better feature extraction or it could be because the camera data has a higher resolution and colour information.

Another factor that may have favoured the 3D lidar row following system in the experiments was the use of the lidar as the sensor for navigation and the sensor for the ground truth. As a result, the computer vision system may have been at a disadvantage since there would have been some pose difference between the camera and ground truth sensor, which may have caused systematic errors in the row following behaviour with the cameras; whereas, the lidar system did not have any such pose difference.

Alternatives to the use of the hand labelled 3D lidar ground truth for the row following experiments were considered. One of these options was using surveying equipment such as a Trimble Total Station [204], which has a fixed surveying station, which can track the position of a prism. Such a piece of equipment could be used to survey posts and trunks and could then be used to track a prism on the robot. However, this equipment is expensive and it was not deemed to be close to cost effective for the results it could give.

Hand labelling for ground truth seemed like a reasonable choice since people can recognise the pattern of the orchard from the data and hence can mouse click the features that belong to the current row. The 3D lidar accuracy seemed sufficient for the ground truth since the accuracy of the lidar measurement at 0.03 m was less than the variation in the features that define the row; for example, this accuracy is an order of magnitude less than the variability in the kinks of the kiwifruit trunk shapes and less than the post variations from vertical. The benefit of hand labelling for ground truth was that it added no additional equipment cost. However, it was time consuming and this issue restricted the number of ground truth measurements generated.

## 4.8 AMMP Navigation Conclusions

Four main approaches were presented for navigation in kiwifruit orchards, which were 2D lidar free space angle search, 3D lidar row following, camera row following and 3D lidar feature extraction with particle filter based localisation and mapping methods. The conclusions drawn for these methods were:

- The 2D lidar free space angle search worked well in tree-wall structured orchards.
- The 3D lidar row following method used multiple steps to remove data from the ground, weeds and canopy to form a robust method that drove for over 30 km.
- Extracting vertical objects and plane selection were feature extraction methods for 3D lidar data, which allowed particle filter localisation and mapping methods to be used in kiwifruit orchards; however, this is an area that will require future work to develop a robust navigation system.
- The best of the camera row following methods presented here performed segmentation of the traversable space and then fitted a line to the segmented result.

# 5 AMMP Safety

The AMMP is a large and powerful mobile robot with a fully loaded mass of 1.9 tonnes and a top operational speed of 3 $ms^{-1}$. If the AMMP was to impact or crush a person, it could cause serious harm. In addition, the AMMP carries modules with moving parts, including the pollination and harvesting components. As with the AMMP, these moving parts could impact or crush body parts and hence cause serious harm to people.

Steps had to be taken as a part of this research in order to reduce the risk of hazards occurring during the operation of the AMMP. The approach taken in this research was to use the method given by ISO 13849-1:2015 [1], which is a general machine safety standard. The approach that this standard describes is to follow a risk assessment and risk reduction process. The risk assessment and risk reduction performed for this research is included here because it is a contribution that may be used as a reference by others performing similar development for similar machines [1].

During the term of this research, there was no established specifically applicable standard to follow for the development of the safety system for the AMMP. The lack of an established standard prompted a review of Automated Guided Vehicle (AGV) standards in relation to the AMMP [205], since standards for AGVs are well established and the AMMP is basically an AGV that is designed to operate in orchards. Key outcomes of this review were that existing autonomous industrial vehicle safety systems were not sufficient for the variability of the orchard environment. This work also concluded that the most appropriate approach was to use a generic machine safety standard, such as ISO 13849-1:2015 [1] to design the safety system for the AMMP.

ISO 18497:2018 [206], which aims to deal with the safety of highly autonomous agricultural machines, was under development during and was released after the MBIE Multipurpose Orchard Robotics project. However, even this standard states that a risk assessment and risk reduction process should be used.

## 5.1 Risk Assessment for the AMMP

The risk assessment presented here is an abbreviated form that is focused on hazards that relate to human to robot interaction with the AMMP. In particular, the hazards related to autonomous driving are considered, as opposed to the hazards present while the AMMP remains stationary. The stages of a risk assessment using the process from ISO 13849-1:2015 [1] are:

1. Determination of the Limits of the Machinery.
2. Hazard Identification.

3. Risk Estimation.

4. Risk Evaluation.

Each of these stages is presented in turn in the following subsections. The outputs of this process are the hazards and the risk reduction required for the hazards. The required risk reduction is expressed as a Performance Level ($PL_r$), as defined in ISO 13849-1:2015 [1].

### 5.1.1 Determination of the Limits of the Machinery

The purpose of the Determination of the Limits of the Machinery is to provide information for the later stages of the risk assessment. The machine, its operation and its operating environment should be described so that it is apparent what hazards and risks relate to the machine. This information is provided in Table 36.

*Table 36: The Determination of the Limits of the Machinery for the AMMP.*

| Parameter | Value/ Description |
|---|---|
| AMMP Width | 2.2 m |
| AMMP Length | 3.8 m |
| AMMP Gross Mass | 1900 kg |
| AMMP Top Operating Speed | 3 $ms^{-1}$ |
| AMMP Track Width | 1.8 m |
| AMMP Wheelbase | 2.7 m |
| AMMP Turning Radius | 3.2 m |
| Maximum Gradient | 20° |
| Operating Environment | The operating environment is the rows, headlands and truck loading areas of kiwifruit orchards. |
| Minimum Design Row Width | 5 m |
| Minimum Design Headland Width | 4 m |
| Level of Maintenance | Branches may be hanging from the canopy and some weeds may be as high as 0.8 m. |
| Ground Condition | The ground in the orchard can be soft and muddy. The ground can also be firm, but even so there can be pot holes, fruit, branches or weeds underfoot, which can make the ground slippery. |
| Presence of People | Most of the time, orchards are sparsely populated. However, it may be possible that people will be drawn towards the AMMP out of curiosity. It is unclear if untrained people will be around the AMMP during harvesting in a commercial deployment. |

### 5.1.2 Hazard Identification

The purpose of the Hazard Identification is to produce a list of hazards relating to the machine described in the Determination of the Limits of the Machinery. The hazards considered here specifically relate to human and AMMP interaction and are given in Table 37.

*Table 37: Hazards identified related to human to AMMP interaction.*

| Hazardous Situation | Variations to Consider | Hazard Description | Hazard Type |
|---|---|---|---|
| There is a person in the path of the AMMP. | The person may be in any pose, may be unaware of the AMMP, could be occluded by some object or may run into the path of the AMMP at the last moment. The AMMP may have started unexpectedly. | The AMMP may come into contact with the person. | Impact/ crushing |
| The AMMP moves off its path into an area, where there is a person | The AMMP may have errors in its localisation, path planning or path following algorithms. A human operator may have made an error in instructions sent to the AMMP. The AMMP may experience a hardware failure or may lose control because of environmental conditions. | The AMMP may come into contact with the person. | Impact/ crushing |

### 5.1.3 Risk Estimation

The purpose of Risk Estimation is to assign a measure of risk to the hazards. This risk measure guides the downstream engineering in terms of design and processes used in the development of the risk reduction measures. The risk estimation process adopted here is taken from ISO 13849-1:2015 [1] and is illustrated in Figure 176.

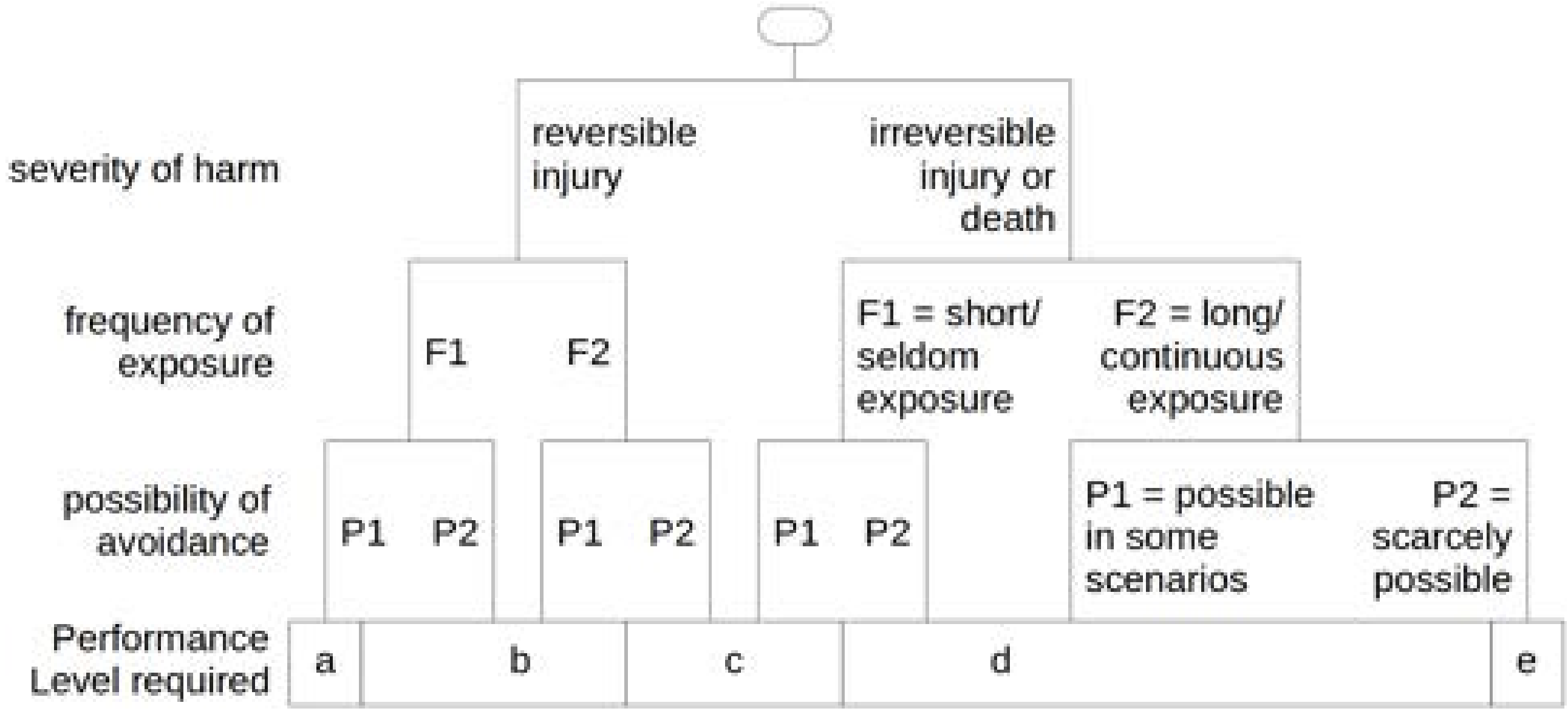

*Figure 176: The Risk Estimation process that was used in this research, based on an example from ISO 13849-1:2015.*

During risk estimation, the hazards identified were treated as a single hazard, where the AMMP may collide with a person. Two stages of risk estimation were performed. The first stage was for the MBIE Multipurpose Orchard Robotics project research, development and testing. The second stage was for an envisioned final product that would be deployed commercially. For the risk estimation for the project, the following risk parameters were chosen:

- Severity of harm: S2, serious, in the case of crushing injuries.
- Frequency of exposure: F2, frequent to continuous, because one or more people would be present and within 30 m of the AMMP during testing.
- Possibility of avoiding the hazard: P1, possible under specific conditions, since the AMMP is relatively slow, the natural tendency would be to ensure that people keep their distance and there are multiple escape paths in the pergola structure of the kiwifruit orchard, where the AMMP will be used the most.

For the risk estimation considering the commercial deployment, the risk parameters selected were different. The parameters selected were:

- Severity of harm: S2, serious, in the case of crushing injuries.
- Frequency of exposure: F2, frequent to continuous, because especially around harvest season there may be multiple people in the orchard and it is Reasonably Foreseeable Misuse that people will be curious and may approach the AMMP.
- Possibility of avoiding the hazard: P2, scarcely possible, because people in the orchard may be untrained, so they will not understand the driving behaviour of the AMMP or how aware to be around the AMMP and how to act in a hazardous situation in order to avoid harm. The AMMP may not be as supervised as it would be during development and testing.

### 5.1.4 Risk Evaluation

The risk parameters from risk estimation give a required Performance Level of *d* during the project and a required Performance Level of *e* in a commercial deployment, according to the method shown in Figure 176. These are very significant levels of risk that require the development of robust solutions. Based on these results, it was decided to employ multiple layers of risk reduction measures. It was thought that some of these layers should relate to operator procedures and training. In addition, it was thought that serious consideration should be given to limiting the exposure of people to the AMMP during a commercial deployment in order to reduce the required risk reduction from a Performance Level of *e*.

## 5.2 Inherent Safety for the AMMP

One general approach for reducing the risk associated with a robot like the AMMP would be to aim to design for inherent safety. For the AMMP this might mean having a smaller, lighter weight, slower and less powerful platform. Furthermore, only a singular and minimally powerful arm might be used for harvesting and pollination.

However, for the purposes of the work considered here, the AMMP was a large robot as detailed in Table 36 and hence designing for inherent safety was not an option. Hence, instead other engineering sensing and control solutions were investigated.

## 5.3 Survey of Safety Systems for Orchard Robots

It seemed prudent to research what other safety systems had been developed for robots similar to the AMMP. Griepentrog et al. [207] used an earlier version of ISO 13849-1 to develop a safety system for an autonomous tractor. The resulting safety system used a horizontal lidar on the front of the tractor to detect obstacles and trigger a relay in the safety circuit. A bumper acted as a back-up for the lidar. There was also a stereo camera system for further obstacle detection capability. For localisation, there was a GPS and IMU. In addition, there were monitors of the tractor states and procedural controls for improving the safety of the system. This system had limitations in that it had to be supervised and public access had to be restricted; these may not be reasonable requirements for some orchards, especially during the busy harvesting season. Furthermore, the obstacle detection system on the autonomous tractor had a range of blind-spots under the forward detection systems as well as to the sides and the rear of the tractor.

The architecture described by Griepentrog et al. [207] used multiple sensors but did not use multiple channels in order to increase the Performance Level (PL) or Safety Integrity Level (SIL) of the safety system. A more suitable architecture would use multiple channels with cross-monitoring, as in the dual channel Category 3 architecture [1] shown in Figure 177.

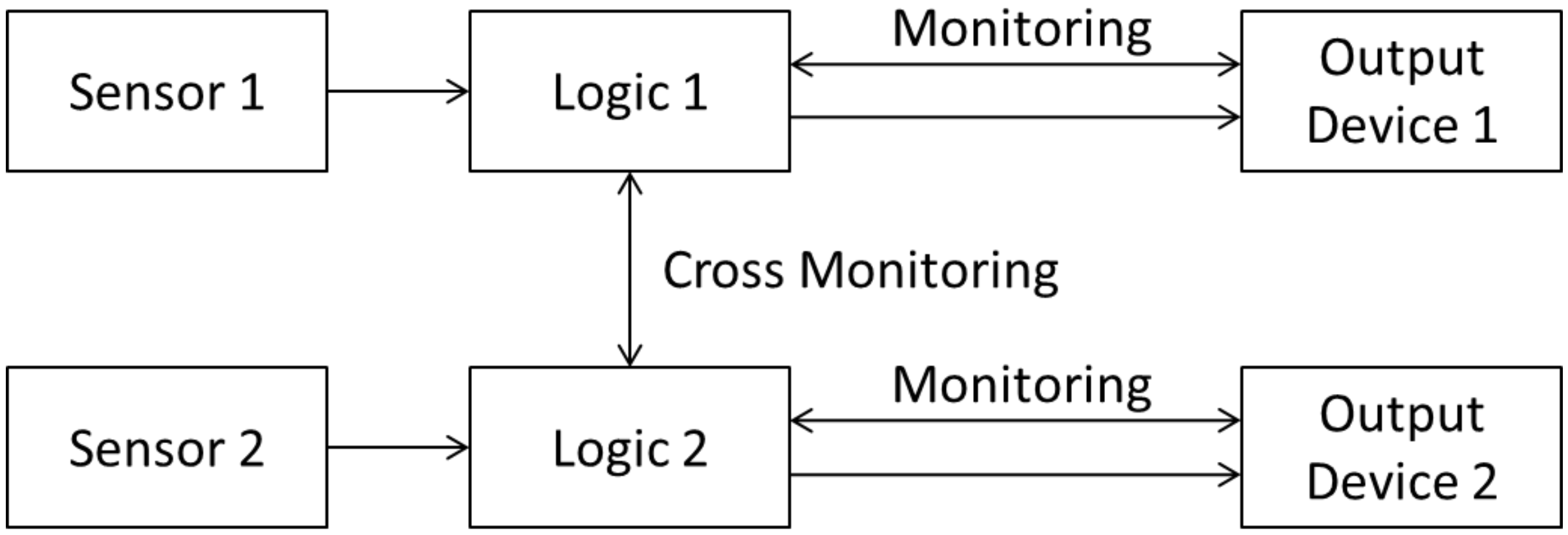

*Figure 177: A Category 3 designated architecture.*

Freitas et al. [186] used an angled down 2D lidar for obstacle detection. Potential obstacle points were defined as points that were a threshold distance or angle higher than adjacent points. Obstacle points were clustered using threshold distances. Clusters were classified based on thresholds for the numbers of points in the clusters. The experimental results showed that the length of the grass influenced the detection of smaller objects, which may be a problem in orchards or areas that are not well maintained. The method was successfully tested with people in upright poses; however, it is unclear if the method would be as successful if a person was tying their shoe laces or had fallen. Throughout the testing, the thresholds were changed; however, it is unclear if it would be practical and safe to be regularly changing the thresholds in real world applications. Furthermore, the angled down lidar configuration restricts the field of view of the lidar and reduces the number of features that can be detected for row or free space detection.

Tabor et al. [208] compared the performance of Aggregate Channel Features, Deformable Parts Models and Convolutional Neural Networks for performing pedestrian detection using camera images in orange orchards. They found that their CNN worked relatively well compared to the other methods, in the presence of occlusions and for unusual poses but was less accurate for bounding box regression for some overlap thresholds. This finding regarding bounding box regression for their CNN would not necessarily apply to all CNNs. The CNN described by Tabor et al. [208] was relatively small, with just 4 convolutional layers with pooling and leaky ReLU layers. It took in small input images of 60 pixels by 45 pixels. This classification network was applied with a sliding window at different scales and then bounding box clustering was applied. This CNN was just one instance of a CNN and did not represent the state of the art of the time or later. Therefore, it may well be that the finding that the CNN did not perform well for bounding box regression with some threshold settings is not valid for other CNNs.

Moorehead et al. [156] used a nodding lidar and cameras to perform object classification and obstacle detection. The output of the obstacle avoidance system was a safe speed. If the safe speed reduced to zero, a remote operator was required to intervene. Additional measures in place for the safety of the system included infrastructure and processes, which were designed to restrict access by people- especially when the autonomous tractors were operating. A key issue with this system was its architecture. Lidar data was used both on its own and with the camera; however, there was no channel in the system that did not use the lidar. Therefore a single fault in the lidar could cause a system fault. Because the lidar was in a nodding configuration, a fault in the nodding mechanism or control system could also cause a system fault. A more reliable architecture for this system could include a camera only classifier that might check and be checked against results from other channels, which would be running on different processors. A stated issue with the lidar processing algorithm was failure in the presence of occlusion due to weeds or other objects. Obstacle detection systems that use cameras or lidar can fail to detect objects that are partially occluded [186], [209]. This is a concern in orchards, where a person moving into the path of an autonomous vehicle may be partially occluded by trees, posts, weeds, bins, tractors or other objects.

Corke [210] discussed collision avoidance for agriculture robots and proposed avoiding something that is not classified as crop or the typical environment as one possible approach. He suggested using radar as a method for detecting objects through vegetation. This approach may be useful for detecting objects that are occluded by weeds in kiwifruit orchards.

Only a limited amount of work was found regarding pedestrian detection in orchards and so pedestrian detection in other environments was considered also. Navarro-Serment et al. [211] detected pedestrians from a moving vehicle using a 3D lidar sensor. The first step taken was to register new data to existing data points, which were organised into a grid of cells with the vehicle at the centre, while applying a higher weight to newer points and a lower weight to older points. The mean and standard deviation of data point heights in each cell were calculated. Then for each cell, a height threshold was calculated as the mean height for the cell with one standard deviation subtracted; only points above this height threshold were retained for later stages of processing at the current time-step. The resulting points were projected onto a horizontal 2D plane and ordered according to their angle, in order to mimic a 2D lidar scan. Consecutive points were considered to be on the same object if the distance between them was less than a threshold [212]. Bounding boxes were used to summarise the points in each resulting object. Bounding boxes from the previous time step were projected forward in time using the object velocity, which was estimated by a Kalman filter. Data association was then performed by searching for overlapping projected and current bounding boxes from consecutive lidar frames. The displacement between matched bounding boxes was used to update the Kalman filter. The number of iterations before an estimate and the standard deviation of an estimate were checked. Detection of pedestrians was partially performed by calculating a set of motion features, which were:

- The size of detected objects, which helped to distinguish people from larger objects, such as other vehicles.
- The distance travelled, which distinguished people from stationary objects.
- The variations in size and velocity because vegetation tends to have high variability in lidar data since adjacent measurements may, for example, detect the outer extremities and inner structures of plants.

In addition to these motion features, a set of geometric features was extracted. Calculation of the geometric features started with a Principal Components Analysis and transformation of the data for alignment to the two principal planes. Then the features calculated from the points of each segmented object were:

- The covariance matrix and the normalised moment of inertia.
- The covariance matrices of the top half, bottom left quarter and bottom right quarter of the points in the object.
- 2D histograms for each of the two principal planes.

These geometric features were the input for a Support Vector Machine (SVM). The output of this first SVM was used as input for a second SVM, along with the motion features. The output of the second SVM was used for pedestrian detection. Such work in lidar pedestrian detection has been superseded by the use of CNNs.

Li et al. [213] ordered the planes from 64 plane Velodyne lidar sensors into rows and consecutive angles into columns to form images. The resulting range images were processed using a Fully Convolutional Network (FCN) with two heads. One head was trained to detect if each pixel was a part of an object of interest in order to provide a mask for the other head, which performed regression of bounding box coordinates per pixel. Non-maximum suppression was used to select from the resulting bounding boxes. This overall method produced better results than other algorithms, when performing object detection of cars on the KITTI dataset [124], [125].

Zhou and Tuzel [214] randomly sampled a fixed number of points from occupied voxels in 3D lidar data and processed vectors of the point features through a series of fully connected, pooling, skip connection and concatenation stages. The rest of their “VoxelNet” network was fully convolutional with Batch Normalisation layers, ReLU layers and two heads for regression and classification. When evaluated on the KITTI dataset 2017 3D object detection challenges [215], the VoxelNet produced leading results for the lidar only methods at the time.

Chen et al. [216] used multiple inputs to their neural network. The primary input was a stack of 3D lidar data projected into a bird’s eye view; this input had channels corresponding to the height of points in horizontal slices of the lidar data as well as the density and the reflectance of points. There was also another separate input which was the lidar data projected into a 2D image from the sensor’s point of view; again, there were distance, density and intensity channels for this input. The other input, which was optionally used, was a camera image of the forward scene. Each input was processed separately with Convolutional Neural Networks. The output of the bird's eye view CNN had its own bounding box regression and classification heads. The resulting bounding boxes were transformed to the coordinate systems of the other inputs. Then following upsampling and ROI pooling layers, different fusion networks were proposed that ended with a 3D bounding box regressor and a multi-class classifier. The approach of fusing the different streams of data may have improved the performance of the detections; however, such an architecture would be vulnerable to Common Cause Failures, since there are no independent channels that can be used for cross-checking and redundancy.

Ondruska et al. [217] developed a recurrent neural network to perform segmentation, classification and tracking of objects in 2D lidar data. Training of the tracking began by having a current lidar frame as the latest input to the recurrent neural network and the next lidar frame was used as the ground truth label; hence, the network was trained to produce a future lidar frame from the existing lidar frames. This approach to unsupervised learning seems like it could be applied in kiwifruit orchards and possibly with 3D lidar data. The network of Ondruska et al. [217] was capable of tracking occluded objects and projecting the paths of objects in future time steps. The system was tested at a busy intersection with people, cyclists, buses, cars and static objects detected. This work was later extended to work from moving platforms by Dequaire et al. [218].

Mosberger et al. [219] detected high visibility vests using a monochrome stereo camera system with an optical band pass filter, which was matched to the wavelengths from the Near Infrared flash. Two pairs of images were taken; one image from each camera was taken without the flash and the other pair of images was taken with the flash. Blobs were extracted in the camera images. Blobs with high reflectivity were detected in the camera images using the differences in pixel values between the images with and without the flash. In addition, different feature descriptors were trialled in Support Vector Machine and Random Forrest classifiers to distinguish the blobs belonging to high visibility vests by supervised learning. This work was similar to the high visibility vest detection described in Subsection 5.4.1, which was developed independently. The benefit of the high visibility vest detection conducted in Subsection 5.4.1 was that it used a sensor that was already on the vehicle and it did not have the issue of objects moving between collections of the two pairs of images.

Leonard [220] discussed various difficulties with outdoor autonomous navigation safety in the real world. He highlighted difficult situations, including people running out in front of robots at the last moment and issues with perception reliability when there are unexpected changes in the environment, which could include different weather conditions and glare from sunlight. The case of people running out in front of a mobile robot is a long established problem that has been recognised by manufacturers of mobile industrial robots [221]. This is a problem that should be considered further in the development of orchard robots. Bad weather conditions are also a problem that should be considered for orchard robots; strategies may involve detecting the fault conditions, such as glare or water on a lense, and entering a safer state accordingly.

Urmson [222], [223] and Eustice [176] separately discussed some approaches that may be used to gain confidence in Safety Related Parts of the Control System for autonomous navigation. Both propose testing simpler scenarios to begin with to build confidence in systems before gradually advancing to specially created test environments where rare and challenging conditions are created for testing the safety systems. Applying these ideas to the AMMP might involve row following to start with, before exposing the AMMP to simulated pedestrians, simulated farm animals, farm vehicles, beehives, bins and other things that may be seen in and around an orchard.

## 5.4 Pedestrian Detection in Orchards

Pedestrian detection was performed in order to implement and experiment with pedestrian avoidance measures. Some of these measures included:

1. Slowing down or stopping when pedestrians were present. It was thought that the safety case for some applications of the AMMP in orchards might require that people could not be present in order for the operation to be safe and therefore stopping when pedestrians were detected would be an appropriate action. Similarly, slowing down might be an appropriate reaction when people were detected, because it could reduce the risk to the people. In addition, it was thought that slowing down could increase safety by reducing the stopping distance of the AMMP and allowing the Safety Related Parts of the Control System more program cycles to process data from sensors in order to detect any hazards more reliably.

2. Stopping when pedestrians were detected in fixed zones around the AMMP. This feature was a subset of the first measure, where the position of pedestrians was measured and, if a pedestrian was in a vulnerable position with respect to the AMMP, a safety stop would be triggered.

3. Logging the presence of pedestrians. When the risk assessment was performed, assumptions were made about the frequency of human to robot interactions. Logging the presence of pedestrians could allow for checking these assumptions and hence it could enable appropriate follow up actions, which could include additional safety measures, if the assumed frequency of interactions was found to be too low.

These measures required at least the detection of pedestrians. However, there were also requirements for measuring the position of pedestrians. All of the methods presented here allow for fulfilling these requirements. The methods considered for pedestrian detection and position measurement included methods using different sensors, which were repurposed from autonomous driving. There was also the reuse of software from the harvesting and pollination systems. The pedestrian detection methods were:

- Pedestrian detection using stereo camera images and an image segmentation algorithm. In this case, the position of a pedestrian was measured using stereovision.
- Pedestrian detection using camera images and an algorithm which inferred the position of the pedestrian. In this case, the position of a pedestrian relative to a zone around the AMMP was inferred using a classification neural network or by the overlap between segmented ground and segmented pedestrians.
- Pedestrian detection using neural networks and capsule networks for processing the 3D lidar data.
- Pedestrian detection using the intensity channel of a 3D lidar sensor for the detection of high visibility clothing retro-reflectors and position measurement of the pedestrian using the range measurement channel of the same sensor.

### *5.4.1 Pedestrian Detection by High Visibility Vest Reflection Intensity*

Some of the project team members, from the University of Waikato and Robotics Plus, put together rules and procedures for operating the AMMP and doing work in the orchards, in order to mitigate risks for all people around the AMMP [224]. One of the personnel protection measures stipulated for working in the orchards and around the AMMP was that everyone must wear high visibility vests (Figure 178). Besides the bright colours, a key feature of these high visibility vests is that they have horizontal and vertical retro-reflective strips.

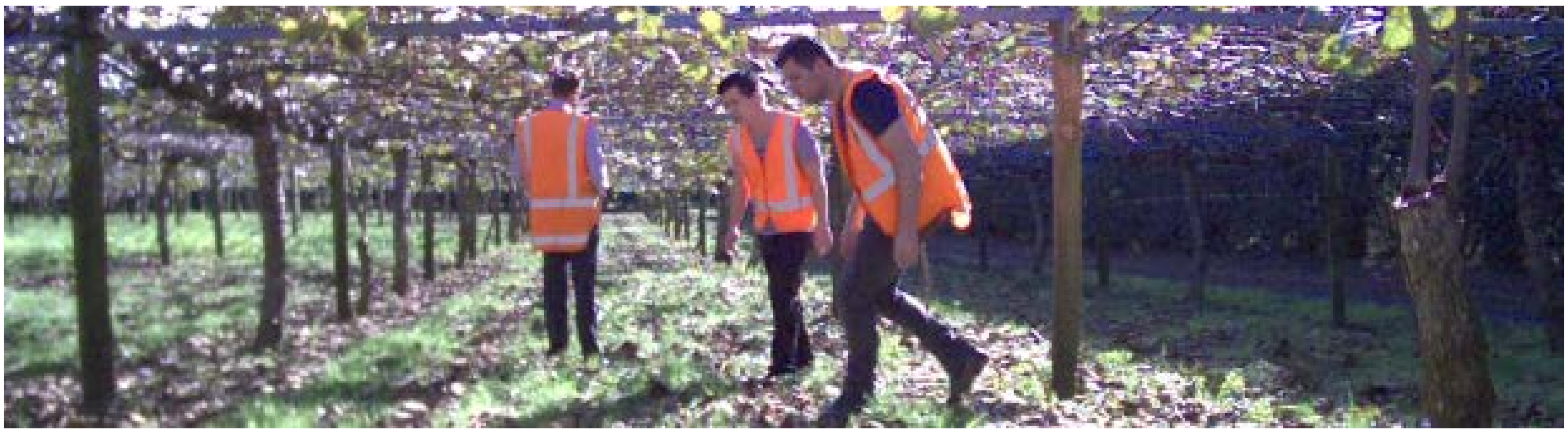

*Figure 178: High visibility vests used in the orchard around the AMMP.*

When viewing the intensity channel of the data from a Velodyne VLP-16 [130] 3D lidar, the reflective material stands out with high values, as shown with the lighter grey areas to the left in Figure 179. These high values are partly due to the higher intensity of the reflections from the reflective material. In addition, the Velodyne VLP-16 intensity data is structured so that values from 0 to 100 are reserved for the intensity of diffuse reflections and readings from 101 to 255 are reserved for values of reflected intensity from retro-reflectors. Note that Figure 179 is the raw intensity data, which has been stretched vertically for easier visual inspection.

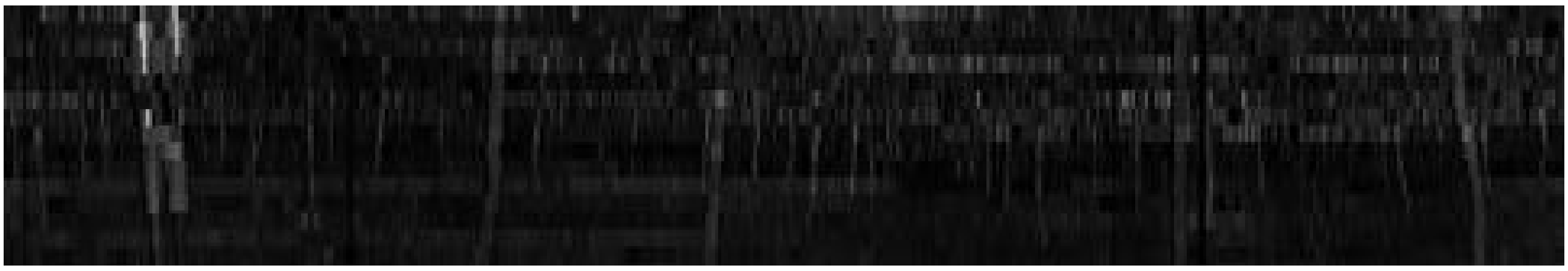

*Figure 179: Intensity data from a Velodyne VLP-16 sensor in a kiwifruit orchard, with a person wearing a high visibility vest to the left.*

Two general approaches were considered for detecting the pedestrians and high visibility vests from the 3D lidar data. It seemed that the range and intensity data could be used as a two channel input for an object detection or segmentation CNN, such as DetectNet [44], FCN-8s [45] or Mask R-CNN [48]. However, given how much the high visibility vests stand out in the intensity data, it was thought that it may be possible to create a reliable hand crafted algorithm also. For the hand crafted algorithm for high visibility vest detection, the following steps were used:

1. The intensity data was thresholded. The threshold used was for values greater than 100, since this is the reserved range for retro-reflectors for the Velodyne VLP-16 [130] sensors used. An example thresholded image is shown in Figure 180, which is the thresholded result of Figure 179. Note that Figure 180 is stretched vertically for ease of viewing; however, the processing was performed on unstretched data in order to avoid redundant processing.

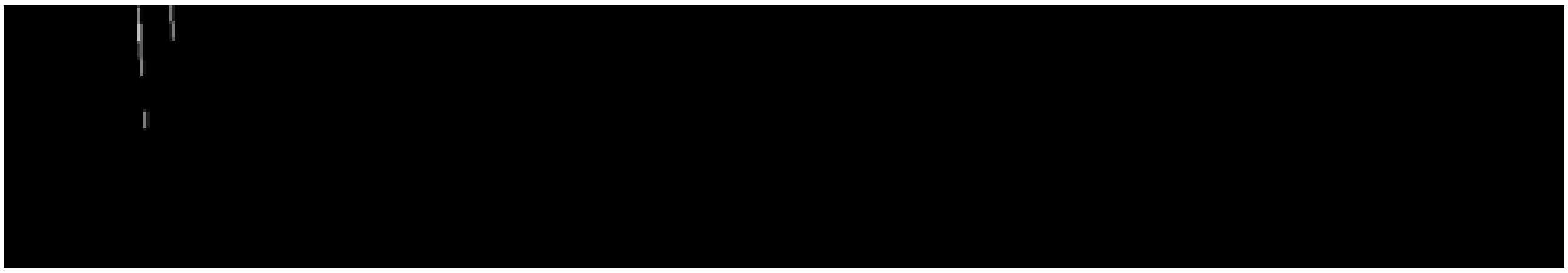

*Figure 180: Intensity data of a 3D lidar, thresholded to leave the data from the retro-reflective strips on a high visibility vest.*

2. It was found that the range measurements from the high visibility vest reflective strips were often missing from the data; when this was completely the case, in order to measure the position of the high visibility vests, the points around the reflective strips were used. To do this a mask was created by setting values above zero to the maximum pixel value and then performing dilation, as shown in Figure 181, which is the result of processing Figure 180.

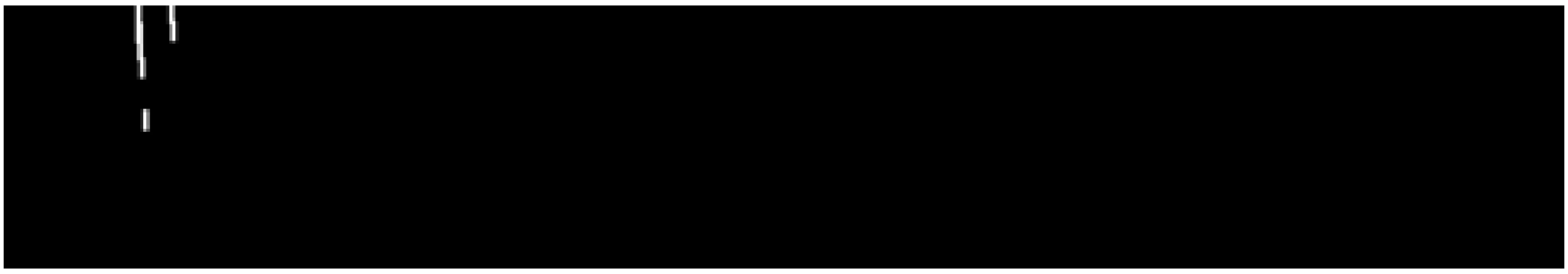

*Figure 181: Mask for extracting high visibility vest range data, created by thresholding 3D lidar intensity data to a binary image and then performing dilation.*

3. The mask created in step 2 was used to extract the corresponding 3D coordinates of points, which were band pass filtered based on percentile thresholds of the ranges. However, if range values corresponding to the high visibility vest reflective strips were present then these points were extracted instead.
4. The 3D coordinates from step 3 were checked against a set of boundaries, making up a volume around the sensor. If a detected retro-reflective point was found to be inside this volume, deceleration behaviour was triggered.
5. Step 4 was repeated for a smaller volume in order to trigger emergency stopping behaviour.

Testing was initially conducted with people attempting to trigger an emergency stop by walking in close proximity to the AMMP with high visibility vests on when the AMMP was stationary and also by throwing high visibility vests at the AMMP, with the AMMP moving. This initial testing indicated that the algorithm seemed to work well, with the AMMP emergency stopping 50 times out of 50 triggering attempts.

However, there was a question as to how many false positives could be produced by objects other than high visibility vests in the kiwifruit orchard environment. To answer this question, the AMMP was driven around an orchard and the points in the intensity data above the retro-reflector threshold value of 100 were counted. Across over 20,000 frames of lidar data, there were zero retro-reflectors detected by the lidar. There were high intensity values, close to 100, from leaves that came close to the lidar; however, there were no false positives for retro-reflectors in the orchard. It was postulated that perhaps steel beams, wires in the canopy or wet surfaces around the orchard might cause false positives; however, this was not the case.

The Velodyne VLP-16 [130] 3D lidar, as it was used for pedestrian detection, had vertical resolution of only 16 layers, separated by 2 degrees, compared to the horizontal resolution of 450 slices, separated by 0.8 degrees. Hence, the vertical retro-reflector strips were more likely to be detected than the horizontal strips on the high visibility vests. However, even the vertical reflector strips had a limited range over which they were likely to be detected. Assuming that half of a lidar beam reflecting from a retro-reflector would be enough to register a retro-reflector in the lidar data, the maximum range for guaranteed detection, $r_d$ (Figure 182), of an unoccluded vertical retro-reflector facing the lidar with width, $w_s$, from a lidar with horizontal angular resolution, $\alpha_l$, would approximately be:

$$r_d = \frac{w_s}{2\tan(0.5\alpha_l)} \tag{24}$$

With a retro-reflector width of approximately 0.05 of a metre, this equation predicts a guaranteed detection range of approximately 3.6 metres. This range is more than sufficient for the AMMP, which has a maximum stopping distance of 1.4 m, including the distance travelled during the reaction time of the safety system. For other applications with a longer maximum stopping distance, options include using a higher angular resolution. Furthermore, to make the high visibility vest detection more robust, using wider or more retro-reflector strips may be a useful option.

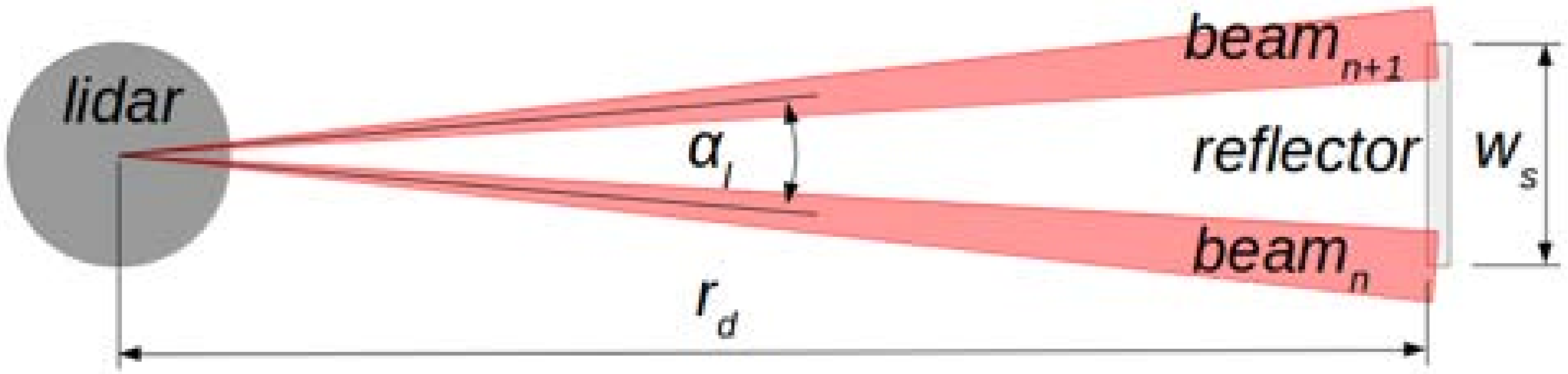

*Figure 182: A sketch of a lidar sensor and reflector from a bird's eye view, illustrating consecutive beams from the lidar with half of the beams incident on the reflector.*

A test dataset was created from frames of 3D lidar data. In each frame of data the outlines of pedestrians were segmented and classified manually. The distance to the closest point on each pedestrian was measured using the range channel of the data. The dataset included 863 instances of pedestrians wearing high visibility vests at ranges from 1 m to 8 m. Pedestrians beyond a distance of 8 m were difficult to positively identify as pedestrians and difficult to segment because there were so few data points from the pedestrians at the longer distances. The high visibility vest detection was tested with a true positive counted for each segmented pedestrian where there was a pixel above the high visibility vest detection threshold in the intensity channel of the data. Segmented pedestrians without pixels above the high visibility vest threshold were counted as false negatives. As for the previous survey of lidar data, there were zero false positives for reflective material from orchard objects. The results of the testing on the test dataset are shown in Figure 183.

These results closely match the calculated guaranteed detection range of 3.6 m. Past this distance, the detection rate drops, which is also consistent with equation 24.

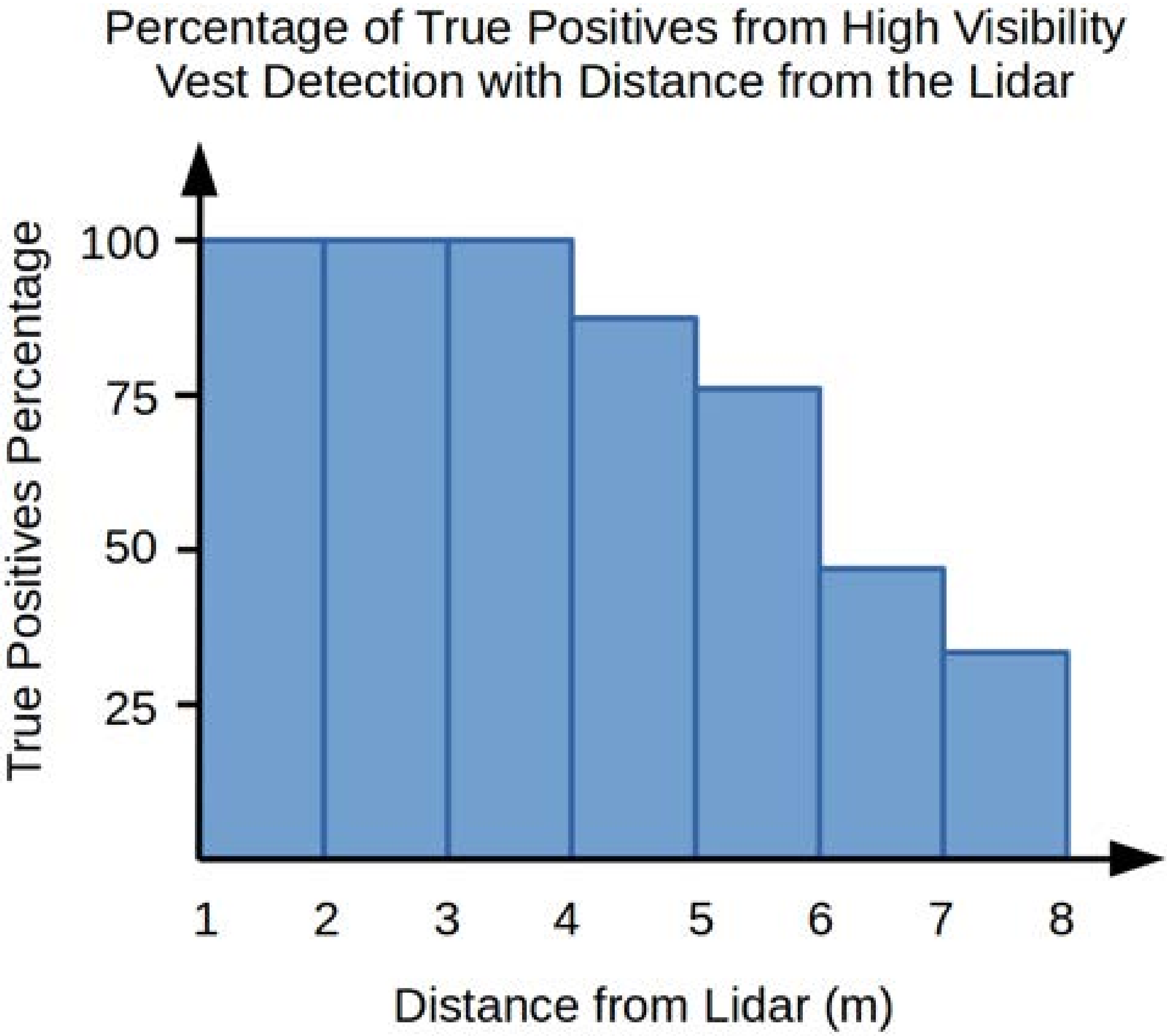

*Figure 183: The performance of the high visibility vest detector with distance from the 3D lidar.*

### 5.4.2 Lidar Segmentation by Thresholded Range Difference

It was thought that additional methods should be developed for lidar pedestrian detection and that segmenting objects in the lidar data would be an important step to achieve this goal. A hand engineered algorithm was created based on the hypothesis that in the sparse array of trunks and posts in kiwifruit orchards that adjacent lidar data points on the same object would have similar ranges. Also, it was thought that in kiwifruit orchards, in particular, adjacent lidar data points on different objects would generally have dissimilar ranges. However, it was postulated that there could be important exceptions to this, such as a person standing close to a trunk or post.

Nevertheless, it was decided to process the lidar data to cluster adjacent points that had similar ranges. The high level steps for this processing were:

1. The data was thresholded based on the magnitude of the range data.
2. Missing range data from retro-reflectors were filled in using surrounding pixels.
3. Searches were conducted in the lidar data for clusters with adjacent points that had similar ranges.
4. Clusters were eliminated if the dimensions of the objects were below thresholds.

The data was thresholded based on the range because points that were too far away were not of interest for pedestrian detection- partly because a pedestrian far away did not represent an immediate hazard and partly because it was difficult to recognise a pedestrian that was far away in the lidar data, since objects further away had less data points. In addition, only considering a limited range and less data reduced the processing time. Figure 184 shows a frame of lidar range data and Figure 185 shows the result of applying a threshold of 10 m to the data from Figure 184.

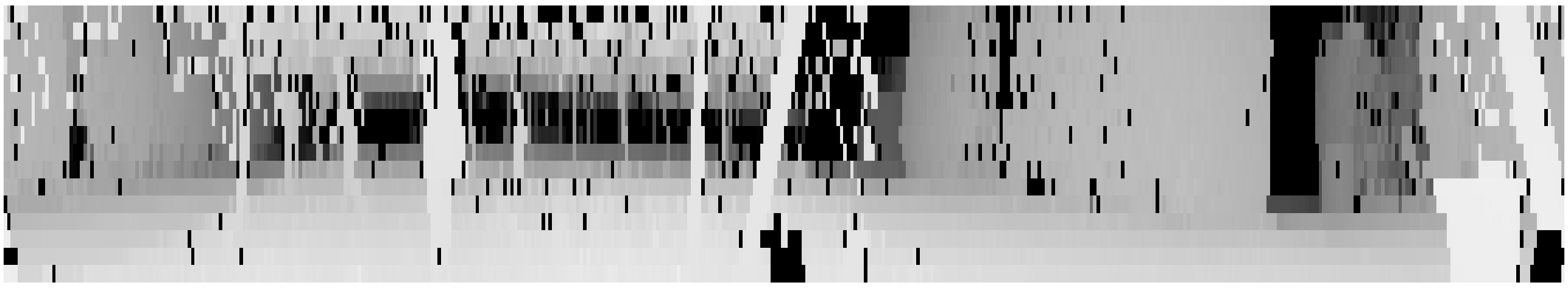

*Figure 184: A frame of range data from a 3D lidar, where closer pixels are brighter and the image has been stretched vertically for visualisation.*

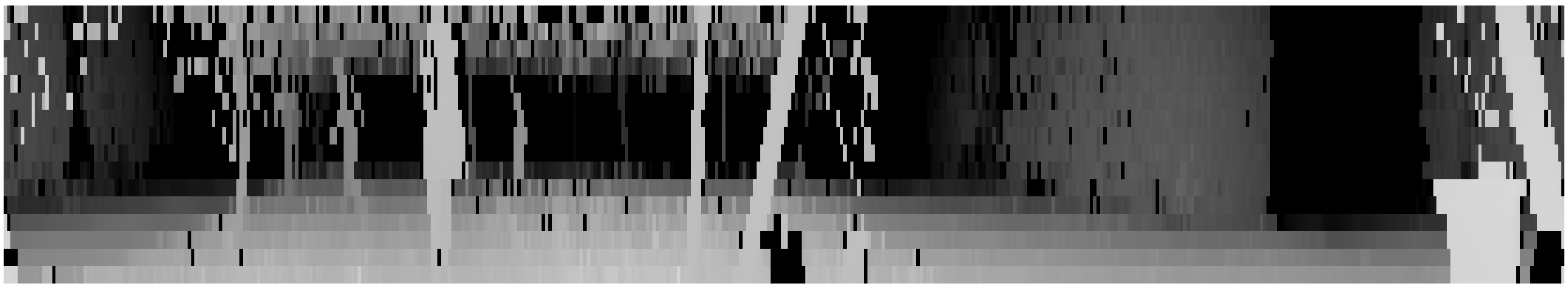

*Figure 185: The range data of Figure 184, which has been thresholded with a value of 10 m and with the contrast increased for visualisation.*

Retro-reflectors were detected in the intensity data using a threshold of 100, as for the high visibility vest detection. An example of this step is shown with the intensity data in Figure 186, thresholded to give the result in Figure 187. Some of the retro-reflector points in Figure 187 correspond to empty range data in Figure 184. It was found that having such points empty could affect the segmentation so a step was added to artificially fill these points. For each retro-reflective point, if the corresponding range value was zero, the range value was filled by the median of the surrounding range values. An example of performing this step is shown in Figure 188.

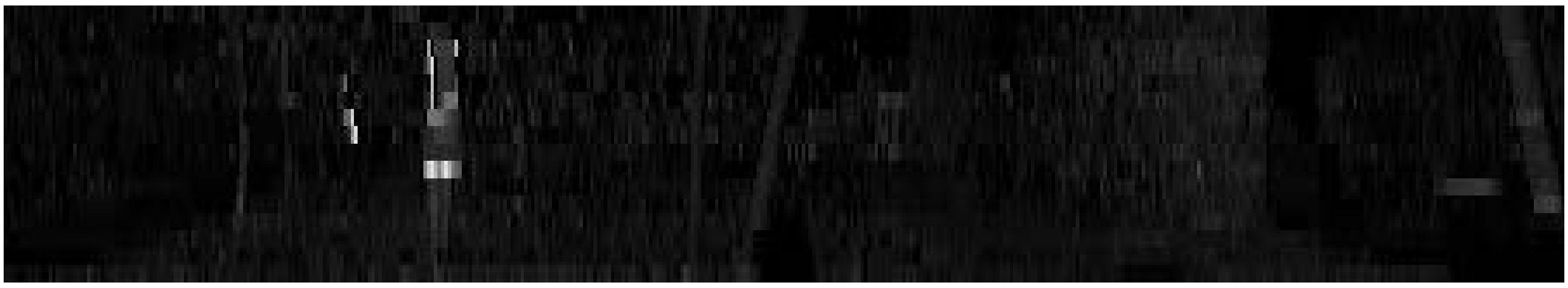

*Figure 186: The intensity data, corresponding to Figure 184.*

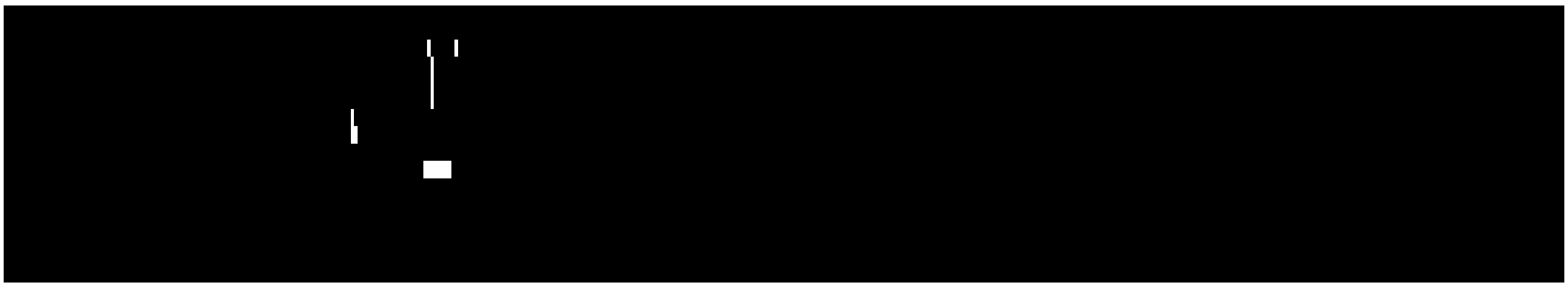

*Figure 187: The intensity data of Figure 186, thresholded to highlight the retro-reflectors.*

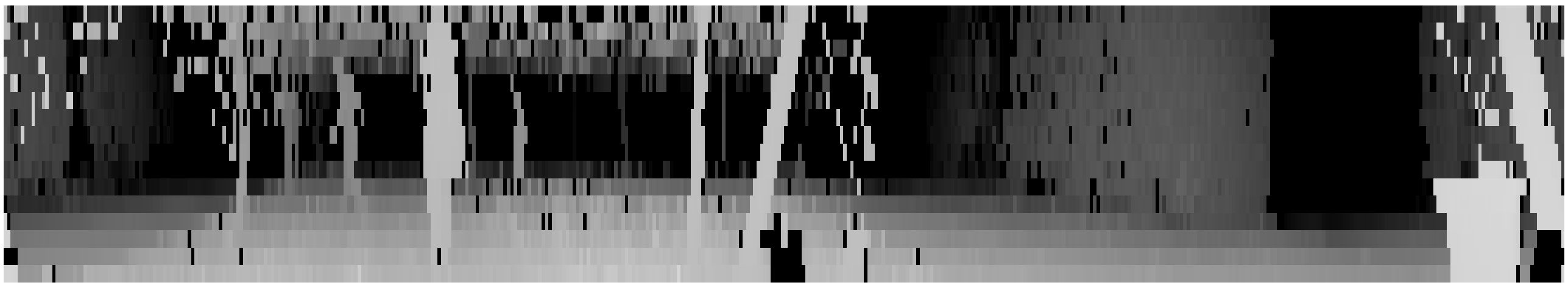

*Figure 188: Result of filling retro-reflector points using the median of surrounding range points.*

Points clustering based on the range difference between adjacent points was performed on the range data. The key steps for clustering points by range similarity were:

1. For every pixel, *start_pixel*, in the image:

2. Push a pointer to *start_pixel* into a container, *current_object*

3. For every pixel, *current_pixel*, in container *current_object*:

4. Read the range at *current_pixel* and assign the value to a variable, *current_range*

5. For each pixel, *next_pixel*, above/ below/ diagonally above/ diagonally below *current_pixel*:

6. Read the range at *next_pixel* and assign the value to a variable, *next_range*

7. Calculate: *range_difference = current_range – next_range*

8. If the magnitude of r*ange_difference* is less than a constant threshold:

9. Push a pointer to the pixel, *next_pixel*, into the container, *current_object*

10. Push the container, *current_object*, to a container of containers

In addition to these steps, the pixels already explored for an object were logged and, in order to avoid redundant search operations, pixels already explored for an object were not queried repeatedly for the same object. Other steps taken to reduce the number of redundant operations were to not start a search at pixels in the range image that were black, since these corresponded to the background, and also pixels that were already a part of an object were not considered further for another object. The output of the steps for segmenting objects based on similar ranges was a container of containers, with each inner container being a list of pointers to pixels from a single object or more than one object at similar ranges.

Having segmented objects, the heights of the objects were found. Objects that were too short were eliminated. An example result after this step is shown in Figure 189.

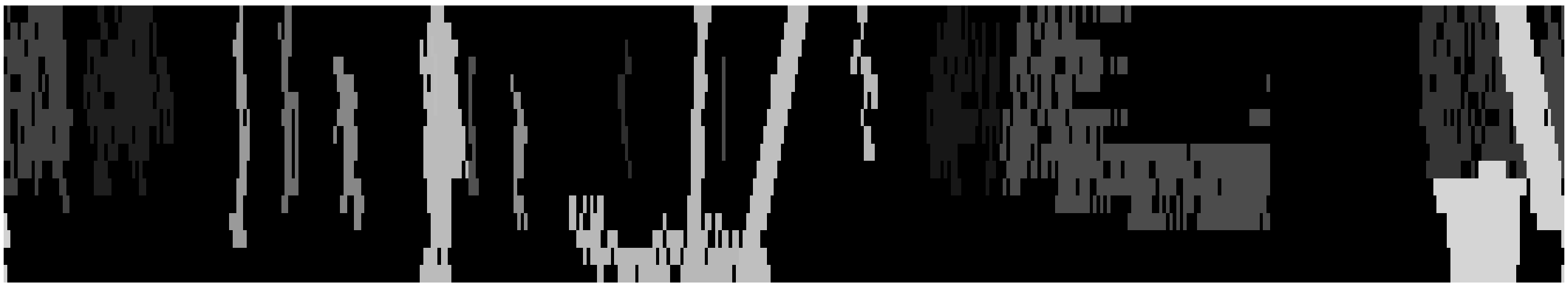

*Figure 189: Results of clustering points, based on a threshold range difference between adjacent points, and eliminating small clusters.*

In the lidar data, points on the ground and canopy tend to have similar ranges and can be adjacent to each other in the rows of the data. In order to reduce the amount of canopy and ground points extracted, the points to the left and right of the *current_pixel* were not considered, since these points could be canopy or ground. Hence, in step 5 of the clustering algorithm, during the search for connected points, only the three points above and below the *current_pixel* were considered for being connected to the *current_pixel*. If all of the points around the *current_pixel* were considered, the result can look like Figure 190, with many points from the ground and canopy that can join potential objects of interest.

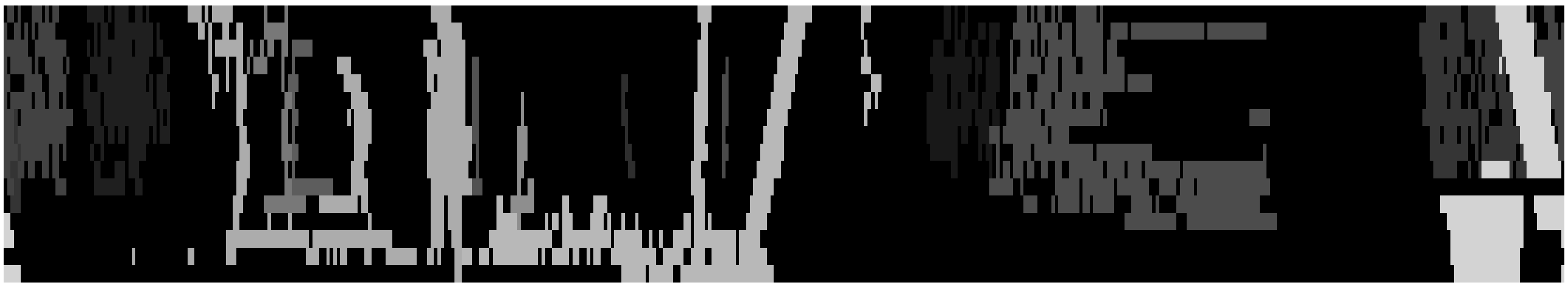

*Figure 190: Results from clustering when searching all points around current_pixel on the object, as opposed to just the 3 points above and 3 points below current_pixel on the object.*

### 5.4.3 Segmented Lidar Data Classification

In Subsection 5.4.2, a method was described to segment 3D lidar data based on the range similarity of adjacent pixels. It seemed that it may be possible to use this segmented output for general obstacle detection by determining the position of the objects and checking that none of these objects were close to the path of the vehicle. It seemed that this approach may be sufficient for pedestrian detection also, since all objects of sufficient size near the AMMP path, including pedestrians, could cause a vehicle stop. However, this approach would not allow for safety related behaviour where a pedestrian is given extra layers of protection, such as larger safety zones or slower speeds, specifically when a pedestrian is detected.

In order to use the segmentation output for pedestrian detection, classification of the segmented outputs was tested. To create a data set for this classification, an application was created which conducted the following process:

1. A single frame of 3D lidar data was read by the application and the segmented output was produced using the method from Subsection 5.4.2.
2. The segmented output was displayed by the application as an image.
3. A person used a mouse to click on the clusters in the image that appeared to be people and then pressed the “enter” key.
4. All of the blobs that were clicked on were extracted by the application into separate images, which were just big enough to fit each blob.
5. The application then resized the blob images to a resolution of 32 pixels wide by 32 pixels high and saved the images to a folder, corresponding to the “people” class.
6. Steps 4 and 5 were repeated for all blobs that were not clicked on by the labeller but the blob images were saved to another folder corresponding to the “not people” class.
7. Steps 1-6 were repeated for a number of 3D lidar data frames, many of which contained people.

The output of this process was two folders for the two classes, “people” and “not people”. The “people” folder contained the resized images of the segmented people and the “not people” folder contained the resized images of the other segmented objects. Three different datasets were generated automatically by the labelling application, which were:

- A dataset that had the lidar range data as one channel of each image and had the lidar intensity data as another channel of the same images. In this dataset, each image in the folders for the two classes had two channels- one channel for lidar range data and one channel for lidar intensity data.
- A dataset of single channel images, which just used the lidar range data.
- A dataset of single channel images, which just used the lidar intensity data.

It was noticed that the intensity object images, in particular, had poor contrast with the black background. To overcome this, it was decided to create additional datasets where the channels were scaled. For the non zero pixels, $p_{ijk}$, using the maximum value of a pixel for the image format, $p_{max}$, the values of all pixels, $P_k$, in channel $k$ and a bias, $b_k$, to differentiate between the minimal values and the background, the formula used for scaling was:

$$p'_{ijk} = b_k + \frac{(p_{max} - b_k)(p_{ijk} - min(P_k))}{max(P_k) - min(P_k)} \tag{25}$$

An example of applying the scaling is shown in Figure 191. For each of the three existing datasets, a scaled version was created. Hence, there were six dataset variants. For the two channel images, the range data and intensity data was scaled separately; this was the equivalent of combining the scaled range dataset and the scaled intensity dataset to create the scaled two channel image dataset, without any further scaling.

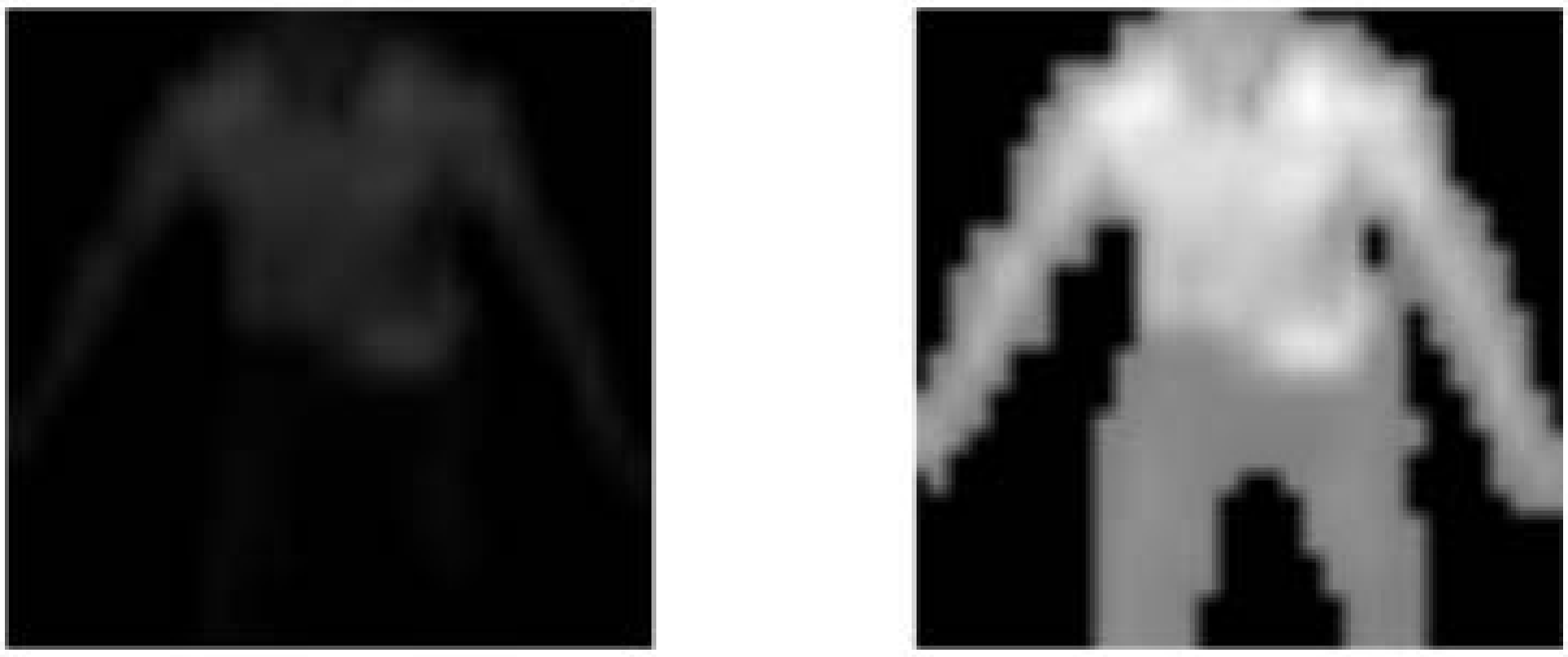

*Figure 191: An example of an intensity image from a dataset before scaling (left) and after scaling (right).*

A key issue identified during labelling was that it was difficult for the labeller to recognise people as people in the lidar data, when the people were standing more than 6 metres from the lidar and this effect was worse the further the people were from the lidar. This was due to the small number of points per person at these distances. The sensor used for data collection was a Velodyne VLP-16 [130], which was mounted at a height of 0.8 of a metre from the ground during data collection. This sensor has 16 planes, separated by 2 degrees. Figure 192 shows discretised values from equation 13 for the Velodyne VLP-16 for a person 1.7 m tall for distances up to 50 m from the lidar. The number of planes per person drops from 16 to 5 within the first 10 metres; within this range some people and other objects appeared to be quite similar (Figure 193). The data labeller claimed that their ability to recognise people as people from a single frame of lidar data dropped from almost 100 percent certainty at 1 metre to close to 0 percent certainty at 10 metres. The labeller explained that at ranges close to 10 metres, they were only able to recognise people by tracking objects between multiple consecutive frames. Because of this, the limit of the range considered for labelled data was set at 10 metres. This range was more than sufficient for the application of the AMMP driving in a kiwifruit orchard, where the maximum stopping distance of the AMMP was measured to be 1.4 m at a speed of 1.4 $ms^{-1}$.

The data augmentation methods used on the people class of the segmented object datasets were horizontal mirroring and 8 crops per object. The sizes of the datasets after data augmentation were more balanced and are given in Table 38. The distributions of the sizes of the pedestrians that were labelled are given in Figure 194, which shows how the pedestrians in the lidar data have very low resolution, even within 10 metres from the lidar. For example, the median number of points per pedestrian in the test dataset was 51 pixels and the lower quartile was 35 pixels.

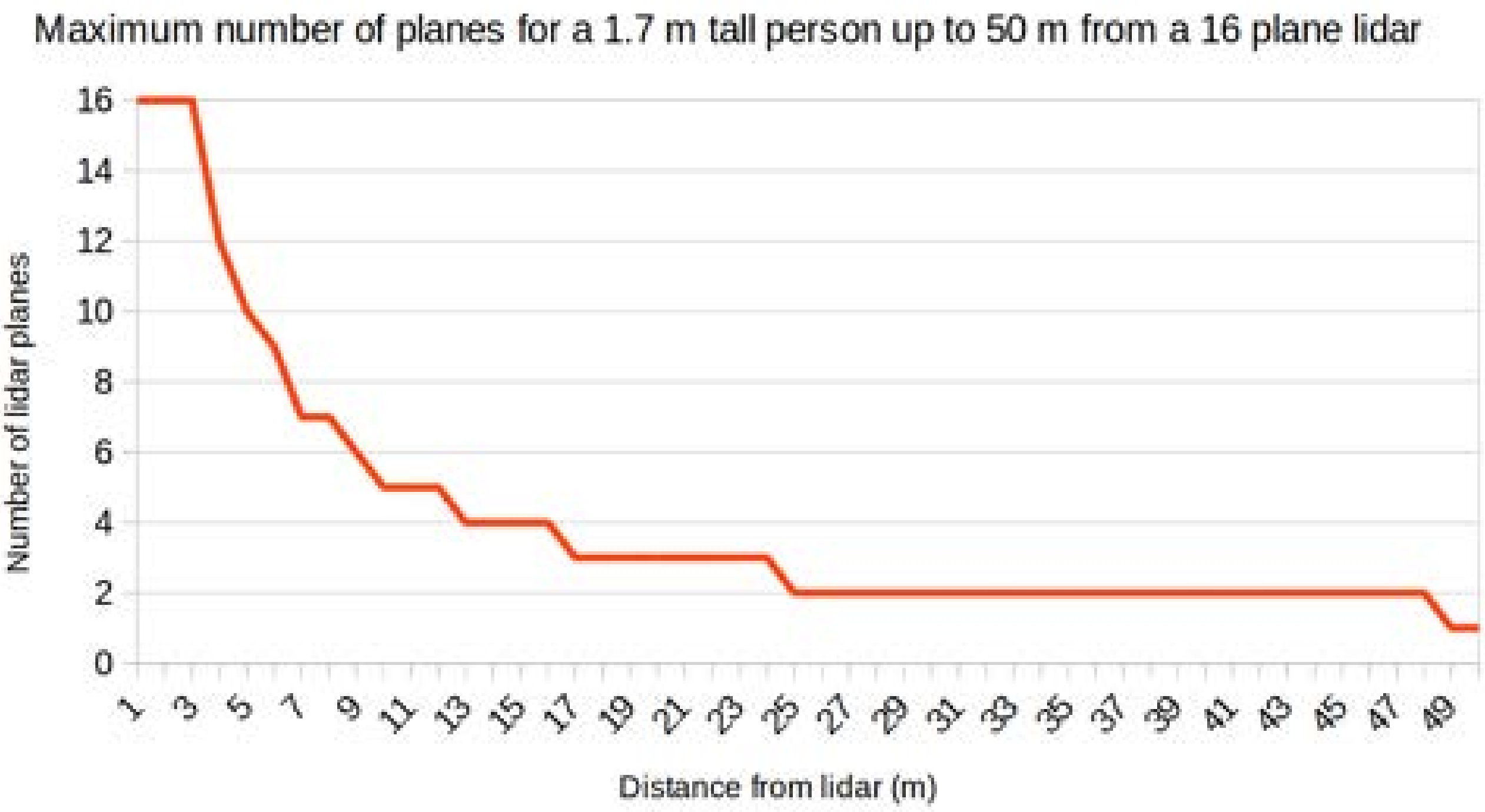

*Figure 192: Equation 13 represented in a graph, showing the maximum number of planes that may land on a person 1.7 m tall for a range of distances from a 16 plane lidar with 2 degrees between planes.*

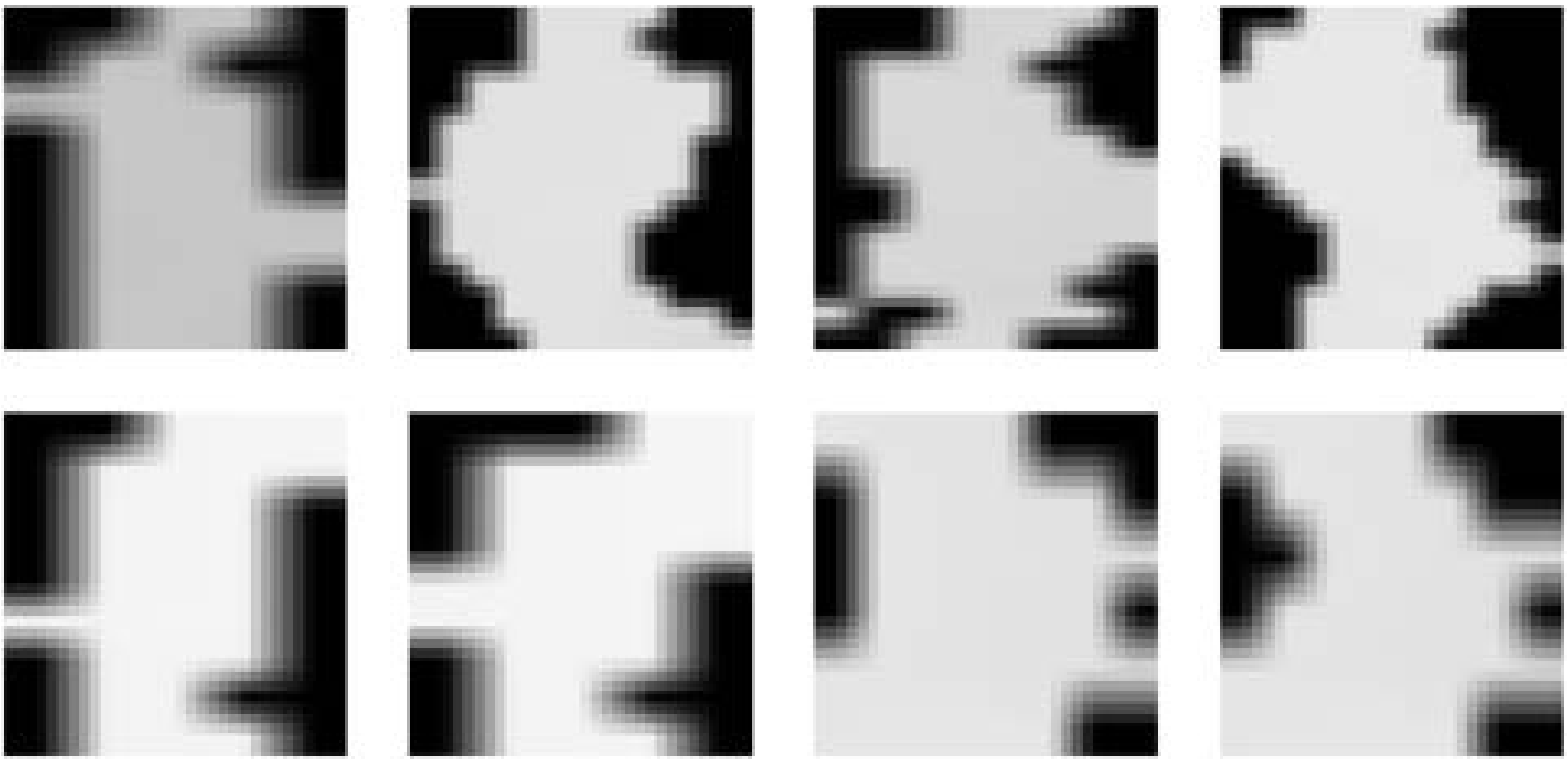

*Figure 193: Some images of people (top row) and images of other objects (bottom row) from the segmented lidar classification dataset.*

*Table 38: Dataset sizes for segmented lidar clusters classification after data augmentation.*

| | **Pedestrian Images** | **Non-Pedestrian Images** |
|---|---|---|
| **Training Dataset** | 70,002 | 55,808 |
| **Test Dataset** | 34,326 | 27,292 |

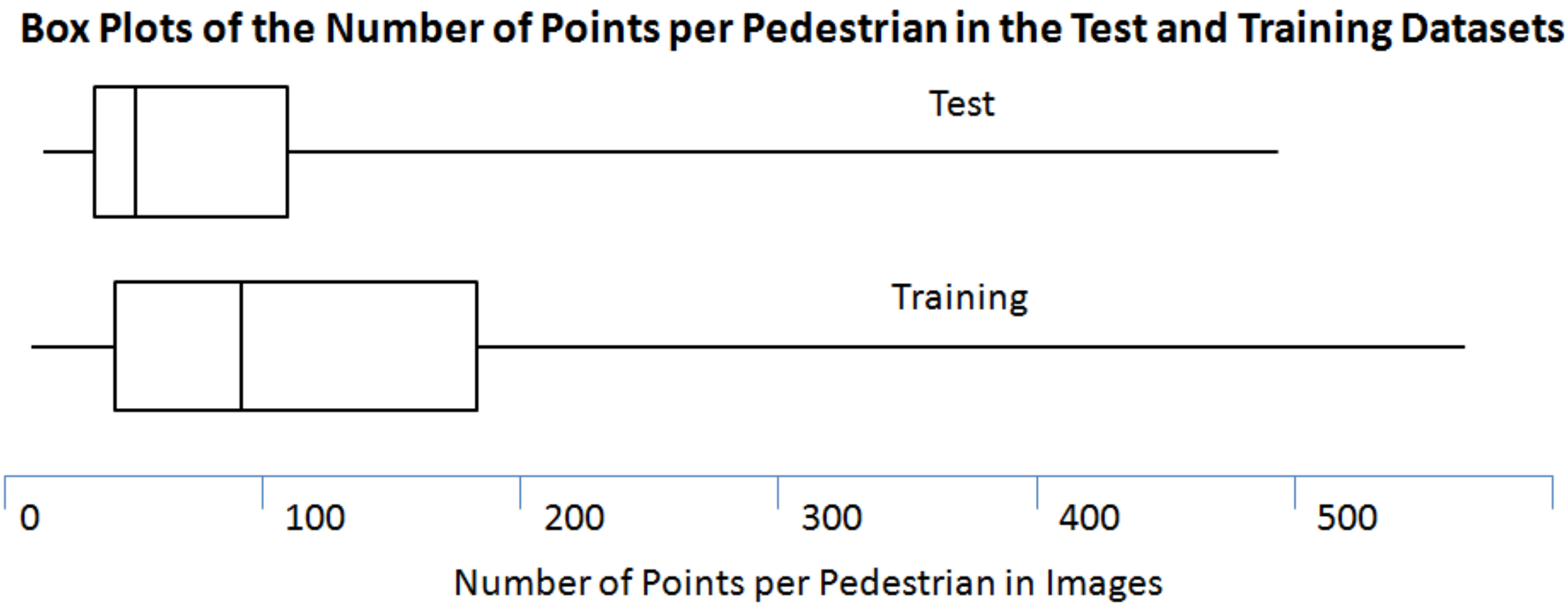

*Figure 194: The distributions of the number of points per labelled pedestrian in the test and training datasets.*

For classification of the segmented objects from the lidar data, CNNs were used, including an implementation of LeNet [46] from Caffe [81] and the “Baseline CNN” from Sabour et al. [225] (Table 39). In addition to classical CNN architectures, it was decided to also test capsule networks as the classifier for the segmented images. Capsule networks have some properties that are desirable for an algorithm for processing lidar data in a safety system, including:

- Capsule networks have been shown to perform well as classifiers for small grayscale images [225], [226]. This is relevant to the segmented lidar data, which can also be represented as small grayscale images (Figure 191).
- Capsule networks are designed to represent part to whole relationships, which compared to other CNNs may increase interpretability [227], which could be a useful property for an algorithm in a safety system.

The capsule network used was an implementation of the dynamic routing by agreement variant from Sabour et al. [225] with two routing iterations. The hyperparameters used for training the CNN variants and the capsule network are given in Table 40.

*Table 39: CNN layers used for segmented lidar data classification.*

| **Caffe [81] LeNet [46]** | **Baseline CNN from Sabour et al. [225]** |
|---|---|
| Convolutional, kernel 5x5, stride 1, depth 20 | Convolutional, kernel 5x5, stride 1, depth 256 |
| Max pooling, kernel 2x2, stride 2 | Relu |
| Convolutional, kernel 5x5, stride 1, depth 50 | Convolutional, kernel 5x5, stride 1, depth 256 |
| Max pooling, kernel 2x2, stride 2 | Relu |
| Fully connected, 500 units | Convolutional, kernel 5x5, stride 1, depth 128 |
| Fully connected, 2 outputs | Relu |
| | Fully connected, 328 units |
| | Fully connected, 192 units |
| | Dropout, 50% |
| | Fully connected, 2 outputs |

*Table 40: Hyperparameters used for training the segmented lidar classification neural networks.*

| **Hyperparameter** | **Value** |
|---|---|
| Batch Size | 480 |
| Number of Training Epochs | 20 |
| Solver | Adam |
| Learning Rate | 0.001 |

The results from the test datasets are given in Table 41. The different neural networks give similar performance. The scaled datasets performed better than the equivalent raw data variants. The intensity channel images and the double channel images performed worse than the range channel equivalents. Based on these results and out of the methods tested, using scaled range channel images with a LeNet model seemed to be the best option for segmented 3D lidar objects classification.

*Table 41: Classification accuracy of different neural networks on the variants of the segmented 3D lidar objects classification datasets.*

| Dataset Variant | Caffe [81] LeNet [46] | Baseline CNN [225] | Capsule Network [225] |
|---|---|---|---|
| **Intensity channel images** | 88.8 % | 87.4 % | 91.6 % |
| **Image with range and intensity channels** | 91.3 % | 92.6 % | 92.4 % |
| **Range channel images** | 93.5 % | 94.8 % | 93.8 % |
| **Scaled intensity channel images** | 94.4 % | 94.3 % | 93.0 % |
| **Image with range and intensity channels scaled** | 94.4 % | 94.9 % | 93.3 % |
| **Scaled range channel images** | 95.4 % | 95.2 % | 94.5 % |

At run time the pedestrian detection algorithm proceeded similar to the data creation procedure. The steps in the pedestrian detection algorithm were:

1. Segmentation, using the range similarity between adjacent pixels method from Subsection 5.4.2 produced object blobs.
2. A blob was extracted and drawn into an image that was just big enough to contain the blob.
3. The blob was resized to resolution 32 pixels by 32 pixels.
4. The non-zero pixels in the resulting image were scaled according to equation 25.
5. The blob was used as input to the classification neural network.
6. If the blob was classified as a person, the Cartesian coordinates of the range data, enclosed by the blob, were calculated. For each set of coordinates that was within a set rectangular volume, referred to as the stop volume, a counter was incremented. If the counter exceeded a threshold, a stop signal was generated.
7. Steps 2 to 6 were repeated for all object blobs, which were segmented in the frame of lidar data.

### 5.4.4 3D Lidar Pedestrian Segmentation

It was noted earlier that the hand engineered segmentation method of Subsection 5.4.2 could tend to segment people and adjacent objects together. It was thought that such errors in segmentation could affect the accuracy of the classification. It was also thought that a segmentation neural network might not have the same issues with segmenting people and adjacent objects together and so this approach was tested.

The dataset creation method from Subsection 5.4.3 was modified so that segmentation labels were created. The original method segmented lidar data by adjacent point range similarity and blobs were clicked for classification. The modification made was that each clicked blob, corresponding to a person, was copied into a segmentation label. The input and segmentation label images were stretched vertically by a factor of 5 to ensure that they were large enough vertically to be processed by the CNN.

When creating a dataset for segmentation by hand, it is common to use a thick line at the border of objects and define the line as an ambiguous region. However, it was unclear if this would be required for creating a segmentation dataset for the lidar data, since the objects were not entirely hand labelled. Furthermore, there was a concern that the objects in the lidar data were so small in terms of the number of pixels that taking some of those pixels to create a border would remove details that were important for training- especially in groups of pedestrians. As a result it was decided to initially test without using a border around objects. Two variants of the segmentation dataset were created: one that just used range data in the input image and another that used range and intensity data in the input image. An example of the input variants and a label are shown in Figure 195.

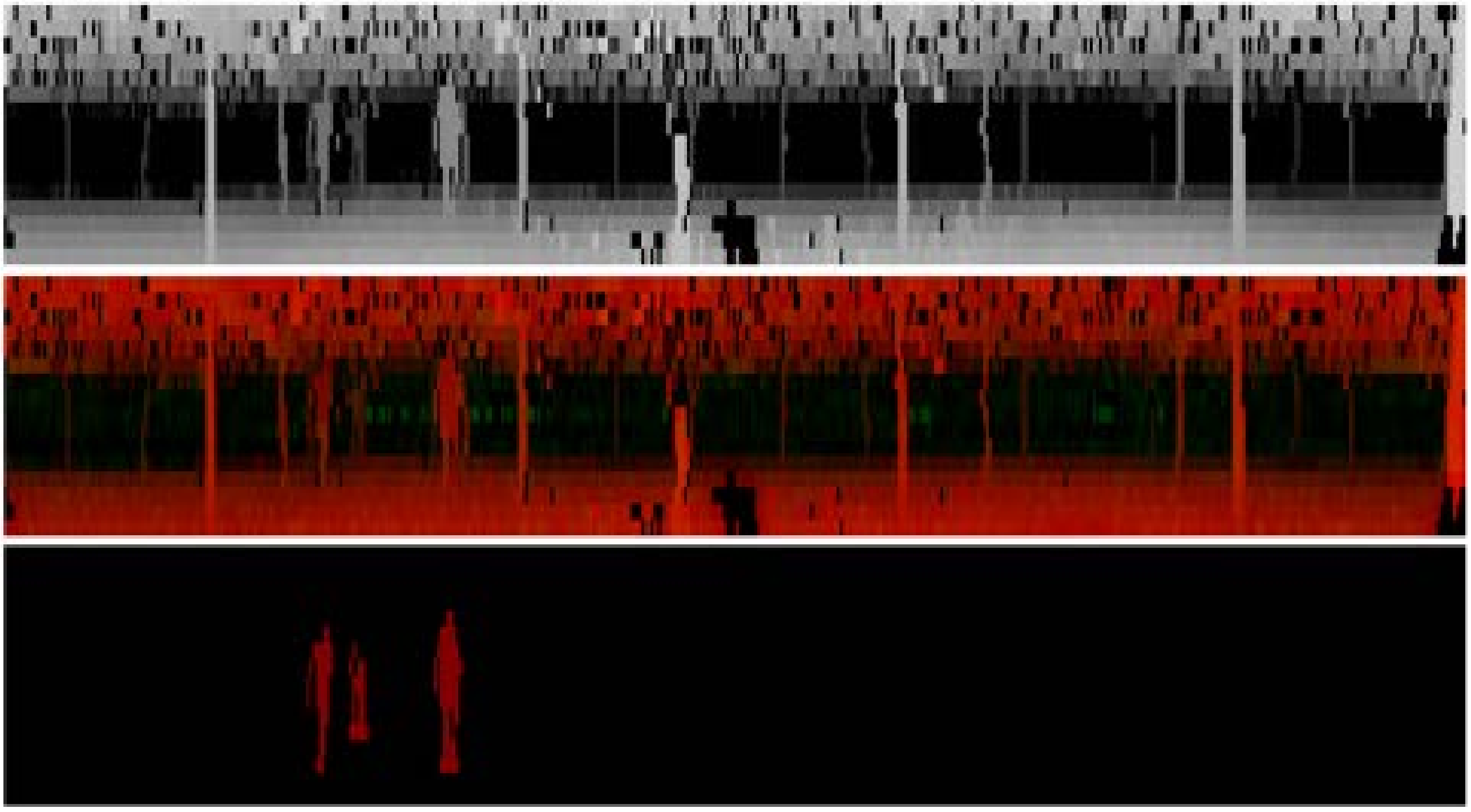

*Figure 195: Examples taken from the lidar CNN segmentation dataset with a range input image (top), range and intensity two channel input image (middle) and the segmentation label (bottom).*

The segmentation CNN used was Mask R-CNN [48] with a GoogLeNet [34] backbone. A pretrained model from the TensorFlow Object Detection API [84], [85] was used for transfer learning. 1000 images were used for training, 250 were used for validation and 298 were used for testing. The hyperparameters used for training are given in Table 5. The metrics used for assessing the performance of the segmentation were the Average Precision and Average Recall as defined by the COCO competition detection evaluation criteria [86]. The results for the range channel dataset are given in Table 42 and the results for the dual channel range and intensity data are given in Table 43. The results for the two datasets are similar. It is interesting to note that the performance on small objects was relatively poor, which was as expected, especially since some of the pedestrians in the dataset consisted of as few as 16 pixels.

The performance of the two CNNs was further analysed by defining the metrics in terms more relevant for pedestrian detection. These definitions were:

- A true positive was any person that was more than half segmented in the inference results.
- A false negative was any person that was less than half segmented in the inference results.
- A false positive was a segmentation mask that did not overlap with a person.

Using these definitions and a threshold class probability for a segmentation mask of 0.5, the results in Table 44 were found. The most concerning aspect of these results was the high numbers of false negatives. It seems from these results that the CNN that was trained on the range data only would be preferable because of its lower rate of false negatives.

Figure 196 shows four inference results for two frames of lidar data. Two of the inference results used just the range channel of the two frames of lidar data as the input. The other two inference results used the range and intensity channels of the same two frames of lidar data as the input. These inference results show how a true positive for one of the CNNs may be a false negative for the other CNN. This suggests that one method for reducing the number of false negatives might be to use the output of both of the CNNs.

Figure 197 shows examples of the most common false positives encountered in the test dataset. The most common false positive was a rubbish bin, which did not appear in the training dataset. For the range data there was a relatively large number of false positives (Table 44) and 94 percent of these were the rubbish bin; hence, this rubbish bin had a large effect on the relatively low precision from the CNN that was trained on the range data only. Most of the other false positives were kiwifruit trunks.

*Table 42: Test dataset inference results from a Mask R-CNN model, trained with 16 plane 3D lidar range channel data for pedestrian detection at up to 10 m from the lidar.*

| Metric Description | | | | Metric Value |
|---|---|---|---|---|
| **Type** | **IoU** | **Object Pixel Area** | **Given Detections per Image** | |
| Average Precision | 0.50 | Any | 100 | 0.80 |
| | 0.75 | Any | 100 | 0.29 |
| | 0.50:0.05:0.95 | Any | 100 | 0.38 |
| | | $< 32^2$ | 100 | 0.22 |
| | | $> 32^2, < 96^2$ | 100 | 0.62 |
| | | $> 96^2$ | 100 | 0.40 |
| Average Recall | | Any | 1 | 0.21 |
| | | Any | 10 | 0.44 |
| | | Any | 100 | 0.44 |
| | | $< 32^2$ | 100 | 0.32 |
| | | $> 32^2, < 96^2$ | 100 | 0.65 |
| | | $> 96^2$ | 100 | 0.40 |

*Table 43: Test dataset inference results from a Mask R-CNN model, trained with 16 plane 3D lidar range and intensity channel data for pedestrian detection at up to 10 m from the lidar.*

| Metric Description | | | | Metric Value |
|---|---|---|---|---|
| **Type** | **IoU** | **Object Pixel Area** | **Given Detections per Image** | |
| Average Precision | 0.50 | Any | 100 | 0.81 |
| | 0.75 | Any | 100 | 0.28 |
| | 0.50:0.05:0.95 | Any | 100 | 0.38 |
| | | $< 32^2$ | 100 | 0.23 |
| | | $> 32^2, < 96^2$ | 100 | 0.61 |
| | | $> 96^2$ | 100 | 0.30 |
| Average Recall | | Any | 1 | 0.21 |
| | | Any | 10 | 0.45 |
| | | Any | 100 | 0.45 |
| | | $< 32^2$ | 100 | 0.34 |
| | | $> 32^2, < 96^2$ | 100 | 0.64 |
| | | $> 96^2$ | 100 | 0.30 |

*Table 44: Pedestrian detection results for Mask R-CNN segmentation of 3D lidar data.*

| Dataset | True Positives | False Positives | False Negatives | Precision | Recall |
|---|---|---|---|---|---|
| **Range** | 642 | 101 | 77 | 0.86 | 0.89 |
| **Range & Intensity** | 598 | 11 | 126 | 0.98 | 0.83 |

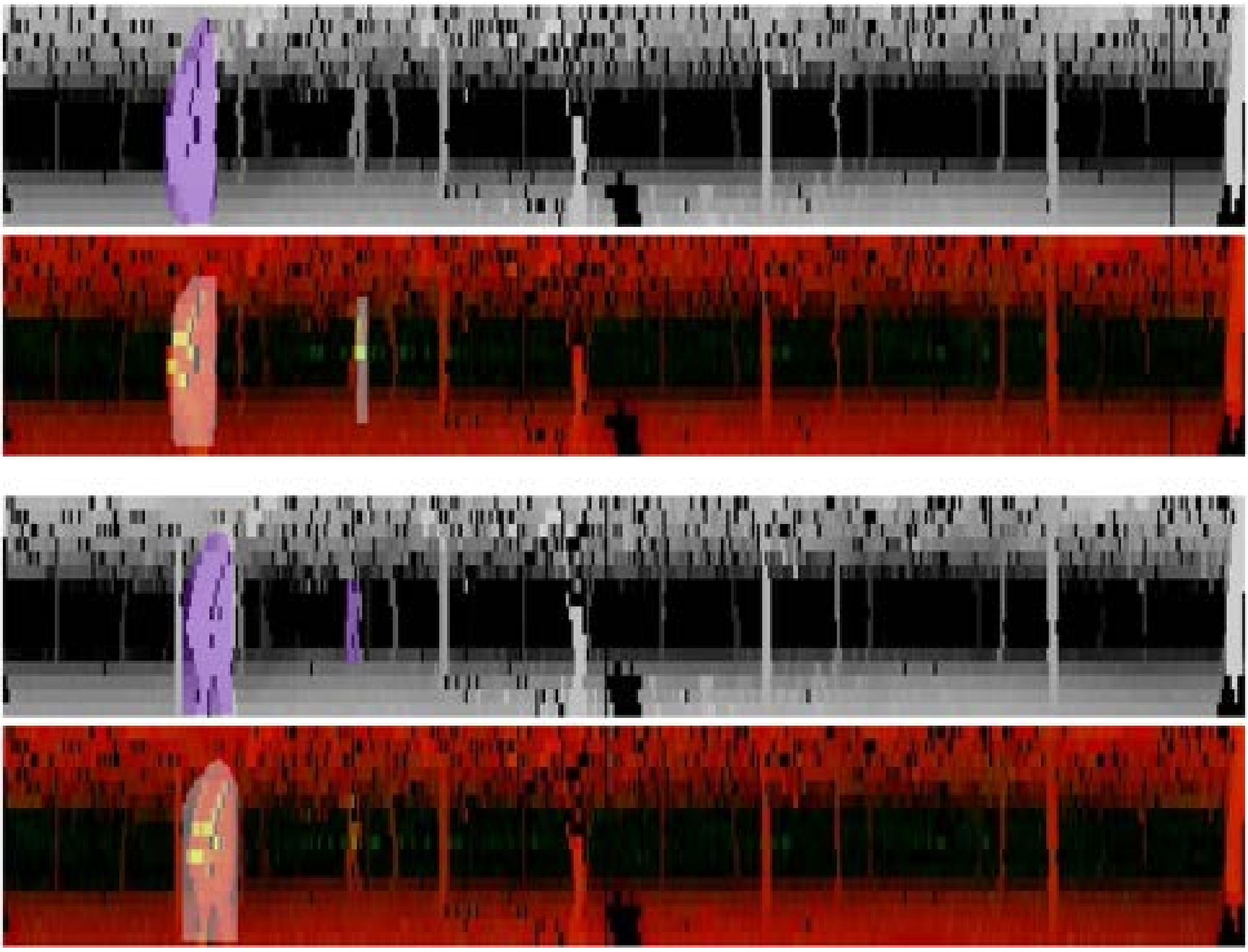

*Figure 196: Two pairs of range and range with intensity inference results showing how a true positive with one dataset and CNN can be a false negative for the other dataset and CNN.*

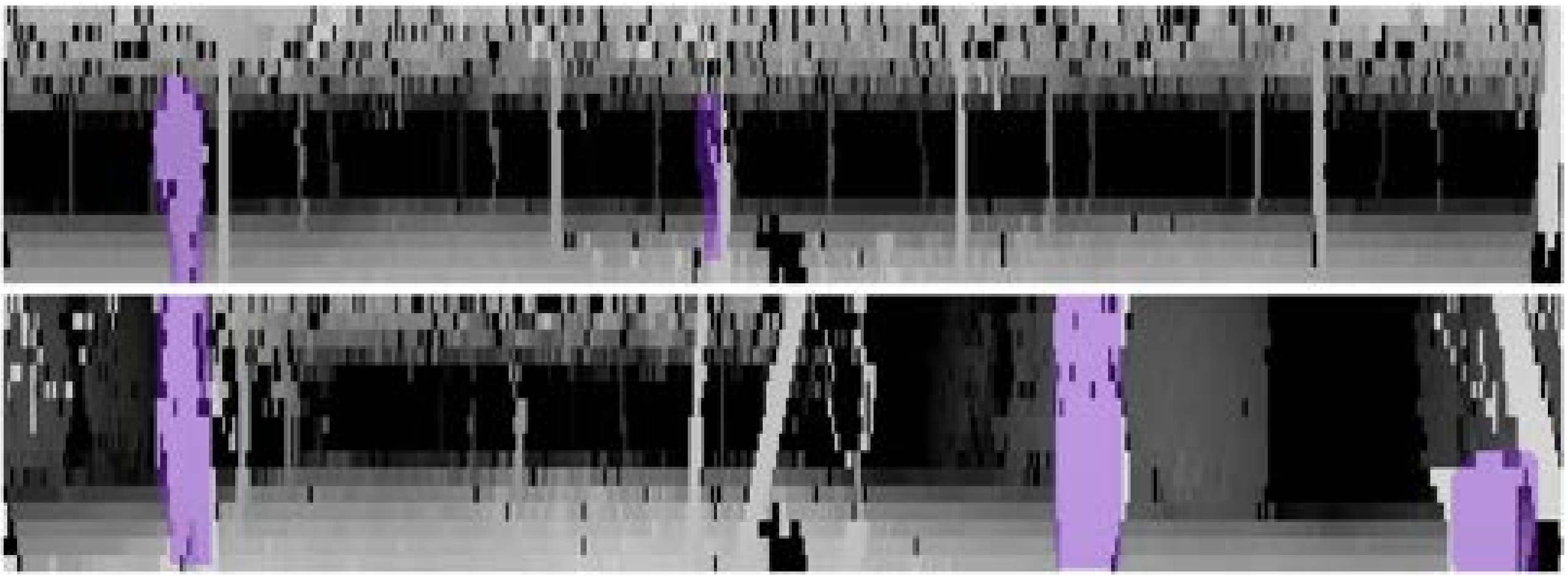

*Figure 197: Examples of the common types of false positives in the test dataset with a trunk mistaken for a person (top) and a rubbish bin mistaken for a person (bottom).*

### 5.4.5 Camera Image Pedestrian Hazard Classification

In addition to detecting pedestrians in lidar data, methods were developed to detect pedestrians in camera images. One of the approaches used was to classify an image according to whether there was a hazard present in the image or not. To do this, a dataset of images was collected in different kiwifruit orchards, where in some of the images there were people present and then in other images there were no people present. Then the dataset was manually labelled with three classes:

- One class was for images where there was at least one person in the “traversable space”, where “traversable space” was defined to be an area where it appeared that the robot could move to in a straight line without obstruction or crossing a treeline.
- Another class was for images where there were no people in the traversable space, but there were people present in other areas of the image.
- The last class was for images where there were no people present anywhere in the image.

Examples of images from each of the classes are shown in Figure 198. The images were taken using different cameras at different times of the day and in different seasons. These raw images were cropped and resized to a resolution of 224 by 224 pixels. The sizes of the training, validation and test datasets are given in Table 45.

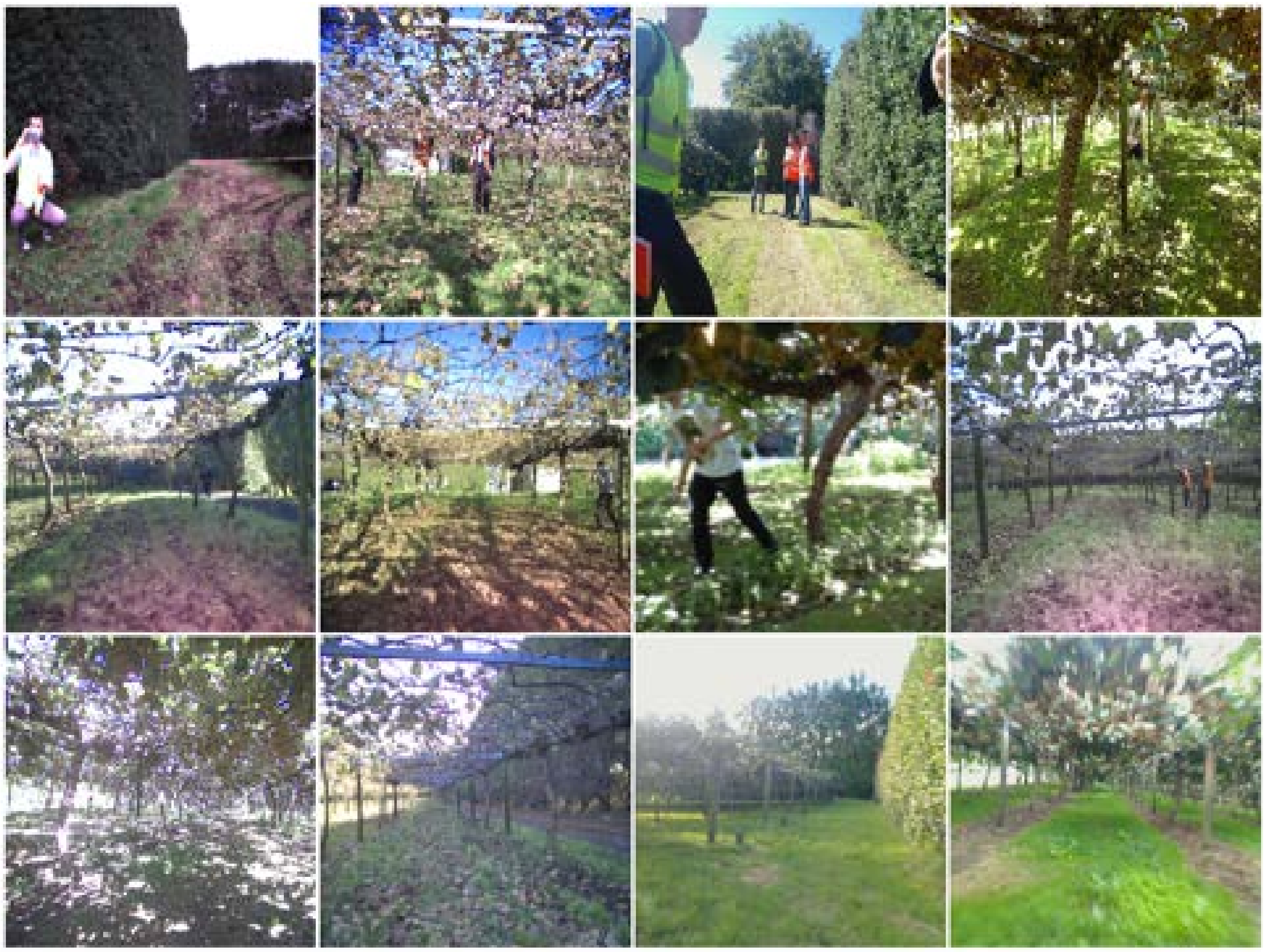

*Figure 198: Examples of the person in traversable space class (top row), person present but not in the traversable space class (middle row) and the no people present class (bottom row).*

*Table 45: Parameters of the pedestrian in traversable space classification dataset.*

| **Class Description** | **Number of Training Images** | **Number of Validation Images** | **Number of Test Images** |
|---|---|---|---|
| Pedestrians in the traversable space in the image | 5010 | 2504 | 2504 |
| Pedestrians present but only outside the traversable space in the image | 5531 | 2765 | 2765 |
| No pedestrians present anywhere in the image | 5823 | 2911 | 2911 |

*Table 46: Hyperparameters used for training a GoogLeNet model for classification of images for the presence of pedestrians in traversable space.*

| **Hyperparameter Description** | **Hyperparameter Value** |
|---|---|
| Number of training epochs | 50 |
| Batch size | 50 |
| Solver type | Stochastic Gradient Descent |
| Initial learning rate | 0.001 |
| Learning rate multiplier | 0.1 |
| Epochs between learning rate reduction | 10 |

A pretrained GoogLeNet [34] CNN from Caffe [81] was trained using the hyperparameters given in Table 46. After training, the validation accuracy was 99.2% and the accuracy on the test dataset was 99.4%. These results were encouraging and so the trained model was tested for inference in a kiwifruit orchard. However, it was found that during field testing the performance was poor; in particular, for the pedestrian present but outside the traversable space class there were just 2 detections from 300 images. It was speculated that this class might be particularly difficult to classify correctly because pedestrians can appear as being relatively small, as can be seen in Figure 198. It was also thought that perhaps the training dataset did not capture the conditions seen during field testing. Regardless, based on such poor field test results and with a list of other possible options, it was decided to investigate the use of other methods for pedestrian detection.

### 5.4.6 Pedestrian Detection Using DetectNet on Camera Images

Rather than using classification of images to infer the position of pedestrians, it was decided to try to use an object detection CNN for pedestrian detection. Since there were already public datasets with labelled pedestrians, it seemed to make sense to try to leverage existing datasets to train a CNN for pedestrian detection. However, it was unclear what would be the best way to use existing datasets, while applying them to the problem of pedestrian detection in orchards. For initial trials, images were selected from the KITTI dataset [124], [125].

The subset of the KITTI dataset that was used in this research was the left colour images of the object data set in the object detection evaluation of 2012 [228]. Inspecting a selection of these images from the KITTI dataset, it seemed that there were many images without any people. It seemed that these images could be used as true negatives; however, these images were mostly of street or urban scenes, so it was postulated that, given the difference of these scenes from the orchard environment, they may not be useful true negative examples. Furthermore, the images with pedestrians contained street scenes as background so regions of these images would become true negatives for an object detection algorithm; making the KITTI images without pedestrians somewhat redundant.

Another issue raised when inspecting the KITTI images was the presence of cyclists. It was unclear whether a cyclist labelled in an image as a cyclist, or not labelled at all, would affect the pedestrian detection performance in an orchard; especially considering that the KITTI bounding box label would include the bike, which could affect the training of the bounding box regression. An alternative considered was relabeling cyclists as pedestrians; however, again it was unclear if this would improve or degrade pedestrian detection performance. To save time and err on the side of caution, it was decided to not use images with cyclists.

Another issue considered with the KITTI images was the scale of some pedestrians. Object detection CNNs can require that the detected objects have pixel dimensions greater than some minimum. Hence it was decided to select images that only contained pedestrians with height and width greater than a set minimum. The selection process involved determining that there were pedestrians of sufficient scale for an object detection CNN and there were no cyclists in the images, according to the following logic:

```
If( (pedestrian_width < PEDESTRIAN_DIMENSION_THRESHOLD) OR
    (pedestrian_height < PEDESTRIAN_DIMENSION_THRESHOLD) OR
    (object_label == “Cyclist”) )
{
        rejectKittiImageAndLabel();
}
```

The variable object_label was read from the KITTI data. The variables pedestrian_width and pedestrian_height were trivially calculated from the KITTI data bounding boxes. The constant PEDESTRIAN_DIMENSION_THRESHOLD was set at 50. After filtering the KITTI images for the above conditions, the resulting dataset size was 510 images and labels.

The 510 KITTI images of pedestrians were used to train a DetectNet CNN [44]. The training was performed using the hyperparameters given in Table 47.

*Table 47: Hyperparameters used for training DetectNet for pedestrian detection.*

| **Hyperparameter Description** | **Value** |
|---|---|
| Number of Training Epochs | 500 |
| Batch Size | 2 |
| Solver Type | ADAM |
| Initial Learning Rate | 0.0001 |
| Learning Rate Multiplier | 0.8 |
| Epochs between Learning Rate Reduction | 25 |

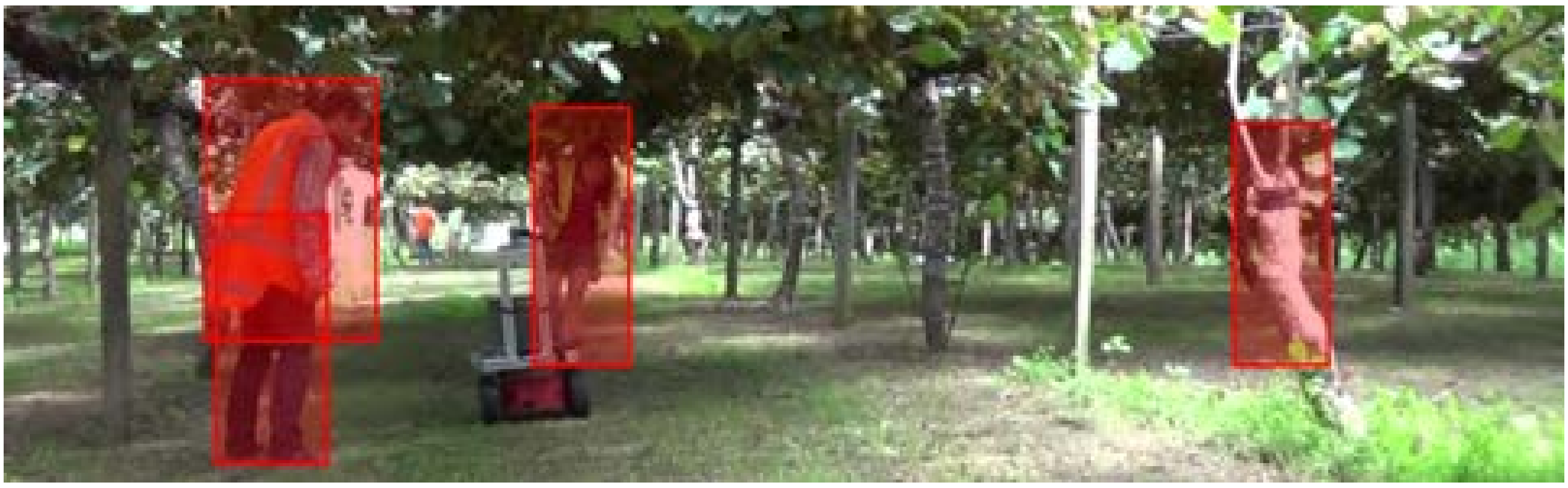

*Figure 199: Pedestrian detection results from a DetectNet model trained with KITTI data, illustrating incorrectly overlapping bounding boxes and a false positive on a kiwifruit trunk.*

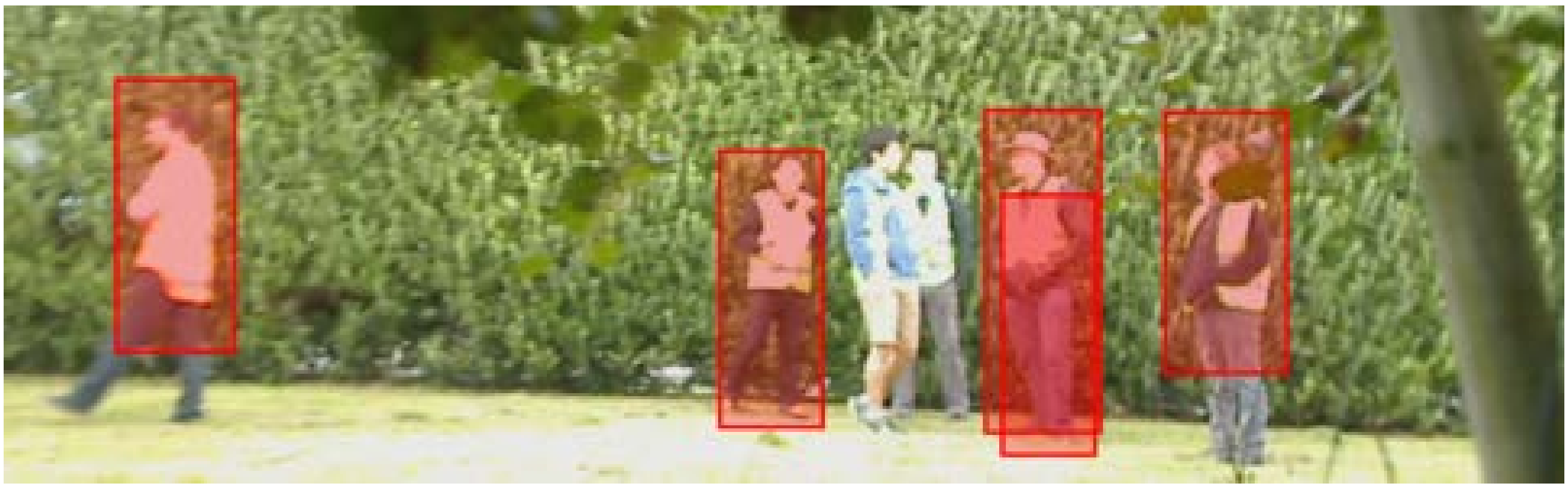

*Figure 200: Pedestrian detection results from a DetectNet model trained with KITTI data, illustrating inaccurate bounding box regression and false negatives.*

For validation, 100 images of people in kiwifruit orchards were used. These images were collected at different times of the day in three different orchards. The images were labelled by hand for all pedestrians. Images that contained pedestrians with a height or width less than 50 pixels were not included in the dataset.

After training the CNN separately two times, the average mean average precision values on the validation dataset was 18. Inspecting inference results revealed that false positives could occur on kiwifruit trunks (Figure 199). In addition, inaccurate bounding box regression, including incorrectly overlapping bounding boxes, and false negatives were observed in inference results (Figure 200). However, there were enough true positive results to encourage continuing with the approach of starting with the KITTI data. It seemed that the false positives could be a result of the CNN not having been trained on kiwifruit trunks and hence the CNN had not been trained to differentiate the trunks from people. It seemed that performing the training with kiwifruit trunks in the dataset would be a useful approach to reduce the occurrence of false positives.

It was hypothesized that images taken from orchards would be better true negatives for pedestrian detection in orchards than images taken from dissimilar environments, such as the street scenes in the KITTI dataset. To test this hypothesis, two new datasets were created. Both datasets included the 510 KITTI images with pedestrians. One dataset included an additional 510 images from the KITTI dataset; these images were selected to contain no pedestrians or cyclists. The other dataset had true negatives that were 510 images taken from kiwifruit orchards- again with no pedestrians.

The two datasets were used to train DetectNet CNNs. The validation dataset was the same as that used for the 510 KITTI images with pedestrians. The mean average precision using just the KITTI dataset was 18, compared to 23 when using the orchard images for true negatives.

Inspecting the inference results after the training runs, when images from the kiwifruit orchards were used as true negatives, it seemed that there were no false positives but also a high rate of false negatives. This could be explained by the CNN learning that the orchard scenes generally do not contain people, according to the training data. Whether this is true or not, the fact that there was a high rate of false negatives was an undesirable result.

### 5.4.7 Orchard Pedestrian Detection Using Instance Segmentation

The computer vision row following system, described in Subsection 4.7.8, performed segmentation of the traversable space. It was realised that pedestrian detection could be added to the traversable space segmentation CNN with minimal extra processing. This would present an advantage over a system using an object detection CNN, such as the DetectNet system trained with the KITTI data.

Again, the CNN used for segmentation was Mask R-CNN [48] with a GoogLeNet [34] backbone and a pretrained model, made available by the TensorFlow Object Detection API [84], [85], was used for transfer learning. 720 images were used for training, 125 images were used for validation and 153 images were used for testing. Only two classes were used: people and traversable space, as previously defined in Subsection 4.7.8. The image widths ranged from 400 to 414 pixels. The image heights were from 125 to 160 pixels. The hyperparameters used for training are given in Table 5. The test dataset results are given in Table 48, with high values for the overall average precision, compared to results from other datasets [85] and the results from the lidar segmentation (Table 42, Table 43). Using the same definitions for true positives, false positives and false negatives, from Subsection 5.4.4, results for pedestrian detection were collected, as given in Table 49. Again, these results are high compared to the results from segmentation of the lidar data (Table 44). An example inference result is shown in Figure 201.

*Table 48: Test dataset inference results from a Mask R-CNN model, trained for pedestrian and traversable space detection in camera images.*

| Metric Description | | | | Metric Value |
|---|---|---|---|---|
| **Type** | **IoU** | **Object Pixel Area** | **Given Detections per Image** | |
| Average Precision | 0.50 | Any | 100 | 0.88 |
| | 0.75 | Any | 100 | 0.51 |
| | 0.50:0.05:0.95 | Any | 100 | 0.51 |
| | | $< 32^2$ | 100 | 0.15 |
| | | $> 32^2, < 96^2$ | 100 | 0.44 |
| | | $> 96^2$ | 100 | 0.72 |
| Average Recall | | Any | 1 | 0.38 |
| | | Any | 10 | 0.57 |
| | | Any | 100 | 0.57 |
| | | $< 32^2$ | 100 | 0.20 |
| | | $> 32^2, < 96^2$ | 100 | 0.54 |
| | | $> 96^2$ | 100 | 0.74 |

*Table 49: Pedestrian detection results from using Mask R-CNN for segmentation of camera images.*

| True Positive | False Positive | False Negative | Precision | Recall |
|---|---|---|---|---|
| 267 | 2 | 18 | 0.99 | 0.94 |

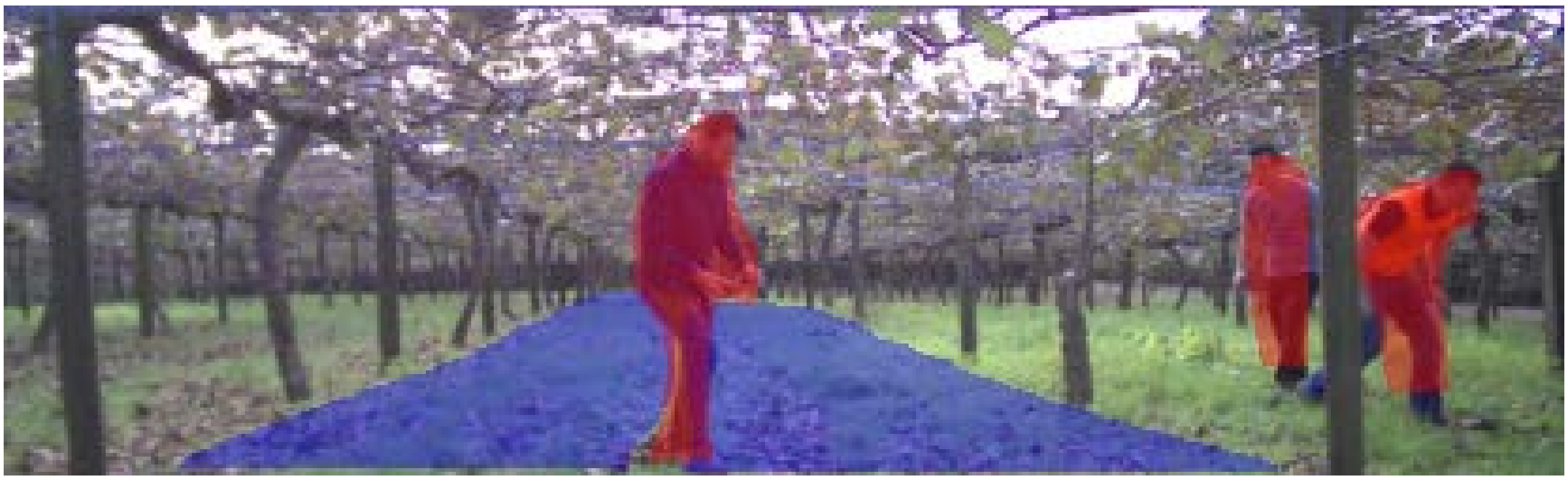

*Figure 201: Example inference result for segmentation of people and traversable space.*

### *5.4.8 Segmented Pedestrian Stereopsis*

In order to measure the position of segmented pedestrians, dense stereopsis was used. The approaches considered for using stereopsis included:

- Performing segmentation on a pair of camera images and performing stereopsis on the two segmented outputs. This would be a more computationally expensive option because segmentation would be performed twice.
- Performing stereopsis on two camera images and then performing segmentation on the disparity map. This option has the issue that pretrained CNNs trained on disparity maps are less readily available than pretrained CNNs trained on camera images.
- Performing segmentation on one image, performing stereopsis on two images and then finding the pixels corresponding to pedestrians from the segmented output in the disparity output. This option has neither of the issues of the first two options, so this was the method tested.

The pipeline for the pedestrian detection system using segmentation and stereovision is given in Figure 202. The pipeline for pedestrian detection was similar to the pipeline used for spray boom navigation with stereovision (Figure 70). The difference was that the final module in the pedestrian detection system checked if points corresponding to pedestrians were within a set volume.

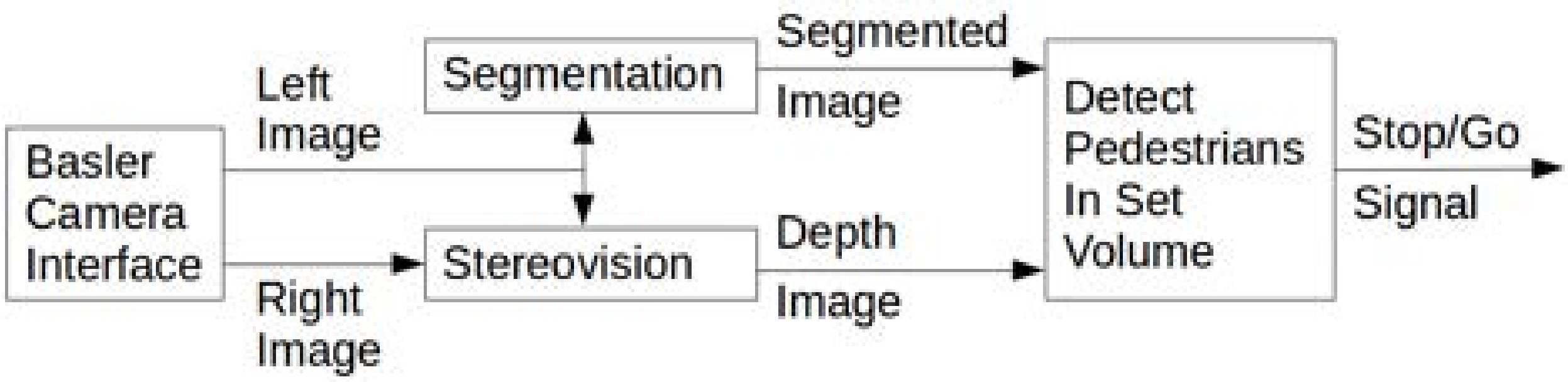

*Figure 202: Pipeline for the segmentation stereovision pedestrian detection system.*

### *5.4.9 Traversable Space Intrusion by Pedestrians*

A stereovision system requires both cameras to be functional in order to produce reliable outputs. If either camera has a fault, the output of the stereovision system may be erroneous. In the kiwifruit orchards, issues that were observed included a camera being occluded by leaves and images being affected with glare and water on the lenses. Other possible issues could include lense focus changes, lense aperture changes, fasteners loosening, causing changes to stereo calibration extrinsic parameters, and damage to the optics or sensors. Even when the stereovision system is compromised in such ways, it may still be possible to detect hazards from single camera images.

One method of detecting hazards in single images was described in Subsection 5.4.5; however, it was found that this method could be unreliable. Another algorithm for detecting hazards in single images was developed using the segmentation output. The key insight behind this algorithm was that if a segmented pedestrian was close to the segmented traversable space, then in the real world the pedestrian was close to the traversable space. This insight allowed the pedestrian's intrusion of the traversable space to be detected from a single camera, even though the position of the pedestrian and the position of points in the traversable space had not been measured. The steps used to determine the intrusion of a pedestrian into the traversable space were as follows:

1. From the raw segmentation output (Figure 203), two separate binary images were created for the traversable space segment and the pedestrian segments (Figure 204).
2. Both binary images were dilated to increase the size of the segments (Figure 205).
3. A pixel-wise Boolean-AND operation was performed with the two images, to leave the regions of overlap of the segments (Figure 206).
4. The sum of high pixels in the overlap image was calculated.
5. The sum of the overlap was compared to a threshold and, if the threshold was exceeded, a pedestrian hazard detected signal was sent.

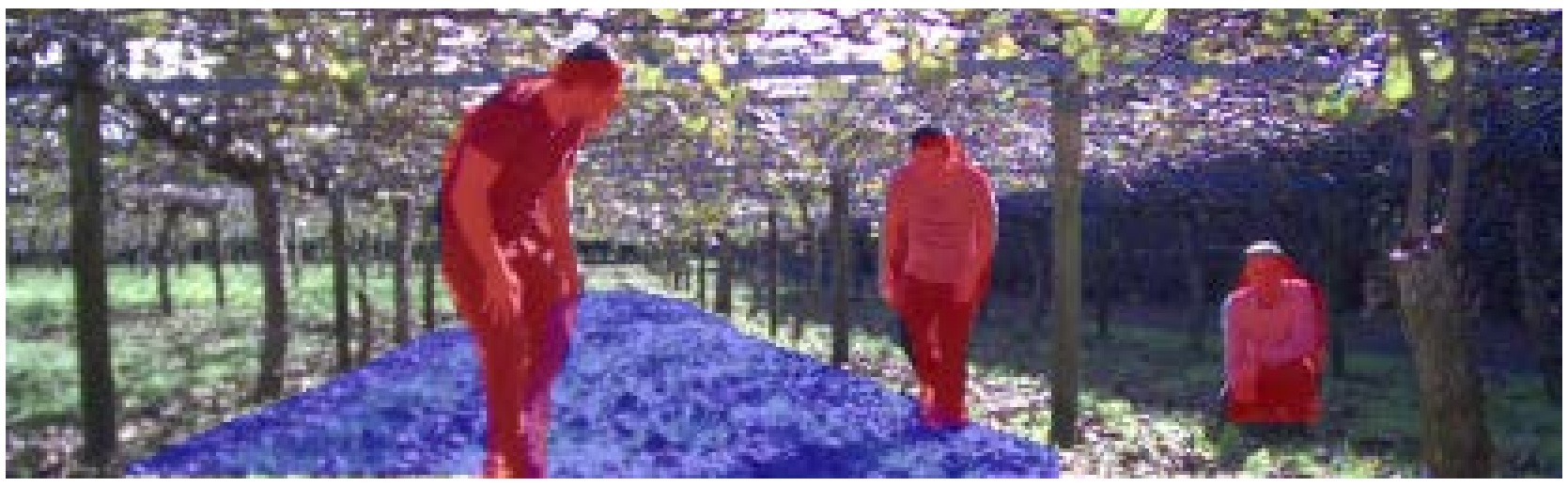

*Figure 203: An inference output of people and traversable space segmentation.*

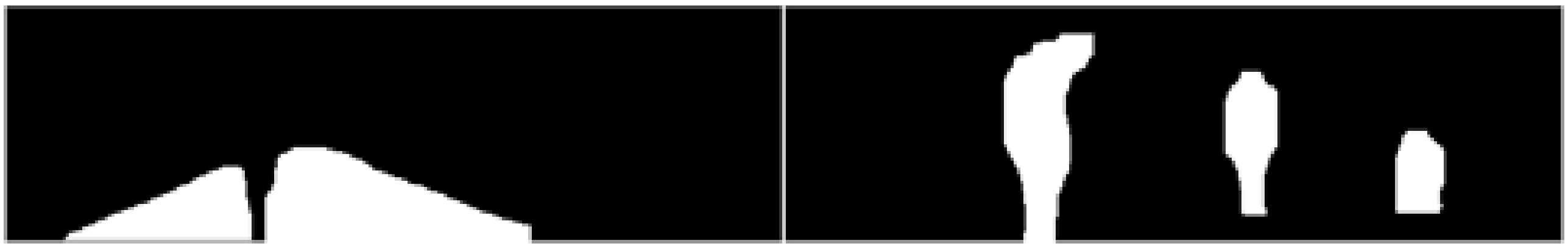

*Figure 204: The segmentation masks from Figure 203 for traversable space and people extracted.*

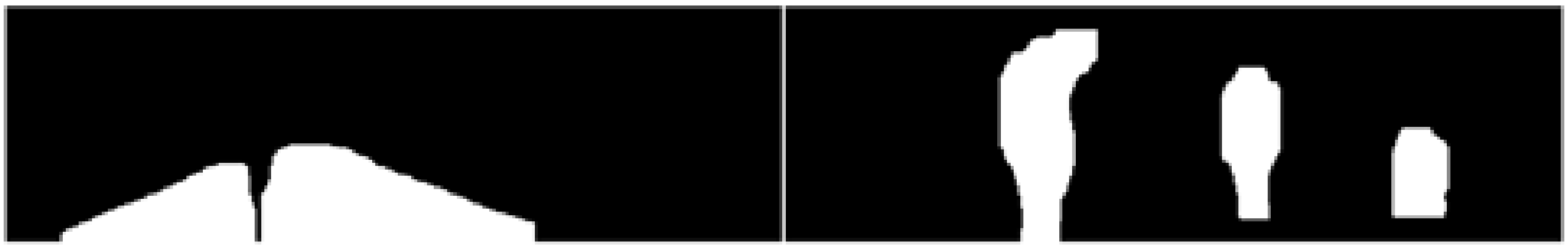

*Figure 205: The segmentation masks from Figure 204 dilated.*

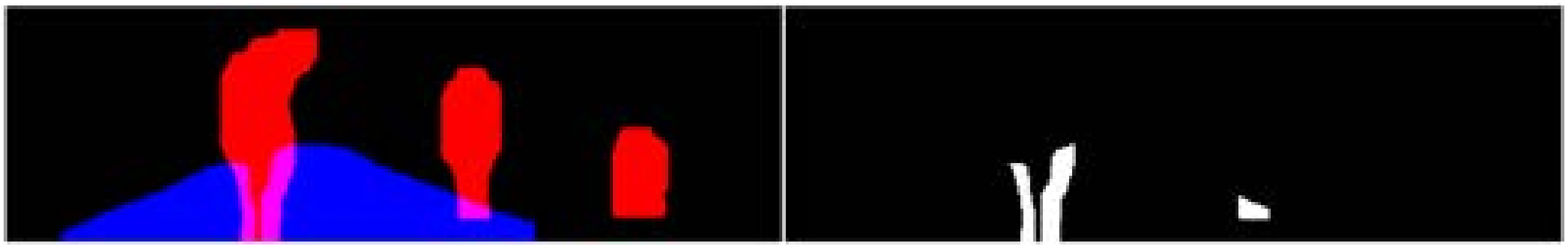

*Figure 206: Showing the overlap of the dilated segmentation masks for people and traversable space from Figure 205; this overlap was used to detect the presence of people in the traversable space.*

Because an overlap with the traversable space is required for detection of a pedestrian in the traversable space, the definitions for true and false positives and negatives (from Subsections 5.4.4 and 5.4.7) could not be reused for traversable space intrusion. Instead the following definitions were used:

- True positives were pedestrians, which were more than half visible and were in the true traversable space, where the pedestrian segmentation output formed an overlap with the segmentation output for the traversable space.
- False negatives were pedestrians, which were more than half visible and were in the true traversable space, where the pedestrians did not have a segmentation output that formed an overlap with the segmentation output for the traversable space.
- False positives were pedestrians outside of the traversable space, where the segmentation output of the pedestrian formed an overlap with the segmentation output for the traversable space.

Using these definitions, the results in Table 50 were found. Because of the different definitions for true positives, false positives and false negatives, it is not fair to directly compare these results to the other pedestrian detection results. Nevertheless, the values for precision and recall for traversable space intrusion are similar to the results for the pedestrian detection using just the segmentation (Table 49). The difference in these results may be due to the requirement of traversable space intrusion to segment both the people and the traversable space accurately.

*Table 50: The results from testing traversable space intrusion.*

| True Positives | False Positives | False Negatives | Precision | Recall |
|---|---|---|---|---|
| 201 | 4 | 14 | 0.98 | 0.93 |

## 5.5 Obstacle Detection Architecture

Multiple pedestrian detection methods were proposed in Subsection 5.4. In addition to these methods, simpler algorithms were developed for triggering emergency stops. These algorithms checked if points in a point cloud were inside a set volume; if more than a threshold number of points were detected in the set volume, an emergency stop was triggered. These algorithms are referred to here as object in volume detection.

Given the multitude of methods proposed here, the question arises as to how the different methods should be used together. To do this a concept described in ISO 13849-1:2015 [1] is used, which says that the risk of Common Cause Failures may be reduced with diversity and redundancy. Redundancy was addressed by using multiple sensors, multiple algorithms for each sensor and independent channels for each sensor. Diversity was achieved by using multiple sensor modalities, using different types of algorithms and using different types of processors. Table 51 captures some of this redundancy and diversity. Note that some of the algorithms are very simple and so can run on simple reliable processors. In addition, some algorithms are somewhat redundant where multiple methods have been proposed to perform the same task. Furthermore, it was proposed that diversity may be increased by not using the same segmentation CNN for the lidar and camera data. Also, diversity may be increased by using different software frameworks for different CNN instances in the Safety Related Parts of the Control System.

*Table 51: The proposed lidar and camera Safety Related Parts of the Control System.*

| **Method Description** | **Suggested Processor** | **Comments** |
|---|---|---|
| Lidar high visibility vest detection | Microcontroller | Algorithm simple enough to be run on a microcontroller |
| Lidar object in volume detection | Microcontroller | Algorithm simple enough to be run on a microcontroller |
| Lidar segmentation by range similarity with CNN classification | ARM processor with FPGA | This CNN may use Caffe |
| Lidar pedestrian detection using segmentation CNN | ARM processor with FPGA | Suggest not using this because this method uses a similar algorithm to stereovision segmentation and there is another lidar pedestrian detection method |
| Camera image hazard presence classification | CPU with GPU | Suggest not using this because it did not work well in field testing |
| DetectNet with camera images | CPU with GPU | Suggest not using this because it is somewhat redundant if the segmentation CNN is used |
| Stereovision object in volume detection | CPU with GPU | |
| Stereovision segmentation | CPU with GPU | This CNN may use TensorFlow for diversity |
| Monocular segmentation traversable space intrusion | CPU with GPU | This is intended to be used in addition to or when there is an issue with stereovision |

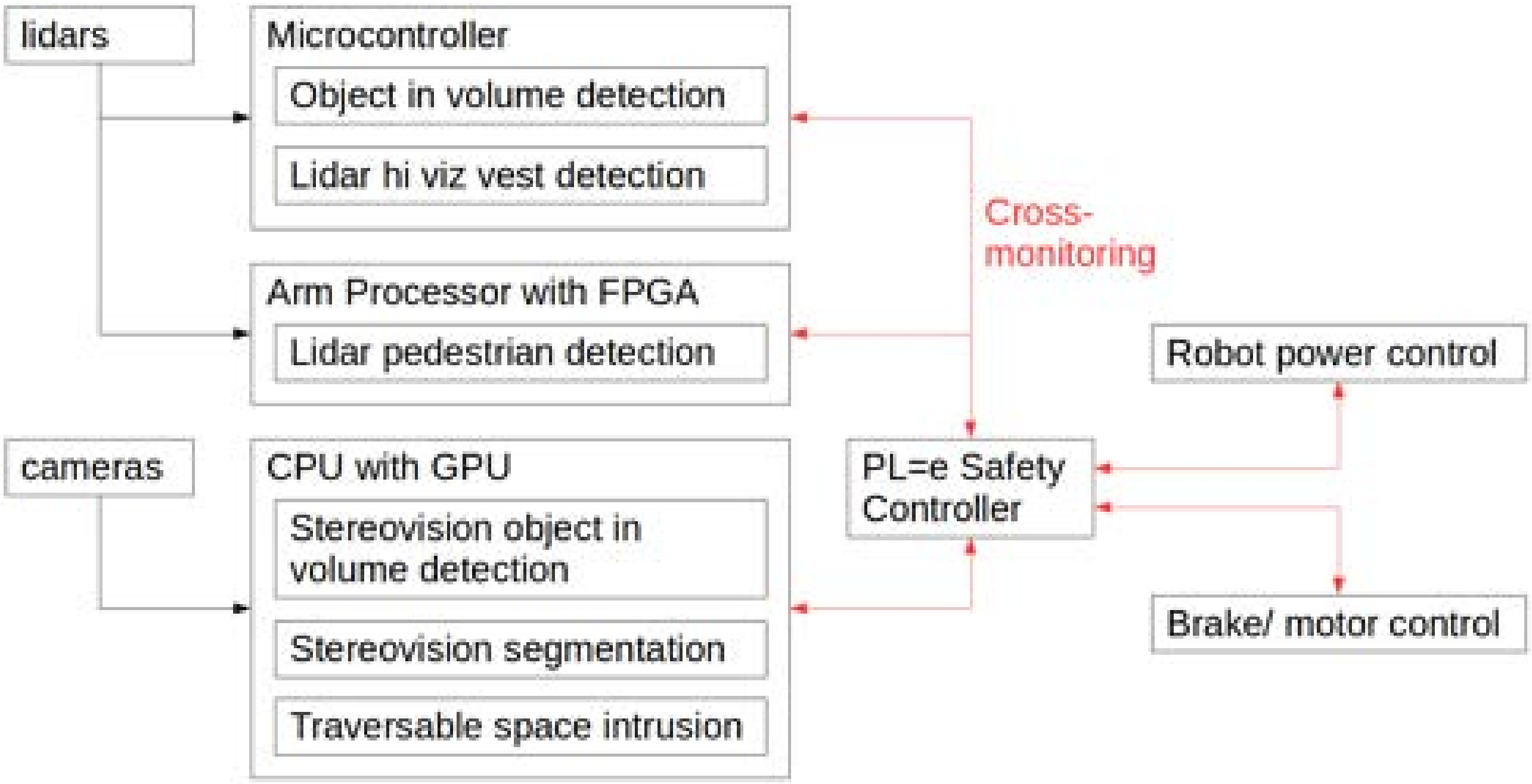

*Figure 207: Proposed architecture for integrating camera and lidar modules into the Safety Related Parts of the Control System.*

Figure 207 shows the architecture for the integration of the camera and lidar components into the overall safety system. The outputs of the camera and lidar processing are fed to a safety controller, with a Performance Level of e and a Safety Integrity Level of 3, which facilitates cross-monitoring between parts of the system and controls emergency stops. In addition to using cameras and lidar, multiple other risk reduction measures were proposed as a part of this project, including the use of emergency stop buttons, wireless emergency stop controllers, bumpers, radars, lights, sirens and access control [205]. Such components would also be integrated into the safety system through the safety controller.

The cross-monitoring illustrated in Figure 207 could take many forms besides heartbeats, handshakes and CRC checking. For example, if a high visibility vest was not detected within a range of 4 m but a person was detected using another channel of the system, this might be evidence of a person not wearing a high visibility vest as required and this could be used to trigger a stop. In addition, if traversable space intrusion detected a person in the current row but the stereovision segmentation system did not detect a hazard, this could be due to an issue with the stereovision system. Similarly, dissimilar systems could be used to cross-check results and build statistics about possible faults.

## 5.6 AMMP Safety Discussion

The risk assessment for the AMMP found that the required risk reduction was Performance Level d or Performance Level e, depending on how much time that people would be around the AMMP in a commercial deployment. These are very demanding risk reduction requirements and so a comprehensive safety system is required for the AMMP. This will be achieved through a combination of many different types of sensors, covering the envelope of the AMMP. It is recommended here that other options, including inherently safer smaller vehicles, should be considered also.

Multiple pedestrian detection methods were presented. The high visibility vest detection worked well within a range of 4 m, for the vests and lidar used. The performance of this system could be enhanced using more reflective material on the vests and using a higher resolution lidar. An obvious issue with the high visibility vest detection is that it will not work if the reflective strips are occluded. In addition, some key concerns with using the reflectivity of the high visibility vest reflective strips were that some people in the orchard may not wear the vests or that different vests might not be reflective enough to cause a stop signal. The concern of not wearing high visibility vests could, to some extent, be managed with access control to the orchard and standard operating procedures for those working around the AMMP. In terms of the variations of different vests, it seems that this could also be tested and controlled with procedures. Furthermore, a person not wearing a high visibility vest could be automatically detected using a combination of the high visibility vest detection and the other lidar pedestrian detection methods.

Two methods of lidar pedestrian detection without using the reflectivity of high visibility vests were presented. These methods are potentially interchangeable; however, it was thought that more diversity in algorithms could be achieved by favouring the method that used segmentation by range similarity with CNN classification.

An issue that was identified during the labelling of the lidar data was that there were only a small number of points per pedestrian, even at a distance of only 10 m. It seems that this could have affected the inference results. For future work, using higher resolution lidars seems like a practical solution to this problem.

Classifying camera images, according to whether there was a person present and whether a person was in the path of the vehicle, seemed to work with test datasets but then not so well in the field. It is possible that this method could work better with more comprehensive datasets and more data augmentation. Nevertheless, using Mask R-CNN with camera data worked well for pedestrian detection. Furthermore, using the segmentation output for stereovision and traversable space intrusion seemed like a more elegant solution than the other camera pedestrian detection methods tested. In addition, the measured performance of the camera image segmentation and traversable space intrusion algorithms was high, especially compared to the lidar pedestrian detection methods.

## 5.7 AMMP Safety Conclusions

The methods presented here mostly related to pedestrian detection, while other aspects of machine safety for the AMMP were deferred as future work. A connection was made between the pedestrian detection methods presented here and a future safety system, through the risk assessment and proposed system architecture. The conclusions that were drawn with respect to the pedestrian detection methods were:

- The lidar intensity channel high visibility vest detection was very reliable within a range from 1 metre to 4 metres from the sensor and this range could be improved with wider reflectors or more measurements per lidar plane.
- Various neural network architectures, including Mask RCNN and capsule networks, work well for pedestrian detection in kiwifruit orchards for camera and lidar data.
- The overlap between the segmented traversable space and a segmented pedestrian may be used to detect a pedestrian in a hazardous zone from a single camera.

# 6 Future Kiwifruit Orchard Robots

The second kiwifruit harvester addressed some of the issues with the original kiwifruit harvester. It seems that there is potential for a system like the second kiwifruit harvester to be able to pick more fruit than the original kiwifruit harvester in the future. Key areas for the improvement of the second kiwifruit harvester include the path planning for the robot arm and there is also potential to experiment more with the use of different sensors in different poses.

The pollination problem always seemed like a difficult problem because of the speed at which detection and spraying is performed. The flower detection, calibration and boom navigation methods presented seemed adequate. However, dry pollination also seems like an intriguing option for future work. The development of the dry pollination system could benefit from developments on the second kiwifruit harvester. In future studies, it would also be interesting to understand how dry robotic pollination can be complemented by bees.

The kiwifruit orchard navigation system presented worked well for navigating orchard blocks once the row end turn parameters had been found. Work was also presented on extracting structure defining features from 3D lidar data, with the intention of using these features for creating and navigating from maps. This work is ongoing.

The safety system will be key to the success of the AMMP and the deployment of all of the work described in this thesis. In addition to the pedestrian detection methods presented using cameras and lidars, multiple other risk reduction measures should be used including bumpers, radar, emergency stop buttons, wireless emergency stop controllers, lights, sirens, signage, standard operating procedures and access control. Smaller pollination and harvesting robots should also be considered for inherent safety.

# 7 Appendix A: Reinforcement Learning Review

The general overall approach adopted by the team for kiwifruit harvesting was to firstly detect kiwifruit in images and then calculate the location of the fruit. Then the robot hand was sent to the kiwifruit location for picking. Other paradigms for kiwifruit harvesting were considered, including reinforcement learning. This review of reinforcement learning methods was focused on practical methods for creating an agent that might be able to perform harvesting. This review is included in this appendix as reference material.

Levine et al. [229] used images from a single camera as input to a CNN to predict the probability of successful grasping by a seven degree-of-freedom arm, using a set of motor commands, which were input to the CNN. The output of the CNN was used for continuous closed loop servoing towards an object. The CNN was trained using over 800,000 grasp attempts from 14 similar hardware setups, consisting of the camera, robot arm and a tray of objects to be picked up. Interesting features of this work included that the camera and robot arm pairs were not calibrated relative to each other and the camera-arm pairs were slightly differently mounted between individual setups. This would be a useful feature for a fleet of kiwifruit harvesting robots since the same CNN could be used despite manufacturing differences and new individual robots added to the fleet would not have to be initially or regularly recalibrated. In addition, the approach of Levine at al. allows the robot to grasp many different types of objects. This could be useful for harvesting because it could allow the robot to generalise across multiple types of fruit. In addition, the approach of Levine et al. was tested with a cluttered arrangement of objects and learnt unexpected strategies for grasping the objects. Hence this approach could be useful for harvesting since the canopy is a clutter of leaves, branches, fruit and wires, so the ability to operate in clutter and learn different strategies could be useful. Training the CNN for grasp success could possibly cause the CNN to learn that fruit behind wires cannot be approached directly and could possibly allow an algorithm to determine what fruit in a cluster will have the highest probability of successfully being picked. Possibly, grasping kiwifruit or apples could be a simpler task than the setup used by Levine et al., since each robot only has to grasp one variety of fruit in a given session. However, in an orchard environment, additional criteria of success may be required, such as:

- Not accidentally hitting and dislodging adjacent fruit in a cluster.
- Not hitting solid obstacles such as wires and branches.

Deep reinforcement learning approaches were considered for training a robot arm and robot hand to pick kiwifruit. However, model-free reinforcement learning is known to be sample inefficient [230] and so might be impractically slow to train in a real world application without the benefit of many parallel systems. However, recent advances such as Hindsight Experience Replay [231] allow some deep reinforcement learning methods to train much quicker than previously possible. The key idea behind Hindsight Experience Replay is that if an agent fails to reach a goal state, the state that the agent reaches can be used as a pretend goal and the actions used to reach the pretend goal can form a part of the training set as a successful instance for reaching the pretend goal.

Another approach to accelerating deep reinforcement learning is to store the actions and rewards for different inputs and then use this stored experience to guide the selection of new actions [230]. It seems that this approach might work well for harvesting if there is only limited variability in the input, dictated by a limited number of arrangements of fruit, branches and leaves, that needs to be learnt.

Finn et al. [232] describe a meta-learning approach where a model is trained to adapt to new data with few samples. The key idea is to repeatedly train a copy of the last saved model on a training dataset and then test on a test dataset, computing an update for the model from the test losses only and then discarding the version of the model that has just been trained. This process is repeated, without actually updating the last saved model, for a batch of training dataset and test dataset pairs. After gathering the updates from the batch is complete, the last saved model is finally updated and saved again, before repeating with another batch of training dataset and test dataset pairs. Finn et al. showed that this meta-learning approach could be used for classification, regression and reinforcement learning. It seems that this approach could be used to train an agent that might be good at learning to pick with relatively few samples so that at training time the datasets may consist of a mix of picking tasks from simulation, lab trials and possibly some real world data. Then the agent might be trained with more real world data and it might be able to quickly adapt to the real world setting.

Heess et al. [233] trained different agents in simulation to perform locomotion in different environments using simple reward functions with a reinforcement learning policy optimisation variant. Their experiments demonstrated that agents learned quicker when trained on increasingly difficult environments as opposed to environments with randomly varying difficulty. It seemed that applying this idea to kiwifruit harvesting might mean initially targeting low hanging and unobstructed fruit, before progressing to partially occluded and then fully occluded fruit. Heess et al. [233] also demonstrated that using more varied environments tended to give more robust performance when the test environments contained variations that were not present in the training environments. This idea might be applied to kiwifruit harvesting by training with multiple orchards or different types of fruit.

Another option for accelerating deep reinforcement learning for harvesting might be to leverage simulation. James and Johns [234] used simulation and transferred their algorithm to the real world when training and testing a CNN to control a robot arm to perform grasping and lifting of an object. However, unlike the work of Levine et al. [229], James and Johns used a Q-learning loss to perform deep reinforcement learning with intermediate rewards based on the distance to the target, successful grasping and the height of the grasped object when lifting. In addition, James and Johns were successful in implementing transfer learning from simulation to the real world by getting a robot arm to move towards the target object; however, the real world arm would fail to grasp the object.

James et al. [235] later extended this work by using domain randomisation in simulation to improve the performance of a CNN that was trained end-to-end to control a robot arm to locate a randomly placed cube, pick-up the cube and drop the cube into a randomly placed basket. The main sensor input was a camera in the scene. The CNN was only trained in simulation, using hundreds of thousands of demonstrations from a hand engineered algorithm, which controlled the simulated robot arm to perform the picking and dropping operation, using the environment state from the simulator. The domain randomisation methods used in the simulation included varying the colour and texture of different objects; varying the initial pose of all objects, including the parts of the robot arm; and also adding various non-target objects to the environment. These measures were shown to improve performance in the real world tests, which used a similar robot arm camera, target cube and basket setup. In addition, it was shown that improvements could be achieved by using an LSTM module, concatenating the joint angles of the robot into a layer of the CNN and including the position of the target and gripper as an output and in the loss of the CNN during training.

Progressive Nets is another approach used for simulation to real world transfer [236]. To perform this transfer, a neural network is trained to perform a task in simulation and then the weights of that neural network are frozen. Then a separate neural network is instantiated for training in the real world with only the output layer weights transferred from the frozen model. When running the real world neural network in a forward-pass during training, the simulator neural network is also run, with the outputs of layer L-1 in the simulator neural network feeding into layer L of the real world neural network. This process may be repeated across a curriculum of tasks of increasing difficulty by freezing the latest neural network and adding another separate neural network which takes inputs from the frozen networks.

Concepts Networks [237] might be another approach for improving simulation to real world transfer and can also allow for integration of expert knowledge, which may improve the sample efficiency of the overall algorithm. In Concept Networks there is a hierarchy of algorithms contained in independent modules: some of the low level algorithms may be expert systems and some can be neural networks. These lower level modules can be controllers, which take in some data and output an action, or can be data processing modules, which transform data into a type that is used by higher level layers. At the higher layers of the hierarchy, there are Deep Q Network (DQN) trained agents that select actions based on the state and inputs feeding in from the lower level modules. By training different modules independently, there can be substantial gains in sample efficiency. Breaking down a problem into modules also allows for simplified reward functions for the module selecting agents and simplified transfer of modules between different tasks. For simulation to real world transfer, some of the lower level modules could be trained in simulation or an agent trained in simulation might be deployed in the real world by swapping a lower level module with one trained in the real world. For harvesting, the hierarchy might be different to Concept Networks; however, the general idea of componentising the system so that different modules can be swapped in and out or can be trained/ tuned independently of other parts of the system may be a useful approach for getting harvesting working with a reinforcement learning trained agent.

Imitation learning, also known as learning from demonstration, may also be a useful approach for getting harvesting working in real kiwifruit orchards, without excessive time invested in supervised learning. Nair et al. [238] combined imitation learning with an off-policy model-free reinforcement learning method, Deep Deterministic Policy Gradients (DDPG). An off-policy method was used so that the demonstrations could be included in a replay buffer for training. In addition, to integrate the demonstrations, the actor was trained with a loss function that measured the difference between the policy selected action and the demonstration action, if and only if the demonstration action led to higher rewards. In addition, to improve sample efficiency, resetting the environment to a mid-demonstration state during sampling and Hindsight Experience Replay (HER) [231] were used.

# 8 Appendix B: YouBot Middle Joints Kinematics Equations

The control of the Kuka YouBot arm for harvesting and pollination involved initially turning joint 0 of the arm so that the plane of joint 1, joint 2 and joint 3 was in line with the target. Then inverse kinematics was used for the calculation of the angles for joint 1, joint 2 and joint 3, which were $\alpha_1$, $\alpha_2$ and $\alpha_3$ respectively. The calculation of these joint angles is included in this appendix for reference. The coordinate systems and variable definitions are shown in Figure 208.

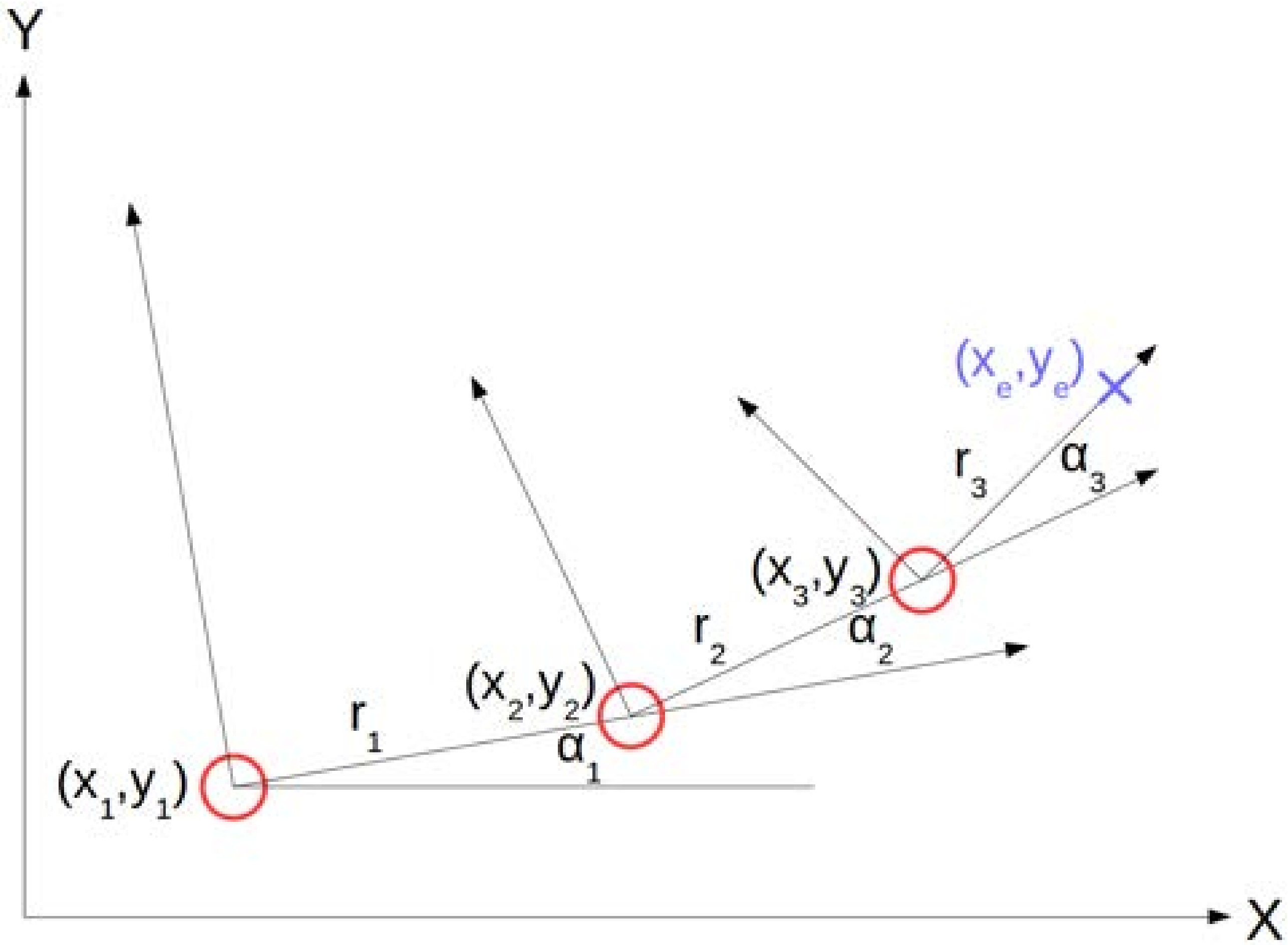

*Figure 208: The variables used for calculating the angles of joint 1, joint 2 and joint 3, which are represented by red circles here.*

The length of the links 1, 2 and 3 are $r_1$, $r_2$ and $r_3$. The coordinates of the end of the arm are $(x_e,y_e)$. The unknowns are $x_2$, $y_2$, $x_3$, $y_3$, $\alpha_1$, $\alpha_2$ and $\alpha_3$. By using Figure 208:

$$x_2 = x_1 + r_1 \cos(\alpha_1) \tag{26}$$

$$y_2 = y_1 + r_1 \sin(\alpha_1) \tag{27}$$

$$x_3 = x_2 + r_2 \cos(\alpha_1 + \alpha_2) \tag{28}$$

$$y_3 = y_2 + r_2 \sin(\alpha_1 + \alpha_2) \tag{29}$$

$$x_e = x_3 + r_3 \cos(\alpha_1 + \alpha_2 + \alpha_3) \tag{30}$$

$$y_e = y_3 + r_3 \sin(\alpha_1 + \alpha_2 + \alpha_3) \tag{31}$$

Considering the length of the links:

$$(x_2 - x_1)^2 + (y_2 - y_1)^2 = r_1^2 \tag{32}$$

$$(x_3 - x_2)^2 + (y_3 - y_2)^2 = r_2^2 \tag{33}$$

$$(x_e - x_3)^2 + (y_e - y_3)^2 = r_3^2 \tag{34}$$

Let β be the angle of the last link, which is known:

$$\alpha_1 + \alpha_2 + \alpha_3 = \beta \tag{35}$$

$$x_3 = x_e - r_3 \cos(\beta) \tag{36}$$

$$y_3 = y_e - r_3 \sin(\beta) \tag{37}$$

Expanding (32) and rearranging:

$$x_2^2 - 2x_1 x_2 + x_1^2 + y_2^2 - 2y_1 y_2 + y_1^2 = r_1^2 \tag{38}$$

$$x_2^2 + (-2x_1)x_2 + (x_1^2 + y_2^2 - 2y_1 y_2 + y_1^2 - r_1^2) = 0 \tag{39}$$

Expanding (33) and rearranging:

$$x_2^2 - 2x_3 x_2 + x_3^2 + y_2^2 - 2y_3 y_2 + y_3^2 = r_2^2 \tag{40}$$

$$x_2^2 + (-2x_3)x_2 + (x_3^2 + y_2^2 - 2y_3 y_2 + y_3^2 - r_2^2) = 0 \tag{41}$$

(39) = (41), then rearrange:

$$x_2^2 + (-2x_1)x_2 + (x_1^2 + y_2^2 - 2y_1 y_2 + y_1^2 - r_1^2) = x_2^2 + (-2x_3)x_2 + (x_3^2 + y_2^2 - 2y_3 y_2 + y_3^2 - r_2^2) \tag{42}$$

$$(2x_3 - 2x_1)x_2 + (2y_3 - 2y_1)y_2 = x_3^2 + y_3^2 - r_2^2 - x_1^2 - y_1^2 + r_1^2 \tag{43}$$

$$(2x_3 - 2x_1)x_2 = x_3^2 + y_3^2 - r_2^2 - x_1^2 - y_1^2 + r_1^2 + (2y_1 - 2y_3)y_2 \tag{44}$$

$$x_2 = \frac{x_3^2 + y_3^2 - r_2^2 - x_1^2 - y_1^2 + r_1^2}{2x_3 - 2x_1} + \frac{(2y_1 - 2y_3)y_2}{2x_3 - 2x_1} \tag{45}$$

Let

$$D=\frac{2y_1-2y_3}{2x_3-2x_1}=\frac{y_1-y_3}{x_3-x_1} \quad (46)$$

$$E=\frac{x_3^2+y_3^2-r_2^2-x_1^2-y_1^2+r_1^2}{2x_3-2x_1} \quad (47)$$

Then

$$x_2=Dy_2+E \quad (48)$$

Substitute (48) into (39):

$$(Dy_2+E)^2+(-2x_1)(Dy_2+E)+(x_1^2+y_2^2-2y_1y_2+y_1^2-r_1^2)=0 \quad (49)$$

$$D^2y_2^2+2DEy_2+E^2-2Dx_1y_2-2Ex_1+x_1^2+y_2^2-2y_1y_2+y_1^2-r_1^2=0 \quad (50)$$

$$(1+D^2)y_2^2+(2DE-2Dx_1-2y_1)y_2-2Ex_1+x_1^2+y_1^2-r_1^2+E^2=0 \quad (51)$$

Let

$$A=1+D^2 \quad (52)$$

$$B=2DE-2Dx_1-2y_1 \quad (53)$$

$$C=-2Ex_1+x_1^2+y_1^2-r_1^2+E^2 \quad (54)$$

Solve (51) using the quadratic formula:

$$y_2=\frac{-B\pm\sqrt{B^2-4AC}}{2A} \quad (55)$$

Rearranging (26) and (27):

$$\cos(\alpha_1)=\frac{x_2-x_1}{r_1} \quad (56)$$

$$\alpha_1=acos\left(\frac{x_2-x_1}{r_1}\right) \quad (57)$$

$$\sin(\alpha_1)=\frac{y_2-y_1}{r_1} \quad (58)$$

$$\alpha_1 = asin\left(\frac{y_2 - y_1}{r_1}\right) \tag{59}$$

Rearranging (28) and (29):

$$\alpha_2 = acos\left(\frac{x_3 - x_2}{r_2}\right) - \alpha_1 \tag{60}$$

$$\alpha_2 = asin\left(\frac{y_3 - y_2}{r_2}\right) - \alpha_1 \tag{61}$$

Rearranging (35):

$$\alpha_3 = \beta - \alpha_1 - \alpha_2 \tag{62}$$

# 9 Appendix C: Untested Harvesting Algorithms

This appendix describes some ideas and algorithms, which were partially implemented but not tested as part of a harvesting system. These algorithms stem from a pipeline proposed for kiwifruit pose detection with stereo cameras, as illustrated in Figure 209. A key part of this pipeline was the neural network (Mask R-CNN [48]) which was intended to detect keypoints on the kiwifruit. The harvesting system hardware considered for these algorithms was more like the original kiwifruit harvester, with a pair of stereo cameras facing vertically up towards the canopy for each robot arm.

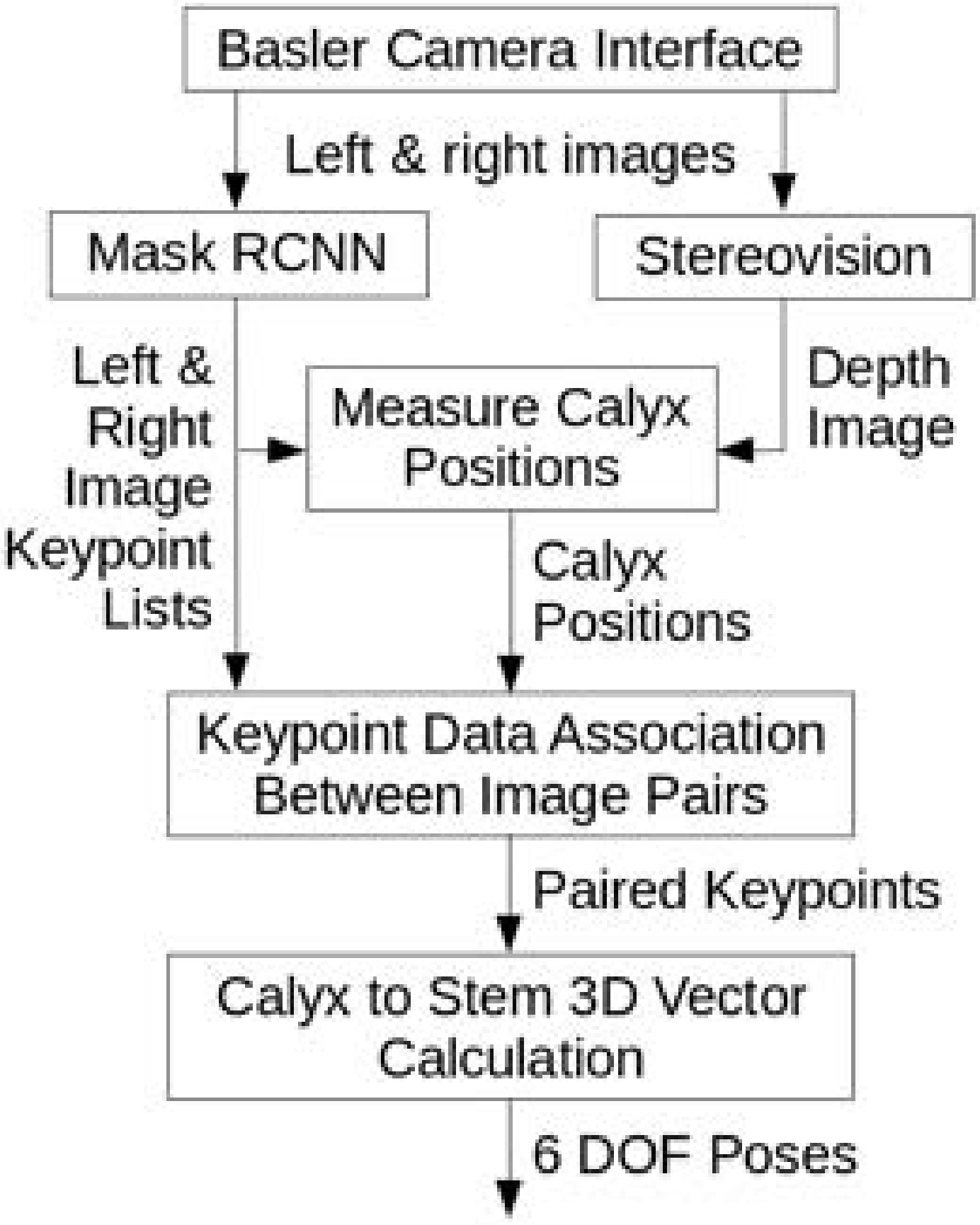

*Figure 209: Pipeline proposed for kiwifruit pose detection with stereo cameras.*

In the proposed pipeline, the neural network would detect the centre of the calyx, at one end of the kiwifruit, and also detect the stem attachment point at the other end of the fruit. However, if the harvesting sensors were placed well below the fruit looking directly upwards, the stem attachment point would not be visible in the images. To solve the issue of not seeing the stem attachment point, it was postulated that there might be a deep learning solution and so the following ideas were proposed:

- Let the labeller attempt to label the stem attachment points, even though these points are not visible in the images, and train the CNN to predict the stem attachment points. However, it was unclear if people would be able to accurately label a point that they were not able to see and then verification of accuracy would be difficult.

- Another idea was to attach kiwifruit to the end of a robot arm and have the arm move the kiwifruit to multiple recorded poses, while taking images of the fruit. This would automatically produce a dataset of image and kiwifruit pose pairs that might be used to train a CNN. It was thought that this approach would take considerable time to set up, considering not every pose of the arm is valid for taking images of the fruit and there would need to be coordination and calibration between arms and the cameras. In addition, the data collection system would not be completely automatic since the fruit would need to be replaced regularly, in order to get data for different fruit samples. Furthermore, it was unclear if the resulting neural network would be effective in the real world; that is to say, if training on the robot arm and kiwifruit would transfer to working with kiwifruit in an orchard.

Because of the identified risks for each of these approaches and the time pressure of preparing for the harvesting season, it was decided to use a hand engineered approach to detect the pose of the kiwifruit. It seemed that the useful features available to use for the pose detection algorithm, could include:

- The shape of the segmented kiwifruit flesh. For example, a vertical fruit, directly above the sensor would appear approximately elliptical with a low eccentricity, whereas a more horizontal fruit would have a higher eccentricity.
- The position of the segmented kiwifruit calyx, relative to the segmented kiwifruit flesh. For example, a vertical fruit, directly above the sensor would have the calyx at the centre of the kiwifruit flesh.
- The position and shape of the point cloud from the depth data. For example, it could be possible to fit a template of a kiwifruit to the segmented point cloud of the kiwifruit.

It seemed that the benefit of using the point cloud data would be that there would be real world dimensions to use to calculate the pose and it seemed that a template point cloud could be registered with the sensor data using an existing algorithm such as Iterative Closest Point (ICP). However, the segmentation of the point cloud will generally not be perfect so there could be extra points that contribute erroneous measurements. In addition, the distance measurements contain noise. Furthermore, the kiwifruit model used may not match the real kiwifruit well, which could contribute to errors. For these reasons, an alternative approach was explored.

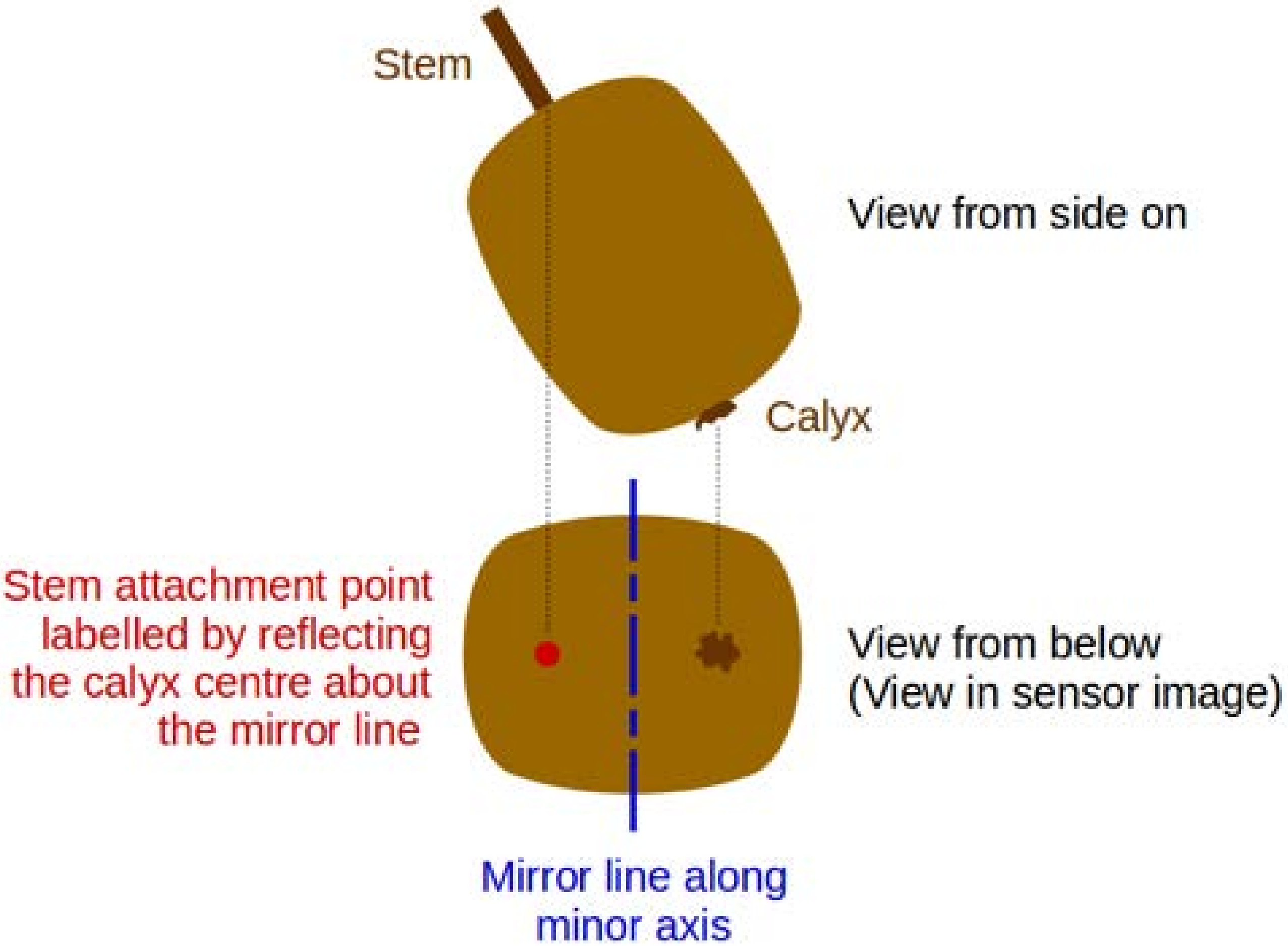

*Figure 210: Illustration of an observed method of a person labelling a stem attachment point which is not visible.*

In order to develop an algorithm for kiwifruit pose detection, some inspiration was taken from how a person attempted to manually label the stem attachment point in images. It was noted that when a person tried to label the stem attachment point of the fruit, when they could not actually see this point, they modelled the fruit as an ellipse in order to project an imaginary mirror line along the minor axis of the fruit and then reflected the calyx about the mirror line to get the stem attachment point (Figure 210). An algorithm using a similar procedure on the instance segmentation output was developed, with the following steps for finding the stem attachment point:

1. In order to find calyx-skin pairs, the calyx segmentation outputs were dilated. Then the overlaps between calyx instances and skin instances were found. For each calyx instance, the skin instance with the maximum overlap was assigned to the calyx instance.
2. The centre of each calyx instance was found using the median x-coordinate and y-coordinate in the calyx segmentation output.

3. For each calyx-skin pair, Principal Components Analysis (PCA) of the skin instance points was used to find the centre of the skin and the directions of the principal components.
4. For each calyx-skin pair, the calyx centre was reflected about the second principal component.

It was thought that this procedure might work well for unoccluded fruit. However, for occluded fruit it was postulated that PCA might not find the principal components of the fruit accurately. Although, for occluded fruit, it was thought that the collision detection might detect the occlusion and rank the fruit as a lower prospect for picking.

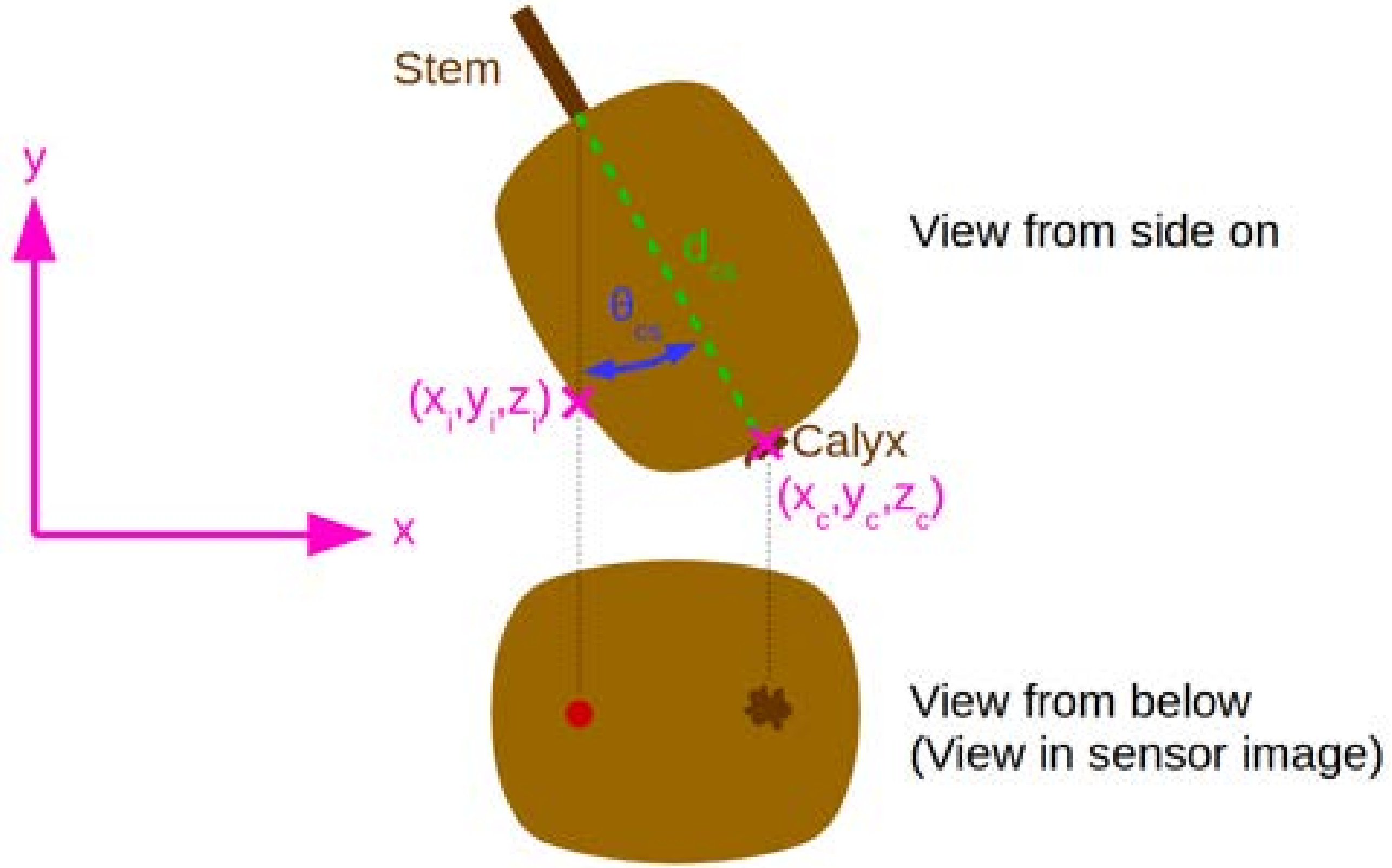

*Figure 211: Key points and variables used in the calculation of kiwifruit pose.*

In order to find the angle of the kiwifruit from vertical, three approximations were made:

- That the kiwifruit has a length, $d_{cs}$, illustrated in Figure 211.
- That the sensor was directly below the fruit.
- That the sensor was a long distance below the fruit, so that light from the fruit was parallel going into the sensor.

With these approximations and the definitions from Figure 211, the angle of the kiwifruit might be found using:

$$\theta_{cs} = asin\left(\frac{x_c - x_i}{d_{cs}}\right) \quad (63)$$

In order to find points potentially in collision with the robot arm during a picking manoeuvre, the gripper was modelled as a rectangular volume and points from the depth sensor that were detected to be within this volume, with the gripper in position to perform detachment, were classified as in collision with the gripper. In order to do this, the following steps were used:

1. A fixed size region of interest around the centre of a detected fruit was set.
2. The points in the region of interest were transformed to the robot arm coordinate system.
3. The points were transformed into the coordinate system of the gripper, for the pose of the gripper at the instant of fruit detachment.
4. Points within the fixed volume of the gripper were classified as in collision.

It was decided that the output for collision checking should be a single score for each fruit instance. This score was supposed to represent the current risk from collisions, if the fruit was picked. The potential harms from collision were identified to be a "soft collision" with non-rigid parts of the canopy, which might cause fruit damage, and a "hard collision" with rigid parts of the canopy, which might cause damage to the canopy or the robot. The metrics that seemed relevant for computing the score for each fruit were:

- The number of points, *n*, detected as in collision with the picking mechanism for the fruit.
- The class of the points, where different classes were assigned different risk values. The way classes were represented in the collision score calculation were as indices, *c*. The risk values were stored in an array, *R*, so that *R[c]* was the risk value for class, *c*. The risk values used are given in Table 52.

It was not clear how these metrics should be combined in order to calculate a score, so it was decided that initially a simple approach could be used so that the effect of each metric could be observed. In order to do this, the collision score was calculated according to:

$$\sum_{i=0}^{n} R[c_i] \quad (64)$$

*Table 52: Risk values used in the calculation of the harvesting collision score.*

| **Classes** | **Risk Value** |
|---|---|
| Fruit | 3 |
| Wires, steel beams, solid branches | 9 |
| No class | 1 |

Fruit with a collision risk score above a threshold were not targeted for picking. This did not necessarily mean that these fruit would never be picked because it is possible that the action of harvesting less obstructed fruit may result in a more obstructed fruit becoming less obstructed.

With kiwifruit poses measured and collisions for each fruit assessed, the question arose: in what order should the fruit be picked? There were multiple factors that could be taken into consideration, including minimising the time to harvest the fruit and minimising the risk of collisions. For the work described here, more emphasis was put on minimising the risk of collisions, since fruit damage was an issue with the original kiwifruit harvester. In order to do this, the harvesting pick order was based on the collision score.

# 10 Appendix D: Further Test with Camera Row Following

This appendix describes an exploratory experiment that was performed for camera row following. The aim of this initial experiment was to determine if a classification CNN would be able to take an image from a robot and correctly classify whether the robot should drive left or right. Ground truth for this experiment was taken from human driving. The hardware used for data collection was similar to that used for driving by traversable space segmentation with a Clearpath Husky robot [96] that had a single Basler Dart daA1600-60uc [92] camera with a 5.5 mm focal length lense, mounted front and centre at a height of 0.8 m off the ground. The camera had 1600 pixels wide by 1200 pixels high resolution- although the images were resized to 256 pixels by 256 pixels resolution. The data was collected using a laptop and a joystick was plugged into the laptop to allow a person to manually drive the Husky robot.

As discussed in existing literature, it is not enough during data collection to collect images from when a robot is driving its path correctly because then most of the data is of path following without drifting off the path and the neural network does not learn to correct errors in the driving [171]. The result is that the robot will tend to drift of its path and will not correct any errors. To counter this problem, the premise of the data collection for driving in kiwifruit orchards was to capture images and label them with steering commands, while a person remote controlled the robot to perform row following manoeuvres, where an initial linear and angular offset were corrected. The data was collected in this way to ensure that the classification network would be trained to give steering commands that would tend to align the robot with the row centreline. The procedure used during data collection was:

1. Manually driving to create a linear and angular offset to correct.
2. Pushing a button on the joystick to trigger the recording of images. Each image was saved with the steering command that was being given at the same time, as well as a sequence identification number to ensure that the name of each image was unique.
3. Manually driving to correct the linear and angular offset.
4. Pushing another button on the joystick to stop the recording of images.
5. Repeating steps 1-4 multiple times.

The camera images collected were post-processed into 2 classes for classification; “turn left” and “turn right”. The resulting data was used to train an AlexNet CNN, using the hyperparameters given in Table 53. The resulting average validation accuracy from 3 training runs was 88%. This result seemed quite poor for a 2 class classifier that is intended to perform autonomous driving; however, it was somewhat encouraging, given the relatively small amount of data used for training, without data augmentation. Furthermore, a pre-trained network was not used for transfer learning, so the only information that the neural network had been trained on was the collected training images.

*Table 53: Hyperparameters for training a steering angle direction classifier.*

| Hyperparameter Description | Value |
|---|---|
| Number of Training Images | 143 |
| Number of Validation Images | 15 |
| Number of Training Epochs | 99 |
| Number of Training Attempts | 3 |
| Solver Type | Adam |
| Static Learning Rate | 0.001 |

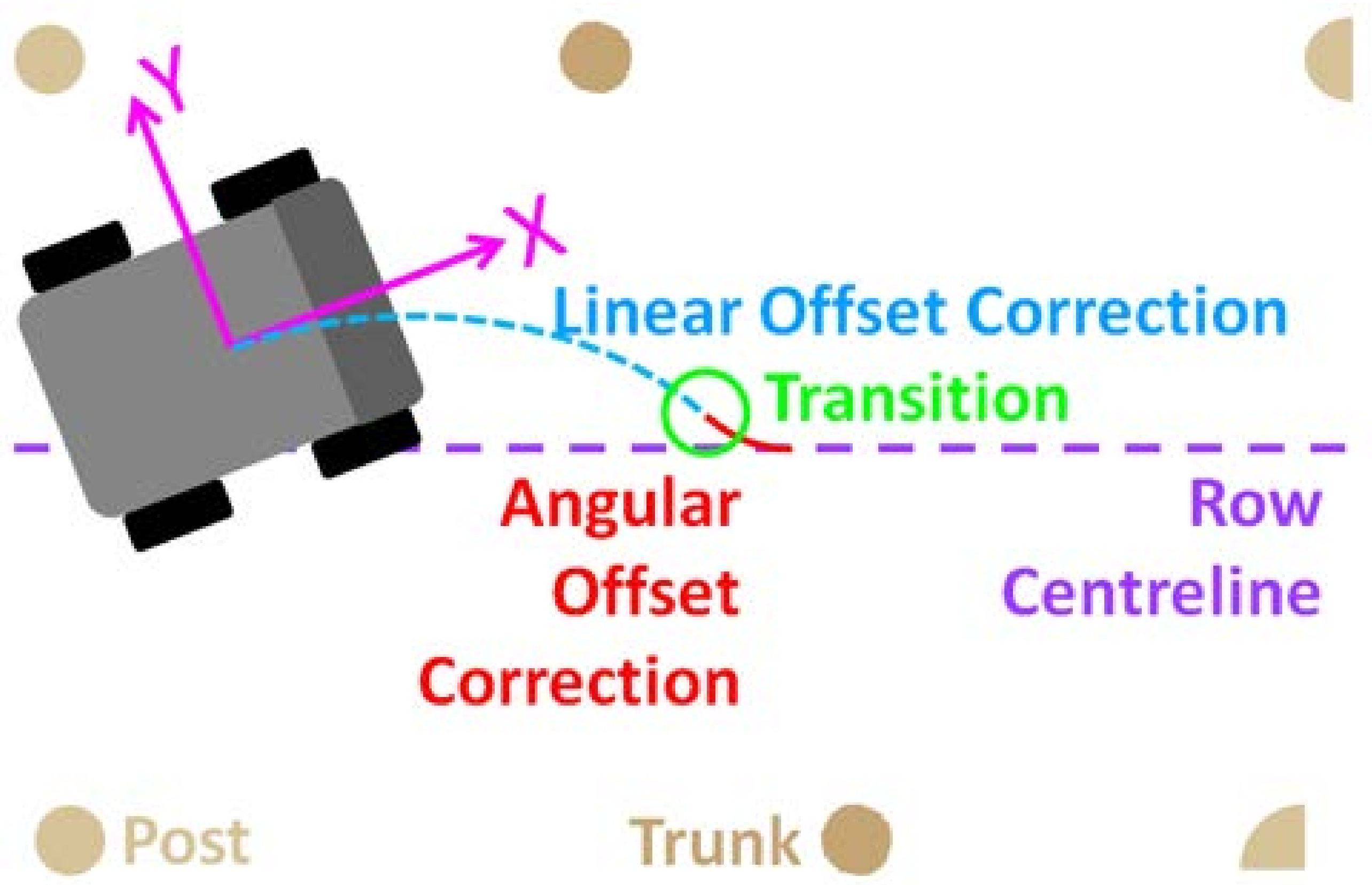

*Figure 212: When driving manually, the manoeuvres performed started with a linear offset correction phase followed by an angular offset correction.*

In addition to the amount of training data, it was noted that there was an ambiguous region in a human controlled row following correction manoeuvre in the transition between the first curve, where the linear offset is corrected, and the second curve in the opposite direction, where the angular offset is corrected (Figure 212). In this transition phase, it was common to get some variability in the manual driving; sometimes the transition was started earlier or later and lasted different amounts of time. As a result the labels in this region may have been inconsistent.

There is another effect that may affect the classifier. It can be observed in Figure 212 that the angular offset correction phase of the row correction manoeuvre tends to be much shorter than the linear offset correction phase. This could result in there being more data for the linear offset correction phase and hence the resulting neural network may be biased towards producing a linear offset correction.

# 11 Appendix E: Publication Contributions